%% file: book.tex
\documentclass[oneside,openany]{book}

\usepackage{lscape}
\usepackage{fancyvrb}
\usepackage[round]{natbib}
\usepackage[dvips]{graphicx}
\usepackage{amssymb}
\usepackage[dvipdfm, bookmarks=true]{hyperref}
\usepackage{makeidx}
\usepackage{sectsty}

\newcommand*{\implies}{\mathbin{\Rightarrow}}

\allsectionsfont{\raggedright}
\subsubsectionfont{\itshape}

\makeindex

\begin{document}

\bibliographystyle{plainnat}

\frontmatter
  
\include{title_page}
\include{dedication}
\include{poem}
\include{acknowledgements}

\pdfbookmark[0]{Table of Contents}{contents}
\tableofcontents

\mainmatter

\include{introduction}

\include{foundations}

\include{theory}

\include{computing}

\include{language}

\include{rr}

\include{pr}

\include{pps}

\include{learning}

\include{maths_logic}

\include{neural}

\include{psychology}

\include{future}

\include{conclusion}

\appendix

\include{matching}

\include{physics}

\backmatter

\raggedright
\pdfbookmark[0]{Bibliography}{bibliography}

\input{book.bbl}
\index{language!semantics|see{grammar, semantics}}
\index{retrieval|see{information, retrieval}}
\index{language!learning|see{learning, language}}
\index{category|see{class}}
\index{redundancy|see{information, redundancy}}
\index{generalisation|see{grammar, generalisation}}
\index{minimum message length|see{minimum length encoding}}
\index{minimum description length|see{minimum length encoding}}
\index{adaptation|see{neural, adaptation}}
\index{inhibition|see{neural, inhibition}}
\index{semantics|see{grammar, semantics}}
\index{rules|see{reasoning, rules}}
\index{deduction|see{reasoning}}
\index{inference|see{reasoning}}
\index{family resemblance|see{class, polythetic}}
\index{language!parsing|see{parsing, language}}
\index{recognition|see{perception, recognition}}

\pdfbookmark[0]{Index}{index}
\printindex

\end{document}

%% file: title_page.tex
\begin{titlepage}

\begin{center}

\vspace*{0.75in}

\huge

{\bf UNIFYING COMPUTING AND COGNITION: THE SP THEORY AND ITS APPLICATIONS}

\vspace{0.5in}

\large

{\bf J Gerard Wolff}\footnote{Dr J G Wolff, CognitionResearch.org.uk. Telephone: +44 (0)1248 712962. \\
\indent \space \space Email: jgw@cognitionresearch.org.uk.}

\vspace{0.5in}

\normalsize

\today

\vspace{0.5in}

\huge

DRAFT

\end{center}

\end{titlepage}

%% file: dedication.tex
\vspace*{2in}

\noindent To Marianne, Daniel and Esther, with my love, and in memory of my parents, Leslie and Agnes.

%% file: poem.tex
\begin{center}

\vspace*{2in}

\section*{Too blue for logic}

My axioms were so clean-hewn, \\
The joins of `thus' and `therefore' neat \\
But, I admit \\
Life would not fit \\
Between straight lines \\
And all the cornflowers said was `blue,' \\
All summer long, so blue. \\
So when the sea came in and with one wave \\
Threatened to wash my edifice away - \\
I let it. \\

\vspace{0.2in}

{\em Marianne Jones} \\

\end{center}

%% file: acknowledgements.tex
\section*{Acknowledgements}

\sloppy I am very grateful to the following people for discussions, comments and suggestions relating to the SP project:
Mike Anderson,
Peter Apostoli,
Jim Baldwin,
Horace Barlow,
Manos Batsis,
Dorrit Billman,
Geof Bishop,
Howard Bloom,
Erik Borgers,
Bob Borsley,
Alan Bundy,
John Winston Bush,
Gordon Brown,
John Campbell,
Chuck Carlson,
Nick Chater,
Andy Chipperfield,
Frans Coenen,
James Crook,
Jim Cunningham,
Alain Cymm,
Louise Dennis,
Colin de la Higuera,
Lorraine Dodd,
David Dowe,
Dave Elliman,
Nick Ellis,
Michael Flor,
Richard Forsyth,
Michele Friend,
Alex Gammerman,
Antony Galton,
Ross Gayler,
Felix Goldberg,
Frank Gooding,
Pat Hayes,
Adrian Hopgood,
John Hornsby,
Geoffrey Hunter,
Jason Hutchens,
Alan Hutchinson,
Christian Huyck,
Miguel Jimenez-Montano,
Bob Jones,
Phil Jones,
Simon Jones,
Mark Lawson,
Vahe Karamian,
Krishna Krishnamurthy,
Lucy Kuncheva,
Pat Langley,
Bill Majoros,
Paul Mather,
Adrian Mathias,
Jim Maxwell Legg,
Will McKee,
Chris Mellish,
Peter Milner,
Detlef Morganstern,
Stephen Muggleton,
Ajit Narayanan,
Sergio Navega,
Craig Nevill-Manning,
Freddy Offenga,
Thomas Packer,
Alex Paseau,
John Pickering,
Tim Porter,
Emmanuel Pothos,
Friedemann Pulverm{\"u}ller,
Theofaris Raptis,
Edward Remler,
Steve Robertshaw,
Stephen Schmidt,
Oliver Schulte,
Derek Sleeman,
Mathew Smith,
Rod Smith,
Ray Solomonoff,
Graham Stephen,
Heiner Stuckenschmidt,
Simon Tait,
Tichomir Tenev,
Guillaume Thierry,
Robert Thomas,
Chris Thornton,
Menno van Zaanen,
Chris Wallace,
Hong Wang,
Chris Wensley,
Rob Whetter,
Chris Whitaker,
Peter Willett,
Richard Young,
Roger Young.

Special thanks to Simon Tait for his interest and for very useful discussions right at the beginning, to Chuck Carlson for setting up the computing-as-compression mailing list (www.sanna.com/casc/), and to Emmanuel Pothos for his interest and major contribution in relating the SP ideas to current thinking in psychology (the basis for Chapter \ref{psychology_chapter}).

Special thanks too to Ticho Tenev for taking the trouble to read the whole book in draft and give me detailed comments and suggestions, to Peter Milner for detailed comments on Chapter 11, to David MacKay for several useful suggestions, and to Colin de la Higuera for raising some interesting questions that led to some rewriting. The responsibility for all errors and oversights is, of course, my own.

\sloppy The work has also benefitted from lively discussions on the computing-as-compression mailing list, including contributions from:
David Cary,
Chuck Carlson,
Nate Cull,
Chris Ferguson,
Phil Goetz,
Max Little,
James Luberda,
Pabitra Mitra,
Detlef Morgenstern,
Brendan Macmillan,
Sergio Navega,
Remy Nonnenmacher,
Freddy Offenga,
Andrew Seaca,
Andrew Stanworth,
Robert Toews,
Ravi Venkatesan.

I am very grateful to Martin Taylor for allowing me the use of an office and facilities in the School of Informatics, University of Wales Bangor, and---for other support and assistance---to Tim Porter, Chris Wensley, Chris Whitaker, Lucy Kuncheva, John Owen, Dave Whitehead and Matthew Williams in the School, and Dafydd Roberts and Paul Rolfe in Information Services.

Two research grants have assisted this work: a Personal Research Grant from the UK Social Science Research Council (HRP8240/1(A), 1982-3) and a Research Grant from the UK Science and Engineering Research Council (GR/G51565, 1992-4).

Last, but very far from least, I am very much indebted to Marianne for her love and support at home, and to my two grown-up children, Daniel and Esther, for their love and continued interest in the project as it has slowly matured.

Many apologies to anyone whose contribution I may have failed to mention.

%% file: introduction.tex
\chapter{Introduction}\label{introduction_chapter}

\begin{quotation}

\noindent ``Fascinating idea! All that mental work I've done over the years, and what have I got to show for it? A goddamned zipfile! Well, why not, after all?'' {\em John Winston Bush}, 1996.

\end{quotation}

\noindent In the world of computing, the term ``information compression''\index{information!compression} (or ``data compression'') is normally associated with slightly dull utilities like WinZip or PkZip or the Lempel-Ziv algorithms on which they are based. Information compression is useful if you want to economise on disk space or save time in transmitting a file but otherwise it does not seem to have any great significance.

In this book I aim to show that there is much more to information compression than this. The book describes ways in which information compression can illuminate concepts and issues in artificial intelligence and human cognition and, indeed, the nature of `computing' itself. More specifically, the book describes the concept of {\em information compression by multiple alignment, unification and search} and the ways in which this framework can be used to model the workings of a Turing machine and such things as parsing and production of language, `fuzzy' pattern recognition and information retrieval, probabilistic reasoning, planning and problem solving, unsupervised learning, and a range of concepts in mathematics and logic.

Information compression may be interpreted as a process of trying to maximise {\em Simplicity} in information (by removing `redundancy') whilst retaining as much as possible of its non-redundant, descriptive {\em Power}. Hence the name `SP' that has been adopted for these proposals. They will normally be referred to as the `SP theory' but they may also be referred to as the `SP system' because the concepts are developed as an abstract working model. Equivalent expressions are `SP framework', `SP model', `SP scheme' and `SP concepts'.

A relatively brief overview of the SP theory and its applications is available in \citet{wolff_icmaus_overview}.

\section{Beginnings}

My interest in these kinds of ideas was sparked originally by fascinating lecturers about economical coding in the nervous system given by Horace Barlow when I was an undergraduate at Cambridge University.

Some time later, I developed two computer models of language learning: MK10 which demonstrates how a knowledge of the segmental structure of language (words, phrases etc) can be bootstrapped from unsegmented text without error-correction or `supervision' by a `teacher', and SNPR---an augmented version of MK10---that demonstrates how grammars can be learned without external supervision \citep[see][and earlier papers cited there]{wolff_1988}. In the course of developing these models, the importance of economical coding became increasingly clear.

At about that time, I became acquainted with Prolog\index{Prolog} and I was struck by the parallels that seemed to exist between that system, designed originally for theorem proving, and my computer models of language learning. A prominent feature of my learning models is a process of information compression by searching for patterns that match each other together with a process of merging or `unifying' patterns that are the same. Although information compression is not a recognised feature of Prolog, a process of searching for patterns that match each other is prominent in that system and the merging of matching patterns is an important part of `unification' as that term is understood in logic. It seemed possible that information compression might have the same fundamental importance in logic as it has in inductive learning.

These observations led to the thought that it might be possible to integrate inductive learning and logical inference within a single system, dedicated to information compression by pattern matching, unification and search. Further thinking suggested that the scope of this integrated system might be expanded to include such things as information retrieval, pattern recognition, parsing and production of language, and probabilistic inference.

Development of these ideas has been underway since 1987. It was evident quite early that the new system would need to be organised in a way that was rather different from the organisation of the MK10 and SNPR models. And, notwithstanding the development of Inductive Logic Programming,\index{inductive logic programming} it seemed that Prolog\index{Prolog}, in itself, was not suitable as a vehicle for the proposed developments---largely because of unwanted complexity in the system and because of the relative inflexibility of the search processes in Prolog. It seemed necessary to build the proposed new integrated system from new and `deeper' foundations.

Initial efforts focussed on the development of an improved version of `dynamic programming' for finding full matches and good partial matches between pairs of patterns. About 1994, it became apparent that the scope of the system could be greatly enhanced by replacing the concept of `pattern matching' with the more specific concept of `multiple alignment', similar to that concept in bioinformatics but with important differences.

\section{Goals of the research and potential benefits}

The main aim of the research has been to develop a theory of information processing in computers and in brains that integrates and simplifies concepts and observations in those domains, as discussed in the next section. A subsidiary aim has been to develop a new kind of computing system, based on the theory, with potential advantages over the current generation of computers, especially in artificial intelligence.

Like any good theory, the SP theory should simplify our ideas, provide new insights, make predictions and suggest new avenues for research. These aspects of the theory are discussed at appropriate points throughout the book. Potential applications of the proposed new computing system will be evident from examples and discussion throughout the book. There is a summary and discussion of the anticipated applications and benefits in Chapter \ref{future_chapter}.

\section{Creating a good theory}\label{creating_a_good_theory}

The SP theory originated in research on human psychology but its scope now extends to information processing in {\em any} kind of system, either natural or artificial. How should such a theory be constructed and how should it be evaluated? At the risk of doing great violence to some complex and subtle issues in the philosophy of science, I discuss those questions briefly in the following subsections, and describe how the SP theory has been developed. 

\subsection{Breadth and depth}\label{breadth_and_depth}

Some time ago, Allen Newell wrote that:

\begin{quotation}

``Psychology, in its current style of operation, deals with phenomena. ... Every time we find a new phenomenon---every time we find PI release, or marking, or linear search, or what-not---we produce a flurry of experiments to investigate it. We explore what it is a function of, and the combinatorial variations flow from our experimental laboratories. ... [These] phenomena form a veritable horn of plenty for our experimental life---the spiral of the horn itself growing all the while it pours forth the requirements for secondary experiments. ...

``Psychology also attempts to conceptualize what it is doing, as a guide to investigating these phenomena. How do we do that? Mostly, so it seems to me, by the construction of oppositions---usually binary ones. We worry about nature versus nurture, about central versus peripheral, about serial versus parallel, and so on. ... Suppose that in the next thirty years we continued as we are now going. Another hundred phenomena, give or take a few dozen, will have been discovered and explored. Another forty oppositions will have been posited and their resolution initiated. Will psychology then have come of age? Will it provide the kind of encompassing of its subject matter ... that we all posit as a characteristic of a mature science? ...

``As I examine the fate of our oppositions, looking at those already in existence as a guide to how they fare and shape the course of science, it seems to me that clarity is never achieved. Matters simply become muddier and muddier as we go down through time. Thus, far from providing the rungs of a ladder by which psychology gradually climbs to clarity, this form of conceptual structure leads rather to an every increasing pile of issues, which we weary of or become diverted from, but never really settle.'' \citep[][pp. 284--289]{newell_1973}.

\end{quotation}

The gist of Newell's critique, in these quotes and elsewhere in that paper, is:

\begin{itemize}

\item That researchers were focussing too narrowly on single phenomena or `oppositions' between classes of possible mechanisms. This echoes Neisser's earlier critique of `microtheories' in psychology \citeyearpar{neisser_1967}.

\item That boxes and arrows or similar sketchy descriptions of possible mechanisms are not sufficiently precise to know whether or not the proposed mechanisms would work as anticipated.

\end{itemize}

As a remedy, he suggested that processing models should be `complete' rather than partial (which really means that they should be working computer programs), that any one model should deal with a complex task (``a genuine slab of human behaviour'' (p. 303)) or that it should provide an integrated view of a range of smaller tasks. In short, he saw a need to increase both the breadth and the depth of theories in psychology.

\subsection{Falsifiability and the making of predictions}\label{falsifiability_predictions}

Since the publication of Karl Popper's {\em The Logic of Scientific Discovery} \citep{popper_2002} it has been widely accepted that a scientific theory cannot be `good' unless it is falsifiable---meaning that one can conceive of empirical evidence that would cause one to discard the theory or at least revise it (although there may be ethical or methodological hurdles that might make it difficult to obtain that evidence).

Marching hand-in-hand with this principle is the idea that good theories should make predictions that can be subject to empirical test. A classic example is the way Einstein's prediction that light is bent when it passes close to a heavy body like the sun was dramatically confirmed in 1919 by Arthur Eddington's observations of stars and their apparent positions, made in the island of Principe off the west coast of Africa at the time of a solar eclipse.

Although falsifiability is widely accepted as a necessary feature of any good theory, it is also recognised that falsifiability can only be achieved in an informal sense. When a piece of evidence appears to falsify a given theory, it is always possible to find alternative explanations: the instruments were faulty, the materials were not in the right condition, and so on. To cut a long discussion short, our judgement of whether or not a given theory is supported by the evidence has a probabilistic character. If we say that a given theory is falsifiable, we are, in effect, saying that there are kinds of evidence that would cause us to judge that there is a low probability of the theory being correct.

\subsection{Simplicity and power}

Although the medieval English philosopher and Franciscan monk William of Ockham\index{Ockham, William of} lived long before the age of science, his dictum that ``entities should not be multiplied unnecessarily'' is often quoted in support of the idea that scientific theories should be simple or, more precisely, any one theory should be as simple as possible, consistent with the set of observations which the theory is designed to explain. An adjunct to this idea, which is less often articulated but is, nevertheless, widely accepted by people working in science, is that if two theories are equally simple, then the one that explains the widest range of observations is to be preferred. In summary, it is generally acknowledged that a `good' scientific theory should combine `simplicity' with explanatory range or `power'.

These ideas are rather similar to principles of minimum length encoding, developed in connection with grammar induction (grammar discovery or grammatical inference) and themselves part of the foundations of the SP theory. To anticipate Section \ref{mle_section}, where minimum length encoding principles are described more fully, a `good' grammar for a given language is one that combines simplicity with an ability to describe the language in an economical manner. If `explanation' in terms of a given scientific theory is understood to be economical description of the world (or aspects of it) in terms of that theory, then the analogy is good:

\begin{itemize}

\item Some theories are weak `catch-all' theories like `because God wills it' or the over-enthusiastic use of the concept of `instinct' to explain any kind of human behaviour. These theories are like the `promiscuous' grammars described in Section \ref{mle_section} that are too simple to be able to describe anything in an economical way.

\item Other theories are weak because they merely redescribe a set of observations in other terms. These theories are like over-large `{\em ad hoc}' grammars ({\em ibid.}) that repeat the original data.

\end{itemize}

As we shall see in Section \ref{mle_section}, minimum length encoding principles actually boil down to a rather simple idea: ``Given a sample of language or other body of data, compress it as much as possible in a lossless manner''.\footnote{Lossless compression will be explained in Section \ref{randomness_redundancy_structure}.} Likewise, science may be seen as a process of compressing raw data as much as possible. As John Barrow has written: ``Science is, at root, just the search for compression in the world'' \citep[][p. 247]{barrow_1992}.

Although the analogy seems good, it is not yet possible to evaluate scientific theories with the mathematical precision with which minimum length encoding principles can be applied in grammar induction (and in the SP theory). Our assessments of scientific theories currently depend on informal judgements of simplicity and power and it seems likely that this will be true for some time to come.

\subsection{Integration}\label{theory_criteria_integration}

The six criteria that have been described may be reduced to two: simplicity and power.

The need for breadth of scope of scientific theories may be equated with power. A good theory (in psychology and elsewhere) should describe or explain a significant range of phenomena (``a genuine slab of human behaviour'') rather than one or two details.

The criterion of depth may be subsumed by both simplicity and power because a theory that is so simple that it cannot even explain its target range of observations is no theory at all. ``Entities should not be multiplied unnecessarily'' but, equally, they should not be reduced past the point where they are not up to the job.

Falsifiability may be seen as a manifestation of explanatory power. If a theory, like Freud's theory of dreams, can always be twisted to `explain' any conceivable observation within its domain of application, then we sense that its ability to describe the world in an economical way is weaker than theories that cannot be manipulated in this way. Such theories are `catch-all' theories like those mentioned above.

The idea that a theory should make empirical predictions may also be seen to derive from the concept of power: if a prediction of a given theory is confirmed by observation or experiment, then the power of the theory (its explanatory range) is increased without any change in simplicity---thus increasing the rating of the theory in terms of simplicity and power. As with Eddington's confirmation of Einstein's prediction, this kind of evidence can be compelling.

In principle, a theory may combine simplicity with an ability to explain a wide range of existing observations without predicting the kinds of observations, yet to be made, that Eddington confirmed on his expedition to Principe. And most people working in science would acknowledge that such a theory would be useful, if, perhaps, not quite as exciting as when predictions coincide with later observations. But the intimate connection that exists between information compression and inductive inference (to be discussed in Section \ref{probabilities_ic_section}) means that it is difficult to construct a theory that compresses a significant range of observations without, at the same time, making predictions of one kind or another.

\subsection{Theories in computing, mathematics and logic}

Most of the foregoing discussion is about theories in empirical sciences like psychology, biology or chemistry where the aim is to provide explanations of things observed in the world. Although disciplines like computing, mathematics and logic deal with abstractions rather than empirical phenomena, similar principles apply. In these cases, a `good' theory is one that combines simplicity with an ability to integrate or unify a range of pre-existing concepts. In these domains, a theory may be regarded as `falsified' if it fails to accommodate a pre-established construct or operation.

Artificial intelligence---which is an important topic throughout this book---is something of a half-way house. It has empirical content---because it aims for human-like capabilities---but it is not necessary that the imitation should be achieved by human-like means. Of course, many researchers in artificial intelligence borrow extensively from psychology and {\em vice versa}. Apart from being as simple as possible, a `good' theory in artificial intelligence should explain a wide range of phenomena or integrate a wide range of concepts or both these things.

\subsection{Orientation}

It would be nice to report that principles like those described in the preceding subsections had all been fully articulated before work began on developing the SP theory. The reality was that, like most people working in science, I had only an informal, intuitive sense of the kind of theory I was aiming for. That said, I believe the SP theory meets the criteria well: it is not a vacuous catch-all `theory' that explains everything and nothing, and it is not a `theory' that is merely a redescription of the data it purports to `explain'.

When I was working on language learning, I had been persuaded by Neisser's \citeyearpar{neisser_1967} criticism of `microtheories' in psychology to aim for something with relatively wide scope. In the SP programme, this prejudice was reinforced by Newell's later writings in the same vein (some of which were quoted above). As things have turned out, the scope of the theory is much wider than was envisaged originally, with things to say about aspects of human cognition, artificial intelligence, computing, mathematics and logic.

In the research on language learning and in the SP programme, computer models have been an invaluable tool. Attempting to express one's embryonic thoughts in the form of a computer program forces explicitness where vagueness might otherwise prevail, and the process of running a computer model on appropriate data allows one to see very clearly whether or not the often complex implications of a simple idea actually work out as anticipated. Countless armchair speculations have been dumped as a result of this kind of testing.

Computer models help to provide the kind of precision and explicitness that Newell (and others) have called for but the same can also be said of mathematical equations. As readers will see in chapters that follow, the SP theory is founded on well-established concepts from information theory, combinatorics and probability theory, and it incorporates mathematical equations from those areas at appropriate points. But many of the concepts in the SP theory, especially the version of the multiple alignment concept that has been adopted in the theory, are best described and developed in the form of computer models, not mathematical equations.

\subsection{Making haste slowly}\label{making_haste_slowly}

Apart from the unexpected similarities between Prolog and my language learning models, a germinal thought was the goal of finding a `universal' format for different kinds of knowledge. Originally, the focus was on finding a single format for the syntax and semantics of natural languages---to facilitate the development of a uniform learning process for natural language syntax and semantics and their integration \citep[see][]{wolff_1987}. Interestingly enough, Prolog\index{Prolog} proved relevant again as one of the strongest candidates, although it is not well suited to the representation of class hierarchies with inheritance of attributes.

The problem of integrating different kinds of knowledge surfaced again when I was working at Praxis Systems on an `integrated project support environment' for the development of software. In this case, the need was to find a way of organising all the different kinds of knowledge involved in software development---specification of requirements, high-level design, low-level design, source code, object-code, project plans, budgets, and so on---bearing in mind that each `document' or `object' is likely to be divided into a hierarchy of parts and sub-parts, that for each part there may be a hierarchy of versions and sub-versions, and that tight control is needed over associations between specific versions of different parts across the range of different kinds of knowledge.

Given that the notion of a `version' is rather similar to the concept of `class' in object-oriented design, it seemed natural to represent a hierarchy of versions as a hierarchy of classes in an object-oriented language like Simula, Smalltalk or C++. An attraction of this idea is that one could exploit the idea of `inheritance of attributes' to avoid repeating information unnecessarily: anything that was true of a major version would also be true of all the versions and sub-versions below it, without the need to repeat the information in each of those subsidiary versions individually. Since object-oriented languages also allow one to represent any object using a hierarchy of parts and sub-parts, things were looking good.

But, of course, there is a snag! Although an `attribute' (data structure or method) is a part of a class, and attributes can themselves be divided into parts and sub-parts, an attribute cannot be part of an individual object and a part of an object cannot serve as an attribute of a class. In C++ (and all other object-oriented languages that I am familiar with), the attributes of a class are defined in the source code whereas part-whole hierarchies of individual objects are built at run time. This makes it impossible to achieve a full integration of class-hierarchies with part-whole hierarchies or to eliminate the artificial distinction between `attribute' and `part'. An implication of that integration would be that the concept of `class' and the concept of `object' should be merged.

The `SP' language was one of my first attempts at solving this problem and creating a system that could represent and integrate diverse kinds of knowledge. As can be seen from the syntax of the language \citep{wolff_1990}, shown in Figure \ref{sp_syntax_figure}, knowledge was conceived as `objects' comprising a combination of ordered and unordered AND relationships, together with (inclusive) OR relationships. It was envisaged that `objects' would also serve as `classes' so that the system could be used to define a range of different kinds of knowledge, with complete integration amongst part-whole hierarchies and class-inclusion hierarchies. It was also envisaged that the system would be driven by information compression, achieved by the matching and unification of patterns.

\begin{figure}[!hbt]
\begin{center}
\begin{tabular}{l}
{\em Object} $\rightarrow$ {\em Ordered-AND-object} $\vert$ \\
~~~~~ {\em Unordered-AND-object} $\vert$ \\
~~~~~ {\em OR-object} $\vert$ {\em Simple-object}; \\
{\em Ordered-AND-object} $\rightarrow$ `(', {\em body}, `)'; \\
{\em Unordered-AND-object} $\rightarrow$ `[', {\em body}, `]'; \\
{\em OR-object} $\rightarrow$ `\{', {\em body}, `\}'; \\
{\em body} $\rightarrow$ {\em b} $\vert$ NULL; \\
{\em b} $\rightarrow$ {\em Object}, {\em body}; \\
{\em Simple-object} $\rightarrow$ {\em symbol} $\vert$ `\_'; \\
{\em symbol} $\rightarrow$ {\em character}, {\em s}; \\
{\em s} $\rightarrow$ {\em symbol} $\vert$ NULL; \\
{\em character} $\rightarrow$ `a' $\vert$ ... $\vert$ `z' $\vert$ `0' $\vert$ ... $\vert$ `9'; \\
\end{tabular}
\end{center}
\caption{The syntax of an early version of the `SP' language \citep{wolff_1990}.}
\label{sp_syntax_figure}
\end{figure}

From these early beginnings, the current system---described in Chapter \ref{theory_chapter}---has slowly evolved. The strengths and weaknesses of each version of the system have been explored using computer models. At all stages, I have tried to avoid introducing any new constructs into the system without compelling reasons. At the same time, I have looked for opportunities to eliminate constructs if at all possible, and I have looked for ways in which the explanatory range of the system might be increased.

Although the syntax shown in Figure \ref{sp_syntax_figure} is quite simple, it is significantly more complicated than the format for knowledge in the current system:

\begin{itemize}

\item `Ordered-AND-objects' are now the {\em patterns} described in Chapter \ref{theory_chapter}.

\item `Unordered-AND-objects' have been dropped because groupings that have no intrinsic order can be modelled with ordinary patterns (as will be described in Section \ref{ordering_of_symbols_and_patterns}).

\item There is no need for `OR-objects' because the notion of alternatives in the representation of knowledge is implicit in the way the current system works.

\item The concept of a `variable' (shown as `\_' in Figure \ref{sp_syntax_figure}) is not needed as a primitive element of the system because the effect of a variable---and `values' for a variable---can be modelled in the system using symbols and patterns (see Section \ref{variable_value_type_definition}).

\item The concept of a `NULL' entity has been dropped although there may be a case for re-introducing it, as suggested in Section \ref{derive_patterns_section}.

\item Although the syntax in Figure \ref{sp_syntax_figure} does not contain a construct for `negation', there was quite a long period when I thought it would be necessary to introduce one. In the end, it became apparent that negation could be modelled by the use of appropriate patterns in the knowledge supplied to the system, without the need to provide for negation in the system itself (see Section \ref{propositional_logic_and_ic}).

\item There is no explicit provision in the system for repetition or looping, like {\em repeat ... until}, {\em for ...} or {\em while ...}. This is because repetition can be modelled with recursion using only symbols and patterns.

\item There is no need for inbuilt types or a dedicated mechanism for the creation of user-defined types because the same effect can be achieved using only symbols and patterns (see Section \ref{variable_value_type_definition}).

\end{itemize}

Apart from the progressive simplification of the syntax, the main developments have been in the processing that provides the `semantics' of the system:

\begin{itemize}

\item First it was necessary to develop a version of `dynamic programming' for finding full matches and good partial matches between patterns, without the shortcomings of traditional algorithms (see Section \ref{matching_one_pattern_with_another}). The result is described in Appendix \ref{matching_appendix} \citep[see also][]{wolff_1994_scaleable}.

\item As mentioned above, a major insight was that the explanatory scope of the system could be dramatically increased by generalising the concept of `pattern matching' to a version of the concept of `multiple alignment'. 

\end{itemize}

Because the SP version of the multiple alignment concept is different from the bioinformatics version, it seemed necessary to develop a new algorithm for building multiple alignments, different from any of those that had been developed in bioinformatics. The result was a waste of two years trying to develop a technique that ultimately proved abortive! Another 18 months was needed to develop a working model using a modified version of one of the techniques already established in bioinformatics. Happily, the system now works well and has remained stable for some time.

\section{The SP theory and related research}

As an attempt to integrate ideas across several areas, the SP theory naturally has many connections with other research. In the main, connections of that kind are noted or discussed at appropriate points in the chapters that follow. This section first makes a few brief remarks about how the SP theory compares with other attempts at integration and then there is a summary of key features that serve to distinguish the SP theory and research programme from other research in computing, artificial intelligence and cognitive science.

\subsection{Unified theories of cognition}

As a theory of human cognition, the theory belongs in the tradition of {\em Unified Theories of Cognition}, pioneered by Allen Newell \citeyearpar{newell_1990, newell_1992} and others, and motivated by the kinds of concerns about cognitive psychology that were quoted above (Section \ref{breadth_and_depth}). No attempt will be made to make a detailed comparison with other unified theories but I will highlight the main similarities and differences between the SP theory and two of the best-known ones: Soar\index{Soar|(} \citep{newell_1990, rosenbloom_etal_1993} and ACT-R\index{ACT-R|(} \citep{anderson_lebiere_1998}. 

The main differences between the SP theory and the other two theories are:

\begin{itemize}

\item The SP theory is not primarily a theory of human cognition: it is a theory of information processing in any kind of system, either natural or artificial.

\item The SP theory is founded on minimum length encoding principles but these are not recognised as central organising principles in the other unified theories of cognition.

\item The concept of multiple alignment as it has been developed in the SP framework has no counterpart in the other theories.

\end{itemize}

Like Soar and ACT-R, the SP system is an {\em architecture} for cognition, not a model of any specific piece of behaviour. Like the other models---or, indeed, a new-born child---its detailed behaviour will depend on knowledge that it is given or that it acquires by learning. A process of matching patterns is prominent in all three systems, although dynamic programming in the SP system appears to provide for more flexibility than in the other two systems.

Soar stores its permanent knowledge as `production rules' (like ACT-R) and its temporary knowledge as `objects' with `attributes' and `values'. By contrast, the SP system stores all knowledge as arrays of symbols in one or two dimensions called {\em patterns}. Unlike Soar, the SP system has no explicit concept of `goal' or `subgoal' (it is anticipated that concepts of that kind may be modelled with patterns). In Soar, all learning is achieved by `chunking', whereas learning in the SP framework derives from the process of forming multiple alignments and achieves chunking---and other kinds of learning---as a by-product of that process.

In the SP theory, there are no `modules' or `buffers' as there are in the ACT-R theory and there is no formal distinction between `declarative memory' and `procedural memory' (see Section \ref{declarative_procedural}).%
\index{Soar|)}\index{ACT-R|)}

\subsection{Key features of the SP theory and research programme}

As we have noted, the SP theory has many connections with other research. As a guide through the maze, readers may find it useful to keep in mind the following points which, together, serve to distinguish the SP theory and research programme from parallel and connected lines of research:

\begin{enumerate}

\item As previously mentioned, the SP theory is a theory of information processing in {\em any} system, either natural or artificial.

\item As such, it is also a theory of computing. As a theory of computing, the SP theory is `Turing equivalent' in the sense that it can model the operation of a universal Turing machine---but it is distinct from the universal Turing machine and equivalent models such as Lamda Calculus \citep{church_1941} or Post's Canonical System \citep{post_1943} because it is built from different foundations, it has much more to say about the nature of intelligence, and it has other advantages described in Chapter \ref{computing_chapter}.

\item A central idea in the SP theory is that {\em all} kinds of computing or information processing is achieved by information compression in accordance with minimum length encoding principles, as described in Section \ref{mle_section}. The apparent paradox of `decompression by compression' and how it may be resolved is discussed in Section \ref{decompression_by_compression}. 

\item The SP theory is distinct from Kolmogorov complexity theory, minimum length encoding and algorithmic information theory \citep{li_vitanyi_1997}   
 because the latter three inter-related areas of thinking are founded on the Turing model of computing whereas the SP theory is itself a new model of computing, built from new foundations. The SP theory is also distinguished from these areas by the two points that follow.

\item Another important idea in the SP theory is the conjecture that {\em all} kinds of information compression may be understood in terms of the matching and unification of patterns coupled with the creation and use of `code' symbols to encode information economically.

\item More specifically, it is proposed that matching and unification of patterns and economical encoding of information is achieved via a process of {\em multiple alignment} that is similar to that concept in bioinformatics but with important differences.

\end{enumerate}

\section{Presentation}

Chapter \ref{foundations_chapter}, next, describes ideas and observations on which the SP theory is founded, or that have provided motivation for the development of the theory, or are simply part of the background thinking that has influenced the ways in which theory has developed.

Chapter \ref{theory_chapter} describes the theory itself and one of the main computer models in which the theory is embodied. After that, Chapter \ref{computing_chapter} shows how the SP theory can model the operation of a universal Turing machine and explains the advantages of the theory compared with earlier theories of computing.

In Chapters that follow, applications of the SP theory are explored: in the processing of natural languages (Chapter \ref{language_chapter}), in pattern recognition and information retrieval (Chapter \ref{rr_chapter}), probabilistic reasoning (Chapter \ref{pr_chapter}), planning and problem solving (Chapter \ref{pps_chapter}), learning of new knowledge (Chapter \ref{learning_chapter}), and in the interpretation of concepts in mathematics and logic (Chapter \ref{maths_logic_chapter}).

In Chapter \ref{neural_chapter} I describe how the SP theory may be realised with known mechanisms and processes in the brain. Constructs in the theory map quite neatly onto a revised version of Hebb's \citeyearpar{hebb_1949} concept of a `cell assembly' and the revised version overcomes several weaknesses of the original concept.

Chapter \ref{psychology_chapter} considers how the SP theory relates to some current themes in cognitive psychology and Chapter \ref{future_chapter} outlines possibilities in the future development of the SP theory and in its applications.

%% file: foundations.tex
\chapter[Computing, Cognition and IC]%
{Computing, Cognition and Information Compression%
\protect\footnote{Based on \citet{wolff_1993}.}}%
\label{foundations_chapter}

\section{Introduction}

The purposes of this chapter are two-fold:

\begin{itemize}

\item To consider the nature of `information', `redundancy', `information compression', `probabilities'---and their inter-relations. These ideas, discussed in the next section, are centre-stage in the SP theory and throughout this book. 

\item The rest of the chapter provides a broad perspective on the several ways in which information compression may be seen in diverse areas of both computing and cognition. This chapter aims to show how a variety of established ideas may be seen as information compression and to illustrate the broad scope of this perspective in information systems of all kinds, both natural and artificial. 

\end{itemize}

\section{Information, redundancy, compression of information and probabilities}\label{information_ic_probabilities}

\index{information!theory|(}\index{information!compression|(}\index{information!redundancy|(}

The development of `information theory' (originally called `communication theory') in the first half of the 20th century gave mathematical precision to a concept that had, hitherto, been rather vague. Although this precision was and is very useful for calculating the bandwidth of communication channels and other applications, a presentation of the concepts purely in terms of the mathematics can have the effect of obscuring important underlying concepts. In this section, I shall focus mainly on what I perceive as the concepts behind the mathematics. There are many excellent sources that provide the mathematical details \citep[see, for example,][]{cover_thomas_1991}.

\subsection{Information}

Anything that contains recognisable variations may be seen as information. This includes light waves, sound waves, pictures (both static and moving), diagrams, written or spoken language, music, mathematical or logical formulae, Morse code, the bar code on a tin of baked beans, non-verbal nods, winks and smiles, and so on.

In this book, it is assumed that any continuously-varying `analogue' form of information (any kind of wave for example) can be converted into digital form with any desired level of precision, as described in Section \ref{raw_and_encoded_information}, below. The focus will be mainly on information in digital form.

To be more specific, it is assumed that any kind of information may be represented as an array of `symbols' in one or more dimensions, where a {\em symbol} is some kind of `mark' that can be seen to be the `same' as another symbol or `different' from it---but nothing in between. Within any given body of information, $I$, each symbol may be seen to belong to a {\em symbol type} comprising all other symbols within $I$ that are the same as the given symbol.\footnote{Here and throughout this book, the phrase ``(the or a) given body of information'' will be abbreviated as `$I$'.} Associated with every $I$ is an {\em alphabet} of the symbol types that appear in $I$.
 
In this book, the main focus is on arrays of symbols in one dimension. However, it is anticipated that the SP ideas will generalise to arrays of symbols in two dimensions and possibly more. For this reason, the relatively general term {\em pattern} is normally used to describe arrays of symbols in one or two dimensions rather than more specific terms such as `string' or `sequence'.

\subsection{`Direct' and `encoded' kinds of digitisation}\label{raw_and_encoded_information}

\begin{figure}[!hbt]
\centering
\includegraphics[width=0.9\textwidth]{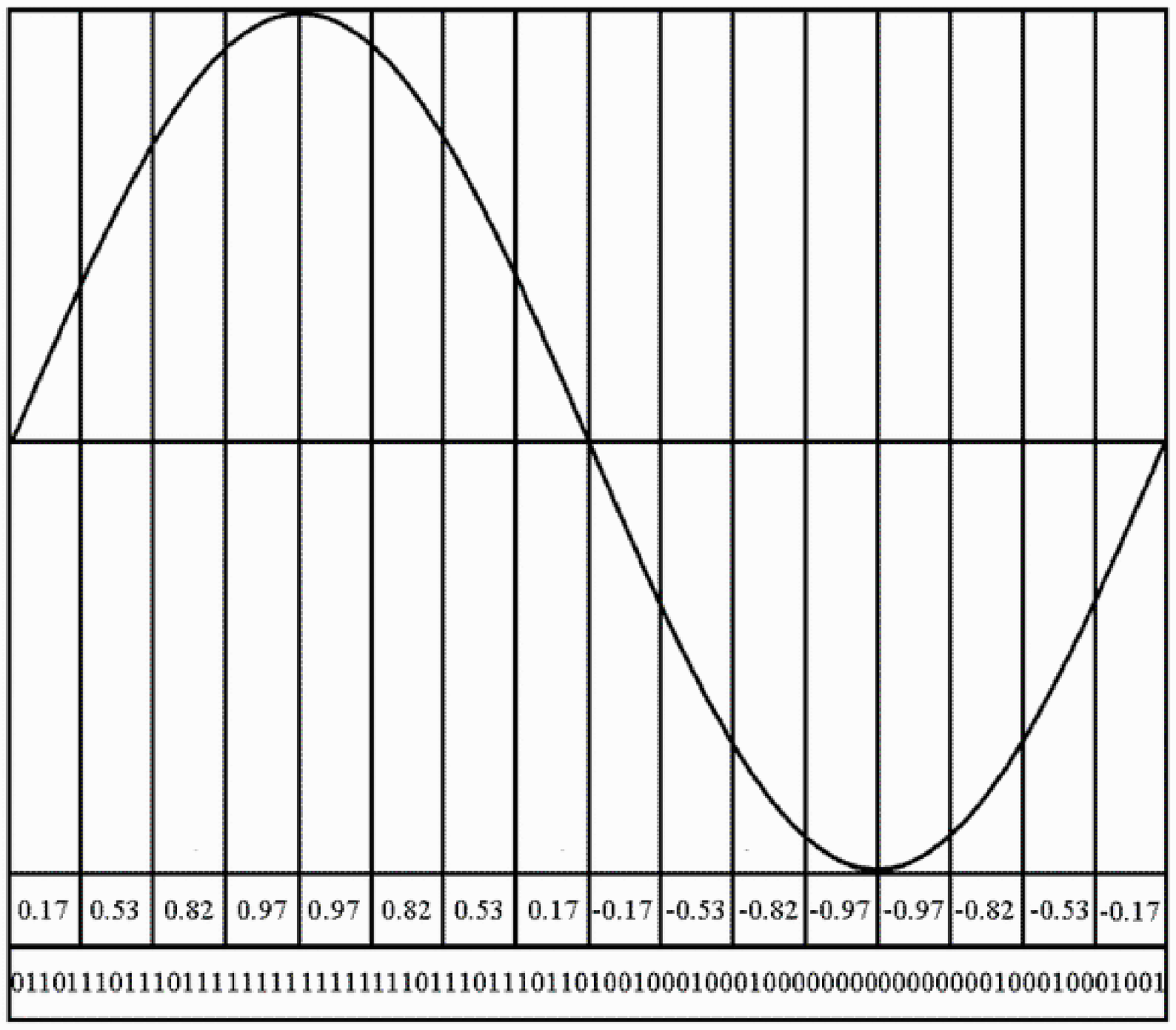}
\caption{Digitisation of an analogue wave as numbers or as densities of `0's and `1's.}
\label{raw_coded_figure}
\end{figure}

Analogue information such as speech or music is normally digitised as a sequence of numbers, each one representing the amplitude of the wave at a given moment. This is represented schematically by the sequence of numbers immediately below the sin wave in Figure \ref{raw_coded_figure}.

Given our familiarity with numbers, this system may seem simple and natural. But it should not be forgotten that, even if these numbers are represented in binary code, the way in which they represent the original analogue information requires a knowledge of the number system and it requires a process of {\em interpretation} in terms of that knowledge. Understanding the nature of the interpretative process---with any kind of knowledge---is one of the main themes of this book.

An alternative to this `encoded' form of digitisation is represented schematically at the bottom of Figure \ref{raw_coded_figure}. Here, the amplitude of the original analogue wave is represented by the {\em density} of the digital symbols `0' and `1'. This kind of digitisation, which is similar to the way in which the density of black dots is used to represent different shades of grey in a newspaper photograph, does not require a knowledge of the number system or a process of interpretation in terms of that system. It is a relatively `direct' form of digitisation which does not depend on numbers or any comparable form of encoding.

\subsection{Randomness, redundancy, structure and compression}\label{randomness_redundancy_structure}

Patterns of information may be {\em random} and, in those cases, we perceive the information to be totally lacking in `structure'---like `snow' on a television screen. Most of the information we encounter in practical applications is not random but is structured in some way. This means that the information contains {\em redundancy}, a term whose technical meaning is quite close to the everyday meaning of `something that is surplus to requirements'. If there is redundancy in $I$, there is, in effect, repetition of information. And information that is repeated is, in terms of communication, unnecessary. Of course, redundancy can be useful in other ways such as guarding against catastrophic loss of information (as in the use of backup copies or mirror disks in computing), or in speeding up processing in databases that are distributed over a wide area, or in aiding the correction of errors when information is corrupted (as in error-correcting codes).

\begin{figure}[!hbt]
\centering
\includegraphics[width=0.9\textwidth]{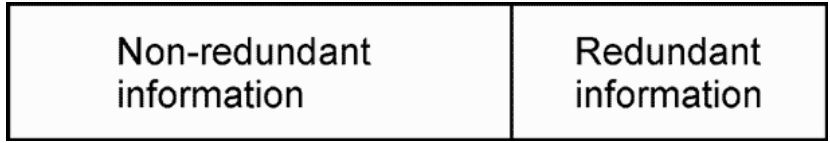}
\caption{A schematic representation of any body of information, $I$, showing mutually-exclusive {\em non-redundant information} and {\em redundant information}.}
\label{redundancy_figure}
\end{figure}

Schematically, any body of information, $I$, may be seen to comprise mutually-exclusive {\em non-redundant information} and {\em redundant information} and nothing else, as shown in Figure \ref{redundancy_figure}. Of course, the redundant and non-redundant parts of $I$ are normally intermingled, not arranged in two blocks as shown in the figure.

Information compression means reducing the size of $I$ by removing information from it. If the only information to be removed is redundant information then in principle and usually in practice it is possible to restore $I$ to its original state with complete fidelity. This is {\em lossless} compression. If some non-redundant information is removed as well as redundant information, this is {\em lossy} compression. Lossy compression can be beneficial in some applications---by speeding up the compression process or increasing the level of compression---but the penalty is that, when non-redundant information has been discarded, it is never possible to restore $I$ exactly to its original form.

In principle, lossy compression could be achieved by discarding non-redundant information and preserving all redundant information. But in practice, lossy techniques, like lossless techniques, are designed to remove as much redundant information as is practically possible. In general, it will be assumed that {\em all} techniques for information compression are designed with the primary emphasis on reducing redundancy in information. 

The foregoing remarks give a broad view of randomness, redundancy and information compression but we need to be more precise about the nature these concepts. They are considered in more detail in the subsections that follow.

\subsection{Shannon's information theory}\label{shannons_information_theory}

\index{probability|(}%
Building on earlier work by Boltzmann, Hartley and others, Claude Shannon \citeyearpar{shannon_weaver_1949} developed the idea that the communicative value of a symbol or other `event' is related to its probability: ``It will rain tomorrow'' is more informative than ``It will rain some time in the coming year''. In this theory, the average quantity of information conveyed by one symbol in a sequence is:

\begin{equation}
H = - \sum_{i = 1}^{i = n} p_i \log p_i,
\label{shannons_information_equation}
\end{equation}

\noindent where $p_i$ is the probability of the $i$th symbol type in the set of $n$ available symbol types. If the base for the logarithm is 2, then the information is measured in `bits'.

If the probabilities of all the symbol types in $I$ are equal, $I$ contains the maximum possible information for the given alphabet of symbol types. In this case, $I$ is {\em random} and contains no redundancy. If some symbols are more probable than others, $I$ is not random and it contains redundancy. If the redundant information is removed, the size of $I$ can be reduced without sacrificing any of its communicative value. 

This analysis of the nature of information, randomness and redundancy was a major advance when it was first described and remains very useful for many purposes. However, it suffers from three main problems:

\begin{itemize}

\item Probabilities of symbols include both {\em absolute} probabilities and {\em conditional} probabilities. The absolute probability of each symbol type can be derived straightforwardly as a normalised measure of the frequency of occurrence of that symbol type in $I$ but conditional probabilities are more problematic. The difficulty arises in defining the condition or context for each conditional probability. Should it be the symbol immediately preceding the target symbol or the symbol immediately following---or both these things? Should it be some string of two or more symbols before or after the target symbol (or both these things)? If so, how long should those strings be? Should there be a gap of one or more symbols between the context and the target symbol or should there be gaps within the context string? If so, how big or numerous should those gaps be? Should the target symbol be a pair of symbols, or three or four ...? In general, there is an explosion of possibilities and it is not obvious what the `correct' answer should be or if, indeed, there is any `correct' answer.

\item In Shannon's theory, any $I$ of finite size is regarded as a sample of the world and probabilities derived from that $I$ are regarded as estimates of probabilities in the world. If $I$ is large, these estimates may be regarded as being tolerably accurate. But if $I$ is small, the probabilities derived from it may be regarded as too unreliable to be trusted.

\item There are some kinds of data that appear to be random in terms of absolute and conditional probabilities of symbols but are, nevertheless, known to have underlying regularities. One example is the decimal expansion of $\pi$ where the seemingly random sequence of digits conforms to a simple formula. Another example is the apparently random stream of digits generated by the kind of {\em random()} function provided in most computer systems. Again, there is an underlying regularity defined by a simple formula. These kinds of redundancy will be referred to as {\em covert} forms of redundancy.

\end{itemize}

\subsection{Algorithmic information theory}\label{ait_section}

An interesting alternative to Shannon's information theory is algorithmic information theory, developed by Gregory Chaitin and others \citep[see, for example,][]{chaitin_1987, chaitin_1988, li_vitanyi_1997}. The key idea here is that, if $I$ can be generated by a computer program that is shorter than $I$ then the information is not random and contains redundancy. If no such program can be found then the information is regarded as random and contains no redundancy.

This approach to understanding randomness, redundancy and information compression neatly sidesteps the second of the two problems associated with Shannon's theory, described above. Each $I$ is regarded as complete in itself and not merely a sample of something larger. If some algorithm can be discovered or invented that is smaller than $I$ but can create $I$, then $I$ is not random. Otherwise, it is. This is true no matter how small $I$ may be.

Another advantage of algorithmic information theory is that it accommodates the covert kinds of redundancy found in the decimal expansions of $\pi$ or the output of a typical {\em random()} function. In each such case, the relevant formula represents an algorithmic compression of the data.

An implication of algorithmic information theory is that, while it may be possible to prove that $I$ is not random in particular cases, it is not possible to prove the converse, except when $I$ is very small. Although a lot of effort may have been expended unsuccessfully in trying to find a means of compressing $I$, there is, for any $I$ of realistic size, always the possibility that a little extra effort would yield a positive result.

Another implication of this view of randomness and redundancy is that, if $I$ has been as fully compressed as is practicable then, with due allowance for the possibility that more resources may allow more compression to be achieved, we may regard $I$ as random.

\subsection{Redundancy as repetition of patterns and information compression by the unification of patterns}\label{ic_repetition_of_patterns}

In Shannon's theory, redundancy is seen in terms of unbalanced probabilities of symbols while algorithmic information theory sees it in terms of the length of a computer program relative to raw data. A third possibility, discussed here, is {\em redundancy as repetition of patterns}.

If, for example, $I$ is the pattern `I N F O R M A T I O N I N F O R M A T I O N', it is clear that it contains a redundant copy of the pattern `I N F O R M A T I O N'. A little less clearly, the same is true if $I$ is `I X Y Z N F O R M A I J K T I O N I N F L M N O R M A T P Q R I O N'. Thus, any instance of a pattern that repeats within $I$ may be a coherent substring of $I$ or it may be a {\em subsequence} of $I$ in which the constituent symbols are not necessarily contiguous within $I$, and the gaps within the subsequence may occur at varying positions from one instance to another. A substring or subsequence that repeats in different contexts is sometimes called a {\em chunk} of information (see Section \ref{techniques_for_ic}, below).

A natural adjunct to redundancy-as-repetition-of-patterns is that redundancy can be reduced---and $I$ can be compressed---by the merging or {\em unification}\index{unification} of two or more patterns so that they are reduced to one. To avoid misunderstanding, the meaning of the term `unification' as it has been used here is different from and simpler than the meaning of that term in logic, although there are recognisable similarities. Unless otherwise indicated, the meaning of the term throughout this book will be a simple merging of identical patterns.

\subsubsection{Preserving non-redundant information by the use of `identifiers' and `references'}\label{identifiers_and_references}

If two or more instances of a pattern within $I$ are reduced, by unification, to a single instance then some non-redundant information is lost, namely the position within $I$ of every one of the original instances, except the one instance that remains---assuming that it has kept its original position within $I$ and has not itself been moved to some other location external to $I$.

For some applications, it may not matter that this non-redundant information has been lost. But if lossless compression is required, it is necessary to preserve the information about the positions of the instances that have been removed from $I$. This can be achieved by attaching some kind of relatively short `name', `label', `tag' or {\em identifier} to the single `unified' instance of the repeating pattern and then inserting a copy of that identifier in the position of each instance that has been removed. Each such copy will be termed a {\em reference} to the unified pattern. Identifiers and references will be referred to collectively as {\em codes}.

\begin{figure}[!hbt]
\centering
\includegraphics[width=0.9\textwidth]{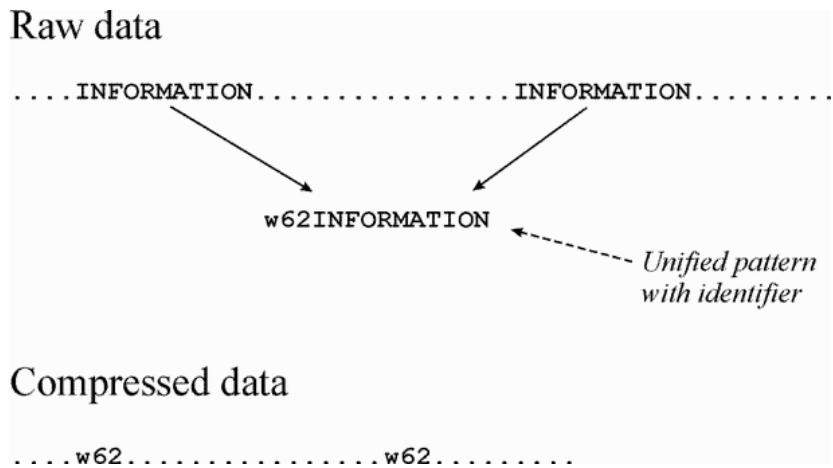}
\caption{A schematic representation of the way two instances of the pattern `INFORMATION' in a body of raw data may be unified to form a single `unified pattern', with `w62' as an identifier assigned by the system. The lower part of the figure shows how the raw data may be compressed by replacing each instance of `INFORMATION' with a copy of the corresponding identifer.}
\label{unification_figure}
\end{figure}

This basic mechanism, illustrated in Figure \ref{unification_figure}, is most straightforward when each instance of the repeating pattern is a coherent substring within $I$ but the technique can be generalised for fragments that repeat as discontinuous subsequences within $I$. Examples will be seen in Chapter \ref{learning_chapter}.

\subsubsection{Huffman coding and related techniques}

An important point about the use of codes is that information compression can be optimised if {\em frequently-occurring patterns are given shorter identifiers than patterns that occur rarely}. This basic idea can be made mathematically precise by using the well-known Huffman coding method or the slightly less efficient Shannon-Fano-Elias coding (both these techniques are described in \citet{cover_thomas_1991}).

\subsubsection{Redundancy, frequency and size}\label{redundancy_frequency_size}

In the redundancy-as-repetition-of-patterns view, it is clear that the amount of redundancy represented by a repeating pattern---and the amount of compression that can be achieved by unification---depends on the {\em frequency} of occurrence of the pattern within $I$ and also on its {\em size}. Patterns that are both large and frequent represent more redundancy and yield more compression than patterns that are small and rare. This follows directly from the compression that can be achieved by the unification of repeating patterns and is only indirectly related to the use of Huffman coding or similar techniques.

A key point in this connection is that, for any given frequency of occurrence of a given pattern, there is a {\em minimum size} below which no (lossless) compression can be achieved. This is because allowance has to be made for the information `cost' of the identifiers and references that are needed for lossless compression. A table of random numbers contains many repeating patterns but none of them are large enough or frequent enough to support lossless compression.

The minimum size required for compression varies with the frequency of the pattern: patterns that occur rarely have a larger minimum size than patterns that occur more frequently. Conversely, useful compression can be achieved with large patterns even if their frequency of occurrence is quite low, whereas small patterns need to be more frequent to yield lossless compression by unification.

\subsubsection{Searching for repeating patterns and the need for constraints}\label{matching_searching_and_constraints}

A question about repeating patterns that has been glossed over so far is how they are to be discovered. At first sight, this is simply a matter of comparing one pattern with another to see whether they match each other or not. But there are, typically, many alternative ways in which patterns within $I$ may be compared---and some are better than others. We are interested in finding those matches between patterns that, via unification, yield most compression---and a little reflection shows that this is not a trivial problem.

Maximising the amount of redundancy found means maximising $R$ where:

\begin{equation}
R = \sum_{i = 1}^{i = n} (f_i - 1) \cdot s_i,
\label{pattern_matching_equation}
\end{equation}

\noindent $f_i$ is the frequency of the $i$th member of a set of $n$ patterns and $s$ is its size in bits. As previously noted, patterns that are both big and frequent are best. This equation applies irrespective of whether the patterns are coherent substrings or patterns that are discontinuous within $I$.

Maximising $R$ means searching the space of possible unifications for the set of big, frequent patterns that gives the best value. For a sequence containing $N$ symbols, the number of possible subsequences (including single symbols and all composite patterns, both coherent and fragmented) is:

\begin{equation}
P = 2^{N} - 1.
\label{unifications_equation}
\end{equation}

The number of possible comparisons is the number of possible pairings of subsequences which is:

\begin{equation}
C = P(P - 1) / 2.
\label{comparisons_equation}
\end{equation}

For all except the very smallest values of $N$ the value of $P$ is very large and the corresponding value of $C$ is huge. In short, the abstract space of possible comparisons between patterns and thus the space of possible unifications is, in the great majority of cases, astronomically large. Since the space is normally so large, it is not feasible to search it exhaustively. For that reason, we cannot normally guarantee to find the theoretically ideal answer and normally we cannot know whether or not we have found the theoretically ideal answer. In general, we must be content with answers that are ``good enough''.

In all practical methods for searching for `good' matches between patterns, there is a need for {\em constraints} that reduce the amount of searching that needs to be done. These constraints come it two main forms:

\begin{itemize}

\item {\em Absolute Constraints}. Search costs may be kept within bounds by searching only within a pre-defined sub-set of the set of possible comparisons. For example, comparisons may be made only between coherent substrings; or the maximum size of patterns may be restricted; or some other arbitrary constraint may be imposed.

\item {\em Heuristic Constraints}. The costs of searching may be reduced without undue sacrifices in effectiveness by applying a measure of redundancy to guide the search. In `heuristic' search techniques (which include `hill climbing' (or `descent'), `beam search', `best-first search', `branch-and-bound search', `simulated annealing', `genetic algorithms' and others), the search is conducted in stages and, at each stage, searching is curtailed in those parts of the search space that have proved sterile and it is concentrated in areas that are indicated by the metric.\footnote{For this reason, these kinds of search technique are sometimes also called `metrics-guided search'.} With this kind of constraint, it is possible in principle to reach any part of the search space. Nevertheless, this kind of constraint can reduce dramatically the amount of searching that needs to be done to achieve acceptable results.

\end{itemize}

\noindent Either or both of these two kinds of constraint may be used.

In general, there is a trade-off between accuracy and speed. A search that is heavily constrained can be achieved quite quickly but---for the kinds of problems that are the focus of interest in artificial intelligence---the results may not be very good. Better results can normally be obtained by relaxing the constraints and taking more time.

As a rough generalisation, conventional computing systems use absolute constraints whereas artificial intelligence applications use heuristic constraints. With certain kinds of problem, conventional systems can produce results relatively fast but they lack the flexibility needed in artificial intelligence applications.

\subsubsection{Can all kinds of redundancy be seen as repeating patterns?}\label{rrp_conjecture}

Another question that arises in connection with redundancy-as-repetition-of-patterns is whether {\em all} kinds of redundancy can be seen in these terms or whether there is a distinction to be drawn between some kinds of redundancy that can be seen as repeating patterns and others that cannot.

At first sight, the answer to this question is clear. Kinds of data such as the previously-mentioned decimal expansions of $\pi$ or the output of {\em random()} functions, do not contain repeating patterns of the kind we have been discussing but they are known to have underlying regularities. And there are techniques for information compression---such as `arithmetic coding'---that work well without any apparent need to search for repeating patterns or to unify patterns that are the same. In short, it seems that, while some kinds of redundancy may be seen as repeating patterns, there are other kinds of redundancy that cannot be seen in these terms.

Notwithstanding the kinds of counter-examples just mentioned, a working hypothesis in this programme of research is that:

\begin{quote}

{\em All kinds of redundancy may be understood as repeating patterns and compression of information by the reduction of redundancy may always be understood in terms of the unification of patterns that match each other.}

\end{quote}

\noindent This working hypothesis will be referred to as the {\em redundancy-as-repetition-of-patterns conjecture}. According to this conjecture, {\em all} the several methods for compressing information---ZIP programs, linear predictive coding, arithmetic coding, Huffman coding, fractal compression, JPEG, MPEG, and so on---may at the most fundamental level be understood in terms of the matching and unification of patterns.

Now that the SP framework is relatively mature, it provides fairly strong evidence in favour of the hypothesis. The main steps in the argument are as follows:

\begin{enumerate}

\item The SP framework is dedicated to the reduction of redundancy by the matching and unification of patterns. There is no other mechanism or process in the framework for reducing redundancy.

\item It is possible to model the operation of a universal Turing machine within the SP framework. This is explained in some detail in Chapter \ref{computing_chapter}.

\item In keeping with the key idea in algorithmic information theory (described in Section \ref{ait_section}), it seems reasonable to assume that {\em any} kind of redundancy---including the covert kinds of redundancy mentioned earlier---can be expressed algorithmically and that {\em any} technique for reducing redundancy can be implemented as a program to run on a universal Turing machine.

\item If the foregoing points are accepted, it follows that {\em any} kind of redundancy and {\em any} technique for compressing information by reducing redundancy may be understood in terms of the matching and unification of patterns. 

\end{enumerate}

Another argument in support of the redundancy-as-repetition-of-patterns conjecture is based on the observation that all methods for information compression depend on measures of the frequency or probability of symbols or patterns. Given that `probability' in this context is a normalised measure of frequency and that frequency implies a process of counting, it is clear that all methods for information compression depend on counting. As will be argued in Section \ref{nature_of_counting}, counting implies recognition of the things to be counted and their assimilation to a single concept---and this means compression of information by the matching and unification of patterns. Hence, all methods for information compression depend on the matching and unification of patterns.

\subsubsection{Redundancy-as-repetition-of-patterns compared with Shannon's theory and algorithmic information theory}

Before leaving the subject of redundancy-as-repetition-of-patterns, a few words are in order about how this concept of redundancy compares with the way in which redundancy is treated in Shannon's theory and in algorithmic information theory.

In `classical' treatments of Shannon's theory, the main focus is on the probabilities of symbol types. It is recognised that symbols do not exist in isolation so conditional probabilities are brought into the picture as well as absolute probabilities. But, as we saw earlier, this leads to difficulties in defining the relevant context or contexts for each symbol type.

In principle, Shannon's theory may be applied to redundancy-as-repetition-of-patterns if the concept of `symbol' is generalised to include sequences of two or more symbols. But any such generalisation of Shannon's theory would need to take account of the way in which subsequences within $I$ may overlap each other (unlike individual symbols) and the analysis would also need to recognise that the sizes of patterns are at least as important as their frequencies. None of this would overcome the difficulty arising from the fact that probabilities derived from $I$ are regarded as estimates and these estimates are unreliable if $I$ is small.

\sloppy Another difficulty with Shannon's theory---that may be described as `psychological'---is that the emphasis on the probability of individual symbols or smallish sequences of fixed size (`$n$-grams'), coupled with the apparent need to derive these probabilities from samples that are as large as possible, seems to have led some researchers to overlook the importance of the sizes of patterns and the way in which an increase in the size of repeating patterns can bring down the minimum frequency or probability required for lossless compression. It is sometimes assumed that high frequencies are needed before redundancy can be detected, despite the fact that lossless compression can be achieved by the unification of patterns of quite modest size even though their frequency of occurrence in $I$ may be as low as 2.

One difference between algorithmic information theory and redundancy-as-repetition-of-patterns is that algorithmic information theory is based on the concept of a universal Turing machine but that concept has no place in redundancy-as-repetition-of-patterns. However, since a universal Turing machine can be modelled within the SP framework, redundancy-as-repetition-of-patterns can borrow the elegant idea that $I$ contains redundancy if it can be compressed, and adapt it in terms of the matching and unification of patterns within the SP framework. Redundancy-as-repetition-of-patterns can also accommodate the covert kinds of redundancy mentioned earlier.

The simple, intuitive idea that information can be compressed by the unification of matching patterns is an obvious implication of redundancy-as-repetition-of-patterns. Neither Shannon's theory nor algorithmic information theory recognise this idea.\index{probability|)}

\subsection{Techniques for compressing information}\label{techniques_for_ic}

There are many techniques for information compression including lossy techniques such as JPEG, MPEG and fractal compression and also lossless techniques such as arithmetic coding and Lempel-Ziv algorithms used in the popular `ZIP' programs for information compression.

As was argued above (Section \ref{rrp_conjecture}), it seems possible that all these techniques may be understood in terms of the matching and unification of patterns but in some this is more obvious than in others. As previously noted, this mechanism is relatively obscure in techniques like arithmetic coding but in Lempel-Ziv algorithms and some other related techniques \citep[see, for example,][]{storer_1988} the reduction of redundancy by the matching and unification of patterns is clear to see.

Three main variants of the technique can be recognised:

\begin{itemize}

\item {\em Chunking-With-Codes}\index{information!compression!chunking-with-codes}. This is the basic technique as described in Section \ref{ic_repetition_of_patterns}, above. Two or more instances of a coherent substring or `chunk' of information are reduced to a single instance. The unified chunk has an identifier and the positions of the original instances may be marked with copies of the identifier (`references'), as described earlier.

This chunking-with-codes technique is so widespread and `natural' that we hardly notice it: we often use the abbreviation `PC' as a short substitute for the relatively long pattern `personal computer'; a citation like `\citet{storer_1988}' may be seen as a reference to the relatively long bibliographic details of the book given in the references section of the article; in everyday speaking and writing, names of people, places and so on may all be regarded as relatively short references to concepts where the full description in each case represents a relatively large body of information. No doubt, readers can think of many other examples.

\item {\em Schema-Plus-Correction}\index{information!compression!schema-plus-correction}. A variant of the basic chunking-with-codes technique is another technique that is often called {\em schema-plus-correction}. In this case, the unified pattern is not a monolithic chunk but is a chunk containing gaps or holes within which a variety of other patterns may appear on different occasions. In this case, the unified pattern is the {\em schema} and the other patterns that fill the gaps are {\em corrections} to or completions of the schema. Identifiers and references are used to connect each such correction with its proper place in the schema.

In everyday life, a common example is a menu in a restaurant. In this case the 
schema is the basic framework of the menu, e.g., `Starter ... main course ... sweet course ...' and the corrections or completions are the dishes chosen to fill the gaps. Another example is any kind of form with fields that will be filled with specific information each time the form is completed. 

\item {\em Run-Length Coding}\index{information!compression!run-length coding}. If a sequence of symbols contains a pattern that repeats in a sequence of instances that are contiguous, one with the next, then information compression can be achieved by reducing the repeating series to one instance with something to mark the repetition or, for fully lossless compression, with something to show the number of repetitions. For example, the pattern `a x y z x y z x y z x y z x y z x y z x y z x y z x y z x y z b' may be reduced to something like `a (x y z)* b' in the case of lossy compression or, for lossless compression, something like `a (x y z)[10] b'.

In programming terms, this kind of {\em run-length coding} can always be expressed using a `recursive' function---one that contains one or more calls to itself, either directly or via calls to other functions. And, as explained in Section \ref{functional_and_structured_programming}, below, functions can themselves be understood as examples of chunking-with-codes or schema-plus-correction.

\end{itemize}

\subsection{Minimum length encoding}\label{mle_section}

\index{minimum length encoding|(}

Another topic to be described in connection with information compression is sometimes known as {\em minimum message length} encoding or {\em minimum description length} encoding. An umbrella term that embraces both those variants is {\em minimum length encoding}, the term that will be used throughout this book.

This area of thinking was pioneered by \citet{solomonoff_1964} and also by \citet{wallace_boulton_1968} and by \citet{rissanen_1978} \citep[see][]{li_vitanyi_1997}. The principle arose in connection with the problem of trying to discover, infer or induce a grammar (or similar knowledge structure) from a sample of `raw' data. The grammar is a set of rules that are intended to describe the raw data but for any given body of data (let us call it $I$) there are infinitely many different grammars that will do.

Describing the discovery of minimum length encoding principles, \citet{solomonoff_1997} writes:

\begin{quotation}

``I was trying to find an algorithm for the discovery of the `best' grammar for a given set of acceptable sentences. One of the things I sought was: Given a set of positive cases of acceptable sentences and several grammars, any of which is able to generate all the sentences, what goodness of fit criterion should be used? It is clear that the `{\em ad hoc} grammar', that lists all of the sentences in the corpus, fits perfectly. The `promiscuous grammar' that accepts any conceivable sentence, also fits perfectly. The first grammar has a long description; the second has a short description. It seemed that some grammar half-way between these, was `correct'---but what criterion should be used?''

\end{quotation}

If $I$ is an alphabetic text, the `promiscuous grammar' is:

\begin{center}
\begin{tabular}{l}
S $\rightarrow$ char S \\
char $\rightarrow$ A \\
char $\rightarrow$ B \\
... \\
char $\rightarrow$ Z
\end{tabular}
\end{center}

\noindent In other words, construct the text from letters chosen from the alphabet, one character at a time, at random. This grammar will describe $I$ but it will also describe many other alphabetic texts as well.

The {\em ad hoc} grammar is one that contains just one rule:

\begin{center}
S $\rightarrow$ {\em a copy of $I$}
\end{center}

\noindent This grammar describes $I$ but it cannot describe anything else.

Between these two extremes are many other grammars, including those that contain any amount of `garbage' in addition to the information needed to describe $I$. As Solomonoff says, some kind of criterion is needed for choosing one or more `good' grammars and avoiding the `bad' ones.

The minimum length encoding principle depends on the idea that a grammar may be used to encode $I$ succinctly. Using the grammar, a `program' may be constructed that describes $I$ in an economical manner. More concretely, each rule in a grammar may be seen as a pattern with a relatively short identifier as described in Section \ref{ic_repetition_of_patterns}. $I$ may be coded economically in terms of those short identifiers, much as was described earlier. The technique needs to be generalised to take account of high-level abstract rules in the grammar but the basic principle is the same as in Section \ref{ic_repetition_of_patterns} (see also Sections \ref{grammar_section} and \ref{ma_evaluation}).

The key idea in minimum length encoding is that, in grammar induction and related kinds of processing, one should seek to minimise $(G + E)$, where $G$ is the size (in bits) of the `grammar' (or comparable knowledge structure) that is under development and $E$ is the size (in bits) of the raw data ($I$) when it has been encoded in terms of the grammar. This principle guards against the induction of trivially small `promiscuous' grammars (where a very small $G$ is offset by a relatively large $E$) and over-large or `{\em ad hoc}' grammars (where $E$ may be small\footnote{If the grammar is very poor it may not even achieve a small $E$.} but this is offset by a relatively large $G$).

Bearing in mind that a complete description of $I$ includes both the `grammar' and `the raw data in its encoded form', then minimum length encoding boils down to a very simple idea: ``Given a sample of language or other body of data, compress it as much as possible in a lossless manner''. As we shall see in Section \ref{grammatical_inference_and_generalisation}, below, lossless compression as just described is entirely compatible with the kind of lossy compression needed to achieve generalisations and inductive predictions.

\subsubsection{Minimum length encoding, algorithmic information theory and Kolmogorov complexity theory}

Ideas that have been developed under the minimum length encoding banner are closely related to algorithmic information theory (Section \ref{ait_section}) and these two areas are also closely related to `Kolmogorov complexity theory' \citep{li_vitanyi_1997}. There is much common ground amongst these three areas but there are also differences in emphasis and focus.

As was noted in Chapter \ref{introduction_chapter}, Kolmogorov complexity theory takes the Turing model of computing as `given'. The same is true of minimum length encoding and algorithmic information theory. By contrast, the SP theory is itself a new theory of computing built on foundations of pattern-matching, unification, and multiple alignment.

The SP theory has adopted the central principle in minimum length encoding---the goal of minimising $(G + E)$ as described in the previous subsection---but it has not adopted the assumption that the Turing model is the reference model of computing. Hence, it is more accurate to say that the SP theory is based on `minimum length encoding principles' than that it is based on `the minimum length encoding theory' or `minimum length encoding'.

\subsubsection{Optimisation and `learnability' theory}\label{optimisation_and_learnability}

The idea that learning is a process of optimisation, guided by minimum length encoding principles, differs sharply from others such as `language identification in the limit'\index{learning!identification in the limit} \citep{gold_1967} or `probably approximately correct learning'\index{learning!probably-approximately-correct} \citep[see][pp. 339--350]{li_vitanyi_1997} because there is no pre-defined `target' grammar or equivalent structure against which the correctness of learning may be judged. The aim of learning is simply to minimise $(G + E)$ (abbreviated hereinafter as $T$), as far as that can be achieved in practice.

One might suppose that the grammar with the smallest possible value for $T$ constitutes the target grammar. But, unlike target grammars in Gold's framework or in probably-approximately-correct learning, this grammar is not pre-defined and, in most cases, it can never be known. The reason it cannot normally be known is that the abstract space of alternative grammars is normally too large to be searched exhaustively. Any practical system must necessarily use heuristic techniques to prune the search tree and this means that, in most cases, we can never be sure that we have found the smallest possible value for $T$ and we can thus never know what the grammar with the smallest possible value for $T$ would be. In minimum length encoding learning, we can compare one grammar with another in terms of $T$ but we have no means of knowing what the `correct' grammar should be.%
\index{minimum length encoding|)}

\subsection{Information, compression of information, inductive inference and probabilities}\label{probabilities_ic_section}

\index{probability|(}\index{reasoning!inductive|(}

Solomonoff \citeyearpar{solomonoff_1964} seems to have been one of the first people to recognise the close connection that exists between information compression and {\em inductive inference}: predicting the future from the past. The connection between them---which may at first sight seem obscure---lies in the redundancy-as-repetition-of-patterns view of redundancy and information compression:

\begin{itemize}

\item Patterns that repeat within $I$ represent redundancy in $I$, and information compression can be achieved by reducing multiple instances of any pattern to one.

\item When we make inductive predictions about the future, we do so on the basis of repeating patterns. For example, the repeating pattern `Spring, Summer, Autumn, Winter' enables us to predict that, if it is Spring time now, Summer will follow.

\end{itemize}

Thus information compression and inductive inference are closely related to concepts of frequency and probability. Here are some of the ways in which these concepts are related:

\begin{itemize}

\item Probability has a key r{\^o}le in Shannon's concept of information (Section \ref{shannons_information_theory}). 

\item Measures of frequency or probability are central in techniques for economical coding such as the Huffman method or the Shannon-Fano-Elias method \citep[see][]{cover_thomas_1991}.

\item In the redundancy-as-repetition-of-patterns view of redundancy and information compression, the frequencies of occurrence of patterns in $I$ is a main factor (with the sizes of patterns) that determines how much compression can be achieved (Section \ref{redundancy_frequency_size}).

\item Given a body of (binary) data that has been `fully' compressed (so that it may be regarded as random or nearly so, as described in Section \ref{ait_section}), its absolute probability may be calculated as $p_{ABS} = 2^{-L}$, where $L$ is the length (in bits) of the compressed data.

\end{itemize}

Probability and information compression may be regarded as two sides of the same coin. That said, they provide different perspectives on a range of problems and, in this work, I have found the information compression perspective---with redundancy-as-repetition-of-patterns---to be more fruitful than viewing the same problems through the lens of probability.%

\subsection{Grammatical inference and generalisation}\label{grammatical_inference_and_generalisation}
\index{grammatical inference|(}\index{grammar!generalisation|(}

An important example of inductive inference is the kinds of predictions that are implicit in a grammar that has been inferred from a finite sample (`corpus') of some language but that is more general than the original sample. It is abundantly clear that this kind of generalisation is involved when we learn our native language or languages. As Chomsky \citeyearpar{chomsky_1965} and others have argued cogently, an adult's knowledge of his or her native language is much more general than the large but finite sample that he or she has heard during childhood. From an early age children show signs of creating general rules such `Add {\em ed} to a verb to give it a past tense' or `Add {\em s} to a noun to make it plural' and, in the early stages, they often overgeneralise such rules and say such things as ``Mummy buyed it'', ``There are some sheeps'' and so on. Interestingly, they learn to correct such overgeneralisations, apparently without the need for explicit error correction by adults or older children (see Section \ref{psychology_language_learning}).

\begin{figure}[!hbt]
\centering
\includegraphics[width=0.9\textwidth]{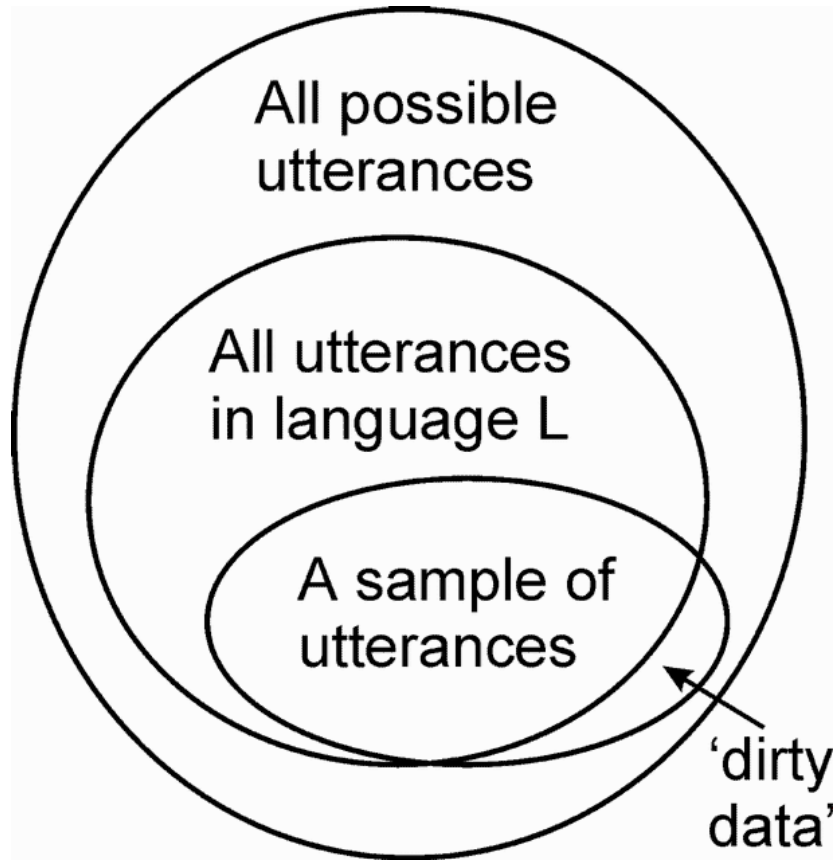}
\caption{Categories of utterances involved in language learning, as described in the text.}
\label{generalisation_figure}
\end{figure}

The learning problem may be represented schematically as shown in Figure \ref{generalisation_figure}. The smallest envelope shows the set of `utterances' that constitute the finite sample of utterances from which a grammar is to be inferred. The middle-sized envelope represents the (infinite) set of utterances in the language being learned. And the largest envelope represents the (infinite) set of all possible utterances. Children also have to cope with `dirty data' meaning `wrong' utterances that are part of the sample from which they learn (see Section \ref{psychology_language_learning}).

The generalisation problem is how, without negative examples or correction by a `teacher' or grading of language samples ({\em cf.} \citet{gold_1967}), to infer a grammar that generates utterances in the middle envelope without creating a grammar that overgeneralises by generating utterances that are outside that envelope.

Solomonoff's solution is very neat: minimise $(G + E)$ as described in Section \ref{mle_section} and then discard the encoding of the sample corresponding to $E$. Minimising $(G + E)$ achieves lossless compression of the sample (as described above) but the grammar itself, without the encoding of the sample, represents a {\em lossy} compression of the sample and, correspondingly, a generalisation beyond the sample. This compromise between the {\em ad hoc} grammar and the promiscuous grammar seems to represent a theoretical ideal in inductive inference.

The question that naturally arises is whether or not this principle applies to human learning and, in particular, the way children learn to talk. There is good evidence that it does (see Section \ref{psychology_language_learning} and \citet{wolff_1988}) but it remains an open question. It seems likely that motivation and emotion have a bearing on what we learn and this may lead us to depart from Solomonoff's ideal (see Section \ref{motivation_and_emotion_in_learning}).

\index{grammatical inference|)}\index{grammar!generalisation|)}
\index{information!theory|)}\index{information!compression|)}\index{information!redundancy|)}\index{probability|)}\index{reasoning!inductive|)}

\section{Coding for reduced redundancy in computing and cognition}

In this section, I describe a variety of observations and ideas from the fields of artificial computing and natural cognition that can plausibly be seen as examples of information compression. Concepts from cognate fields such as theoretical linguistics are included in the discussion.

Earlier examples relate mainly to the brain and nervous system while later examples come mainly from computing. But the separation is not rigid because similar ideas appear in both areas.

Useful reviews of some of the evidence relating information compression to human perception and learning may be found in \citep{chater_1999, chater_1996}.

\subsection{Adaptation and inhibition in the nervous system}\label{adaptation_and_inhibition}

\index{neural!adaptation|(}\index{neural!inhibition|(}

A familiar observation is that we are more sensitive to changes in stimulation than to constant stimulation. We notice a sound which is new in our environment---e.g., the hum of a motor when it is first switched on---but then, as the sound continues, we adapt and cease being aware of it. Later, when the motor is switched off, we notice the change and are conscious of the new quietness for a while until we adapt again and stop giving it attention.

This kind of adaptation at the level of our conscious awareness can be seen also at a much lower level in the way individual nerve cells respond to stimulation. The two studies to be described are of nerve cells in a horseshoe crab ({\em Limulus}) but the kinds of effects that have been observed in this creature have also been observed in single neuron studies of mammals and appear to be widespread amongst many kinds of animal, including humans. There are more complex modes of responding but their existence does not invalidate the general proposition that nervous tissue is relatively sensitive to changes in stimulation and is relatively insensitive to constant stimulation.

\begin{figure}[!hbt]
\centering
\includegraphics[width=0.9\textwidth]{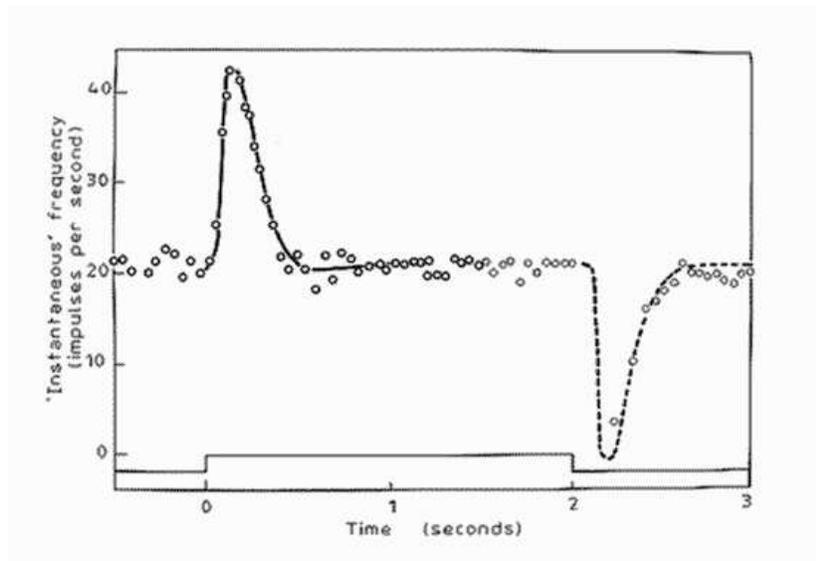}
\caption{Variation in the rate of firing of a single ommatidium of Limulus in response to changing levels of illumination \citep[][, p. 118]{ratliff_etal_1963}.}
\label{ommatidium_firing_figure_1}
\end{figure}

Figure \ref{ommatidium_firing_figure_1} shows how the rate of firing of a single receptor ({\em ommatidium}) in the eye of Limulus changes with the onset and offset of a light \citep{ratliff_etal_1963}. The receptor responds with a burst of spike potentials when the light is first switched on. Although the light stays bright for some time, the rate of firing quickly settles down to a background rate. When the light is switched off, there is a brief dip in the rate of firing followed by a resumption of the background rate.

\begin{figure}[!hbt]
\centering
\includegraphics[width=0.9\textwidth]{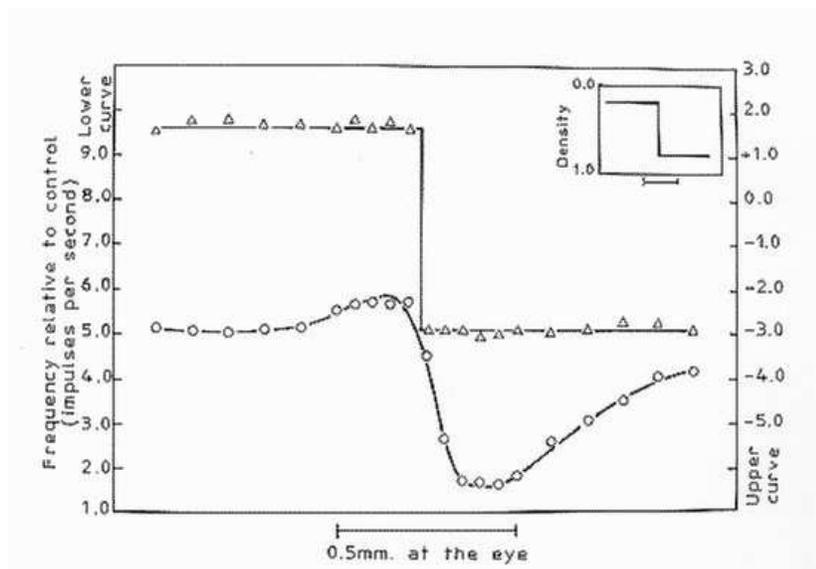}
\caption{Two sets of recordings from a single ommatidium of Limulus \citep[][p. 1248]{ratliff_hartline_1959}.}
\label{ommatidium_firing_figure_2}
\end{figure}

This relative sensitivity to changes in stimulation and relative insensitivity to constant stimulation can be seen also in the spatial dimension. Figure \ref{ommatidium_firing_figure_2} shows two sets of recordings from a single ommatidium of Limulus \citep{ratliff_hartline_1959}. In both sets of recordings, the eye of the crab was illuminated in a rectangular area bordered by a dark rectangle of the same size. In both cases, successive recordings were taken with the pair of rectangles in successive positions across the eye along a line which is at right angles to the boundary between light and bright areas. This achieves the same effect as---but is easier to implement than---keeping the two rectangles in one position and taking recordings from a range of receptors across the bright and dark areas.

In the top set of recordings (triangles) all the ommatidia except the one from which recordings were being taken were masked from receiving any light. In this case, the target receptor responds with frequent impulses when the light is bright and at a sharply lower rate when the light is dark.

In the bottom set of recordings (circles) the mask was removed so that all the ommatidia were exposed to the pattern of bright and dark rectangles. In this situation, the target receptor fires at or near a background rate in areas which are evenly illuminated (either bright or dark) but, near the border between bright and dark areas, positive and negative responses are exaggerated. In the spatial dimension, as with the temporal dimension, changes in stimulation are more significant than constant stimulation.

The kinds of effects, just described, that have been recorded in Limulus have also been observed in the retina and other parts of the visual pathway in mammals \citep[see, for example,][]{nicholls_etal_2001}. Similar effects occur in sensory nerves associated with the skin and the Basilar membrane \citep{von_bekesy_1967}.

It is now widely recognised that this sensitivity to temporal and spatial discontinuities in stimulation is due to the action of inhibitory pathways in the nervous system which counteract the excitation of nerve cells. The way in which `lateral inhibition' and other kinds of inhibition may explain the observations which have just been described and a range of other phenomena (Mach bands, simultaneous contrast, motion sensitivity) is well described and discussed by \citet{von_bekesy_1967} (see also \citet[pp. 65--75]{lindsay_norman_1972}).\footnote{In the eye of the horseshoe crab {\em Limulus}, each receptor cell has an excitatory connection to a ganglion cell which then transmits to the brain. In itself, this arrangement of nerve cells and fibres would mean that a pattern of bright and dark illumination on the eye would be mirrored by a corresponding pattern of excitation in the ganglion cells. However, each receptor cell has an inhibitory connection to one or more of the neighboring ganglion cells. This means that in an area of the eye that is brightly illuminated, the excitation of each ganglion cell is largely cancelled by the inhibitory signals from neighbouring receptor cells, bringing the overall level of excitation down to much the same `background' rate of firing as ganglion cells in the dark regions. However, ganglion cells around the boundary of each bright region receive relatively few inhibitory signals from the neigbouring dark region which means their level of excitation is higher. And the rate of firing of ganglion cells around the boundary of each dark region is lower than those in the body of the dark region due to the inhibitory effect of receptors in the neighbouring bright region. The overall effect is to emphasise boundaries between bright and dark regions, with only a `background' rate of firing in areas that are uniformly illuminated (either light or dark). Similar mechanisms operate in the mammalian retina.}

It has also been recognised for some time that these phenomena may be understood in terms of principles of economy in neural functioning \citep{barlow_1959, barlow_1969, von_bekesy_1967}. In the terms discussed earlier, adaptation and inhibition in nervous functioning have an effect which is similar to that of the run-length coding technique for data compression. Economy is achieved in all these cases by coding changes in information and reducing or eliminating the redundancy represented by sequences or areas that are uniform \citep{barlow_1959, barlow_1969, von_bekesy_1967} (but see \citet[p. 245]{barlow_2001_network} for further thoughts on that idea).%
\index{neural!adaptation|)}\index{neural!inhibition|)}

\subsubsection{Perceptual constancies}\label{perceptual_constancies}

\index{perception!constancies|(}

It has been recognised for a long time that, in our perceptions of the world, we can abstract invariance or `constancies' from the variability of our sensory inputs:

\begin{itemize}

\item {\em Size constancy}. We can recognise an object despite wide variations in the size of its image on the retina---and we judge its size to be constant despite these variations.

\item {\em Brightness constancy}. We can recognise something despite wide variations in the absolute brightness of the image on our retina (and, likewise, we judge its intrinsic brightness to be constant).

\item {\em Colour constancy}. In recognising the intrinsic colour of an object, we can make allowances for wide variations in the colour of the light that falls on the object and the consequent effect on the colour of the light that leaves the object and enters our eyes.

\end{itemize}

By concentrating on aspects of the world that remain constant and discarding the variations that surround them, we can drastically reduce the amount of information that we need to deal with.%
\index{perception!constancies|)}

\subsubsection{`Neural' computing}

\index{neural!network|(}

The idea that principles of economy may apply to neural functioning is recognised in some of the theory associated with artificial `neural' computing, e.g., `Hopfield nets' \citep{hopfield_1982} and `simulated annealing' \citep{hinton_sejnowski_1986}. But with some notable exceptions \citep[e.g.,][]{mahowald_mead_1991}, there seems to have been little attempt in the development of artificial neural networks to exploit the kinds of mechanisms that real nervous systems apparently use to achieve information compression.\index{neural!network|)}

\subsection{Objects and classes in perception and cognition}\label{oo_in_perception_and_cognition}

\index{object|(}\index{class!hierarchy|(}

\subsubsection{Seeing the world as objects}\label{objects_section}

Some things in our everyday experiences are so familiar that we often do not realise how remarkable they are. One of these is the automatic and unconscious way we see the world as composed of discrete objects. Imagine a film taken with a fixed camera of a tennis ball crossing the field of view. Successive frames show the ball in a sequence of positions across a constant background. Taken together, these frames contain a very large proportion of redundancy: the background repeats in every frame (apart from that part of the background that is hidden behind the ball) and the ball repeats in every frame (let's assume that it is not spinning). Uniform areas within each frame also represent redundancy.

Any simple record of this information---on cinema film or digitised images---is insensitive to the redundancy between frames or within frames and has no concept of `ball' or `background'. But people automatically collapse the several images of the ball into one coherent concept and, likewise, see the background as the `same' throughout the sequence of frames.

This is a remarkable piece of information compression, especially since it is performed in real time by nerve cells that are, by electronic standards, exceedingly slow. On this last point, Mahowald and Mead's \citeyearpar{mahowald_mead_1991} work throws useful light on how this kind of compression may be achieved in a simulated mammalian retina and how the necessary speed can be achieved with slow components. In keeping with what was said earlier about neural functioning, inhibitory pathways and processes are significant.

Of course, the concepts we have of real-world objects are normally more complicated than this example suggests. The appearance of a typical object varies significantly depending on orientation or view point. In cases like this, each concept that we form must accommodate several distinct but related views. What we normally think of as a unitary entity should, perhaps, be regarded more accurately as a class of inter-related snapshots or views (more about classes below).

Notwithstanding these complexities, our everyday notion of an object is similar to the previously-described concept of a chunk. Like chunks, objects often have names or tags but again, as with chunks, not every object has a name. An object (with or without a name) may, like a chunk, be seen as the product of processes for extracting redundancy from information.

\subsubsection{Stereoscopic vision and random dot stereograms}\label{stereograms_section_1}

\index{perception!vision|(}\index{stereogram|(}

``In an animal in which the visual fields of the two eyes overlap extensively, as in the cat, monkey, and man, one obvious type of redundancy in the messages reaching the brain is the very nearly exact reduplication of one eye's message by the other eye.'' \citep[][p. 213]{barlow_1969}.

\begin{figure}[!hbt]
\centering
\includegraphics[width=0.9\textwidth]{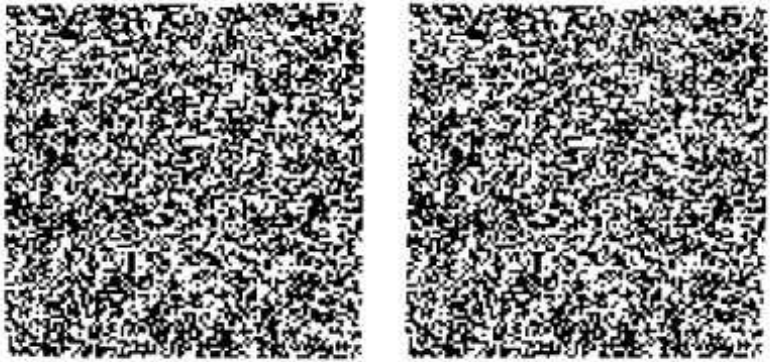}
\caption{A random-dot stereogram \citep[from][p. 21]{julesz_1971}.}
\label{stereogram_figure_1}
\end{figure}

\begin{figure}[!hbt]
\centering
\includegraphics[width=0.9\textwidth]{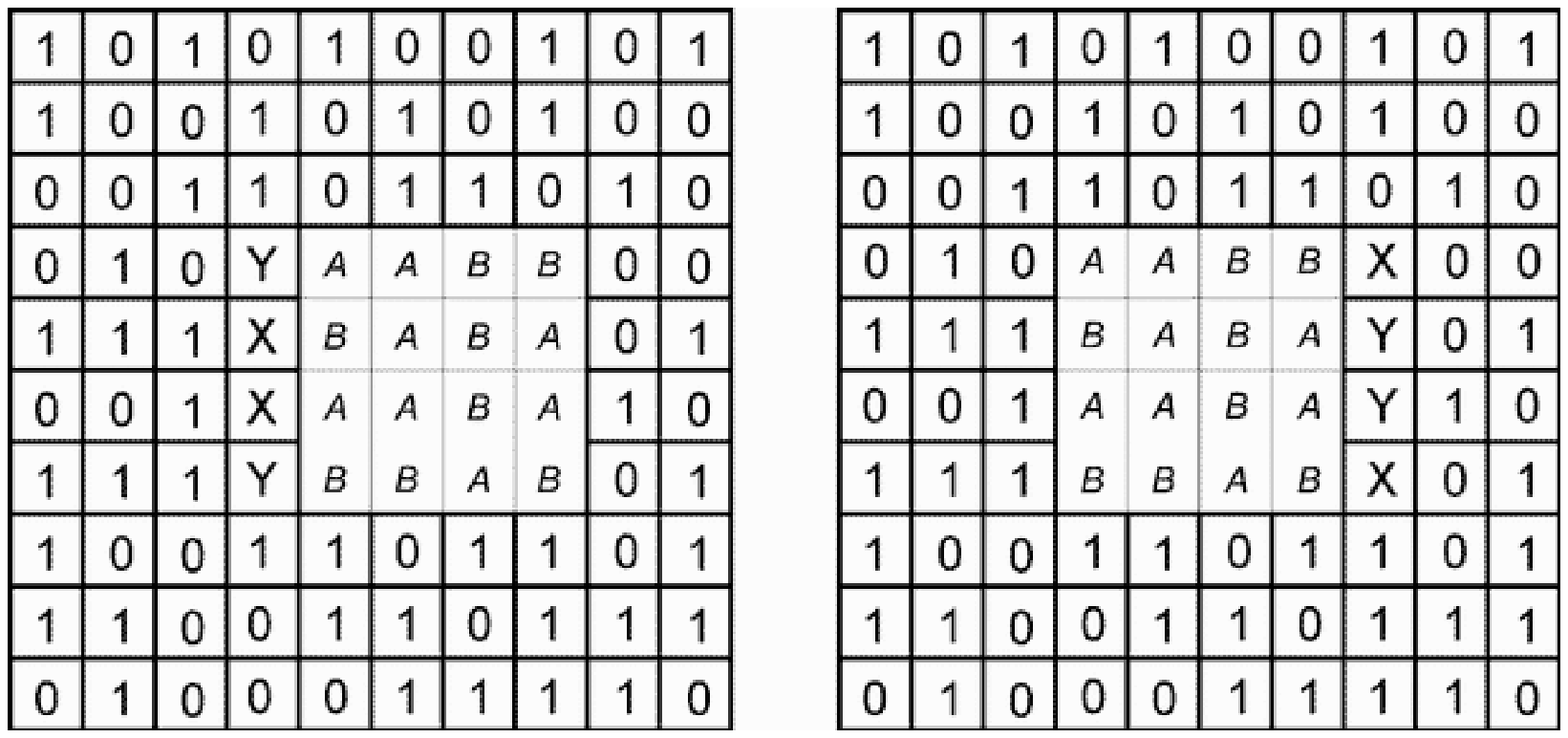}
\caption{Diagram to show the relationship between the left and right images in Figure \ref{stereogram_figure_1} \citep[based on][p. 21]{julesz_1971}.}
\label{stereogram_figure_2}
\end{figure}

Stereoscopic vision and the phenomenon of random dot stereograms provides further evidence of information compression by our nervous systems. It also provides a striking illustration of the connection between redundancy extraction and our tendency to see the world as discrete objects.
Each of the two images in Figure \ref{stereogram_figure_1} is a random pattern of black and white pixels. Each image, in itself, contains little or no redundancy. But when the two images are taken together, there is substantial redundancy because they have been designed so that they are almost, but not quite, the same. The difference is that a square area in the left image has been shifted a few pixels to the right compared with a corresponding square area in the right image, as is illustrated in Figure \ref{stereogram_figure_2}.

When the images are viewed through a stereoscope---so that the left image is seen by the left eye and the right image by the right eye---one's brain fuses the two images so that the two areas around the squares are seen as the `same' and the two square areas are merged into a single square object which appears to stand out vividly in front of its background.

A similar effect can be achieved without the need for a stereoscope by combining the two images in the kind of `magic eye' picture that was popular a few years ago

Random dot stereograms are normally used to illustrate and study human abilities to perceive depth using stereoscopic vision. But they are also good examples of our ability to extract redundancy from information by merging matching patterns.

In Figure \ref{stereogram_figure_1}, the central square, and its background, are chunks in the sense described earlier, each of which owes its perceptual existence to the merging of matching patterns. It is the redundancy that exists between the two images, coupled with our ability to find it, that gives coherence to the objects that we see. The vivid boundary that we can see between the square and its background is the product of search processes that successfully find the maximum possible unification between the two images.

In everyday vision (e.g., the tennis ball example discussed above), recognition of an object may owe something to redundancy within each frame. Since each of the two images in a random dot stereogram contains little or no redundancy, our ability to see coherent objects in such stereograms demonstrates that we do not depend on redundancy within each image but can derive an object concept exclusively from redundancy between images.%
\index{perception!vision|)}\index{stereogram|)}

\subsubsection{Classes, sub-classes and inheritance}\label{classes_subclasses_inheritance}

\index{inheritance of attributes|(}

It is commonly recognised that objects may be grouped into classes and classes may themselves be grouped into higher level classes. Cross-classification with overlapping classes (e.g., `woman' and `doctor') is also seen.

A class may be defined extensionally by listing its constituent objects or classes. It may also be defined intensionally in terms of attributes which members of the class have in common. Although many commonly-used classes are `polythetic'---no single attribute need necessarily be shared by all members of the class and any given attribute may be associated with more than one class---it is similarity amongst members of a class---some degree of sharing of attributes---that gives `natural' classes their coherence.

Grouping things by their similarity gives us a means of compressing the information which describes them. An attribute which is shared by all members of a class (or, in the case of polythesis, a set of alternative attributes which is shared by members of a class) need be recorded only once and not repeated for every member. In the jargon of object-oriented design, each member of a class `inherits' the attributes of the class. Those shared attributes constitute a `schema' in the sense discussed earlier, that may be `corrected' for each individual member by the addition of more specific information about that member.

The widespread use of classes and subclasses in our thinking and in language, coupled with their obvious value in compressing information, strongly suggests that we do store classes and attributes in this way. But it is difficult to obtain direct confirmation of this idea. Attempts to verify the idea experimentally \citep[e.g.,][]{collins_quillian_1969, collins_quillian_1972} have proved inconclusive. This is probably more to do with the difficulties of making valid inferences from experimental studies of human cognition than any intrinsic defect in the idea.%
\index{object|)}\index{class!hierarchy|)}\index{inheritance of attributes|)}

\subsection{Natural languages}\label{compression_in_nl}

\index{language!compression in|(}

Samples of natural language---English, French etc---are normally about 50\% redundant \citep{miller_friedman_1958}. This redundancy often serves a useful purpose in helping listeners or readers to correct errors and to compensate for noise in communication---and this is almost certainly the reason why natural languages have developed in this way.

Despite the existence of redundancy in natural languages, they provide a further example of economical coding in cognition. Every `content' word in a natural language (e.g., noun, verb, adjective or adverb) may be regarded as a reference to the relatively complex chunk of information that is the meaning of the word.

Without a convenient brief label like `table', the concept of `a horizontal platform with four, sometimes three, vertical supports, normally about three feet high, normally used for ...' would have to be long-windedly repeated in every relevant context rather like the slow language of the Ents in Tolkien's {\em The Lord of the Rings}. A sentence is normally a highly `coded' and compressed representation of its meanings.

An even more obvious example of identifiers and references in natural language is the use of citations, as described in connection with the chunking-with-codes technique in Section \ref{techniques_for_ic}. Each citation is a reference to the bibliographic details of a given book or article and these details may themselves be seen as a code for that publication. Like other kinds of reference, citations can circumvent the need to repeat information redundantly in two or more contexts. But it is sometimes convenient to use this device for reasons of consistency and style when a given citation appears only once in a given book or article.

Before leaving this section on natural language, it is relevant to comment on Zipf's extensive studies of the distribution of words in natural languages \citep{zipf_1935} and the {\em Principle of Least Effort} \citep{zipf_1949} that he proposed to explain these observations and others. Zipf's arguments are interesting and quite persuasive but, as \citet{mandelbrot_1957} and others have pointed out, the phenomena described by `Zipf's law' could be due to nothing more profound than a random process for creating words and the boundaries between words in natural languages. However, in George Miller's words \citeyearpar{miller_1965}, ``It is impossible to believe that nothing more is at work to guide our choice of letter sequences than whatever random processes might control a monkey's choice, or that the highly plausible arguments Zipf puts forward have not relevance at all.'' (p. vii). The jury is still out!%
\index{language!compression in|)}

\subsection{Grammars}\label{grammar_section}

\index{grammar|(}

As indicated in Section \ref{mle_section}, a grammar provides a means of encoding data in an economical form. More accurately, the notational conventions that are used in grammars may be regarded as a set of devices that may be used to encode information in an economical form. They are not necessarily used to good effect in any one grammar.

Consider a grammar\index{grammar!phrase-structure ---} like the one shown in Figure \ref{cf_psg_figure}. Each rule in the grammar may be seen as a `chunk' of information as described in Section \ref{ic_repetition_of_patterns} and the symbols to the left of the re-write arrows may be seen as identifiers, as described in Section \ref{identifiers_and_references}.

\begin{figure}[!hbt]
\centering
\begin{tabular}{l}
S $\rightarrow$ NP V ADV \\
NP $\rightarrow$ D N \\
D $\rightarrow$ t h i s \\
D $\rightarrow$ s o m e \\
N $\rightarrow$ t r a i n \\
N $\rightarrow$ a e r o p l a n e \\
V $\rightarrow$ g o e s \\
V $\rightarrow$ t r a v e l s \\
ADV $\rightarrow$ s l o w l y \\
ADV $\rightarrow$ f a s t \\
\end{tabular}
\caption{A context-free phrase-structure grammar.}
\label{cf_psg_figure}
\end{figure}

Notice that each of the symbols `D', `N', `V' and `ADV' do not identify any one chunk uniquely and so, in themselves, they can only achieve lossy compression. However, with the addition of symbols to identify each rule uniquely, a grammar like the one shown can be used for lossless compression of appropriate data. Details of how this kind of encoding may be done are described in Section \ref{ma_evaluation}.

\subsubsection{Recursion in grammars and run-length coding}

\index{recursion!in grammars|(}\index{information!compression!run-length coding|(}

Since grammars incorporate the `chunking-with-codes'\index{information!compression!chunking-with-codes} device for information compression, they can express recursion---which is itself a form of run-length coding. The fragment of grammar shown in Figure \ref{cf_psg_recursion_figure} generates phrases like {\em the very very very tall girl}, {\em a very very short boy} etc. Notice that the number of instances of {\em very} in any one phrase is not specified: recursion like this represents lossy compression of any finite set of terminal strings.

\begin{figure}[!hbt]
\centering
\begin{tabular}{l}
NP $\rightarrow$ D V A N \\
D $\rightarrow$ the \\
D $\rightarrow$ a \\
V $\rightarrow$ NULL \\
V $\rightarrow$ very V \\
A $\rightarrow$ tall \\
A $\rightarrow$ short \\
N $\rightarrow$ boy \\
N $\rightarrow$ girl \\
\end{tabular}
\caption{A cf\_psg with recursion.}
\label{cf_psg_recursion_figure}
\end{figure}

\index{recursion!in grammars|)}\index{information!compression!run-length coding|)}

\subsubsection{More powerful grammars}

It has been recognised for some time (and pointed out most notably by \citet{chomsky_1957}) that context-free phrase-structure grammars are not `powerful' enough to represent the structure of natural languages effectively. As shown by the examples just given, context-free phrase-structure grammars can be used to represent simple sub-sets of English in a succinct form. But the full complexity of English or other natural language can only be accommodated by a context-free phrase-structure grammar, if at all, at the cost of large amounts of redundancy in the grammatical description.

The phenomenon of `discontinuous dependencies'\index{grammar!syntactic dependencies} highlights the shortcomings of context-free phrase-structure grammars. In a sentence like {\em The winds from the West are strong}, there is a `number' dependency between {\em winds} and {\em are}: the plural noun must be followed by a plural verb and likewise for singular forms. The dependency is `discontinuous' because it jumps over the intervening structure ({\em from the West} in the example) and this intervening structure can be arbitrarily large.

To represent this kind of dependency with a context-free phrase-structure grammar requires one set of rules for singular sentences and another set of rules for plurals---and the two sets of rules are very similar. The resulting redundancy in the grammar can multiply substantially when other dependencies of this kind are included.

This problem can be largely overcome by using a more `powerful' kind of grammatical system like a definite clause grammar \citep{pereira_warren_1980}\index{grammar!definite clause ---} or a transformational grammar \citep[see, for example,][]{radford_1988}\index{grammar!transformational ---}. This kind of system may be seen as a superset of a context-free phrase-structure grammar with additional mechanisms that, amongst other things, allow discontinuous dependencies to be represented without undue redundancy.

Grammatical systems like these can be seen as variations on the `schema-plus-correction' idea that was described above. They provide the means of representing sentence structure as a `schema' that may, in the example given earlier, be `corrected' by the addition of singular or plural components at appropriate points in the structure.%
\index{grammar|)}

\subsection{Computer programs}\label{computer_programs_section}

\index{programming|(}

Functions in computing and mathematics are often defined `intensionally' in terms of rules or operations required to perform the function. But functions are also defined `extensionally' by specifying one or more outputs for every input or combination of inputs \citep{sudkamp_1988}. This idea applies to functions of various kinds (`total', `partial' and 'multi') and also to `programs' and information systems in general.

Elsewhere \citep{wolff_1994_software} I have discussed how a computer program or mathematical function may be seen as a compressed representation of its inputs and its outputs, how the process of designing programs and functions may be seen to be largely a process of information compression, and how the execution of programs or functions may also be seen in terms of information compression (see also Chapter \ref{maths_logic_chapter}).

In this section, I view computer programs more conventionally as a set of `operations' and discuss how principles of compression may be seen in the way programs are organised. In the same way that the notational conventions used in grammars may be regarded as a means of compressing linguistic information, the conventions used in computer programs may be seen as devices for representing computing operations in a succinct form. As with grammars, the provision of these facilities does not in itself guarantee that they will be used to best effect.

Compression techniques may be seen in `functional', `structured' and `logic' programming and, {\em a fortiori} in `object-oriented' programming.

\subsubsection{Functional and structured programming}\label{functional_and_structured_programming}

\index{programming!functional|(}\index{programming!structured|(}

{\em Chunking-with-codes}\index{information!compression!chunking-with-codes}. If a set of statements is repeated in two or more parts of a program then it is natural to declare them once as a `function', `procedure' or `sub-routine' within the program and to replace each sequence with a `call' to the function from each part of the program where the sequence occurred. This is an example of compression: the function may be regarded as a chunk and the name of the function is its identifier.

Whether or not a programmer chooses to create a function in a situation like this---or to leave repeated sequences as `macros'---depends, amongst other things, on whether the sequence is big enough to justify the `cost' of giving it a name and whether the run-time overhead which is typically associated with the use of functions in conventional computers is acceptable for the application in hand.

{\em Run-length coding}\index{information!compression!run-length coding}. If a body of code is repeated in one location within the program then it may be declared as a function or marked as a `block' and the fact of repetition may be marked with one of the familiar conventions for showing iteration: {\em repeat ... until}, {\em while ... do}, {\em for ... do}. Each of these is a form of run-length coding.

Recursion, which is available in most procedural programming languages, is another means of showing repetition of program operations. It is essentially the same as recursion in grammars.

{\em Schema-plus-correction}\index{information!compression!schema-plus-correction}. It often happens in software design that two or more sets of statements are similar but not identical. In these cases, it is natural to merge the parts that are the same and to provide conditional statements ({\em if ... then ... else} statements or case statements) to select alternative `paths' within the software. The complete set of statements, including the conditional statements, may be regarded as a `schema' describing a set of behaviours for the program, much as a grammar describes a set of terminal strings.

To specify a particular path through the program, it is necessary to supply the information needed in the conditional statements so that all the relevant choices can be made. This additional information is the `correction' to the schema represented by the program code. If the program statements have been encapsulated in a function then these corrections to the program schema will normally be supplied as arguments or parameters to the function.%
\index{programming!functional|)}\index{programming!structured|)}

\subsubsection{Logic programming}\label{logic_programming_section}

\index{programming!logic|(}

Similar ideas appear in logic programming languages like Prolog. For example, the chunking-with-codes idea can be seen in the structure of a Prolog Horn clause. The predicate in the head of the clause may be seen as an identifier while the body of the clause may be seen as the chunk that it labels. Copies of the identifier that appear in other Horn clauses may be seen as references to the given clause.

As with functional and structured programs, a Prolog program may be seen as a schema that represents the set of possible behaviours of the program. The information supplied in a Prolog query serves as a correction to the schema which reduces the range of possible behaviours of the program.

Repetition in Prolog programs is coded using recursion.%
\index{programming!logic|)}

\subsubsection{Object-oriented programming}\label{oop_section}

\index{programming!object-oriented|(}

Object-oriented programming, as it was originated in Simula and has been developed in Smalltalk, C++ and several other languages, embraces the kinds of mechanisms just described but includes an additional set of ideas that may also be understood in terms of information compression.

One important idea in object-oriented programming, that was introduced in Simula, is that there should be a one-for-one correspondence between objects in the real world and `objects' in the software. For example, an object-oriented program for managing a warehouse will have a software object for each person employed in the warehouse, a software object for each shelf or bay, a software object for each item stored, and so on. In object-oriented terms, a software object is a discrete piece of program: procedural code and associated data structures, or either one of these without the other. For example, an object for a `person' would be a program that can respond to `messages' such as a request for the person's name, an instruction to move an item from one location to another, and so on.

Superficially, this is an extravagant---and redundant---way to design software because it seems to mean that the same code is repeated in every object of a given type. But, of course, this is not how things are done. As with `real world' objects, economy can be achieved by the use of classes and sub-classes.

In object-oriented programming languages, every object belongs to a class (in some object-oriented languages an object can be assigned to more than one class), the code for that class is stored only once within the program and it is inherited by each instance of the class. There can be a hierarchy of two or more classes with inheritance of code from any level down to individual objects.

As with individual objects, a recognised principle of object-oriented programming is that the classes that are defined in an object-oriented program should correspond, one for one, with the classes that we recognise in the real world. In the warehouse example, there would be a class for `person', a class for `shelf' and a class for `item'. Each class---`person', for example---may be divided into sub-classes like `manager', `foreman', `operative' etc.

The ideas that have been described---software objects, classes and inheritance---are further examples of the way in which information compression pervades the organization of computer programs:

\begin{itemize}

\item As discussed earlier, objects and classes as we see them in the real world may be understood in terms of our subjective compression of perceptual information received from the world. By using similar devices in software design we can make the structure of software reflect patterns of redundancy in the real world which our brains are apparently so efficient at exploiting.

\item Within the program code itself, class hierarchies with inheritance of code are powerful mechanisms for information compression. Instances of a class may be multiplied without creating any corresponding redundancy because all the data structures and procedural code which they have in common can be inherited from the class definition. Likewise, sets of classes which have common attributes may inherit these attributes from a higher level class.

\item As was mentioned in connection with natural classes, the mechanisms of class hierarchies with inheritance may be seen as an example of the schema-plus-correction technique for information compression: a class definition is a `schema' and information which is supplied when a new instance of the class is created (e.g., the name of that object) is a `correction' to the schema. In a similar way, a high level class may be seen as a relatively abstract schema which is refined or `corrected' by the more specific information contained in lower level classes.

\end{itemize}

It is perhaps pertinent to comment, in passing, on the advantages of designing software in this way:

\begin{itemize}

\item One advantage is psychological: software that reflects the structure of our established concepts is easier to understand than software that does not.

\item A more subtle, but nonetheless important, advantage is that software designed in this way will normally be easier to modify than otherwise: the fact that the structures reflected in the software are persistent (repeating) patterns in the world means that new versions of software are more likely to be refinements or rearrangements of already established objects and classes than radical reorganizations of the code.

\item There is a third advantage for software designers in using these mechanisms for information compression: if any given piece of information is recorded only once within a program then any change to that information needs be made only once and there is no need to check that it is correctly repeated in other parts of the program.

\end{itemize}

\index{programming|)}\index{programming!object-oriented|)}

\subsection{Other aspects of computing}

We have already seen several uses for identifiers and references in computing---as the names of functions, procedures or sub-routines, as the names for object-oriented programming objects and classes of objects and as names for rules in grammars. Some other examples of codes in computing include:

\begin{itemize}

\item Names of tables or records in databases.

\item Names of fields in tuples or records.

\item Names of files.

\item Names of variables, arrays and other data structures used in computer programs.

\item Labels for program statements (for use with the now shunned `go to' statements).

\end{itemize}

The names of directories (folders) in Unix, MS-DOS and similar operating systems are also identifiers in the sense of this chapter. But an hierarchical directory structure may also be seen as a simple form of class hierarchy and, as such, it may be seen as a restricted form of schema plus correction. The name of a directory is a name that applies to all the files and sub-directories within that directory and to all files and sub-directories at lower levels. Rather than redundantly repeat the name for every file and sub-directory to which it applies, the name is recorded once and, in effect, `inherited' by all the objects at lower levels.

\subsection{The uses of redundancy in computing and cognition}\label{uses_of_redundancy}

\index{information!redundancy|(}

Although information compression has an undoubted r{\^o}le in computing, it is also true that redundancy can be and often is exploited to reduce errors in processing, to speed up processing and to guard against the risk of losing information. It seems likely that the same can be said of brains and nervous systems although this aspect of their workings seems not to have received much attention.

Any user of a computer system knows that it is very unwise to run the system without keeping backup copies of critical information to reduce the risk of catastrophic loss of that information. Since backup copies repeat the original information, they represent redundancy in the system. Redundancy in the form of mirror web sites or replicated databases is also exploited to speed up processing in systems that are distributed across a network. And redundancy provides the key to all computational techniques for reducing errors in the processing or transmission of information.

Despite the apparent conflict between the idea that information compression has an important r{\^o}le in information processing and the uses of redundancy for the purposes just described, they are really independent and can be reconciled very easily. For example, it is entirely possible for a database to be designed to minimise internal redundancies and, at the same time, for redundancies to be used in backup copies or mirror copies of the database or in error-correction techniques when the database is transmitted across a network. Paradoxical as it may sound, knowledge can be compressed and redundant at the same time.%
\index{information!redundancy|)}

\section{Searching for redundancy in computing and cognition}

So far, we have reviewed a variety of ways in which redundancy extraction appears in computing and cognition, focusing mainly on the ways in which coding techniques relate to these two fields. Now we shall look at the dynamic side of the coin---the things in computing and cognition that may be understood as searching for the redundancy which may then be removed by the use of coding techniques.

As we have seen, searching for redundancy can be understood as a search for patterns that match each other and, as we have seen (Section \ref{matching_searching_and_constraints}), there is normally a need to constrain the search by restricting the search space or guiding the search using some measure of redundancy, or both. The sections which follow describe some of the ways in which this kind of search appears in computing and cognition. As with the section on coding, earlier examples are mainly from natural information processing with later examples from computing.

\subsection{Random dot stereograms}\label{stereogram_section_2}

\index{stereogram|(}

A particularly clear example of this search problem is what the brain has to do to enable one to see the figure in the kinds of random dot stereogram described earlier (Section \ref{stereograms_section_1}).

In this case, assuming the left image has the same number of pixels as the right image, the size of the search space is $P^2$ where $P$ is the number of possible patterns in each image, calculated in the same way as was described in Section \ref{matching_searching_and_constraints}. (The fact that the images are two dimensional needs no special provision because the original formula covers all combinations of atomic symbols.)

For any stereogram with a realistic number of pixels, this space is very large. Even with the very large processing power represented by the $10^{11}$ neurons in the brain, it is inconceivable that this space can be searched in a few seconds and to such good effect without the use of metrics-guided searching of the kind described earlier and probably also with some restriction on what comparisons between patterns will be made.

David Marr \citeyearpar[][Chapter 3]{marr_1982} describes two algorithms that solve this problem. In line with what has just been said, both algorithms rely on constraints on the search space and both may be seen as incremental search guided by redundancy-related metrics.%
\index{stereogram|)}

\subsection{Recognition of objects and patterns}

\index{perception!recognition|(}

As we saw in Section \ref{objects_section}, perceptual objects can be understood as `chunks' that promote economy in the storage and processing of perceptual information. We not only create such chunks out of our perceptual experience but we have a very flexible and efficient ability to recognise objects and other patterns which have already been stored.

How we recognise objects and other patterns is, of course, the subject of much research and is certainly not yet well understood. Amongst the complexities of the subject is the problem, mentioned earlier, of how we can recognise objects despite variations in their appearance due to variations in orientation. Likewise, we can recognise patterns such as handwriting despite many variations of style.

Despite these complexities, it is reasonably clear that, in general terms, the phenomenon of recognition can be understood as information compression \citep{watanabe_article_1972}. Recognition of an object or a pattern means the partial or complete unification of perceptual input with a corresponding pattern already in memory---and thus an overall reduction in the information that they jointly contain.

By assimilating new information to previously-stored information (with or without variations) we can drastically reduce the huge quantities of redundancy that exists in normal sensory inputs. It is hard to imagine how we could survive if we did not merge new percepts with old ones in this kind of way. Shutting one's eyes and opening them again would result in a complete new record of what we had been looking at, without any relation to the previous record. Merging successive percepts is essential if we are to maintain a reasonably coherent view of the world.

\subsection{Objects in motion}

A closely-related idea is that the concept of motion implies compression of information \citep{barlow_1961}. Encoding a sequence of images in terms of entities that move requires much less information than simply storing the raw images. The waterfall illusion shows that one's perception of motion can conflict with one's perception of position and this suggests that there are neural mechanisms dedicated to the perception of motion that are distinct from mechanisms that detect position. There is evidence for such motion detectors in the neurons of the `magnocellular' layers of the lateral geniculate body and in neurons to which they project in areas V$_{1}$, V$_{2}$ and V$_{5}$ of the cortex \citep[][434--435]{nicholls_etal_2001}.%
\index{perception!recognition|)}

\subsection{Grammar discovery, language learning and other kinds of learning}

\index{grammar!induction or discovery|(}\index{learning!language|(}

If, as was suggested earlier, we view a grammar as a compressed version of its terminal strings, then it is natural to see the process of discovering or inferring a grammar from examples as being largely a matter of discovering the redundancy in those examples and building a stochastic grammar in accordance with minimum length encoding principles (Section \ref{mle_section}). In line with these expectations, practical techniques for grammar discovery are largely a search for repeating patterns in data with unification of repeating patterns, the assignment of identifiers and, quite often, the discarding of some portion of the non-redundant information in the samples \citep{wolff_1982, wolff_1988}.

The learning of a first language by children is clearly a richer and more complex process than grammar discovery as it is normally understood. But computer models of language learning that I have developed in earlier research, which are based on principles of information compression, show remarkable correspondences in their learning behaviour to well-documented phenomena in the way children learn a first language \citep{wolff_1988}:

\begin{itemize}

\item The unsupervised induction of grammatical structure, including segmental structures such as words \citep{wolff_1977} and phrases \citep{wolff_1980}, and disjunctive structures such as parts of speech \citep{wolff_1982}.

\item Generalisation of rules and correction of overgeneralisations without external error correction \citep{wolff_1982}.

\item The way the rate of acquisition of words and other structures varies as learning proceeds.

\item The order in which words are acquired.

\item Brown's \citeyearpar{brown_1973} ``Law of Cumulative Complexity''.

\item The S-P/episodic-semantic shift.

\item The learning of semantic structures and their integration with syntax.

\item The word frequency effect.

\end{itemize}

The fact that these phenomena can be seen as emergent properties of models that are largely mechanisms for achieving information compression lends support to the proposition that language learning by children may be understood, in large measure, as information compression.

Other kinds of learning have also been analysed in terms of economical coding. Pioneering work on the learning of classifications \citep{wallace_boulton_1968} has been followed by related work on economical description of data (e.g., \citet{wallace_freeman_1987, rissanen_1987, cheeseman_1990, pednault_1991, forsyth_1992}.%
\index{grammar!induction or discovery|)}\index{learning!language|)}

\subsection{Query-by-example}

\index{information!retrieval|(}\index{query-by-example|(}

This section and those that follow discuss areas of computing where pattern matching and search of the kinds described earlier can be seen. Unification, with corresponding compression of information is less obvious but can still be recognised in most cases.

A commonly-used technique for retrieving records from a database is to provide a query in the form of an incomplete record---an `example' of the kind of complete record which is to be retrieved. This is illustrated in Figure \ref{query_by_example_figure} where the query `example' has the general form of the complete records but contains asterisks (`*') where information is missing. The search mechanisms in the database match the query with records in the database to retrieve the zero or more records which fit.

\begin{figure}[!hbt]
\centering
\begin{BVerbatim}
An `example' used as a query:

     Name: John *
     Address: * New Street *

Some records in a database:

     Name: Susan Smith
     Address: 8 New Street, Chicago

     Name: John Jones
     Address: 4 New Street, Edinburgh

     Name: John Black
     Address: 20 Long Street, London

     Name: David Baker
     Address: 41 Avenue des Pins, Paris

     ...
\end{BVerbatim}
\caption{A database query with records in the same format, illustrating query-by-example.}
\label{query_by_example_figure}
\end{figure}

Most systems of this kind are driven by some kind of redundancy-related metric. In Figure \ref{query_by_example_figure}, there is some degree of matching between the query and the first three of the records shown. But John Jones of 4 New Street, Edinburgh is the preferred choice because it gives a better fit than the other records.

Retrieval of a record may be seen as `unification' between the record and the query and a corresponding extraction of the redundancy between them. However, this compression is normally evanescent, appearing only temporarily on the operator's screen. When a query operation has been completed, the records in the database are normally preserved in their original form, while the unification and the query are normally discarded.%
\index{information!retrieval|)}\index{query-by-example|)}

\subsection{De-referencing of identifiers in computing}

Identifiers, names, labels or tags of the kinds described earlier---names of functions, procedures or sub-routines, names of object-oriented programming objects and classes of objects, names of directories or files etc---have a psychological function providing us with a convenient handle on these objects in our thinking or in talking or writing. But a name is at least as important in computing systems as an aid to finding a particular information object amongst the many objects in a typical system.

Finding an object by means of its name---`de-referencing' the identifier---means searching for a match between the `reference' as it appears without its associated object (e.g., a `call' to a program function) and the same pattern as it appears attached to the object which it identifies (e.g., a function name together with the function declaration).

Finding an object by means of its name is like a simple form of query-by-example. The name by itself is the incomplete record which constitutes the query, while the object to be found, together with its name, corresponds with the complete record which is to be retrieved.

As with query-by-example, there is normally some kind of redundancy-related metric applied to the search. In conventional computing systems, a `full' match (including the termination marker) is accepted while all partial matches are rejected. A character-by-character search algorithm may be seen as a simple form of hill-climbing search. The seemingly `direct' technique of hash coding exploits memory access mechanisms which, as is described below, may also be understood in terms of metrics-guided search.

As with query-by-example, unification and compression are less obvious than pattern matching and search. Certainly, a computer program is not normally modified by de-referencing of the identifiers it contains. Unification and compression are confined to the evanescent data structures created in the course of program execution.

\subsection{Memory access in computing systems}

\index{information!retrieval|(}

The mechanisms for accessing and retrieving information from computer memory, which operate at a `lower' level in most computing systems, may also be seen in similar terms.

Information contained in a computer memory---`data' or statements of a program---can be accessed by sending an `address' from the central processing unit to the computer memory along an address bus. The address is a bit pattern which is `decoded' by logic circuits in the computer memory. These have the effect of directing the pattern to the part of the memory where the required information is stored.

The logic circuits in memory which are used to decode a bit pattern of this kind have the effect of labelling the several parts of memory with their individual addresses. Accessing a part of memory by means of its address may be seen as a process of finding a match between the access pattern and the address as it is represented in computer memory---together with a unification of the two patterns. The search process is driven by a metric which leads to a full match via successively improving partial matches.%
\index{information!retrieval|)}

\subsection{Logical deduction}\label{logical_deduction_section}

\index{reasoning!deductive|(}\index{logic|(}

Logical deduction, as it appears in systems like Prolog, is based on Robinson's \citeyearpar{robinson_1965} `resolution' principle. The key to resolution is `unification' in a sense which  means giving values to the variables in two structures which will make them the same. However, it also embraces the simpler sense of the word as it has been used in this chapter because it includes the comparison or matching of patterns and their effective merging or `unification' to make one. The wide scope of unification (in both senses) is recognised in a useful review by \citet{knight_1989}.

As was the case in query-by-example, de-referencing of identifiers and memory access, systems for resolution theorem proving normally look for a full match between patterns and reject all partial matches. This feature may be seen as a crude but effective way of constraining the search to reduce its computational demands (Section \ref{matching_searching_and_constraints}).

\subsubsection{Modus ponens}

Similar ideas may be seen in more traditional treatments of logic \citep[e.g.,][]{copi_1986}. Consider, for example, the {\em modus ponens} form of logical deduction which may be represented in abstract terms like this:

\begin{enumerate}

\item $p \Rightarrow q$.

\item $p$.

\item $\therefore q$.
\end{enumerate}

\noindent Here is an example in ordinary language:

\begin{enumerate}

\item If today is Tuesday then tomorrow will be Wednesday.

\item Today is Tuesday.

\item Therefore, tomorrow will be Wednesday.

\end{enumerate}

The implication `$p \Rightarrow q$' (today being Tuesday implies that tomorrow will be Wednesday) may be seen as a `pattern', much like a record in a database. The proposition $p$ (`today is Tuesday') may be seen as an incomplete pattern, rather like the kind of database query which was discussed earlier. Logical deduction may be seen as a unification of the incomplete pattern, `$p$', with the larger pattern, `$p \Rightarrow q$', with a consequent `marking' of the conclusion, `$q$' (`tomorrow will be Wednesday').%
\index{reasoning!deductive|)}\index{logic|)}

\subsection{Inductive reasoning and probabilistic inference}

\index{reasoning!inductive|(}\index{reasoning!probabilistic|(}

Logicians and philosophers have traditionally made a sharp distinction between deductive reasoning, where conclusions seem to follow with certainty from the premises, and inductive reasoning, where inferences are uncertain and the apparent clockwork certainty of deduction is missing.

A popular example of inductive reasoning is the way (in low latitudes) we expect the sun to rise in the morning because it has always done so in our experience in the past. Past experience, together with the proposition that it is night time now, are the `premises' which lead us to conclude that the sun is very likely to rise within a few hours. Our expectation is strong but there is always a possibility that we may be proved wrong.

There are countless examples like this where our expectations are governed by experience. We expect buds to `spring' every Spring and leaves to `fall' every Fall. We expect fire where we see smoke. We expect water to freeze at low temperatures. And we expect to be broke after Christmas!

The way we mentally merge the repeating instances of each pattern in our experience may be seen as further evidence of chunking in human cognition and, as such, further evidence of how human cognition is geared to the extraction of redundancy from information.

Like query-by-example, de-referencing of identifiers, memory access and logical deduction, inductive reasoning may be seen as the matching and unification of a pattern with a larger pattern of which it is a part. Our experience of day following night is a sequential pattern: `night sunrise'. The observation that it is night time now is another pattern: `night'. The second pattern will unify with the first part of the first pattern and, in effect, `mark' the remainder of that pattern as an expectation or conclusion.

As was noted earlier, there is a significant body of work on economical description and its connection with probabilistic inference. But the connection between these two topics and the topics of pattern matching and the unification of patterns is not yet properly recognised.%
\index{reasoning!probabilistic|)}

\subsubsection{The possible integration of deductive and inductive reasoning}

\index{reasoning!deductive|(}

That deductive and inductive reasoning may both be seen as the matching and unification of a pattern with a larger pattern that contains it suggests that they are not as distinct as has traditionally been thought. The example of {\em modus ponens} deductive reasoning given earlier may also be seen as inductive reasoning: the way Wednesday always follows Tuesday is a (frequently repeated) pattern in our experience and the observation that today is Tuesday leads to an inductive expectation that tomorrow will be Wednesday.%
\index{reasoning!deductive|)}\index{reasoning!inductive|)}

\subsection{Normalisation of databases}

\index{database!normalisation|(}

The process of `normalisation' which is applied in the design of relational databases is essentially a process of removing redundancy from the framework of tables and columns in the database \citep{date_1986}.

If, for example, a database contains tables for `buildings' and for `sites' both containing columns such as `location', `value', `date-acquired', `date-of-disposal', then these columns (perhaps with others) may be extracted and placed in a table for `property'. With suitable provision for linkage between tables, this information may be `inherited' by the tables for `buildings' and `sites' (and other sub-classes of property), much in the manner of object-oriented design.%
\index{database!normalisation|)}

\subsection{Parsing}

\index{parsing!language|(}

The process of parsing a string---the kind of analysis of a computer program which is done by the front end of a compiler, or the syntactic analysis of a natural language text---is a process of relating a grammar to the text which is being analysed. To a large extent, this means searching for a match between each of the terminal elements in the grammar and corresponding patterns in the text, and the unification of patterns which are the same. It also means de-referencing of the identifiers of the rules in the grammar which, as was discussed above, also means matching, unification and search.

That parsing must be guided by some kind of redundancy-related metric is most apparent with the ambiguous grammars and relatively sophisticated parsers associated with natural language processing \citep[see, for example,][]{spark-jones_wilkes_1985}. In these cases, alternative analyses may be graded according to how well the grammar fits the text. But even with the supposedly unambiguous grammars and simple parsing techniques associated with computer languages, there is an implicit metric in the distinction between a `successful', complete parse of the text and all failed alternatives---much like the all-or-nothing way in which identifiers are normally recognised in computing.%
\index{parsing!language|)}

\section{Conclusion}

In this chapter I have tried to show that information compression is a pervasive feature of information systems, both natural and artificial. But, even if this is accepted, it is still pertinent to ask whether the observation is significant or merely an incidental feature of these systems?

\subsection{The significance of information compression}

Since natural and artificial mechanisms for the storage and processing of information are not `free', we should expect to find principles of economy at work to maximise the ratio of benefits to costs. It may be argued that information compression in information systems is nothing more than a reasonable response to the need to get the most out of these systems.

Other considerations suggest that information compression is much more significant than that:

\begin{itemize}

\item We gather information and store it in brains and computers because we expect it to be useful to us in the future. Natural and artificial information processing is founded on the inductive reasoning principle that the past is a guide to the future.

\item But stored information is only useful if new information can be related to it. To make use of stored patterns of information it must be possible to recognize these patterns in new information. Recognition means matching stored patterns with new information and unification of patterns which are the same. And this means information compression in the ways that have been discussed.

\item The inductive principle that the past is a guide to the future may be stated more precisely as the expectation that patterns which have occurred relatively frequently in the past will tend to recur in the future. But `relatively frequent repetition of patterns' means redundancy! It is only possible to see a pattern as having repeated relatively frequently in the past by the implicit unification of its several instances---and this means extraction of redundancy and compression of information in the ways that have been discussed.

\end{itemize}

These arguments point to the conclusion that information compression is not an incidental feature of information systems. It is intimately related to the principle of inductive reasoning which itself provides a foundation or {\em raison d'{\^e}tre} for all kinds of system for the storage and processing information.

\subsection{Implications}

Readers with a pragmatic turn of mind may say ``So what?''. What benefits may there be, now or in the future, from an interpretation of phenomena in information processing which puts information compression centre stage?

In the rest of this book, I describe how this view may be developed into a general theory of computing and cognition that provides a radical simplification, rationalisation and integration of many concepts in these fields. The theory offers new insights and suggests new directions for research. A simplified view of computing can mean benefits for everyone concerned with the application of computers or the development of computer-based systems. It can also be a substantial benefit in the teaching of computer skills.

In more concrete terms, the view which has been described can mean new and better kinds of computer. The chapters that follow describe how logical deduction, probabilistic inference, inductive learning, information retrieval and other capabilities may be derived from information compression.
Chapter \ref{future_chapter} describes a vision of how the SP system may be developed and applied in the future.

%% file: theory.tex
\chapter[The SP Theory]{The SP Theory%
\protect\footnote{The work reported in this chapter was supported in part by a grant from the UK Science and Engineering Research Council (GR/G51565).}}\label{theory_chapter}

\section{Introduction}

This chapter provides a fairly comprehensive description of the SP theory, except for its development for unsupervised learning which is described in Chapter \ref{learning_chapter}.

The overall structure of the system is described first, followed by a description of the simple but versatile format for representing knowledge that is used in the system. The way in which the concept of {\em multiple alignment} has been developed for the SP theory is described followed by an account of the way in which multiple alignments are evaluated in terms of information compression and the way in which absolute and relative probabilities may be calculated for multiple alignments and associated structures. The way in which the system may achieve the apparent paradox of `decompression by compression' is explained. The chapter finishes with an outline description of the two main computer models that have been developed (SP61 and SP70) followed by a detailed description of SP61 (a full description of the SP70 model is given in Chapter \ref{learning_chapter}).

\section{The overall framework}\label{overall_framework}

The SP theory is intended as an abstract model of {\em any} kind of system for processing information, either natural or artificial. It is a development of the conjecture that {\em all kinds of information processing in brains or computers---including `thinking', `reasoning', `computing', and so on---may be understood as information compression by multiple alignment, unification and search}.

In its most general form, the theory is envisaged as a system for the unsupervised inductive learning of grammar-like structures that works like this:

\begin{itemize}

\item Starting with little or no knowledge of the `world', the system receives raw data from the world via its `senses'. These data are designated {\em New}\index{New}.

\item As it is received, New information is transferred to a repository of stored information called {\em Old}\index{Old}. At the same time, the system tries to compress the information as much as possible by finding patterns (or parts of patterns) that match each other and unifying them, introducing new identifiers where appropriate or using existing identifiers to encode the information in an economical manner (as described in Section \ref{identifiers_and_references}).

\item In the process of searching for matching patterns, the system builds {\em multiple alignments} of the kind described below.

\item The entire process is based on minimum length encoding principles, described in Section \ref{mle_section}.

\end{itemize}

In broad terms, this incremental scheme is similar to the well-known and widely-used Lempel-Ziv\index{information!compression!Lempel-Ziv} algorithms for information compression (`ZIP' programs). What is different about the SP scheme is an emphasis on thoroughness of searching rather than speed of processing and the way the concept of multiple alignment has been developed to support the encoding of New information in a hierarchy of `levels'.

As we saw in Section \ref{matching_searching_and_constraints}, the process of searching for `good' matches between patterns is not practical unless the process is constrained in some way. This is especially true for the formation of multiple alignments, where all practical methods use heuristic techniques or arbitrary constraints of one kind or another.

\subsection{New and Old in relation to established concepts}

\index{New|(}\index{Old|(}

The possible ways in which the concepts of New information and Old information may relate to established concepts in computing and cognition are shown in Table \ref{new_old_table}.

\begin{table}[!hbt]
\centering
\begin{tabular}{l l l}
\em Area of &       \em New &           \em Old \\
\em application \\
\\
Unsupervised &      `Raw' data &        Grammar or other \\
inductive learning &    &               knowledge structure \\
&               &               created by learning \\
\\
Parsing &           The sentence &      The grammar used for \\
&               to be parsed. &     parsing. \\
\\
Pattern &           A pattern to be &       The stored knowledge \\
recognition &       recognised &        used to recognise \\
and scene &         or scene to be &        one pattern or several \\
analysis &          analysed. &         within a scene. \\
\\
Databases &         A `query' in SQL or &   Records stored \\   
 &              other query language. & in the database. \\
\\
Expert &            A `query' in the &  The `rules' or other \\
system &            query language for &    knowledge stored in \\
 &              the expert system. &    the expert system. \\
\\
Computer &          The `data' or &         The computer program \\
program &           `parameters' &      itself. \\
 &              supplied to the \\
 &              program on each run. \\
\end{tabular}
\caption{The way in which the concepts of `New' and `Old' in this research appear to relate to established concepts in computing.}
\label{new_old_table}
\end{table}

\index{New|)}\index{Old|)}

\subsection{Computer models}

During the development of the SP theory, computer models have been developed as a means of testing the ideas and also as a means of demonstrating what can be done with the system.

Two main models have been developed to date:

\begin{itemize}

\item SP61\index{SP61} realises the building of multiple alignments and the encoding of New information in terms of user-defined Old information but without any attempt to add to or modify the information in Old. SP61 also contains procedures for calculating probabilities of inferences that may be drawn by the system, as described below. Although SP61 is largely a subset of SP70 (next), it is convenient to run it as a separate model for many applications. This model, which is relatively robust and mature, provides most of the examples of multiple alignments in this book.

\item SP70\index{SP70} realises all the elements of the framework, including the building of the repository of Old information in accordance with minimum length encoding principles. This model demonstrates the unsupervised learning of grammar-like structures from appropriate data but requires further development to realise its full potential.

\end{itemize}

Both models are described in outline later in the chapter. SP61 is also described in more detail at the end of this chapter. A relatively full description of SP70 is given in Chapter \ref{learning_chapter} where unsupervised learning is the main focus of interest.

\section{Representation of knowledge}\label{representation_of_knowledge}

\index{knowledge, representation of|(}

Given the intended wide scope of the theory, a goal of the research has been to devise a `universal' scheme for the representation of knowledge, capable of representing many different kinds of knowledge succinctly and integrating them seamlessly. Naturally, the scheme should be compatible with the various uses to which the knowledge will be put.

What has emerged is almost the simplest conceivable format for knowledge: arrays or {\em patterns}\index{pattern} of atomic {\em symbols}\index{symbol} in one or two dimensions and possibly more. One-dimensional patterns have many applications and have, to date, been the main focus in this research programme. However, it is envisaged that the ideas will at some stage be generalised to patterns in two dimensions, with potential for the representation and processing of pictures, maps and diagrams (see Chapter \ref{future_chapter}). To avoid pre-judging whether arrays of symbols are in one or two dimensions, the relatively general term {\em pattern} is normally used, rather than `sequence' or `string'.

Despite the extreme simplicity of this format for representing knowledge, the way it is processed within the SP system means that it can model a variety of established representational schemes including context-free phrase-structure grammars, context-sensitive grammars, class-inclusion hierarchies, part-whole hierarchies, if-then rules, discrimination networks, and other schemes. Many examples will be found in the chapters that follow. 

As a working hypothesis in this research, it is assumed that principles of information compression provide the key both to the succinct representation of diverse kinds of knowledge and to the manipulation of that knowledge for purposes of learning, reasoning, inference, calculation and the like.

\subsection{Patterns and the neural basis of knowledge}

Part of the inspiration for the idea of representing all kinds of knowledge with patterns was the observation that, in the brain and nervous system, information is often mapped into `sheets' of neurons as, for example, in the retina and in the layers of the lateral geniculate body and the cortex. It seems plausible to suppose that the way in which sensory information is mapped to sensory tissue would have provided a basis in evolution for the storage and processing of all empirical knowledge. The way in which the SP concepts may be realised in terms of neural mechanisms is discussed quite fully in Chapter \ref{neural_chapter}.

\subsection{Syntax and semantics}\label{syntax_and_semantics}

An important feature of the concept of a {\em symbol}\index{symbol}, as it is used in
this research, is that, with one qualification, it has {\em
no hidden meaning}. As described below, a symbol is simply a mark that can be matched with other symbols in a yes/no manner. Symbols such as `6', `22', `+', `-', `$\times$' and `/' may be used in the SP framework but each one, in itself, does not have the meaning that it would have in a mathematical equation (but see Section \ref{maths_and_sp}, below).

Any meanings that may attach to symbols within a given body of knowledge are to be expressed in the form of other symbols and patterns within that knowledge. In accordance with the goal of providing a universal format for diverse kinds of knowledge, {\em there is no formal distinction between `syntax' and `semantics'}. Any pattern, or part of a pattern, may serve either r{\^o}le, in a flexible manner according to need.

The one qualification to the slogan ``no hidden meaning'' is that, within each pattern, there is a distinction between symbols that serve as `identifiers' for that pattern and symbols that represent the `contents' of the pattern (Section \ref{id_and_c_symbols}, below). Labels like `identifier' or `contents' reflect the workings of the system itself and may therefore be regarded as distinct from `user-oriented' meanings that are intrinsic to the material being processed.

\subsection{Declarative and procedural knowledge}\label{declarative_procedural}

In the SP system, there is no formal distinction between declarative knowledge and procedural knowledge as there is in ACT-R \citep{anderson_lebiere_1998}. Furthermore, a given body of knowledge may be seen as `declarative' in some contexts and as `procedural' in others. For example, patterns that describe the structure of a sentence (like those that will be seen in Section \ref{framework_examples_section} and Chapter \ref{language_chapter}) may be seen to be `declarative' if they are used to describe the static structure of a sentence on a printed page but they may be seen as `procedural' if they are used in the dynamic process of speaking the sentence. Likewise, patterns that represent a map of some geographical area may be seen as a static statement of the arrangement of roads, towns and other things in the given area but they may also serve as a procedural guide if one is travelling through that area.

\subsection{Mathematics and the SP framework}\label{maths_and_sp}

\index{mathematics|(}

In Chapter \ref{computing_chapter} it is argued that, while there are important differences between the SP theory and the Turing concept of computing, the computational scope of the SP system is the same as a universal Turing machine: anything that can be computed with a universal Turing machine can also be computed in the SP system and both systems are subject to the same limitations (see Section \ref{scope_of_sp_theory}). Thus the representation and use of mathematical constructs should be possible within the SP system, as it is with a universal Turing machine. Chapter \ref{maths_logic_chapter} provides relevant examples and discussion.

In order to perform arithmetic and other kinds of calculation in the SP system it would be necessary to define relevant concepts with appropriate patterns. The mathematical meanings of symbols such as `6', `22', `+', `-', `$\times$' and `/' would need to be defined. There is probably also a need for constraints that are optimised for mathematical operations (Section \ref{matching_searching_and_constraints}). Given appropriate definitions and constraints, the system should be good for mathematics. 

That said, this aspect of the system has not yet received detailed attention. Pending further work in this area, the main strengths of the SP system are in non-mathematical kinds of symbolic manipulation such as pattern recognition, information retrieval, reasoning and so on.%
\index{mathematics|)}

\subsection{Terms}\label{terms_section}

This section summarises the meanings of terms associated with the main concepts introduced so far.

\subsubsection{`Symbol'}

A {\em symbol}\index{symbol} is some kind of mark that can be compared with any other symbol to decide whether it is the `same' or `different'. No other result is permitted. In the context of pattern matching, a symbol is the smallest unit that can participate in matching. 

\subsubsection{`Pattern'}

A {\em pattern}\index{pattern} is an array of symbols in one, two or more dimensions. In this book, one dimensional patterns ({\em sequences} or {\em strings} of symbols) are the main focus of attention.

\subsubsection{`Symbol type' and `alphabet'}

If two symbols match, we say that they belong to the same {\em symbol type}\index{symbol!type}. In any set of patterns, we normally recognise an {\em alphabet} of symbol types such that every symbol in the system belongs in one and only one of the symbol types in the alphabet, and every symbol type is represented at least once in the set of patterns.

\subsubsection{`Substring'}

A {\em substring}\index{substring} is a sequence of symbols of length $n$ within a sequence of length $m$, where $n \leqslant m$ and where the constituent symbols in the substring are contiguous within the sequence which contains the substring.

\subsubsection{`Subsequence'}

A {\em subsequence}\index{subsequence} is a sequence of symbols of length $n$ within a sequence of length $m$, where $n \leqslant m$ and where the constituent symbols in the subsequence may not be contiguous within the sequence which contains the subsequence. The set of all subsequences of a given sequence includes all the substrings of that sequence.

\subsubsection{`Chunk'}

A substring or subsequence that repeats two or more times within a given sequence is sometimes described informally as a {\em chunk}\index{chunk} of information.%
\index{knowledge, representation of|)}

\section{Multiple alignments}\label{multiple_alignments_section}

\index{multiple alignment|(}

The term {\em multiple alignment} has been borrowed from bioinformatics where
it means the arrangement of two or more sequences of symbols in
rows (or columns) so that, by judicious `stretching'
of sequences where necessary, symbols that match each other from one
sequence to another can be brought into alignment in columns (or rows).

Multiple alignments like these are normally used in the computational
analysis of (symbolic representations of) sequences of DNA bases or
sequences of amino acid residues as part of the process of elucidating
the structure, functions or evolution of the corresponding molecules. An example
of this kind of multiple alignment is shown in Figure \ref{DNA_figure}.

\begin{figure}[!hbt]
\fontsize{10.00pt}{12.00pt}
\centering
\begin{BVerbatim}
  G G A     G     C A G G G A G G A     T G     G   G G A
  | | |     |     | | | | | | | | |     | |     |   | | |
  G G | G   G C C C A G G G A G G A     | G G C G   G G A
  | | |     | | | | | | | | | | | |     | |     |   | | |
A | G A C T G C C C A G G G | G G | G C T G     G A | G A
  | | |           | | | | | | | | |   |   |     |   | | |
  G G A A         | A G G G A G G A   | A G     G   G G A
  | |   |         | | | | | | | |     |   |     |   | | |
  G G C A         C A G G G A G G     C   G     G   G G A
\end{BVerbatim}
\caption{A `good' alignment amongst five DNA sequences.}
\label{DNA_figure}
\end{figure}

A `good' alignment is, in general, one where there is a relatively
large number of {\em hits} (positive matches between symbols) and where any
{\em gaps} (sequences of unmatched symbols between hits) are relatively few
and relatively short. As we shall see (Section \ref{ma_evaluation}), it also possible to
evaluate multiple alignments in terms of minimum length encoding principles. Multiple alignments that are good in those
terms are normally good in terms of numbers of hits and the numbers and sizes of
gaps. A symbol that participates in a hit may be described as a {\em hit symbol}.

\subsection{Multiple alignments in the SP framework}\label{sp_multiple_alignment}

As in bioinformatics, a multiple alignment in the SP system is an arrangement of two or more patterns in rows (or columns), with one pattern in each row (or column). However, in the SP system, the multiple alignment concept has been modified in the following way:

\begin{itemize}

\item One or more of the patterns to be aligned represents New information. All other patterns are Old information.

\item A `good' multiple alignment is one which, through the unification of
symbols in New with symbols in Old, and through unifications amongst
the symbols in Old, leads to a relatively large amount of compression
of New in terms of the sequences in Old. How this may be done is
explained in Section \ref{ma_evaluation}, below.

\item By contrast with `multiple alignment' as normally understood in
bioinformatics, any given pattern in Old may appear two or more times
in any one multiple alignment and, in these cases, it is possible for the given
sequence to be aligned with itself. As explained below (Section \ref{multiple_appearances}), two or more {\em appearances} of a pattern within one multiple alignment is {\em not} the same as including two or more {\em copies} of a pattern in a multiple alignment.

\item Between any two Old patterns in a multiple alignment, there must not be any {\em mismatches}, as described in Section \ref{mismatches_section}, below.

\item Any symbol in one pattern may be aligned with any other symbol from the same pattern or another pattern, providing {\em order constraints} are not violated (Section \ref{multiple_appearances}, below).

\item It is envisaged that, at some point in the future, the concept of multiple alignment as it is understood here will be generalised to alignments of patterns with two dimensions and possibly more.

\end{itemize}

\subsection{Examples}\label{framework_examples_section}

A good introduction to the nature of the current system and its capabilities is the way it can be used for parsing a sentence. The examples of multiple alignment shown here will be referred to in later explanations and discussion.

Like the examples shown here, many of the other multiple alignments shown in this book are relatively simple. This is because simple examples are easier to understand than complicated examples and because large multiple alignments are often too big to fit on one page unless fonts are made so small that they are unreadable. It should be emphasised that {\em small examples do not reflect limitations of the SP system or computer models}. The system is quite capable of producing much larger examples. The computational complexity of the system---which is polynomial---is described in Sections \ref{sp21_computational_complexity}, \ref{sp61_computational_complexity} and \ref{sp70_computational_complexity}.

In the same vein, it should be emphasised that these first examples do {\em not} show the full power of the SP system for representing the structure of natural languages. These examples reproduce the effects of a context-free phrase-structure grammar and it is well-known (since Chomsky's \citeyearpar{chomsky_1957} {\em Syntactic Structures}) that such grammars are not fully adequate for representing the structure of natural languages like English. However, in Chapter \ref{language_chapter} it will be seen that the system can represent `context sensitive' aspects of language structure such as number or gender dependencies that bridge arbitrary amounts of intervening structure and the subtle structure of inter-locking dependencies in English auxiliary verbs. The system also supports the integration of natural language syntax and semantics.

Figure \ref{grammar_1} shows a simple context-free phrase-structure grammar\index{grammar!phrase-structure ---|(}---similar to the one shown in Figure \ref{cf_psg_figure}---describing a fragment of the syntax of English. This grammar generates sentences like `t h i s b o y l o v e s t h a t g i r l', `t h a t b o y h a t e s t h i s g i r l', and so on. Any of these sentences may be parsed in terms of the grammar giving a labelled bracketing like this:

\begin{center}
\fontsize{09.00pt}{10.80pt}
\begin{BVerbatim}
(S(NP(D t h i s)(N b o y))(V l o v e s)(NP(D t h a t)(N g i r l)))
\end{BVerbatim}
\end{center}

\noindent or an equivalent representation in the form of a tree.

\begin{figure}[!hbt]
\centering
\begin{tabular}{l}
S $\rightarrow$ NP V NP \\
NP $\rightarrow$ D N \\
D $\rightarrow$ t h i s \\
D $\rightarrow$ t h a t \\
N $\rightarrow$ g i r l \\
N $\rightarrow$ b o y \\
V $\rightarrow$ l o v e s \\
V $\rightarrow$ h a t e s \\
\end{tabular}
\caption{A context-free phrase-structure grammar describing a fragment of English syntax.}
\label{grammar_1}
\end{figure}

Figure \ref{grammar_2} shows the grammar from Figure \ref{grammar_1}
expressed as a set of SP patterns, in two versions.
Each pattern in either version is like a re-write rule in the context-free phrase-structure grammar notation except that
the rewrite arrow has been removed, some other symbols have been
introduced and there is a number shown in brackets to the right of each rule. Incidentally, the term `grammar' will be used in this book to describe any set of patterns (like those shown in Figure \ref{grammar_2} (a) or (b)) that has a r{\^o}le comparable to that of a grammar expressed in more traditional style.

\begin{figure}[!hbt]
\centering
\begin{BVerbatim}
!S NP #NP V #V NP #NP !#S (500)
!NP D #D N #N !#NP (1000)
!D !0 t h i s !#D (600)
!D !1 t h a t !#D (400)
!N !0 g i r l !#N (300)
!N !1 b o y !#N (700)
!V !0 l o v e s !#V (650)
!V !1 h a t e s !#V (350)

(a)

!< !S < NP > < V > < NP > !> (500)
!< !NP < D > < N > !> (1000)
!< !D !0 t h i s !> (600)
!< !D !1 t h a t !> (400)
!< !N !0 g i r l !> (300)
!< !N !1 b o y !> (700)
!< !V !0 l o v e s !> (650)
!< !V !1 h a t e s !> (350)

(b)
\end{BVerbatim}
\caption{The grammar from Figure \ref{grammar_1} recast as
patterns of symbols. In version (a), each pattern starts and ends with symbols representing a grammatical category. In version (b), angle brackets are used to mark the beginning and end of each pattern and the grammatical category is shown with a symbol following the initial left bracket. The character `!' in both versions serves to mark a symbol as being an `ID-symbol', as explained in Section \ref{id_and_c_symbols}, below.}
\label{grammar_2}
\end{figure}

\index{grammar!phrase-structure ---|)}

For any given rule, the number in brackets on the right is a frequency of occurrence of the rule in a (`good') parsing of a notional sample of the language. These frequencies have a r{\^o}le in the evaluation of multiple alignments in terms of information compression (Section \ref{ma_evaluation}) and in the calculation of probabilities associated with multiple alignments (Section \ref{probabilities_section}).

In examples shown later in the book, patterns are often shown without associated numbers. Unless otherwise indicated, readers may assume in those cases that each pattern has the default frequency of 1.

\subsection{Parsing as an alignment of a sentence and rules in a grammar}\label{parsing_as_alignment}

\index{parsing!language|(}

\begin{figure}[!hbt]
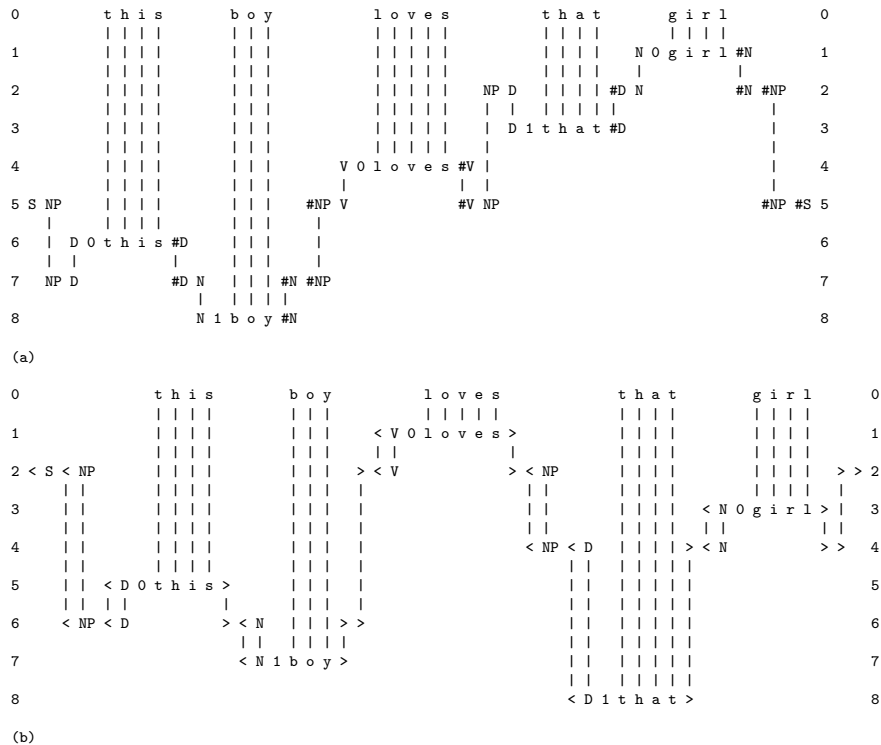

\fontsize{06.00pt}{07.20pt}
\centering
\begin{BVerbatim}
0          t h i s        b o y            l o v e s           t h a t        g i r l           0
           | | | |        | | |            | | | | |           | | | |        | | | |            
1          | | | |        | | |            | | | | |           | | | |    N 0 g i r l #N        1
           | | | |        | | |            | | | | |           | | | |    |           |          
2          | | | |        | | |            | | | | |    NP D   | | | | #D N           #N #NP    2
           | | | |        | | |            | | | | |    |  |   | | | | |                  |      
3          | | | |        | | |            | | | | |    |  D 1 t h a t #D                 |     3
           | | | |        | | |            | | | | |    |                                 |      
4          | | | |        | | |        V 0 l o v e s #V |                                 |     4
           | | | |        | | |        |             |  |                                 |      
5 S NP     | | | |        | | |    #NP V             #V NP                               #NP #S 5
    |      | | | |        | | |     |                    
6   |  D 0 t h i s #D     | | |     |                                                           6
    |  |           |      | | |     |                    
7   NP D           #D N   | | | #N #NP                                                          7
                      |   | | | |                        
8                     N 1 b o y #N                                                              8

(a)

0                t h i s         b o y           l o v e s              t h a t         g i r l       0
                 | | | |         | | |           | | | | |              | | | |         | | | |      
1                | | | |         | | |     < V 0 l o v e s >            | | | |         | | | |       1
                 | | | |         | | |     | |             |            | | | |         | | | |      
2 < S < NP       | | | |         | | |   > < V             > < NP       | | | |         | | | |   > > 2
      | |        | | | |         | | |   |                   | |        | | | |         | | | |   |  
3     | |        | | | |         | | |   |                   | |        | | | |   < N 0 g i r l > |   3
      | |        | | | |         | | |   |                   | |        | | | |   | |           | |  
4     | |        | | | |         | | |   |                   < NP < D   | | | | > < N           > >   4
      | |        | | | |         | | |   |                        | |   | | | | |                    
5     | |  < D 0 t h i s >       | | |   |                        | |   | | | | |                     5
      | |  | |           |       | | |   |                        | |   | | | | |                    
6     < NP < D           > < N   | | | > >                        | |   | | | | |                     6
                           | |   | | | |                          | |   | | | | |                    
7                          < N 1 b o y >                          | |   | | | | |                     7
                                                                  | |   | | | | |                    
8                                                                 < D 1 t h a t >                     8

(b)
\end{BVerbatim}
\caption{(a) The best multiple alignment found by SP61 with `t h i s b o y l o v e s
t h a t g i r l' in New and the grammar from Figure \ref{grammar_2} (a) in Old. (b) The best multiple alignment found using the grammar in Figure \ref{grammar_2} (b).}
\label{alignment_figure_1}
\end{figure}

Figure \ref{alignment_figure_1} (a) shows how a parsing of the sentence `t h i s b o y
l o v e s t h a t g i r l' may be seen as an alignment of patterns
which includes the sentence and relevant rules from the grammar shown
in Figure \ref{grammar_2} (a). Figure \ref{alignment_figure_1} (b) is the corresponding multiple alignment using the grammar shown in Figure \ref{grammar_2} (b).

In multiple alignments like these, the sentence or other
sequence of symbols to be parsed is the New pattern, while the patterns representing rules of the grammar are Old. By convention, the New pattern is always shown at the top of a multiple alignment like the one shown here with the Old patterns in the rows underneath, one pattern per row. Apart from this constraint, the order of the patterns across the rows is entirely arbitrary, without any significance. 

As we shall see, it can be convenient to show multiple alignments rotated through $90^{o}$. In these cases, the New pattern is shown in the left-most column, with Old patterns---in any order---in columns to the right. Of course, when a multiple alignment has been rotated in this way, the meaning of `row' and `column' changes. To avoid introducing clumsy new terms to generalise these concepts across both arrangements of a multiple alignment, the rest of this chapter will refer to rows and columns of a multiple alignment on the assumption that multiple alignments are arranged as shown in Figure \ref{alignment_figure_1} with patterns in the rows and alignments of symbols shown in columns.

The similarity between the first multiple alignment and the conventional parsing may be seen if matching symbols in the multiple alignment are unified so that, in effect, the multiple alignment is `projected' on to a single sequence, thus:

\begin{center}
\begin{BVerbatim}
S NP D 0 t h i s #D N 1 b o y #N #NP V 0 l o v e s #V
     NP D 1 t h a t #D N 0 g i r l #N #NP #S
\end{BVerbatim}
\end{center}

\noindent In this projection, the two instances of `NP' in the second column of
the multiple alignment have been merged or `unified' and likewise for the two
instances of `D' in the third column and so on wherever there are two
or more instances of a symbol in any column.

This projection is the same as the conventional parsing except that `0' and `1' symbols are included, right bracket symbols (`)') are replaced by symbols such as `\#D', `\#N' and `\#NP', and each of the upper-case symbols is regarded both as a `label' for a structure and as a left bracket for that
structure.

The similarity with a conventional parsing can be seen even more clearly in the multiple alignment shown Figure \ref{alignment_figure_1} (b) which projects to a single sequence, thus:

\begin{center}
\begin{BVerbatim}
< S < NP < D 0 t h i s > < N 1 b o y > > < V 0 l o v e s >
     < NP < D 1 t h a t > < N 0 g i r l > > >
\end{BVerbatim}
\end{center}

\noindent Apart from the use of angle brackets instead of round brackets and the addition of symbols like `0' and `1', the projection is exactly the same as an ordinary labelled bracketing.%
\index{parsing!language|)}

\subsection{Identification symbols and contents symbols}\label{id_and_c_symbols}

\index{symbol!identification|(}\index{symbol!contents|(}

Any one symbol in an SP pattern is either a {\em contents} symbol (C-symbol) or it is an {\em identification} symbol (ID-symbol). In general, the former represents the contents or substance of knowledge stored in the pattern while the latter serve to define how a pattern relates to other patterns in a given set of patterns as described below.

In Figure \ref{grammar_2}, each symbol that is preceded by `!' is an ID-symbol and all other symbols are C-symbols. The `!' character is simply a marker for ID-symbols and is not part of the symbol itself. Notice that any given symbol type (e.g., `NP') may be an ID-symbol in some contexts and a C-symbol in others. As a general rule, ID-symbols appear at or near the beginning and end of each pattern (e.g., `!D' `!0' and `!\#D' in the pattern `!D !0 t h i s !\#D') while corresponding C-symbols appear within the body of a pattern (e.g., the pair of symbols `D \#D' within the pattern `!NP D \#D N \#N !\#NP'). Within each pattern, it is normally reasonably clear which symbols are ID-symbols and which ones are C-symbols---so in most of the examples in this book, the `!' marker for ID-symbols is omitted.

An ID-symbol may have either or both of two main functions:

\begin{itemize}

\item It may serve to represent the left or right boundary of a pattern. In Figure \ref{grammar_2} (a), examples include `S' and `\#S' at the left and right of the first pattern, `NP' and `\#NP' at the left and right of the second pattern, and so on. In Figure \ref{grammar_2} (b), left and right angle brackets play the same r{\^o}le. It is sometimes convenient to refer to symbols like these as (left or right) `boundary markers' and to refer to symbols like `\#S', `\#NP' and `$>$' at the right ends of patterns as `termination markers'---but these concepts have no formal status in the system.

In the multiple alignments shown in Figure \ref{alignment_figure_1}, the boundary markers are aligned with matching C-symbols in other patterns. For example, `N' and `\#N' in `N 1 b o y \#N' are aligned with `N \#N' in `NP D \#D N \#N \#NP'. It is these alignments of symbols that ensure that a coherent multiple alignment can be formed. 

In conventional computing, it may seem as if termination markers are not needed but they are often hidden from view, e.g., `EOF' at the end of a file or the invisible `LF' character at the end of a line, or space characters marking the ends of non-space strings of characters.

In some applications of the SP system, some or all of the patterns do not need to have boundary markers. Examples will be seen later in the book. The provision of this kind of symbol is not an essential part of the system.

\item As its name implies, an ID-symbol---either singly or in combination with other ID-symbols---may serve to distinguish a pattern from other patterns and thus identify it within a given set of patterns. In Figure \ref{grammar_2} (a), examples include `S' at the beginning of `S NP \#NP V \#V NP \#NP \#S', `D' and `0' at the beginning of `D 0 t h i s \#D', `N' and `1' at the beginning of `N 1 b o y \#N' and likewise for the other patterns. In these examples, `S', `D' and `N' also serve as left boundary markers. In Figure \ref{grammar_2} (b), patterns are identified with the same symbols but the boundary marking r{\^o}le is been delegated to left and right angle brackets.

ID-symbols like `S', `D' or `N' near the beginnings of patterns are sometimes called `class' symbols because they represent grammatical categories or other kinds of class. Since ID-symbols like `D' or `N' appear in more than one pattern, they need to be combined with symbols like `0' and `1' in order to identify a pattern uniquely. The latter symbols are sometimes referred to as `discrimination' symbols because they enable one to distinguish one member of a class from another. Neither of these terms has any formal status in the system.

A pair of C-symbols like `NP \#NP' in Figure \ref{grammar_2} (a) or a triplet like `$<$ NP $>$' in Figure \ref{grammar_2} (b) is a means of referring to one other pattern in the set of patterns or to two or more patterns in the set. Consequently, this kind of combination of C-symbols is sometimes called a `reference' to a pattern or a class. Again, this term has no formal status in the system.

\end{itemize}

The only distinction amongst symbols with any formal status within the SP system is the distinction between ID-symbols and C-symbols. Otherwise, every symbol has the same status as any other symbol and it makes no difference whether they are written with letters or digits, with upper-case or lower-case letters, with an initial `\#' character or without, and so on.%
\index{symbol!identification|)}\index{symbol!contents|)}

\subsection{Any one pattern may appear two or more times in one alignment}\label{multiple_appearances}

\index{multiple alignment!constraints}

As was noted in Section \ref{sp_multiple_alignment}, any one pattern may appear two or more times in one multiple alignment. Thus, for example, the pattern `NP D \#D N \#N \#NP' appears twice in the multiple alignment shown in Figure \ref{alignment_figure_1} (a). 

It should be stressed that two or more {\em appearances} of a pattern in a multiple alignment is {\em not} the same as having two or more {\em copies} of a given pattern in the multiple alignment. In the former case, it is {\em not} permissible to align a given symbol in one appearance of a pattern with the corresponding symbol in another appearance---because this would mean aligning the given symbol with itself and it is obvious that any such match would be meaningless
in terms of the identification of redundancy. In the latter case, it is permissible to align a given symbol in one copy of a pattern with the corresponding symbol in another copy. However, in most applications, the repository of Old patterns would not normally contain multiple copies of any one pattern. 

In addition to the constraint that any given symbol must not be aligned with itself, it is also necessary to ensure that {\em order constraints} are preserved. This means that the following statement is always true:

\begin{quote}
For any two rows in a multiple alignment, A and B, and any four symbols, A$_1$ and A$_2$ in A, and B$_1$ and B$_2$ in B, if A$_1$ is in the same column as B$_1$, and if A$_2$ is in the same column as B$_2$, and if A$_2$ in A follows A$_1$ in A, then B$_2$ in B must follow B$_1$ in B. 
\end{quote}

\noindent This condition applies when the two rows contain two appearances of one pattern and also when the two rows contain different patterns.

\subsection{`Mismatches'}\label{mismatches_section}

\index{multiple alignment!mismatch|(}

A {\em mismatch} between a pair of patterns in a multiple alignment occurs when, between one pair of aligned symbols and another, or between one pair of aligned symbols and the beginnings or ends of the two patterns, there are no other aligned pairs of symbols and there are unmatched symbols in each of the two patterns.

In the SP61 model and also the SP70 model, multiple alignments are rejected if they contain mismatches between {\em Old} patterns in the multiple alignment but mismatches between the New pattern and any Old pattern are accepted. Thus, for example, a multiple alignment like this:

\begin{center}
\begin{BVerbatim}
0 a   c 0
  |   |
1 a x c 1
  |   |
2 a y c 2
\end{BVerbatim}
\end{center}

\noindent or this:

\begin{center}
\begin{BVerbatim}
0 a b   0
  | |
1 a b x 1
  | |
2 a b y 2
\end{BVerbatim}
\end{center}

\noindent would be rejected. But multiple alignments like this:

\begin{center}
\begin{BVerbatim}
0 a b c 0
  |   |
1 a x c 1
\end{BVerbatim}
\end{center}

\noindent or this:

\begin{center}
\begin{BVerbatim}
0 a b c 0
  | |
1 a b x 1
\end{BVerbatim}
\end{center}

\noindent would be accepted. (In any multiple alignment, row 0 contains the New pattern and all other rows contain Old patterns.)

The rejection of multiple alignments that contain mismatches between Old patterns ensures that all `legal' multiple alignments can be unified to form a single sequence as shown in Section \ref{parsing_as_alignment}. This is true even if there is a mismatch between the New pattern in a multiple alignment and one or more of the Old patterns. Thus, for example, the first of the two multiple alignments just shown would unify to form the sequence `a x c' and the second one would unify to form the sequence `a b x'. In cases like these, the New pattern in the unified sequence is a {\em subsequence} of the New pattern that was originally supplied. The symbols in that subsequence are those that form hits with symbols in the Old patterns. In SP61, the unmatched symbols in the original New pattern are, in effect, ignored. In SP70, unmatched symbols in the New pattern in a multiple alignment provide a basis for learning.%
\index{multiple alignment!mismatch|)}

\subsection{The process of searching for good multiple alignments and the need for constraints}\label{ma_searching_and_constraints}

\index{multiple alignment!constraints|(}

As we saw in Section \ref{matching_searching_and_constraints}, the number of possible ways in which two sequences of symbols may be matched is normally huge. For example, the number of possible subsequences in a sequence of 100 symbols is $1.26 \times 10^{30}$ and the number of possible comparisons between the subsequences in one such sequence and the subsequences in another is $8.034 \times 10^{59}$. For three or more sequences, this number would be even larger.

In work on multiple alignments, it is widely recognised that, with the exception of alignments of patterns that are very small and very few, the number of possible alignments of symbols is too large to be searched exhaustively. For any set of patterns of realistic size, a search which has acceptable speed and acceptable scaling properties can only be achieved if the search is constrained so that (large) parts of the search space are excluded from the search. As discussed in Section \ref{matching_searching_and_constraints}, constraints may {\em absolute} or {\em heuristic} or a combination of the two.

There is now a fairly large literature about methods for finding good alignments amongst two or more sequences of symbols. All practical methods use one or more forms of heuristic search. Some of them are reviewed in \cite{taylor_1988, barton_1990, chan_etal_1992, day_mcmorris_1992}. The ways in which constraints are applied in the SP61 model are described in Section \ref{searching_and_constraints_in_sp61}, below.%
\index{multiple alignment!constraints|)}

\subsection{Ordering of symbols and patterns}\label{ordering_of_symbols_and_patterns}

One-dimensional patterns in the SP scheme are ordered {\em sequences} of symbols, not unordered {\em sets} of symbols and, in the building of multiple alignments, the order of symbols matters. For many applications, such as the parsing of natural language (Section \ref{framework_examples_section} and Chapter \ref{language_chapter}), this is entirely appropriate. However, in some applications, it is useful to be able to treat New information as unordered or partially-ordered sets of symbols.

In order to ensure that `A' and `B' in one pattern can be matched with `B' and `A' in another pattern, we can provide a {\em framework pattern} that defines an arbitrary order for disjunctive classes of symbol that may be found amongst the patterns in Old. Examples will be seen in Sections \ref{medical_diagnosis_section} and \ref{reasoning_with_rules_section} and elsewhere in the book. Provided that classes of symbol are always specified in the defined order, appropriate multiple alignments can be built.

\subsubsection{An alternative method for unordered sets of patterns or symbols}\label{alternative_method_for_unordered_patterns}

At the time of writing, another method for accommodating unordered sets of patterns or symbols is being considered. Instead of specifying the New information as a single pattern, it is possible to specify it as a set of patterns. In many applications, it is likely that many of the patterns would contain only one symbol but there is no reason to exclude patterns containing two or more symbols. 

The SP61 model would need to be modified so that it could build multiple alignments where any one multiple alignment might contain two or more New patterns. It is envisaged that, in any such multiple alignment, the New patterns would normally be arranged in a sequence and they would normally be displayed in row 0 of the multiple alignment. That said, the possibility of overlap amongst the New patterns---shown in two or more rows of the multiple alignment---has not been ruled out. Either way, the formation of a multiple alignment would ensure that the symbols in the New patterns would project into a single sequence.

Although the New patterns in any one multiple alignment would project into a single sequence, the ordering of the patterns could vary from one multiple alignment to another. The only constraint would be that the order of symbols within any one pattern would be preserved. Given a system in which multiple alignments are built in this way then it should be possible to accommodate applications in which New information needs to be treated as an unordered set of patterns or an unordered or partially-ordered set of symbols.%
\index{multiple alignment|)}

\section{Coding and the evaluation of an alignment in terms of compression}\label{ma_evaluation}

\index{multiple alignment!evaluation|(}

The first subsection here describes in outline how multiple alignments are evaluated in the SP system and, more specifically, in SP61 and SP70. This description should be sufficient for many readers but a more detailed description is given in the next subsection with some of the rationale of the method and some of the practicalities.

\subsection{Outline}\label{ma_evaluation_outline}

Associated with each symbol type is a notional {\em code}\index{code} or bit pattern that serves to distinguish the given symbol type from all the others. This is only notional because the bit patterns are not actually constructed. All that is needed for the purpose of evaluating multiple alignments is the size of the notional bit pattern associated with each symbol type.

The sizes of the codes are calculated (as described below) so that frequently-occurring symbols have shorter codes than symbols that occur more rarely. {\em Notice that these bit patterns and their sizes are totally independent of the names for symbols used in written accounts like this one and chosen purely for their mnemonic value}.

Given a multiple alignment like the one shown in Figure \ref{alignment_figure_1}, one can derive a {\em code pattern}\index{code pattern} from the multiple alignment in the following way:

\begin{enumerate}

\item Scan the multiple alignment from left to right looking for columns that contain an ID-symbol by itself, not aligned with any other symbol.

\item Copy these symbols into a code pattern in the same order that they appear in the multiple alignment.

\end{enumerate}

\noindent The code pattern derived in this way from the multiple alignment shown in Figure \ref{alignment_figure_1} (a) is `S 0 1 0 1 0 \#S'. This is, in effect, a compressed representation of those symbols in the New pattern that form hits with Old symbols in the multiple alignment. In this case, the code pattern is a compressed representation of {\em all} the symbols in the New pattern.

Given a code pattern derived in this way, we may calculate a `compression difference'\index{CD}\index{compression difference} as:

\begin{equation}
CD = B_N - B_E
\label{CD_equation}
\end{equation}

\noindent or a `compression ratio'\index{CR}\index{compression ratio} as:

\begin{equation}
CR = B_N / B_E.
\label{CR_equation}
\end{equation}

Each of these measures is an indication of how effectively the New pattern (or parts of it) may be compressed in terms of the Old patterns that appear in the given multiple alignment. The $CD$ of a multiple alignment---which has been found to be more useful than $CR$---is often called the {\em compression score}\index{compression score} of the multiple alignment.

In each of these equations, $B_N$ is calculated as:

\begin{equation}
B_N = \sum_{i=1}^h C_i,
\label{BN_equation}
\end{equation}

\noindent where $C_i$ is the size of the code for $i$th symbol in a sequence, $H_1 ... H_h$, comprising those symbols within the New pattern that form hits with Old symbols within the multiple alignment.

$B_E$ is calculated as:

\begin{equation}
B_E = \sum_{i=1}^s C_i,
\label{BE_equation}
\end{equation}

\noindent where $C_i$ is the size of the code for $i$th symbol in the sequence of $s$ symbols in the code pattern derived from the multiple alignment.

\subsection{Details and rationale}\label{ma_evaluation_details}

This section presents a relatively detailed description of the method for evaluating multiple alignments in terms of compression and describes the reasoning behind the method.

\subsubsection{Encoding individual symbols}\label{encoding_individual_symbols}

The simplest way to encode individual symbols in the New pattern and the set of Old patterns is with a `block' code using a fixed number of bits for each symbol. In the New pattern shown in Figure \ref{alignment_figure_1} together with the grammar shown in Figure \ref{grammar_2} (a), there are 24 symbol types so the minimum number of bits required for each symbol is $\lceil \log_2 24\rceil = 5$ bits per symbol.

In fact, the SP61 model uses variable-length codes for symbols, assigned in accordance with the
Shannon-Fano-Elias coding scheme (described by \citet{cover_thomas_1991}) so
that the shortest codes represent the most frequent symbol types and {\em vice
versa}. Although this scheme is slightly less efficient than the well-known Huffman scheme, it has been adopted because the Huffman scheme can give anomalous results when probabilities are calculated (Section \ref{probabilities_section}).

For the Shannon-Fano-Elias calculation, the frequency of each symbol type ($f_{st}$) is calculated as:

\begin{equation}
f_{st} = \sum_{i=1}^P (f_i \times o_i)
\label{fst_equation}
\end{equation}

\noindent where $f_i$ is the (notional) frequency of the $i$th pattern in
the grammar (illustrated by the numbers on the right of Figure \ref{grammar_2}) (a),
$o_i$ is the number of occurrences of the given symbol in the $i$th
pattern and $P$ is the number of patterns in the grammar.

When the code sizes of symbol types have been calculated, each symbol in the New pattern and each symbol in the set of Old patterns is assigned the code size corresponding to its type. An important point to stress is that, when this process has been completed, the code size of each symbol in the New pattern is multiplied by a {\em cost factor}, normally about 2 or larger. This means that, in effect, each New symbol represents a relatively large chunk of information. When a New symbol is aligned with a matching Old symbol, it may be encoded with the smaller code associated with the Old symbol, in much the same way that `GMT' may be used as an abbreviation for `Greenwich Mean Time' or `WWW' is short for `World Wide Web' in ordinary writing. This arrangement means that compression can occur at the level of individual symbols as well at the level of patterns. This has a bearing on the apparent paradox of `decompression by compression', discussed in Section \ref{decompression_by_compression}, below.

There are many variations and refinements that may be made at the symbol
level but, in general, the choice of coding system for individual
symbols is not critical for the principles to be described below where
the focus of interest is the exploitation of redundancy that may be
attributed to sequences of two or more symbols rather than any
redundancy attributed to individual symbols.

\subsubsection{Encoding words}

In Figure \ref{alignment_figure_1} (a), the word `t h i s' in the New pattern is aligned with the corresponding symbols in the Old pattern `D 0 t h i s \#D'. From this multiple alignment, we derive a code pattern for `t h i s' from the ID-symbols in the Old pattern. The result is the pattern `D 0 \#D'.

In terms of numbers of symbols, this code is only slightly smaller than the original symbols in New. However, owing to the application of the cost factor each New symbol is, in effect, much larger than any of the Old symbols, including the ID-symbols used to create the code pattern. So the size of the code pattern in terms of bits is substantially smaller than the size of the New symbols that it represents.

The other words in the New pattern in Figure \ref{alignment_figure_1} (a)---`b o y', `l o v e s', `t h a t' and `g i r l'---may be encoded in a similar way.

\subsubsection{Encoding phrases}\label{encoding_phrases}

Consider the phrase `t h i s b o y'. If this were encoded with a code
pattern for each word, the result would be `D 0 \#D N 1 \#N' which is
only one symbol smaller than the original. However, we can encode the
phrase with fewer symbols by taking advantage of the fact that the
sequence `D 0 \#D N 1 \#N' has a subsequence, `D \#D N \#N', which is a
substring within the pattern `NP D \#D N \#N \#NP' in the grammar. Notice
that the sequence `D \#D N \#N' is discontinuous within the sequence `D 0
\#D N 1 \#N'.

Since the `noun phrase' pattern `NP D \#D N \#N \#NP' is in the grammar,
we may replace the substring, `D \#D N \#N', by the `code' sequence `NP
\#NP'. But then, to encode the two words within the noun phrase (`t h i
s' and `b o y'), we must add the symbols, `0' and `1' from `D 0 \#D N 1
\#N' so that the final coded sequence is `NP 0 1 \#NP'.

Notice how the symbols `NP' and `\#NP' in the code pattern `NP 0 1 \#NP'
serve as a disambiguating context so that the symbol `0' identifies the
pattern `D 0 t h i s \#D' and the symbol `1' identifies the pattern `N 1
b o y \#N'.  The overall cost of the code pattern `NP 0 1 \#NP' is 4
symbols compared with the original 7 symbols in `t h i s b o y'---a
saving of 3 symbols. In a similar way, the phrase `t h a t g i r l' may
be encoded as `NP 1 0 \#NP' which is 4 symbols smaller than the
original. In terms of the numbers of bits required, the savings are even greater.

\subsubsection{Encoding the sentence}\label{encoding_the_sentence}

Given the two noun phrases in their encoded forms (`NP 0 1 \#NP' for `t
h i s b o y' and `NP 1 0 \#NP' for `t h a t g i r l') and the encoding
of `l o v e s' as `V 0 \#V', the whole sentence may be encoded as `NP 0
1 \#NP V 0 \#V NP 1 0 \#NP'.

However, this sequence contains the subsequence `NP \#NP V \#V NP \#NP'
and this sequence is a substring within the `sentence' pattern `S NP
\#NP V \#V NP \#NP \#S'---and this pattern is in the grammar. So we may
replace the sequence `NP \#NP V \#V NP \#NP' by the `code' sequence `S
\#S'. To discriminate the words in this sentence we must add the symbols
`0 1 0 1 0' from the sequence `NP 0 1 \#NP V 0 \#V NP 1 0 \#NP'. The
overall result is an encoded representation of the sentence as:

\begin{center}
\begin{BVerbatim}
S 0 1 0 1 0 #S.
\end{BVerbatim}
\end{center}

The 7 symbols in this encoding of the sentence represents a substantial
compression compared with the 20 symbols in the unencoded sentence. As before, the saving in terms of bits is even greater.

In general, the procedure for deriving a code for a multiple alignment is exactly as described in Section \ref{ma_evaluation_outline}, above: scan the multiple alignment from left to right looking for ID-symbols that are not aligned with any other symbol and copy those symbols into a code pattern in the same order as they appear in the multiple alignment.

\subsubsection{Discussion}\label{ma_evaluation_discussion}

Each pattern expresses sequential redundancy in the data to be encoded
and this sequential redundancy can be exploited to reduce the number of
symbols which need to be written out explicitly. In the grammar shown
in Figure \ref{grammar_2}, each pattern for an individual word expresses the
sequential redundancy of the letters within that word; the pattern for
a noun phrase expresses the sequential redundancy of `determiner'
followed by `noun'; and the pattern for a sentence expresses the
sequential redundancy of the pattern: `noun phrase' followed by `verb'
followed by `noun phrase'.

Since this principle operates at all levels in the `hierarchy' of
patterns, many of the symbols at intermediate levels may be omitted
completely. A sentence may be specified with symbols marking the start
and end of the sentence pattern together with interpolated symbols
which discriminate amongst alternatives at lower levels.

Notice that these ideas are only applicable to multiple alignments without any `mismatches' between Old patterns (as described in Section \ref{mismatches_section}, above). Multiple alignments with mismatches between Old patterns are rejected by the system.

The method that has been described illustrates the role of context in
the encoding of information. Any one symbol like `0' or `1' is
ambiguous in terms of the patterns in the grammar in Figure
\ref{grammar_2} (a). But in the context of the pattern `S 0 1 0 1 0 \#S'
and the same grammar, it is possible to assign each instance of `0' or
`1' unambiguously to one of the words in the grammar, giving the
sequence of words in the original sentence. As we shall see in Section \ref{decompression_by_compression}, the encoded form of a New pattern may be used within the SP system to recreate that New pattern using exactly the same processes as were used to create the code in the first place.

\subsection{An approximate method}\label{approximate_method}

The foregoing describes how multiple alignments are evaluated in SP61 once they have been completed. However, in the course of building multiple alignments, the program uses an approximate method for evaluating partial alignments that is relatively quick and `cheap' to compute.

As we shall see in Sections \ref{sp61_outline} and \ref{sp61_detail_section}, the SP61 model builds multiple alignments by searching for full and partial matches between sequences, taken two at a time. For any alignment between two sequences, it creates a measure of how good the alignment is from the number of hits in the alignment and the number and sizes of the gaps between hits. The term {\em compression score} may also be used to describe this approximate measure.

The method of allowing for gaps is based on an analogy with the rolling of two $A$-sided dice, where $A$ is the size of the alphabet used in New and Old. The sequence of rolls of one die corresponds with the sequence of symbols in one pattern and the sequence of rolls of the other
die corresponds with the sequence of symbols in the other pattern. The
method is based closely on the method described in \citet{lowry_1989} for
calculating probabilities of various contingencies in problems of this
type. The method, as it has been adapted for the matching of two patterns, is described in Appendix \ref{matching_appendix} which is itself based on \citet{wolff_1994_scaleable}.%
\index{multiple alignment!evaluation|)}

\subsection{Critique of technique for evaluating alignments in terms of compression}

The method for evaluating alignments described in the first parts of this section makes good sense for any alignment in which there are few gaps between hit symbols in the New pattern and few unmatched C-symbols in the Old patterns, but it is less satisfactory for any alignment in which there are many gaps, especially if they are amongst the C-symbols in the Old patterns of the alignment. This is because the ID-symbols of each Old pattern represent {\em all} the C-symbols of the pattern, not some subset that happens to be matched to the symbols in New. In some respects, the approximate method described in the previous subsection is better for alignments containing gaps, because the method does take account of the numbers of gaps and their sizes.

Despite this apparent weakness, the coding method seems to work quite well in practice, as will be seen in examples shown in the next chapter (e.g., Section \ref{fuzzy_pattern_recognition}) and later.

It is envisaged that the problem of gaps and how they should be allowed for will be resolved when the system is fully developed for learning. As will be seen in Chapter \ref{learning_chapter}, the learning process is driven by gaps in hit sequences and each gap leads to the creation of system-generated ID-symbols. When this has been fully worked out, the additional `cost' of newly-created ID-symbols should create a bias against gaps in alignments, in accordance with our intuitions.

\section{Realisation of the three compression techniques in SP}\label{compression_in_sp}

The examples of multiple alignment that we have seen so far give us a 
preliminary view of the way in which the three compression techniques described in 
Section \ref{techniques_for_ic} may be modelled in the SP framework.

\subsection{Chunking-with-codes}

In Figure \ref{alignment_figure_1} (a), a pattern like `N 0 j o h n \#N' may be seen as a chunk of information (`j o h n') together with its `code' (`N 0 \#N'). In this case, the code has a dual function of providing a unique identifier for the chunk and marking the 
beginning and end of the chunk.

Superficially, the code in this example appears to be only slightly smaller than the 
chunk and there does not seem to be much advantage in terms of information compression in recognising the 
chunk. If `j o h n' had been replaced by `j i m', there would appear to be no compression at all. But it must be born in mind that each symbol has its own code and calculations of information compression are based on symbol sizes in bits not simple counts of numbers of symbols.

Sometimes a chunk is smaller than its corresponding code. Although this seems nonsensical, there may be an overall economy within a larger structure, as we shall see in the next subsection.

\subsection{Schema-plus-correction}\label{ma_schema_plus_correction}

In Figure \ref{alignment_figure_1} (a), it should be clear that a pattern like `S N \#N V \#V \#S' is taking the role of a schema that may be corrected or completed by the addition of a noun pattern and a verb pattern. Although sentence patterns in a language like English are relatively abstract concepts, they are patterns that repeat very often in the language and thus represent, in compressed form, relatively large amounts of redundancy.

Given a schema like `S N \#N V \#V \#S', it may make perfect sense in terms of economy to use `corrections' to that schema such as `N 23 I \#N', where `I' is the small `chunk' to be encoded and `N 23 \#N' are the ID-symbols used to encode it. The number of bits in the ID-symbols may be larger than the number of bits in the C-symbol used to represent `I' but, in any reasonably large body of text, there is an overall compression because `S N \#N V \#V \#S' (and other sentence patterns) are such powerful schemata and they need small patterns like `N 23 I \#N' in order to function properly.

\subsection{Run-length coding}

In the SP framework, there is no explicit mechanism for representing iterated structures as there is in computer programming (e.g., the {\em while} loop, the {\em for} loop, or the
{\em repeat ... until} loop). The sole mechanism is recursion in the sense of computer programming. Examples will be seen in Sections \ref{unary_numbers_and_sp}, \ref{language_recursion} and elsewhere in the book.

\section{Calculation of probabilities associated with alignments}\label{probabilities_section}

\index{probability|(}

As we shall see in Chapter \ref{pr_chapter}, the formation of multiple alignments in the SP framework supports a variety of kinds of probabilistic reasoning. The core idea is that any Old symbol in a multiple alignment that is {\em not} aligned with a New symbol represents an inference that may be drawn from the multiple alignment. 

This section describes how absolute and relative probabilities for such inferences may be calculated.

\subsection{Absolute probabilities}\label{absolute-probs}

\index{probability!absolute|(}

Any sequence of $L$ symbols, drawn from an alphabet of $|A|$ symbol types, represents one point in a set of $N$ points where $N$ is calculated as:

\begin{equation}
N = |A|^L.
\label{N_equation}
\end{equation}

\noindent {\em If we assume that the sequence is random or nearly so}, which means that the $N$ points are equi-probable or nearly so, the probability of any one point (which represents a sequence of length $L$) is close to:

\begin{equation}
p_{ABS} = |A|^{-L}.
\label{pABS_equation}
\end{equation}

\noindent This formula can be used to calculate the absolute probability of the code, $C$, derived from the multiple alignment as described in Section \ref{ma_evaluation}. In SP61, the value of $|A|$ is $2$. $p_{ABS}$ may be regarded as the probability of any inferences that may be drawn from the multiple alignment.

\subsubsection{Is it reasonable to assume that new in encoded form is random or nearly so?}\label{new_random}

Why should we assume that the code for a multiple alignment is a random sequence or nearly so? In accordance with algorithmic information theory (Section \ref{ait_section}), a sequence is random if it is incompressible. If we have reason to believe that a sequence is incompressible or nearly so, then we may regard it as random or nearly so.

As noted previously, we cannot prove that no further compression of $I$ is possible (unless $I$ is very small). But we may say that, for a given set of methods and a given amount of computational resources that have been applied, no further compression can be achieved. In short, the assumption that the code for a multiple alignment is random or nearly so only applies to the best encodings found for a given body of information in New and must be qualified by the quality and thoroughness of the search methods which have been used to create the code.%
\index{probability!absolute|)}

\subsection{Relative probabilities}\label{rel_probs}

\index{probability!relative|(}

The absolute probabilities of multiple alignments, calculated as described in the last subsection, are normally very small and not very interesting in themselves. From the standpoint of practical applications, we are normally interested in the {\em relative} values of probabilities, not their {\em absolute} values.

A point we may note in passing is that the calculation of relative probabilities from $p_{ABS}$ will tend to cancel out any general tendency for values of $p_{ABS}$ to be too high or too low. Any systematic bias in values of $p_{ABS}$ should not have much effect on the values which are of most interest to us.

If we are to compare one multiple alignment and its probability to another multiple alignment and its probability, {\em we need to compare like with like}. An multiple alignment can have a high value for $p_{ABS}$ because it encodes only one or two symbols from New. It is not reasonable to compare a multiple alignment like that to another multiple alignment which has a lower value for $p_{ABS}$ but which encodes more symbols from New. Consequently, the procedure for calculating relative values for probabilities ($p_{REL}$) is as follows:

\begin{enumerate}

\item For the multiple alignment which has the highest $CD$ (which we shall call the {\em reference multiple alignment}), identify the symbols from New which are encoded by the multiple alignment. We will call these symbols the {\em reference set of symbols in New}.

\item Compile a {\em reference set of multiple alignments} which includes {\em the multiple alignment with the highest $CD$ and all other multiple alignments (if any) which encode exactly the reference set of symbols from New, neither more nor less}.\footnote{There may be a case for defining the reference set of multiple alignments as those multiple alignments which encode the reference set of symbols {\em or any super-set of that set}. It is not clear at present which of those two definitions is to be preferred.}

\item The multiple alignments in the reference set are examined to find and remove any rows which are redundant in the sense that all the symbols appearing in a given row also appear in another row in the same order.\footnote{If Old is well compressed, this kind of redundancy amongst the rows of a multiple alignment should not appear very often.} Any multiple alignment which, after editing, matches another multiple alignment in the set is removed from the set.

\item Calculate the sum of the values for $p_{ABS}$ in the reference set of multiple alignments:

\begin{equation}
p_{A\_SUM} = \sum_{i = 1}^{i = R} p_{ABS_i}
\label{pA_SUM_equation}
\end{equation}

\noindent where $R$ is the size of the reference set of multiple alignments and $p_{ABS_i}$ is the value of $p_{ABS}$ for the $i$th multiple alignment in the reference set.

\item For each multiple alignment in the reference set, calculate its relative probability as:

\begin{equation}
p_{REL_i} = p_{ABS_i} / p_{A\_SUM}.
\label{pREL_equation}
\end{equation}

\end{enumerate}

The values of $p_{REL}$ calculated as just described seem to provide an effective means of comparing the multiple alignments in the reference set (normally, this will be those multiple alignments which encode the same set of symbols from New as the multiple alignment which has the best overall $CD$).

It is not necessary always to use the multiple alignment with the best $CD$ as the basis of the reference set of symbols. It may happen that some other set of symbols from New is the focus of interest. In this case a different reference set of multiple alignments may be constructed and relative values for those multiple alignments may be calculated as described above.

\subsection{Relative probabilities of patterns and symbols}\label{rel_probs_patts}

It often happens that a given pattern from Old or a given symbol type within patterns from Old appears in more than one of the multiple alignments in the reference set. In cases like these, one would expect the relative probability of the pattern or symbol type to be higher than if it appeared in only one multiple alignment. To take account of this kind of situation, SP61 calculates relative probabilities for individual patterns and symbol types in the following way:

\begin{enumerate}

\item Compile a set of patterns from Old, each of which appears at least once in the reference set of multiple alignments. No single pattern from Old should appear more than once in the set.

\item For each pattern, calculate a value for its relative probability as the sum of the $p_{REL}$ values for the multiple alignments in which it appears. If a pattern appears more than once in a multiple alignment, it is only counted once for that multiple alignment.

\item Compile a set of symbol types which appear anywhere in the patterns identified in step 2.

\item For each symbol type identified in step 3, calculate its relative probability as the sum of the relative probabilities of the patterns in which it appears. If it appears more than once in a given pattern, it is only counted once.

With regard to symbol types, the foregoing applies only to symbol types which do not appear in New. Any symbol type which appears in New necessarily has a probability of $1.0$---because it has been observed, not inferred.

\end{enumerate}

\index{probability!relative|)}

\subsection{Comparison of alignments that do not encode the same symbols from new}

It is true that, when we compare multiple alignments, we should compare like with like in terms of 
the symbols from New which are encoded by the multiple alignment. But, nevertheless, there may be 
occasions when we wish to compare multiple alignments that do not encode the same symbols from New. 

In cases like that, $CD$ or $CR$ can be used. It may be possible to develop a principled method for calculating probabilities of multiple alignments that occur different subsets of the symbols in New but this has not, so far, been investigated.%
\index{probability|)}

\section{Decompression by compression}\label{decompression_by_compression}

\index{information!compression!decompression by|(}\index{language!production|(}

As was described in Section \ref{ma_evaluation}, a `good' multiple alignment in the SP system is one in which the New pattern may be encoded economically in terms of the Old patterns in the multiple alignment.\footnote{As applied to multiple alignments and patterns, the words `bad' and `good' are shorthand for ``bad/good in terms of minimum length encoding principles and perhaps also in terms of one's intuitions about what is or is not an appropriate grammar for the data''. Quote marks will be dropped in the remainder of the book.} Given a good multiple alignment like the one shown in Figure \ref{alignment_figure_1} (a), the sentence `t h i s b o y l o v e s t h a t g i r l' may be encoded by the relatively short pattern `S 0 1 0 1 0 \#S'. The saving is even more pronounced if the lengths of these patterns are measured in bits rather than numbers of symbols.

Despite the fact that the SP system is dedicated to information compression, it provides a means of decompressing a code like `S 0 1 0 1 0 \#S' so that the original uncompressed sentence is recreated. Although this may seem like a paradox, there is no inconsistency or contradiction at all.

\begin{figure}[!hbt]
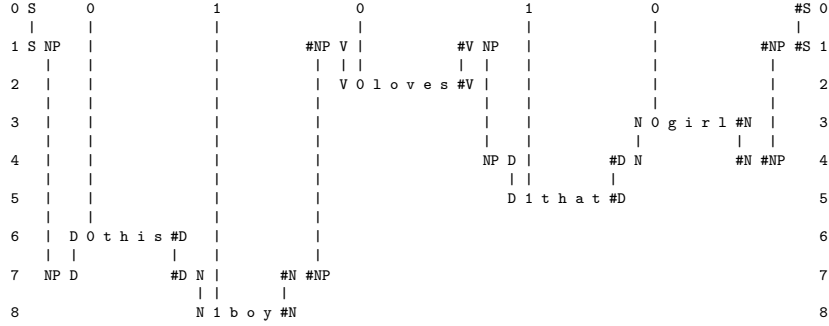

\fontsize{06.00pt}{07.20pt}
\centering
\begin{BVerbatim}
0 S      0              1                0                   1              0                #S 0
  |      |              |                |                   |              |                |   
1 S NP   |              |          #NP V |           #V NP   |              |            #NP #S 1
    |    |              |           |  | |           |  |    |              |             |      
2   |    |              |           |  V 0 l o v e s #V |    |              |             |     2
    |    |              |           |                   |    |              |             |      
3   |    |              |           |                   |    |            N 0 g i r l #N  |     3
    |    |              |           |                   |    |            |           |   |      
4   |    |              |           |                   NP D |         #D N           #N #NP    4
    |    |              |           |                      | |         |                         
5   |    |              |           |                      D 1 t h a t #D                       5
    |    |              |           |                    
6   |  D 0 t h i s #D   |           |                                                           6
    |  |           |    |           |                    
7   NP D           #D N |       #N #NP                                                          7
                      | |       |                        
8                     N 1 b o y #N                                                              8
\end{BVerbatim}
\caption{The best multiple alignment found by SP61 with `S 0 1 0 1 0 \#S' in New
and the grammar from Figure \ref{grammar_2} in Old.}
\label{alignment_figure_2}
\normalsize
\end{figure}

If Old contains the grammar shown in Figure \ref{grammar_2} (a) (the same grammar as was used to create the multiple alignment shown in Figure \ref{alignment_figure_1} (a)) and if New contains the code sequence `S 0 1 0 1 0 \#S', then the best multiple alignment found by the system is the one shown in Figure \ref{alignment_figure_2}. This multiple alignment contains the same words as the original sentence, in the same order as the original. In effect, it recreates the original sentence. This can be seen even more clearly if the patterns in the multiple alignment are unified to create a single sequence like this:

\begin{center}
\begin{BVerbatim}
S NP D 0 t h i s #D N 1 b o y #N #NP V 0 l o v e s #V
     NP D 1 t h a t #D N 0 g i r l #N #NP #S
\end{BVerbatim}
\end{center}

\noindent This is the same as the unified sequence derived from Figure \ref{alignment_figure_1} (a). If we strip out the symbols that represent grammatical categories, the sequence is exactly the same as the original sentence. Readers who are familiar with Prolog, will recognise that this process of recreating the original sentence is similar in some respects to the way in which an appropriately-constructed Prolog program may be run `backwards', deriving `data' from `results'.

How is it possible to decompress the compressed code for the original sentence by using information compression? How is the apparent paradox to be resolved or explained? The answer is  that when a code pattern such as `S 0 1 0 1 0 \#S' is introduced as New information, with the original grammar as Old information, there is normally a small but usable amount of redundancy between the New pattern and the Old patterns and it is the detection and extraction of this small amount of redundancy that allows the system to construct a multiple alignment that is the same as the original, except for the first row. If there was no detectable redundancy in the body of information represented by the combination of the New pattern with the Old patterns, then that body of information would be totally random and the process would not work.

Why should there be any redundancy between the code pattern and the Old patterns? If the Old patterns were fully compressed and if the original sentence were fully compressed in terms of the Old patterns then there should not be any redundancy in the body of information represented by code pattern and the Old patterns combined.

However, it is normally difficult to achieve such perfect compression. There are normally approximations and inefficiencies which mean that compression is less than perfect. And if the system is to be used in `reverse'---as in Figure \ref{alignment_figure_2}---it is important that there should be residual redundancies to allow the system to work.

\subsection{Evaluation of alignments and the cost factor}\label{evaluation_and_cost_factor}

If the multiple alignment shown in Figure \ref{alignment_figure_2} is evaluated using the method described in Section \ref{ma_evaluation_outline} we arrive at a curious result: the code derived from the multiple alignment contains no symbols and the value of $B_E$ calculated by Equation \ref{BE_equation} is 0!

At first sight, this is entirely reasonable. The original sentence was compressed to produce the relatively small code `S 0 1 0 1 0 \#S' and then, in the process of retrieval, the residual redundancy is further reduced to zero. But without any code from a multiple alignment like the one shown in Figure \ref{alignment_figure_2}, we have no record of the structure of the multiple alignment.

In the SP61 and SP70 models, a partial answer has been provided by the provision of the `cost factor' described in Section \ref{encoding_individual_symbols}, above. This causes symbols in the New pattern to behave as if they were relatively large chunks of information, each with its own relatively small code. It is rather as if `S 0 1 ...' had been converted into `Sentence nought one ...' with corresponding patterns added to Old such as `$<$ S Sentence $>$', `$<$ 0 nought $>$', `$<$ 1 one $>$', and adjustments to the original Old patterns so that `D 0 t h i s \#D' becomes `D $<$ 0 $>$ t h i s \#D', and so on.

If the effective sizes of New symbols are boosted in this way, then it would make sense for the multiple alignment in Figure \ref{alignment_figure_2} to yield `S 0 1 0 1 0 \#S' as the code for the boosted version of that pattern in New. So far, there has been no pressing need to make provision for this in the SP models but this would be a relatively simple thing to do, if required.

Although there is not currently any particular need for the cost factor in this connection, the facility has been retained in both models because there are other situations where it is useful to be able to treat New symbols as if they were relatively large chunks of information.%
\index{information!compression!decompression by|)}\index{language!production|)}

\section{Outline of computer models}\label{computer_models_outline}

This section describes the SP61 model and the SP70 model in outline and the next section describes SP61 in more detail. A full description of SP70 is given in Chapter \ref{learning_chapter}.

\subsection{Outline of SP61}\label{sp61_outline}

\index{SP61|(}

The SP61 model requires the user to supply a set of Old patterns and one New pattern. After these have been read in, the model builds multiple alignments in a pairwise manner, starting with the original patterns. At each stage, multiple alignments that contain mismatches between Old patterns are rejected. This means that each of the multiple alignments that are not rejected can be treated as if it was a single sequence of symbols, as described in Section \ref{mismatches_section}, above.

On each cycle, the operations are as follows:

\begin{enumerate}

\item Identify a set of `driving' patterns and a set of `target' patterns. On the first cycle, the New pattern is the sole driving pattern and the Old patterns are the target patterns. On all subsequent cycles, the best of the multiple alignments formed so far (in terms of their $CD$ scores) are chosen to be driving patterns and the target patterns are the Old patterns together with a larger selection of the best multiple alignments formed so far that includes all of those that are driving patterns. 

\item Compare each driving pattern with each of the target patterns to find full matches and good partial matches between patterns. This is done with a process that is essentially a form of `dynamic programming' \citep[see, for example,][]{sankoff_kruskall_1983}. A fairly full description of how the process works is given in Appendix \ref{matching_appendix}.

\item From the best of the matches found in the current cycle, create corresponding multiple alignments and add them to the repository of multiple alignments created by the program.

\end{enumerate}

This process of matching driving patterns against target patterns and building multiple alignments is repeated until no more multiple alignments can be found. The best of the multiple alignments created since the start of processing are printed out and probabilities are calculated, as described in Section \ref{probabilities_section}.%
\index{SP61|)}

\subsection{Outline of SP70}\label{sp70_outline}

\index{SP70|(}

The SP70 model realises all elements of the SP system, as described in Section \ref{overall_framework}. Initially, the repository of Old patterns is empty and New contains a set of patterns representing `raw' data about the world.

The New patterns are processed one at a time in the following way:

\begin{enumerate}

\item Using the same processes as in SP61, a set of multiple alignments is built, each one containing the current pattern from New and one or more patterns from Old.

\item At the same time, the New pattern is transferred to Old and ID-symbols are added to it. The transfer is done in such a way that the New pattern can be aligned with itself with the restrictions described in Section \ref{multiple_appearances}. 

\item From amongst the best multiple alignments formed, derive new patterns as described below and add them to Old.

\end{enumerate}

To clarify the last point, SP70 processes a multiple alignment like this:

\begin{center}
\begin{BVerbatim}
0        t h e t a l l t r e e   0
         | | |         | | | |  
1 < %1 0 t h e b i g   t r e e > 1
\end{BVerbatim}
\end{center}

\noindent to create Old patterns like these:

\begin{center}
\begin{BVerbatim}
< %2 0 t h e >
< %3 0 t r e e >
< %4 0 b i g >
< %4 1 t a l l >
< %5 0 < %2 > < %4 > < %3 > >.
\end{BVerbatim}
\end{center}

\noindent The first four of these patterns correspond to coherent substrings of matched symbols or unmatched symbols in the multiple alignment, each substring with ID-symbols that are added by the system. The last pattern is created by the system to represent the sequence of substrings in the multiple alignment. Notice that `b i g' and `t a l l' are alternatives in this sequence because they have the same `class' symbol (`\%4').

Many of the good multiple alignments are not as tidy as the one just shown and many of the patterns derived from those multiple alignments do not coincide neatly with words or other `correct' grammatical structures. However, SP70 continues to process all the patterns from New in this way and accumulates a relatively large collection of Old patterns, some of them good and many of them bad.

At this stage, the program performs a heuristic search through the space of possible subsets of the patterns in Old, aiming to discover subsets that minimise $T$ in accordance with minimum length encoding principles as described in Section \ref{mle_section}. The best subset or the best few subsets may be retained and all the other bad patterns may be discarded. It is envisaged that the program will be developed so that New patterns are processed in batches, with a purging of bad patterns at the end of each batch.%
\index{SP70|)}

\section{Detailed description of SP61}\label{sp61_detail_section}

\index{SP61|(}

This section describes the SP61 model in more detail, starting with its overall organisation.

\subsection{Organisation of the SP61 model}

Figure \ref{sp61_structure} presents a high-level view of the
organisation of the SP61 model using pseudocode while Figure
\ref{compress_figure} shows the {\em compress()} function
within the model. The text below describes how the model works together
with details of its organisation that are not included in the
pseudocode.

\begin{figure}[!hbt]
\fontsize{08.00pt}{09.60pt}
\centering
\begin{BVerbatim}
SP61()
{
     1 Read the 'New' pattern and store it in New.
     2 Read the set of 'Old' patterns, each one with a frequency
          of occurrence in a notional sample of the world.
          Store the patterns with their frequencies in Old.
     3 Derive a frequency for each symbol type in New and Old and
          calculate a code size for each symbol type.
     4 To each symbol in New and Old, assign the code size of its
          symbol type. For each symbol in New, multiply the code
          size by the 'cost factor'.
     5 Select the New pattern and add it as the first
          'driving pattern' to an otherwise empty set of
          driving patterns. Assign the Old patterns to
          an initial set of 'target patterns'.
     6 while (new multiple alignments are being formed) 
          COMPRESS()
     7 Out of all the multiple alignments that have been formed,
          print the ones with the best CDs. Calculate
          and print relative probabilities of multiple alignments,
          patterns and symbols.
}
\end{BVerbatim}
\caption{A high level view of the organisation of the SP61 model.}
\label{sp61_structure}
\normalsize
\end{figure}

\begin{figure}[!hbt]
\fontsize{07.00pt}{08.40pt}
\centering
\begin{BVerbatim}
COMPRESS()
{
     1 Clear the 'hit structure' (described in the text).
     2 while (there are driving patterns that have not yet been processed)
     {
          2.1 Select the first or next driving pattern in the set of
               driving patterns.
          2.2 while (there are more symbols in the current driving pattern)
          {
               2.2.1 Working left to right through the current driving
                    pattern, select the first or next symbol in the pattern.
               2.2.2 'Broadcast' this symbol to make a yes/no match with
                    every symbol in the set of 'target patterns'.
               2.2.3 Record each positive match (hit) in the hit structure.
                    As more symbols are broadcast, the hit structure builds
                    up a record of sequences of hits between the driving
                    pattern and the several target patterns. As each hit
                    sequence is extended, the compression score of the
                    corresponding multiple alignment is estimated using a
                    `cheap to compute' method of estimation.
               2.2.4 If the space allocated for the hit structure is filled
                    at any time, the system 'purges' the worst 50% of the
                    hit sequences from the hit structure to release more
                    space. The selection uses the estimates of compression
                    scores assigned to each hit sequence in Step 2.2.3.
          }
     }

     3 For each hit sequences that has an estimated compression score above
          some threshold value and which will 'project' into a single
          sequence (as described in the text), convert the hit sequence
          into the corresponding multiple alignment. Discard this multiple alignment
          if it is identical with any multiple alignment already in Old. Otherwise,
          compute the CD value for the multiple alignment and its absolute probability.
          If no new multiple alignments are formed, quit the COMPRESS() function.
     4 Examine all the multiple alignments that have been created since the beginning
          of processing and choose a subset of these multiple alignments using the
          method described in the text. Discard all the multiple alignments that have
          not been selected.
     5 Clear the list of driving patterns and then, using the same method as
          is used in 4 but (usually) with a more restrictive parameter,
          select a subset of the multiple alignments chosen in 4 to be the new list
          of driving patterns.
}
\end{BVerbatim}
\caption{The organisation of the {\em compress()} function of the SP61 model.}
\label{compress_figure}
\normalsize
\end{figure}

\subsection{Preliminary processing}

\subsubsection{Calculating the information cost of each symbol}

Each pattern in Old has an associated frequency of occurrence in some domain. In Step 3 of {\em
SP61()} in Figure \ref{sp61_structure}, the model derives the frequency
of occurrence of each symbol type as described in Appendix \ref{ma_evaluation}.

These frequencies are then used (in Step 3 of {\em SP61()}) to calculate the
minimum number of bits needed to represent each symbol type using the
Shannon-Fano-Elias coding scheme (see \cite{cover_thomas_1991}), as described in Section \ref{encoding_individual_symbols}. The resulting sizes for each symbol type are then assigned to the corresponding symbols in New and Old. The code size of each symbol in New is multiplied by the `cost factor', as discussed in Section \ref{evaluation_and_cost_factor}.

\subsection{Building the hit structure}

The {\em compress()} function shown in Figure \ref{compress_figure} is the heart
of the SP61 model. This subsection and the ones that follow supplement
the description in the figure.

As can be seen from the figure and the outline description in Section \ref{sp61_outline},
the {\em compress()} function is applied iteratively. On the first cycle, the `driving'
pattern is simply the New pattern. On subsequent cycles, the
list of driving patterns is a subset of the multiple alignments formed in
preceding cycles. Iteration stops when no new multiple alignments can be found
that satisfy conditions described below.

\subsubsection{Fuzzy matching of one pattern with another}\label{matching_one_pattern_with_another}

Step 2 of the {\em compress()} function is based on the central
process in SP21. It is a form of dynamic programming \citep{wagner_fischer_1974, sankoff_kruskall_1983} designed, like other dynamic programming algorithms, to find full matches and good partial matches between one sequence of symbols and another. This process, described in Appendix \ref{matching_appendix} and \citet{wolff_1994_scaleable}, has three main advantages compared with `standard' forms of dynamic programming:

\begin{itemize}

\item It can match arbitrarily long sequences without excessive demands on memory.

\item For any two sequences, it can find a set of alternative matches (each with a measure of how good it is) instead of a single `best' match.

\item The `depth' or thoroughness of the searching can be controlled by parameters.

\end{itemize}

Taking the driving patterns one by one, the process `broadcasts' each symbol in each driving pattern to make a `same' or `different' match with each symbol in the set of target patterns. Single hits and sequences of hits are recorded in a {\em hit structure}. Each {\em hit sequence} represents a multiple alignment between the driving pattern and one of the target patterns.

As is described in Appendix \ref{matching_appendix} and \cite{wolff_1994_scaleable}, the hit structure has the form of a list-processing tree with each node representing a hit and each path
from the root to a leaf node representing a hit sequence.

\subsubsection{No one instance of a symbol should ever be matched with
itself}\label{no_self_matching}

Since any driving pattern can also be a target pattern, any one pattern may
be aligned with itself. That being so, a check is made to ensure that
no instance of a symbol is ever matched against itself (Section
\ref{multiple_appearances}).

Since any symbol in the driving pattern and any symbol in the target
pattern may have been derived by the unification of two or more other
symbols, a check is also made to exclude all hits where the set of
symbols from which one of the hit symbols was derived has one or more
symbols in common with the set of symbols from which the other hit
symbol was derived. In short, while any given pattern from the grammar
may appear two or more times in one multiple alignment, no symbol in any of the
original patterns in Old ever appears in the same column as itself in
any multiple alignment.

\subsubsection{The order of symbols in new must be preserved}\label{preserve_symbol_order_in_new}

As the matching process has been described so far, it would be entirely
possible for the system to align a pattern like `NP D 1 t h a t \#D N 1
g i r l \#N \#NP' in the example considered in Section \ref{framework_examples_section}, with the first `NP \#NP' in the `sentence pattern, `S NP \#NP V 0 l o v e s \#V NP \#NP \#S', and to align `NP D 0 t h i s \#D N 0 b o y \#D \#NP' with the second `NP \#NP' in that pattern. To avoid the formation of multiple alignments like this which violate the order of the symbols in New, the system makes checks to
ensure that, at all stages, the order of the symbols in New is honoured.

\subsubsection{Estimation of compression scores}

While the hit structure is being built, the compression score for the
multiple alignment corresponding to each hit sequence may be calculated at every
stage but only at the cost of a lot of processing which would slow the
model down. Consequently, a simple method of estimating the compression
score is used in Step 2.2.3 of Figure \ref{compress_figure} which is
computationally `cheap'. Although it gives results that do not
correspond exactly with the values calculated using the method described in Section \ref{ma_evaluation}, the differences appear not to be critical for the purposes of purging the hit structure (Step 2.2.4 in Figure \ref{compress_figure} and Section
\ref{purging_of_hit_structure}) or determining the threshold for
converting hit sequences into multiple alignments (Step 3 in
Figure \ref{compress_figure} and Section \ref{creating_alignmments}).

\subsubsection{Purging the hit structure}\label{purging_of_hit_structure}

If the space allocated to the hit structure is exhausted at any time,
the hit structure is `purged' or, more literally, `pruned' to remove
branches corresponding to the worst 50\% of the hit sequences (where the
meaning of `worst' is determined using the estimates of compression
scores calculated in Step 2.2.3 of the {\em compress()} function). In this
way, space is released in which new sequences of hits can be stored.

\subsection{Building, scoring and selecting alignments}\label{creating_alignmments}

\subsubsection{Building alignments and scoring them}

When the hit structure for a set of driving patterns has been built,
the best hit sequences are converted into the corresponding multiple alignments,
excluding all multiple alignments which will not `project' on to a single
sequence (as described in Section \ref{sp61_outline}).

The process of converting a hit sequence into a multiple alignment achieves two
things: it creates a one-dimensional sequence of symbols which is a
unification of the driving pattern or patterns with the target pattern
and it creates a list structure in two dimensions representing the multiple alignment
itself. For each multiple alignment, the structure occupies a portion of memory of
exactly the right size, allocated dynamically at the time the multiple alignment
is formed.

The one-dimensional sequence may enter into matching and unification in
later iterations of the {\em compress()} function, while the two-dimensional
list structure allows the full structure of the multiple alignment to be seen. This structure can be
used in later checks to ensure that no instance of a symbol is ever
matched with itself (Section \ref{no_self_matching}) and to ensure that
the order of symbols in New is not violated (Section
\ref{preserve_symbol_order_in_new}).

From time to time, identical multiple alignments are formed via different
routes. The program checks each of the newly-formed multiple alignments against
multiple alignments already formed. Any multiple alignment which duplicates one already
formed is discarded. The process of comparing multiple alignments is indifferent
to the order (from top to bottom) in which patterns appear in the
multiple alignment ({\em cf.} Section \ref{parsing_as_alignment}, above).

Every new multiple alignment which survives the several hurdles is added to a store of multiple alignments and its $CD$ score is computed as described in Section \ref{ma_evaluation}.

\subsubsection{Selection of alignments: a quota for each hit symbol in
New}\label{selection_of_alignments}

Apart from purging the hit structure when space is exhausted, the
main way in which the SP61 model narrows its search space is a
two-fold selection of multiple alignments at the end of every cycle of the
{\em compress()} function:

\begin{itemize}

\item The program examines the multiple alignments created since the start of
processing and selects a subset by a method to be described. All the
other multiple alignments are discarded.

\item Using the same method, the program selects a subset of the
remaining multiple alignments to be used as driving patterns on the
next cycle. These multiple alignments also function as target patterns.

\end{itemize}

At first sight it seems natural to select multiple alignments purely on the
basis of their compression scores. However, it can easily happen that,
at intermediate stages in processing, the best multiple alignments are trivial
variations of each other and involve the same subset of the symbols
from New. If selection is made by choosing multiple alignments with a $CD$ above a
certain threshold, the multiple alignments which are chosen may all involve the
same subset of the symbols in New, while other multiple alignments, containing
symbols from other parts of New, may be lost. If this happens, the
model cannot ever build a multiple alignment which contains all or most the
symbols in New and may thus never find the `correct' answer.

A solution to this problem which seems to work well is to make
selections in relation to the symbols in New which appear in the
multiple alignments. Each symbol in New is assigned a {\em quota} (the same for all
symbols) and, for each symbol, the best multiple alignments up to the quota are
identified. Any multiple alignment which appears in one or more of the quotas is
preserved. All other multiple alignments are purged. The merit of this
technique is that it can `protect' any multiple alignment which is the best
multiple alignment for a given subsequence of the symbols in New (or is second
or third best etc) but which may, nevertheless, have a relatively low
$CD$ compared with other multiple alignments in Old.

\subsubsection{Processing New in stages}\label{windows_section}

A feature of the SP61 model that has been omitted
from Figure \ref{sp61_structure} (to avoid clutter) is that New may be divided into
`windows' of any fixed size (determined by the user) and the model can
be set to process New in stages, one window at a time, from left to
right. This feature of the model was introduced for two reasons:

\begin{itemize}

\item It seems to bring the model closer to the way people seem to operate,
processing sentences stage by stage as they are heard or read, not waiting
until the whole of a sentence has been seen before attempting to analyse
it.

\item Since it is possible to discard all but the best intermediate results
at the end of each window, this mode of processing has the advantage of
reducing peak demands for storage of information and it also has the effect
of reducing the size of the search space.

\end{itemize}

\subsection{Calculation of probabilities}

As indicated in Figure \ref{compress_figure} (step 3), SP61 calculates an absolute probability for each multiple alignment; the method of calculation is described in Section \ref{absolute-probs}. When no more multiple alignments can be found, the program derives a reference set of best multiple alignments and calculates relative probabilities and probabilities for symbols and patterns as described in Sections \ref{rel_probs} and \ref{rel_probs_patts} (step 7 of Figure \ref{sp61_structure}).

\subsection{Computational complexity}\label{sp61_computational_complexity}

\index{complexity, computational|(}

Given that many of the examples in this book are fairly small, and given the well-known computational demands of multiple alignment problems, readers may reasonably ask whether the SP61 model (and the SP70 model) would scale up to handle realistically large amounts of knowledge. This section considers the time complexity and space complexity of the SP61 model and Section \ref{sp70_computational_complexity} considers the same aspects of the SP70 model.

\subsubsection{Matching patterns}

The critical operation at the heart of the SP61 model is the process of comparing one pattern with another to find full matches and good partial matches (Appendix \ref{matching_appendix}). As described in Section \ref{sp21_computational_complexity}, the time complexity of this process in a serial processing environment is estimated to be O$(n \cdot m)$, where $n$ is the number of symbols in one pattern and $m$ is the number of symbols in the other. In a parallel processing environment, the time complexity may approach O$(n)$, depending on how well the parallelism is applied. In serial or parallel processing environments, the space complexity is estimated to be O$(m)$.

It appears that the big O values just given will also apply when a {\em set} of one or more patterns is to be compared with another set of one or more patterns. In this case, $n$ is the number of symbols in the first set and $m$ is the total number of symbols in the second. 

\subsubsection{The complete process}

In SP61, multiple alignments are built by pairwise alignment of patterns and multiple alignments. The 
number of such pairings required to build a multiple alignment for a whole sentence appears to be 
approximately:

\begin{equation}
P = \log_2 (s_N / c),
\label{pairings_equation}
\end{equation}

\noindent where $s_N$ is the number of symbols in the New pattern and $c$ is a constant.

At any given stage of processing, the value of $n$ is the total number of symbols in the driving patterns and the value of $m$ is:

\begin{equation}
m = s_O + s_A,
\label{m_equation}
\end{equation}

\noindent where $s_O$ is the total number of symbols in the Old patterns and $s_A$ is the total number of symbols in the repository of multiple alignments, assuming that each multiple alignment is treated as a single sequence of symbols.

The values of $n$ and $m$ for successive pairings will both vary as processing proceeds. At the beginning, when the New pattern is the sole driving pattern, $n = s_N$, but the value of $n$ will increase at later stages when there are two or more driving patterns. Likewise, the value of $m$ will increase when multiple alignments have been created, with a corresponding increase in $s_A$. However, the selection process described in Section \ref{selection_of_alignments} ensures that, after the initial growth, the sizes of $n$ and $m$ will remain relatively constant. It is true that multiple alignments in the later stages are longer than in the early stages but there are fewer of them. For the purpose of assessing computational complexity, it seems reasonable to assume that variations in the sizes of $n$ and $m$ from the start to finish of processing will be approximately constant and can therefore be discounted.

Overall, the time complexity of SP61 in a serial processing environment should be 
approximately O$(\log_2 s_{N} \times s_{N}s_{O})$, where, as before, $s_N$ is the number of symbols in the New pattern and $s_O$ is the total number of symbols in the patterns in Old. In a parallel processing environment, the time complexity may approach O$(\log_2 s_N \times s_N)$, depending on how well the parallel processing is applied. The space complexity in serial or parallel environments should be O$(s_O)$.

For most applications, there will normally be a relatively small maximum value of $s_N$. If we regard this as a constant, then the value of $s_N$ may be discounted. Accordingly, the time complexity of SP61 in a serial processing environment will be O$(s_O)$ and its time complexity in a parallel processing environment may approach O(1). The space complexity will be unchanged as O$(s_O)$.

In summary, the time complexity of SP61 in serial and parallel processing environments is polynomial and well within acceptable limits.%
\index{complexity, computational|)}

\subsection{The use of angle brackets in Old patterns}\label{angle_brackets_constraint}

An aspect of SP61 that has not so far been mentioned is that, if angle brackets (`$<$' and `$>$')are used in any of the Old patterns supplied to the model, they are always treated as ID-symbols and, more importantly, the model is constrained to accept only `legal' matches between brackets---meaning that, if two left angle brackets have been matched then their corresponding right angle brackets {\em must} be matched, and {\em vice versa}.

Without this constraint, the program can waste a lot of time forming multiple alignments that will not ultimately yield economical encodings of New. The constraint may be justified on the grounds that, in the fully-developed SP system (and the SP70 model), angle brackets (with other ID-symbols) are supplied by the system as part of the process of encoding raw data so there is no need to consider possible multiple alignments that are not consistent with that encoding.

When angle brackets are used in Old patterns, this constraint speeds up processing. But, given the current design of the hit structure and the way it is updated, the application of the constraint consumes computational resources. This means that the program runs even faster if knowledge is represented in the style shown in Figure \ref{grammar_2} (a) and angle brackets are not used at all.

\subsection{Searching and constraints in SP61}\label{searching_and_constraints_in_sp61}

So far, SP61 has been described in relatively concrete terms without much reference to the way it realises the principles of search and constraint described earlier in Sections \ref{matching_searching_and_constraints} and \ref{ma_searching_and_constraints}.

Searching in the model is embodied in the way it matches patterns (Appendix \ref{matching_appendix} and step 2 of the {\em compress()} function
in Figure \ref{compress_figure}) and builds alternative multiple alignments (step 3 of Figure \ref{compress_figure}).

Heuristic constraints are applied as follows:

\begin{itemize}

\item If the memory available for the hit structure is filled at any time, hit sequences are sorted by their compression score and the worst 50\% are discarded thus making space for more hit sequences (Step 2.2.4 in Figure \ref{compress_figure} and Section \ref{purging_of_hit_structure}). This purging of memory, which may be repeated many times as the program runs, is a simple but effective way of pruning away large parts of the search space.

The amount of memory available for storing the hit structure can be set as a parameter for the program. If the amount of memory available is small, the program will normally run faster but it may fail to find some good alignments of patterns. With more memory available, a `deeper' search can be performed but this normally takes more time.

\item In Step 4 of the {\em compress()} function, a selection of the multiple alignments formed since the start of processing are chosen to be target patterns for the next cycle, as described in Section \ref{selection_of_alignments}. The size of the quota for target patterns (described in Section \ref{selection_of_alignments}) is set as a parameter to the program. It is normally about 10 or 20.

\item In Step 5, a subset of the multiple alignments chosen in Step 4 are selected by the same method as driving patterns for the next cycle. The size of the quota for driving patterns is set as a parameter to the program and it is normally about 3 or 5.

\end{itemize}

Absolute constraints that always operate in the model are these:

\begin{itemize}

\item As described in Section \ref{angle_brackets_constraint}, the matching of angle brackets must be `legal'. Multiple alignments containing angle brackets that are not matched correctly are discarded.

\item All multiple alignments containing mismatches between Old patterns are always discarded (Section \ref{mismatches_section}).

\end{itemize}

Absolute constraints that may be applied optionally are these:

\begin{itemize}

\item As described in Section \ref{windows_section}, it is possible to process the New pattern in a succession of windows. This reduces the range of alternative alignments that need to be considered at any one time. The size of the window may be set as a parameter to the program. Normally, this value is set to be large so that the New pattern is processed in a single window.

\item It is possible to limit the size of any unmatched gap between two hit symbols in a New pattern, and to set similar maxima for unmatched gaps in Old patterns, driving patterns and target patterns. These parameters of the program are almost always set to be very large so that, in effect, there is no constraint at all.

\item With certain combinations of New and Old patterns, compression scores for multiple alignments may rise to a maximum as processing proceeds and then gradually fall while the program continues to produce multiple alignments that are worse than ones that it has already created. In cases like this, it is convenient to set a maximum of about 5 or 10 for the number of cycles of the {\em compress()} function that produce multiple alignments with compression scores that are lower than those of the best multiple alignments produced since the program was started.

\item With certain kinds of application, the program may be constrained so that ID-symbols are only ever matched with C-symbols and {\em vice versa}. This can speed up processing but it does not normally make any difference to the final `best' multiple alignments delivered by the program.

\item There is a limit to the number of multiple alignments produced on any one cycle of the {\em compress()} function. Normally, this parameter is given a high value so that, in effect, there is no limit at all.

\end{itemize}

\index{SP61|)}

\section{Conclusion}

This completes the description of the SP system, except for the description of how the system has been developed for unsupervised learning, described in Chapter \ref{learning_chapter}. Apart from that chapter, most of the examples and discussion in this book will relate to the SP system as it has been realised in the SP61 model.

%% file: computing.tex
\chapter[The SP Theory as a Theory of Computing]{The SP Theory as a Theory of Computing\protect\footnote{Based on \citet{wolff_1999_comp}.}}%
\label{computing_chapter}

\index{computing|(}

\section{Introduction}

The nature of `computing' was the focus of much interest in the 1930s and early '40s but, since then, it has been widely accepted that the essentials of this concept have been captured in Alan Turing's `Universal Turing Machine' \citet{turing_1936}\index{Turing machine} and that other models of computing (such as `Lambda Calculus'\index{lamda calculus} (Church and Kleene, see \citet{rosser_1984}), `Recursive Function'\index{recursive function} \citep{kleene_1936}, `Normal Algorithm'\index{normal algorithm} \citep{markov_nagorny_1988} and Post's `Canonical System'\index{Post canonical system} \citep{post_1943}) are equivalent.

In this chapter I will show that the operation of a Post canonical system---and, consequently, the operation of a universal Turing machine---may be interpreted within the SP system, and I will argue that the SP theory is not merely another variant on ideas that were developed 60 or more years ago but that it provides new insights into the nature of computing and artificial intelligence, and new opportunities for rationalisation and integration in the design of computing systems.

In what follows, Section \ref{utm_pcs_section} describes, briefly, the concept of a universal Turing machine and, more fully, the equivalent concept of a Post canonical system; Section \ref{sp_and_pcs} shows how the operation of a Post canonical system may be modelled within the SP system; and Section \ref{computing_discussion} discusses the relationship between the SP theory and earlier models of computing, the scope of the theory, and its implications for the design of practical computing systems.

\section{Universal Turing machine and Post canonical system}\label{utm_pcs_section}

\index{Turing machine|(}\index{Post canonical system|(}

In its standard form, a universal Turing machine comprises a `tape' on which symbols may appear in sequence, a `read-write head' and a `transition function' \citep{turing_1936}. In each of a sequence of steps, a symbol is read from the tape at the current position of the head and then, depending on that symbol, on the transition function and the current `state' of the head, a new symbol is written on the tape, the state of the head is changed, and the head moves one position left or right.

A more detailed description of universal Turing machines is not needed here because the arguments to be presented take advantage of the fact \citep[chapters 10 to 14]{minsky_1967} that the operation of any universal Turing machine can be modelled with a Post canonical system \citep{post_1943}. In this section, I shall describe the workings of a Post canonical system and then, in Section \ref{sp_and_pcs}, I shall try to show how the operation of a Post canonical system, and thus a universal Turing machine, may be understood in terms of the SP theory.

\subsection{The structure of a Post canonical system}\label{structure_pcs_section}

A Post canonical system comprises:

\begin{itemize}

\item An {\em alphabet} of primitive {\em symbols} (`letters' in Post's terminology),

\item One or more {\em primitive assertions} or {\em axioms}. These can often be regarded as `input'.

\item One or more {\em productions} which can often be regarded as a `program'.

\item Although this is not normally included in formal definitions of the Post canonical system concept, it is clear from how the system is intended to work (see Section \ref{how_pcs_works} below), that the Post canonical system also includes mechanisms for {\em matching patterns} and for {\em searching} for matches between leading substrings in the input pattern and the left-hand ends of the productions.

\end{itemize}

Each production has this general form:\footnote{Spaces between symbols here and in other examples have been inserted for the sake of readability and because it allows us to use atomic symbols where each one comprises a string of two or more characters (with spaces showing the start and finish of each string). Otherwise, spaces may be ignored.}

\[g_0\ \$_1\ g_1\ \$_2\ ...\ \$_n\ g_n \rightarrow  h_0\ \$^{\prime}_1\ h_1\ \$^{\prime}_2\ ...\ \$^{\prime}_m\ h_m\]

\noindent where ``Each $g_i$ and $h_i$ is a certain fixed string; $g_0$ and $g_n$ are often null, and some of the $h$'s can be null. Each $\$_i$ is an `arbitrary' or `variable' string, which can be null. Each $\$^{\prime}_i$ is to be replaced by a certain one of the $\$_i$.'' \citep[pp. 230--231]{minsky_1967}.

In its simplest `normal' form, a Post canonical system has one primitive assertion and each production has the form:

\[g\ \$ \rightarrow  \$\ h\]

\noindent where $g$ and $h$ each represent a string of zero or more symbols, and both instances of `\$' represent a single `variable' which may have a `value' comprising a string of zero or more symbols.

It has been proved \citep{post_1943} that any kind of Post canonical system can be reduced to a Post canonical system in normal form \citep[see also chapter 13 in][]{minsky_1967}. That being so, a Post canonical system in this form will be the main (but not exclusive) focus of our attention.

\subsection{How the Post canonical system works}\label{how_pcs_works}

When a Post canonical system (in normal form) processes an `input' string, the first step is to find a match between that string and the left-hand side of one of the productions in the given set. The input string matches the left hand side of a production if a match can be found between leading symbols of the input string and the fixed string (if any) at the start of that left-hand side, with the assignment of any trailing substring within the input string to the variable within the left-hand side of the production.

Consider, for example, a Post canonical system comprising the alphabet `a ... z', an axiom or input string `a b c b t', and productions in normal form like this:

\begin{center}
\begin{tabular}{l}
a \$ $\rightarrow$ \$ a \\
b \$ $\rightarrow$ \$ b \\
c \$ $\rightarrow$ \$ c
\end{tabular}
\end{center}

In this example, the first symbol of the input string matches the first symbol in the first production, while the trailing `b c b t' is considered to match the variable which takes that string as its value. The result of a successful match like this is that a new string is created in accordance with the configuration on the right hand side of the production which has been matched. In the example, the new string would have the form `b c b t a', derived from `b c b t' in the variable and `a' which follows the variable on the right hand side of the production.

After the first step, the new string is treated as new input which is processed in exactly the same way as before. In this example, the first symbol of `b c d t a' matches the first symbol of the second production, the variable in that production takes `c d t a' as its value and the result is the string `c b t a b'.

This cycle is repeated until matching fails. It should be clear from this example that the effect would be to `rotate' the original string until it has the form `t a b c b'. The `t' which was at the end of the string when processing started has been brought round to the front of the string. This is an example of the ``rotation trick'' used by \cite[Chapter 13]{minsky_1967} in demonstrating how a Post canonical system in normal form can model any kind of Post canonical system.

With some combinations of alphabet, input and productions, the process of matching strings to productions never terminates. With some combinations of alphabet, input and productions, the system may follow two or more `paths' to two or more different `conclusions' or may reach a given conclusion by two or more different routes. The `output' of the computation is the set of strings created as the process proceeds.

\subsection{Other examples}\label{other_computing_examples}

For readers unfamiliar with this kind of system, two other examples are included here to show more clearly how a Post canonical system may achieve `computing'. These examples are based on examples in \cite[Chapter 12]{minsky_1967}. Other examples may be found in the same chapter.

\subsubsection{Creation and recognition of numbers in unary notation}\label{generating_unary_numbers}

\index{unary numbers|(}

In the unary number system, 0 = 0, 1 = 01, 2 = 011, 3 = 0111, and so on. The nature of the unary number system can be described with a Post canonical system like this:

\begin{itemize}
\item Alphabet: the symbols 0 and 1.
\item Axiom: 0.
\item Production: If any string `\$' is a number, then so is the string `\$ 1'. 
This can be expressed with the production:
\end{itemize}

\[\$ \rightarrow \$\ 1\]

It should be clear from the description of how Post canonical systems work, above, that this Post canonical system is recursive\index{recursion!in Post canonical system} and that it can be used to create the infinite series of unary strings: `0', `0 1', `0 1 1', `0 1 1 1', `0 1 1 1 1' etc, as far as resources allow.

Slightly less obviously, the Post canonical system can also be used to recognise a string of symbols as being an example of a unary number. This is done by using the production in `reverse', matching a character string to the right hand side of the production, taking the left hand side as the `output' and then repeating the right-to-left process until only the axiom will match.%
\index{unary numbers|)}

\subsubsection{Creation and recognition of palindromes}\label{generation_of_palindromes}

\index{palindromes|(}

To quote \citet[p. 228]{minsky_1967}: ``The `palindromes' are the strings that read the same backwards and forwards, like {\em cabac} or {\em abcbbcba}. Clearly, if we have a string that is already a palindrome, it will remain so if we add the same letter to the beginning and end. Also, clearly, we can obtain all palindromes by building in this way out from the middle.'' Here is his example:

\begin{itemize}
\item Alphabet: a, b, c.
\item Axioms: a, b, c, aa, bb, cc.
\item Productions:
\end{itemize}

\begin{center}
\begin{tabular}{l}
\$ $\rightarrow$ a \$ a \\
\$ $\rightarrow$ b \$ b \\
\$ $\rightarrow$ c \$ b \\
\end{tabular}
\end{center}

Although this example is not in normal form, it should be clear how it generates all the palindromes for the given alphabet.\footnote{Here and elsewhere in this book, the word {\em generate} will normally be used in its `linguistic' sense to mean ``define precisely'' rather than ``create'' or ``produce''. A generative definition is or should be neutral between the production or recognition of patterns of symbols.} As with the previous example, this Post canonical system may be used to produce a palindrome and it may also be used to recognise whether or not a string of characters is a palindrome by using the productions in `reverse'.%
\index{Turing machine|)}%
\index{palindromes|)}

\section{SP and the operation of a Post canonical system}\label{sp_and_pcs}

It is not hard to see that a successful match between a leading substring in an input string and the leading substring in a Post canonical system production may be interpreted in terms of the alignment and unification of patterns, as described in Chapter \ref{theory_chapter}. What is less obvious is how concepts of alignment and unification might be applied to the processes of matching a trailing string of symbols to the variable, assigning it as a value to the variable and incorporating it in an output string.

\subsection{Modelling one step of the `rotation trick'}\label{sp_rotation_trick}

In the SP system, the first Post canonical system described above (for the `rotation' of a string) may be modelled using the patterns shown in Figure \ref{rotation_trick_patterns}. Regarding these patterns:

\begin{itemize}

\item The one under the heading `New' corresponds to the axiom or `input' in the Post canonical system.

\item The four patterns on the left under the heading `Old' may be seen as a representation of the Post canonical system alphabet. 

\item The first three patterns on the right correspond to the three productions of the Post canonical system. As we shall see, the pair of symbols, `X \#X', in each pattern function as if they were a `variable'\index{variable}. There is no rewrite arrow (`$\rightarrow$') and only one instance of the `variable'.

\item As we shall see, the last pattern on the right (`X L \#L X \#X \#X') is, in effect, a recursive definition of the concept of `any string of the letters in the given alphabet'. This is comparable with the production shown in the example in Section \ref{generating_unary_numbers} which expresses the recursive nature of the unary number system. It is, in effect, a `type definition' for the variable defining the infinite set of alternative strings (composed from the given alphabet) which the variable may take.\footnote{Readers will note that the pattern
`X L \#L X \#X \#X' contains two `variables': the pair of symbols `X \#X', already considered, and also the pair of symbols `L \#L' which can take any of `a', `b' or `c' as its `value'.}

\end{itemize}

In accordance with what was said about symbols in Section \ref{syntax_and_semantics}, none of the symbols in Figure \ref{rotation_trick_patterns} have any `hidden' meaning. In particular, the symbols `X', `\#X', `L', `\#L', `P' and `\#P' (which may be described, informally, as `service' symbols) have exactly the same formal status as other symbols (`a', `b' etc) and enter into matching and unification in the same way.

\begin{figure}
\centering
\begin{tabular}{l l}
{\em New} \\
\\
a b c b t \\
\\
{\em Old} \\
\\
L\ a\ \#L\ \ \ \ \ \ \ & P\ a\ X\ \#X\ a\ \#P \\
L\ b\ \#L\ \ \ \ \ \ \ & P\ b\ X\ \#X\ b\ \#P \\
L\ c\ \#L\ \ \ \ \ \ \ & P\ c\ X\ \#X\ c\ \#P \\
L\ t\ \#L\ \ \ \ \ \ \ & X\ L\ \#L\ X\ \#X\ \#X \\
\end{tabular}

\caption{SP patterns modelling the first example of a Post canonical system described in Section \ref{structure_pcs_section}. An interpretation of these patterns is given in the text.}
\label{rotation_trick_patterns}
\end{figure}

\subsubsection{One step}

Figure \ref{rotation_alignment} shows the best multiple alignment (in terms of information compression) found by SP61 with the `input' pattern in New and the other patterns in Old. In the figure, the repeated appearances of the pattern `X L \#L X \#X \#X' represent a single instance of that pattern in accordance with the generalisation of the multiple alignment problem described in Section \ref{multiple_appearances}. The same is true of the two appearances of the pattern `L b \#L'.

\begin{figure}[!hbt]
\centering
\begin{BVerbatim}
0   a     b        c        b        t                          0
    |     |        |        |        |                           
1 P a X   |        |        |        |                  #X a #P 1
      |   |        |        |        |                  |        
2     X L | #L X   |        |        |               #X #X      2
        | | |  |   |        |        |               |           
3       L b #L |   |        |        |               |          3
               |   |        |        |               |           
4              X L | #L X   |        |            #X #X         4
                 | | |  |   |        |            |              
5                L c #L |   |        |            |             5
                        |   |        |            |              
6                       X L | #L X   |         #X #X            6
                          | | |  |   |         |                 
7                         L b #L |   |         |                7
                                 |   |         |                 
8                                X L | #L X #X #X               8
                                   | | |                         
9                                  L t #L                       9
\end{BVerbatim}
\caption{The best multiple alignment (in terms of information compression) found by SP61 using the patterns shown in Figure \ref{rotation_trick_patterns} as described in the text.}
\label{rotation_alignment}
\end{figure}

Unification of the matching symbols in Figure \ref{rotation_alignment} has the effect of `projecting' the multiple alignment into a single sequence, thus:

\begin{center}
P a X L b \#L X L c \#L X L b \#L X L t \#L X \#X \#X \#X \#X \#X a \#P.
\end{center}

\noindent Notwithstanding the interpolation of instances of `service' symbols like `P', `X', `L', `\#L' etc, this sequence contains the subsequence `a b c b t' corresponding to the `input' and also the subsequence `b c b t a' corresponding to the `output' of the first step of our first example of a Post canonical system.

In this example, the subsequence `b c b t' corresponding to the contents of the `variable' is shared by the input and the output thus removing the need for the rewrite arrow in the Post canonical system. An analogy for this idea is the way Welsh and English are sometimes conflated in bilingual signs in Wales. The phrase `Cyngor Gwynedd' in Welsh means `Gwynedd Council' in English. So it is natural, in bilingual signs put up by Gwynedd Council, to represent both the Welsh and the English forms of the name using the words `Cyngor Gwynedd Council'. In this example, the word `Gwynedd' serves both the Welsh version of the name and the English version in much the same manner that, in our example above, the contents of the variable serves both the input string and the output string.

\subsubsection{Repetition of steps}

The second and subsequent steps may proceed in the same way. At the end of each step, we may suppose that the `service' symbols are stripped out and the sequence from the contents of the `variable' onwards is presented to the system again as New. When `t' reaches the front of the string, there is no production with a leading `t' which means that New at that stage cannot be fully matched. In this example, this may be taken as a signal that the rotation is complete.

In Minsky's example (which takes too much space to reproduce here), the presence at the beginning of the string of `T1' (which may be equated with `t' in our example) causes other productions to `fire', leading the processing into other paths.

\subsection{Other examples}

This subsection shows how the second and third examples of a Post canonical system which were described above may be modelled by the SP system.

\subsubsection{Unary numbers and the SP system}\label{unary_numbers_and_sp}

\index{unary numbers|(}\index{recursion!in multiple alignments|(}\index{recursion!unary numbers}

Figure \ref{unary_number_patterns} shows a New pattern and two Old patterns to model the example of a Post canonical system for the creation or recognition of unary numbers (Section \ref{generating_unary_numbers}, above).

\begin{figure}
\centering
\begin{tabular}{l}
{\em New} \\
\\
0 (Other options for New are described in the text) \\
\\
{\em Old} \\
\\
X\ a\ 0\ \#X \\
X\ b\ X\ \#X 1 \#X \\
\end{tabular}
\caption{SP patterns corresponding to a Post canonical system for the creation or recognition of unary numbers (Section \ref{generating_unary_numbers}).}
\label{unary_number_patterns}
\end{figure}

The New pattern corresponds to the axiom in the Post canonical system while the pattern `X b X \#X 1 \#X' is equivalent to the production. The pattern `X a 0 \#X' represents the number `0' corresponding to `0' in the alphabet of the Post canonical system (`1' is implicit in `X b X \#X 1 \#X'). Incidentally, `a' and `b' in the two Old patterns are needed merely to accommodate the scoring system in SP61 and may otherwise be ignored.

The pair of symbols `X \#X' in the pattern `X b X \#X 1 \#X' may be read as ``any unary number'' and the whole pattern may be read as ``a unary number is any unary number followed by `1''', a recursive description much like the production `$\$ \rightarrow \$\ 1$'.

Given the pattern `0' in New and the other two patterns in Old, SP61 creates a succession of good multiple alignments, one example of which is shown in Figure \ref{unary_alignments} (a). If it is not stopped, the program will continue producing multiple alignments like this until the memory of the machine is exhausted.

If we project the multiple alignment in Figure \ref{unary_alignments} (a) into a single sequence and then ignore the `service' symbols (`a', `b', `X' and `\#X'), we can see that the system has, in effect, generated the unary number `0 1 1 1 1'. We can see from this example how the multiple alignment has captured the recursive nature of the unary number definition.

\begin{figure}[!hbt]
\centering
\begin{BVerbatim}
0                     0                        0
                      |                       
1                 X a 0 #X                     1
                  |     |                     
2             X b X     #X 1 #X                2
              |              |                
3         X b X              #X 1 #X           3
          |                       |           
4     X b X                       #X 1 #X      4
      |                                |      
5 X b X                                #X 1 #X 5

(a)

0                     0    1    1    1    1    0
                      |    |    |    |    |   
1                 X a 0 #X |    |    |    |    1
                  |     |  |    |    |    |   
2             X b X     #X 1 #X |    |    |    2
              |              |  |    |    |   
3         X b X              #X 1 #X |    |    3
          |                       |  |    |   
4     X b X                       #X 1 #X |    4
      |                                |  |   
5 X b X                                #X 1 #X 5

(b)
\end{BVerbatim}
\caption{(a) One of many good multiple alignments produced by SP61 with the pattern `0' in New and the `Old' patterns from Figure \ref{unary_number_patterns} in Old. (b) The best multiple alignment produced by SP61 with `0 1 1 1 1' in New and the same patterns in Old.}
\label{unary_alignments}
\end{figure}

Figure \ref{unary_alignments} (b) shows the best multiple alignment produced by SP61 when `0' in New is replaced by the `axiom' or `input' string `0 1 1 1 1'. This multiple alignment is, in effect, a recognition of the fact that `0 1 1 1 1' is a unary number. It corresponds to the way a Post canonical system may be run `backwards' to recognise input patterns but, since there is no left-to-right arrow in the SP scheme, the notion of `backwards' processing does not apply. Other unary numbers may be recognised in a similar way.%
\index{unary numbers|)}

\subsubsection{Palindromes and the SP system}

\index{palindromes|(}\index{recursion!palindromes}

Figure \ref{palindrome_patterns} shows patterns that may be used with SP61 to model the Post canonical system described in Section \ref{generation_of_palindromes}. The first six patterns in Old may be seen as analogues of the six axioms in the Post canonical system. The last pattern in Old expresses the recursive nature of palindromes. The pattern above it serves to link the single letter patterns (the first three in Old) to the recursive pattern.

\begin{figure}
\centering
\begin{tabular}{l}
{\em New} \\
\\
a c b a b a b c a \\
\\
(Other options for New are described in the text) \\
\\
{\em Old} \\
\\
L a \#L \\
L b \#L \\
L c \#L \\
L1 a \#L1 L2 a \#L2 \\
L1 b \#L1 L2 b \#L2 \\
L1 c \#L1 L2 c \#L2 \\
X L \#L \#X \\
X L1 \#L1 X \#X L2 \#L2 \#X \\
\end{tabular}
\caption{Patterns for processing by SP61 to model a Post canonical system to produce or recognise palindromes (Section \ref{generation_of_palindromes}).}
\label{palindrome_patterns}
\end{figure}

\begin{figure}[!hbt]
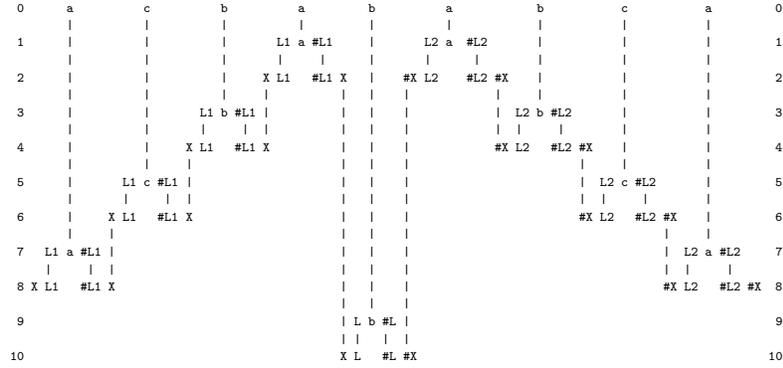

\fontsize{05.50pt}{06.60pt}
\centering
\begin{BVerbatim}
 0      a          c          b          a         b          a            b           c           a         0
        |          |          |          |         |          |            |           |           |          
 1      |          |          |       L1 a #L1     |       L2 a  #L2       |           |           |         1
        |          |          |       |     |      |       |      |        |           |           |          
 2      |          |          |     X L1   #L1 X   |    #X L2    #L2 #X    |           |           |         2
        |          |          |     |          |   |    |            |     |           |           |          
 3      |          |       L1 b #L1 |          |   |    |            |  L2 b #L2       |           |         3
        |          |       |     |  |          |   |    |            |  |     |        |           |          
 4      |          |     X L1   #L1 X          |   |    |            #X L2   #L2 #X    |           |         4
        |          |     |                     |   |    |                        |     |           |          
 5      |       L1 c #L1 |                     |   |    |                        |  L2 c #L2       |         5
        |       |     |  |                     |   |    |                        |  |     |        |          
 6      |     X L1   #L1 X                     |   |    |                        #X L2   #L2 #X    |         6
        |     |                                |   |    |                                    |     |          
 7   L1 a #L1 |                                |   |    |                                    |  L2 a #L2     7
     |     |  |                                |   |    |                                    |  |     |       
 8 X L1   #L1 X                                |   |    |                                    #X L2   #L2 #X  8
                                               |   |    |         
 9                                             | L b #L |                                                    9
                                               | |   |  |         
10                                             X L   #L #X                                                  10
\end{BVerbatim}
\caption{\sloppy The best multiple alignment (in terms of compression) produced by SP61 with the patterns from Figure \ref{palindrome_patterns} in New and Old, as shown in that figure.}
\label{palindrome_alignment}
\end{figure}

Figure \ref{palindrome_alignment} shows the best multiple alignment produced by SP61 with `a c b a b a b c a' in New and, in Old, the patterns under the heading `Old' in Figure \ref{palindrome_patterns}. The multiple alignment may be seen as a recognition that the pattern in New is indeed a palindrome. Other palindromes (using the same alphabet) may be recognised in a similar way.

As with the example of unary numbers, the same patterns from Old may be used to generate palindromes. In this case, some kind of `seed' is required in New like a single letter (`a', `b' or `c') or a pair of letters (`a a', `b b' or `c c'). Given one of these `seeds' in New, the system can generate palindromes until the memory of the machine is exhausted.

Readers may wonder whether the patterns from Old in Figure \ref{palindrome_patterns} might be simplified. Would it not be possible to substitute `X a \#X' for `L a \#L' (and likewise for the other single letter `axioms') and to replace `L1 a \#L1 L2 a \#L2' with `X a X \#X a \#X' (and likewise for the other pairs) and then remove `X L \#L \#X' and `X L1 \#L1 X \#X L2 \#L2 \#X'? This is possible but the penalty is that, for reasons that would take too much space to explain, searching for the best multiple alignments takes very much longer and leads along many more `false trails'.

\index{Post canonical system|)}\index{palindromes|)}\index{recursion!in multiple alignments|)}

\section{Discussion}\label{computing_discussion}

Section \ref{sp_and_pcs} has described, with examples, how the operation of a Post canonical system in normal form may be understood in terms of the SP theory. Since it is known that any Post canonical system may be modelled by a Post canonical system in normal form (\citet{post_1943}, \citet{minsky_1967}, Chapter 13), we may conclude that the operation of any Post canonical system may be interpreted in terms of the SP theory. Since we also know that any universal Turing machine may be modelled by a Post canonical system \citep[Chapter 14]{minsky_1967} we may also conclude that the operation of any universal Turing machine may be interpreted in terms of the SP theory.

What then is the relationship between the SP theory and earlier models of computing---the universal Turing machine, Post canonical system, lamda calculus and others? Is the SP theory subject to the same limitations as earlier models? What are the implications for the design of practical computing systems? These and related questions are discussed in the subsections that follow. In this discussion, earlier models of computing will be referred to collectively as the `universal Turing machine' since they are known to be equivalent to each other, and the universal Turing machine is the best known of those models.

\subsection{The SP theory in relation to earlier models of computing}\label{sp_and_other_models_of_computing}

A slightly inaccurate view of the relationship between the SP theory and the universal Turing machine is that:

\begin{center}
\begin{tabular}{l}
SP = universal Turing machine + X,
\end{tabular}
\end{center}

\noindent where `X' is software that makes the universal Turing machine behave like an SP system. If a conventional computer is seen as being roughly equivalent to a universal Turing machine, then `SP = universal Turing machine + X' is a description of the SP61 and SP70 models of the SP system.

These models have been constructed in that way because, with current technology, this is the only practical option for creating working models. However, the SP theory itself---distinct from the SP models---is a new theory of computing that is built from foundations that are different from any of the earlier theories. None of those theories are incorporated within the SP theory. Concepts such as the matching and unification of patterns, and the SP version of multiple alignment, are bedrock in the SP theory whereas such concepts are, at best, only implicit in the other theories. Perhaps more importantly, information compression and minimum length encoding principles are fundamental in the SP theory but these concepts have no counterpart in any of the earlier theories.

While it may be accepted that the SP theory is indeed different from any earlier theory of computing, a sceptic may object that the theory is merely another variant on ideas that were developed 60 or more years ago and that it contributes nothing to our understanding that had not already been learned from earlier theories.

In answer to this possible objection, a key difference between the SP theory and earlier theories is that the former has a lot to say about the nature of intelligence (Chapters \ref{language_chapter} to \ref{learning_chapter}), and it provides a novel interpretation for a range of concepts in logic and mathematics (Chapter \ref{maths_logic_chapter}), while the universal Turing machine is largely silent in these areas. Notwithstanding Alan Turing's own vision of the possibility of artificial intelligence \citep{turing_1950}, a `raw' universal Turing machine, without any software, can do very little.

It may be argued that, while a universal Turing machine is not intelligent in itself, it can be {\em programmed} to behave intelligently. But a universal Turing machine that has been programmed is not a universal Turing machine! The addition of software to a universal Turing machine makes it into a new and different system:

\begin{center}
\begin{tabular}{l}
SP = universal Turing machine + Y,
\end{tabular}
\end{center}

\noindent where `Y' is non-SP software needed to make the universal Turing machine behave with the same kind of intelligence as an SP system. If this point is accepted, the further question arises: ``Is there anything to choose between the SP theory and `universal Turing machine + Y'?''

In answer to this question, the SP system appears to be very much simpler, at least with respect to alternatives that are currently available. As described in Section \ref{simplification_of_computing_systems}, below, the SP theory is more complex than the core universal Turing machine model but, when we take account of the software that is needed to do anything useful, an overall reduction in complexity may be achieved. In keeping with the minimum length encoding principles on which the theory is founded, the SP theory is a rationalisation and simplification of concepts that combines conceptual {\em Simplicity} with explanatory or descriptive {\em Power}.

\subsection{The scope of the SP theory}\label{scope_of_sp_theory}

It is natural to ask whether the SP theory, as a new theory of computing, might be free from some or all of the limitations of existing systems. Is it possible that an SP system might be able to compute things that are not computable with a universal Turing machine, by-passing the conclusions of G{\"o}del's theorems or computational complexity\index{complexity, computational} theory?

Although no formal proofs have been attempted in this area, the answer to such questions is almost certainly ``No''. Computation in the SP system is, almost certainly, subject to the same limitations as other systems.

However, we can say with some confidence that the scope of the SP system is {\em no less} than a universal Turing machine. Anything that is computable with a universal Turing machine is also computable in the SP system. This follows from the fact that any universal Turing machine can be modelled in the SP system, as described above.

\subsection{Constraints and alternative styles of computation}\label{styles_of_computation}

\index{constraint|(}

The SP theory is a theory of computation of {\em any} kind, in any kind of system, either natural or artificial. In terms of the theory, {\em all} computations may be understood as a process of searching for patterns that match each other, with compression of information by the unification of patterns that are the same.

The generality of the theory provides scope for alternative `styles' of computation. As we have seen (Sections \ref{matching_searching_and_constraints} and \ref{ma_searching_and_constraints}), it is not feasible to conduct an exhaustive search of the space of possible matches between patterns within $I$, unless $I$ is very small. In any practical system for computation, there is a need for constraints---and the choice of constraints can influence the style of computation:

\begin{itemize}

\item In `conventional' kinds of computing---arithmetic, compilers, accounts programs and so on---the process of searching for matches between patterns is heavily constrained. In the main, this takes the form of absolute constraints, the most prominent of which is that, normally, patterns must match each other exactly and all kinds of partial matching are excluded. This kind of constraint speeds up processing dramatically but the penalty is the notoriously unforgiving and `brittle' nature of conventional computer systems. It seems likely that this style of computing was adopted in the early days of computing because early machines were not powerful enough for anything more ambitious.

\item In `artificial intelligence' kinds of computing---fuzzy pattern recognition, unsupervised learning, probabilistic reasoning and so on---the process of searching is less heavily constrained and `heuristic' kinds of constraint are favoured so that no part of the search space is ruled out {\em a priori}. This kind of searching is more demanding than the more heavily constrained searching of conventional computing because larger areas of the search space is examined. However, the benefit is a more flexible, human-like `intelligence', and `graceful' failure with less of the brittleness of conventional systems.

\end{itemize}

\index{constraint|)}

\subsection{Rationalisation and simplification of computing systems}\label{simplification_of_computing_systems}

In the words of Kurt Lewin, ``There is nothing so practical as a good theory.''\footnote{{\em Field Theory in Social Science}, 1951, p.~20.} Rationalisation and simplification of concepts in the SP theory should translate into corresponding benefits in the design of computing systems.

\begin{figure}[!hbt]
\centering
\includegraphics[width=0.9\textwidth]{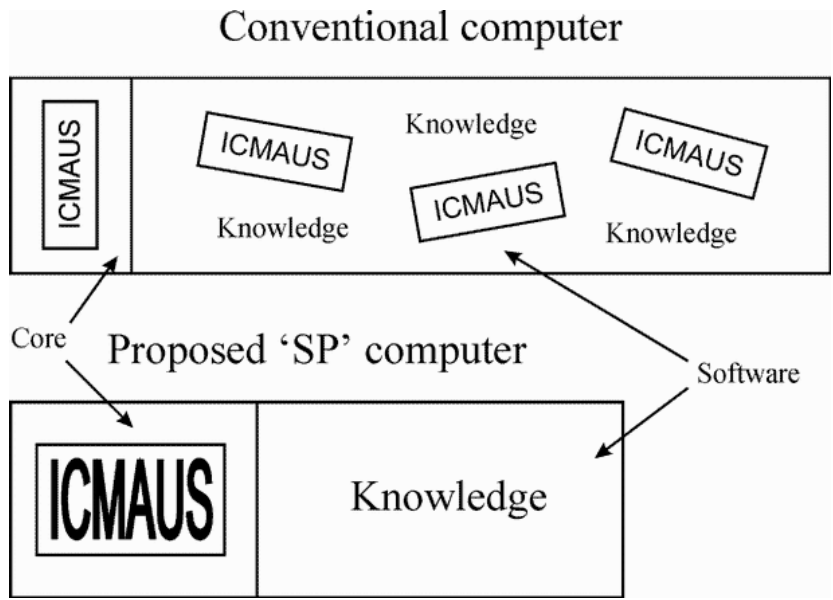}
\caption{Schematic representation of a conventional computer and the proposed `SP' computer, as discussed in the text.}
\label{conventional_and_sp_computers}
\end{figure}

Figure \ref{conventional_and_sp_computers} shows a schematic representation of a conventional computer and the proposed `SP' computer, based on the SP theory. The computational `core' of the computer (the central processing unit with its associated mechanisms) is shown to the left of the vertical line in each representation, while software for a range of applications is shown on the right. Data to be processed is not shown in either figure because it is assumed to be the same in both cases.

In the conventional computer, relatively restricted or constrained forms of the mechanisms for information-compression-by-multiple-alignment-unification-and-search (represented as `ICMAUS' in small letters in a box) may be seen in the core of the computer. For example, although the process of finding a location in computer memory is normally described in terms of the operation of logic circuits, it may also be seen in more abstract terms as a process of comparing the required address with actual addresses until an exact match is found. In a similar way, the operation of an `arithmetic and logical unit' may be understood in terms of pattern matching although it is normally described more concretely in terms of logic gates.

Software in a conventional computing system may be seen as domain-specific knowledge (shown in the figure as `knowledge') combined with SP mechanisms (shown as three instances of `ICMAUS' in a box). SP mechanisms range from the relatively constrained kinds of searching and matching which is done in a typical compiler or database management system to the more ambitious kinds of searching and matching performed by artificial intelligence programs. Knowledge about some domain of application such as accountancy or gardening may be integrated with other software, or it may take the form of rules or records loaded into a pre-existing expert-system shell or database management system.

The key point here is that, although the core of a conventional computer is relatively simple, the complete system, including software for a range of applications, is relatively complex. This is because software systems are unnecessarily complex. Notwithstanding the existence of suites of programs that are designed to work together, there is very poor integration amongst software systems. In particular, mechanisms for comparing patterns and searching for good matches between patterns appear repeatedly and redundantly in many different applications.

What is envisaged for the proposed SP computer is that a single, relatively sophisticated SP mechanism would be developed for the core of the computer. This is shown in Figure \ref{conventional_and_sp_computers} as `ICMAUS' in large letters in a box. It is anticipated that this would eliminate the need for any SP mechanisms in the software. All that would be necessary would be to supply the system with the knowledge that is needed for each domain of application. This is shown as `knowledge' in the right-hand part of the schematic representation of the SP system.

Although the core of the proposed new system would be more complex than the computational core of a conventional system, there would be an overall simplification of the system owing to the elimination of complexity arising from the redundant programming of SP mechanisms in diverse applications, especially in artificial intelligence. In the figure, this can be seen as the relatively small size of the SP system compared with the conventional computer.

A single SP mechanism does not preclude alternative styles of computing. The mechanism would be designed so that it could operate in `Rolls Royce' mode for artificial intelligence applications but it should also be possible to apply constraints so that it would operate more like a conventional system, trading flexibility for speed.

\subsubsection{Generalisation of computing systems}

The kind of rationalisation and simplification of computing systems that has been described is not a new idea. In the early days of computing, each new database was written from scratch. But then computer scientists realised that every database needs mechanisms for storing knowledge and editing it, and every database needs a mechanism for finding knowledge in the store. A lot of effort could be saved by writing a generalised 'database management system' that could be loaded with different kinds of knowledge according to need.

The SP project extends this idea. Instead of creating a generalised framework for databases alone, the aim has been to create a generalised framework for things like learning, reasoning, information storage and retrieval, pattern recognition, and so on.

\section{Conclusion}\label{computing_conclusion}

The SP theory is a new theory of computing, built from new foundations and distinct from any of the earlier theories. Its computational scope is almost certainly the same as earlier theories, neither more nor less. However, it has much more to say about the nature of intelligence than any of the earlier theories, it provides a novel interpretation for concepts in logic and mathematics, and it offers opportunities for the rationalisation and simplification of the design of computing systems.

These aspects of the theory are developed in the chapters that follow.%
\index{computing|)}

%% file: language.tex

\chapter[Natural Language Processing]{Natural Language Processing%
\protect\footnote{Based on \citet{wolff_2000}.}}%
\label{language_chapter}

\section{Introduction}

This chapter continues the linguistic theme of the examples presented in Chapter \ref{theory_chapter}. By contrast with those introductory examples, this chapter shows how the SP system can model more subtle aspects of language such as recursive structures, ambiguity in parsing, and `context sensitive' features of syntax.

The main emphasis in this chapter is on the representation and processing of syntax in natural language. The ways in which the system can model non-syntactic, `semantic' kinds of knowledge such as class-inclusion hierarchies, part-whole hierarchies, if-then rules and so on are described and discussed in several of the chapters that follow. However, this chapter does contain an example showing how syntax and semantics may be integrated within the system.

\subsection{Novelty of the proposals}

Aspects of these proposals that appear to be novel---compared with other systems for natural language processing---are described in the following subsections.

\subsubsection{Representing syntax with patterns}

As we saw in Section \ref{framework_examples_section}, the SP system allows the rules of a context-free phrase-structure grammar to be represented with patterns in a manner that is distinct from ordinary re-write rules but with recognisable similarities. However, as we shall see, patterns can also be used to represent discontinuous dependencies in syntax in a manner which is rather different from the way in which this aspect of syntax is handled in other systems. Arguably, it is more direct and more transparent than alternative systems.

\subsubsection{Parsing (with choices at many levels) as multiple alignment}

\index{parsing!language|(}

Perhaps the most novel feature of the present proposals appears to be the idea
that parsing, in the sense understood in theoretical and computational
linguistics and natural language processing, may be understood as multiple alignment.

A concept of parsing is already well-established
in the literature on data compression \citep[see, for example,][]{storer_1988}.
In that context, it means a process of analysing data into segments corresponding to those in a pre-defined dictionary of segments.

But this kind of parsing is simpler than `linguistic' kinds of parsing.
In the first case, alternatives can be chosen only at one level (although segments may have internal hierarchical structure). In the second kind of parsing, which is the focus of interest in this chapter, sets of alternatives (from which selections must be made) may appear at {\em arbitrarily many levels} in the grammar which is used to guide the parsing.

\subsubsection{Parsing as information compression}

\sloppy Research on parsing and related topics within computational
linguistics and AI does not normally consider these topics in terms of
information compression \citep[but see, for example,][]{berger_etal_1996, hu_etal_1997}.
However, there is a well-developed tradition of parsing and linguistic
analysis in terms of probabilities, with associated concepts such as
`stochastic grammars', `maximum-likelihood', `Bayesian inference' and
`statistical analysis' \citep[see, for example,][]{abney_1997, black_etal_1993,
dreuth_ruber_1997, garside_etal_1987, takahashi_sagayama_1997, wu_1997, lucke_1995} and, as we saw in Section \ref{probabilities_ic_section}, there is a close connection between probabilities and information compression.%
\index{parsing!language|)}

\subsubsection{Production of language in the SP framework}

\index{language!production}

As we saw in Section \ref{decompression_by_compression}, a sentence may be constructed within the SP system by compression of information in exactly the same way as a sentence may be parsed. This is a novel feature of the system, not shared by other systems for natural language processing, although it is similar in some respects to the way in which an appropriately-constructed Prolog program may be run `forwards' or `backwards' according to need.

As a system for natural language production, the system contrasts with other systems in much the same way as it does for parsing: in the representation of syntax with patterns, in the use of the multiple alignment concept, and in the focus on compression.

\section{Ambiguities and recursion}

\subsection{Ambiguities in parsing}\label{ambiguity_in_parsing}

\index{ambiguity|(}

Natural languages are notoriously ambiguous, not only in their meanings but also in their syntax. An example which includes syntactic ambiguity is the second sentence in Groucho Marks's ``Time flies like an arrow. Fruit flies like a banana''.

\begin{figure}[!hbt]
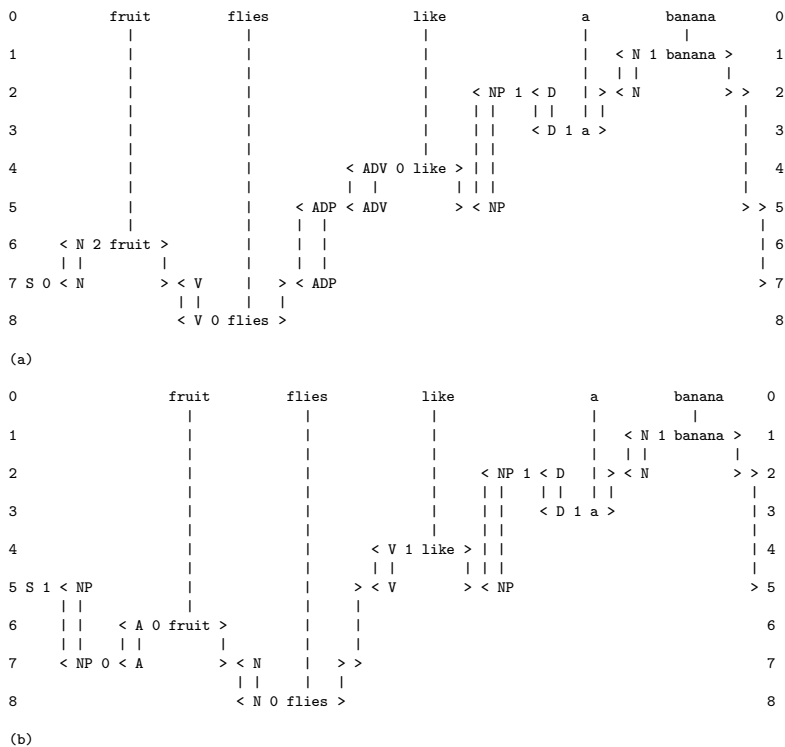

\fontsize{06.00pt}{07.20pt}
\centering
\begin{BVerbatim}
0           fruit         flies                 like                a         banana       0
              |             |                    |                  |           |         
1             |             |                    |                  |   < N 1 banana >     1
              |             |                    |                  |   | |          |    
2             |             |                    |     < NP 1 < D   | > < N          > >   2
              |             |                    |     | |    | |   | |                |  
3             |             |                    |     | |    < D 1 a >                |   3
              |             |                    |     | |                             |  
4             |             |           < ADV 0 like > | |                             |   4
              |             |           |  |         | | |                             |  
5             |             |     < ADP < ADV        > < NP                            > > 5
              |             |     |  |                                                   |
6     < N 2 fruit >         |     |  |                                                   | 6
      | |         |         |     |  |                                                   |
7 S 0 < N         > < V     |   > < ADP                                                  > 7
                    | |     |   |                                                         
8                   < V 0 flies >                                                          8

(a)

0                  fruit         flies           like                a         banana     0
                     |             |              |                  |           |       
1                    |             |              |                  |   < N 1 banana >   1
                     |             |              |                  |   | |          |  
2                    |             |              |     < NP 1 < D   | > < N          > > 2
                     |             |              |     | |    | |   | |                |
3                    |             |              |     | |    < D 1 a >                | 3
                     |             |              |     | |                             |
4                    |             |       < V 1 like > | |                             | 4
                     |             |       | |        | | |                             |
5 S 1 < NP           |             |     > < V        > < NP                            > 5
      | |            |             |     |                                               
6     | |    < A 0 fruit >         |     |                                                6
      | |    | |         |         |     |                                               
7     < NP 0 < A         > < N     |   > >                                                7
                           | |     |   |                                                 
8                          < N 0 flies >                                                  8

(b)
\end{BVerbatim}
\caption{The two best alignments found by SP61 with patterns in Old representing
grammatical rules and the ambiguous sentence `fruit flies like a banana' in New.}
\label{ambiguity_alignments}
\end{figure}

Figure \ref{ambiguity_alignments} shows how SP61 can accommodate the ambiguity of that example, given an appropriate grammar.\footnote{By contrast with the alignments shown in Figures \ref{alignment_figure_1} and \ref{alignment_figure_2}, the alignment in Figure \ref{ambiguity_alignments} shows words and word stems as single symbols, not divided into letters. This change has been made to save space and improve readability. It makes no difference to the overall structure created by the program.} In this example, the two parsings shown have $CD$s that are close in value and higher than the compression scores of other alignments formed for the same sentence. 

\subsubsection{Another example}

As a further example, consider the phoneme sequence `ae i s k r ee m' (which can be read as ``ice cream'' or ``I scream'').\footnote{The symbols used are an alphabetic adaptation of the International Phonetic Alphabet phoneme symbols.} A grammar that relates to this example is shown in Figure \ref{icecream_grammar}.

\begin{figure}[!hbt]
\centering
\begin{BVerbatim}
S 0 NP #NP V #V ADV #ADV #S (100)
S 1 NP #NP VB #VB A #A #S (200)
NP 0 w ee #NP (100)
NP 1 ae i #NP (50)
NP 2 A #A N #N #NP (150)
A 0 ae i s #A (100)
A 1 h o t #A (80)
A 2 k o l d #A (70)
N 0 k r ee m #N (30)
N 1 m i l k #N (20)
V 0 s k r ee m #V (150)
V 1 sh ae w t #V (50)
VB 0 i z #VB (200)
ADV 0 l ae w d l i #ADV (40)
ADV 1 k w ae i e t l i #ADV (60)
\end{BVerbatim}
\caption{A simple grammar for phoneme patterns which allows two main parsings for the phoneme sequence `ae i s k r ee m'.}
\label{icecream_grammar}
\end{figure}

Figure \ref{icecream_alignment_1} shows the two best alignments obtained by SP61 with `ae i s k r ee m' in New and the grammar from Figure \ref{icecream_grammar} in Old. These two alignments, which are the two `correct' parsings of the sample, have compression scores which are close in value and higher than any others. 

\begin{figure}[!hbt]
\fontsize{10.00pt}{12.00pt}
\centering
\begin{BVerbatim}
0          ae i s        k r ee m        0
           |  | |        | | |  |       
1          |  | |    N 0 k r ee m #N     1
           |  | |    |            |     
2 NP 2 A   |  | | #A N            #N #NP 2
       |   |  | | |                     
3      A 0 ae i s #A                     3

(a)

0          ae i         s k r ee m                0
           |  |         | | | |  |               
1          |  |     V 0 s k r ee m #V             1
           |  |     |              |             
2 S 0 NP   |  | #NP V              #V ADV #ADV #S 2
      |    |  |  |                               
3     NP 1 ae i #NP                               3

(b)
\end{BVerbatim}
\caption{An alignment showing the two `best' parsings of the phoneme sequence `ae i s k r ee m' created by SP52 using the grammar shown in Figure \ref{icecream_grammar}.}
\label{icecream_alignment_1}
\end{figure}

\subsubsection{Disambiguating context}
If something is ambiguous, we normally look for some kind of context to resolve the ambiguity. This effect of context fits neatly into the SP system as can be seen from the example shown here. 

Figure \ref{icecream_alignment_2} (a) shows the best alignment found by SP61 with `ae i s k r ee m l ae w d l i' in New and the grammar from Figure \ref{icecream_grammar} in Old. The addition of `l ae w d l i' (``loudly'') to the ambiguous pattern `ae i s k r ee m' shifts the balance in favour of ``I scream'' rather than ``ice cream''.

In a similar way, the addition of `i z k o l d' (``is cold'') to the same ambiguous pattern yields the alignment shown in Figure \ref{icecream_alignment_2}, shifting the balance in the other direction.

\begin{figure}[!hbt]
\fontsize{08.00pt}{09.60pt}
\centering
\begin{BVerbatim}
0          ae i         s k r ee m          l ae w d l i         0
           |  |         | | | |  |          | |  | | | |        
1          |  |         | | | |  |    ADV 0 l ae w d l i #ADV    1
           |  |         | | | |  |     |                  |     
2 S 0 NP   |  | #NP V   | | | |  | #V ADV                #ADV #S 2
      |    |  |  |  |   | | | |  | |                            
3     |    |  |  |  V 0 s k r ee m #V                            3
      |    |  |  |                                              
4     NP 1 ae i #NP                                              4

(a)

0              ae i s        k r ee m             i z         k o l d       0
               |  | |        | | |  |             | |         | | | |      
1              |  | |        | | |  |             | |     A 2 k o l d #A    1
               |  | |        | | |  |             | |     |           |    
2 S 1 NP       |  | |        | | |  |    #NP VB   | | #VB A           #A #S 2
      |        |  | |        | | |  |     |  |    | |  |                   
3     |        |  | |    N 0 k r ee m #N  |  |    | |  |                    3
      |        |  | |    |            |   |  |    | |  |                   
4     NP 2 A   |  | | #A N            #N #NP |    | |  |                    4
           |   |  | | |                      |    | |  |                   
5          A 0 ae i s #A                     |    | |  |                    5
                                             |    | |  |                   
6                                            VB 0 i z #VB                   6

(b)
\end{BVerbatim}
\caption{(a) An alignment showing the best parsing found by SP61 for the phoneme sequence `ae i s k r ee m' when it is included within the larger sequence `ae i s k r ee m l ae w d l i'---using the grammar shown in Figure \ref{icecream_grammar}. (b) An alignment showing the best parsing of the phoneme sequence `ae i s k r ee m' when it is included within the larger sequence `ae i s k r ee m i z k o l d'---using the same grammar as for (a).}
\label{icecream_alignment_2}
\end{figure}

\index{ambiguity|)}

\subsection{Recursion}\label{language_recursion}

\index{recursion|(}

Recursion is a prominent feature of natural languages, illustrated classically by the traditional nursery rhyme {\em The House that Jack Built} whose last verse begins: ``This is the farmer sowing his corn, That kept the cock that crowed in the morn, That waked the priest all shaven and shorn, That married the man all tattered and torn, ...'' and so on.\footnote{From {\em Mother Goose Nursery Rhymes}, London: Heinemann, 1994.}

In all the diverse manifestations of recursion (so brilliantly described by Douglas Hofstadter \citeyearpar{hofstadter_1980}), the key feature is that there is at least one structure which contains a reference to itself, either immediately or at some lower `level' within its (hierarchically organised) constituents. 

Examples showing how the system can handle recursive structures outside the realm of natural language may be found in Chapter \ref{computing_chapter}. Here, Figure \ref{recursion_grammar} shows an SP grammar for a fragment of English where the second pattern `S 1 PN \#PN V \#V DPN \#DPN S \#S \#S' contains a recursive reference to itself (near the end of the pattern) via the left and right boundary symbols `S \#S'. 

\begin{figure}[!hbt]
\centering
\begin{BVerbatim}
S 0 PN #PN V #V ADV #ADV #S (1000)
S 1 PN #PN V #V DPN #DPN S #S #S (700)
DPN that #DPN (300)
PN 0 we #PN (400)
PN 1 you #PN (700)
PN 2 he #PN (350)
PN 3 it #PN (250)
V 0 says #V (200)
V 1 say #V (210)
V 2 said #V (300)
V 3 thinks #V (250)
V 4 think #V (200)
V 5 goes #V (300)
V 6 go #V (240)
ADV 0 fast #ADV (400)
ADV 1 away #ADV (250)
ADV 2 later #ADV (350)
\end{BVerbatim}
\caption{A fragment of English grammar with recursion.}
\label{recursion_grammar}
\end{figure}

\begin{figure}[!hbt]
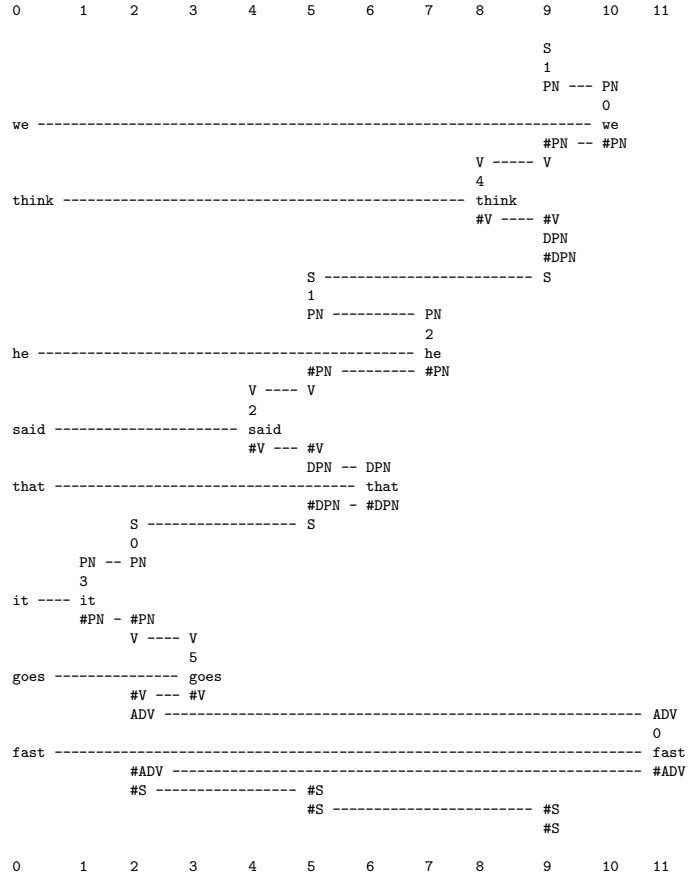

\fontsize{06.00pt}{07.20pt}
\centering
\begin{BVerbatim}
0       1     2      3      4      5      6      7     8       9      10    11  

                                                               S                
                                                               1                
                                                               PN --- PN        
                                                                      0         
we ------------------------------------------------------------------ we        
                                                               #PN -- #PN       
                                                       V ----- V                
                                                       4                        
think ------------------------------------------------ think                    
                                                       #V ---- #V               
                                                               DPN              
                                                               #DPN             
                                   S ------------------------- S                
                                   1                                            
                                   PN ---------- PN                             
                                                 2                              
he --------------------------------------------- he                             
                                   #PN --------- #PN                            
                            V ---- V                                            
                            2                                                   
said ---------------------- said                                                
                            #V --- #V                                           
                                   DPN -- DPN                                   
that ------------------------------------ that                                  
                                   #DPN - #DPN                                  
              S ------------------ S                                            
              0                                                                 
        PN -- PN                                                                
        3                                                                       
it ---- it                                                                      
        #PN - #PN                                                               
              V ---- V                                                          
                     5                                                          
goes --------------- goes                                                       
              #V --- #V                                                         
              ADV --------------------------------------------------------- ADV 
                                                                            0   
fast ---------------------------------------------------------------------- fast
              #ADV -------------------------------------------------------- #ADV
              #S ----------------- #S                                           
                                   #S ------------------------ #S               
                                                               #S               

0       1     2      3      4      5      6      7     8       9      10    11  
\end{BVerbatim}
\caption{The best alignment created by SP61 with `we think he said that it goes fast' in New and the grammar shown in Figure \ref{recursion_grammar} in Old.}
\label{recursion_alignment}
\end{figure}

Figure \ref{recursion_alignment} shows how the recursive sentence ``We think he said that it goes fast'' may be parsed by multiple alignment, using the grammar shown in Figure \ref{recursion_grammar}.\footnote{This alignment is the first example in this book of alignments rotated by $90^o$ compared with the examples shown in Chapter \ref{theory_chapter}. In the majority of cases, alignments are shown like this because they fit more easily onto the page. In accordance with the convention that has been adopted, column 0 contains the pattern from New while the one or more other columns in each alignment show patterns from Old, one pattern per column, in an arbitrary order.} The fact that any pattern in the grammar may appear one or more times in an alignment (Section \ref{multiple_appearances}) means that the second pattern in the grammar may provide a framework for the whole sentence (in column 9 of the alignment) and may also provide a framework for the embedded sentence ``he said that ...''. Within this second sentence is the sentence ``it goes fast'' which is modelled on the pattern in the first line in the grammar. 

\index{recursion|)}

\section{Syntactic dependencies in French}\label{dependencies_in_french}

\index{grammar!syntactic dependencies|(}\index{grammar!French|(}

The examples considered up to now may have given the impression that the SP system is merely a trivial variation on context-free phrase-structure grammar. This section and the one that follows presents alignments from two areas of syntax showing how the SP system as it is realised in the SP61 model may accommodate `context sensitive' aspects of syntax.

It often happens in natural languages that there are syntactic
dependencies between one part of a sentence and another. For example,
there is usually a `number' dependency between the subject of a
sentence and the main verb of the sentence: if the subject has a {\em
singular} form then the main verb must have a singular form and
likewise for {\em plural} forms of subject and main verb.

A prominent feature of these kinds of dependency is that they are often
`discontinuous' in the sense that the elements of the dependency can be
separated, one from the next, by arbitrarily large amounts of
intervening structure. For example, the subject and main verb of a
sentence must have the same number (singular or plural) regardless of
the size of qualifying phrases or subordinate clauses that may come
between them.

Another interesting feature of syntactic dependencies is that one kind
of dependency (e.g., number dependency) can overlap other kinds of
dependency (e.g., gender ({\em masculine}/{\em feminine}) dependency),
as can be seen in the following example.

In the French sentence {\em Les plumes sont vertes} (``The feathers are
green'') there are two sets of overlapping syntactic dependencies like
this:

\begin{center}
\begin{BVerbatim}
 P        P  P          P        Number dependencies
Les plume s sont vert e s
      F               F          Gender dependencies
\end{BVerbatim}
\end{center}

\noindent In this example, there is a number dependency, which is
plural (`P') in this case, between the subject of the sentence, the
main verb and the following adjective: the subject is expressed with a
plural determiner ({\em Les}) and a noun ({\em plume}) which is marked
as plural with the suffix ({\em s}); the main verb ({\em sont}) has a
plural form and the following adjective ({\em vert}) is marked as plural by the suffix ({\em s}).  Cutting right across these number
dependencies is the gender dependency, which is feminine (`F') in this
case, between the feminine noun ({\em plume}) and the adjective ({\em
vert}) which has a feminine suffix ({\em e}).

{\sloppy For many years, linguists puzzled about how these kinds of syntactic
dependency could be represented succinctly in grammars for natural
languages. But then elegant solutions were found in transformational
grammar \citep{chomsky_1957} and, later, in systems like definite clause
grammars \citep[][, based on Prolog]{pereira_warren_1980} \citep[see also][]{gazdar_mellish_1989}.}

The solution proposed here is different from any established system and
is arguably simpler and more transparent than other systems. It will be
described and illustrated with a fragment of the grammar of French
which can generate the example sentence just shown. This fragment of
French grammar, shown in Figure \ref{fr10_grammar}, is expressed with
`patterns' in the same manner as the grammar in Figure \ref{grammar_2}
and others in this article.

\begin{figure}[!hbt]
\centering
\begin{BVerbatim}
S NP #NP VP #VP #S (500)
NP D #D N #N #NP (700)
VP 0 V #V A #A #VP (300)
VP 1 V #V P #P NP #NP #VP (200)
P 0 sur #P (50)
P 1 sous #P (150)
V SNG est #V (250)
V PL sont #V (250)
D SNG M 0 le #D (90)
D SNG M 1 un #D (120)
D SNG F 0 la #D (130)
D SNG F 1 une #D (110)
D PL 0 les #D (125)
D PL 1 des #D (125)
N NR #NR NS1 #NS1 #N (450)
NS1 SNG - #NS1 (250)
NS1 PL s #NS1 (200)
NR M papier #NR (300)
NR F plume #NR (400)
A AR #AR AS1 #AS1 AS2 #AS2 #A (300)
AS1 F e #AS1 (100)
AS1 M - #AS1 (200)
AS2 SNG - #AS2 (175)
AS2 PL s #AS2 (125)
AR 0 noir #AR (100)
AR 1 vert #AR (200)
NP SNG SNG #NP (450)
NP PL PL #NP (250)
NP M M #NP (450)
NP F F #NP (250)
N SNG V SNG A SNG (250)
N PL V PL A PL (250)
N M V A M (300)
N F V A F (400)
\end{BVerbatim}
\caption{A fragment of French grammar with patterns
for number dependencies and gender dependencies.}
\label{fr10_grammar}
\end{figure}

Apart from the use of patterns as the medium of expression, this
grammar differs from systems like transformational grammar or definite clause grammars because the parts of the
grammar which express the forms of `high level' structures like
sentences, noun phrases and verb phrases (represented by the first four
patterns in Figure \ref{fr10_grammar}) do not contain any reference to
number or gender.

Instead, the grammar contains patterns like `NP SNG SNG \#NP' and `N M
V A M' (the last eight patterns in Figure \ref{fr10_grammar}). The
first of these says, in effect, that between the symbols `NP' and
`\#NP' there are two structures marked as singular (`SNG'). In this
simple grammar, there is no ambiguity about what those two structures
are:  they can only be a determiner (`D') followed by a noun (`N'). In
a more complex grammar, there would need to be disambiguating
context to establish the `correct' alignments of symbols. The second
pattern says, in effect, that in a sentence which contains the
(discontinuous) sequence of symbols `N V A', the noun (`N') is
masculine (`M') and the adjective (`A') is also masculine.

\subsection{An alignment}

The alignment in Figure \ref{alignment_fr10} shows the best alignment found by SP61 with our
example sentence in New and the grammar from Figure \ref{fr10_grammar} in Old. The main constituents of the sentence are marked in an appropriate manner and dependencies for number and gender are marked by patterns appearing in columns 13, 14 and 15 of the alignment.

\begin{figure}[!hbt]
\fontsize{06.00pt}{07.20pt}
\centering
\begin{BVerbatim}
0       1      2      3      4      5     6      7      8       9     10    11    12     13    14  15

                                                                            S                        
                                                                      NP -- NP --------- NP          
                                                                D --- D                              
                                                                PL --------------------- PL          
                                                                0                                    
les ----------------------------------------------------------- les                                  
                                                                #D -- #D                             
                                                 N ------------------ N ---------------------- N - N 
                                                 NR --- NR                                           
                                                        F ------------------------------------ F     
plume ------------------------------------------------- plume                                        
                                                 #NR -- #NR                                          
                                          NS1 -- NS1                                                 
                                          PL ------------------------------------------- PL ------ PL
s --------------------------------------- s                                                          
                                          #NS1 - #NS1                                                
                                                 #N ----------------- #N                             
                                                                      #NP - #NP -------- #NP         
                                    VP ------------------------------------ VP                       
                                    0                                                                
                             V ---- V -------------------------------------------------------- V - V 
                             PL ------------------------------------------------------------------ PL
sont ----------------------- sont                                                                    
                             #V --- #V                                                               
               A ------------------ A -------------------------------------------------------- A - A 
               AR --- AR                                                                             
                      1                                                                              
vert ---------------- vert                                                                           
               #AR -- #AR                                                                            
        AS1 -- AS1                                                                                   
        F ------------------------------------------------------------------------------------ F     
e ----- e                                                                                            
        #AS1 - #AS1                                                                                  
               AS2 -------------------------------------------------------------- AS2                
                                                                                  PL ------------- PL
s ------------------------------------------------------------------------------- s                  
               #AS2 ------------------------------------------------------------- #AS2               
               #A ----------------- #A                                                               
                                    #VP ----------------------------------- #VP                      
                                                                            #S                       

0       1      2      3      4      5     6      7      8       9     10    11    12     13    14  15
\end{BVerbatim}
\caption{The best alignment found by SP61 with `les plume s sont
vert e s' in New and the grammar from Figure \ref{fr10_grammar}
in Old.}
\label{alignment_fr10}
\end{figure}

\subsection{Discussion}

Readers may wonder why, in the example just shown, the pattern `NP PL
PL \#NP' is separate from the pattern `N PL V PL A PL'. Why not simply
merge them into something like `NP PL N PL \#NP V PL A PL'. The reason
for separating the number dependencies in noun phrases (`NP') from the
other number dependencies is that they do no always occur together. For
example, noun phrases may be found within one of the two verb-phrase
(`VP') patterns shown in Figure \ref{fr10_grammar} (the fourth pattern
in the grammar) and this context does not contain the `N ... V ... A
...' pattern.

Another question that may come to mind is what happens when there are one or more subordinate clauses between the subject of a sentence and the corresponding verb, given that each such subordinate clause has its own subject and verb? In a sentence such as ``Those that we like win'', it is possible that the number dependency (plural) between `those' and `win' could become confused with the number dependency (also plural) between `we' and `like'.

If patterns representing long-range dependencies are completely separated from the syntactic patterns to which they apply---as in Figures \ref{fr10_grammar} and \ref{alignment_fr10}---this can indeed be a problem. A solution is to mark dependencies within the patterns to which they apply, as shown in Figure \ref{dependencies_with_subordinate_clause_alignment}. Here, the number dependency between `those' and `win' is distinct from the dependency between `we' and `like', despite the fact that both these dependencies are plural.

\begin{figure}[!hbt]
\fontsize{07.00pt}{08.40pt}
\centering
\begin{BVerbatim}
0       1       2     3     4      5      6     7     8     9      10    11    12 

                      S                                                           
                NP -- NP                                                          
                1                                                                 
        PN ---- PN                                                                
        PL ---------- PL                                                          
        2                                                                         
those - those                                                                     
        #PN --- #PN                                                               
                #NP - #NP                                                         
                      QL --------- QL                                             
                                   0                                              
                            DPN -- DPN                                            
                            0                                                     
that ---------------------- that                                                  
                            #DPN - #DPN                                           
                                   S ---------------- S                           
                                                NP -- NP                          
                                                1                                 
                                          PN -- PN                                
                                          PL -------- PL                          
                                          0                                       
we -------------------------------------- we                                      
                                          #PN - #PN                               
                                                #NP - #NP                         
                                                      QL                          
                                                      #QL                         
                                                      V ---------- V              
                                                      PL --------- PL             
                                                                   0              
                                                            VR --- VR             
                                                            2                     
like ------------------------------------------------------ like                  
                                                            #VR -- #VR            
                                                      #V --------- #V             
                                   #S --------------- #S                          
                      #QL -------- #QL                                            
                      V ------------------------------------------------------ V  
                      PL ----------------------------------------------------- PL 
                                                                               0  
                                                                         VR -- VR 
                                                                         0        
win -------------------------------------------------------------------- win      
                                                                         #VR - #VR
                      #V ----------------------------------------------------- #V 
                      #S                                                          

0       1       2     3     4      5      6     7     8     9      10    11    12 
\end{BVerbatim}
\caption{An alignment showing how number dependencies in a sentence may be distinguished from number dependencies in a subordinate clause.}
\label{dependencies_with_subordinate_clause_alignment}
\end{figure}

This method of showing dependencies within the SP system overcomes the problem of subordinate clauses but it loses some of the apparent elegance of the earlier scheme. It is necessary to maintain singular and plural versions of each main structure with corresponding redundancy owing to the similarities amongst these patterns. And this redundancy is compounded if gender dependencies are integrated with main syntactic structures in a similar way.%
\index{grammar!French|)}

\section{Dependencies in the syntax of English auxiliary verbs}

\index{grammar!English auxiliary verbs|(}

This section presents a grammar and examples showing how the syntax of
English auxiliary verbs may be described in the SP system.
Before the grammar and examples are presented, the syntax of this part
of English is described and alternative formalisms for describing the
syntax are briefly discussed.

In English, the syntax for main verbs and the `auxiliary' verbs which
may accompany them follows two quasi-independent patterns of constraint
which interact in an interesting way.

The {\em primary pattern of constraint} may be expressed with this sequence of symbols,

\begin{center}
\begin{BVerbatim}
M H B B V,
\end{BVerbatim}
\end{center}

\noindent which should be interpreted in the following way:

\begin{itemize}

\item Each letter represents a category for a single word:

\begin{itemize}

\item `M' stands for `modal' verbs like `will', `can', `would' etc.

\item `H' stands for one of the various forms of the verb `to have'.

\item Each of the two instances of `B' stands for one of the various
forms of the verb `to be'.

\item `V' stands for the main verb which can be any verb except a modal verb.

\end{itemize}

\item The words occur in the order shown but any of the words may be
omitted.

\item \sloppy Questions of `standard' form follow exactly the same pattern as
statements except that the first verb, whatever it happens to be (`M',
`H', the first `B', the second `B' or `V'), precedes the subject noun
phrase instead of following it.

\end{itemize}

Here are two examples of the primary pattern with all of the words
included:

\begin{center}
\begin{BVerbatim}
It will have been being washed  
    M    H    H     B     V   
 
Will it have been being washed?  
 M       H    H     B     V   
\end{BVerbatim}
\end{center}

The {\em secondary constraints} are these:

\begin{itemize}

\item Apart from the modals, which always have the same form, the first
verb in the sequence, whatever it happens to be (`H', the first `B',
the second `B' or `V'), always has a `finite' form (the form it would
take if it were used by itself with the subject).

\item If an `M' auxiliary verb is chosen, then whatever follows it
(`H', first `B', second `B', or `V') must have an `infinitive' form
(i.e., the `standard' form of the verb as it occurs in the context `to
...', but without the word `to').

\item If an `H' auxiliary verb is chosen, then whatever follows it (the
first `B', the second `B' or `V') must have a past tense form such as
`been', `seen', `gone', `slept', `wanted' etc. In Chomsky's {\it
Syntactic Structures} \citeyearpar{chomsky_1957}, these forms were characterised as
{\em en} forms and the same convention has been adopted here.

\item If the first of the two `B' auxiliary verbs is chosen, then
whatever follows it (the second `B' or `V') must have an {\em ing} form,e.g., `singing', `eating', `having', `being' etc.

\item If the second of the two `B' auxiliary verbs is chosen, then
whatever follows it (only the main verb is possible now) must have a
past tense form (marked with {\em en} as above).

\item The constraints apply to questions in exactly the same way as they
do to statements.

\end{itemize}

Figure \ref{english_sentences} shows a selection of examples with the
dependencies marked.

\begin{figure}[!hbt]
\centering
\begin{BVerbatim}
           H------en  B2---------en
          ----    --  --         --
It  will  have  been  being  washed
    ----  ----  --      ---  ----
     M----inf   B1------ing   V

          B1------ing
          --      ---
Will  he  be  talking?
----      --  ----
 M-------inf   V

              V
            ------
They  have  finished
      ----        --
       H----------en
      fin

Are  they  gone?
---        ----
B2----------en
fin         V

         B1--------ing
         --        ---
Has  he  been  working?
---        --  ----
 H---------en   V
fin
\end{BVerbatim}
\caption{A selection of example sentences in English with
markings of dependencies between the verbs. {\em Key:} {\em M} = modal,
{\em H} = forms of the verb `have', {\em B1} = first instance of a form
of the verb `be', {\em B2} = second instance of a form of the verb
`be', {\em V} = main verb, {\em fin} = a finite form, {\em inf} = an
infinitive form, {\em en} = a past tense form, {\em ing} = a verb
ending in `ing'.}
\label{english_sentences}
\end{figure}

\subsection{Transformational grammar and English auxiliary verbs}

In Figure \ref{english_sentences} it can be seen that in many cases but
not all, the dependencies which have been described may be regarded as
discontinuous because they connect one word in the sequence to the
suffix of the following word thus bridging the stem of the following
word. Dependencies that are discontinuous can be seen most clearly in questions (e.g., the second, fourth and fifth sentences in Figure \ref{english_sentences}) where the verb before the subject influences the form of the verb that follows immediately after the subject. 

In {\em Syntactic Structures}, \citet{chomsky_1957} showed that this kind of
regularity in the syntax of English auxiliary verbs could be described
using transformational grammar. For each pair of symbols linked by
a dependency (`M inf', `H en', `B1 ing', `B2 en') the two symbols could
be shown together in the `deep structure' of a sentence and then moved
into their proper position or modified in form (or both) using
`transformational rules'.

This elegant demonstration argued persuasively in favour of TG compared
with alternatives which were available at that time. However, later
research has shown that the same kinds of regularities in the syntax of
English auxiliary verbs can be described quite well without recourse to
transformational rules, using definite clause grammars or other
systems which do not use that type of rule \citep[see, for example,][]{pereira_warren_1980, gazdar_mellish_1989}. An example showing how English auxiliary verbs
may be described using the definite clause grammar formalism may be found in
\citet[pp. 183-184]{wolff_1987}).

\subsection{English auxiliary verbs in the SP system}

Figure \ref{auxiliary_verbs_1} shows an `SP' grammar for English auxiliary verbs which exploits several of the ideas described earlier in this article. Figures
\ref{three_parsings_1}, \ref{three_parsings_2} and Figure \ref{three_parsings_3} show the
best alignments in terms of information compression for three different sentences produced
by the SP61 model using this grammar. In the following paragraphs,
aspects of the grammar and of the examples are described and discussed.

\begin{figure}[!hbt]
\fontsize{06.00pt}{07.20pt}
\centering
\begin{BVerbatim}
S ST NP #NP X1 #X1 XR #S (3000)
S Q X1 #X1 NP #NP XR #S (2000)
NP SNG it #NP (4000)
NP PL they #NP (1000)
X1 0 V M #V #X1 XR XH XB XB XV #S (1000)
X1 1 XH FIN #XH #X1 XR XB XB XV #S (900)
X1 2 XB1 FIN #XB1 #X1 XR XB XV #S (1900)
X1 3 V FIN #V #X1 XR #S (900)
XH V H #V #XH XB #S (200)
XB XB1 #XB1 XB #S (300)
XB XB1 #XB1 XV #S (300)
XB1 V B #V #XB1 (500)
XV V #V #S (5000)
M INF (2000)
H EN (2400)
B XB ING (2000)
B XV EN (700)
SNG SNG (2500)
PL PL (2500)
V M 0 will #V (2500)
V M 1 would #V (1000)
V M 2 could #V (500)
V H INF have #V (600)
V H PL FIN have #V (400)
V H SNG FIN has #V (200)
V H EN had #V (500)
V H FIN had #V (300)
V H ING hav ING1 #ING1 #V (400)
V B SNG FIN 0 is #V (500)
V B SNG FIN 1 was #V (400)
V B INF be #V (400)
V B EN be EN1 #EN1 #V (600)
V B ING be ING1 #ING1 #V (700)
V B PL FIN 0 are #V (300)
V B PL FIN 1 were #V (500)
V FIN wrote #V (166)
V INF 0 write #V (254)
V INF 1 chew #V (138)
V INF 2 walk #V (318)
V INF 3 wash #V (99)
V ING 0 chew ING1 #ING1 #V (623)
V ING 1 walk ING1 #ING1 #V (58)
V ING 2 wash ING1 #ING1 #V (102)
V EN 0 made #V (155)
V EN 1 brok EN1 #EN1 #V (254)
V EN 2 tak EN1 #EN1 #V (326)
V EN 3 lash ED #ED #V (160)
V EN 4 clasp ED #ED #V (635)
V EN 5 wash ED #ED #V (23)
ING1 ing #ING1 (1883)
EN1 en #EN1 (1180)
ED ed #ED (818)
\end{BVerbatim}
\caption{A grammar for the syntax of English auxiliary verbs.}
\label{auxiliary_verbs_1}
\end{figure}

\subsection{The primary constraints}

\begin{landscape}
\begin{figure}[!hbt]
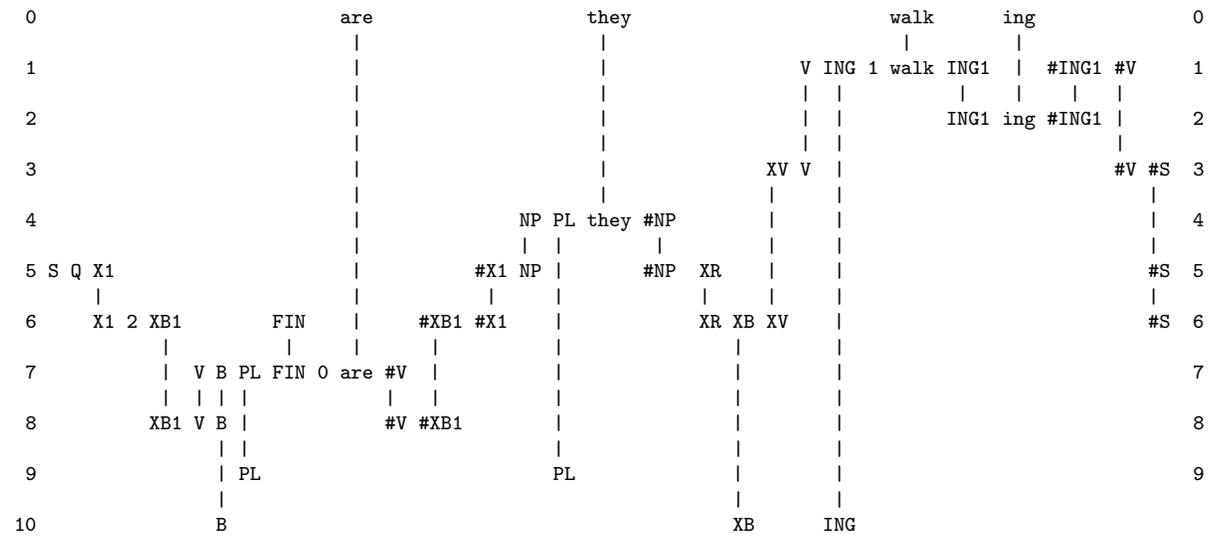

\fontsize{08.00pt}{09.60pt}
\centering
\begin{BVerbatim}
 0                           are                   they                       walk      ing              0
                              |                     |                          |         |                
 1                            |                     |                 V ING 1 walk ING1  |  #ING1 #V     1
                              |                     |                 |  |          |    |    |   |       
 2                            |                     |                 |  |         ING1 ing #ING1 |      2
                              |                     |                 |  |                        |       
 3                            |                     |              XV V  |                        #V #S  3
                              |                     |              |     |                           |    
 4                            |              NP PL they #NP        |     |                           |   4
                              |              |  |        |         |     |                           |    
 5 S Q X1                     |          #X1 NP |       #NP  XR    |     |                           #S  5
       |                      |           |     |            |     |     |                           |    
 6     X1 2 XB1        FIN    |     #XB1 #X1    |            XR XB XV    |                           #S  6
             |          |     |      |          |               |        |                                
 7           |  V B PL FIN 0 are #V  |          |               |        |                               7
             |  | | |            |   |          |               |        |                                
 8          XB1 V B |            #V #XB1        |               |        |                               8
                  | |                           |               |        |                                
 9                | PL                          PL              |        |                               9
                  |                                             |        |                                
10                B                                             XB      ING                       
\end{BVerbatim}
\caption{The best alignment found by SP61 with `are they
walk ing' in New and the grammar from Figures \ref{auxiliary_verbs_1} in Old.}
\label{three_parsings_3}
\end{figure}
\end{landscape}

The first line in the grammar is a sentence pattern for a statement
(marked with the symbol `ST') and the second line is a sentence pattern
for a question (marked with the symbol `Q'). Apart from these markers,
the only difference between the two patterns is that, in the statement
pattern, the symbols `X1 \#X1' follow the noun phrase symbols (`NP
\#NP'), whereas in the question pattern they precede the noun phrase
symbols. As can be seen in the examples in Figures
\ref{three_parsings_1}, \ref{three_parsings_2} and Figure \ref{three_parsings_3}, the pair of
symbols, `X1 \#X1', has the effect of selecting the first verb in the
sequence of auxiliary verbs and ensuring its correct position with
respect to the noun phrase. In Figure \ref{three_parsings_1} it follows
the noun phrase, while in Figures \ref{three_parsings_2} and Figure \ref{three_parsings_3}
it precedes the noun phrase.

\begin{landscape}
\begin{figure}[!hbt]
\fontsize{08.00pt}{09.60pt}
\centering
\begin{BVerbatim}
 0             it                            is                              wash    ed            0
               |                             |                                |      |              
 1             |                             |                        V EN 5 wash ED |  #ED #V     1
               |                             |                        | |         |  |   |  |       
 2             |                             |                        | |         ED ed #ED |      2
               |                             |                        | |                   |       
 3             |                             |                     XV V |                   #V #S  3
               |                             |                     |    |                      |    
 4             |                 B           |                     XV   EN                     |   4
               |                 |           |                     |                           |    
 5             |               V B SNG FIN 0 is #V                 |                           |   5
               |               | |  |   |       |                  |                           |     
 6             |           XB1 V B  |   |       #V #XB1            |                           |   6
               |            |       |   |           |              |                           |    
 7             |      X1 2 XB1      |  FIN         #XB1 #X1 XR XB  XV                          #S  7
               |      |             |                    |  |                                  |    
 8      NP SNG it #NP |             |                    |  |                                  |   8
        |   |      |  |             |                    |  |                                  |    
 9 S ST NP  |     #NP X1            |                   #X1 XR                                 #S  9
            |                       |                               
10         SNG                     SNG                                                            10
\end{BVerbatim}
\caption{The best alignment found by SP61 with `it is wash ed' in
New and the grammar from Figures \ref{auxiliary_verbs_1} in Old.}
\label{three_parsings_1}
\end{figure}
\end{landscape}

\begin{landscape}
\begin{figure}[!hbt]
\fontsize{06.00pt}{07.20pt}
\centering
\begin{BVerbatim}
 0                will               it                   have                          be     en                        brok     en             0
                   |                 |                     |                            |      |                          |       |               
 1                 |                 |                     |                            |  EN1 en #EN1                    |       |              1
                   |                 |                     |                            |   |      |                      |       |               
 2                 |                 |                     |                     V B EN be EN1    #EN1 #V                 |       |              2
                   |                 |                     |                     | | |                 |                  |       |               
 3                 |                 |                     |                 XB1 V B |                 #V #XB1            |       |              3
                   |                 |                     |                  |    | |                     |              |       |               
 4                 |                 |                     |                  |    | |                     |      V EN 1 brok EN1 |  #EN1 #V     4
                   |                 |                     |                  |    | |                     |      | |          |  |   |   |       
 5                 |                 |                     |                  |    | |                     |   XV V |          |  |   |   #V #S  5
                   |                 |                     |                  |    | |                     |   |    |          |  |   |      |    
 6                 |                 |                     |              XB XB1   | |                    #XB1 XV   |          |  |   |      #S  6
                   |                 |                     |              |        | |                         |    |          |  |   |      |    
 7          V M 0 will #V            |                     |              |        | |                         |    |          |  |   |      |   7
            | |        |             |                     |              |        | |                         |    |          |  |   |      |    
 8     X1 0 V M        #V #X1        |      XR XH          |          XB  XB       | |                         XV   |          |  |   |      #S  8
       |      |            |         |      |  |           |          |            | |                         |    |          |  |   |      |    
 9     |      |            |  NP SNG it #NP |  |           |          |            | |                         |    |          |  |   |      |   9
       |      |            |  |          |  |  |           |          |            | |                         |    |          |  |   |      |    
10 S Q X1     |           #X1 NP        #NP XR |           |          |            | |                         |    |          |  |   |      #S 10
              |                                |           |          |            | |                         |    |          |  |   |      |    
11            |                                |  V H INF have #V     |            | |                         |    |          |  |   |      |  11
              |                                |  | |  |       |      |            | |                         |    |          |  |   |      |    
12            |                                XH V H  |       #V #XH XB           | |                         |    |          |  |   |      #S 12
              |                                     |  |                           | |                         |    |          |  |   |           
13            |                                     |  |                           | |                         |    |         EN1 en #EN1       13
              |                                     |  |                           | |                         |    |                             
14            M                                     | INF                          | |                         |    |                           14
                                                    |                              | |                         |    |                             
15                                                  H                              | EN                        |    |                           15
                                                                                   |                           |    |                             
16                                                                                 B                           XV   EN                          16
\end{BVerbatim}
\caption{The best alignment found by SP61 with `will it have
be en brok en' in New and the grammar from Figures \ref{auxiliary_verbs_1} in Old (Part 1).}
\label{three_parsings_2}
\end{figure}
\end{landscape}

Each of the next four patterns in the grammar have the form `X1 ... \#X1
XR ... \#S'. The symbols `X1' and `\#X1' align with the same pair of
symbols in the sentence pattern. The symbols `XR ... \#S' encode the
remainder of the sequence of verbs.

The first `X1' pattern encodes verb sequences which start with a modal
verb (`M'), the second one is for verb sequences beginning with a
finite form of the verb `have' (`H'), the third is for sequences
beginning with either of the two `B' verbs in the primary sequence (see
below), and the last `X1' pattern is for sentences which contain a main
verb without any auxiliaries.

In the first of the `X1' patterns, the subsequence `XR ... \#S' encodes
the remainder of the sequence of auxiliary verbs using the symbols `XH
XB XB XV'. In a similar way, the subsequence `XR ... \#S' within each of
the other `X1' patterns encodes the verbs which follow the first verb
in the sequence.

Notice that the pattern `X1 2 XB1 FIN \#XB1 \#X1 XR XB XV \#S' can
encode sentences which start with the first `B' verb and also contains
the second `B' verb. And it also serves for any sentence which starts
with the first or the second `B' verb with the omission of the other
`B' verb. In the latter two cases, the `slot' between the symbols `XB'
and `XV' is left vacant. Figure \ref{three_parsings_1} illustrates the
case where the verb sequence starts with the first `B' verb with the
omission of the second `B' verb. Figure \ref{three_parsings_3}
illustrates the case where the verb sequence starts with the second `B'
verb (and the first `B' verb has been omitted).

\subsection{The secondary constraints}

The secondary constraints are represented using the patterns `M INF',
`H EN', `B XB ING' and `B XV EN'. Singular and plural dependencies are
marked in a similar way using the patterns `SNG SNG' and `PL PL'.

Examples appear in all three alignments in Figures
\ref{three_parsings_1}, \ref{three_parsings_2} and Figure \ref{three_parsings_3}. In every
case except one (row 4 in Figure \ref{three_parsings_1}), the patterns
representing secondary constraints appear in the bottom rows of the
alignment. These examples show how dependencies bridging arbitrarily
large amounts of structure, and dependencies that overlap each other,
can be represented with simplicity and transparency in the medium of
multiple alignments.

Notice, for example, how dependencies between the first and second verb
in a sequence of auxiliary verbs are expressed in the same way
regardless of whether the two verbs lie side by side (e.g., the
statement in Figure \ref{three_parsings_1}) or whether they are
separated from each other by the subject noun-phrase (e.g., the
question in Figure \ref{three_parsings_2} and in Figure
\ref{three_parsings_3}).  Notice, again, how the overlapping
dependencies in Figure \ref{three_parsings_2} and their independence from each other are
expressed with simplicity and clarity in the SP system.

Readers may wonder why the two patterns representing dependencies
between a `B' verb and whatever follows it (`B XB ING' and `B XV EN')
contain three symbols rather than two. One reason is that, when two (or
more) patterns begin with the same symbol (or sequence of symbols), the
scoring method for evaluating alignments requires that the two patterns
can be distinguished from each other by one (or more) symbols in each
pattern which does not include the terminal symbol in each pattern. A
second reason is that the second symbol in each pattern helps to
determine whether the `B' at the start of the pattern corresponds to
the first or the second `B' verb in the primary sequence:

\begin{itemize}

\item `B XB ING'. The inclusion of `XB' in this pattern means that the
`B' verb is the first of the two `B' verbs in the primary sequence and
the following verb must be `ING'.

\item `B XV EN'. The inclusion of `XV' in this pattern means that the
`B' verb may be the first or the second of the two `B' verbs. However,
since the first case is already covered by 'B XB ING', this pattern
covers the constraint between the second `B' verb and verbs of the
category `EN'.

\end{itemize}

\index{grammar!English auxiliary verbs|)}

\section{Cross serial dependencies}\label{cross_serial_dependencies}

\index{grammar!cross-serial dependencies|(}

Recursion appears again in a syntactic phenomenon---`cross-serial dependency'---found in languages like Swiss German and Dutch. This kind of dependency is difficult or impossible to represent using a basic context-free phrase-structure grammar  although it can be expressed in a reasonably straightforward manner with augmented forms of phrase-structure grammar that provide Prolog-like facilities for carrying information from one part of a structure to another and the construction of Lisp-like lists.

The example shown here is an attempt to represent cross-serial dependencies in the SP system. It is not correct but it has been included in this chapter because the problem is interesting and may suggest the need for changes in the SP system. An alternative possibility is that I have simply failed to see how this kind of structure should be represented in the SP system. I would be very pleased to hear any insights that readers may have.

In Swiss German, a sentence which means ``We let the children help Hans paint the house'' 
may be expressed as {\em Mer d'chind em Hans es huus lond halfed aastriiche}. This corresponds approximately with the following sequence of words in English: ``We the children Hans the house let help paint'' \citep[see][section 2.3]{borsley_1996}. In the standard English form, this type of sentence is clearly recursive because it is easily extended, without limit, to become ``We let the children help Hans help Jim help Mary ... paint the house'', and so on. The same appears to be true of the cross-serial form.

\begin{figure}[!hbt]
\centering
\includegraphics[width=0.9\textwidth]{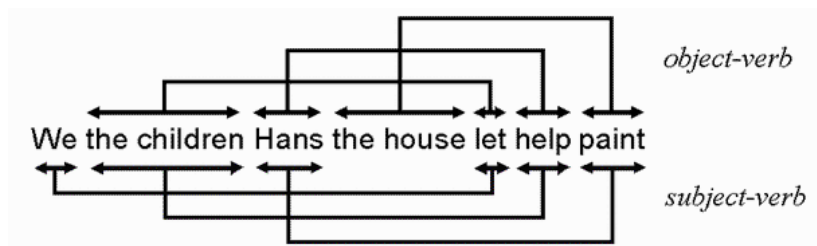}
\caption{A schematic representation of cross-serial dependencies in the sentence ``We the children Hans the house let help paint''. Dependencies between subjects and verbs are shown below the sentence and dependencies between objects and verbs are shown above.}
\label{cross_serial_schema}
\end{figure}

In this type of structure, the dependencies between verbs and their subjects and between the 
verbs and their objects are discontinuous and they overlap each other as can be seen
schematically in Figure \ref{cross_serial_schema}.

In both the English form and the cross-serial form, the relationship between one sentence 
and the next in the recursive sequence is curious because the object of every sentence
except the last becomes the subject of the next.

Figure \ref{cross_serial_grammar} shows a grammar for sentences like ``We the children
Hans the house let help paint'' except that ``the children'' has been replaced
by ``them'' and ``the house'' has been replaced by ``it''. These substitutions have
been made so that alignments are more easily displayed within the confines of
one page. Figure \ref{cross_serial_alignment} shows one of the best alignments created by the SP61 model with `we them hans it let help paint' in New and the grammar in Figure \ref{cross_serial_grammar} in Old.

\begin{figure}[!hbt]
\centering
\begin{BVerbatim}
S NP #NP S NP #NP : : #S V #V #S (1000)
NP 0 him #NP (500)
NP 1 hans #NP (200)
NP 2 them #NP (300)
NP 3 we #NP (150)
NP 4 it #NP (250)
V 0 help #V (200)
V 1 let #V (100)
V 2 encourage #V (300)
V 3 run #V (100)
V 4 paint #V (125)
V 5 walk #V (175)
\end{BVerbatim}
\caption{A grammar for cross-serial dependencies.}
\label{cross_serial_grammar}
\end{figure}

\begin{figure}[!hbt]
\fontsize{05.00pt}{06.00pt}
\centering
\begin{BVerbatim}
0       1     2     3     4     5       6     7      8     9      10    11    12    13    14    15  

                                                                                    S               
                                                                                    NP              
                                                                                    #NP             
              S ------------------------------------------------------------------- S               
              NP ------------------------------------------------------------------ NP              
              #NP ----------------------------------------------------------------- #NP             
              S --------- S                                                                         
              NP -- NP -- NP                                                                        
                    3                                                                               
we ---------------- we                                                                              
              #NP - #NP - #NP                                                                       
                          S ------------------------ S                                              
                          NP ----------------------- NP --------------------------------------- NP  
                                                                                                2   
them ------------------------------------------------------------------------------------------ them
                          #NP ---------------------- #NP -------------------------------------- #NP 
                                        S ---------- S                                              
                                        NP --------- NP                                             
                                        #NP -------- #NP                                            
                                        S ----------------------- S                                 
                                        NP --------------- NP --- NP                                
                                                           1                                        
hans ----------------------------------------------------- hans                                     
                                        #NP -------------- #NP -- #NP                               
                                                                  S --------- S                     
                                                                  NP -- NP -- NP                    
                                                                        4                           
it -------------------------------------------------------------------- it                          
                                                                  #NP - #NP - #NP                   
                                                                              S --------- S         
                                                                              NP -------- NP        
                                                                              #NP ------- #NP       
                                                                                          S         
                                                                                          NP        
                                                                                          #NP       
                                                                                          :         
                                                                              : --------- :         
                                                                  : --------- :                     
                                        : ----------------------- :                                 
                                        : ---------- :                                              
                          : ------------------------ :                                              
              : --------- :                                                                         
              : ------------------------------------------------------------------- :               
                                                                                    :               
                                                                                    #S              
                                                                                    V               
                                                                                    #V              
              #S ------------------------------------------------------------------ #S              
        V --- V                                                                                     
        1                                                                                           
let --- let                                                                                         
        #V -- #V                                                                                    
              #S -------- #S                                                                        
                          V                                                                         
                          #V                                                                        
                          #S ----------------------- #S                                             
                                              V ---- V                                              
                                              0                                                     
help ---------------------------------------- help                                                  
                                              #V --- #V                                             
                                        #S --------- #S                                             
                                V ----- V                                                           
                                4                                                                   
paint ------------------------- paint                                                               
                                #V ---- #V                                                          
                                        #S ---------------------- #S                                
                                                                  V                                 
                                                                  #V                                
                                                                  #S -------- #S                    
                                                                              V                     
                                                                              #V                    
                                                                              #S -------- #S        
                                                                                          V         
                                                                                          #V        
                                                                                          #S        

0       1     2     3     4     5       6     7      8     9      10    11    12    13    14    15  
\end{BVerbatim}
\caption{One of the best alignments produced by the SP61 model with `we them hans it let help paint' in New and the grammar in Figure \ref{cross_serial_grammar} in Old.}
\label{cross_serial_alignment}
\end{figure}

This alignment expresses the recursive nature of the structure but it fails to capture the `subject object ... verb' dependency that exists between `we, `them' and `let', between `them', `hans' and `help' or between `hans', `it' and `paint'.%
\index{grammar!syntactic dependencies|)}
\index{grammar!cross-serial dependencies|)}

\section{The integration of syntax with semantics}

\index{grammar!semantics|(}

\begin{figure}[!hbt]
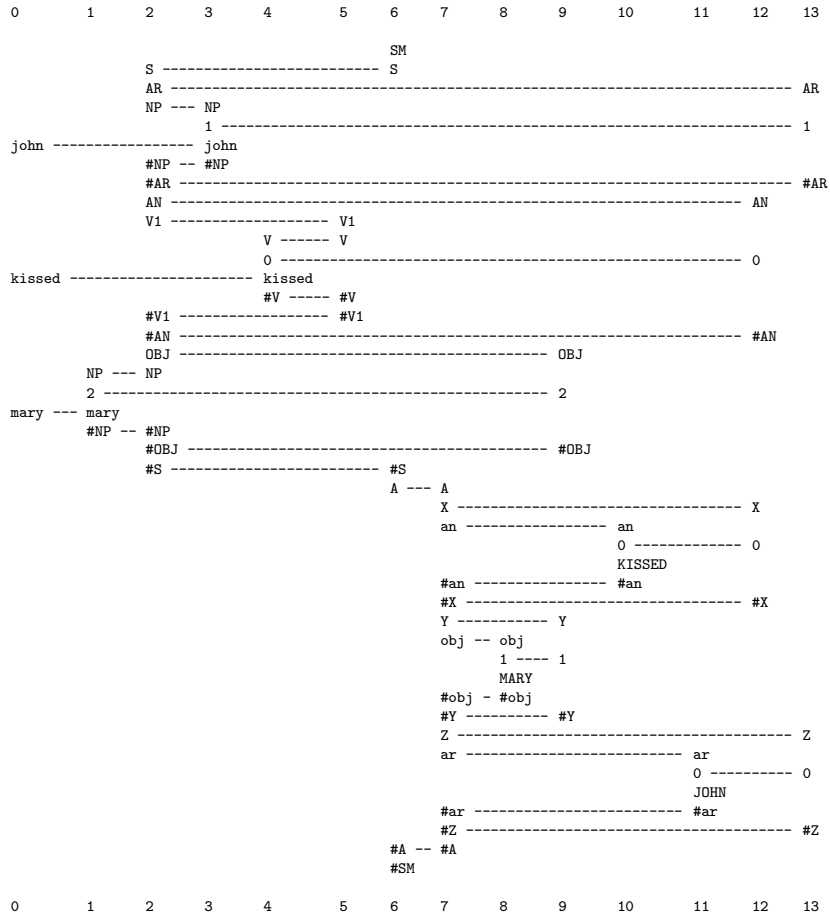

\fontsize{06.00pt}{07.20pt}
\centering
\begin{BVerbatim}
0        1      2      3      4        5     6     7      8      9      10       11     12    13 

                                             SM                                                  
                S -------------------------- S                                                   
                AR -------------------------------------------------------------------------- AR 
                NP --- NP                                                                        
                       1 -------------------------------------------------------------------- 1  
john ----------------- john                                                                      
                #NP -- #NP                                                                       
                #AR ------------------------------------------------------------------------- #AR                
                AN -------------------------------------------------------------------- AN       
                V1 ------------------- V1                                                        
                              V ------ V                                                         
                              0 ------------------------------------------------------- 0        
kissed ---------------------- kissed                                                             
                              #V ----- #V                                                        
                #V1 ------------------ #V1                                                       
                #AN ------------------------------------------------------------------- #AN      
                OBJ -------------------------------------------- OBJ                             
         NP --- NP                                                                               
         2 ----------------------------------------------------- 2                               
mary --- mary                                                                                    
         #NP -- #NP                                                                              
                #OBJ ------------------------------------------- #OBJ                            
                #S ------------------------- #S                                                  
                                             A --- A                                             
                                                   X ---------------------------------- X        
                                                   an ----------------- an                       
                                                                        0 ------------- 0        
                                                                        KISSED                   
                                                   #an ---------------- #an                      
                                                   #X --------------------------------- #X       
                                                   Y ----------- Y                               
                                                   obj -- obj                                    
                                                          1 ---- 1                               
                                                          MARY                                   
                                                   #obj - #obj                                   
                                                   #Y ---------- #Y                              
                                                   Z ---------------------------------------- Z  
                                                   ar -------------------------- ar              
                                                                                 0 ---------- 0  
                                                                                 JOHN            
                                                   #ar ------------------------- #ar             
                                                   #Z --------------------------------------- #Z 
                                             #A -- #A                                            
                                             #SM                                                 

0        1      2      3      4        5     6     7      8      9      10       11     12    13 
\end{BVerbatim}
\caption{The best alignment created by SP61 with the sentence `john kissed mary' in New and a grammar in Old that represents natural language syntax, semantics and their integration.}
\label{syntax_semantics_alignment_1}
\end{figure}

In keeping with the remarks in Section \ref{representation_of_knowledge}, it has been anticipated that the SP system would not only support the representation of syntactic and non-syntactic (`semantic') kinds of knowledge but that it would facilitate their integration. A preliminary example of how this might be done is shown in Figure \ref{syntax_semantics_alignment_1}. This is the best alignment produced by SP61 with `john kissed mary' as the New pattern and a grammar in Old that contains patterns representing syntax, `semantics' and the connections between them. The scare quotes are intended to indicate that the representations of semantic structures in this example are, at best, crude. That point made, the quote marks for `semantics' or `meanings' will be dropped in the remainder of this section.

In the alignment, the sentence appears in column 0. In the remaining columns, Old patterns with the main r{\^o}les are as follows:

\begin{itemize}

\item The pattern in column 2 represents the overall syntactic structure of the sentence: `S AR NP \#NP \#AR AN V1 \#V1 \#AN OBJ NP \#NP \#OBJ \#S'. Notice that this pattern differs from comparable patterns shown in previous examples because each constituent within the pattern is marked with its semantic r{\^o}le. Thus the first noun phrase (`NP \#NP') is enclosed by the pair of symbols `AR ... \#AR' (representing the `actor' r{\^o}le), the verb (`V1 \#V1') is marked as an `action' with the symbols `AN ... \#AN', and the second noun phrase is marked as an `object' (`OBJ ... \#OBJ').

\item In column 7, the pattern `A X an \#an \#X Y obj \#obj \#Y Z ar \#ar \#Z \#A' may be seen as a generalised description of the association between an `action' (`an \#an'), the `object' of the action (`obj \#obj') and the `actor' or performer of the action (`ar \#ar'). Notice that the order in which these concepts are specified is different from the order of the corresponding markers in the syntax pattern. Notice also that these three slots are also marked, in order, as `X ... \#X', `Y ... \#Y' and `Z ... \#Z'. The reason for this marking will be seen shortly.

\item In column 6, the pattern `SM S \#S A \#A \#SM' provides a link between the pattern in column 2 representing the syntactic structure of the sentence (`S ... \#S') and the action-object-actor pattern in column 7 (`A ... \#A').

\item In columns 9, 12 and 13 are three more patterns that link the syntax with the semantics. In column 9, the pattern `OBJ 2 \#OBJ Y 1 \#Y' connects `NP 2 mary \#NP' in the object position of the syntax with `MARY' in the `obj \#obj' slot of the semantic structure. Here, `MARY' is intended to represent some kind of conceptual structure that is the meaning of the word `mary'. More precisely, it is intended to represent a `code' for that structure (see Section \ref{codes_meanings_and_nl_production}, below). In a similar way, the pattern in column 12 connects `kissed' with `KISSED' in the `an \#an' semantic slot; and the pattern in column 13 connects `john' with `JOHN' in the `ar \#ar' semantic slot. 

\end{itemize}

The key idea in this example is that the SP system allows information to be carried from the syntactic part of the knowledge structure to the semantic part and it allows the ordering of information to change from one part to the other. There seems no reason to suppose that this basic capability could not also be applied to examples in which the syntax and the semantics are more elaborate and more realistic.

\subsection{Codes, meanings and the production of language from meanings}\label{language_production_from_meanings}\label{codes_meanings_and_nl_production}

\index{language!production|(}

In Section \ref{decompression_by_compression}, we saw how the SP system could be used to produce a sentence, given a short code for that sentence supplied as New. That example shows in general terms how the system may be used for language production as well as language analysis but it seems unlikely that there would be many applications where there would be a requirement for the production of sentences purely in terms of their syntax and encodings of that syntax. In practice, it is more likely that one would wish to create sentences on the basis of intended {\em meanings}.

One possibility is that meanings might serve as codes for syntax and be used for language production in the way described in Section \ref{decompression_by_compression}. In support of this view, we seem to remember what people have said in terms of meanings that were expressed rather than the words that were used to express them. And we can often reconstruct the words that people have used from the meanings that we remember---although there may be an element of lossy compression here because the reconstruction is not always accurate.

Although this view has attractions, it conflicts with the idea described in Section \ref{compression_in_nl} that many words in natural language, especially `content' words like `table', `house' etc, may be seen as relatively short identifiers for relatively complex semantic concepts and that a sentence may be seen as a compressed representation of its meanings.

How might these two views be reconciled? A working hypothesis in this programme of research is that there are {\em internal} codes for non-syntactic semantic concepts that are distinct from the words and other surface forms of natural language that serve as {\em external} codes for meanings in written or spoken communication. In this view, when we remember what someone has said in terms of meanings, we remember internal codes for those meanings, distinct from the concepts themselves. If someone tells us that ``The horse is in the field'', it is unlikely that we copy our entire knowledge of `the horse' into association with our knowledge of `the field'. It is much more likely that we construct a representation of that situation in terms of {\em codes} for those concepts, not the concepts themselves. And, by hypothesis, those codes are distinct from the verbal `codes' used in the original sentence.

\begin{figure}[!hbt]
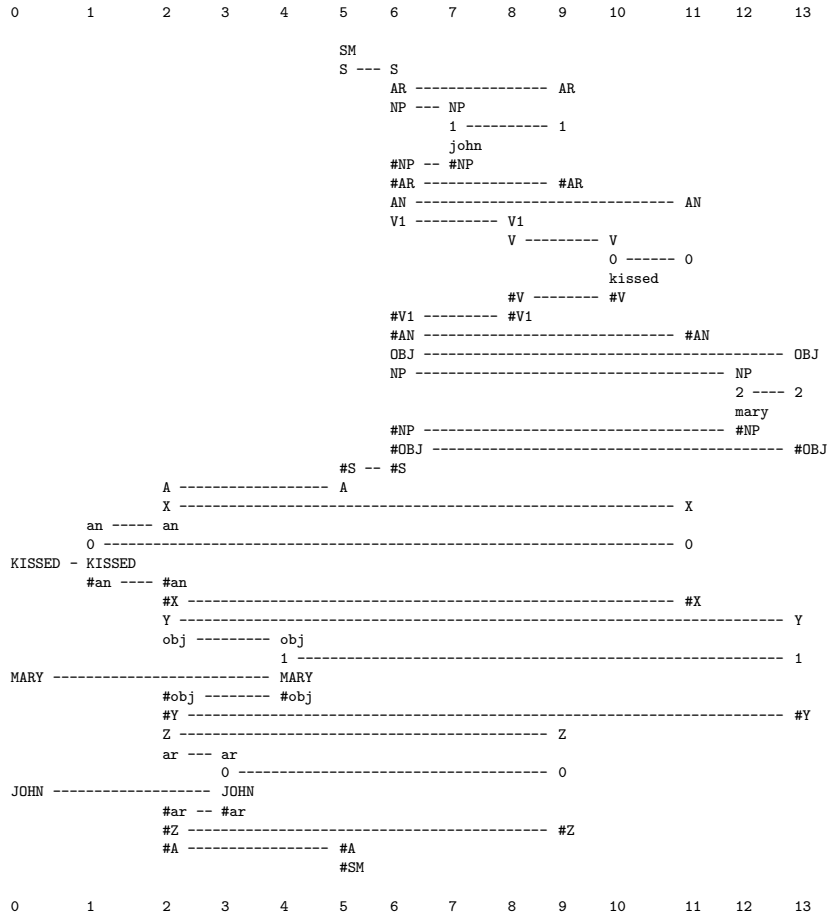

\fontsize{06.00pt}{07.20pt}
\centering
\begin{BVerbatim}
0        1        2      3      4      5     6      7      8     9     10       11    12     13  

                                       SM                                                        
                                       S --- S                                                   
                                             AR ---------------- AR                              
                                             NP --- NP                                           
                                                    1 ---------- 1                               
                                                    john                                         
                                             #NP -- #NP                                          
                                             #AR --------------- #AR                             
                                             AN ------------------------------- AN               
                                             V1 ---------- V1                                    
                                                           V --------- V                         
                                                                       0 ------ 0                
                                                                       kissed                    
                                                           #V -------- #V                        
                                             #V1 --------- #V1                                   
                                             #AN ------------------------------ #AN              
                                             OBJ ------------------------------------------- OBJ 
                                             NP ------------------------------------- NP         
                                                                                      2 ---- 2   
                                                                                      mary       
                                             #NP ------------------------------------ #NP        
                                             #OBJ ------------------------------------------ #OBJ
                                       #S -- #S                                                  
                  A ------------------ A                                                         
                  X ----------------------------------------------------------- X                
         an ----- an                                                                             
         0 -------------------------------------------------------------------- 0                
KISSED - KISSED                                                                                  
         #an ---- #an                                                                            
                  #X ---------------------------------------------------------- #X               
                  Y ------------------------------------------------------------------------ Y   
                  obj --------- obj                                                              
                                1 ---------------------------------------------------------- 1   
MARY -------------------------- MARY                                                             
                  #obj -------- #obj                                                             
                  #Y ----------------------------------------------------------------------- #Y  
                  Z -------------------------------------------- Z                               
                  ar --- ar                                                                      
                         0 ------------------------------------- 0                               
JOHN ------------------- JOHN                                                                    
                  #ar -- #ar                                                                     
                  #Z ------------------------------------------- #Z                              
                  #A ----------------- #A                                                        
                                       #SM                                                       

0        1        2      3      4      5     6      7      8     9     10       11    12     13  
\end{BVerbatim}
\caption{The best alignment created by SP61 with `KISSED MARY JOHN' in New and the same grammar in Old as was used for the example shown in Figure \ref{syntax_semantics_alignment_1}.}
\label{syntax_semantics_alignment_2}
\end{figure}

Figure \ref{syntax_semantics_alignment_2} shows how, via the building of an multiple alignment, a sentence may be derived from the pattern `KISSED MARY JOHN', representing an `internal' code for the meaning to be expressed. The alignment is the best alignment created by SP61 with that pattern in New and the same grammar in Old as was used for the example in Figure \ref{syntax_semantics_alignment_1}. In the top part of the alignment, the words `john', `kissed', and `mary' appear, in that order. If we strip out the `service' symbols in the alignment, we have the sentence corresponding to the semantic representation in column 0.

\index{grammar!semantics|)}\index{language!production|)}

\section{Conclusion}

This chapter has shown how the SP system may be applied to the representation of natural language structures and the processing of those structures in the analysis and production of sentences. The framework can accommodate recursion, ambiguity and context-sensitive aspects of natural language syntax and it shows promise for the integration of syntax with semantics.

Outstanding questions include how best to represent discontinuous dependencies in syntax, and whether or how the system may handle cross-serial dependencies in syntax.

%% file: rr.tex
\chapter[Recognition and Retrieval]{Pattern Recognition and Information Retrieval}\label{rr_chapter}

\index{perception!recognition|(}\index{information!retrieval|(}

\section{Introduction}

From at least as far back as the 1970s, it has been recognised that pattern recognition may be understood as information compression \citep{watanabe_article_1972}. At first sight, pattern recognition and information compression may seem unrelated but a little reflection shows the connection. In order to recognise something, incoming sensory data must be matched with a stored `engram' or record of the thing to be recognised. It is possible, of course, that the incoming data is stored independently of the record with which it has been matched. However, it seems much more likely that we merge or unify the sensory data with the stored record and encode it in terms of that record. This achieves information compression as described in Section \ref{ic_repetition_of_patterns}.

This chapter includes a (shortish) section on information retrieval because there are many similarities between pattern recognition and information retrieval. When we recognise something, we necessarily retrieve stored information about the thing that we have recognised. Any kind of `query' must make a connection with stored information in much the same way that a pattern to be recognised must connect with stored records of things that may be recognised. This can be seen most clearly in the case of `query-by-example' in which the query has the same general form as the stored records but with gaps where information is needed.

As we shall see in this chapter and the one that follows, recognition and retrieval are intimately related to (probabilistic) reasoning. This three-way connection may be referred to as `recognition-retrieval-and-reasoning'.

The chapter presents a selection of examples of the ways in which pattern recognition and information retrieval may be modelled in the SP system, including simple `fuzzy' pattern recognition, best-match information retrieval, and indirection in information retrieval, recognition with class-inclusion relations and part-whole relations, polythetic (family resemblance) concepts, and medical diagnosis.

\section{Scene analysis, fuzzy pattern recognition, best-match information retrieval and multiple alignment}\label{scene_analysis_recognition_retrieval}

\index{scene analysis|(}

A prominent feature of human perception is our ability to recognise things despite distortions or `noise' of various kinds. Consider, for example, how we perceive a typical scene. We may see a tree, behind which is parked a car which is itself in front of a house. With no conscious effort we immediately recognise the car as a car and the house as a house despite the fact that part of the car is obscured by the tree and part of the house is obscured by the tree and the car.

Not only can we recognise a whole object or pattern from a sub-set of its parts but, within 
wide limits, it normally does not matter very much which parts we see and which are obscured. We can also cope with other kinds of distortion in things that we recognise. Even if the car had an extra wheel at the front or back, we would probably still recognise it as a car provided it retained a reasonable number of car-like features. And we would still recognise a tree as a tree even if some of the branches had been replaced by balloons. In short, we have a robust ability to recognise things in a `fuzzy' manner that can cope with errors of omission, commission or substitution.

\begin{figure}[!hbt]
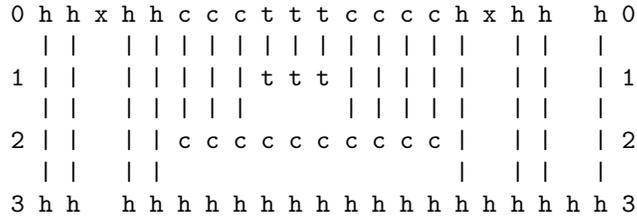

\centering
\begin{BVerbatim}
0 h h x h h c c c t t t c c c c h x h h   h 0
  | |   | | | | | | | | | | | | |   | |   |
1 | |   | | | | | t t t | | | | |   | |   | 1
  | |   | | | | |       | | | | |   | |   |
2 | |   | | c c c c c c c c c c |   | |   | 2
  | |   | |                     |   | |   |
3 h h   h h h h h h h h h h h h h h h h h h 3
\end{BVerbatim}
\caption{A multiple alignment illustrating schematically the recognition of a house, car and tree. The top row represents the scene as it is projected onto the retina of one eye. Row 1 (`t') represents a record in memory for the appearance of a tree. Rows 2 and 3 represent similar records for a car (`c') and a house (`h'). The two instances of `x' in the figure represent errors of addition and substitution.}
\label{house_tree_car_figure}
\end{figure}

This process of recognising objects in a scene using information that is incomplete or distorted---so familiar and so apparently effortless---may be interpreted in terms of multiple alignment. Figure \ref{house_tree_car_figure} shows a multiple alignment which represents, schematically, how the scene described above is recognised.

The symbols `t', `c' and `h' represent elements of the tree, car and house, respectively. 
The top row represents an image of the scene as it appears on the retina of one eye. The three rows below represent three stored patterns corresponding to the three types of object. The two instances of `x' represent errors of addition and substitution. There is also an error of omission on the right.

There is, of course, much more to the recognition process than is conveyed by the figure. 
But, for recognition to occur, there must be a mapping between elements of the image and elements of the stored patterns. The multiple alignment shown in Figure \ref{house_tree_car_figure} captures the overall structure of that mapping.

Current computer models of the SP system are restricted to one-dimensional patterns. For applications like scene analysis, the SP concepts would need to be generalised to patterns in two dimensions with corresponding developments in the computer models. This is something for the future (see Chapter \ref{future_chapter}).%
\index{scene analysis|)}

\subsection{Fuzzy recognition of words}\label{fuzzy_pattern_recognition}

Given a dictionary of correctly-spelled words, SP61 is able to recognise incorrectly-spelled words in fuzzy manner much like a system for spelling checking and correction. However, by contrast with most spelling checkers, the model can bridge arbitrarily large sequences of unmatched letters. 

In the example shown here, words are treated as simple flat patterns without any internal structure. This is not strictly correct (many words may be seen to contain more or less discrete `morphemes' and there are always smaller structures such as `phonemes') but for a preliminary illustration of fuzzy pattern recognition in the SP system, it is convenient to treat words as simple patterns. There are many examples throughout this book (starting with the examples in Section \ref{framework_examples_section}) that demonstrate how the system can accommodate patterns with rich internal structure.

SP61 has been run using the thoroughly mis-spelled `word', `e p e r i m a n t a t x p i u n' as New and, as Old, a small dictionary of words that vary in their similarity with that pattern. Each word in Old was assigned a notional frequency of occurrence so that relative values corresponded roughly with their subjective rate of use in ordinary English.

Figure \ref{fuzzy_recognition_figure} shows a selection of the multiple alignments formed by SP61 in order of their compression scores (calculated as described in Section \ref{ma_evaluation}). Notice that the system is able to find good matches between patterns despite the omission, addition or substitution of symbols. 

\begin{figure}[!hbt]
\centering
\begin{BVerbatim}
0      e   p e r i m a n t a t x p i u n   0
       |   | | | | |   | | | |     |   |  
1 < E3 e x p e r i m e n t a t     i o n > 1

(a) CD = 926.18

0            e p e r i m       a n t a t x p i u n   0
             |     | | |       |       |     |   |  
1 < E3 e x p e     r i m e n t a       t     i o n > 1

(b) CD = 573.08

0 e          p e r i m a n t a t x p i u n   0
             |             | | |     |   |  
1 < C1 c o m p u           t a t     i o n > 1

(c) CD = 467.21

0 e p e        r i m a n t a t x p i u n   0
               |   |       | |     |   |  
1 < I3 i n f o r   m       a t     i o n > 1

(d) CD = 434.59

0 e p e r i m a n t a t x p i u n                 0
                              | |                
1 < U1                        u n t a n g l e s > 1

(e) CD = 182.38
\end{BVerbatim}
\caption{The best four multiple alignments formed by SP61 between `e p e r i m a n t a t x p i u n' and words in a small dictionary. The $CD$ for each multiple alignment is shown below the multiple alignment.}
\label{fuzzy_recognition_figure}
\end{figure}

It is pertinent to point out that, in the small dictionary of words which was used as Old, the 
words `e x p e r i m e n t a t i o n' and `c o m p u t a t i o n' were assigned the same notional frequency of occurrence (10,000). Thus the relatively high $CD$ of multiple alignment (a) compared with multiple alignment (c) is not due to differences in frequency between the two words from Old. It is due to the fact that the best multiple alignment encodes more symbols from New than the other multiple alignment.

In general, these results accord with our intuition that the `correct' form of `e p e r i m a n t a t x p i u n' is `e x p e r i m e n t a t i o n', that `c o m p u t a t i o n' and `i n f o r   m a t i o n' are similar in some respects, and that `u n t a n g l e s' is much less similar.

\section{Best-match information retrieval}

SP61 can also be used for the retrieval of information from a `database' in the manner of query-by-example. Figure \ref{best_match_ir_figure} shows the best three multiple alignments found with a `query' pattern in New (`Name John Jones \#Name Address New \#Address') and patterns in Old that are analogous to records in a database.

\begin{figure}[!hbt]
\fontsize{08.00pt}{09.60pt}
\centering
\begin{BVerbatim}
0          1               0          1              0          1         

Name ----- Name            Name ----- Name           Name ----- Name      
John ----- John            John ----- John           John       Mary      
Jones ---- Jones           Jones      Smith          Jones ---- Jones     
#Name ---- #Name           #Name ---- #Name          #Name ---- #Name     
Address -- Address         Address -- Address        Address -- Address   
           54                         6              New        20        
New ------ New             New ------ New                       Old       
           Road                       Street                    Street    
#Address - #Address        #Address - #Address       #Address - #Address  
           Town                       Town                      Town      
           Anytown                    Oldsville                 Countytown
           #Town                      #Town                     #Town     
           County                     County                    County    
           Coastshire                 Ruralshire                Urbanshire
           #County                    #County                   #County   
           Telephone                  Telephone                 Telephone 
           05718                      01234                     09876     
           638945                     567890                    543210    
           #Telephone                 #Telephone                #Telephone

0          1               0          1              0          1         

(a) CD = 558.41           (b) CD = 470.11            (c) CD = 381.81
\end{BVerbatim}
\caption{The three best multiple alignments found by SP61 between the New pattern `Name John Jones \#Name Address New \#Address' in column 0 (the `query' pattern) and three different Old patterns in column 1 (`records' in a `database'). The $CD$ for each multiple alignment is shown at the bottom.}
\label{best_match_ir_figure}
\end{figure}

As in a conventional database, retrieval of a complete record from an incomplete query enables one to read off information that was not in the query. From the first multiple alignment in the figure, we can obtain the full street address for John Jones, his home town and county, and his telephone number.

\subsection{Indirection in information retrieval}\label{indirection_in_ir}

A common problem with the simpler kinds of system for information retrieval is that relevant material will not be found if, in the query which is used to initiate the search, the query term (or terms) does not appear in the index, title or body of that relevant material. For example, the focus of interest may be `parallel processing', but this term will not retrieve relevant material if that material is indexed by or uses the term `concurrency' instead of `parallel processing'.

Figure \ref{thesaurus_knowledge_base} shows a small knowledge base in which the first four patterns represent bodies of information about `sleep', `theory of knowledge', `luggage' and `concurrency'. At the bottom is shown four patterns from a thesaurus where each pattern contains one of the four terms just mentioned, together with one or more synonyms for that term. 

\begin{figure}[!hbt]
\centering
\begin{BVerbatim}
sleep ... information_about_sleep ... #sl (500)
theory of knowledge ... information_about_TOK ... #tk (800)
luggage ... information_about_luggage ... #lg (200)
concurrency ... information_about_concurrency ... #cy (400)
...
sleep #sl nap #np kip #kp (1)
epistemology #ey theory of knowledge #tk (1)
luggage #lg baggage #bg (1)
concurrency #cy parallel processing #pp (1)
\end{BVerbatim}
\caption{A small knowledge base, including patterns representing entries in a thesaurus.}
\label{thesaurus_knowledge_base}
\end{figure}

The best multiple alignment formed with this knowledge base in Old and the pattern `parallel 
processing' in New is shown at the top of Figure \ref{indirect_retrieval_alignments}. Below it is the one other multiple alignment in the reference set of multiple alignments for the best multiple alignment. The relative probabilities of the two multiple alignments are 0.9878 and 0.0122 respectively.

\begin{figure}[!hbt]
\fontsize{09.00pt}{10.80pt}
\centering
\begin{BVerbatim}
0                 parallel processing 0
                     |         |     
1 concurrency #cy parallel processing 1

(a)

0                                                parallel processing 0
                                                    |         |     
1 concurrency                                #cy parallel processing 1
       |                                      |                     
2 concurrency ... info_about_concurrency ... #cy                     2

(b)
\end{BVerbatim}
\caption{The best multiple alignment and the one other multiple alignment in its reference set formed by SP61 with the patterns from Figure \ref{thesaurus_knowledge_base} in Old and the pattern `parallel processing' in New.}
\label{indirect_retrieval_alignments}
\end{figure}

The best multiple alignment shows that `concurrency' is associated with `parallel 
processing'. This is not entirely without interest or usefulness but it does not tell us anything about the technical or other aspects of parallel processing. However, the second multiple alignment in 
Figure \ref{indirect_retrieval_alignments} contains a pattern with `information\_about\_concurrency' which, we may suppose is much closer to what we are looking for.

Although the relative probability of the second multiple alignment is only 0.0122 (and likewise for 
the the pattern `concurrency ... information\_about\_concurrency ... \#cy' and the symbol 
`information\_about\_concurrency'), this is the best of the `sub-optimal' inferences made by the 
system.

\section{Class-inclusion relations, part-whole relations and inheritance of attributes}\label{class_part_inheritance}

\index{class!hierarchy|(}\index{part-whole hierarchy|(}\index{inheritance of attributes|(}\index{object|(}

Figure \ref{animal_classes_figure} shows some patterns representing a hierarchy of classes from `animal' at the top level through the classes `mammal' and `bird' to relatively low-level classes such as `cat' and `robin'. At the lowest level are the individual animals `Tibs' and `Tweety'.

As in object-oriented computer languages, each class in the figure is linked to its parent 
class (if any) by including the relevant symbol or symbols for the parent class in the definition of the child class. The pattern for `mammal' contains the symbols `A' and `\#A'. These match the ID-symbols for the pattern that describes the class `animal' and they express the fact that a mammal is an animal. Likewise, the pattern for `cat' contains the symbols `M' and `\#M' showing that a cat is a mammal. And the pattern for `Tibs' contains the symbols `C' and `\#C' showing that it is a cat. 

Apart from symbols of that kind, each pattern contains symbols representing attributes of the corresponding class. An animal (in this simplified scheme) has a head, body and legs and it eats, breathes and has senses. Apart from being an animal, a mammal is furry and warm-blooded and, apart from being a mammal, a cat has such attributes as carnassial teeth and retractile claws. Three dots in each pattern (`...') show that there are other attributes that have been omitted to avoid cluttering the example unnecessarily. 

\begin{figure}[!hbt]
\centering
\begin{BVerbatim}
!A animal head #head body #body legs #legs eats
     breathes has-senses ... !#A
!M mammal A #A furry warm-blooded ... !#M
!C cat M !head carnassial-teeth !#head !legs
     retractile-claws !#legs #M purrs ... !#C
!T Tibs C !body white-bib !#body #C tabby ... !#T
!R reptile A #A cold-blooded scaly-skin ... !#R
!B bird A body wings feathers #body #A  can-fly ... !#B
!Rb robin B #B red-breast ... !#Rb
!Tw Tweety R #R ... !#Tw
\end{BVerbatim}
\caption{Patterns representing the class `animal' with some of its sub-classes, including individual animals `Tibs' and `Tweety'.}
\label{animal_classes_figure}
\end{figure}

\subsection{Class-inclusion relations, part-whole relations and multiple alignment}\label{class_inclusion_part_whole}

\begin{figure}[!hbt]
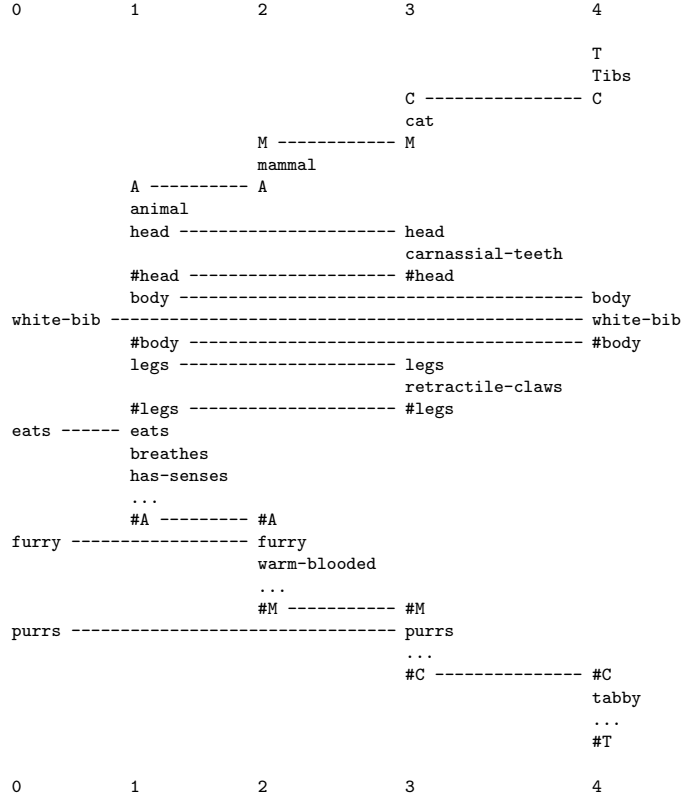

\fontsize{07.00pt}{08.40pt}
\centering
\begin{BVerbatim}
0           1            2              3                  4        

                                                           T        
                                                           Tibs     
                                        C ---------------- C        
                                        cat                         
                         M ------------ M                           
                         mammal                                     
            A ---------- A                                          
            animal                                                  
            head ---------------------- head                        
                                        carnassial-teeth            
            #head --------------------- #head                       
            body ----------------------------------------- body     
white-bib ------------------------------------------------ white-bib
            #body ---------------------------------------- #body    
            legs ---------------------- legs                        
                                        retractile-claws            
            #legs --------------------- #legs                       
eats ------ eats                                                    
            breathes                                                
            has-senses                                              
            ...                                                     
            #A --------- #A                                         
furry ------------------ furry                                      
                         warm-blooded                               
                         ...                                        
                         #M ----------- #M                          
purrs --------------------------------- purrs                       
                                        ...                         
                                        #C --------------- #C       
                                                           tabby    
                                                           ...      
                                                           #T       

0           1            2              3                  4        
\end{BVerbatim}
\caption{The best multiple alignment found by SP61, between the pattern `white-bib eats furry purrs' in New and the patterns shown in Figure \ref{animal_classes_figure} in Old.}
\label{class_part_whole_alignment}
\end{figure}

Figure \ref{class_part_whole_alignment} shows the best multiple alignment found by SP61 between the pattern `white-bib eats furry purrs' in New with the patterns from Figure \ref{animal_classes_figure} in Old.

A multiple alignment like this may be interpreted as the result of a process of recognition. An unknown entity with the attributes `white-bib eats furry purrs' has been recognised as `Tibs' (column 4). At the same time, it is recognised as a cat (column 3), as a mammal (column 2), and as an animal (column 1).

\subsubsection{Inheritance of attributes}

From a multiple alignment like the one shown in Figure \ref{class_part_whole_alignment}, it is possible to infer that the entity that has been recognised has a tabby colouration (from the pattern for `Tibs') and retractile claws (from the pattern for `cat'), that it is warm-blooded (from the pattern for `mammal') and that it breathes and has senses (from the pattern for `animal'). In short, the process of recognition achieves the effect of {\em inheritance of attributes} in the manner of object-oriented design.

This kind of inheritance of attributes in a class hierarchy is a simple but important form of inference or reasoning---part of the recognition-retrieval-and-reasoning constellation mentioned at the beginning of this chapter. In Chapter \ref{pr_chapter}, we shall see how several other kinds of probabilistic reasoning may be modelled in the SP framework. And Chapter \ref{maths_logic_chapter} discusses how similar principles may provide an interpretation for the `exact' forms of reasoning associated with mathematics and logic. 

\subsubsection{Integration of class-inclusion relations and part-whole relations}\label{integration_class_part_whole}

The multiple alignment in Figure \ref{class_part_whole_alignment} not only shows a hierarchy of classes with `animal' at the highest level and `Tibs' at the bottom but it also shows the division of the recognised entity into parts and sub-parts. At the level of `animal', it has a `head', a `body' and some `legs'. Additional information about the structure of those parts is provided at lower levels in the class hierarchy. At the level of `cat', the head contains `carnassial-teeth' and associated with the legs are the `retractile-claws'. At the level of `Tibs' the body has a white bib. It should be clear that the breakdown of an entity into its parts and sub-parts may be elaborated in much more detail than is shown in this simple example.

Notice that there is no artificial distinction between `class' and `object' or between the `attribute' of a class and a `part' of an object, as described in Section \ref{making_haste_slowly}. The multiple alignment scheme allows the {\em integration} of class-inclusion relations with part-whole relations as envisaged at the outset of this project (see also Section \ref{objects_classes_metaclasses}, below).

\subsection{Cross classification and multiple inheritance}\label{cross_classification_multiple_inheritance}

\index{class!cross-classification|(}

`Natural' objects and classes that we recognise in everyday speaking and thinking often 
have more than one parent class: one person may be both a `woman' and a `doctor', another person 
may be both a `man' and a `footballer', and so on. In this kind of `cross classification' or `multiple inheritance', the parent classes overlap each other without one being a sub-class of the other.

\begin{figure}[!hbt]
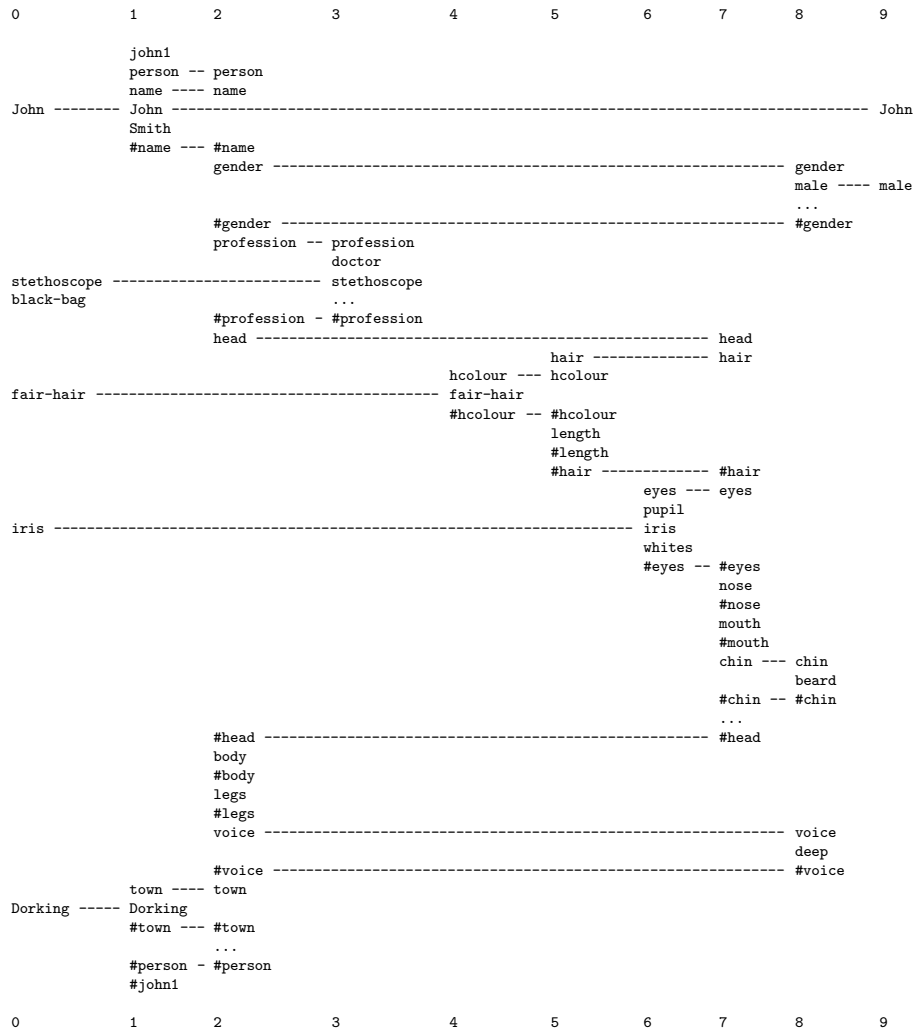

\fontsize{06.00pt}{07.20pt}
\centering
\begin{BVerbatim}
0             1         2             3             4           5          6        7        8         9   

              john1                                                                                        
              person -- person                                                                             
              name ---- name                                                                               
John -------- John ----------------------------------------------------------------------------------- John
              Smith                                                                                        
              #name --- #name                                                                              
                        gender ------------------------------------------------------------- gender        
                                                                                             male ---- male
                                                                                             ...           
                        #gender ------------------------------------------------------------ #gender       
                        profession -- profession                                                           
                                      doctor                                                               
stethoscope ------------------------- stethoscope                                                          
black-bag                             ...                                                                  
                        #profession - #profession                                                          
                        head ------------------------------------------------------ head                   
                                                                hair -------------- hair                   
                                                    hcolour --- hcolour                                    
fair-hair ----------------------------------------- fair-hair                                              
                                                    #hcolour -- #hcolour                                   
                                                                length                                     
                                                                #length                                    
                                                                #hair ------------- #hair                  
                                                                           eyes --- eyes                   
                                                                           pupil                           
iris --------------------------------------------------------------------- iris                            
                                                                           whites                          
                                                                           #eyes -- #eyes                  
                                                                                    nose                   
                                                                                    #nose                  
                                                                                    mouth                  
                                                                                    #mouth                 
                                                                                    chin --- chin          
                                                                                             beard         
                                                                                    #chin -- #chin         
                                                                                    ...                    
                        #head ----------------------------------------------------- #head                  
                        body                                                                               
                        #body                                                                              
                        legs                                                                               
                        #legs                                                                              
                        voice -------------------------------------------------------------- voice         
                                                                                             deep          
                        #voice ------------------------------------------------------------- #voice        
              town ---- town                                                                               
Dorking ----- Dorking                                                                                      
              #town --- #town                                                                              
                        ...                                                                                
              #person - #person                                                                            
              #john1                                                                                       

0             1         2             3             4           5          6        7        8         9   
\end{BVerbatim}
\caption{\small The best multiple alignment found by SP61 with patterns representing classes related to `person' in Old and the pattern `John stethoscope black-bag fair-hair iris Dorking' in New.}
\label{cross_classification_alignment}
\end{figure}

The SP system can accommodate this kind of relationship between classes quite simply. If an entity or class belongs in two or more parent classes, it is merely necessary for the corresponding pattern to include references to those parent classes. This can be seen in Figure \ref{cross_classification_alignment} where someone called `John' is recognised as being a member of the class `doctor' and also a member of the class `male'. 

As with our previous example, various attributes can be inferred from the multiple alignment that were not directly observed as `New'. We can infer that John has the qualifications and other attributes of his profession (represented only by `...' in this multiple alignment) and that he has male attributes such as a beard and a deep voice (no attempt has been made here to model the possibility that John might be a child).

Notice how the attribute `beard' has been tied in to the `chin' part of `head' (column 2), while the attribute `deep' is linked to `voice' in the pattern for `person' (column 4). Notice also how John's male gender (column 5) has, in effect, been inferred from the association between the name `John' and the attribute `male' (column 6).%
\index{class!cross-classification|)}

\subsection{Polythetic or `family resemblance' classes}\label{polythetic_categories}

\index{class!polythetic|(}

It is widely recognised that many of the `natural' categories that we use in everyday thinking are `polythetic' meaning that, for each such category, no single attribute is necessarily present in every member of the given category and there need be no single attribute that is exclusive to that category \citep{sokal_sneath_1973}. In Wittgenstein's phrase, members of this type of category have a `family resemblance' to each other.

The SP system can accommodate this type of category in two different ways that may operate individually or together:

\begin{itemize}

\item As we saw in Section \ref{fuzzy_pattern_recognition}, the dynamic programming that is built into the system (Appendix \ref{matching_appendix}) enables it to recognise patterns despite errors of omission, commission or substitution. This means that recognition does not depend on the presence (or absence) of any one feature. Because the system looks for the best overall match between patterns, recognition does not require that any one category should have any feature that is exclusive to that category.

\item As we saw in Section \ref{framework_examples_section}, SP patterns can function in a manner that is equivalent to a context-free phrase-structure grammar. Any system with that capability (or more) can be used to define categories in which no single attribute need be found in all members of the category and no attribute need be exclusive to that category. 

\end{itemize}

To illustrate this last point, consider the following simple grammar:

\begin{center}
\begin{tabular}{l}
S $\rightarrow$ 1 2 \\
1 $\rightarrow$ A \\
1 $\rightarrow$ B \\
2 $\rightarrow$ C \\
2 $\rightarrow$ D \\
\end{tabular}
\end{center}

This defines the set of two-letter strings \{AC, AD, BC, BD\}. Notice that no single attribute (letter) is found in all the members of this category. And none of the letters need be exclusive to this category. Since the SP system can model this type of grammar, it can model polythetic categories.

In this connection, it is pertinent to say that, while the system can accommodate categories of that kind, it is not constrained to operate only in that way. It is easy enough within the SP system to create categories in which one or more features (or combination of features) are necessary for the recognition of a category or sufficient for that recognition or both these things.

\subsubsection{Objects, classes and metaclasses}\label{objects_classes_metaclasses}

\index{metaclass|(}

By contrast with most object-oriented systems, the SP system makes no formal distinction between `class' and `object'. This accords with the observation that what we perceive to be an individual object, such as `Tibs', can itself be seen to represent a variety of possibilities: `Tibs playing with a ball', `Tibs catching a mouse', `Tibs when he is sleeping', and so on. A pattern like the one shown in column 4 of Figure \ref{class_part_whole_alignment} could easily function as a class with vacant slots to be filled at a more specific level by details of the activity that Tibs is engaged in, whether he is well-fed or hungry, happy or sad, and so on. This flexibility is lost in systems that do make a formal distinction between classes and objects.

Making a formal distinction between objects and classes points to the need for the concept of a `metaclass':

\begin{quotation}

``If each object is an instance of a class, and a class is an object, the [object-oriented] model should provide the notion of {\em metaclass}. A metaclass is the class of a class.'' \cite[p. 43]{bertino_etal_2001}.

\end{quotation}

\noindent This construct has been introduced in some AI systems so that classes can be derived from metaclasses in the same way that objects are derived from classes. Of course, this logic points to the need for `metametaclasses', `metametametaclasses', and so on without limit. Because the SP system makes no distinction between `object' and `class', there is no need for the concept of `metaclass' or anything beyond it. All these constructs are represented by patterns.%
\index{metaclass|)}\index{class!hierarchy|)}\index{part-whole hierarchy|)}%
\index{inheritance of attributes|)}\index{class!polythetic|)}\index{object|)}

\section{Medical diagnosis}\label{medical_diagnosis_section}

\index{diagnosis!medical|(}

The problem of providing computational support for medical diagnosis has been approached from many directions including logical reasoning, set theory, rough set theory, if-then rules, Bayesian networks, artificial neural networks, case-based reasoning, possibility theory, and others. This section describes how medical diagnosis, viewed as a problem of pattern recognition, may be modelled in the SP system.

In the SP scheme, knowledge about diseases may be stored as patterns in Old and the signs and symptoms for an individual patient (abbreviated hereinafter as `symptoms') may be represented as a New pattern. A pattern in Old may represent one disease and its associated symptoms or it may represent a cluster of symptoms that tend to occur together in two or more different diseases (see Section \ref{clusters_of_symptoms}, below). In addition, Old may include patterns that play supporting r{\^o}les (see Section \ref{uncertainties_in_diagnosis}, below).

Associated with each pattern in Old is an integer value reflecting the frequency with which a given disease or cluster of symptoms is found in a given population. These figures may be derived from population surveys or they may be estimated by medical experts.

The process of diagnosis may be modelled by the building of one or more multiple alignments. Figure \ref{medical_diagnosis_alignment} shows the best multiple alignment created by SP61 with a New pattern describing a fictional `John Smith' and his symptoms and a set of patterns in Old, provided by the user, that represent diseases or aspects of diseases.

\begin{figure}[!hbt]
\fontsize{05.00pt}{06.00pt}
\centering
\begin{BVerbatim}
0                    1                     2                  3              4                 5       

                                           disease                                                     
                     D ------------------- D                                                           
                     flu                                                                               
                     : ------------------- :                                                           
patient ---------------------------------- patient                                                     
john_smith                                                                                             
#patient --------------------------------- #patient                                                    
                     diagnosis ----------- diagnosis                                                   
poor_appetite        influenza                                                                         
                     #diagnosis ---------- #diagnosis                                                  
                     R1 ------------------ R1                                                          
                     flu_symptoms --------------------------- flu_symptoms                             
                     #R1 ----------------- #R1                                                         
                                           R2 --------------- R2                                       
                                                              fever -------- fever                     
                                           #R2 -------------- #R2                                      
                     ap ------------------ ap                                                          
                     normal_appetite                                                                   
                     #ap ----------------- #ap                                                         
                                           br ------------------------------ br                        
rapid_breathing ------------------------------------------------------------ rapid_breathing           
                                           #br ----------------------------- #br                       
                     chst ---------------- chst                                                        
                     normal_chest                                                                      
                     #chst --------------- #chst                                                       
                                           chl -------------- chl                                      
has_chills -------------------------------------------------- has_chills                               
                                           #chl ------------- #chl                                     
                                           cgh -------------- cgh                                      
                                                              has_cough                                
                                           #cgh ------------- #cgh                                     
                     dh ------------------ dh                                                          
                     no_diarrhoea                                                                      
                     #dh ----------------- #dh                                                         
                                           fc ------------------------------ fc                        
flushed_face --------------------------------------------------------------- flushed_face              
                                           #fc ----------------------------- #fc                       
                     ftg ----------------- ftg                                                         
tired                no_fatigue                                                                        
                     #ftg ---------------- #ftg                                                        
                                           hd --------------- hd                                       
                                                              has_headache                             
                                           #hd -------------- #hd                                      
                     ln ------------------ ln                                                          
normal_lymph_nodes - normal_lymph_nodes                                                                
                     #ln ----------------- #ln                                                         
                     mls ----------------- mls                                                         
no_malaise --------- no_malaise                                                                        
                     #mls ---------------- #mls                                                        
                                           msl -------------- msl                                      
muscle_aches ------------------------------------------------ muscle_aches                             
                                           #msl ------------- #msl                                     
                                           ns --------------- ns                                       
runny_nose -------------------------------------------------- runny_nose                               
                                           #ns -------------- #ns                                      
                     skn ----------------- skn                                                         
                     normal_skin                                                                       
                     #skn ---------------- #skn                                                        
                                           tmp ----------------------------- tmp                       
                                                                             tclass1 --------- tclass1 
t37-39 --------------------------------------------------------------------------------------- t37-39  
                                                                             #tclass1 -------- #tclass1
                                           #tmp ---------------------------- #tmp                      
                                           thr -------------- thr                                      
sore_throat ------------------------------------------------- sore_throat                              
                                           #thr ------------- #thr                                     
                     wt ------------------ wt                                                          
                     no_weight_change                                                                  
                     #wt ----------------- #wt                                                         
                     causative_agent ----- causative_agent                                             
                     influenza_virus                                                                   
                     #causative_agent ---- #causative_agent                                            
                     treatment ----------- treatment                                                   
                     influenza_treatment                                                               
                     #treatment ---------- #treatment                                                  
                     #D ------------------ #D                                                          
                                           #disease                                                    

0                    1                     2                  3              4                 5       
\end{BVerbatim}
\caption{The best multiple alignment found by SP61 with a pattern describing a patient's symptoms in New and a set of patterns in Old describing a range of different diseases and named clusters of symptoms, together with the `framework' pattern shown in column 2.}
\label{medical_diagnosis_alignment}
\end{figure}

A multiple alignment like this may be interpreted as the result of a process of recognition. In this case, the symptoms that have been recognised are those of influenza, as shown in column 1. The following subsections discuss aspects of the multiple alignment and of this interpretation.  

\subsection{Ordering of symptoms}\label{ordering_of_symptoms}

As noted in Section \ref{ordering_of_symbols_and_patterns}, there are some applications where one might wish to treat New information as an unordered set of patterns or as an unordered or partially-ordered set of symbols. Medical diagnosis seems to be one such application because it is not obvious that there is any intrinsic order to the symptoms of a disease: `high temperature' may precede `runny nose' but, equally well, it might come after.

Given that a patient's symptoms are specified in a single pattern in New, and given that that pattern represents a sequence of symbols, one can achieve the effect of unordered symptoms by using a `framework' pattern (mentioned in Section \ref{ordering_of_symbols_and_patterns}) that defines an arbitrary order for disjunctive classes of symptom that may be found across the range of diseases described by the patterns in Old. An example of such a framework pattern is shown in Figure \ref{medical_diagnosis_alignment}, column 2. Within that pattern, a pair of symbols such as `ap \#ap' provides a slot for alternative states of a patient's appetite. Likewise, the pair of symbols `br \#br' provides a slot for different kinds of breathing---`rapid', `laboured', `normal' and so on. Provided that corresponding pairs of symbols are used in other patterns in Old, and provided that classes of symptoms are always specified in the defined order, appropriate multiple alignments can be built.

\subsection{Clusters of symptoms}\label{clusters_of_symptoms}

It is quite usual in medicine for a cluster of two or more symptoms that occur in two or more different diseases to be given a name. For example, the name `fever' provides a conveniently brief shorthand for the symptoms `high temperature', `flushed face' and `rapid breathing' that feature in several different diseases.

Two such clusters are shown in Figure \ref{medical_diagnosis_alignment}. The symptoms of fever are shown in column 4 while column 3 shows a cluster of symptoms associated with influenza and other diseases such as smallpox (see below). Notice that fever is itself part of the influenza cluster.

Notice also how the provision of a named cluster saves the need to specify the corresponding symptoms redundantly in each of the diseases where that cluster appears.

\subsection{Uncertainties in diagnosis}\label{uncertainties_in_diagnosis}

Diagnosis is not an exact process. Although some symptoms may be necessary or sufficient for the diagnosis of a given disease, the majority are merely `characteristic' in the sense that they are associated with the disease but any one such symptom need not be present in every case and any of them may be associated with other diseases. There may also be errors in the observation or recording of symptoms.

In broad terms, SP61 can accommodate this kind of uncertainty in diagnosis. This is because it looks for a global best match amongst patterns and does not depend on the presence or absence of any particular symptom. At the same time, particular symptoms can have a major impact on diagnosis, as described in Section \ref{explaining_away}, below.

Within the SP system, it is not necessary for every symptom of a disease to be recorded as a specific value. In column 4 of the multiple alignment in Figure \ref{medical_diagnosis_alignment}, the pair of symbols `tclass1 \#tclass1' represents a set of alternative values for the temperature associated with fever. In this case, there are just two values, represented in Old by the patterns `tclass1 t37-39 \#tclass1' (high temperature) and `tclass1 t39+ \#tclass1' (very high temperature). The first of these patterns is shown in column 5 of the multiple alignment, matched to the temperature of the patient shown in column 0.

\subsection{Weighing alternative hypotheses}\label{medical_alternative_hypotheses}

In medical diagnosis, it is quite usual for the physician to consider alternative hypotheses about what disease or diseases the patient may be suffering from. The SP system provides a model for this process in the way the system builds alternative multiple alignments for any given pattern in New, each with a compression score calculated as described in Section \ref{ma_evaluation}. Notice that the compression score  depends in part on the frequency values for patterns mentioned at the beginning of Section \ref{medical_diagnosis_section}.

As described in Section \ref{probabilities_section}, each multiple alignment has an absolute probability. And for each multiple alignment in a set of alternative multiple alignments that encode the same symbols from New, a relative probability ($p_{REL}$) may also be calculated.

Where alternative multiple alignments encode different subsets of the symbols in New, $CD$ seems to be the most appropriate measure for the purpose of comparison. Where two or more multiple alignments encode exactly the same symbols from New, they may be compared using relative probabilities.

When SP61 formed the multiple alignment shown in Figure \ref{medical_diagnosis_alignment}, it also formed a similar multiple alignment, matching exactly the same symbols in New, in which column 2 contained a pattern representing the symptoms of smallpox, instead of the pattern for influenza. The $p_{REL}$ values calculated in this case were 0.99950 for influenza and 0.00049 for smallpox, reflecting the relative frequencies of those two diseases in the world today.

\subsubsection{`Explaining away'}\label{medical_explaining_away}

\index{reasoning!explaining away|(}

The symptoms of influenza and smallpox are quite similar, except for the very distinctive rash and blisters that occur in smallpox. The example shown in Figure \ref{medical_diagnosis_alignment} is silent about whether John Smith had a rash and blisters or not.\footnote{If a rash and blisters had been seen to be absent, this would be represented as `normal\_skin'.} Given this lack of information about those symptoms, the probabilities shown above are reasonable.

If `rash\_with\_blisters' is added to the symptoms recorded in New, and SP61 is run again, the best multiple alignment found by the system is similar to that shown in Figure \ref{medical_diagnosis_alignment} but with the pattern for smallpox in column 2 and with a match shown for the two instances of `rash\_with\_blisters'. However, in this case there is no other multiple alignment that matches the same symbols in New. So the value of $p_{REL}$ for the best multiple alignment is 1.0.

From this result, we may conclude that the patient certainly has smallpox and that his aching muscles and runny nose are due to smallpox, not influenza. This is the phenomenon of `explaining away' discussed in Section \ref{bayesian_network}.%
\index{reasoning!explaining away|)}

\subsubsection{A patient may suffer from two or more diseases at the same time}

It is entirely possible, of course, for a patient to suffer from two or more diseases at the same time. In the SP system, a combination of diseases may be treated as a single disease, represented as a pattern in Old. The pattern may be constructed economically as references to the component diseases, as described in Section \ref{clusters_of_symptoms}.

As with single diseases, frequency values for a combinations of diseases may be obtained from population surveys or by the judgement of experts. Where diseases are statistically independent, these values will of course be very low. But they can be used for the calculation of $CD$ values and probabilities in exactly the same way as for single diseases.

\subsection{Inference and diagnosis}

Hypotheses formed in the course of diagnosis often lead to inferences about the condition of the patient. These inferences may suggest the need to gather more information from blood tests, X rays or the like. Or they may provide the basis for prescription, treatment or advice to the patient.

In a multiple alignment like the one shown in Figure \ref{medical_diagnosis_alignment}, any symbol within an Old pattern that is {\em not} matched to a symbol in New represents an inference that may be drawn from the multiple alignment. In this example, we may infer from the multiple alignment {\em inter alia} that the patient is likely to have a cough and a headache and that the standard treatment for influenza is required. Probabilities of these inferences can be calculated as described in Section \ref{probabilities_section}.

\subsubsection{Causal reasoning}\label{medical_causal_reasoning}

Apart from the kinds of inference just described, medical diagnosis often seems to involve a `deeper' kind of reasoning about the causes of symptoms and diseases, using knowledge of anatomy and physiology.

No attempt has yet been made with the SP system to model this kind of medical reasoning. However, as we shall see in Section \ref{causal_diagnosis_section}, the SP system can be used to model causal diagnosis in other contexts and there seems no reason in principle why it should not also be applied to reasoning about the causes of medical conditions.%
\index{diagnosis!medical|)}

\section{Conclusion}

As we have seen, the SP system allows the recognition or retrieval of patterns despite errors of omission, commission or substitution. It also allows recognition or retrieval with class-inclusion relations (and cross-classification) and their integration with part-whole relations. The system can model polythetic categories and it supports indirection in information retrieval. An example has been presented showing how the system may provide computational support for medical diagnosis, viewed as a problem of pattern recognition.

We have noted the close recognition-retrieval-and-reasoning connection between recognition, retrieval and reasoning and we have touched briefly on the last of these three in Sections \ref{medical_alternative_hypotheses}, \ref{medical_explaining_away} and \ref{medical_causal_reasoning}. In the next chapter, we look in more detail at the ways in which the SP system supports probabilistic reasoning.%
\index{perception!recognition|)}\index{information!retrieval|)}

%% file: pr.tex
\chapter[Probabilistic Reasoning]{Probabilistic Reasoning%
\protect\footnote{Based in part on \citet{wolff_1999_prob}.}}\label{pr_chapter}

\index{reasoning!probabilistic|(}

\section{Introduction}

Quoting Benjamin Franklin (``Nothing is certain but death and taxes''), Ginsberg \citeyearpar{ginsberg_1994} writes (p. 2) that: ``The view that Franklin was expressing is that virtually every conclusion we draw [in reasoning] is an uncertain one.'' He goes on to say: ``This sort of reasoning in the face of uncertainty... has ... proved to be remarkably difficult to formalise.''

This chapter shows how various kinds of probabilistic reasoning may be performed within the SP system.

\subsection{`Reasoning' and `inference'}\label{reasoning_and_inference_section}

Before we proceed, some clarification is necessary for the meanings of the terms {\em reasoning} 
and {\em inference} as they will be used in this book.

`Reasoning' will be used to mean any process of ``going beyond the information 
given''. In most forms of `deductive' reasoning, conclusions are either `true' or `false', whereas in `probabilistic reasoning', conclusions have some kind of `probability' or level of confidence attaching to them, e.g., ``Black clouds mean that it will probably rain''.

All-or-nothing deductive reasoning as it appears in logic will not be considered in this 
chapter, except in passing. Chapter \ref{maths_logic_chapter} considers how logic and mathematics may be understood in terms of information compression and, more specifically, the SP theory.

As it is normally used, the term `inference' has three distinct but related meanings:

\begin{itemize}

\item It is sometimes used to mean a process of constructing a grammar or other kind of 
knowledge structure by `induction' from a body of `raw' data.

\item It is also often used with essentially the same meaning as probabilistic reasoning as 
described above.

\item It may also be used to refer to the result or product of a process of probabilistic reasoning.

\end{itemize}

Inference of the first kind is the main subject of Chapter \ref{learning_chapter}. Only the second and third of these notions of inference will be considered in this chapter. 

\subsection{Related research and novelty of the present proposals}

There is now a huge literature on probabilistic reasoning and related ideas ranging over `standard' parametric and non-parametric statistics, {\em ad hoc} uncertainty measures in early expert systems, Bayesian statistics, Bayesian/belief/causal networks, Markov networks, Self-Organising Feature Maps, fuzzy set theory and `soft' computing', the Dempster-Shaffer theory, abductive reasoning, nonmonotonic reasoning and reasoning with default values, autoepistemic logic, defeasible logic, possibilistic and other kinds of logic designed to accommodate uncertainty, MLE, algorithmic probability and algorithmic complexity theory, truth maintenance systems, decision analysis, utility theory, and more.

A well-known and authoritative survey of the field, with an emphasis on Bayesian networks, is provided by \citet{pearl_1988} although this book is now, perhaps, in need of some updating \citep[but see][]{pearl_2000}. A relatively short but useful review of ``uncertainty handling formalisms'' appears in \citet{parsons_hunter_1998}. Regarding the application of different kinds of `logic' to nonmonotonic and uncertain reasoning, there is a mine of useful information in the articles in \cite{gab_hog_rob_1994} covering such things as `default logic', `autoepistemic logic', `circumscription', `defeasible logic', `uncertainty logics' and `possibilistic logic'. In that volume, the chapter by \citet{ginsberg_1994} provides an excellent introduction to the problems of nonmonotonic reasoning. Papers by \cite{bondarenko_etal_1997, kern-isberner_1998, kohlas_etal_1998} are also relevant as are the papers in \citet{gammerman_1996}.

Amongst these many approaches to probabilistic reasoning, the multiple alignment concept---as it has been developed in the SP system---is distinctive. Also distinctive is the attempt to integrate probabilistic reasoning with a wide range of concepts in artificial intelligence, computing, logic and mathematics. 

\subsection{Information compression and probabilistic reasoning}

Naturally enough, much of the literature on probabilistic reasoning deals directly with 
concepts of probability, especially conditional probability. Since, however, there is a close 
connection between probability and compression (as described in Section \ref{probabilities_ic_section}), concepts of probability imply corresponding concepts of compression.

That said, a primary emphasis on compression rather than probability provides an alternative perspective on the subject which may prove useful. Relevant sources include \citet{dagu_luby_1997, grunwald_1997, van_der_gaag_1996, watanabe_article_1972}.

\section{Probabilistic reasoning, multiple alignment and information compression}\label{prob-reason}

What connection is there between the formation of a multiple alignment and probabilistic reasoning? This section describes the connection and the way in which probabilities of inferences may be derived from multiple alignments.

In the simplest terms, probabilistic reasoning arises from partial pattern recognition: if a pattern is recognised from a subset of its parts (something that people and animals are very good at doing), then, in effect, an inference is made that the unseen part or parts are `really' there. We might, for example, recognise a car from seeing only the front half because the rear half is hidden behind another car or a building. The inference that the rear half is present is probabilistic because there is always a possibility that the rear half is absent or, in some surreal world, replaced by the front half of a horse, or something equally bizarre.

In terms of multiple alignment, probabilistic reasoning may be understood as the formation of a multiple alignment in which one or more symbols in the Old patterns are not aligned with any matching symbol or symbols in the New pattern. The probabilities of these symbols (and the inferences that they represent) are calculated as described in Section \ref{probabilities_section}. As a working hypothesis, all kinds of probabilistic reasoning may be understood in these terms.

\subsection{Confidence in inferences}\label{confidence_in_inferences}

The formulae and calculations presented in Section \ref{probabilities_section} may suggest that, given a system which is equipped with this mathematical machinery, we can calculate precisely the level of confidence that should be placed in inferences that the system makes.

This is, of course, nonsense. Any knowledge-based system (including human brains!) is 
subject to the law ``rubbish in means rubbish out''. In the case of human experts, we do or should ask whether they were given all the relevant information about a given case, how well-trained they are, how much experience they have, whether they have given full attention to the case, and so on. Artificial systems are no different.

In general, the confidence which we may place in probabilistic inferences made by a given 
system should be influenced by several factors including the accuracy and coverage of the 
information supplied about a particular case, the accuracy and coverage of the knowledge stored 
in the system, the effectiveness of the search methods used by the system, and the thoroughness of the search which has been made for a given set of data and inferences.

\subsection{A simple example}\label{simple_pr_example}

In order to illustrate the kinds of values that may be calculated for absolute and relative probabilities, this subsection presents a very simple example: the inference of `fire' from `smoke'. Here, we shall extend the concept of `smoke' to include anything, like mist or fog, which looks like smoke. Also, `fire' has been divided into three categories: the kind of fire used to heat a house or other building, dangerous fires that need a fire extinguisher or more, and the kind of fire inside a burning cigarette or pipe. 

Given a New pattern containing the single symbol `smoke' and the Old patterns shown in Figure \ref{associations_figure}, SP61 forms the five obvious multiple alignments of New with each of the patterns which contain the symbol `smoke'.

\begin{figure}[!hbt]
\centering
\begin{tabular}{l}
clouds black rain (15,000) \\
dangerous fire smoke (500) \\
heating fire smoke (7,000) \\
tobacco fire smoke (10,000) \\
fog smoke (2,000) \\
stage smoke (100) \\
thunder lightning (5,000) \\
strawberries cream (1,500) \\
\end{tabular}
\caption{A small knowledge base of associations. The numbers in brackets show an imaginary frequency of occurrence of each pattern in some notional reference environment.}
\label{associations_figure}
\end{figure}

The absolute and relative probabilities of the five multiple alignments, calculated as described in Section \ref{probabilities_section}, are shown in Table \ref{smoke_alignment_probabilities}.

\begin{table}[!hbt]
\centering
\begin{tabular}{l l l}
\em Alignment & \em Absolute & \em Relative \\
& \em probability & \em probability \\
\\
smoke/tobacco fire smoke & 0.08718 & 0.51020 \\
smoke/heating fire smoke & 0.06103 & 0.35714 \\
smoke/fog smoke & 0.01744 & 0.10204 \\
smoke/dangerous fire smoke & 0.00436 & 0.02551 \\
smoke/stage smoke & 0.00009 & 0.00510 \\
\end{tabular}
\caption{Absolute and relative probabilities of each of the five reference multiple alignments formed between `smoke' in New and the patterns shown in Table \ref{associations_figure} in Old. In this example, the relative probability of each pattern from Old is the same as the multiple alignment in which it appears.}
\label{smoke_alignment_probabilities}
\end{table}

In this very simple example, the relative probability of each pattern from Old is the same as for the multiple alignment in which it appears. However, the same cannot be said of individual symbol types. The relative probabilities of the symbol types that appear in any of the five reference multiple alignments are shown in Table \ref{smoke_symbol_probabilities}. The main points to notice about the relative probabilities shown in this table are:

\begin{itemize}

\item The relative probability of `smoke' is 1.0. This is because it is a `fact' which appears in New, so there is no uncertainty attaching to it.

\item Of the other symbol types from Old, the one with the highest probability relative to the other symbols is `fire', and this relative probability is higher than the relative probability of any of the patterns from Old (Table \ref{smoke_alignment_probabilities}). This is because `fire' appears in three of the reference multiple alignments.

\end{itemize}

\begin{table}[!hbt]
\centering
\begin{tabular}{ll}
\em Symbol & \em Relative \\
 & \em probability \\
\\
smoke & 1.00000 \\
fire & 0.89286 \\
tobacco & 0.51020 \\
heating & 0.35714 \\
fog & 0.10204 \\
dangerous & 0.02551 \\
stage & 0.00510 \\
\end{tabular}
\caption{The relative probabilities of the symbol types from Old that appear in any of the reference set of multiple alignments shown in Table \ref{smoke_alignment_probabilities}.}
\label{smoke_symbol_probabilities}
\end{table}

In this example, we have ignored all the subtle cues that people would use in practice to infer the origin of smoke: the smell, colour and volume of smoke, associated noises, behaviour of other people, and so on. Allowing for this, and allowing for the probable inaccuracy of the frequency values which have been used, the relative probabilities of multiple alignments, patterns and symbols seem to reflect the subjective probability which we might assign to the five alternative sources of smoke-like matter in everyday situations.

\section{One-step `deductive' reasoning}\label{one_step_deductive_reasoning}

\index{reasoning!deductive|(}

Consider a `standard' example of {\em modus ponens} syllogistic reasoning:

\begin{enumerate}

\item $\forall x$: bird($x$) $\implies$ canfly($x$).

\item bird(Tweety).

\item $\therefore$ canfly(Tweety).

\end{enumerate}

\noindent which, in English, may be interpreted as:

\begin{enumerate}

\item If something is a bird then it can fly.

\item Tweety is a bird.

\item Therefore, Tweety can fly.

\end{enumerate}

In classical logic, a `material implication' like $(p \implies q)$ (``If something is a bird then it can fly'') is equivalent to $\neg(p \land \neg q)$ (``It is not true that something is a bird and it cannot fly'') and to $(\neg q \implies \neg p)$ (``If something cannot fly then it is not a bird'') and also to $(\neg p \lor q)$ (``Either something is not a bird or it can fly'').

However, there is a more relaxed, `everyday' kind of `deduction' which, in terms of our example, may be expressed as: ``If something is a bird then, {\em probably}, it can fly. Tweety is a bird. Therefore, {\em probably}, Tweety can fly.''

This kind of probabilistic `deduction' differs from material implication because it does not have the same equivalencies as the classical form. If our focus of interest is in describing and reasoning about the real world rather than exploring the properties of abstract systems of symbols, the probabilistic kind of `deduction' seems to be more appropriate. With regard to birds, we know that there are flightless birds, and for most other examples of a similar kind, an ``all or nothing'' logical description would not be an accurate reflection of the facts.

With a pattern of symbols, we may record the fact that birds can fly and, in a very natural way, we may record all the other attributes of a bird in the same pattern. The pattern may look something like this:

\begin{center}
\begin{BVerbatim}
Bd bird name #name canfly warm-blooded wings feathers ... #Bd,
\end{BVerbatim}
\end{center}

\noindent or the attributes of a bird may be described in the more elaborate way described in Section \ref{class_part_inheritance}.

\begin{figure}[!hbt]
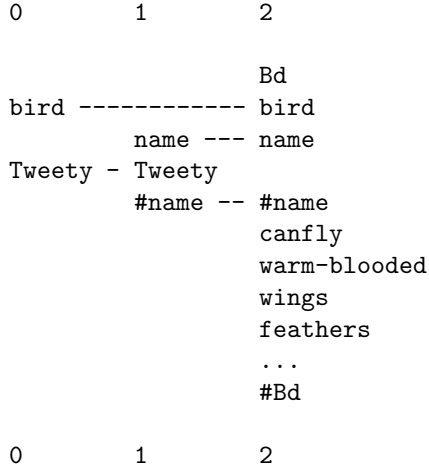

\centering
\begin{BVerbatim}
0        1        2           

                  Bd          
bird ------------ bird        
         name --- name        
Tweety - Tweety               
         #name -- #name       
                  canfly      
                  warm-blooded
                  wings       
                  feathers    
                  ...         
                  #Bd         

0        1        2           
\end{BVerbatim}
\caption{The best multiple alignment found by SP61 with the pattern `bird Tweety' in New and other patterns in Old as described in the text.}
\label{bird_tweety_alignment}
\end{figure}

This pattern and others of a similar kind may be stored in `Old', together with patterns like `name Tweety \#name', `name George \#name', `name Susan \#name' and so on which define the range of possible names. Also, the pattern, `bird Tweety', corresponding to the proposition ``Tweety is a bird'' may be supplied as New. Given patterns like these in New and Old, the best multiple alignment found by SP61 is the one shown in Figure \ref{bird_tweety_alignment}.

As before, the inferences which are expressed by this multiple alignment are represented by the unmatched symbols in the multiple alignment. The fact that Tweety is a bird allows us to infer that Tweety can fly but it also allows us to infer that Tweety is warm-blooded, has wings and feathers and all the other attributes of birds. These inferences arise directly from the pattern describing the attributes of birds.

In this case, there is only one multiple alignment which encodes all the symbols in New. Therefore, the relative probability of the multiple alignment is 1.0, the relative probability of `canfly' is 1.0, and likewise for all the other symbols in the multiple alignment, both those which are matched to New and those which are not.

At this point readers may wonder whether the SP scheme can handle nonmonotonic reasoning: the fact that additional information about penguins, kiwis and other flightless birds would invalidate the inference that something being a bird means that it can fly. The way in which the SP system can perform nonmonotonic reasoning is described in Section \ref{nonmonotonic_reasoning_section}, below.%
\index{reasoning!deductive|)}

\section{Abductive reasoning}\label{abductive_reasoning_section}

\index{reasoning!abductive|(}

In the SP system, any subsequence of a pattern may function as what is `given' in reasoning, with the complementary subsequence functioning as the inference. Thus, it is just as easy to reason in a `backwards', abductive manner as it is to reason in a `forwards', deductive manner. We can also reason from the middle of a pattern outwards, from the ends of a pattern to the middle, and many other possibilities. In short, the SP system allows seamless integration of probabilistic `deductive' reasoning with abductive reasoning and other kinds of reasoning which are not commonly recognised.

\begin{figure}[!hbt]
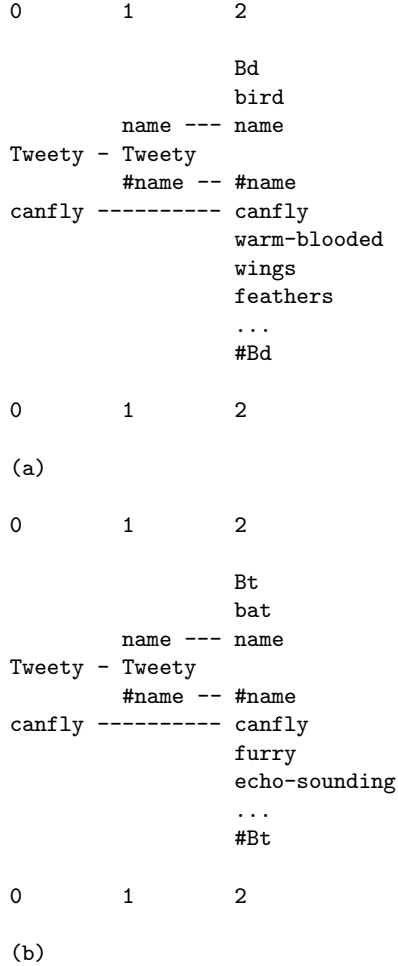

\fontsize{09.00pt}{10.80pt}
\centering
\begin{BVerbatim}
0        1        2           

                  Bd          
                  bird        
         name --- name        
Tweety - Tweety               
         #name -- #name       
canfly ---------- canfly      
                  warm-blooded
                  wings       
                  feathers    
                  ...         
                  #Bd         

0        1        2           

(a)

0        1        2            

                  Bt           
                  bat          
         name --- name         
Tweety - Tweety                
         #name -- #name        
canfly ---------- canfly       
                  furry        
                  echo-sounding
                  ...          
                  #Bt          

0        1        2            

(b)
\end{BVerbatim}
\caption{The best multiple alignment (a), and the other member of its reference set of multiple alignments (b), formed by SP61 with patterns in Old as described in Section \ref{one_step_deductive_reasoning} and with `Tweety canfly' in New.}
\label{tweety_canfly_alignment}
\end{figure}

Figure \ref{tweety_canfly_alignment} shows the best multiple alignment and the other member of its reference set of multiple alignments which are formed by SP61 with the same patterns in Old as were used in the example of `deductive' reasoning (Section \ref{one_step_deductive_reasoning}) and with the pattern `Tweety canfly' in New.

By contrast with the example of `deductive' reasoning, there are two multiple alignments in the reference set of multiple alignments that encode all the symbols in New. These two multiple alignments represent two alternative sets of abductive inferences that may be drawn from this combination of New and Old.

With regard to the first multiple alignment (Figure \ref{tweety_canfly_alignment} (a)), `Tweety' could be a bird with all the attributes of birds, including the ability to fly. The relative probability of the multiple alignment is 0.74, as is the relative probability of the pattern for `bird' and every other symbol in that pattern (apart from the `name' and `\#name' symbols where the relative probability is 1.0).

Alternatively, we may infer from the second multiple alignment (Figure \ref{tweety_canfly_alignment} (b)) that `Tweety' could be a bat. But in this case the relative probability of the multiple alignment, the pattern for `bat' and all the symbols in that pattern (apart from the `name \#name' symbols) is only 0.26.%
\index{reasoning!abductive|)}

\section{Reasoning with probabilistic decision networks and decision trees}\label{probabilistic_decision_network}

\index{decision network or tree|(}

So far, we have considered examples of reasoning in a single step. One of the simplest kinds of system that supports reasoning in more than one step (as well as single step reasoning) is a `decision network' or a `decision tree'. In such a system, a path is traced through the network or tree from a start node to two or more alternative destination nodes depending on the answers to multiple-choice questions at intermediate nodes. Any such network or tree may be given a probabilistic dimension by attaching a value for probability or frequency to each of the alternative answers to questions at the intermediate nodes.

Figure \ref{decision_tree_rules} shows a set of patterns, each of which represents a non-terminal node of a decision network from a car maintenance manual for diagnosing faults in a car engine. To save space, the text associated with each pattern has been omitted. Terminal nodes have also been omitted because, without the text, each one would contain nothing but a number symbol. Figure \ref{decision_tree_rules_sample} shows a sample of the patterns for non-terminal and terminal nodes of the network with the text for each node included.

In Figure \ref{decision_tree_rules}, each pattern except the pattern for the start node is identified by the number symbol which appears at the beginning of the pattern, together with the `yes' or `no' answer to the question in the parent node. The number at the end of each pattern identifies the two children of the node.

As usual with an SP knowledge base, each pattern in Figure \ref{decision_tree_rules} has a frequency of occurrence, shown to the right of each pattern. In the present case, each frequency is a guestimated frequency of a symptom or symptoms in the domain of car repair.

\begin{figure}[!hbt]
\fontsize{07.00pt}{08.40pt}
\centering
\begin{BVerbatim}
Start 43 (1202)
   43 yes 44 (1043)
      44 yes 19 (1009)
         19 yes 59 (46)
         19 no 1 (963)
             1 yes 2 (691)
                2 yes 4 (92)
                   4 yes 58 (36)
                   4 no 23 (56)
                2 no 5 (599)
                   5 yes 58 (62)
                   5 no 27 (537)
                      27 yes 21 (293)
                         21 yes 61 (84)
                         21 no 22 (209)
                            22 yes 24 (24)
                            22 no 22a (87)
                               22a yes 35 (44)
                                  35 yes 58 (19)
                                  35 no 26 (25)
                               22a no 36 (43)
                                  36 yes 58 (21)
                                  36 no 26 (22)
                      27 no 60 (244)
             1 no 3 (272)
                3 yes 6 (36)
                3 no 7 (236)
                   7 yes 8 (71)
                      8 yes 8a (55)
                      8 no 9 (16)
                   7 no 10 (165)
                      10 yes 11 (115)
                         11 yes 13 (14)
                         11 no 14 (101)
                            14 yes 15 (21)
                            14 no 16 (80)
                      10 no 12 (50)
      44 no 51 (34)
         51 yes 56 (14)
         51 no 59 (20)
   43 no 45 (159)
      45 yes 46 (145)
         46 yes 52 (120)
            52 yes 53 (86)
            52 no 54 (34)
               54 yes 55 (22)
               54 no 50 (12)
         46 no 62 (25)
      45 no 47 (14)
         47 yes 48 (8)
         47 no 49 (6)
\end{BVerbatim}
\caption{A set of patterns representing the non-terminal nodes of a decision tree for the diagnosis of 
faults in a car engine. To save space, the text associated with each pattern has been 
omitted. The full version of some of these patterns can be seen in Figure \ref{decision_tree_rules_sample}.}
\label{decision_tree_rules}
\end{figure}

\begin{figure}[!hbt]
\fontsize{10.00pt}{12.00pt}
\centering
\begin{BVerbatim}
Start 43 Does the starter turn the engine?
   43 yes 44 Does starter turn the engine briskly?
      44 yes 19 Slowly press throttle to floor, return 
            choke and try again. Does engine start?
         19 yes 58 Problem is solved.
         19 no 1 Is there a spark at the plug leads?
            1 yes 2 Remove and examine plugs.
                  Are they black, oily or wet?
               2 yes 4 Thoroughly clean and
                     dry [plugs], or renew, check
                     gap and refit. Try again.
                     Does engine start?
                  4 yes 58 Problem is solved.
                  4 no 23 Check for water in fuel.
                        Drain off to clear.
               2 no 5 Clean, check gap and replace plugs.
                     Does engine start?
(etc)
\end{BVerbatim}
\caption{A sample of the patterns representing nodes of the decision network for diagnosing faults 
in car engines, including the text associated with each pattern.}
\label{decision_tree_rules_sample}
\end{figure}

\subsection{Networks and trees: convergence and divergence of links}

The patterns in the full network of non-terminal and terminal nodes represent a network 
rather than a tree because there is both divergence and convergence of links. Divergence appears 
when a node has two or more children while convergence appears where a node has two or more 
parents (e.g., node 58). Clearly, a decision tree can be represented by using simple patterns in the 
same manner as this example of a decision network, except that there must not be any patterns 
which have two or more parents.

\subsection{Forming multiple alignments: sequential processing of New}

How can this network be used to diagnose a fault in a car engine? In principle, this means 
putting the `Start' symbol into New followed by a sequence of `yes' and `no' symbols, putting 
the patterns from Figure \ref{decision_tree_rules} into Old, and then searching for the multiple alignment which represents the best encoding of New in terms of Old.

This abstract description of how the system may be used is not very practical---because it 
leads to inefficient searching (in terms both of speed and success in finding the `correct' multiple alignment) 
and because it does not allow the questions which are posed by the system to be answered in the 
kind of progressive interactive way which is natural for this kind of system.

For these reasons, SP61 has been designed so that, where necessary or appropriate, it can 
be set to search for a good multiple alignment by processing any pattern in New in sections or `windows', 
one window at a time, in left-to-right order, forming intermediate multiple alignments at each stage, as described in Section \ref{windows_section}. The size of the window can vary from one symbol to the whole of New.

In this mode of operation, with each window containing only one symbol, it is possible to 
start the search with only the `Start' symbol in New and then add `yes' and `no' symbols to it, 
one symbol at a time, in response to the questions which are posed in each of the best intermediate 
multiple alignments.

\subsection{Multiple alignment and engine fault diagnosis}

\begin{figure}[!hbt]
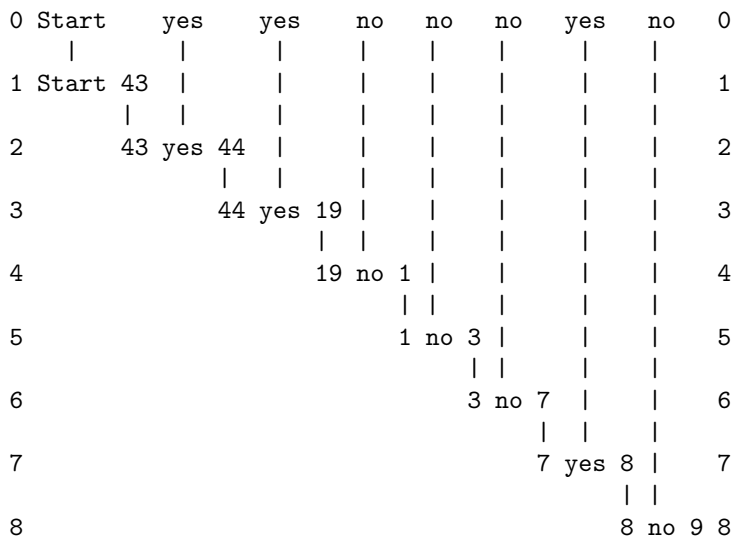

\fontsize{10.00pt}{12.00pt}
\centering
\begin{BVerbatim}
0 Start    yes    yes    no   no   no   yes   no   0
    |       |      |     |    |    |     |    |   
1 Start 43  |      |     |    |    |     |    |    1
        |   |      |     |    |    |     |    |   
2       43 yes 44  |     |    |    |     |    |    2
               |   |     |    |    |     |    |   
3              44 yes 19 |    |    |     |    |    3
                      |  |    |    |     |    |   
4                     19 no 1 |    |     |    |    4
                            | |    |     |    |   
5                           1 no 3 |     |    |    5
                                 | |     |    |   
6                                3 no 7  |    |    6
                                      |  |    |   
7                                     7 yes 8 |    7
                                            | |   
8                                           8 no 9 8
\end{BVerbatim}
\caption{The best multiple alignment found by SP61 with the pattern `Start yes yes no no no yes 
no' in New and the patterns from Figure \ref{decision_tree_rules} in Old. SP61 was set to operate in the 
progressive manner described in the text.}
\label{decision_tree_alignment_1}
\end{figure}

Figure \ref{decision_tree_alignment_1} shows the best multiple alignment found by SP61 with the patterns from Figure \ref{decision_tree_rules} in Old and the pattern `Start yes yes no no no yes no', supplied symbol-by-symbol to New.

As usual, every symbol in a pattern from Old which does not form a hit with a 
symbol in New represents an inference which can be drawn from the multiple alignment. However, the most 
important of these is the last one, the symbol `9' which represents a terminal node containing the 
text ``There is an HT fault or, possibly, a capacitor fault''. This is the diagnosis which the system 
offers in response to the sequence of yes/no symbols in New representing answers to questions 
posed by the system. All the other `inferences'---the symbols `43', `44', `19' etc---merely 
represent the questions which have been answered by the yes/no symbols in New.

\subsection{Probabilities}

What about probabilities? In this case, the reference set of multiple alignments (which encode all 
and only the same symbols from New as the best multiple alignment) contains only one member---the best 
multiple alignment itself. In this case, the relative probability of the diagnosis is 1.0. In other words, this 
probabilistic version of the system provides an all-or-nothing answer in the same manner as the 
non-probabilistic chart from which it was derived. Of course, this probability value, like any other, is subject to the qualifications that were noted in Section \ref{confidence_in_inferences}.

\subsection{So what?}

Regarding the example which has been discussed, readers may object that the SP scheme is a 
long-winded way to achieve something which is done perfectly adequately with a conventional 
expert system or even the original flow chart on paper from which the example was derived. Has 
anything been gained by re-casting the example in an unfamiliar form?

In this case, the answer is ``probably not''. The main reason for including the example in 
this chapter is to show that multiple alignment as it has been developed in the SP system 
has a much broader scope than may, at first sight, be assumed.

For any particular domain, a system with this kind of broad scope may not have any 
particular advantage over a system which is dedicated to that domain and can function effectively 
only in that domain. The benefits of using a generic system rather than different systems 
for each area of application are described in Section \ref{simplification_of_computing_systems}.

\subsection{Information which is incomplete}\label{information_which_is_incomplete}

Another possible response to the ``So what?'' question, above, is that the SP system, unlike 
most conventional systems, does not depend exclusively on input which is both complete and 
accurate. One of the strengths of the SP system is that it can bridge gaps in information 
supplied to it (as New) and can compensate for symbols which have been added or substituted in 
the input, provided there are not too many.

It is not immediately obvious why this kind of capability might be useful for a diagnostic 
system like the example above but it is interesting to see that SP61 can produce a plausible result 
even when parts of the input (in New) are missing or when there is addition or substitution of 
`wrong' symbols.

SP61 has been run in sequential processing mode with the patterns from Figure \ref{decision_tree_rules} in Old and the pattern `Start 43 yes 44 7 yes no' in New. Apart from the addition of some 
`correct' number symbols (for a reason which is given in a moment), this pattern in New is the same 
as in Figure \ref{decision_tree_alignment_1} except that the middle sequence of yes/no symbols (`yes no no no') is missing. By contrast with the example in Figure \ref{decision_tree_alignment_1}, it is necessary in this case to include in New some of the number symbols from the relevant patterns in Old, otherwise there is too much ambiguity about which questions are being answered by the symbols `yes' and `no'.

\begin{figure}[!hbt]
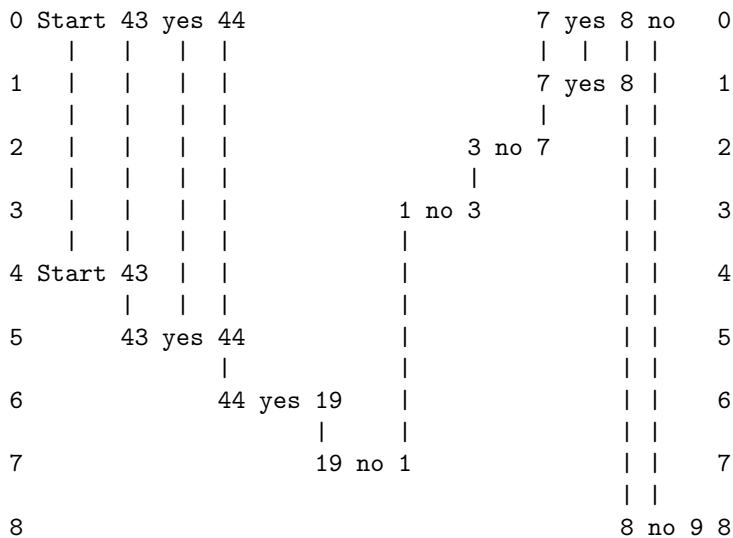

\fontsize{10.00pt}{12.00pt}
\centering
\begin{BVerbatim}
0 Start 43 yes 44                     7 yes 8 no   0
    |   |   |  |                      |  |  | |   
1   |   |   |  |                      7 yes 8 |    1
    |   |   |  |                      |     | |   
2   |   |   |  |                 3 no 7     | |    2
    |   |   |  |                 |          | |   
3   |   |   |  |            1 no 3          | |    3
    |   |   |  |            |               | |   
4 Start 43  |  |            |               | |    4
        |   |  |            |               | |   
5       43 yes 44           |               | |    5
               |            |               | |   
6              44 yes 19    |               | |    6
                      |     |               | |   
7                     19 no 1               | |    7
                                            | |   
8                                           8 no 9 8
\end{BVerbatim}
\caption{The best multiple alignment produced by SP61 with the patterns from Figure \ref{decision_tree_rules} in Old and the pattern `Start 43 yes 44 7 yes no' in New.}
\label{decision_tree_alignment_2}
\end{figure}

Figure \ref{decision_tree_alignment_2} shows the best multiple alignment produced by SP61 with Old and New as just described. No other multiple alignment is produced which encodes all the symbols in New. The multiple alignment in this figure successfully bridges four steps in the multiple alignment where information from New is missing and 
arrives at the same diagnostic conclusion as the multiple alignment in Figure \ref{decision_tree_alignment_1}. By bridging this gap in New, the multiple alignment has, in effect, inferred what questions should have been asked in places where relevant information is missing and what the answers to those questions should have been.

\subsection{Information containing additions or substitutions}

What about adding `wrong' symbols to New or substituting `wrong' symbols for `correct' 
symbols? With respect to the pattern in New shown in Figure \ref{decision_tree_alignment_1}, adding `no' or `yes' at any point or substituting `no' for `yes' (or {\em vice versa}), will either lead to a different conclusion which is correct in terms of the modified version of New or, if the sequence of `yes' and `no' symbols does not correspond to any possible sequence in the decision network, then the results are unpredictable because of the many alternative ways in which partially `correct' multiple alignments may be formed.

The kind of `errors' that the system can cope with most easily are ones which are clearly 
wrong such as, for example, the addition or substitution in New of symbols which do not match 
any of the symbols in Old. Provided there not too many of them, SP61 can bridge these kinds of 
error in much the same way as we saw in the example of `spelling checking and correction' in 
Section \ref{fuzzy_pattern_recognition}.

As noted above, bridging gaps in information or by-passing inaccurate information is not 
the kind of thing that conventional decision networks or decision trees can do. Although this may 
be nothing more than a `party trick' at present, there may turn out to be situations where these kinds 
of capability would be useful in a decision network or tree.%
\index{decision network or tree|)}

\section{Reasoning with `rules'}\label{reasoning_with_rules_section}

\index{reasoning!rules|(}

The rules in a typical expert system express associations between things in the form `IF condition THEN consequence (or action)'. As we saw with the example in Section \ref{simple_pr_example}, we can express an association quite simply as a pattern like `fire smoke' without the need to make a formal distinction between the `condition' and the `consequence' or `action'. And, as we saw in Sections \ref{one_step_deductive_reasoning} and \ref{abductive_reasoning_section}, it is possible to use patterns like these quite freely in both a `forwards' and a `backwards' direction. As was noted in Section \ref{abductive_reasoning_section}, the SP system allows inferences to be drawn from patterns in a totally flexible way: any subsequence of the symbols in a pattern may function as a condition, with the complementary subsequence as the corresponding inference.

It is easy to form a chain of inference like ``If A then B, if B then C'' from patterns like `A B' and `B C'. But if the patterns are `A B' and `C B', the relative positions of `A' and `C' in the multiple alignment are undefined and they form a `mismatch' as described in Section \ref{mismatches_section}. This means that, in the SP61 model, the multiple alignment is treated as being illegal and is discarded.

A way round this problem which can be used with the current models is to adopt a convention that the symbols in every pattern are arranged in some arbitrary sequence, e.g., alphabetical, and to include a `framework' pattern as described in Sections \ref{ordering_of_symbols_and_patterns} and \ref{ordering_of_symptoms}. This has the effect of ensuring that every symbol type always has a column to itself. This avoids the kind of problem described above and allows patterns to be treated as if they were unordered associations of symbols.

Figure \ref{arson_associations} shows a small set of patterns representing well-known associations---such as the association of fire with smoke or the association of black clouds with rain---together with one pattern (`1 suspect 2 7 petrol 8') representing the fact that a  `suspect' person has been seen with petrol, another pattern (`4 destroy 5 11 the\_barn 12') representing the fact that `the barn' has been destroy(ed) and a framework pattern (`1 2 3 4 5 6 7 8 9 10 11') as mentioned above. In every pattern except the last, the alphabetic symbols are arranged in alphabetical order. Every alphabetical symbol has its own slot in the framework, e.g., `black\_clouds' has been assigned to the position between the service symbols `1' and `2'. Every alphabetical symbol is flanked by the service symbols representing its slot.

\begin{figure}[!hbt]
\fontsize{10.00pt}{12.00pt}
\centering
\begin{BVerbatim}
1 suspect 2 7 petrol 8 (1)
4 destroy 5 11 the_barn 12 (1)
5 fire 6 9 smoke 10 (500)
4 destroy 5 fire 6 (100)
5 fire 6 matches 7 petrol 8 (300)
2 black_clouds 3 8 rain 9 (2000)
2 black_clouds 3 cold 4 10 snow 11 (1000)
1 2 3 4 5 6 7 8 9 10 11 12 (7000)
\end{BVerbatim}
\caption{Patterns in Old representing well-known associations, together with two `facts' (`1 suspect 2 7 petrol 8' and `4 destroy 5 11 the\_barn 12') and a `framework' pattern (`1 2 3 4 5 6 7 8 9 10 11') as described in the text.}
\label{arson_associations}
\end{figure}

\begin{figure}[!hbt]
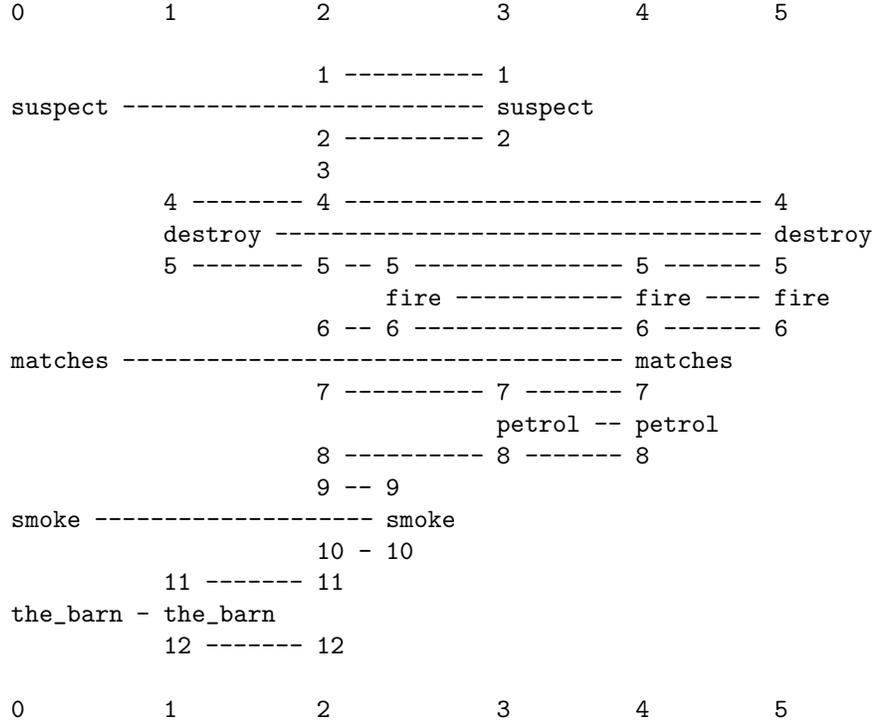

\fontsize{10.00pt}{12.00pt}
\centering
\begin{BVerbatim}
0          1          2            3         4         5

                      1 ---------- 1                          
suspect -------------------------- suspect                    
                      2 ---------- 2                          
                      3                                       
           4 -------- 4 ------------------------------ 4      
           destroy ----------------------------------- destroy
           5 -------- 5 -- 5 --------------- 5 ------- 5      
                           fire ------------ fire ---- fire   
                      6 -- 6 --------------- 6 ------- 6      
matches ------------------------------------ matches          
                      7 ---------- 7 ------- 7                
                                   petrol -- petrol           
                      8 ---------- 8 ------- 8                
                      9 -- 9                                  
smoke -------------------- smoke                              
                      10 - 10                                 
           11 ------- 11                                      
the_barn - the_barn                                           
           12 ------- 12                                      

0          1          2            3         4         5
\end{BVerbatim}
\caption{The best multiple alignment formed by SP61 with a pattern in New representing allegations about someone suspected of arson and patterns in Old as shown in Figure \ref{arson_associations}.}
\label{arson_alignment}
\end{figure}

Figure \ref{arson_alignment} shows the best multiple alignment found by SP61 with the pattern `suspect matches smoke the\_barn' in New and the patterns from Figure \ref{arson_associations} in Old. The pattern in New may be taken to represent the key points in an allegation that a person suspected of arson has been seen with petrol near `the\_barn' (which has been destroyed) and that smoke was seen at the same time.

The alleged facts about the suspect do not, in themselves, show that he/she is guilty of arson. For the jury to find the suspected person guilty, they must understand the connections between the suspect, petrol, smoke and the destruction of the barn. With this example, the inferences are so simple (for people) that the prosecuting lawyer would hardly need to spell them out. But the inferences still need to be made.

The multiple alignment shown in Figure \ref{arson_alignment} may be interpreted as a piecing together of the argument that the suspect used matches to start a fire (which was not witnessed by anyone), and that the fire explains why smoke was seen and why the barn was destroyed. Part of the argument is that the suspect was known to be in possession of petrol. Of course, in a more realistic example, there would be many other clues to the existence of a fire (e.g., charred wood) but the example, as shown, gives an indication of the way in which evidence and inferences may be connected together in the SP paradigm.%
\index{reasoning!rules|)}

\section{Nonmonotonic reasoning and reasoning with default values}\label{nonmonotonic_reasoning_section}

\index{reasoning!nonmonotonic|(}

The concepts of {\em monotonic} and {\em nonmonotonic} reasoning are well explained by \citet{ginsberg_1994}. In brief, conventional deductive inference is {\em monotonic} because deductions made on the strength of current knowledge cannot be invalidated by new knowledge. The conclusion that ``Socrates is mortal'', deduced from ``All humans are mortal'' and ``Socrates is human'' remains true for all time, regardless of anything we learn later. By contrast, the inference that ``Tweety can probably fly'' from the propositions that ``Most birds fly'' and ``Tweety is a bird'' is {\em nonmonotonic} because it may be changed if, for example, we learn that Tweety is a penguin (unless he/she is an astonishing new kind of penguin that can fly).

This section presents some simple examples which show how the SP system can accommodate nonmonotonic reasoning.

\subsection{Typically, birds fly}\label{typically_birds_fly}

In Sections \ref{one_step_deductive_reasoning} and \ref{abductive_reasoning_section}, the idea that (all) birds can fly was expressed with the pattern `Bd bird name \#name canfly warm-blooded wings feathers ... \#Bd'. This, of course, is an oversimplification of the real-world facts because, while it true that the majority of birds fly, we know that there are also flightless birds like ostriches, penguins and kiwis.

In order to model these facts more closely, we need to modify the pattern that describes birds to be something like this:

\begin{center}
\begin{BVerbatim}
Bd bird name #name f #f warm-blooded wings feathers ... #Bd.
\end{BVerbatim}
\end{center}

\noindent And, to our database of Old patterns, we need to add patterns like this:

\begin{center}
\begin{BVerbatim}
Default Bd f canfly #f #Bd #Default
P penguin Bd f cannotfly #f #Bd ... #P
O ostrich Bd f cannotfly #f #Bd ... #O.
\end{BVerbatim}
\end{center}

Now, the pair of symbols `f \#f' in `Bd bird name \#name f \#f warm-blooded wings feathers ... \#Bd' functions like a `variable' that may take the value `canfly' if a given class of birds can fly and `cannotfly' when a type of bird cannot fly. The pattern `P penguin Bd f cannotfly \#f \#Bd ... \#P' shows that penguins cannot fly and, likewise, the pattern `O ostrich Bd f cannotfly \#f \#Bd ... \#O' shows that ostriches cannot fly. The pattern `Default Bd f canfly \#f \#Bd \#Default', which has a substantially higher frequency than the other two patterns, represents the default value for the variable which is `canfly'. Notice that all three of these patterns contains the pair of symbols `Bd ... \#Bd' showing that the corresponding classes are all subclasses of birds.

\subsection{Tweety is a bird so, probably, Tweety can fly}\label{tweety-flies}

\begin{figure}[!hbt]
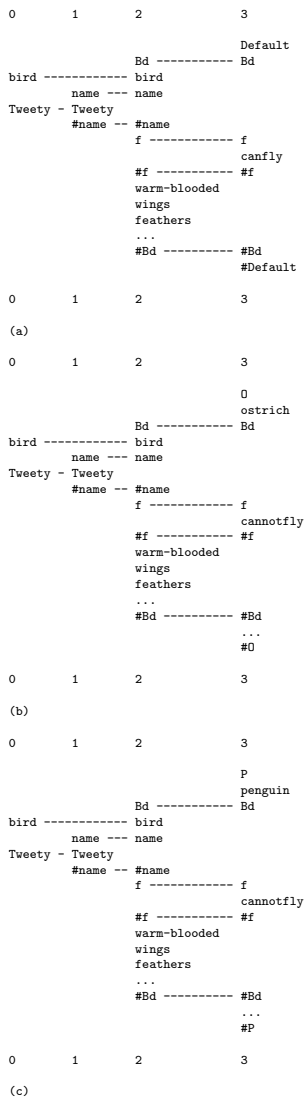

\fontsize{05.00pt}{06.00pt}
\centering
\begin{BVerbatim}
0        1        2              3       

                                 Default 
                  Bd ----------- Bd      
bird ------------ bird                   
         name --- name                   
Tweety - Tweety                          
         #name -- #name                  
                  f ------------ f       
                                 canfly  
                  #f ----------- #f      
                  warm-blooded           
                  wings                  
                  feathers               
                  ...                    
                  #Bd ---------- #Bd     
                                 #Default

0        1        2              3       

(a)

0        1        2              3        

                                 O        
                                 ostrich  
                  Bd ----------- Bd       
bird ------------ bird                    
         name --- name                    
Tweety - Tweety                           
         #name -- #name                   
                  f ------------ f        
                                 cannotfly
                  #f ----------- #f       
                  warm-blooded            
                  wings                   
                  feathers                
                  ...                     
                  #Bd ---------- #Bd      
                                 ...      
                                 #O       

0        1        2              3        

(b)

0        1        2              3        

                                 P        
                                 penguin  
                  Bd ----------- Bd       
bird ------------ bird                    
         name --- name                    
Tweety - Tweety                           
         #name -- #name                   
                  f ------------ f        
                                 cannotfly
                  #f ----------- #f       
                  warm-blooded            
                  wings                   
                  feathers                
                  ...                     
                  #Bd ---------- #Bd      
                                 ...      
                                 #P       

0        1        2              3        

(c)
\end{BVerbatim}
\caption{The best three multiple alignments formed by SP61 with `bird Tweety' in New and patterns in Old as described in the text. The relative probabilities of (a), (b) and (c) are 0.66, 0.22 and 0.12, respectively.}
\label{nonmon_figure_1}
\end{figure}

When SP61 is run with `bird Tweety' in New and the same patterns in Old as before, modified as just described, the three best multiple alignments found are those shown in Figure \ref{nonmon_figure_1}. The first multiple alignment, which has a relative probability of 0.66, confirms that Tweety is a bird and tells us that he or she can fly.

The second multiple alignment, with a relative probability of 0.22, tells us that Tweety might be an ostrich and, as such, he or she would not be able to fly. Likewise, the third multiple alignment tells us that, with a relative probability of 0.12, Tweety might be a penguin and would not be able to fly. The values for probabilities in this simple example are derived from frequencies that are, almost certainly, not ornithologically correct.

\subsection{Tweety is a penguin, so Tweety cannot fly}\label{tweety_is_penguin}

\begin{figure}[!hbt]
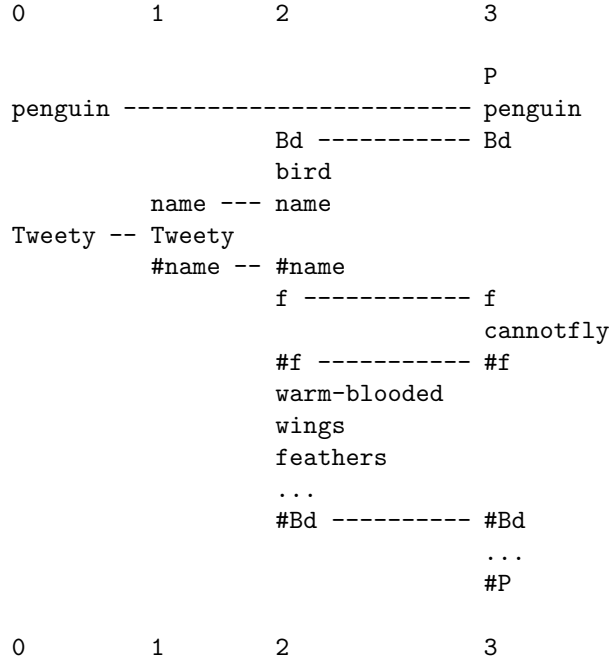

\fontsize{10.00pt}{12.00pt}
\centering
\begin{BVerbatim}
0         1        2              3        

                                  P        
penguin ------------------------- penguin  
                   Bd ----------- Bd       
                   bird                    
          name --- name                    
Tweety -- Tweety                           
          #name -- #name                   
                   f ------------ f        
                                  cannotfly
                   #f ----------- #f       
                   warm-blooded            
                   wings                   
                   feathers                
                   ...                     
                   #Bd ---------- #Bd      
                                  ...      
                                  #P       

0         1        2              3        
\end{BVerbatim}
\caption{The best multiple alignment formed by SP61 with `penguin Tweety' in New and patterns in Old as described in the text. The relative probability of this multiple alignment is 1.0.}
\label{nonmon_figure_2}
\end{figure}

Figure \ref{nonmon_figure_2} shows the best multiple alignment found by SP61 when it is run again, with `penguin Tweety' in New instead of `bird Tweety'. This time, there is only one multiple alignment in the reference set and its relative probability is 1.0. Correspondingly, all inferences that we can draw from this multiple alignment have a probability of 1.0. In particular, we can be confident, within the limits of the available knowledge, that Tweety cannot fly.

\index{reasoning!nonmonotonic|)}

\section{Explaining away `explaining away': the SP system as an alternative to Bayesian networks}\label{explaining_away}

\index{reasoning!explaining away|(}\index{reasoning!causal|(}

In recent years, {\em Bayesian networks} (otherwise known as {\em causal nets}, {\em influence diagrams}, {\em probabilistic networks} and other names) have become popular as a means of representing probabilistic knowledge and for probabilistic reasoning \citep[see][]{pearl_1988}.

A Bayesian network is a directed, acyclic graph like the one shown in Figure \ref{alarm_bayesian_network} (below) where each node has zero or more `inputs' (connections with nodes that can influence the given node) and one or more `outputs' (connections to other nodes that the given node can influence).

Each node contains a set of conditional probability values, each one the probability of a given output value for a given input value. With this information, conditional probabilities of alternative outputs for any node may be computed for any given {\em combination} of inputs. By combining these calculations for sequences of nodes, probabilities may be propagated through the network from one or more `start' nodes to one or more `finishing' nodes.

This section shows how the SP system provides alternative to the Bayesian network explanation of the phenomenon of ``explaining away''.

\subsection{A Bayesian network explanation of ``explaining away''}\label{bayesian_network}

In the words of Judea Pearl \citeyearpar[p. 7]{pearl_1988}, the phenomenon of `explaining away' may be characterised as: ``If A implies B, C implies B, and B is true, then finding that C is true makes A {\em less} credible. In other words, finding a second explanation for an item of data makes the first explanation less credible.'' (his italics). Here is an example:

\begin{quotation}

``Normally an alarm sound alerts us to the possibility of a burglary. If somebody calls you at the office and tells you that your alarm went off, you will surely rush home in a hurry, even though there could be other causes for the alarm sound. If you hear a radio announcement that there was an earthquake nearby, and if the last false alarm you recall was triggered by an earthquake, then your certainty of a burglary will diminish.'' \citep[][pp. 8-9]{pearl_1988}.

\end{quotation}

Although it is not normally presented as an example of nonmonotonic reasoning, this kind of effect in the way we react to new information is similar to the example we considered in Section \ref{nonmonotonic_reasoning_section} because new information has an impact on inferences that we formed on the basis of information that was available earlier.

\begin{figure}[!hbt]
\centering
\includegraphics[width=0.9\textwidth]{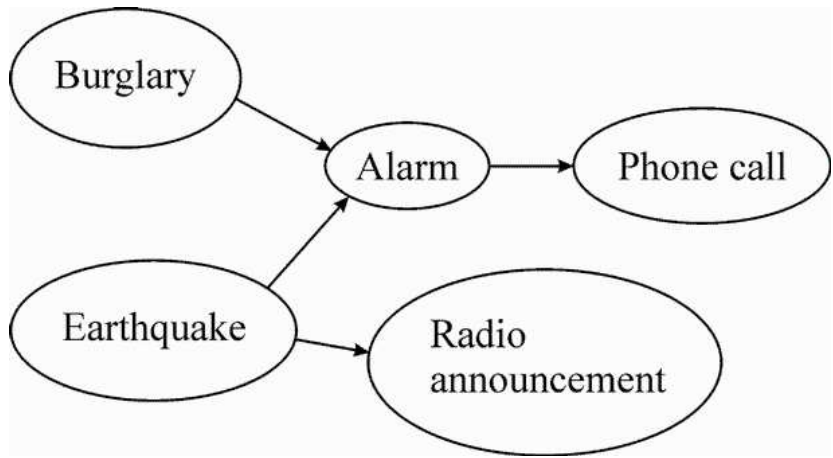}
\caption{A Bayesian network representing causal relationships discussed in the text. In this diagram, ``Phone call'' means ``a phone call about the alarm going off'' and ``Radio announcement'' means ``a radio announcement about an earthquake''.}
\label{alarm_bayesian_network}
\end{figure}

The causal relationships in the example just described may be captured in a Bayesian network like the one shown in Figure \ref{alarm_bayesian_network}.

Pearl argues that, with appropriate values for conditional probabilities, the phenomenon of ``explaining away'' can be explained in terms of this network (representing the case where there is a radio announcement of an earthquake) compared with the same network without the node for ``radio announcement'' (representing the situation where there is no radio announcement of an earthquake).

\subsection{Representing contingencies with patterns and frequencies}\label{contingencies}

To see how this phenomenon may be understood in terms of the SP theory, consider, first, the set of patterns shown in Figure \ref{alarm_patterns}, which are to be stored in Old. The first four patterns in the figure show events which occur together in some notional sample of the `World' together with their frequencies of occurrence in the sample.

Like other knowledge-based systems, an SP system would normally be used with a `closed-world' assumption that, for some particular domain, the knowledge stored in the knowledge base is comprehensive. Thus, for example, a travel booking clerk using a database of all flights between cities will assume that, if no flight is shown between, say, Edinburgh and Paris, then no such flight exists. Of course, the domain may be only `flights provided by one particular airline', in which case the booking clerk would need to check databases for other airlines. In systems like Prolog, the closed-world assumption is the basis of `negation as failure': if a proposition cannot be proved with the clauses provided in a Prolog program then, in terms of that store of knowledge, the proposition is assumed to be false.

In the present case, we shall assume that the closed-world assumption applies so that the absence of any pattern may be taken to mean that the corresponding pattern of events did not occur, at least not with a frequency greater than one would expect by chance.

\begin{figure}[!hbt]
\fontsize{10.00pt}{12.00pt}
\centering
\begin{BVerbatim}
alarm phone_alarm_call (980)
earthquake alarm (20)
earthquake radio_earthquake_announcement (40)
burglary alarm (1000)
e1 earthquake e2 (40)
\end{BVerbatim}
\caption{A set of patterns to be stored in Old in an example of `explaining away'. The symbol `phone\_alarm\_call' is intended to represent a phone call conveying news that the alarm sounded; `radio\_earthquake\_announcement' represents an announcement on the radio that there has been an earthquake. The symbols `e1' and `e2' represent other contexts for `earthquake' besides the contexts `alarm' and `radio\_earthquake\_announcement'.}
\label{alarm_patterns}
\end{figure}

The fourth pattern shows that there were 1000 occasions when there was a burglary and the alarm went off and the second pattern shows just 20 occasions when there was an earthquake and the alarm went off (presumably triggered by the earthquake). Thus we have assumed that burglaries are much more common than earthquakes. Since there is no pattern showing the simultaneous occurrence of an earthquake, burglary and alarm, we shall infer from the closed-world assumption that this constellation of events was not recorded during the sampling period.

The first pattern shows that, out of the 1020 cases when the alarm went off, there were 980 cases where a telephone call about the alarm was made. Since there is no pattern showing telephone calls (about the alarm) in any other context, the closed-world assumption allows us to assume that there were no false positives (including hoaxes): telephone calls about the alarm when no alarm had sounded.

Some of the frequencies shown in Figure \ref{alarm_patterns} are intended to reflect the two probabilities suggested for this example in \citet[p. 49]{pearl_1988}: ``... the [alarm] is sensitive to earthquakes and can be accidentally (P = 0.20) triggered by one. ... if an earthquake had occurred, it surely (P = 0.40) would be on the [radio] news.''

In our example, the frequency of `earthquake alarm' is 20, the frequency of `earthquake radio\_earthquake\_announcement' is 40 and the frequency of `earthquake' in other contexts is 40. Since there is no pattern like `earthquake alarm radio\_earthquake\_announcement' or `earthquake radio\_earthquake\_announcement alarm' representing cases where an earthquake triggers the alarm and also leads to a radio announcement, we may assume that cases of that kind have not occurred. As before, this assumption is based on the closed-world assumption that the set of patterns is a reasonably comprehensive representation of non-random associations in this small world.

The pattern at the bottom, with its frequency, shows that an earthquake has occurred on 40 occasions in contexts where the alarm did not ring and there was no radio announcement.

\subsection{Approximating the temporal order of events}\label{temporal-order}

In these patterns and in the multiple alignments shown below, the left-to-right order of symbols may be regarded as an approximation to the order of events in time. Thus, in the first pattern, `phone\_alarm\_call' (a phone call to say the alarm has gone off) follows `alarm' (the alarm itself); in the second pattern, `alarm' follows `earthquake' (the earthquake which, we may guess, triggered the alarm); and so on. A single dimension can only approximate the order of events in time because it cannot represent events which overlap in time or which occur simultaneously. However, this kind of approximation has little or no bearing on the points to be illustrated here.

\subsection{Other considerations}\label{other_explaining_away_considerations}

Other points relating to the patterns shown in Figure \ref{alarm_patterns} include:

\begin{itemize}

\item No attempt has been made to represent the idea that ``the last false alarm you recall was triggered by an earthquake'' \citep[][p. 9]{pearl_1988}. At some stage in the development of the SP system, there will be a need to take account of recency (see Section \ref{recency_section}).

\item With these imaginary frequency values, it has been assumed that burglaries (with a total frequency of occurrence of 1160) are much more common than earthquakes (with a total frequency of 100). As we shall see, this difference reinforces the belief that there has been a burglary when it is known that the alarm has gone off (but without additional knowledge of an earthquake).

\item In accordance with Pearl's example (p. 49) (but contrary to the phenomenon of looting during earthquakes), it has been assumed that earthquakes and burglaries are independent. If there was some association between them, then, in accordance with the closed-world assumption, there should be a pattern in Figure \ref{alarm_patterns} representing the association.

\end{itemize}

\subsection{Formation of alignments: the burglar alarm has sounded}\label{burglar_alarm_sounded}

\begin{figure}[!hbt]
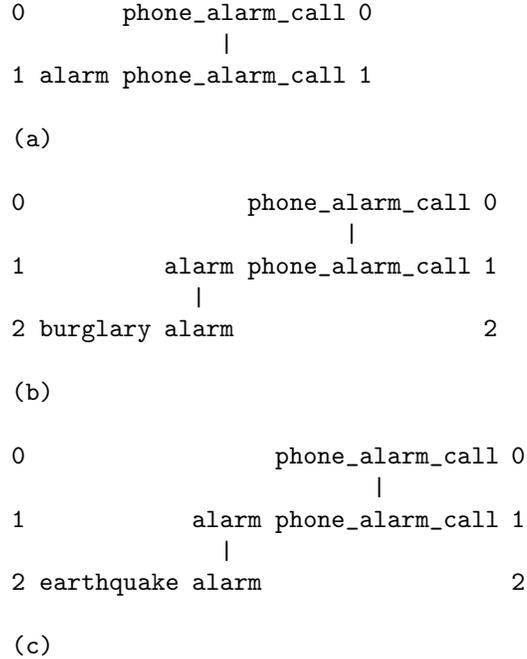

\fontsize{10.00pt}{12.00pt}
\centering
\begin{BVerbatim}
0       phone_alarm_call 0
               |        
1 alarm phone_alarm_call 1

(a)

0                phone_alarm_call 0
                        |        
1          alarm phone_alarm_call 1
             |                   
2 burglary alarm                  2

(b)

0                  phone_alarm_call 0
                          |        
1            alarm phone_alarm_call 1
               |                   
2 earthquake alarm                  2

(c)
\end{BVerbatim}
\caption{The best multiple alignment (at the top) and the other three multiple alignments in its reference set formed by SP61 with the pattern `phone\_alarm\_call' in New and the patterns from Figure \ref{alarm_patterns} in Old. In order from the top, the values for $CD$ with relative probabilities in brackets are: 19.91 (0.6563), 18.91 (0.3281), 14.52 (0.0156).}
\label{alarm_alignments_1}
\end{figure}

Receiving a phone call to say that one's house alarm has gone off may be represented by placing the symbol `phone\_alarm\_call' in New. Figure \ref{alarm_alignments_1} shows, at the top, the best multiple alignment formed by SP61 in this case with the patterns from Figure \ref{alarm_patterns} in Old. The other two multiple alignments in the reference set are shown below the best multiple alignment, in order of CD value and relative probability. The actual values for $CD$ and relative probability are given in the caption to Figure \ref{alarm_patterns}.

The unmatched symbols in these multiple alignments represent inferences made by the system. The probabilities for these inferences which are calculated by SP61 (using the method described in Section \ref{probabilities_section}) are shown in Table \ref{symbol_probabilities_table}. These probabilities do not add up to 1 and we should not expect them to because any given multiple alignment can contain two or more of these symbols.

The most probable inference is the rather trivial inference that the alarm has indeed sounded. This reflects the fact that there is no pattern in Figure \ref{alarm_patterns} representing false positives for telephone calls about the alarm. Apart from the inference that the alarm has sounded, the most probable inference (p = 0.3281) is that there has been a burglary. However, there is a distinct possibility that there has been an earthquake---but the probability in this case (p = 0.0156) is much lower than the probability of a burglary.

\begin{table}
\centering
\begin{tabular}{ll}
\em Symbol & \em Probability \\
\\
alarm & 1.0 \\
burglary & 0.3281 \\
earthquake & 0.0156 \\
\end{tabular}
\caption{The probabilities of unmatched symbols, calculated by SP61 for the four multiple alignments shown in Figure \ref{alarm_alignments_1}. The probability of `phone\_alarm\_call' is 1.0 because it is supplied as a `fact' in New.}
\label{symbol_probabilities_table}
\end{table}

These inferences and their relative probabilities seem to accord quite well with what one would naturally think following a telephone call to say that the burglar alarm at one's house has gone off (given that one was living in a part of the world where earthquakes were not vanishingly rare).

\subsection{Formation of alignments: the burglar alarm has sounded and there is a radio announcement of an earthquake}\label{radio-announcement}

In this example, the phenomenon of `explaining away' occurs when you learn not only that the burglar alarm has sounded but that there has been an announcement on the radio that there has been an earthquake. In terms of the SP model, the two events (the phone call about the alarm and the announcement about the earthquake) can be represented in New by a pattern like this:

\begin{center}
\begin{BVerbatim}
phone_alarm_call radio_earthquake_announcement
\end{BVerbatim}
\end{center}

\noindent or `radio\_earthquake\_announcement phone\_alarm\_call'. The order of the two symbols does not matter because it makes no difference to the result, except for the order in which columns appear in the best multiple alignment.

\begin{figure}[!hbt]
\fontsize{09.00pt}{10.80pt}
\centering
\begin{BVerbatim}
0                  phone_alarm_call radio_earthquake_announcement 0
                          |                       |              
1            alarm phone_alarm_call               |               1
               |                                  |              
2 earthquake alarm                                |               2
      |                                           |              
3 earthquake                        radio_earthquake_announcement 3

(a)

0 phone_alarm_call radio_earthquake_announcement 0
                                 |              
1 earthquake       radio_earthquake_announcement 1

(b)

0       phone_alarm_call radio_earthquake_announcement 0
               |                                      
1 alarm phone_alarm_call                               1

(c)

0                phone_alarm_call radio_earthquake_announcement 0
                        |                                      
1          alarm phone_alarm_call                               1
             |                                                 
2 burglary alarm                                                2

(d)

0                  phone_alarm_call radio_earthquake_announcement 0
                          |                                      
1            alarm phone_alarm_call                               1
               |                                                 
2 earthquake alarm                                                2

(e)
\end{BVerbatim}
\caption{At the top, the best multiple alignment formed by SP61 with the pattern `phone\_alarm\_call radio\_earthquake\_announcement' in New and the patterns from Figure \ref{alarm_patterns} in Old. Other multiple alignments formed by SP61 are shown below. From the top, the $CD$ values are: 74.64, 54.72, 19.92, 18.92, and 14.52.}
\label{alarm_alignments_2}
\end{figure}

In this case, there is only one multiple alignment (shown at the top of Figure \ref{alarm_alignments_2}) that can `explain' all the information in New. Since there is only this one multiple alignment in the reference set for the best multiple alignment, the associated probabilities of the inferences that can be read from the multiple alignment (`alarm' and `earthquake') are 1.0.

These results show broadly how `explaining away' may be explained in terms of the SP theory. The main point is that the multiple alignment or multiple alignments that provide the best `explanation' of a telephone call to say that one's burglar alarm has sounded is different from the multiple alignment or multiple alignments that best explain the same telephone call coupled with an announcement on the radio that there has been an earthquake. In the latter case, the best explanation is that the earthquake triggered the alarm. Other possible explanations have lower probabilities.

\subsection{Other possible alignments}\label{other_possible_alignments}

The foregoing account of `explaining away' in terms of the SP theory is not entirely satisfactory because it does not say enough about alternative explanations of what has been observed. This subsection tries to plug this gap. What is missing from the account of `explaining away' in the previous subsection is any consideration of such other possibilities as, for example:

\begin{itemize}

\item A burglary (which triggered the alarm) and, at the same time, an earthquake (which led to a radio announcement), or

\item An earthquake that triggered the alarm and led to a radio announcement and, at the same time, a burglary that did not trigger the alarm.

\item And many other unlikely possibilities of a similar kind.

\end{itemize}

\begin{figure}[!hbt]
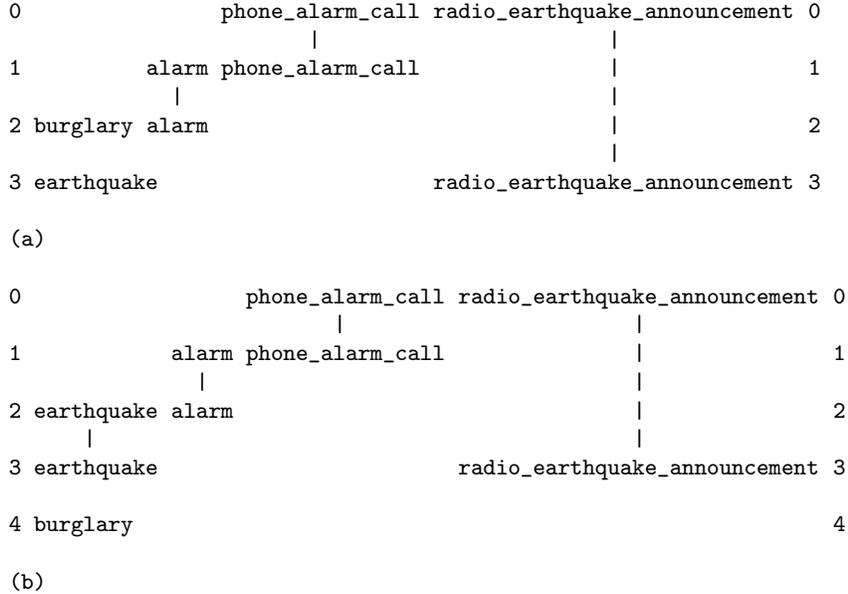

\fontsize{09.00pt}{10.80pt}
\centering
\begin{BVerbatim}
0                phone_alarm_call radio_earthquake_announcement 0
                        |                       |              
1          alarm phone_alarm_call               |               1
             |                                  |              
2 burglary alarm                                |               2
                                                |              
3 earthquake                      radio_earthquake_announcement 3

(a)

0                  phone_alarm_call radio_earthquake_announcement 0
                          |                       |              
1            alarm phone_alarm_call               |               1
               |                                  |              
2 earthquake alarm                                |               2
      |                                           |
3 earthquake                        radio_earthquake_announcement 3

4 burglary                                                        4

(b)
\end{BVerbatim}
\caption{Two multiple alignments discussed in the text. (a) A multiple alignment created by combining the second and fourth multiple alignment from Figure \ref{alarm_alignments_2}. $CD$ = 73.64, Absolute P = 5.5391e-5. (b) A multiple alignment created from the first multiple alignment in Figure \ref{alarm_alignments_2} and the symbol `burglary'. $CD$ = 72.57, Absolute P = 2.6384e-5.}
\label{alarm_alignments_3}
\end{figure}

Alternatives of this kind may be created by combining multiple alignments shown in Figure \ref{alarm_alignments_2} with each other, or with patterns or symbols from Old, or both these things. The two examples just mentioned are shown in Figure \ref{alarm_alignments_3}.

Any multiple alignment created by combining multiple alignments as just described may be evaluated in exactly the same way as the multiple alignments formed directly by SP61. $CD$s and absolute probabilities for the two example multiple alignments are shown in the caption to Figure \ref{alarm_alignments_3}.

Given the existence of multiple alignments like those shown in Figure \ref{alarm_alignments_3}, values for relative probabilities of multiple alignments will change. The best multiple alignment from Figure \ref{alarm_alignments_2} and the two multiple alignments from Figure \ref{alarm_alignments_3} constitute a reference set because they all `encode' the same symbols from New. However, there are probably several other multiple alignments that one could construct that would belong in the same reference set.

Given a reference set containing the first multiple alignment in Figure \ref{alarm_alignments_2} and the two multiple alignments in Figure \ref{alarm_alignments_3}, values for relative probabilities are shown in Table \ref{absolute_and_relative_probabilities}, together with the absolute probabilities from which they were derived. Whichever measure is used, the multiple alignment which was originally judged to represent the best interpretation of the available facts has not been dislodged from this position.

\begin{table}
\centering
\begin{tabular}{lll}
\em Alignment & \em Absolute & \em Relative \\
 & \em probability & \em probability \\
\\
(a) in Figure \ref{alarm_alignments_2} & 1.1052e-4 & 0.5775 \\
(a) in Figure \ref{alarm_alignments_3} & 5.5391e-5 & 0.2881 \\
(b) in Figure \ref{alarm_alignments_3} & 2.6384e-5 & 0.1372 \\
\end{tabular}
\caption{\small Values for absolute and relative probability for the best multiple alignment in Figure \ref{alarm_alignments_2} and the two multiple alignments in Figure \ref{alarm_alignments_3}.}
\label{absolute_and_relative_probabilities}
\end{table}

\index{reasoning!explaining away|)}

\section{Causal diagnosis}\label{causal_diagnosis_section}

\index{diagnosis!causal|(}\index{diagnosis!fault finding|(}

As we saw in Section \ref{medical_diagnosis_section}, medical diagnosis may be viewed as a process of pattern recognition but the diagnostic process may also involve reasoning about the causes of a patient's symptoms (Section \ref{medical_causal_reasoning}). Causal reasoning is, perhaps, even more prominent in the process of diagnosing faults in artificial systems (cars, televisions etc), probably because these systems are simpler than the human body and better understood.

\begin{figure}[!hbt]
\centering
\includegraphics[width=0.9\textwidth]{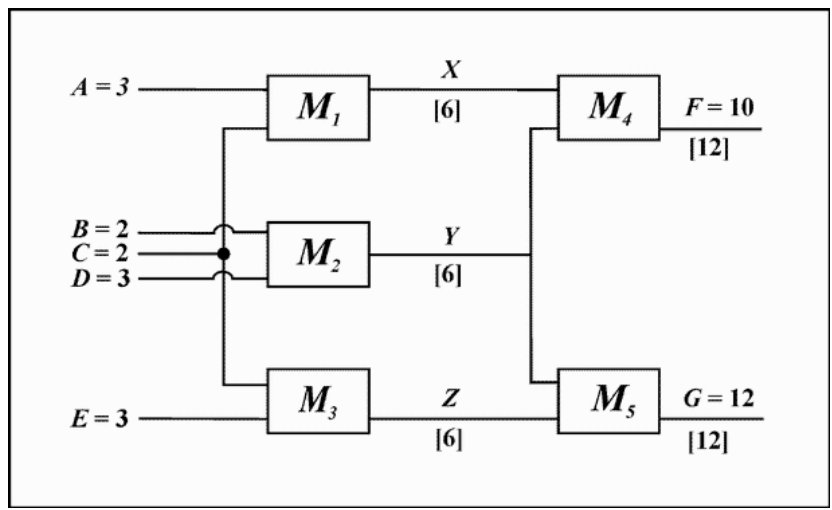}
\caption{An electronic circuit containing three multipliers, $M_1$, $M_2$, and $M_3$, and two adders, $M_4$ and $M_5$ (Redrawn from \citet[p. 263]{pearl_1988}).}
\label{electronic_circuit_figure}
\end{figure}

In this section, we consider a simple example of fault diagnosis in an electronic circuit---described by \citet[pp. 263--272]{pearl_1988}. Figure \ref{electronic_circuit_figure} shows the circuit with inputs on the left, outputs on the right and, in between, three multipliers ($M_1$, $M_2$, and $M_3$) and two adders ($M_4$ and $M_5$). For the given inputs on the left, it is clear that output F is false and output G is correct.

\begin{figure}[!hbt]
\centering
\includegraphics[width=0.9\textwidth]{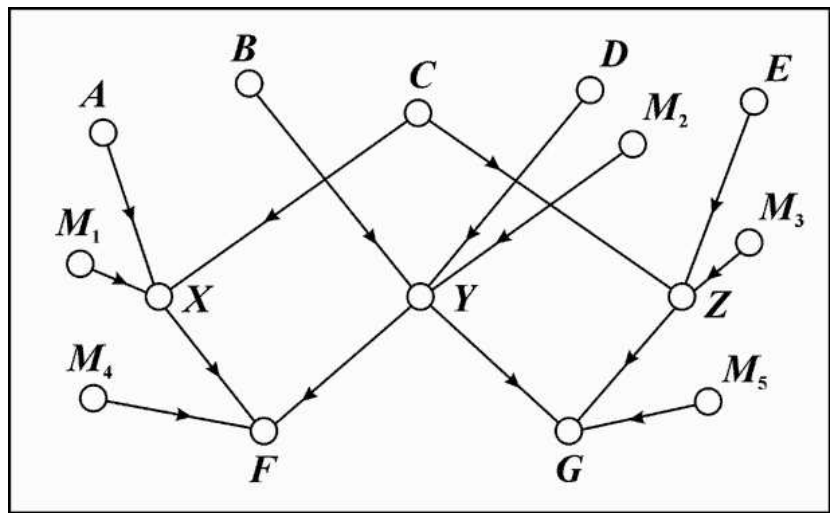}
\caption{A causal network derived from the electronic circuit in Figure \ref{electronic_circuit_figure} (Redrawn from \citet[p. 264]{pearl_1988}).}
\label{electronic_circuit_network}
\end{figure}

Figure \ref{electronic_circuit_network} shows a causal network derived from the electronic circuit in Figure \ref{electronic_circuit_figure} (from \citet[p. 264]{pearl_1988}). In this diagram, each of the nodes $X$, $Y$, $Z$, $F$ and $G$ represent the outputs of components $M_1$, $M_2$, $M_3$, $M_4$ and $M_5$, respectively. In each case, there are three causal influences on the output: the two inputs to the component and the state of the component which may be `good' or `bad'. These influences are shown by lines with arrows connecting the source of the influence to the target node. Thus, for example, the two inputs of component $M_1$ are represented by $A$ and $C$ in Figure \ref{electronic_circuit_network}, the good or bad state of component $M_1$ is represented by the node labelled $M_1$, and their causal influences on node $X$ are shown by the three arrows pointing at that node.

Given a causal analysis like this, and given appropriate information about conditional probabilities, it is possible to derive one or more alternative diagnoses of which components are good and which are bad. In Pearl's example, it is assumed that components of the same type have the same prior probability of failure and that the probability of failure of multipliers is greater than for adders. Given these and some subsidiary assumptions together with the inputs and outputs (but not the intermediate values) shown in Figure \ref{electronic_circuit_figure}, the best diagnosis derived from the causal network is that the $M_1$ component is bad and the second best diagnosis is that $M_4$ is bad. Pearl indicates that some third-best interpretations may be retrievable (e.g., $M_2$ and $M_5$ are bad) ``... but in general, it is not guaranteed that interpretations beyond the second-best will be retrievable.'' (p. 272). 

\subsection{An SP approach to causal diagnosis}\label{sp_causal_diagnosis}

The main elements of the SP analysis presented here are as follows:

\begin{itemize}

\item The input-output relations of any component may be represented as a set of patterns, each one with a measured or estimated frequency of occurrence.

\item With suitable extensions, these patterns may serve to transfer the output of one component to the input of another.

\item A framework pattern (as described in Sections \ref{ordering_of_symbols_and_patterns} and elsewhere) is needed to ensure that appropriate multiple alignments can be built.

\end{itemize}

Figure \ref{sp_causal_diagnosis_patterns} shows a set of patterns for the circuit shown in Figure \ref{electronic_circuit_figure}. In the figure, the patterns that start with the symbol `M1' represent I/O relations for component $M_1$, those that start with `M2' represent I/O relations for the $M_2$ component and likewise for the other patterns except the last one (starting with the symbol `frame') which is the framework pattern mentioned above. For each initial symbol there is a corresponding `terminating' symbol with an initial `\#' character. For reasons explained shortly, there may be other symbols following the `terminating' symbol.

\begin{figure}[!hbt]
\fontsize{10.00pt}{12.00pt}
\centering
\begin{BVerbatim}
M1 M1GOOD TM1I1 TM1I2 TM1O #M1 TM4I2 (500000)
M1 M1BAD TM1I1 TM1I2 TM1O #M1 TM4I2 (4)
M1 M1BAD TM1I1 TM1I2 FM1O #M1 FM4I2 (96)
M2 M2GOOD TM2I1 TM2I2 TM2O #M2 TM4I1 TM5I2 (500000)
M2 M2BAD TM2I1 TM2I2 TM2O #M2 TM4I1 TM5I2 (4)
M2 M2BAD TM2I1 TM2I2 FM2O #M2 FM4I1 FM5I2 (96)
M3 M3GOOD TM3I1 TM3I2 TM3O #M3 TM5I1 (500000)
M3 M3BAD TM3I1 TM3I2 TM3O #M3 TM5I1 (4)
M3 M3BAD TM3I1 TM3I2 FM3O #M3 FM5I1 (96)
M4 M4GOOD TM4I1 TM4I2 TM4O #M4 (250000)
M4 M4GOOD TM4I1 FM4I2 FM4O #M4 (250000)
M4 M4GOOD FM4I1 TM4I2 FM4O #M4 (250000)
M4 M4GOOD FM4I1 FM4I2 FM4O #M4 (250000)
M4 M4BAD TM4I1 TM4I2 FM4O #M4 (24)
M4 M4BAD TM4I1 FM4I2 FM4O #M4 (24)
M4 M4BAD FM4I1 TM4I2 FM4O #M4 (24)
M4 M4BAD FM4I1 FM4I2 FM4O #M4 (24)
M4 M4BAD TM4I1 TM4I2 TM4O #M4 (1)
M4 M4BAD TM4I1 FM4I2 TM4O #M4 (1)
M4 M4BAD FM4I1 TM4I2 TM4O #M4 (1)
M4 M4BAD FM4I1 FM4I2 TM4O #M4 (1)
M5 M5GOOD TM5I1 TM5I2 TM5O #M5 (250000)
M5 M5GOOD TM5I1 FM5I2 FM5O #M5 (250000)
M5 M5GOOD FM5I1 TM5I2 FM5O #M5 (250000)
M5 M5GOOD FM5I1 FM5I2 FM5O #M5 (250000)
M5 M5BAD TM5I1 TM5I2 FM5O #M5 (24)
M5 M5BAD TM5I1 FM5I2 FM5O #M5 (24)
M5 M5BAD FM5I1 TM5I2 FM5O #M5 (24)
M5 M5BAD FM5I1 FM5I2 FM5O #M5 (24)
M5 M5BAD TM5I1 TM5I2 TM5O #M5 (1)
M5 M5BAD TM5I1 FM5I2 TM5O #M5 (1)
M5 M5BAD FM5I1 TM5I2 TM5O #M5 (1)
M5 M5BAD FM5I1 FM5I2 TM5O #M5 (1)
frame M1 #M1 M2 #M2 M3 #M3 M4 #M4 M5 #M5 #frame (1)
\end{BVerbatim}
\caption{A set of SP patterns modelling I/O relations in the electronic circuit shown in Figure \ref{electronic_circuit_figure}. They were supplied as Old patterns to SP61 for the building of  the multiple alignment shown in Figure \ref{sp_causal_diagnosis_alignment}. {\em Key}: T = true (information is correct); F = false (information is incorrect); M1, M2, M3, M4, M5 = components of the circuit; GOOD, BAD indicates whether a component is good or bad; I1, I2 = First and second inputs of a component; O = Output of a component.}
\label{sp_causal_diagnosis_patterns}
\end{figure}

Let us now consider the first pattern in the figure (`M1 M1GOOD TM1I1 TM1I2 TM1O \#M1 TM4I2')     
 representing I/O relations for component $M_1$ when that component is good, as indicated by the symbol `M1GOOD'. In this pattern, the symbols `TM1I1', `TM1I2' and `TM1O' represent the two inputs and the output of the component, `\#M1' is the terminating symbol, and `TM4I2' serves to transfer the output of $M_1$ to the second input of component $M_4$ as will be explained. In a symbol like `TM1I1', `T' indicates that the input is true, `M1' identifies the component, and `I1' indicates that this is the first input of the component. Other symbols may be interpreted in a similar way, following the key given in the caption of Figure \ref{sp_causal_diagnosis_patterns}. In effect, this pattern says that, when the component is working correctly, true inputs yield a true output. The pattern has a relatively high frequency of occurrence (500000) reflecting the idea that the component will normally work correctly.

The other two patterns for component $M_1$ (`M1 M1BAD TM1I1 TM1I2 TM1O \#M1 TM4I2' and
`M1 M1BAD TM1I1 TM1I2 FM1O \#M1 FM4I2') describe I/O relations when the component is bad. The first one describes the situation where true inputs to a faulty component yield a true result, a possibility noted by Pearl (p. 265). The second pattern---with a higher frequency---describes the more usual situation where true inputs to a faulty component yield a false result. Both these bad patterns have much lower frequencies than the good pattern.

The other patterns in Figure \ref{sp_causal_diagnosis_patterns} may be interpreted in a similar way. Components $M_1$, $M_2$ and $M_3$ have only three patterns each because it is assumed that inputs to the circuit will always be true so it is not necessary to include patterns describing what happens when one or both of the inputs are false. By contrast, there are 4 good  patterns and 8 bad patterns for each of $M_4$ and $M_5$ because either of these components may receive faulty input.

For each of the five components, the frequencies of the bad patterns sum to 100. However, for each of the components $M_1$, $M_2$, and $M_3$, the total frequency of the good patterns is 500,000 compared with 1,000,000 for the set of good patterns associated with each of the component $M_4$ and $M_5$. These figures accord with the assumptions in Pearl's example that components of the same type have the same probability of failure and that the probability of failure of multipliers ($M_1$, $M_2$, and $M_3$) is greater than the probability of failure of adders ($M_4$ and $M_5$).

\subsection{Multiple alignments in causal diagnosis}

\begin{figure}[!hbt]
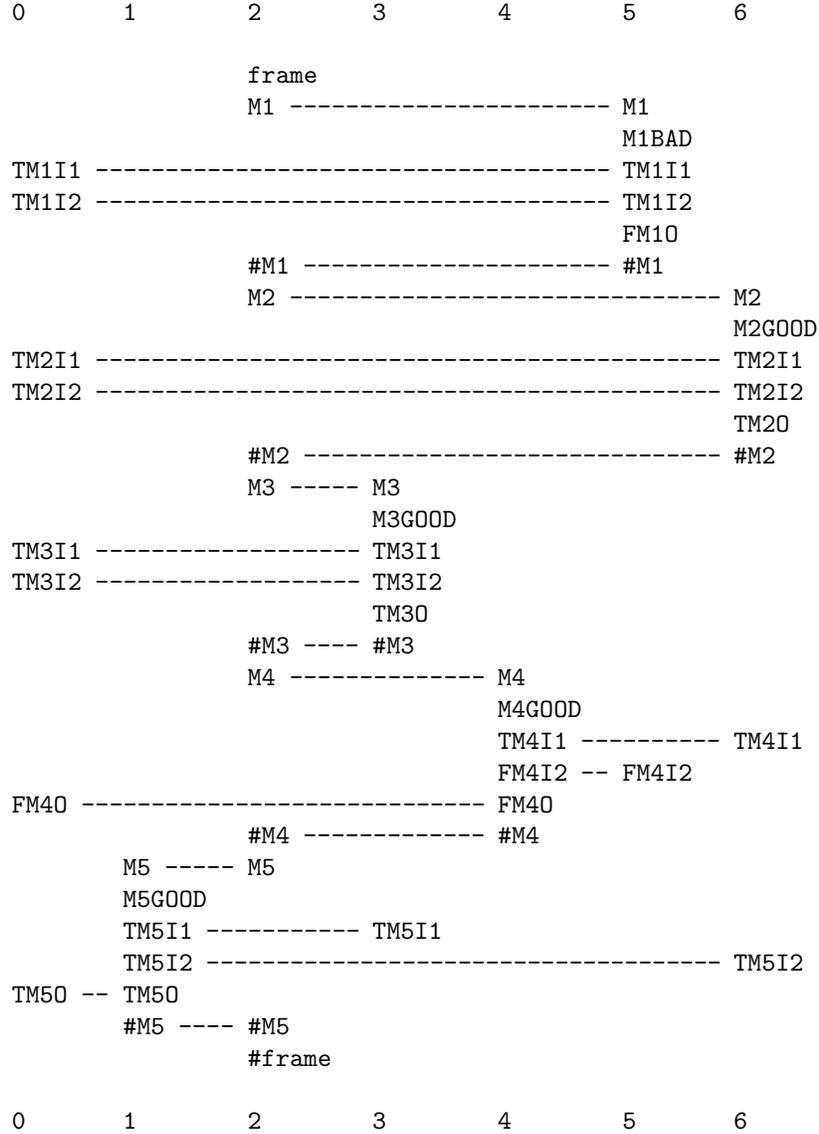

\fontsize{10.00pt}{12.00pt}
\centering
\begin{BVerbatim}
0       1        2        3        4        5       6     

                 frame                                    
                 M1 ----------------------- M1            
                                            M1BAD         
TM1I1 ------------------------------------- TM1I1         
TM1I2 ------------------------------------- TM1I2         
                                            FM1O          
                 #M1 ---------------------- #M1           
                 M2 ------------------------------- M2    
                                                    M2GOOD
TM2I1 --------------------------------------------- TM2I1 
TM2I2 --------------------------------------------- TM2I2 
                                                    TM2O  
                 #M2 ------------------------------ #M2   
                 M3 ----- M3                              
                          M3GOOD                          
TM3I1 ------------------- TM3I1                           
TM3I2 ------------------- TM3I2                           
                          TM3O                            
                 #M3 ---- #M3                             
                 M4 -------------- M4                     
                                   M4GOOD                 
                                   TM4I1 ---------- TM4I1 
                                   FM4I2 -- FM4I2         
FM4O ----------------------------- FM4O                   
                 #M4 ------------- #M4                    
        M5 ----- M5                                       
        M5GOOD                                            
        TM5I1 ----------- TM5I1                           
        TM5I2 ------------------------------------- TM5I2 
TM5O -- TM5O                                              
        #M5 ---- #M5                                      
                 #frame                                   

0       1        2        3        4        5       6     
\end{BVerbatim}
\caption{The best multiple alignment found by SP61 with `TM1I1 TM1I2 TM2I1 TM2I2 TM3I1 TM3I2 FM4O TM5O' in New and the patterns shown in Figure \ref{sp_causal_diagnosis_patterns} in Old.}
\label{sp_causal_diagnosis_alignment}
\end{figure}

Given appropriate patterns, SP61 constructs multiple alignments from which diagnoses may be obtained. Figure \ref{sp_causal_diagnosis_alignment} shows the best multiple alignment created by SP61 with the Old patterns shown in Figure \ref{sp_causal_diagnosis_patterns} and `TM1I1 TM1I2 TM2I1 TM2I2 TM3I1 TM3I2 FM4O TM5O' as the New pattern. The first six symbols in this pattern express the idea that all the inputs for components $M_1$, $M_2$ and $M_3$ are true. The penultimate symbol (`FM4O') shows that the output of $M_4$ is false and the last symbol (`TM5O') shows that the output of $M_5$ is true---in accordance with the outputs shown in Figure \ref{electronic_circuit_figure}.

From the multiple alignment in Figure \ref{sp_causal_diagnosis_alignment} it can be inferred that component $M_1$ is bad and all the other components are good. A total of seven alternative diagnoses can be derived from those multiple alignments created by SP61 that encode all the symbols in New. These diagnoses are shown in Table \ref{circuit_diagnoses_and_probabilities}, each with its relative probability.

\begin{table}
\centering
\begin{tabular}{ll}
\em Bad Component(s) & \em Relative Probability \\
\\
M1 & 0.6664 \\
M4 & 0.3332 \\
M1, M3 & 0.00013 \\
M1, M2 & 0.00013 \\
M1, M4 & 6.664e-5 \\
M3, M4 & 6.664e-5 \\
M1, M2, M3 & 2.666e-8 \\
\end{tabular}
\caption{Seven alternative diagnoses of faults in the circuit shown in Figure \ref{electronic_circuit_figure}, derived from multiple alignments created by SP61 with `TM1I1 TM1I2 TM2I1 TM2I2 TM3I1 TM3I2 FM4O TM5O' in New and the patterns from Figure \ref{sp_causal_diagnosis_patterns} in Old. The relative probability of each diagnosis is shown in the second column.}
\label{circuit_diagnoses_and_probabilities}
\end{table}

It is interesting to see that the best diagnosis derived by SP61 ($M_1$ is bad) and the second best diagnosis ($M_4$ is bad) are in accordance with first two diagnoses obtained by Pearl's method. The remaining five diagnoses derived by SP61 are different from the one obtained by Pearl's method ($M_2$ and $M_5$ are bad) but this is not altogether surprising because detailed figures are different from Pearl's example and there are differences in assumptions that have been made.%
\index{diagnosis!causal|)}\index{diagnosis!fault finding|)}\index{reasoning!causal|)}

\section{Reasoning which is not supported by evidence}

In Section \ref{reasoning_and_inference_section}, `reasoning', including `probabilistic reasoning', was characterised as a process of ``going beyond the information given''. All the examples of reasoning in the SP framework that we have considered thus far have exhibited this feature in the form of symbols and sequences of symbols from Old that are not matched to anything in New.

What happens if there is little or no information in New or if the process of `reasoning' goes 
so far beyond the information in New that alternative lines of reasoning lose their support? 
Something like this seems to happen in ordinary thinking when, in considering alternative 
scenarios in the future, we think and sometimes worry about possibilities which are very unlikely 
to occur (winning the lottery and what we might do with the `loot', falling under the proverbial bus, etc). Hence the patronising advice, ``Don't worry, it may never happen!'' or, a little more helpfully, ``Let's climb that mountain [or `cross that bridge'] when we get to it''---the problem may never arise.

\begin{figure}[!hbt]
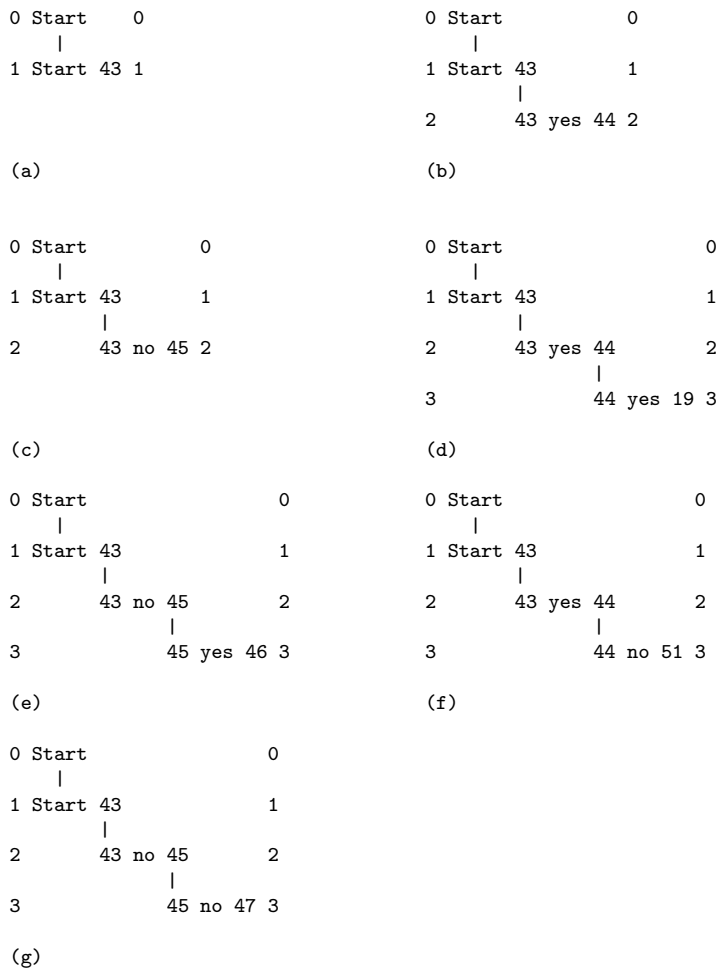

\fontsize{08.00pt}{09.60pt}
\centering
\begin{BVerbatim}
0 Start    0                         0 Start           0
    |                                    |   
1 Start 43 1                         1 Start 43        1      
                                             | 
                                     2       43 yes 44 2

(a)                                  (b)

0 Start          0                   0 Start                  0
    |                                    | 
1 Start 43       1                   1 Start 43               1
        |                                    | 
2       43 no 45 2                   2       43 yes 44        2
                                                    | 
                                     3              44 yes 19 3

(c)                                  (d)

0 Start                 0            0 Start                 0
    |                                    |
1 Start 43              1            1 Start 43              1
        |                                    |
2       43 no 45        2            2       43 yes 44       2
              |                                     |
3             45 yes 46 3            3              44 no 51 3

(e)                                  (f)

0 Start                0
    |                 
1 Start 43             1
        |             
2       43 no 45       2
              |       
3             45 no 47 3

(g)
\end{BVerbatim}
\caption{Alignments formed by SP61 with the patterns from Figure \ref{decision_tree_rules} in Old and only the symbol `Start' in New. SP61 was set to stop searching when two `unsupported' inferential steps had been made. Without this check, many more multiple alignments would be formed.}
\label{start_alignments}
\end{figure}

Figure \ref{start_alignments} shows the first seven multiple alignments formed by SP61 with the patterns from Figure \ref{decision_tree_rules} in Old and only the symbol `Start' in New. If it is allowed to proceed without any check that the multiple alignments formed are actually or potentially useful, the program would carry on creating multiple alignments corresponding to the large number of possible multiple alignments which are implicit in the patterns in Old.

Out of all these many multiple alignments, the first one (at the top of Figure \ref{start_alignments}) has the highest $CD$ because it encodes New completely and contains the fewest patterns from Old. Amongst the other multiple alignments, $CD$ values decrease as the number of patterns from Old in each multiple alignment increases.

\subsection{Escaping from `local peaks' in the search space}

Is there any use for this kind of `reasoning' without supporting evidence? In many cases, 
``no'', but in any kind of problem where there are `local peaks' in the search space (if we regard heuristic search as a form of `hill climbing'), the program must be able to explore regions of the search space which are sub-optimal from a local perspective but which may provide the means of escaping from a local peak and finding a result elsewhere which is better from a broader perspective (a higher `peak').

The example described in Section \ref{information_which_is_incomplete} (Figure \ref{decision_tree_alignment_2}) illustrates this kind of reasoning. Compared with in Figure \ref{decision_tree_alignment_1}, the example contains a gap in New because the substring `yes no 
no no' is missing. Working left-to-right in sequential mode, SP61 is able to bridge this gap in 
New because, after it has found a good multiple alignment for the first part of New, it is able to continue forming relatively poor multiple alignments until it finds one which bridges the gap and allows the right-hand section of New can be included. The multiple alignment shown in Figure \ref{decision_tree_alignment_2} has a higher $CD$ than any of the earlier multiple alignments, including the best of the multiple alignments for the left-hand section of New.

\section{Conclusion}

In this chapter we have seen how probabilistic inferences can be drawn from any multiple alignment which contains one or more Old symbols that are not matched to any symbol in New. Probabilities can be derived using the method described in Section \ref{probabilities_section}. The versatility of the SP system has been seen in examples showing how the system can model one-step `deductive' reasoning, abductive reasoning, reasoning with probabilistic networks and trees, reasoning with `rules', and nonmonotonic reasoning with default values. The system provides an alternative to causal networks in modelling the phenomenon of `explaining away' and in the diagnosis of faults in electronic circuits and similar systems. The system can also model kinds of conceptual exploration that are not constrained by empirical evidence.

It is in the nature of the SP system that it blurs many distinctions that are prevalent in computing and artificial intelligence. In particular, there is no clear boundary in the SP scheme between fuzzy pattern recognition, information retrieval and probabilistic reasoning. As we see in chapters that follow, much the same can be said about the boundary between this recognition-retrieval-and-reasoning amalgam and other areas of artificial intelligence.%
\index{reasoning!probabilistic|)}

%% file: pps.tex
\chapter{Planning and Problem Solving}\label{pps_chapter}

\index{planning|(}\index{problem solving|(}

\section{Introduction}

Although `planning' and `problem solving' are often linked in artificial intelligence, it is not entirely obvious that they are any more closely related than several other pairs of artificial intelligence topics. Nevertheless, the tradition will be maintained in this chapter with two relatively brief sections showing how the SP system may be applied in these two areas. As in other chapters, the aim is to show the broad scope of the SP theory rather than demonstrating the bells and whistles of a completed product.

\section{Planning}

The example that we shall consider in this section is how to work out a flying route from one city to another using information about flights that are available between pairs of cities. It is generally accepted (at least in artificial intelligence) that this kind of operation is a form of `planning' but this is no guarantee that methods for solving this kind of problem will transfer without modification to other kinds of `planning' such as deciding what to do on holiday or planning a major construction project.

In its essentials, an available flight between two cities may be represented by a pattern containing the names of the two cities. Thus, in our first example, an available flight from Delhi to Cape Town is represented with the pattern `Delhi 15a Cape\_Town' and a flight from Cape Town to Paris is represented with the pattern `Cape\_Town 14 Paris'. The symbol in the middle of each pattern is an ID-symbol needed for the purpose of scoring multiple alignments. No attempt has been made in this example to represent other information about flights such as times, distances, costs, and so on (see Section \ref{times_distances_costs}, below).

\begin{figure}[!hbt]
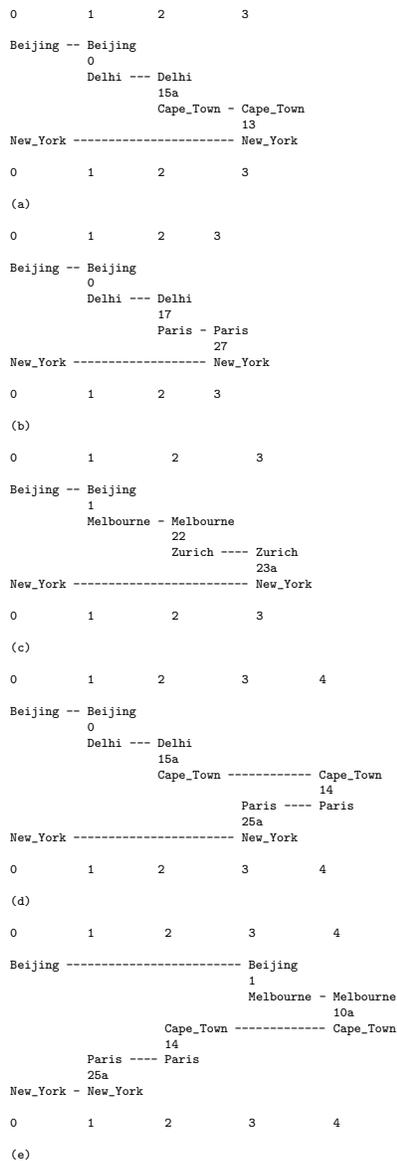

\fontsize{05.00pt}{06.00pt}
\centering
\begin{BVerbatim}
0          1         2           3        

Beijing -- Beijing                        
           0                              
           Delhi --- Delhi                
                     15a                  
                     Cape_Town - Cape_Town
                                 13       
New_York ----------------------- New_York 

0          1         2           3        

(a)

0          1         2       3       

Beijing -- Beijing                   
           0                         
           Delhi --- Delhi           
                     17              
                     Paris - Paris   
                             27      
New_York ------------------- New_York

0          1         2       3       

(b)

0          1           2           3       

Beijing -- Beijing                         
           1                               
           Melbourne - Melbourne           
                       22                  
                       Zurich ---- Zurich  
                                   23a     
New_York ------------------------- New_York

0          1           2           3       

(c)

0          1         2           3          4        

Beijing -- Beijing                                   
           0                                         
           Delhi --- Delhi                           
                     15a                             
                     Cape_Town ------------ Cape_Town
                                            14       
                                 Paris ---- Paris    
                                 25a                 
New_York ----------------------- New_York            

0          1         2           3          4        

(d)

0          1          2           3           4        

Beijing ------------------------- Beijing              
                                  1                    
                                  Melbourne - Melbourne
                                              10a      
                      Cape_Town ------------- Cape_Town
                      14                               
           Paris ---- Paris                            
           25a                                         
New_York - New_York                                    

0          1          2           3           4        

(e)
\end{BVerbatim}
\caption{A selection of routes between Beijing and New York found by SP61 with `Beijing New\_York' in New and patterns showing one-way air links between individual cities as described in the text.}
\label{planning_figure_1}
\end{figure}

SP61 may be used to find one or more routes between two cities if Old contains the set of patterns representing available flights and New contains a pattern that specifies the cities at the beginning and end of the journey. Figure \ref{planning_figure_1} shows a selection of five of the routes found by SP61 with `Beijing New\_York' in New and a set of patterns about available flights in Old. The program also forms multiple alignments that do not represent sensible routes between Beijing and New York. All these other multiple alignments have lower $CD$ values than the ones that represent plausible routes, as discussed below.

\subsection{Evaluation of alignments}

In the example just described, all three symbols in each pattern has the status of ID-symbol. This means that in a multiple alignment like this:

\begin{center}
\begin{BVerbatim}
0          1         2       3          4       

Beijing -- Beijing                              
           0                                    
           Delhi --- Delhi                      
                     17                         
                     Paris - Paris              
                             27                 
New_York ------------------- New_York - New_York
                                        25      
                                        Paris   

0          1         2       3          4       
\end{BVerbatim}
\end{center}

\noindent the unmatched symbol `Paris' has the effect of increasing the value of $B_E$ for the multiple alignment and thus reducing the value of $CD$ (Section \ref{ma_evaluation}). This means that multiple alignments of this kind, that do not represent sensible routes between Beijing and New York, have lower $CD$s than those in which all the names of cities in Old patterns have been matched.

In a multiple alignment like this:

\begin{center}
\begin{BVerbatim}
0          1          2           3        

Beijing                           Bombay   
                                  3        
                      Cape_Town - Cape_Town
                      14                   
           Paris ---- Paris                
           25a                             
New_York - New_York                        

0          1          2           3        
\end{BVerbatim}
\end{center}

\noindent the unmatched symbol `Beijing' in the New pattern lowers the value of $B_N$ for the multiple alignment and this lowers the value of $CD$. So, in general, multiple alignments that do not connect the start and end cities of the route have lower $CD$ than those that do.

As we have noted already, the second symbol in each Old pattern is also an ID-symbol. Since these symbols are never matched to any other symbol, they always add to $B_E$ for each multiple alignment and lower the corresponding $CD$ value. This means that multiple alignments with many columns (representing routes with many links) have lower $CD$s than those with few columns (representing routes with few links). Thus, for example, each of the first three multiple alignments in Figure \ref{planning_figure_1} have $CD = 140.18$, while each of the last two multiple alignments have $CD = 133.76$.

In general, when SP61 has been supplied with patterns like those that have been described, the best multiple alignments created by the program are those that represent coherent routes that link the start and end cities. And, in general, multiple alignments representing routes with few connections have higher $CD$s than those representing routes with many connections.

\subsection{One-way and two-way connections}

In the scheme that has been described, each pattern represents a one-way connection between two cities with the left-to-right order of the cities corresponding to the direction of travel. This means that, for every pair of cities with two-way connections, it is necessary to include two patterns in the flights database: one from A to B and the other from B to A.

If there are many cases where it is possible to fly from A to B but not from B to A, then this scheme makes good sense. But if all the connections are two-way, or even the majority, then there is redundancy. This would be reduced if each two-way connection could be represented by a single pattern.

\begin{figure}[!hbt]
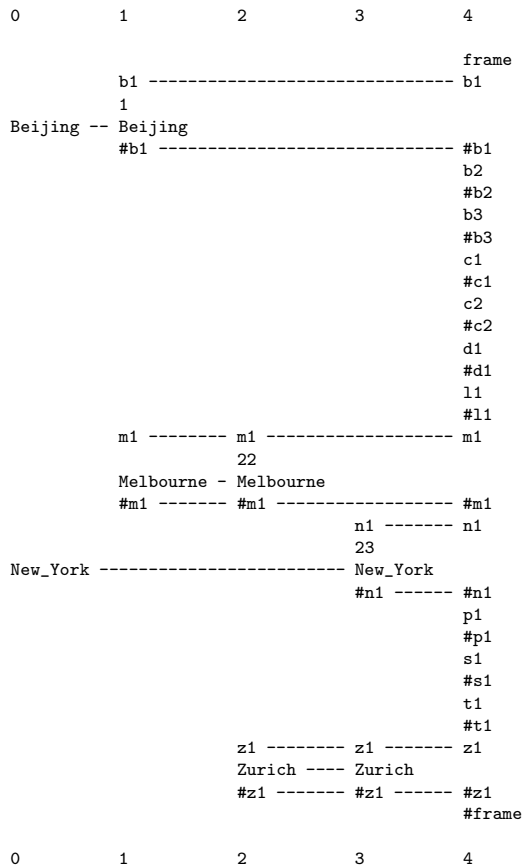

\fontsize{07.00pt}{08.40pt}
\centering
\begin{BVerbatim}
0          1           2           3          4     

                                              frame 
           b1 ------------------------------- b1    
           1                                        
Beijing -- Beijing                                  
           #b1 ------------------------------ #b1   
                                              b2    
                                              #b2   
                                              b3    
                                              #b3   
                                              c1    
                                              #c1   
                                              c2    
                                              #c2   
                                              d1    
                                              #d1   
                                              l1    
                                              #l1   
           m1 -------- m1 ------------------- m1    
                       22                           
           Melbourne - Melbourne                    
           #m1 ------- #m1 ------------------ #m1   
                                   n1 ------- n1    
                                   23               
New_York ------------------------- New_York         
                                   #n1 ------ #n1   
                                              p1    
                                              #p1   
                                              s1    
                                              #s1   
                                              t1    
                                              #t1   
                       z1 -------- z1 ------- z1    
                       Zurich ---- Zurich           
                       #z1 ------- #z1 ------ #z1   
                                              #frame

0          1           2           3          4     
\end{BVerbatim}
\caption{An example of a route between Beijing and New York found by SP61 with `Beijing New\_York' in New and, in Old, a framework pattern and patterns showing two-way air links between individual cities as described in the text.}
\label{planning_figure_2}
\end{figure}

If we are to adopt this idea then, for much the same reasons as were described in Section \ref{ordering_of_symptoms}, it is necessary with SP61 to provide a framework pattern in addition to the patterns representing connections between cities and to adapt those patterns so that they can be aligned with the framework pattern. Figure \ref{planning_figure_2} shows an example of a multiple alignment formed with patterns that have been adapted in this way. From this multiple alignment we can see that a possible route between Beijing and New York is via Melbourne and Zurich, in that order. Notice that this ordering of cities in the route can be derived unambiguously from the multiple alignment despite the fact that the order of the symbols in the multiple alignment is different.

\subsection{Times, distances and costs}\label{times_distances_costs}

There is, of course, a lot more to any practical system for route planning than what has been described. Normally, there would be a need to take account of departure and arrival times, the duration of each flight and intervals between flights, distances travelled and the costs of flights. This subsection offers a few remarks about whether or how these aspects of route planning may be accommodated in the SP system.

In conventional systems, information about departure and arrival times, time intervals, distances and costs would all be expressed as numbers. Although the SP system can, in principle, support the representation and use of numbers, current models are weak in this area (Section \ref{maths_and_sp}). One can either attempt to work around this weakness or wait until the system has been more fully developed for mathematics (Section \ref{sp_machine_other_developments}).

In planning a flying route, it is of course necessary to ensure that the time of arrival for one leg of the journey comes before the time of departure for the next leg, and that the time between arrival and departure is adequate to allow transfers. These things mean comparisons such as `greater than' or `less than' and such comparisons seem to demand proper definitions of numbers and the operations for comparing them.

If one is wishing to minimise the overall flying time, or the distance flown, or the cost of the journey, there is a possible work around that takes advantage of the fact that the SP system is designed to look for multiple alignments that allow the New pattern to be encoded with as little redundancy as possible. For each pattern like `Delhi 15a Cape\_Town', the second symbol may be assigned a number of bits that represents the duration or cost of the flight or the distance flown. Then multiple alignments delivered by the system would represent routes that tend to minimise whichever of those measures has been chosen. No account would be taken of waiting times between flights.

This work around is really nothing but a fudge, without any theoretical foundation. In terms of practicalities, it is probably better to develop the system for mathematical operations.

\section{Solving geometric analogy problems}\label{geometric_analogy_problems}

\index{analogy|(}

\begin{figure}[!hbt]
\centering
\includegraphics[width=0.9\textwidth]{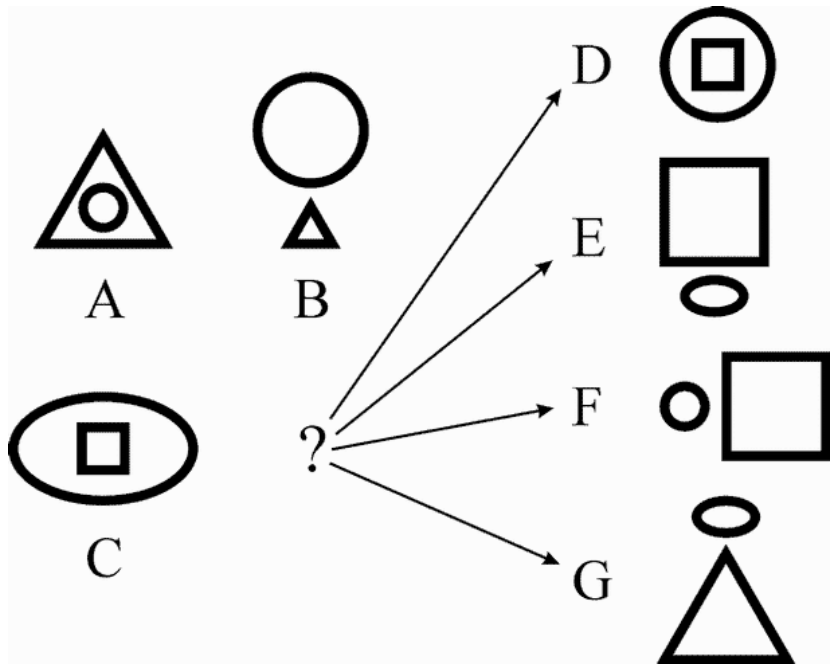}
\caption{A geometric analogy problem.}
\label{geometric_analogy_figure}
\end{figure}

Figure \ref{geometric_analogy_figure} shows an example of a well-known type of simple puzzle---a geometric analogy problem. The task is to complete the relationship ``A is to B as C is to ?'' using one of the figures `D', `E', `F' or `G' in the position marked with `?'. For this example, the `correct' answer is clearly `E'. Quote marks have been used for the word `correct' because in many problems of this type, there may be two or even more alternative answers for which cases can be made and there is a corresponding uncertainty about which answer is the right one. 

Computer-based methods for solving this kind of problem have existed for some time (e.g., Evans' \citeyearpar{evans_1968} well-known heuristic algorithm). In more recent work \citep{belloti_gammerman_1996, gammerman_1991}, minimum length encoding principles have been applied to good effect. The proposal here is that, within the general framework of minimum length encoding, this kind of problem may be understood in terms of the SP concepts.

As in some previous work \citep{belloti_gammerman_1996, gammerman_1991}, the proposed solution assumes that some mechanism is available which can translate the geometric forms in each problem into patterns of alpha-numeric symbols like the patterns in other examples in this article. For example, item `A' in Figure \ref{geometric_analogy_figure} may be described as `small circle inside large triangle'.

How this kind of translation may be done is not part of the present proposals (one such translation mechanism is described in \citet{evans_1968}). As noted elsewhere \citep{gammerman_1991}, successful solutions for this kind of problem require consistency in the way the translation is done. For this example, it would be unhelpful if item `A' in Figure \ref{geometric_analogy_figure} were described as `large triangle outside small circle' while item `C' were described as `small square inside large ellipse'. For any one puzzle, the description needs to stick to one or other of ``X outside Y'' or ``Y inside X''---and likewise for `above/below' and `left-of/right-of'.

Given that the diagrammatic form of the problem has been translated into patterns as just described, this kind of problem can be cast as a problem of partial matching, well within the scope of SP61. To do this, symbolic representations of item A and item B in Figure \ref{geometric_analogy_figure} are treated as a single pattern, thus:

\begin{center}
\begin{BVerbatim}
small circle inside large triangle ;
     large circle above small triangle
\end{BVerbatim}
\end{center}

\noindent and this pattern is placed in New. Four other patterns are constructed by pairing a symbolic representation of item C (on the left) with symbolic representations of each of D, E, F and G (on the right), thus:

\begin{center}
\begin{BVerbatim}
C1 small square inside large ellipse ;
     D small square inside large circle #C1
C2 small square inside large ellipse ;
     E large square above small ellipse #C2
C3 small square inside large ellipse ;
     F small circle left-of large square #C3
C4 small square inside large ellipse ;
     G small ellipse above large rectangle #C4.
\end{BVerbatim}
\end{center}

\noindent These four patterns are placed in Old, each with an arbitrary frequency value of 1.

Figure \ref{geometric_analogy_alignment} shows the best multiple alignment found by SP61 with New and Old as just described. The multiple alignment is a partial match between the pattern in New and the second of the four patterns in Old. This corresponds with the `correct' result (item E) as noted above.

\begin{figure}[!hbt]
\centering
\begin{BVerbatim}
           C2     
small ---- small  
circle     square 
inside --- inside 
large ---- large  
triangle   ellipse
; -------- ;     
           E      
large ---- large  
circle     square 
above ---- above  
small ---- small  
triangle   ellipse
           #C2
\end{BVerbatim}
\caption{The best multiple alignment found by SP61 for the patterns in New and Old as described in the text.}
\label{geometric_analogy_alignment}
\end{figure}

\index{analogy|)}

\section{Conclusion}

This chapter has given a brief description of the ways in which the SP system might support one kind of planning and one kind of problem solving. There are many other kinds of planning and problem solving that have not been examined and plenty of scope for further development of the applications that have been described.%
\index{planning|)}\index{problem solving|)}

%% file: learning.tex
\chapter{Unsupervised Learning}\label{learning_chapter}

\section{Introduction}

\index{learning|(}

As described in Chapter \ref{introduction_chapter}, this entire programme of research grew out of an earlier programme of research on grammar induction and language learning \citep{wolff_1988, wolff_1982, wolff_1980, wolff_1977, wolff_1975}. Until relatively recently, the focus has been on the development of a framework that could integrate aspects of artificial intelligence other than learning, bearing in mind that learning should eventually form part of the picture. Now, in the SP70 model, this goal has been realised, bring the research full circle to learning again but with a new and much broader perspective than the earlier research, focussed as it was specifically on learning.

Given the new broad perspective, the goal has been to develop the SP system as a generalised system for the unsupervised learning of {\em any} kind of knowledge, in keeping with the working hypothesis that any kind of knowledge may be represented in the SP system as {\em patterns} of symbols in one or more dimensions (Section \ref{representation_of_knowledge}). That said, the examples in this chapter have a `linguistic' flavour and the automatic learning of realistic grammars for natural languages provides a challenging goal for this phase of the research and a touchstone of success.

\subsection{Relationship to other research on unsupervised learning}

In terms of broad categories of research in machine learning, development of the SP system for learning is more closely related to research on grammar induction using PSGs (augmented or otherwise) than it is to research on the learning of finite-state grammars, n-grams or Markov models---which are known to be less adequate for representing the structure of natural languages \citep{chomsky_1957}. Here, learning is conceived as a process of optimisation using minimum length encoding principles, by contrast with research on language learning conceived as ``language identification in the limit'' \citep{gold_1967} (Section \ref{optimisation_and_learnability}). The research is closer in spirit to other research on unsupervised learning than it is to research on learning with external error correction or negative samples or the grading of language samples from simple to complex \citep[{cf.}][]{gold_1967}. And there is a stronger affinity with systems that are designed to learn explicit symbolic structures that are readable and comprehensible than with systems---like most artificial neural networks---that learn knowledge in an implicit form.

Compared with other work on unsupervised learning of grammar-like structures, the most distinctive features of this approach to machine learning are:

\begin{itemize}

\item The integration of learning with other areas of artificial intelligence and computation.

\item The multiple alignment concept as it has been developed in the SP system.

\end{itemize}

With respect to the second point, there is an affinity with recent work by \citet{van_zaanen_thesis_2002, van_zaanen_2002, solan_etal_2002} in which learning is based on the alignment of pairs of patterns. Otherwise, studies that are, perhaps, most closely related to the present research include: \citet{adriaans_et_al_2000, allison_wallace_yee_1992, clark_2001, denis_2001, henrichsen_2002, johnson_reizler_2002, klein_manning_2001, nevill-manning_witten_1997, oliveira_sv_1996, rapp_et_al_1994, watkinson_manandhar_2001}. 

\section{SP70}\label{sp70_section}

\index{SP70|(}

All the main components of the SP system outlined in Section \ref{overall_framework} are now realised within the SP70 software model (version 9.2).

The model gives results in the area of unsupervised learning that are good enough to show that the framework is sound. As we shall see, the model is able to abstract plausible grammars\footnote{In accordance with the remarks in Section \ref{framework_examples_section}, the term {\em grammar} will be used for sets of SP patterns that function like a grammar cast in more conventional form.} from sets of simple sentences without prior knowledge of word segments or the classes to which they belong (Section \ref{learning_examples_section}) and the computational complexity of the model appears to be acceptable (Section \ref{sp70_computational_complexity}).

However, in its current form, the model has at least two significant shortcomings and some other deficiencies. A programme of further development, experimentation and refinement is still needed to realise the full potential of the model for unsupervised learning. 

This model is governed by mathematical principles---explained at pertinent points below---but these are remarkably simple and in accordance with established theory. The main focus in what follows is on the organisation of the model and the computational techniques employed within it.

\subsection{Objectives}

The main problems addressed in the development of this model have been:

\begin{itemize}

\item How to identify significant segments in the `corpus' of data which is the basis of learning, given that the boundary between one segment and the next is not marked explicitly.

\item How to identify disjunctive classes of syntactically-equivalent segments (e.g., parts of speech such as `noun', `verb' or `adjective').

\item How to combine the learning of segmental structure with the learning of disjunctive classes.

\item How to achieve the learning of segments and classes through two or more levels of abstraction. In a linguistic context, this means how to learn abstract structures representing phrases, clauses and sentences---and classes of these entities---as well as lower-level segments such as words.

\item \sloppy How to generalise grammatical rules beyond the data and how to correct over-generalisations without feedback from a `teacher' or the provision of `negative' samples or the grading of the data from `easy' to `hard'.\footnote{In the learning framework considered by \citet{gold_1967}, learning is not possible without some kind of feedback or negative samples or grading of samples. The minimum length encoding framework circumvents the need for these sources of information but it is still necessary to consider exactly how the generalisation of rules and the correction of over-generalisations is to be achieved.}

\end{itemize}

Solutions to these problems were found in the SNPR model \citep{wolff_1988, wolff_1982} and in the MK10 model \citep{wolff_1980, wolff_1977, wolff_1975} but, as noted earlier, the organisation of these models is quite unsuited to the wider goals of the present research: integration of diverse functions within one framework. Finding new solutions to these problems within the SP system has been a significant challenge.

The SP70 model (v.~9.2) provides solutions to the first three problems and partial solutions to the fourth and fifth problems. Further development will be required to achieve robust learning of structures with more than two levels of abstraction and more work is required on the generalisation of grammatical rules and the correction of overgeneralisations (see Sections \ref{grammatical_inference_and_generalisation} and \ref{learning_discussion}).

\subsection{Overall structure of the model}

Figure \ref{sp70_figure} shows the high-level organisation of the SP70 model. The program starts with a set of New patterns and with a repository of Old patterns that is initially empty.

In broad terms, the model comprises two main phases:

\begin{itemize}

\item Create a set of patterns that may be used to encode the patterns from New in an economical manner (operations 1 to 3 in Figure \ref{sp70_figure}).

\item From the patterns created in the first phase, compile one or more alternative grammars for the patterns in New in accordance with minimum length encoding principles (operation 4 in Figure \ref{sp70_figure}).

\end{itemize}

After reading a set of patterns into New, the system compiles an `alphabet' of the different types of symbols appearing in those patterns, counts their frequencies of occurrence and calculates encoding costs as described below. These values will be needed for the evaluation of multiple alignments. As was noted in Section \ref{terms_section}, the term {\em symbol type} is used to mean a representative template or example of a set of identical symbols.

\begin{figure}[!hbt]
\fontsize{09.00pt}{10.80pt}
\centering
\begin{BVerbatim}
SP70()
{
     1 Read a set of patterns into New. Old is initially empty.
     2 Compile an alphabet of symbol types in New and, for each type,
          find its frequency of occurrence and the number of bits
          required to encode it.
     3 While (there are unprocessed patterns in New)
     {
          3.1 Identify the first or next pattern from New as the
               `current pattern from New'.
          3.2 Apply the function CREATE_MULTIPLE_ALIGNMENTS() to
               create multiple alignments, each one between the
               current pattern from New and one or more patterns from Old.
          3.3 During 3.2, the current pattern from New is copied into Old,
               one symbol at a time, in such a way that the current pattern
               from New can be aligned with its copy but that any one
               symbol in the current pattern from New cannot be aligned
               with the corresponding symbol in the copy.
          3.4 Sort the multiple alignments formed by this function in order
               of their compression scores and select the best
               few for further processing.
          3.5 Process the selected multiple alignments with the function
               DERIVE_PATTERNS(). This function derives encoded
               patterns from multiple alignments and adds them to Old.
     }

     4 Apply the function SIFTING_AND_SORTING() to create one or
          more alternative grammars for the patterns in New, each
          one scored in terms of minimum length encoding principles.
          Each grammar is a subset of the patterns in Old.
}
\end{BVerbatim}
\caption{The organisation of SP70. The workings of the functions {\em create\_multiple\_alignments()}, {\em derive\_patterns()} and {\em sifting\_and\_sorting()} are explained in Sections \ref{create_multiple_alignments_section}, \ref{derive_patterns_section} and \ref{sifting_and_sorting_section}, respectively.}
\label{sp70_figure}
\end{figure}

Next (operation 3), each pattern from New is processed by searching for multiple alignments that allow the given pattern to be encoded economically in terms of patterns in Old (as outlined in Section \ref{framework_examples_section}). From a selection of the best multiple alignments found, the program `learns' new patterns, as explained below, and adds them to Old. A copy of each pattern from New is also added to Old, marked with new ID-symbols as will be explained.

When all the patterns from New have been processed in this way, there is a process of sifting and sorting to create one or more alternative grammars for the patterns from New (operation 4). Each grammar comprises a subset of the patterns in Old and each one is scored in terms of minimum length encoding principles.

\subsection{Creating multiple alignments}\label{create_multiple_alignments_section}

The function {\em create\_multiple\_alignments()} referred to in Figure \ref{sp70_figure} creates zero or more multiple alignments, each one comprising the current pattern from New and one or more patterns from Old. Apart from some minor modifications and improvements, this function is essentially the same as the main component of the SP61 model, described in Sections \ref{sp61_outline} and \ref{sp61_detail_section}.

\subsubsection{Calculation of compression scores}\label{calculation_of_compression_scores}

The compression score for an multiple alignment is calculated as described in Section \ref{ma_evaluation}, except for an approximation that is needed because the contents of Old changes.

As learning proceeds, patterns are added to Old (operation 3.5 in Figure \ref{sp70_figure}). Within those patterns, some of the symbols are derived from New and their encoding costs are already known. However, other symbols---ID-symbols and copies of them---are created by the system in operation 3.5 (Figure \ref{sp70_figure}) and the variety of types of these symbols and their frequencies of occurrence are constantly changing. For this reason, it is difficult at this stage to calculate encoding costs using the Shannon-Fano-Elias method. Accordingly, the encoding costs of symbols created by the system are initially set at a fixed arbitrary value. As we shall see, more precise values are calculated in the {\em sifting\_and\_sorting()} phase of processing. The approximation at this stage does not seem to be a serious impediment to learning, perhaps because the selection of multiple alignments depends on relative values for compression scores, not absolute values.

\subsubsection{Minimum length encoding principles in the formation of multiple alignments}

\index{minimum length encoding|(}

The search for multiple alignments that maximise $CD$ conforms with minimum length encoding principles as described in Section \ref{mle_section}. The repository of Old patterns may be taken to be the current grammar for encoding the current pattern from New and, for any given pattern from New, the size of this grammar, $G$, is constant.\footnote{As indicated in operation 3.3 of Figure \ref{sp70_figure} and described in Section \ref{copying_cpfn}, below, a copy of the current pattern from New is added to Old during the process of building multiple alignments. However, for the purpose of encoding the current pattern from New, the entire copy (with ID-symbols that are added to the copy) counts as part of Old and thus, for any given pattern from New, $G$ is indeed constant.} Since $G$ is constant, the goal of minimising $T$ is equivalent to a goal of minimising $E$. For any given pattern from New, $E$ is the same as $B_E$ (Equation \ref{BE_equation}). Thus, since $B_N$ is constant for any given pattern from New, seeking to minimise $E$ is equivalent to the attempt to maximise $CD$.%
\index{minimum length encoding|)}

\subsection{Copying the current pattern from New into Old}\label{copying_cpfn}

In its bare essentials, `learning' in SP70 is achieved by the addition of patterns to Old. This occurs in two ways: by copying each pattern from New into Old (operation 3.3 in Figure \ref{sp70_figure}) and by deriving patterns from multiple alignments in the function {\em derive\_patterns()} (operation 3.5 in the same figure). The first of these is described here and the second is described in the next subsection.

\sloppy During the matching process in the first cycle of the {\em create\_multiple\_alignments()} function, the current pattern from New is copied, one symbol at a time, into Old in such a way that any symbol in the current pattern from New can be matched with any earlier symbol in the copy but it cannot be matched with the corresponding symbol in the copy or any subsequent symbol. When the transfer is complete, ID-symbols are added to the copy to provide a `code' for the pattern, as described below.

The aim here is to detect any redundancy that may exist {\em within} each pattern from New (e.g., the repetition that can be seen in the pattern `a b c d x y z a b c d') but to avoid detecting the redundancy resulting from the fact that the current pattern from New has been copied into Old. This constraint is imposed for very much the same reason as the constraint (described in Section \ref{multiple_appearances}) which prevents any one symbol within an multiple alignment being matched with itself.

The reason for copying each pattern from New into Old rather than simply moving it is that each such pattern (with its ID-symbols) is a candidate for inclusion in one or more of the best grammars selected by the {\em sifting\_and\_sorting()} function and it cannot be evaluated properly unless it is a copy of the corresponding pattern from New, not the pattern itself (see Section \ref{sifting_and_sorting_section}, below). 

The ID-symbols that are added to the copy of each pattern from New comprise left and right brackets (`$<$' and `$>$') at each end of the pattern together with symbols immediately after the left bracket that serve to identify the pattern uniquely amongst the patterns in Old. For the sake of consistency with the {\em derive\_patterns()} function (see Section \ref{derive_patterns_section}, next), two ID-symbols follow the left bracket. Thus, for example, a pattern from New like `t h a t b o y r u n s' might become `$<$ \%1 9 t h a t b o y r u n s $>$' when ID-symbols have been added.

\subsection{Deriving patterns from alignments}\label{derive_patterns_section}

In operation 3.5 in Figure \ref{sp70_figure}, the {\em derive\_patterns()} function is applied to a selection of the best multiple alignments formed and, in each case, it looks for sequences of unmatched symbols within the multiple alignment and also sequences of matched symbols.

Consider the multiple alignment shown in Figure \ref{learning_alignment_1}. From an multiple alignment like that, the function finds the unmatched sequences `g i r l' and `b o y' and, within row 1, it also finds the matched sequences `t h a t' and `r u n s'. With respect to row 1, the focus of interest is the matched and unmatched sequences of C-symbols---ID-symbols are ignored.

\begin{figure}[!hbt]
\centering
\begin{BVerbatim}
0        t h a t g i r l r u n s   0
         | | | |         | | | |  
1 < %1 9 t h a t b o y   r u n s > 1
\end{BVerbatim}
\caption{A simple multiple alignment from which other patterns may be derived.}
\label{learning_alignment_1}
\end{figure}

A copy of each of the four sequences is made, ID-symbols are added to each copy (as described in Section \ref{assigning_id_symbols}, below) and the copy is added to Old. In addition, another `abstract' pattern is made that records the sequence of matched and unmatched patterns within the multiple alignment. The result in this case is five patterns like those shown in Figure \ref{patterns_figure_1}.

\begin{figure}[!hbt]
\centering
\begin{BVerbatim}
< %7 12 t h a t >
< %9 14 b o y >
< %9 15 g i r l >
< %8 13 r u n s >
< %10 16 < %7 > < %9 > < %8 > >
\end{BVerbatim}
\caption{Patterns derived from the multiple alignment shown in Figure \ref{learning_alignment_1}.}
\label{patterns_figure_1}
\end{figure}

It should be clear that the set of patterns in Figure \ref{patterns_figure_1} is, in effect, a simple grammar for the two sentences in Figure \ref{learning_alignment_1}, with patterns representing grammatical rules in much the same style as those shown in Figure \ref{alignment_figure_1} (b). The abstract pattern `$<$ \%10 220 $<$ \%7 $>$ $<$ \%9 $>$ $<$ \%8 $>$ $>$' describes the overall structure of this kind of sentence with slots that may receive individual words at appropriate points in the pattern.

Notice how the symbol `\%9' serves to mark `b o y' and `g i r l' as alternatives in the middle of the sentence. This is a grammatical class in the tradition of distributional or structural linguistics \citep[see, for example,][]{fries_1952, harris_1951}.

With multiple alignments like this:

\begin{center}
\begin{BVerbatim}
0        t h e g r e e n a p p l e   0
         | | |           | | | | |
1 < %1 2 t h e           a p p l e > 1 
\end{BVerbatim}
\end{center}

\noindent or this:

\begin{center}
\begin{BVerbatim}
0        t h e           a p p l e   0
         | | |           | | | | |  
1 < %1 2 t h e g r e e n a p p l e > 1
\end{BVerbatim}
\end{center}

\noindent the system derives patterns very much as before except that the unmatched sequence (`g r e e n') is assigned to a class by itself, without any alternative pattern that may appear in the same context. Arguably, there should be some kind of `null' alternative to `g r e e n' in cases like this in order to capture the idea that ``the apple'' and ``the green apple'' are acceptable variants of the same phrase. This is a possible refinement of the model in the future.

Readers may wonder why the grammar shown in Figure \ref{patterns_figure_1} was not simplified to something like this:

\begin{center}
\begin{BVerbatim}
< %9 14 b o y >
< %9 15 g i r l >
< %10 16 t h a t < %9 > r u n s >
\end{BVerbatim}
\end{center}

The main reason for adopting the style shown in Figure \ref{patterns_figure_1} is that the overall organisation of the model is simpler if each newly-derived pattern is automatically referenced from the contexts or contexts in which it may appear. Another reason is that it is anticipated that, with realistically large corpora, most of the patterns that will ultimately turn out to be significant in terms of minimum length encoding principles will appear in two or more contexts and, in that case, minimum length encoding principles are likely to dictate that each pattern should be referenced from each of its contexts rather than written out redundantly in each of the two or more places where it appears.

\subsubsection{Assignment of identification symbols}\label{assigning_id_symbols}

\index{symbol!identification|(}

Apart from the terminating brackets, each pattern in Figure \ref{patterns_figure_1} has two ID-symbols:

\begin{itemize}

\item A `class' symbol (e.g., `\%7' or `\%9') that normally starts with the `\%' character. The class symbol is, in effect, a reference to the context or contexts in which the given pattern may appear. Thus, for example, the symbol `\%7' in the first pattern in Figure \ref{patterns_figure_1} shows that that pattern may appear where the matching symbol occurs in the pattern `$<$ \%10 220 $<$ \%7 $>$ $<$ \%9 $>$ $<$ \%8 $>$ $>$'. Any one pattern may belong in more than one class and should contain a class symbol for each of the classes it belongs to (see Section \ref{avoiding_duplication}).

\item A `discrimination' symbol (e.g., `12', `14') that serves to distinguish the pattern from any others that may belong in the same class. At this stage, the discrimination symbol is simply a unique identifier for the given pattern amongst the other patterns and multiple alignments created by the program.

\end{itemize}

While multiple alignments are being built and patterns are being added to Old, new class symbols and new discrimination symbols are created quite liberally. However, many of these symbols are weeded out during the {\em sifting\_and\_sorting()} phase of processing and those that remain are renamed in a tidy manner.%
\index{symbol!identification|)}

\subsubsection{Avoiding duplication}\label{avoiding_duplication}

In the course of deriving patterns from multiple alignments and adding them to Old, it can easily happen that a newly-derived pattern has the same C-symbols as one that is already in Old. For this reason, each newly-derived pattern is checked against patterns already stored in Old and it is discarded if an existing pattern is found with the same C-symbols.

Although the discarded pattern has the same C-symbols as a pre-existing pattern, it comes from a different context. So a new symbol type is created to represent that context, a copy of the symbol type is added to the pre-existing pattern, and another copy of the symbol type is added to the abstract pattern in the appropriate position. In this way, any one sequence of C-symbols may appear in a pattern containing several different class symbols, each one representing one of the contexts where the C-symbols may appear.

As the program stands, there is a one-to-one relation between contexts and classes. But it can easily happen that the set of patterns that may appear in one context is the same as the set of patterns that may appear in another. At some stage, it is intended that the program will be augmented to check for this kind of redundancy and to merge classes that turn out to be equivalent. 

\subsubsection{Deriving patterns from alignments containing three or more rows}\label{deriving_from_three_plus_rows}

For any given multiple alignment, the {\em derive\_patterns()} function works by looking for one or more unmatched sequences of symbols in the current pattern from New, or one or more sequences of unmatched C-symbols in a pattern from Old, or both these things. What happens if an multiple alignment contains two or more patterns from Old?

Consider the multiple alignment shown in Figure \ref{learning_alignment_2}. In a case like this, it is necessary to identify {\em one} of the patterns from Old for the purpose of deriving patterns from the multiple alignment. The pattern that is chosen is the one that is deemed to be the most `abstract' pattern amongst those in the rows below the top row.

\begin{figure}[!hbt]
\fontsize{08.00pt}{09.60pt}
\centering
\begin{BVerbatim}
0               t h e   r e d         a p p l e          f a l l s     0
                | | |                 | | | | |          | | | | |    
1               | | |                 | | | | |   < %4 5 f a l l s >   1
                | | |                 | | | | |   | |              |  
2 < %5 7 < %1   | | | > < %2 > < %3   | | | | | > < %4             > > 2
         | |    | | | |        | |    | | | | | |                     
3        | |    | | | |        < %3 3 a p p l e >                      3
         | |    | | | |                                               
4        < %1 0 t h e >                                                4
\end{BVerbatim}
\caption{An multiple alignment (between a pattern from New and four patterns from Old) from which other patterns may be derived.}
\label{learning_alignment_2}
\end{figure}

The most abstract row in any multiple alignment is the row below the top row that starts furthest to the left within the multiple alignment, e.g., row 2 in Figure \ref{learning_alignment_2}. Typically, this is also the row that finishes furthest to the right. In general, this is the row within any multiple alignment that, directly or indirectly, encodes the largest number of symbols from the current pattern from New. Of course, if an multiple alignment contains only two rows, then row 1 is the most abstract row.

To be a suitable candidate for processing by the {\em derive\_patterns()} function, the only row below the top row that may contain unmatched C-symbols is the most abstract row. Also, there must be at least one unmatched C-symbol somewhere within the current pattern from New and the most abstract row. Any multiple alignment that does not meet these conditions is discarded for the purpose of deriving new patterns.

From the multiple alignment shown in Figure \ref{learning_alignment_2}, the function creates patterns like those shown in Figure \ref{patterns_figure_2} and adds them to Old. 

\begin{figure}[!hbt]
\centering
\begin{BVerbatim}
< %2 9 r e d >
< %7 10 < %3 > < %4 > >
< %8 11 < %1 > < %2 > < %7 > >
\end{BVerbatim}
\caption{A set of patterns derived from the multiple alignment shown in Figure \ref{learning_alignment_2}.}
\label{patterns_figure_2}
\end{figure}

Here, the unmatched sequence `r e d' from the current pattern from New has been converted into the pattern `$<$ \%2 9 r e d $>$'. The system recognises that `r e d' lies opposite the sequence `$<$ \%2 $>$' within row 2 of Figure \ref{learning_alignment_2} and that this sequence is a reference to the class `\%2'. Accordingly, the pattern `r e d' has been assigned to that class. If the unmatched sequence opposite `r e d' could not be recognised as a reference to a class, or if there was no unmatched sequence opposite `r e d', then the system would create a new class and new patterns in the same manner as we saw in the examples at the beginning of Section \ref{derive_patterns_section}.

The second pattern in Figure \ref{patterns_figure_2} (`$<$ \%7 10 $<$ \%3 $>$ $<$ \%4 $>$ $>$') is derived from the sequence `$<$ \%3 $>$ $<$ \%4 $>$' within the abstract pattern in row 2 of Figure \ref{learning_alignment_2}. The third pattern (`$<$ \%8 11 $<$ \%1 $>$ $<$ \%2 $>$ $<$ \%7 $>$ $>$') is a new version of that abstract pattern that references the class of the second pattern (`\%7').

\subsubsection{Redundancy in Old}

Given that the system is dedicated to information compression, it may seem strange that, at this stage of processing, there may be considerable replication of information (redundancy) amongst the patterns in Old. Patterns are added to Old but, at this stage, nothing is removed from Old. In the example just considered, the pattern `$<$ \%7 10 $<$ \%3 $>$ $<$ \%4 $>$ $>$' coexists with the pattern `$<$ \%5 7 $<$ \%1 $>$ $<$ \%2 $>$ $<$ \%3 $>$ $<$ \%4 $>$ $>$' even though they both contain the sequence `$<$ \%3 $>$ $<$ \%4 $>$'. In the example from Figures \ref{learning_alignment_1} and \ref{patterns_figure_1}, the patterns `$<$ \%7 12 t h a t $>$', `$<$ \%9 14 b o y $>$' and `$<$ \%8 13 r u n s $>$ coexist with the pattern `$<$ \%1 9 t h a t b o y   r u n s $>$' despite the obvious duplication of information amongst these patterns.

The reason for designing the system in this way is that there is no guarantee that any given pattern derived from an multiple alignment will ultimately turn out to be `correct' in terms of minimum length encoding principles or one's intuitions about what the correct grammar should be. Indeed, many of the patterns abstracted by the system are clearly `wrong' in these terms. Retention of older patterns in the store alongside patterns that have been derived from them leaves the door open for the system to create `correct' patterns at later stages regardless of whether `wrong' patterns had been created earlier. In effect, the system is able to explore alternative paths through the abstract space of possible patterns.

\subsection{Sifting and sorting of patterns}\label{sifting_and_sorting_section}

Identification of `wrong' patterns occurs in the {\em sifting\_and\_sorting()} stage of processing (operation 4 in Figure \ref{sp70_figure}), where the system develops one or more alternative grammars for the patterns in New in accordance with minimum length encoding principles. Figure \ref{sifting_and_sorting_figure} shows the overall structure of the {\em sifting\_and\_sorting()} function.

Each pattern in Old has an associated frequency of occurrence and, at the start of the function, all these values are set to zero. Then, all the patterns in New are reprocessed with the {\em create\_multiple\_alignments()} function, building multiple alignments as before, each one between one pattern from New and one or more patterns in Old. 

The difference on this occasion is that, for each pattern from New, the best multiple alignments are filtered to remove any that contain unmatched symbols in the current pattern from New or unmatched C-symbols in any pattern from Old. The remaining `full' multiple alignments provide the basis for further processing.

\begin{figure}[!hbt]
\fontsize{09.00pt}{10.80pt}
\centering
\begin{BVerbatim}
SIFTING_AND_SORTING()
{
     1 For each pattern in Old, set its frequency of occurrence to 0.
     2 While (there are still unprocessed patterns in New)
     {
          2.1 Identify the first or next pattern from New as the
               current pattern from New.
          2.2 Apply the function CREATE_MULTIPLE_ALIGNMENTS() to
               create multiple alignments, each one between the
               current pattern from New and one or more patterns
               from Old.
          2.3 From amongst the best of the multiple alignments
               formed, select `full' multiple alignments in which
               all the symbols of the current pattern from New are
               matched and all the C-symbols are matched in each
               pattern from Old.
          2.4 For each pattern from Old, count the maximum number of
               times it appears in any one of the full multiple
               alignments selected in operation 2.3. Add this count
               to the frequency of occurrence of the given pattern.
     }
     3 Compute frequencies of symbol types and their encoding costs.
          From these values, compute encoding costs of patterns in
          Old and new compression scores for each of the full
          multiple alignments created in operation 2.
     4 Using the multiple alignments created in 2 and the values computed in
          operation 3, COMPILE_ALTERNATIVE_GRAMMARS().
}
\end{BVerbatim}
\caption{The organisation of the {\em sifting\_and\_sorting()} function. The {\em compile\_alternative\_grammars()} function is described in Section \ref{compile_grammars}.}
\label{sifting_and_sorting_figure}
\end{figure}

In this phase of the program, we can be confident of finding at least one full multiple alignment for each pattern from New because, in the previous phase, each unmatched portion of the given pattern from New led to the creation of patterns in Old that would provide an appropriate match in the future.

When all the patterns from New have been processed in this way, there is a set $A$ of full multiple alignments, divided into $b_1 ... b_m$ disjoint subsets, one for each pattern from New. From these multiple alignments, the function computes the frequency of occurrence of each of the $p_1 ... p_n$ patterns in Old as:

\begin{equation}
f_i = \sum_{j = 1}^{j = m} max(p_i, b_j)
\label{frequency_of_patterns_equation}
\end{equation}

\noindent where $max(p_i, b_j)$ is the maximum number of times that $p_i$ appears in any {\em one} multiple alignment in subset $b_j$. Using the maximum value for any {\em one} multiple alignment for a given pattern from New is necessary because the multiple alignments in each $b_j$ are {\em alternative} analyses of the corresponding pattern from New. If we simply counted the number of times each pattern appeared in all the multiple alignments for a given pattern from New, the frequency values would be too high.

The function also compiles an alphabet of the symbol types, $s_1 ... s_r$, in the patterns in Old and, following the principles just described, computes the frequency of occurrence of each symbol type as:

\begin{equation}
F_i = \sum_{j = 1}^{j = m} max(s_i, b_j)
\label{frequency_of_symbol_types_equation}
\end{equation}

\noindent where $max(s_i, b_j)$ is the maximum number of times that $s_i$ appears in any {\em one} multiple alignment in subset $b_j$.

From these values, the encoding cost of each symbol type is computed using the Shannon-Fano-Elias method as before \citep{cover_thomas_1991}. As usual (Section \ref{encoding_individual_symbols}), the encoding cost of each of the `data' symbol types (those that appears in New) is weighted so that data symbols behave as if they were relatively large chunks of information.

Each symbol in each pattern in New and Old is then assigned the frequency and encoding cost of its type. With these values in place, the compression score of each multiple alignment in the set of full multiple alignments is recalculated.

Finally, in operation 4 of Figure \ref{sifting_and_sorting_figure}, a set of one or more alternative grammars is compiled, as described in Section \ref{compile_grammars}.

As the program stands, these alternative grammars are simply presented to the user for inspection. However, it is intended that the patterns in Old should be purged of all its patterns except those in the best grammar that has been found. It is anticipated that the program will be developed so that patterns from New will be processed in batches and that this kind of purging of Old will occur at the end of each batch to remove the `rubbish' and retain only those patterns that have proved useful in encoding the cumulative set of patterns from New.

\subsubsection{Compiling a set of alternative grammars}\label{compile_grammars}

\sloppy The set of alternative grammars for the patterns in New are derived (in the {\em compile\_alternative\_grammars()} function) from the full multiple alignments created in operation 2 of Figure \ref{sifting_and_sorting_figure}.

For any given pattern from New, a grammar for that pattern can be derived from one of its full multiple alignments by simply listing the patterns from Old that appear in that multiple alignment, counting multiple appearances of any pattern as one. Any such grammar may be augmented to cover an additional pattern from New by selecting {\em one} of the full multiple alignments for the second pattern from New and adding patterns from Old that appear within that multiple alignment and are not already present in the grammar (counting multiple appearances as one, as before). In this way, taking one pattern from New at a time, a grammar may be compiled for all the patterns from New.

A complication, of course, is that there are often two or more full multiple alignments for any given pattern from New. This means that, for a given set of patterns from New, one can generate a tree of alternative grammars with branching occurring wherever there are two or more alternative multiple alignments for a given pattern from New. Without some constraints, this tree can become unmanageably large.

In the {\em compile\_alternative\_grammars()} function, the tree of alternative grammars is pruned periodically to keep it within reasonable bounds. The tree is grown in successive stages, at each stage processing the multiple alignments for {\em one} of the patterns from New and, for each grammar, processing only {\em one} multiple alignment for each pattern from New. Values for $G$, $E$ and $T$ are calculated for each grammar and, at each stage, grammars with high values for $T$ are eliminated.

For a given grammar comprising patterns $p_1 ... p_g$, the value of $G$ is calculated as:

\begin{equation}
G = \sum_{i=1}^{i=g}(\sum_{j=1}^{j=L_i}s_j)
\label{size_of_grammar_equation}
\end{equation}

\noindent where $L_i$ is the number of symbols in the $i$th pattern and $s_j$ is the encoding cost of the $j$th symbol in that pattern.

Given that each grammar is derived from a set $a_1 ... a_n$ of multiple alignments (one multiple alignment for each pattern from New), the value of $E$ for the grammar is calculated as:

\begin{equation}
E = \sum_{i=1}^{i=n}e_i
\label{size_of_encoded_data_equation}
\end{equation}

\noindent where $e_i$ is the size, in bits, of the code string derived from the $i$th multiple alignment (Section \ref{ma_evaluation}).

When the set of alternative grammars has been completed, each grammar is `cleaned up' by removing code symbols that have no function in the grammar and by renaming code symbols in a tidy manner (see Section \ref{learning_example_1}, below).

Before leaving this section, it is worth pointing out a minor anomaly in the way values for $G$ and $E$ are calculated. These values depend on the encoding costs of symbols which themselves depend on frequencies of symbol types. At present, these frequencies are derived from the entire set of patterns in Old but it would probably be more appropriate if frequency values were derived from each grammar individually at each stage in the process of compiling grammars. The results are likely to be similar in both cases since encoding costs depend on relative frequency values, not absolute values and, for the symbol types appearing in any one grammar, it is likely that the ranking of frequency values derived from the grammar would be similar to the ranking derived from the entire set of patterns in Old.

\section{Evaluation of the model}

Criteria that may be used to evaluate a learning model like SP70 include:

\begin{itemize}

\item If applications are scaled up to realistic size, is the model likely to make unreasonable demands for processing power or computer memory?

\item Does it produce knowledge structures that look `natural' or `reasonable'?

\item Can the model discover the grammar that was used to create the data from which it has learned?

\item Given that the system aims to find grammars with relatively small values for $T$, does it succeed in this regard?

\item Does it produce knowledge structures that successfully support other operations such as reasoning, making `expert' judgements, playing chess, and so on?

\end{itemize}

The evaluation of SP70 in terms of the first of these criteria is discussed in the next subsection. The second and third criteria are discussed in Section \ref{learning_plausible_structures} and at various points in Sections \ref{learning_examples_section} and \ref{learning_discussion}. Evidence bearing on the fourth criterion is presented in Section \ref{plotting_values}, below. No attempt has yet been made to evaluate SP70 in terms of the fifth criterion.

\subsection{Computational complexity}\label{sp70_computational_complexity}

\index{complexity, computational|(}

In common with other programs for unsupervised learning (and, indeed, other programs for finding good multiple alignments), SP70 does not attempt to find theoretically ideal solutions. This is because the abstract space of possible grammars (and the abstract space of possible multiple alignments) is, normally, too large to be searched exhaustively. In general, heuristic techniques like hill climbing, genetic algorithms, simulated annealing etc must be used. By using these techniques, one can normally convert an intractable computation into one with computational complexity that is within acceptable limits.

\sloppy In SP70, the critical operation is the formation of multiple alignments ({\em create\_multiple\_alignments()}). Other operations (e.g., the {\em derive\_patterns()} function) are, in comparison, quite trivial in their computational demands.

\sloppy In a serial processing environment, the time complexity of the {\em create\_multiple\_alignments()} function has been estimated \citep{prob_reason_report} to be approximately O$(\log_2 n \times nm)$, where $n$ is the size of the pattern from New (in bits) and $m$ is the sum of the lengths of the patterns in Old (in bits). In a parallel processing environment, the time complexity may approach O$(\log_2 n \times n)$, depending on how well the parallel processing is applied. In serial and parallel environments, the space complexity should be O$(m)$.

In SP70, the function is applied (twice) to the set of patterns in New so we need to take account of how many patterns there are in New. It seems reasonable to assume that the sizes of patterns in New are approximately constant.\footnote{There is no requirement in the model that patterns in New should, for example, be complete sentences. They may equally well be arbitrary portions of incoming data, perhaps measured off by some kind of input buffer.}

Old is initially empty and grows as learning proceeds. The size of Old (before purging) is, approximately, a linear function of the size of New. Given this growth in the size of Old, the time required to create multiple alignments for any given pattern from New will grow as learning proceeds. Again, the relationship is approximately linear. 

So if we ignore operations other than the {\em create\_multiple\_alignments()} function, we may estimate the time complexity of the program (in a serial environment) to be O$(N^2)$ where $N$ is the number of patterns in New. In a parallel processing environment, the time complexity may approach O$(N)$, depending on how well the parallel processing is applied. In serial or parallel environments, the space complexity should be O$(N)$.%
\index{complexity, computational|)}

\subsection{Learning plausible structures}\label{learning_plausible_structures}

One way to assess a learning system is to look at the rules that it creates or the way it parses its corpus of raw data and to make a judgement of whether the rules or the parsings look natural. This `looks-good-to-me' approach \citep{van_zaanen_thesis_2002} may seem crude but we should remember that the human brain is currently the best learning system in existence so human judgements of whether a grammar or a parsing are well-structured should not be dismissed too readily. 

This approach can be formalised. For example, in \citet{wolff_1977}, the parsings of an natural language text produced by the MK10 model are compared with the segmentation into words that would be approved by an editor or printer. Using the Chi Square test it was found that the correspondence was highly significant and very unlikely to have been obtained by random processes. Likewise, in \citet{wolff_1980}, the phrase-structure parsings obtained by MK10 were found to have a statistically-significant correspondence with parsings produced independently by an expert in linguistics.

A related approach to the evaluation of grammar learning systems is to create an artificial grammar, use it to produce a sample of artificial language, and then see whether the learning system, using the sample, is able to discover or infer the original grammar. This may seem more rigorous than an informal judgement of whether the system has produced a plausible set of rules or a plausible parsing but it should not be forgotten that the artificial grammar is itself the product of human judgement. In many respects, the two approaches are equivalent.

As we shall see in the next section, SP70 can learn simple grammars that look plausible but further work will be needed before it will be appropriate to assess the system in a more formal way with statistical tests or the like. 

\section{Examples}\label{learning_examples_section}

This section presents two examples to illustrate how SP70 works and what it can do. The examples are fairly simple, partly for the sake of clarity and partly because of shortcomings in the model (discussed in Section \ref{learning_discussion}, below).

\subsection{Example 1}\label{learning_example_1}

The four short sentences supplied to SP70 as New for this first example are shown in Figure \ref{learning_example_1_patterns}.

\begin{figure}[!hbt]
\centering
\begin{BVerbatim}
j o h n r u n s
m a r y r u n s
j o h n w a l k s
m a r y w a l k s
\end{BVerbatim}
\caption{Four patterns supplied to SP70 as New.}
\label{learning_example_1_patterns}
\end{figure}

The two best grammars found by the program for these sentences (with 10 as the cost factor) are shown in Figure \ref{learning_example_1_grammars}. In (a) the best grammar is shown in the form that it is first compiled and in (b) the same grammar is shown after it has been cleaned up. Figure \ref{learning_example_1_grammars} (c) shows the second-best grammar after it has been cleaned up.

\begin{figure}[!hbt]
\centering
\begin{BVerbatim}
< %4 15 s >
< %7 %9 %11 152 m a r y >
< %9 %14 162 j o h n >
< %24 406 r u n >
< %24 %27 407 w a l k >
< %25 412 < %9 > < %24 > < %4 > >

(a)

< %2 1 s >
< %3 2 m a r y >
< %3 3 j o h n >
< %1 4 r u n >
< %1 5 w a l k >
< 6 < %3 > < %1 > < %2 > >

(b)

< %2 2 m a r y >
< %2 3 j o h n >
< %1 1 r u n s >
< %1 5 w a l k s >
< 4 < %2 > < %1 > >

(c)
\end{BVerbatim}
\caption{Grammars found by SP70 with the four patterns shown in Figure \ref{learning_example_1_patterns} supplied as New. (a) The best grammar without cleaning up. (b) The same grammar after cleaning up. (c) The second best grammar after cleaning up.}
\label{learning_example_1_grammars}
\end{figure}

Cleaning up grammars means removing class symbols that have no referents (e.g., `\%7' and `\%11' in the second pattern in Figure \ref{learning_example_1_grammars} (a)) and renumbering the class symbols and the discrimination symbols, starting from 1 for each set. The renumbering is a purely cosmetic matter and makes no difference to the encoding cost calculated for each symbol.

The two grammars shown in Figure \ref{learning_example_1_grammars} are both reasonably plausible grammars for the four original sentences. In the best grammar `m a r y' and `j o h n' are picked out as discrete words and assigned to the same grammatical class (`\%3'). In a similar way, `r u n' and `w a l k' are each picked out as a discrete entity---corresponding to the `stem' of a verb---and both are assigned to the class `\%1'. The suffix for these verb stems (`s') is picked out as a distinct entity and the overall sentence structure is captured in the pattern `$<$ 6 $<$ \%3 $>$ $<$ \%1 $>$ $<$ \%2 $>$ $>$'. The second best grammar is the same except that the suffix of each verb is not separated from the stem.

\subsubsection{Intermediate results}

The simplicity of the results shown in Figure \ref{learning_example_1_grammars} disguises what the program has done to achieve them. The flavour of this processing may be seen from the selection of intermediate results presented here.

When the first pattern from New is processed, Old is empty except for a copy of that first pattern that is added to Old, one symbol at a time, as the pattern is processed. The only multiple alignment formed at this stage is shown in Figure \ref{learning_alignment_3}. Remember that the symbols in the pattern from New (`j o h n r u n s') must not be aligned with the corresponding symbols in the pattern in Old because, in effect, this would mean matching each symbol with itself (Section \ref{copying_cpfn}).

\begin{figure}[!hbt]
\centering
\begin{BVerbatim}
0 j o h n r u  n s         0
               |          
1 < %1 5 j o h n r u n s > 1
\end{BVerbatim}
\caption{The only multiple alignment formed when the first pattern from New is processed.}
\label{learning_alignment_3}
\end{figure}

It is evident that, at this stage, opportunities to gain any useful insights into the overall structure of the patterns in New are quite limited. From the `bad' multiple alignment shown in Figure \ref{learning_alignment_3} the program abstracts the `bad' patterns `$<$ \%2 7 n $>$', `$<$ \%3 10 j o h $>$', `$<$ \%3 11 j o h n r u $>$', `$<$ \%4 14 r u n s $>$', `$<$ \%4 15 s $>$' and `$<$ \%5 19 $<$ \%3 $>$ $<$ \%2 $>$ $<$ \%4 $>$ $>$'.

When the next pattern from New (`m a r y r u n s') is processed, the program is able to form more sensible multiple alignments like

\begin{center}
\begin{BVerbatim}
0 m a r y r u n s   0
          | | | |  
1 < %4 14 r u n s > 1
\end{BVerbatim}
\end{center}

\noindent and

\begin{center}
\begin{BVerbatim}
0 m a r y        r u n s   0
                 | | | |  
1 < %1 5 j o h n r u n s > 1.
\end{BVerbatim}
\end{center}

The first of these multiple alignments yields the patterns `$<$ \%7 152 m a r y $>$' and `$<$ \%8 157 $<$ \%7 $>$ $<$ \%4 $>$ $>$'. It would have created a pattern for `r u n s' but this is suppressed because the program detects that a pattern with those C-symbols already exists (`$<$ \%4 14 r u n s $>$').

From the second multiple alignment, the system derives the pattern `$<$ \%9 162 j o h n $>$' (assigned to the class `\%9') and it would create a pattern for `m a r y' but it detects that a pattern with these C-symbols already exists (`$<$ \%7 152 m a r y $>$'). However, since `j o h n' and `m a r y' occur in the same context (`---r u n s'), they should be assigned to the same context-defined class. Accordingly, the program adds the class symbol `\%9' to the pattern `$<$ \%7 152 m a r y $>$' so that it becomes `$<$ \%7 \%9 152 m a r y $>$' and it creates the abstract pattern `$<$ \%10 168 $<$ \%9 $>$ $<$ \%4 $>$ $>$', tying the whole structure together.

As processing proceeds in the pattern-generation phase (operation 3 in Figure \ref{sp70_figure}), the program forms good multiple alignments like

\begin{center}
\begin{BVerbatim}
0        j o h n w a l k s   0
         | | | |         |  
1 < %1 5 j o h n r u n   s > 1
\end{BVerbatim}
\end{center}

\noindent and relatively bad ones like

\begin{center}
\begin{BVerbatim}
0        j o h       n w a l k s   0
         | | |       |         |  
1 < %1 5 j o h n r u n         s > 1.
\end{BVerbatim}
\end{center}

\noindent From multiple alignments like these it derives correspondingly good and bad patterns.

By the time the last pattern from New is processed (`m a r y w a l k s') there are enough good patterns in Old for the program to start forming quite plausible multiple alignments like

\begin{center}
\fontsize{08.00pt}{09.60pt}
\begin{BVerbatim}
0                          m a r y                        w a l k s     0
                           | | | |                        | | | | |    
1                          | | | |   < %22 %29 %4 %31 394 w a l k s >   1
                           | | | |   |         |                    |  
2 < %8 157 < %7            | | | | > <         %4                   > > 2
           | |             | | | | |                                   
3          < %7 %9 %11 152 m a r y >                                    3.
\end{BVerbatim}
\end{center}

At the end of this phase of processing, Old contains a variety of patterns including several good ones and quite a lot of bad ones.

In the second {\em sifting\_and\_sorting()} phase of the program (operation 4 in Figure \ref{sp70_figure}) the program compiles a set of alternative grammars the best of which are shown in Figure \ref{learning_example_1_grammars}.

\subsection{Example 2}\label{learning_example_2}

When New contains the eight sentences shown in Figure \ref{learning_example_2_patterns}, the best grammar found by SP70 (after cleaning up) is the one shown in Figure \ref{learning_example_2_grammar}.

\begin{figure}[!hbt]
\centering
\begin{BVerbatim}
t h a t b o y r u n s
t h a t g i r l r u n s
t h a t b o y w a l k s
t h a t g i r l w a l k s
s o m e b o y r u n s
s o m e g i r l r u n s
s o m e b o y w a l k s
s o m e g i r l w a l k s
\end{BVerbatim}
\caption{Eight sentences supplied to SP70 as New.}
\label{learning_example_2_patterns}
\end{figure}

\begin{figure}[!hbt]
\centering
\begin{BVerbatim}
< %2 2 s o m e >
< %2 3 t h a t >
< %1 5 b o y >
< %1 6 g i r l >
< %3 4 r u n s >
< %3 7 w a l k s >
< 1 < %2 > < %1 > < %3 > >
\end{BVerbatim}
\caption{The best grammar found by SP70 (after cleaning up) when New contains the eight sentences shown in Figure \ref{learning_example_2_patterns}.}
\label{learning_example_2_grammar}
\end{figure}

This result looks reasonable but, in the light of the best grammar found in Example 1, one may wonder why the terminal `s' of `r u n s' and `w a l k s' has not been identified as a discrete entity, separate from the verb stems `r u n' and `w a l k'.

In the pattern-generation phase of processing, SP70 does form multiple alignments like this

\begin{center}
\begin{BVerbatim}
0        t h a t b o y w a l k s   0
         | | | | | | |         |  
1 < %1 9 t h a t b o y r u n   s > 1
\end{BVerbatim}
\end{center}

\noindent which clearly recognises the verb stems and `s' as distinct entities. But for reasons that are still not entirely clear, the program does not build these entities into plausible versions of the full sentence structure. The program isolates the pattern `$<$ \%25 599 t h a t b o y $>$' from the multiple alignment just shown but for some reason it fails to find internal structure within that pattern---although it recognises `t h a t' and `b o y' in other contexts. This issue is discussed in Section \ref{finding_internal_structure}, below.

\subsubsection{Plotting values for $G$, $E$ and $T$}\label{plotting_values}

The value of $T$ for the best grammar is 18695 bits before cleaning up and 18515 bits after cleaning up. By contrast, the total number of bits in the original 8 patterns is 69172.

\begin{figure}[!hbt]
\centering
\includegraphics[width=0.9\textwidth]{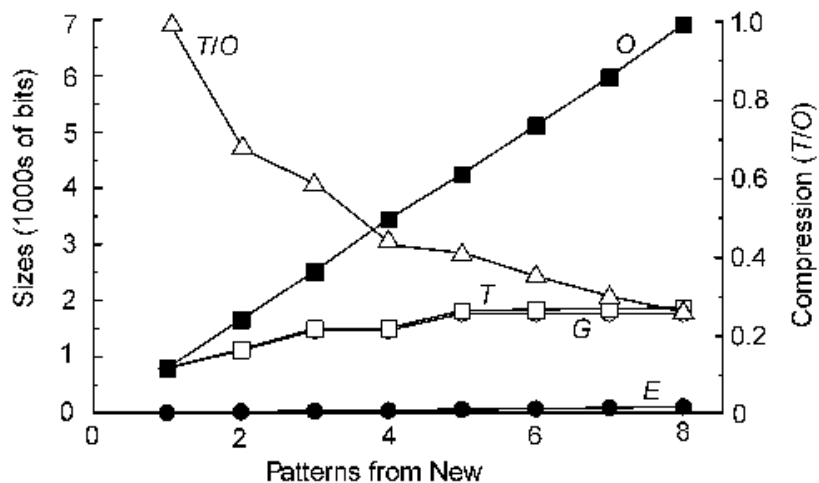}
\caption{The lower three curves show values of $G$, $E$ and $T$ (on the scale to the left) for the best grammar found (before cleaning up) as successive patterns from New are processed in {\em compile\_alternative\_grammars()} with Example 2. The curve marked `$O$' shows cumulative values (on the left scale) for the numbers of bits in the original patterns from New. The curve marked `$T/O$' plots the ratio of $T$ to $O$, representing the amount of compression achieved (on the scale to the right).}
\label{plotting_figure}
\end{figure}

Figure \ref{plotting_figure} shows how the values of $G$, $E$ and $T$ change as successive patterns from New in Example 2 are processed when the set of alternative grammars is compiled. Each point on each of the lower three graphs represents the relevant value (on the scale at the left) from the best grammar found after the full multiple alignments for a given pattern from New have been processed. The graph labelled `$O$' shows cumulative values on the scale at the left for the original patterns from New. The graph labelled `$T/O$' shows the amount of compression achieved (on the scale to the right).

Notice how the value of $G$ does not change for the last 4 points. This is because the grammar developed for the first four sentences provides an efficient of encoding of the remaining sentences.

\section{Discussion}\label{learning_discussion}

This section discusses a selection of issues relating to the SP70 model, especially shortcomings of the current version and how they may be overcome.

\subsection{Finding internal structure}\label{finding_internal_structure}

The failure of the model to find some of the internal structure within a sentence in Example 2 seems to be a manifestation of a more general shortcoming. Although the model in its current form can isolate basic segments and tie them together in an overall abstract structure, it is not good at finding intermediate levels of abstraction.

One possible solution is some kind of additional reprocessing of the patterns in Old, including the abstract patterns that have been added to Old, to discover partial matches not detected in the initial processing of the patterns from New. This should allow the system to detect intermediate levels of structure such as phrases or clauses or structures that may exist within smaller units such as words.

Another possible solution is to modify the process of building multiple alignments so that, instead of discarding multiple alignments containing mismatches between Old patterns, the system derives new patterns from the mismatches and adds them to Old, much as described in Section \ref{derive_patterns_section}. This should avoid the awkwardness involved in deriving new patterns from multiple alignments containing three or more rows (Section \ref{deriving_from_three_plus_rows}) and should also allow a closer integration between parsing and learning.

\subsection{Finding discontinuous dependencies}

\index{grammar!syntactic dependencies|(}

In the development of the model to date, no attempt has been made to enable the system to detect discontinuous dependencies in syntax such as the number and gender dependencies discussed in Section \ref{dependencies_in_french} (and, perhaps, intra-word dependencies in Arabic and some other languages).

Although this kind of capability may seem like a refinement that we can afford to do without at this stage of development, a deficiency in this area seems to have an impact on the program's performance at an elementary level. Even in quite simple structures, dependencies can exist that bridge intervening structure and, in its current form, the program does not encode this kind of information in a satisfactory manner.

There do not seem to be any insuperable obstacles to solving this problem within the SP system:

\begin{itemize}

\item The format for representing knowledge accommodates these kinds of structure quite naturally.

\item Finding partial matches that bridge intervening structure is bread-and-butter for the {\em create\_multiple\_alignments()} function.

\end{itemize}

\noindent What seems to be required is some revision of the way in which patterns are derived from multiple alignments so that the system creates patterns containing breaks as well as coherent substrings.%
\index{grammar!syntactic dependencies|)}

\subsection{Generalisation of grammatical rules and the correction of overgeneralisations}\label{generalisation}

\index{grammar!generalisation|(}

As we noted in Section \ref{grammatical_inference_and_generalisation}, a well-documented phenomenon in the way young children learn language is that they say things like ``The ball was hitted'' or ``Look at the gooses'', apparently applying general rules for constructing words but applying them too generally. How is it that children eventually learn to avoid these kinds of overgeneralisations? It is tempting to suppose that children are corrected by parents or other adults but the weight of empirical evidence is that, while such corrections may be helpful, they are not actually necessary for language learning.

Minimum length encoding principles provide an elegant solution to this puzzle. Without the kind of feedback or supplementary information postulated by \citet{gold_1967}, it is possible to search for grammars that are good in minimum length encoding terms and these are normally ones that steer a path between generalisations that are, intuitively, `correct' and others that appear to be `wrong'. This kind of effect has been demonstrated with the SNPR model \citep{wolff_1982}.

When SP70 is run on the first three of the four patterns shown in Figure \ref{learning_example_1_patterns}, the best grammar found is exactly the same as before (Figure \ref{learning_example_1_grammars} (b)). This grammar generates the missing sentence (`m a r y w a l k s') as well as the other three sentences, but it does not generate anything else.

In this example, the model generalises in a way that seems intuitively to be correct and avoids creating overgeneralisations that are glaringly wrong. However, relatively little attention has so far been given to this aspect of the model and further work is required. In particular, a better understanding is needed of alternative ways in which grammatical rules may be generalised.%
\index{grammar!generalisation|)}

\subsection{Learning recursive structures}

\index{recursion!learning}

Recursive structures are prominent in natural languages (Section \ref{language_recursion}) and many other kinds of knowledge. As we saw in Section \ref{techniques_for_ic}, recursive functions provide a convenient means of achieving `run-length coding', one of our three basic techniques for information compression.

As we have seen in Sections \ref{unary_numbers_and_sp} and \ref{language_recursion} and elsewhere in this book, recursive structures and functions fit comfortably in the SP framework because they can be modelled using exactly the same mechanisms as are provided for encoding other kinds of knowledge. It is anticipated that such structures may be learned using the same principles and processes are required for the learning of other kinds of knowledge, without the need for any special or {\em ad hoc} mechanisms. That said, it will be necessary to ensure that future versions of the SP70 model can indeed learn recursive structures in an appropriate manner using the same mechanisms as are provided for the learning of non-recursive structures.

\subsection{Symmetry and mirror images}

\index{symmetry}

A form of redundancy that is prominent in many kinds of knowledge is that represented by symmetrical structures and mirror images. It is clear that significant compression can be achieved if this kind of redundancy can be recognised and extracted when it is found. This is outside the scope of SP61 and SP70 but at some stage in the future this issue will need to be addressed. With 1-D patterns, this means developing an ability to match patterns in two directions instead of one and it also means developing an appropriate method for encoding symmetry wherever it is found.

\subsection{Learning other kinds of knowledge}

The emphasis in this chapter and in the development of SP70 has been on the learning of grammar-like structures from language-like input. The reason for this emphasis is partly my long-standing interest in language learning \citep[][and earlier papers]{wolff_1988} and partly because grammar induction, as one of the most challenging kinds of learning, is likely to yield insights and solutions that may be applied to other kinds of learning. If an SP model can be developed with robust capabilities for grammar induction, it seems likely that it may be applied with little or no modification to the learning of other kinds of knowledge such as if-then rules, class hierarchies, part-whole hierarchies, and associations between syntactic structures and the non-linguistic `meanings' of those structures (see next section).

In all cases, the patterns to be learned reflect redundancies in the data. The associations amongst the elements of a coherent or discontinuous syntactic pattern are similar to the associations in an if-then rule. The way in which a sentence may be broken into parts and sub-parts is essentially the same as the way in which a car, house or other non-syntactic entity may be described in terms of a hierarchy of components. And the levels of abstraction in a typical grammar are essentially the same as the levels of abstraction in a hierarchy of classes and subclasses.

At some stage, it will be interesting to try to generalise the concepts that have been developed for 1-D patterns to the learning of structures in a two-dimensional environment. Some possibilities are sketched in Section \ref{2d_patterns_section}.

\subsection{Learning associations between words and meanings}\label{syntax_semantics_learning}

\index{grammar!semantics|(}

Learning associations between syntactic forms such as words and their corresponding meanings seems puzzling because, in any given situation, there is a large number of possibilities. If, for example, a child hears ``look there is a cat'' as a cat goes by, this could refer to the colour of the cat's paw, the act of walking, the time of day, or any other aspect of what the child would see. Quine \citeyearpar{quine_1960} has called this problem ``the scandal of induction''.

Although this seems puzzling at first sight, an association between a word and any one of its meanings is no different in principle from any other kind of association, including the kinds of association embodied in the syntax of any language. Notwithstanding the huge numbers of possibilities, the use of heuristic techniques should make it feasible to search successfully for associations that are good in minimum length encoding terms. This has been amply demonstrated with the MK10 and SNPR models and also with SP70. In short, ``the scandal of induction'' appears to be well within the scope of established principles.

Another problem that is potentially very challenging is that many meanings are quite abstract. We may represent the meaning of `cat' with some complex of cat-like features. But representing the meaning of `justice', `liberty' or `democracy' is a much more slippery eel! Pending any serious attempt to apply the SP theory to problems of this sort, I shall assume that abstract concepts like the ones just mentioned may be accommodated in essentially the same way that other kinds of abstractions may be represented, as described in Section \ref{class_part_inheritance}.

In the context of the SP theory and its current state of development, a more immediate problem may be that words and meanings appear to be in different `domains': a word in spoken or written form may need to be associated with a meaning that may be some complex of sound, vision, touch and so on that does not fall neatly into the same auditory or visual stream as the given word. This is similar to the problem of learning `spelling rules'---meaning associations between letter combinations and their sound values \citep[see][]{wolff_spelling_rules}.

Given that current SP models are restricted to 1-D patterns, a way to imitate the relationship between words and their potential meanings in New information might be to create groupings in which several words and potential meanings can appear and the potential meanings would be scattered randomly within each group.

A possible snag here is that there are two possible orders for a word and its meaning and this would dilute the association and lead to the creation of two patterns for each word-meaning association, instead of the one that seems preferable. There may be a similar snag with 2-D patterns because there are many possible arrangements of `cat' and `CAT' (representing the meaning of `cat') in a 2-D array and it is not clear how their association could be abstracted from their specific arrangements.

A possible answer to this problem might be to generalise the matching process in the SP models so that---in the case of 1-D patterns---any two patterns may be compared in both of the two possible directions---and in the case of 2-D patterns---comparisons may be made in many possible orientations of the patterns. Generalisations like these will, in any case, be needed to account for our ability to recognise `n o i t a m r o f n i' as being the same as `i n f o r m a t i o n', but reversed, and our ability to recognise someone in a photograph even though the person, or the photograph, may be upside down.%
\index{grammar!semantics|)}

\subsection{Motivation and emotion}\label{motivation_and_emotion_in_learning}

\index{motivation|(}\index{emotion|(}

Although learning has been considered in this chapter primarily as an engineering problem, the theory that has been described may be viewed as a possible theory of learning in people and other animals.

In the theory as it has been developed to date, compression of information is the driving force in learning. But it is clear that motivational factors also have a r{\^o}le to play as ``occurs in `flash-bulb memories,' memories of emotionally charged events that are recalled in detail, as if a complete picture had been instantly and powerfully etched in the brain.'' \citep[][p. 1034]{kandel_2001}. How would this fit in with the kinds of learning mechanisms that have been described?

The tentative suggestion here is that motivations and emotions may have an impact when patterns are purged from the system. Other things being equal, we may suppose that minimum length encoding principles govern the choice of which patterns should be retained and which should be discarded. But any pattern that represents something that has special significance may be retained by the system even if it does not score well in terms of minimum length encoding measures.%
\index{motivation|)}\index{emotion|)}

\subsection{Other developments}

Other areas where further work is planned include:

\begin{itemize}

\item As was indicated in Section \ref{sifting_and_sorting_section}, it is anticipated that the program will be developed so that it processes patterns from New in batches, purging bad patterns from Old at the end of each batch.

\item As was noted in Section \ref{avoiding_duplication}, it is possible for two or more context-defined classes to be the same. The program needs to check for this possibility and merge identical classes whenever they are found.

\item At present, the program applies the {\em create\_multiple\_alignments()} function twice to each pattern from New, once as part of the process of generating patterns to be added to Old and once in the {\em sifting\_and\_sorting()} phase. It seems possible that the two phases could be integrated so that the {\em create\_multiple\_alignments()} function need only be applied once to each pattern from New.

\item As was suggested in Section \ref{derive_patterns_section}, there may be a case for introducing a `null' pattern to allow for the encoding of optional elements in a syntactic structure.

\item Although the computational complexity of the model on a serial machine is within acceptable limits, improvements in that area, and higher absolute speeds, may be obtained by the application of parallel processing. If or when residual problems in the model have been solved, it is envisaged that the system will be developed as a software virtual machine on existing high-parallel hardware or perhaps on new forms of hardware dedicated to the needs of the model. It may be possible to exploit optical techniques to achieve high-parallel matching of patterns in the core of the model (see Chapter \ref{future_chapter}).

\end{itemize}

\section{Conclusion}

Although SP70 is still some way short of an `industrial strength' system for unsupervised learning, the results obtained so far are good enough to show that the general approach is sound. Problems that have been identified appear to be soluble.

A particular attraction of this approach to learning is that the SP system provides a unified view of a variety of issues in artificial intelligence thus facilitating the integration of learning with other aspects of intelligence.%
\index{learning|)}\index{SP70|)}

%% file: maths_logic.tex
\chapter{Mathematics and Logic}\label{maths_logic_chapter}

\index{mathematics|(}\index{logic|(}

\section{Introduction}

This chapter considers the conjecture that mathematics, logic and related disciplines may usefully be understood as information compression and, more specifically, as information-compression-by-multiple-alignment-unification-and-search. Incidentally, the phrase `mathematics and logic and related disciplines' will be shortened to `mathematics and logic'.

Chapter \ref{computing_chapter}, showed how the intuitive concept of `computing' may be understood in terms of the SP theory. If the arguments in that chapter are accepted, and if `computing' is seen to include mathematics and logic, then the present chapter might be thought to be redundant. However, the purpose of the previous chapter was to show that the workings of a universal Turing machine and the workings of a Post canonical system may be understood in terms of the SP theory. By contrast, the purpose of this chapter is to show how aspects of mathematics and logic other than universal Turing machines or Post canonical systems may be understood in terms information compression or, more specifically, in terms of the SP theory.

There are no formal proofs in this chapter and indeed formal proofs in this context 
are probably inappropriate since the argumentation depends on one's sense of analogy and 
what is or has potential to be intellectually fruitful. The treatment of the 
proposals is very far from being exhaustive: the chapter merely introduces 
some ideas in support of the proposals and presents a selection of examples. Whether or how the ideas may be applied to the many areas of mathematics and logic not discussed in this chapter are matters that readers may like to consider.

Section \ref{ml_preliminaries}, next, is a preparation for two main sections that follow. The first of these (Section \ref{ml_structures}) discusses the conjecture in relation to `static' forms and structures in mathematics and logic, while the second (Section \ref{ml_processes}) discusses it in relation to the `dynamics' of inferences in mathematics and logic: `calculation', `proof', `deduction' etc. Section \ref{discussion_conclusion} briefly considers a range of issues that relate to the ideas that have been described.

\section{Preliminaries}\label{ml_preliminaries}

\subsection{Representation, semantics and Platonism}\label{repr_semantics_platonism}

In mathematics and logic, it is often convenient to make a distinction between the representation of an entity and the thing itself (the meaning or `semantics' associated with the representation). Thus the concept {\em four} may be represented by `four', `4', `IV', 
`1111' (in unary arithmetic), `100' (in binary arithmetic), and so on.

In mathematical Platonism\index{Platonism}, mathematical entities ``are not merely formal or quantitative structures imposed by the human mind on natural phenomena, nor are they only mechanically present in phenomena as a brute fact of their concrete being. Rather, they are numinous and transcendent entities, existing independently of both the phenomena they order and the human mind that perceives them.'' \citep[][pp. 95--96]{hersh_1997}. Such ideas are ``invisible, apprehensible by intelligence only, and yet can be discovered to be the formative causes and regulators of all empirical visible objects and processes.'' ({\em ibid.} p. 95).

The view of mathematical (and logical) concepts that has been adopted here is quite different. It is assumed that both the representations of mathematical or logical concepts and the concepts themselves are `information', as described in Section \ref{information_ic_probabilities}. If it is accepted that brains and computers are valid vehicles for mathematical and logical concepts and processes of calculation and inference, it is difficult to resist the conclusion that such concepts must exist within those systems in some relatively concrete form---arrays of binary digits in digital computers or, in brains, something like `cell assemblies', patterns of nerve impulses or, perhaps, DNA or RNA sequences. There is no place for ``numinous and transcendent entities'' in a Platonic world outside the brain or computer.

Do G{\"o}del's incompleteness theorems pose a problem for this non-Platonist view? Discussion of that issue would take us well beyond the scope of this chapter. We can at least say that any such problem is neither more nor less than it would be for any artificial computing system.

\subsection{The proposals in relation to established ideas}\label{ml_established_ideas}

\sloppy{The various `isms' in the philosophy and foundations of mathematics and logic---
foundationism, logicism, intuitionism (constructivism), formalism, Platonism, neo-Fregeanism, humanism, structuralism and so on---are well described in various sources (see, for example, \citet{potter_2000, hersh_1997, craig_1998, hart_1996, barrow_1992, eves_1990, epstein_carnielli_1989, boolos_jeffrey_1980}) and there is no intention to describe them again here.}

Insofar as the embryonic ideas to be presented constitute any kind of philosophy of 
mathematics and logic, it seems that they do not fit easily into any of the existing isms or schools of thought about the philosophy and foundations of mathematics and logic.

\subsubsection{Possible connections}\label{ml_possible_connections}

Devlin's interesting ideas about logic and information \citep{devlin_1991} might be thought to 
be related to the present proposals because a concept of information is fundamental in the
SP theory. However, Devlin develops a concept of information that is different from the 
`standard' concepts of Hartley-Shannon information theory and algorithmic information theory that have been adopted in 
the present work. Perhaps more important is the fact that information compression does not have any role in his proposals in contrast to the SP proposals where information compression has a key significance.

Two books about mathematics suggest in their titles that mathematics is a 
``science of patterns'' \citep{devlin_1997, resnik_1997}. Given the use of the word 
`pattern' in the present research, one might think that these two books give an account of mathematics that is closely related to the SP concepts. However, the word 
`pattern' in the present work is merely a generic name for an array of atomic symbols in 
one or more dimensions whereas in \citet{devlin_1997} and \citet{resnik_1997}, the word `pattern' has more abstract meanings.

That said, Devlin's book---which is intended as a `popular' presentation of 
mathematics rather than an academic treatise---gives a hint of a connection with the present work because the term `pattern' is used in a way that seems to be roughly equivalent to `regularity' and there is a connection between regularities and information compression. Resnik uses the term `pattern' to mean the same as the abstract concept of `structure' as it has been developed in the {\em structuralist} view of mathematics \citep[see also][]{shapiro_2000}. As with Devlin's book, there is a hint of a connection with the present work because of an apparent link between structuralist concepts and the intuitive concept of structure which itself seems to reflect redundancy.

Despite these possible links with the present work, neither of the books cited 
develops any connection with established concepts of information, redundancy or information compression and 
there is no mention of any concept of multiple alignment. Readers may wish to view the present proposals as an extension or development of structuralism. Alternatively, the proposals may be seen as an application of an independent intellectual tradition---{\em information theory}---to concepts in mathematics and logic, with some points of similarity with structuralism.

\subsection{Parsing in mathematics and logic}\label{maths_logic_parsing}

\index{parsing!mathematics and logic|(}

To bring us a little closer to the main themes of this chapter, this subsection presents an example showing how expressions in logic and mathematics may be parsed within the SP system. This, in itself, is not a process of mathematical or logical inference such as addition or multiplication but some kind of parsing is necessary for the interpretation of mathematical or logical symbols prior to the application of such processes. The example provides a bridge between the `linguistic' examples shown in Section \ref{framework_examples_section} and other examples presented later in this chapter showing how mathematical and logical structures and processes may be understood in terms of the SP theory.

Both mathematics and logic make frequent use of `expressions' that need to be analysed into their constituent parts before they can be evaluated. An example from arithmetic is $(((6 + 4)(16 - 4)) - 6)$. In logic one might write a composite proposition like $\sim((p \wedge q) \Rightarrow ((x \wedge y) \vee (s \wedge p)))$, where the letters are simple `atomic propositions' like ``Today is Tuesday'' or an equivalent `atomic formula' like `today(Tuesday)' and the other symbols have their standard meanings in logic (see the key to Figure \ref{logic_grammar}). Any such analysis is essentially a form of parsing in the same sense as was used in Section \ref{framework_examples_section}. And, like `linguistic' kinds of parsing, it requires an appropriate grammar.

It may be objected that the examples are already parsed because they contain the 
kinds of brackets that may be used to represent the structure of a sequence of symbols. 
The reason that parsing is needed in examples like these is that, until the constituent parts of a sequence are recognised by the system, the sequence is merely a stream of symbols and the role of the brackets in marking structures has not yet been recognised. Brackets are, in any case, not a necessary part of the examples---they might equally well be presented in Polish notation without the use of brackets.

Figure \ref{logic_grammar} shows part of a grammar for expressions in symbolic logic. Readers will see that it contains three main rules (beginning with `F'). The first one creates an `atomic formula' like `today(Tuesday)'. In a fuller grammar there would be rules for the creation of a variety of such atomic formulae. The second rule is a recursive rule covering cases like `proposition implies proposition', where `proposition' can be simple or complex. The third rule is also recursive, covering all forms of `not proposition'. The patterns beginning with `R' show some of the symbols that may be put between two propositions in the second rule in the grammar.

\begin{figure}[!hbt]
\centering
\begin{tabular}{l}
F $\rightarrow$ a \\
F $\rightarrow$ ( F R F ) \\
F $\rightarrow$ ( $\sim$ F ) \\
R $\rightarrow$ $\Rightarrow$ \\
R $\rightarrow$ $\wedge$ \\
R $\rightarrow$ $\vee$ \\
\\
etc
\end{tabular}
\caption{Part of a grammar for symbolic logic. {\em Key:} `a' = `atomic formula', 
`$\sim$' = `not', `$\wedge$' = `and', `$\vee$' = `or', `$\Rightarrow$' = `implies'.}
\label{logic_grammar}
\end{figure}

Figure \ref{logic_grammar_sp} shows the same grammar as the one in Figure \ref{logic_grammar} but it is recast into a form that is appropriate for the SP system. The main differences are that the grammar in SP style leaves out the rewrite arrow, it uses numbers to differentiate the three `F' rules and the three `R' rules, and it puts in a `terminating' symbol for each rule and each `call' to a rule. Each `terminating' symbol is formed by copying the initial symbol and adding `\#' to the front of it.

\begin{figure}[!hbt]
\centering
\begin{tabular}{l}
F 1 a \#F \\
F 2 ( F \#F R \#R F \#F ) \#F \\
F 3 ( $\sim$ F \#F ) \#F \\
R 1 $\Rightarrow$ \#R \\
R 2 $\wedge$ \#R \\
R 3 $\vee$ \#R \\
\end{tabular}
\caption{The grammar shown in Figure \ref{logic_grammar} recast in `SP' style.}
\label{logic_grammar_sp}
\end{figure}

Finally, Figure \ref{expression_parsing} shows the best multiple alignment that was found by SP61 with the pattern `( ( $\sim$ ( a $\Rightarrow$ a) ) $\wedge$ a )' in New and the patterns from Figure \ref{logic_grammar_sp} in Old. The multiple alignment may be interpreted as a parsing of the logical expression in terms of the given grammar and, as such, appears to be `correct'.

\begin{figure}[!hbt]
\fontsize{07.00pt}{08.40pt}
\centering
\begin{BVerbatim}
0     (     ( ~     (     a        =>        a    )    )        ^        a    )    0
      |     | |     |     |        |         |    |    |        |        |    |   
1     | F 3 ( ~ F   |     |        |         |    | #F ) #F     |        |    |    1
      | |       |   |     |        |         |    | |    |      |        |    |   
2 F 2 ( F       |   |     |        |         |    | |    #F R   | #R F   | #F ) #F 2
                |   |     |        |         |    | |       |   | |  |   | |      
3               F 2 ( F   | #F R   |  #R F   | #F ) #F      |   | |  |   | |       3
                      |   | |  |   |  |  |   | |            |   | |  |   | |      
4                     F 1 a #F |   |  |  |   | |            |   | |  |   | |       4
                               |   |  |  |   | |            |   | |  |   | |      
5                              R 1 => #R |   | |            |   | |  |   | |       5
                                         |   | |            |   | |  |   | |      
6                                        F 1 a #F           |   | |  |   | |       6
                                                            |   | |  |   | |      
7                                                           |   | |  F 1 a #F      7
                                                            |   | |                
8                                                           R 2 ^ #R               8
\end{BVerbatim}
\caption{The best multiple alignment found by SP61 with the expression 
`( ( $\sim$ ( a $\Rightarrow$ a) ) $\wedge$ a )' in New and the patterns shown in Figure \ref{logic_grammar_sp} in Old.}
\label{expression_parsing}
\end{figure}

\index{parsing!mathematics and logic|)}

\subsection{Mathematics, logic, information, redundancy and structure}

Before we proceed to the two main sections of this chapter, it may be useful to take a `global' view of mathematics and logic in terms of concepts that were introduced in Section \ref{information_ic_probabilities}.

The concept of `information', as previously described, is a highly abstract concept with a very wide scope. Anything and everything of which we may conceive, including mathematics, logic, brains, computers, trees, houses, and so on, may be seen as information. We cannot learn about the world without absorbing information and any inborn knowledge that we may have may also be seen as information.

Information that is totally random---without any detectable redundancy---is totally lacking in structure. All persistent entities, concepts, laws, theorems or regularities, including the concepts of mathematics and logic, represent redundancy or structure in information. Any kind of repeatable calculation or deduction is a manifestation of redundancy in the substance of mathematics and logic. And information compression is the mechanism by which we detect and manipulate these redundancies.

In short, it is difficult to escape from the conclusion that concepts in mathematics and logic are intimately related at a fundamental level to concepts of information, redundancy, structure and the compression of information.

\section{Information compression and structures in mathematics and logic}\label{ml_structures}

The preceding parts of this chapter have prepared the ground for what is really the main 
substance of the chapter:

\begin{itemize}

\item This section describes the idea that notations and structures in mathematics and logic may be seen as 
a set of devices for representing information in a compressed form. 

\item The next section presents the idea that processes of mathematical and logical 
inference may also be understood in terms of information compression.

\end{itemize} 

As was noted earlier, the arguments and examples to be presented are not intended 
as any kind of proof of the proposition that mathematics and logic is based on information compression. They are merely suggestive indications of what may turn out to be a fruitful way to view mathematics, logic and related disciplines---an avenue to new insights and understanding.

Within this main section, Section \ref{ml_suggestive_evidence} considers some general features of mathematics and logic that suggest that there is a close connection between mathematics and logic and information compression. Next, we consider some aspects of number systems that seem to relate to information compression. Then in Sections \ref{ml_chunking_with_tags}, \ref{ml_schema_plus_correction} and \ref{ml_run_length_coding}, we consider how the information compression principles of chunking-with-codes, schema-plus-correction and run-length coding may be seen to operate in the forms and structures of mathematics and logic. Section \ref{ml_modelling_in_sp} considers some of the ways in which these concepts may be modelled in the SP system.

\subsection{Suggestive evidence}\label{ml_suggestive_evidence}

Considering, first, the idea that notations and structures in mathematics and logic might be seen as 
devices for representing information in a compressed form, a general indication that 
something like this might be true is provided by three aspects of mathematics discussed in 
the next three subsections.

\subsubsection{Mathematics as the ``language of science''}

In the context of science as a search for succinct descriptions of the world, it is often 
remarked that mathematics provides a remarkably and `mysteriously' convenient way to 
express scientific truths. Thus \citet{penrose_1989}, who takes a Platonic view of mathematics, 
writes: 

\begin{quotation}

\noindent ``It is remarkable that {\em all} the SUPERB theories of Nature have proved to be extraordinarily fertile as sources of mathematical ideas. There is a deep and 
beautiful mystery in this fact: that these superbly accurate theories are also 
extraordinarily fruitful simply as {\em mathematics}.''
(pp. 225--226, emphasis as in the original).

\end{quotation} 

\noindent In a similar vein, \citet{barrow_1992} writes: 

\begin{quotation}

\noindent ``For some mysterious reason mathematics has proved itself a reliable guide 
to the world in which we live and of which we are a part. Mathematics 
works: as a result we have been tempted to equate understanding of the 
world with its mathematical encapsulization. ... Why is the world found to 
be so unerringly mathematical?.'' (Preface, p. vii).

\end{quotation} 

\noindent This last question is partly answered later: 

\begin{quotation}

\noindent ``Science is, at root, just the search for compression in the world. ... In short, the world is surprisingly compressible and the success of mathematics in 
describing its workings is a manifestation of that compressibility.'' ({\em ibid.}, p. 
247).

\end{quotation}

Thus he recognises that mathematics provides a good medium for expressing scientific 
observations in a compressed form but he stops short of making the connection, which is 
the subject of this chapter, between mathematical concepts and standard concepts of 
information and information compression.

\subsubsection{The nature of mathematical `functions'}\label{maths_functions}

\index{function|(}

A mathematical `function' is a mapping from one set ($X$) to another set ($Y$) and, in 
the `strict' form of a function, each element of $X$ is mapped to at most one element of $Y$ 
\citep[][p. 7]{sudkamp_1988} although it is sometimes useful to consider more `relaxed' kinds of `function' where each element of $X$ may be mapped to more than one element of $Y$. In essence, a function is a table with one or more columns representing the `input' and one or more columns representing the `output'. Thus an element of set $X$ may be one or more cells from one row of the table and the corresponding element of set $Y$ is the complementary set of one or more cells from the same row of the table.

Very simple functions, such as the exclusive-OR function, can be and often are represented literally as a table. This and other examples will be considered later. But, in the vast majority of cases, the definition of a function (the `body' of the function in computer-science jargon) does not look anything like a table. Nevertheless, whatever the definition of a function might look like, all functions behave as if they were `look up' tables that yield an output value or set of values in response to an appropriate input.

The key point here is that, with regard to the more complex kinds of functions that 
do not look like a table, they are almost invariably smaller than the abstract table they 
represent and, in most cases, they are smaller by a large factor.

This feature of the more complex kinds of functions means that they can provide a 
very effective means of compressing information. Consider, for example, a function 
relating the distance travelled by an object to the time since the object was dropped from a high place and allowed to fall freely under the influence of gravity (ignoring the effect of resistance from air). If the function is represented as a table and if there is an entry in the table for each microsecond or nanosecond, then for a fall of 100 metres, say, the table will be very large. By comparison, Newton's formula, $s = (gt^{2}) / 2$ (where $s$ is the distance travelled, $g$ is the acceleration due to gravity and $t$ is the time since the object was allowed to fall), is extremely compact.

This feature of the more complex kinds of function goes a long way to explaining 
why mathematics is such an effective ``language of science''. However, as we shall see, 
there is more to be said than this about information compression and the nature of functions.%
\index{function|)}

\subsubsection{The nature of counting}\label{nature_of_counting}

\index{counting|(}

Given its close association with the number system, the process of counting things is clearly central in mathematics.

A little reflection shows that it is not possible to count things without some kind of process of recognition. If we are asked to count the number of apples in a bowl of fruit, we need to be able to recognise each apple and distinguish it from other fruits in the bowl---oranges, bananas and so on. Even if we are asked to count the number of `things' in some location, we need to recognise each one as a `thing' and distinguish it from the environment in which it lies.

Recognition is a complex process which can be applied at various levels of abstraction. But notwithstanding this complexity it seems clear that it necessarily involves a process of matching patterns. And, if the things to be counted are all to be assimilated to a single concept (e.g., `apple' or `thing' or other category), this implies the unification of matching patterns.

Given the connection between information compression and the matching and unification of patterns (described in Section \ref{ic_repetition_of_patterns}), it seems that counting involves a process of compressing information. Given the central importance of counting in mathematics, this is one strand of evidence pointing to the importance of information compression in mathematics.%
\index{counting|)}

\subsection{Chunking-with-codes}\label{ml_chunking_with_tags}

\index{information!compression!chunking-with-codes|(}

As we saw in Section \ref{techniques_for_ic}, the idea of giving a name, `tag', `code' or `identifier' to a relatively large chunk of information is a means of compressing information that we use so often and that comes so naturally that we scarcely realise that we are compressing information.

In mathematics and logic, as in many other areas, names are widespread. We use names for mathematical and logical functions, for sets and members of sets, for numbers, for variables, for matrices and for geometric forms. In computer programming, names are used for functions and subroutines, for methods, classes and objects (in object-oriented design), for arrays and variables, for tables and columns (in databases) and, in the bad old days of spaghetti programming, names were used copiously for individual statements or lines where the program could `go to'.

In almost every case, it seems that a name has the role of being a relatively short 
code representing a relatively large chunk of information. As such, most or all of these 
names contribute to the achievement of information compression.

In the following subsections we shall consider these ideas in a little more detail.%

\subsubsection{The concept of a discrete object in mathematics and logic}

\index{object|(}

Historically, it seems likely that the concept of number originated with the need to 
keep a tally on concrete objects like sheep, goats, oranges etc. Arguably, the earliest 
number system was the system of unary natural numbers where $1 = 01, 2 = 011, 3 = 
0111$ etc and other number systems were invented or discovered later as a generalisation of that simple kind of counting.\footnote{A concept of zero was, historically, a relatively late development.}

Whether or not that historical speculation is accepted, most people will accept that 
there is a connection between the concept of natural numbers, especially unary numbers, 
and the concept of discrete objects in the world. The latter do not depend on the former (because animals---and people without any knowledge of numbers---can recognise discrete objects). But each stroke in a unary number is itself recognised as a discrete entity and, even if we are using numbers with bases higher than one, we know that it is often convenient to make a one-to-one association between strokes in the underlying unary number and objects in the real world.

In a similar way, the concept of a set, with elements of the set being discrete entities 
(some of which may themselves be sets), provides a link with the way we see the world as 
composed of discrete objects.

Thus the concept of a discrete object has a significant role in both the concept of 
number and of a set. And the concepts of number and set are both key parts of mathematics and logic. Bearing in mind that seeing the world in terms of discrete objects is almost certainly a reflection of the way our perceptions and concepts are governed by information compression principles (Section \ref{objects_section}), it is apparent that information compression plays a significant part in the foundations of mathematics and logic.%
\index{object|)}

\subsubsection{Chunking-with-codes in number systems?}

With regard to names or codes as relatively short identifiers for chunks, let us begin 
with a type of name which, at first sight, does not appear to have a role in information compression. It can be argued that the names (words or numerals) that we give to numbers merely serve to 
differentiate one number from another and are {\em not} codes for relatively large chunks of 
information. We may argue that this is true even for names like `one billion' or `$10^9$' that represent relatively large numbers and would take a lot of space if they were written out using unary arithmetic or even binary arithmetic. The reason we may give is that the base of a number (unary, binary etc) is irrelevant to the calculation of the amount of information it represents so that any number in any base is totally equivalent to the same number in another base in terms of information, apart from errors arising from any rounding of numbers that we may choose to make.

This line of reasoning is valid for conversions between two numbers both of which 
have a base of 2 or more, but it appears to be wrong for conversions between numbers 
where one of the two numbers is unary. 

Although unary numbers are conventionally considered to have a base of 1, there is 
a sense in which they have no base at all. Each stroke in a unary number seems to be a 
primitive representation of a discrete entity and the stroke cannot be analysed into anything 
simpler.

By contrast, any number that uses a base greater than $1$ introduces a new principle: 
the idea that groups of symbols may be given a relatively short name or code, and that this may be done recursively through any number of levels. In the decimal system, for example, $8$ is a relatively short name for the unary number `$0 1 1 1 1 1 1 1 1$'. At this point, a sceptic might object that, to accommodate decimal digits from $0$ to $9$, each one must be assigned at least $\lceil \log_2 10\rceil = 4$ bits of space in a computer memory and that this is equivalent to the amount of information in the corresponding unary number.

In the case of the decimal digits from $0$ to $9$, this objection may carry some force. 
But it appears not to be valid when we consider numbers of $10$ or more. The decimal 
number $100$, for example, uses digits within the range $0$ to $9$ and so, on the assumption that we are not needing to represent anything other than numbers, we could accommodate the three digits in `$100$' with only $3 \times 4 = 12$ bits (or, if we started counting from $0$ and thus finished counting at $99$, we could accommodate the number with only $2 \times 4 = 8$ bits). This is very much less than the number of strokes in the corresponding unary number and thus, {\em prima facie}, seems to be a clear case of information compression.

In Section \ref{randomness_redundancy_structure} we saw that lossless compression of $I$ is only possible if we can detect redundancy in $I$. Where is the redundancy in unary arithmetic that might allow us to use the chunking-with-codes technique as suggested above? The answer seems to lie in the apparent need to provide a symbol (normally `$0$') to represent zero in the unary number system.\footnote{Apart from the need to represent the concept of zero, a system containing nothing but unary numbers would need some means of marking the beginnings and ends of numbers. This can be achieved if every number starts with `0'.} If the symbols `$0$' and `$1$' are both provided in the unary number system then it is clear that in all the many unary numbers where the number of `$1$'s is large, the probability of `$1$' is very much higher than `$0$' and so, in accordance with Hartley-Shannon information theory, these numbers generally contain considerable amounts of redundancy. Thus the conversion of any unary number (except very small ones) into the equivalent number using a base of $2$ or more seems to be a genuine example of chunking-with-codes and seems to achieve genuine information compression.

\subsubsection{Grammars and re-write rules}

As we saw in Section \ref{grammar_section}, the notational conventions that are used in grammars may be regarded as a set of devices that may be used to encode information in an economical form. The information on the right-hand side of a re-write rule may be seen as a `chunk' of information and the symbol or symbols on the left-hand side may be seen as a relatively short label or identifier for that information.

\subsubsection{Named functions and operators}\label{named_functions}

Again, as we saw in Section \ref{computer_programs_section}, the `body' or definition of a `function' or `operator' in a computer program may be seen as a relatively large chunk of information and its name may be seen as a relatively short identifier for the chunk. Very much the same can be said about functions in mathematics and, as was noted in Section \ref{logic_programming_section}, about clauses in logic programming.%
\index{information!compression!chunking-with-codes|)}

\subsection{Schema-plus-correction}\label{ml_schema_plus_correction}

\index{information!compression!schema-plus-correction|(}

This subsection considers some constructs in mathematics and logic that seem to be examples of the 
information compression principle of schema-plus-correction.

\subsubsection{Structures containing variables}\label{sp_variables}

The concept of a `variable' is widespread in mathematics, logic, computer 
programming and theoretical linguistics. In all its uses it is understood as some kind of 
`receptacle' that may receive a `value' by `assignment' or by `instantiation'. In many 
systems, each variable has a `type' with a `type definition' that defines the range of values that the variable may receive.

In the majority of cases, a variable has a name, although unnamed variables are used 
sometimes in systems like Prolog. If the size of the value of a named variable is relatively large compared with the name (and values in Prolog can be very large, complex structures), then, in some cases, the name may be understood as a `code' and the value may be seen as a corresponding chunk so that the two together may be seen to make a contribution to information compression via the chunking-with-codes technique. But in many cases, the value and the name are similar in size or the value is smaller than the name. Surely, in cases like these, or in cases where the variable has no name, there cannot be any contribution to information compression?

The answer to this question seems to be much as was described in Section \ref{functional_and_structured_programming}. A structure that contains variables may usually be seen as a schema, the variables may be seen as `holes' or `slots' in the schema and the values may be seen as `corrections' or `completions' of the schema. In cases like these, the schema may often be seen as a unification of two or more patterns with a corresponding gain in information compression. In short, the contribution that variables can make to information compression is not normally a function of the variables in themselves but because they form part of a larger structure---a schema---that contributes to information compression. Some examples are described in the subsections that follow.

\subsubsection{Functions with arguments or parameters}

As we noted in Section \ref{named_functions}, a function need not be a monolithic chunk of information: most functions allow for variations in the data to which they will be applied by providing `arguments' or `parameters' that may take different values on different invocations of the function.

This kind of function may be seen as a schema with holes or slots (the arguments or parameters) that may be filled with a variety of `corrections' or `completions' on different occasions. Provided the function is called two or more times in a mathematical or logical text, it can be seen as a means of avoiding undue redundancy in that text.

\subsubsection{Classes, subclasses and objects}\label{classes_subclasses_objects}

As we saw in Section \ref{oop_section}, any one `class' in object-oriented programming or object-oriented design may be seen as a schema and the additional details provided in subclasses of that class or objects derived from that class may be seen as `corrections' or completions of the schema. This organisation allows software to be represented in a compressed form because the attributes of a class need be recorded only once instead of being repeated in each member of the class.

\subsubsection{Sets}

Perhaps the closest relative of the OO concepts just described in mathematics and 
logic is the concept of a set.

A set can be defined extensionally---by listing the elements in the set (e.g., $\{sun, 
wind, rain\}$)---or it may be defined intensionally---by listing the defining characteristics of the class (e.g., \{$n | n$ is prime\}). An intensionally-defined set is similar to a class as described in Section \ref{class_part_inheritance} and, when one set includes one or more other sets recursively through an arbitrary number of levels, the structure is similar to a class hierarchy as described above. But, to my knowledge, little or no attempt has been made to develop concepts like inheritance of attributes or the instantiation of a class by one or more objects.

With regard to information compression, an intensionally defined set that describes two or more 
elements may be regarded as a compressed representation of the relevant attributes of those elements.

\subsubsection{Propositions in logic with the universal quantifier}

Here are two examples of propositions expressed in English and in logical notation:

\begin{itemize}

\item \raggedright ``All humans are mortal.''

\indent $\forall x: human(x) \Rightarrow mortal(x)$.

\item ``For any integer which is a member of the set of natural numbers ($N$) and which is 
even, it is also true that the given integer with 1 added is odd.''

\begin{center}
$\forall i\ \epsilon\ N \cdot is\_even(i) \Rightarrow is\_odd(i + 1)$.
\end{center}

\end{itemize}

Each example is a proposition containing a universal quantifier and a variable (all three 
instances of $x$ refer to a single variable and likewise for $i$). The first statement is a 
generalisation about what it is to be human and the second one is a generalisation about 
natural numbers. Both of them represent repeating (redundant) patterns. Each may be 
regarded as a schema rather than a chunk because, notwithstanding its status as `bound', the variable in each example can receive a variety of values.%
\index{information!compression!schema-plus-correction|)}

\subsection{Run-length coding}\label{ml_run_length_coding}

\index{information!compression!run-length coding|(}

When something repeats in a series of instances, each one contiguous with the next, we find it very natural to find some way to reduce the repetition to a single instance with something to mark the repetition. As noted previously, lossless compression is achieved if the number of 
repetitions is marked; otherwise the information compression is lossy.

This section identifies some examples of run-length coding and further examples will be described in Section \ref{ml_processes}.

\subsubsection{Iteration and recursion in computer programming}\label{iteration_and_recursion}

In computer programming, repetition of a part of a program can be shown using 
forms like a {\em while} loop or a {\em for} loop or a {\em repeat ... until} loop. Generally speaking, there is some change in the value of one or more variables on each iteration of the block of program. So this kind of repetition may be regarded as a combination of run-length coding and schema-plus-correction.

As we have seen earlier (Section \ref{techniques_for_ic}), repetition can also be expressed with recursion (in the sense understood in computer programming). Here is an example (in the C programming language) of the factorial function:

\begin{center}
\begin{BVerbatim}
int factorial(int x)
{
     if (x == 1) return(1) ;
     return(x * factorial(x - 1)) ;
}
\end{BVerbatim}
\end{center}

\subsubsection{Repetition in mathematics}

Mathematics uses a variety of ways to show repetition in a succinct manner:

\begin{itemize}

\item Multiplication is a shorthand for repeated addition.

\item The power notation (e.g., $10^9$) is a shorthand for repeated multiplication.

\item A factorial (e.g., $10!$) is a shorthand for repeated multiplication and subtraction.

\item The bounded summation notation (`$\sum$') and the bounded power notation (`$\prod$') are 
shorthands for repeated addition and repeated multiplication, respectively. In both 
cases, there is normally a change in the value of a variable on each iteration, so these 
devices may be seen as a combination of run-length coding and schema-plus-correction.

\item As in our previous examples, repetition of a function may be achieved by a direct 
or indirect `recursive' call to the function.

\end{itemize}

As we noted previously (Section \ref{techniques_for_ic}), the use of shorthands is so automatic and `natural' that it is easy to overlook the fact that they are examples of information compression.%
\index{information!compression!run-length coding|)}

\subsection{Modelling these concepts in the SP system}\label{ml_modelling_in_sp}

If it is accepted that our three basic techniques for information compression can be modelled in the SP system
(as outlined in Section \ref{compression_in_sp}) and if it is accepted that many forms and structures in mathematics and logic may be understood in terms of information compression (as described in the preceding parts of this main section), then one might expect that many of the forms and structures of mathematics and logic could be modelled in the SP system. The majority of examples will be seen in Section \ref{ml_processes}. As a preparation for these later examples, this subsection focuses on some basic structures in mathematics and logic and how they may be modelled in the SP system.

\subsubsection{Variable, value and type definition}\label{variable_value_type_definition}

\index{variable|(}\index{value|(}\index{type definition|(}

The basic idea of a named variable can be represented quite simply in the SP system by 
a pair of symbols that are contiguous within a pattern. In principle, any pair of symbols 
may be seen to function as a variable but in most examples we shall be considering the left 
and right symbols will correspond to the initial and terminal symbols of patterns. Thus, in 
Figure \ref{alignment_figure_1}, each of the pairs of symbols `N \#N' and `V \#V' in the pattern `S N \#N V \#V \#S' may be seen to function as variables with the first symbol of each pair acting as a name for the variable and the second serving to mark the end of whatever `value' goes between the two symbols. These `values' are patterns representing a noun in the case of `N \#N' and a verb between the symbols `V' and `\#V'. In a similar way in Figure \ref{expression_parsing}, each of the pairs of symbols `F \#F' and `R \#R' in the pattern `F 2 ( F \#F R \#R F \#F ) \#F' function as a variable that can take other patterns in the grammar as values.

In these examples, each variable may be seen to have a `type definition' in the form 
of the set of patterns that has initial and terminal symbols that match the left and right 
symbols of the variable. For the variable `N \#N', the type definition is the two `noun' 
patterns that can fit in that slot and for the variable `V \#V' the type definition is the two 
patterns for verbs. Of course, in a realistic system, those two types would have many more 
members.

Although the examples shown seem to capture the essentials of the concepts of 
`variable', `value' and `type definition', they lack two of the features that are commonly 
found in computer systems:

\begin{itemize}

\item {\em Scoping}: In most computer systems, a variable can have a `scope' meaning a part 
of a program within which repeated instances of a variable name refer to a single 
variable. 

\item {\em Assignment of a type to a variable}: In many computer systems, any variable can be 
assigned a type that is independent of its name. This contrasts with the examples 
given above where the name of the variable is also the name of its type definition.

\end{itemize}

\noindent It seems likely that these two ideas can be modelled in the SP system but not in the SP61 
model in its current state of development.%
\index{variable|)}\index{value|)}\index{type definition|)}

\section{Information compression and processes in mathematics and logic}\label{ml_processes}

The previous section presented a range of examples showing that many notations 
and forms in mathematics and logic may be seen as devices for representation of mathematical or logical concepts in a succinct 
form. This section presents a range of examples in support of the idea that, very often, the 
computational or inferential {\em processes} in mathematics and logic may also be understood in terms of information compression. More 
specifically, it is suggested that, very often, they may be understood in terms of the SP system.

As we have seen in Chapter \ref{computing_chapter}, the concept of `computing', as defined by the structure and functioning of the universal Turing machine and Post canonical system, may be understood in terms of the SP system. Section \ref{unary_numbers_and_sp} of that chapter showed how the SP system may model a Post canonical system for the creation and recognition of unary numbers. The arguments and examples here may be seen as a further development of the theme in other areas of mathematics, logic and related disciplines.

\subsection{Sets}

\index{set|(}

This subsection considers briefly how some of the basic features of sets may be 
understood in terms of the alignment and unification of patterns.

\subsubsection{Creating a set from a bag or multiset}

The concept of `New' as it has been described in this chapter allows two or more 
patterns within New to be identical and, in that respect, New is like a `bag' or `multiset' in 
logic.

\sloppy If New were to contain a range of one-symbol patterns like this: \{(A)(B)(C)(D)(A)(D)(B)(A)(C)(C)(A)(C)\} (or if New were to contain multi-symbol patterns that repeated in a similar way) it is not hard to see how, with Old initially empty, the alignment and unification of patterns in a process like that outlined in Section \ref{overall_framework} would reduce New to a set of patterns in Old something like this: \{({\bf A})({\bf B})({\bf C})({\bf D})\}.

Thus, the process of converting a bag into the corresponding set containing a single 
example of each type of element in the bag may be seen as a process of information 
compression by the alignment and unification of patterns.%

\subsubsection{Union and intersection of sets}

If the set \{(B)(C)(D)(F)(G)\} were in New and the set \{(A)(B)(C)(E)(F)\} were 
the entire contents of Old, it is not hard to see that, in much the same way as just described, the result would be: \{(A)({\bf B})({\bf C})(D)(E)({\bf F})(G)\}. This set is the {\em union} of the first two sets and their intersection is the one-symbol patterns show in bold type---the set \{({\bf B})({\bf C})({\bf F})\}.

In general, we can see that the union and intersection of two sets may be seen as information compression 
by alignment and unification amongst the elements of the sets.

\subsubsection{Intensionally-defined sets}

An intensionally-defined set such as \{$x | x > 1$\} may be regarded as a 
representation of a set in compressed form. Unless the set has only a few members (or is 
empty), this representation is clearly more economical than the extensional definition 
that itemises every element of the set. The process of deriving an intensional definition 
from the corresponding extensional definition may be seen as the alignment and unification 
of the defining attributes in the elements of the set.%
\index{set|)}

\subsection{Execution of simple functions}

\index{function|(}

The concept of a `function', that we have already considered as an example of 
chunking-with-codes (Sections \ref{techniques_for_ic} and \ref{named_functions}) and as a mediator of information compression (Section \ref{maths_functions}), is one of the most distinctive features of mathematics and logic, especially in mathematics and computer programming.

It is often convenient to think of a function as some kind of `black box' that can 
receive some kind of `input' and, after some processing, will deliver some `output'. As we 
noted in Section \ref{maths_functions}, a function can, in principle, be regarded as a lookup table and, in simple cases, functions can often be implemented in that way. But in more complex cases, a simple lookup table would not be appropriate, either because the size of the table would be finite but too large to be practical or because the table would be infinitely large.

The remainder of this subsection describes with an example how the execution of 
simple functions may be modelled in the SP system and the next subsection considers how the 
execution of the more complicated kinds of functions may be understood within the SP system.

\subsubsection{A simple function as a look-up table: a one-bit adder}

Where a simple function is structured as a table, it can be processed by searching 
for a row in the table where the input matches the input side of the row and then reading off 
the corresponding output.

In the SP system, a table may be represented with a set of patterns like those shown in 
Figure \ref{half_adder_definition}. These patterns define the function for the addition of two one-bit numbers in binary arithmetic. The two digits immediately following `A' in each row are the two one-bit numbers to be added together, the bit following `S' is the sum of the two input bits, and the bit following `C' is the {\em carry-out} bit, to be carried to the next column. Thus, reading the results in each pattern from right to left and writing them in the usual manner, the four possible results are `1 0', `0 1', `0 1' and `0 0', in order from the top row to the bottom.

\begin{figure}[!hbt]
\centering
\begin{tabular}{l}
A 1 1 S 0 C 1 \\
A 1 0 S 1 C 0 \\
A 0 1 S 1 C 0 \\
A 0 0 S 0 C 0 \\
\end{tabular}
\caption{Four patterns representing a table to define the function for the addition of two one-bit numbers in binary arithmetic, with provision for the carrying out of one bit. The 
meanings of the letters are described in the text.}
\label{half_adder_definition}
\end{figure}

Apart from their function in conveying the meanings of the bits to human readers, 
service symbols (like `A', `S' and `C') are needed to constrain the search for possible 
multiple alignments and, in effect, to tell the system which digits represent the `input' digits and 
which are `output'.\footnote{A distinction between `content' symbols and `code' symbols provides additional constraint but the details are not relevant to the subject of this chapter and need not be spelled out here.}

Figure \ref{one_bit_adder_alignment} shows the best multiple alignment that has been found by SP61 with the pattern `A 0 1 S C' in New and the patterns shown in Figure \ref{half_adder_definition} in Old. (Notice that the first and last digit in each row is a row number, not part of the pattern for that row.) Generally speaking, in the SP system the `result' of a multiple alignment is the symbols in patterns from Old in the multiple alignment that have not been aligned with any symbols from New. The result in this case is `1' for the sum bit and `0' for the carry-out bit. In a very straightforward way, the matching process has functioned as a process of `table lookup' and has `selected' the pattern in Figure \ref{half_adder_definition} that yields the `correct' addition of `0' and `1' in binary arithmetic.

\begin{figure}[!hbt]
\centering
\begin{BVerbatim}
0 A 0 1 S   C   0
  | | | |   |  
1 A 0 1 S 1 C 0 1
\end{BVerbatim}
\caption{The best multiple alignment found by SP61 with `A 0 1 S C' in New and the patterns shown in Figure \ref{half_adder_definition} in Old. The first and last digit in each row is a row number, not part of the pattern for that row.}
\label{one_bit_adder_alignment}
\end{figure}

In general, any function that can be represented as a look-up table can be 
represented in the SP system as a set of patterns and processed in essentially the same way as 
the example of one-bit addition shown here.

\subsection{Composite functions and their execution}

We saw in Section \ref{maths_functions} that, although all functions in mathematics and logic are defined abstractly as a table of inputs and outputs, many functions do not look much like a table and are very much more compact than the table that they represent. In later parts of Section \ref{ml_structures} we have seen that the succinctness of many functions, and the succinctness of mathematics and logic in general, can often be attributed to the exploitation of our three simple compression techniques: chunking-with-codes, schema-plus-correction and run-length coding.

So far, all of this seems to be reasonably straightforward. A complication, however, 
is that when two or more functions are part of a higher-level function, or when there is a 
repetition of a function two or more times, it is often necessary for the output of one 
function to become the input of another function or for the output of a function to become 
the input for another invocation of the same function. Does this passing of information 
between functions represent any kind of information compression and, in particular, can it be modelled with 
the SP system?

The next subsection shows how, in a simple case, it can be modelled in the SP system. 
And the following subsection considers this issue in more general terms.

\subsubsection{Adding two-bit numbers or larger}\label{adding_two_bit_numbers}

Conceptually, the simplest way to add together a pair of numbers containing two or 
more bits is to apply the one-bit adder recursively to successive bits in each of the two 
numbers which are to be added. And this means that it should be possible for any carry-out 
bit from one application of the adder to become the {\em carry-in} bit for the next application of the adder in the next higher column of the addition.

In the jargon of computer engineering, a one-bit adder like the one shown in Figure \ref{half_adder_definition} is called a {\em half adder} \citep[see][pp. 83--89]{richards_1955} because it lacks any means of receiving a carry-in bit from the column immediately below. This deficiency is made up in a {\em full adder} \citep[see][pp. 89--95]{richards_1955}, a version of which is represented by the set of patterns shown in Figure \ref{full_adder_definition}. Here, the `C' bit at the beginning of each pattern is the carry-in bit and all the other bits are the same as in the half adder (Figure \ref{half_adder_definition}).

Notice that the pair of symbols `C ... \#C' marking the carry-in bit at the beginning 
of each pattern is the same as the symbols `C ... \#C' that mark the carry-out bit at the end 
of the pattern. Matching of these pairs of symbols allows the carry-out bit at any one level to become the carry-in bit at the next higher level so that values can, in effect, be passed from level to level. This can be seen in the example described next.

Consider, for example, how we may add together two three-bit numbers like `1 1 0' 
and `0 1 1'. The two numbers need to be arranged so that the bits for each level lie together 
and are marked in the same way as in the New pattern in Figure \ref{one_bit_adder_alignment}. The order needs to be reversed to take account of the way the SP61 model searches for multiple alignments.\footnote{In principle, SP61 can search for good multiple alignments by processing all the symbols in New together. But in practice, some types of good multiple alignment (including the ones described here) are much easier to find if the symbols from New are processed one at a time in left-to-right order. And, in addition, it is natural to process columns in sequence from the lowest order to the highest.} Thus the sequence of symbols for these two numbers should be: `A 1 0 S A 1 1 S A 0 1 S'. However, the symbols `C 0 \#C' need to be added at the beginning of New to ensure that the carry-in bit for the first addition is `0'. So the final result is `C 0 \#C A 1 0 S A 1 1 S A 0 1 S'.

The best multiple alignment found by SP61 with `C 0 \#C A 1 0 S A 1 1 S A 0 1 S' in New 
and the patterns shown in Figure \ref{full_adder_definition} in Old is shown in Figure \ref{three_bit_addition_alignment}. Ignoring the `service' symbols, the sum of the two numbers can be read from this multiple alignment as the three `sum' bits (each one between an `S' and a `C' column), with the order reversed, and preceded on the left by the last carry-out bit on the right of the multiple alignment. The overall result is `1 0 0 1'. Readers will verify easily enough that this is the correct sum of the binary numbers `1 1 0' and `0 1 1'.

\begin{figure}[!hbt]
\centering
\begin{tabular}{l}
C 1 \#C A 1 1 S 1 C 1 \#C \\
C 1 \#C A 0 1 S 0 C 1 \#C \\
C 1 \#C A 1 0 S 0 C 1 \#C \\
C 0 \#C A 1 1 S 0 C 1 \#C \\
C 1 \#C A 0 0 S 1 C 0 \#C \\
C 0 \#C A 0 1 S 1 C 0 \#C \\
C 0 \#C A 1 0 S 1 C 0 \#C \\
C 0 \#C A 0 0 S 0 C 0 \#C \\
\end{tabular}
\caption{Eight patterns representing a table to define the function for the addition of two one-bit numbers in binary arithmetic, with provision for the carrying in of a digit as well as the carrying out of a digit. The meanings of the symbols are described in the text.}
\label{full_adder_definition}
\end{figure}

\begin{figure}[!hbt]
\fontsize{10.00pt}{12.00pt}
\centering
\begin{BVerbatim}
0 C 0 #C A 1 0 S          A 1 1 S          A 0 1 S          0
  | | |  | | | |          | | | |          | | | |         
1 C 0 #C A 1 0 S 1 C 0 #C | | | |          | | | |          1
                   | | |  | | | |          | | | |         
2                  C 0 #C A 1 1 S 0 C 1 #C | | | |          2
                                    | | |  | | | |         
3                                   C 1 #C A 0 1 S 0 C 1 #C 3
\end{BVerbatim}
\caption{The best multiple alignment found by SP61 with the pattern `C 0 \#C A 1 0 S A 1 1 S A 0 1 S' in New and the patterns shown in Figure \ref{full_adder_definition} in Old.}
\label{three_bit_addition_alignment}
\end{figure}

\subsubsection{Other composite functions}

Compared with conventional systems, the method just shown for passing 
information from one invocation of a function to another is somewhat inflexible. This is 
because it requires the `output' variable to have the same name as the `input' variable. In 
conventional systems there is a variety of notations for showing that the output of one 
function becomes the input to another and none of them are restricted in this way. 

Modelling these more flexible methods in the SP system probably requires an ability to 
model `scoping' of variables and, possibly, an ability for `learning'. Both these things are beyond what the SP61 model can do as it has been developed to date.%
\index{function|)}

\subsection{Propositional logic and information compression}\label{propositional_logic_and_ic}

Figure \ref{xor_definition} shows four patterns representing a `truth table' for the `exclusive OR' logical proposition ($p \vee q$. ``Either $p$ is true or $q$ is true but they are not both true''). Here, `1' means TRUE and `0' means FALSE, the first two digits in each pattern are the truth values of $p$ and $q$ (individually) and, for each pattern, the last digit, between the symbols `R .. \#R' , is the truth value of $p \vee q$.

Readers will notice that the XOR truth table is very similar to the definition of 
one-bit addition shown in Figure \ref{half_adder_definition}, not merely in the arrangement of the digits but, more importantly, in the fact that both things can be represented as a simple lookup table that can be represented by a set of patterns. It should be no surprise that the XOR function can be evaluated by SP61 in essentially the same way as with the function for one-bit addition.

\begin{figure}[!hbt]
\centering
\begin{tabular}{l}
XOR 1 1 R 0 \#R \\
XOR 0 1 R 1 \#R \\
XOR 1 0 R 1 \#R \\
XOR 0 0 R 0 \#R \\
\end{tabular}
\caption{Four patterns representing the `exclusive OR' (XOR) truth table as described in the text.}
\label{xor_definition}
\end{figure}

Again, one might expect to be able to evaluate combinations of simple propositions 
in the SP system in essentially the same way as was done for the addition of numbers containing 
two or more bits (Section \ref{adding_two_bit_numbers}). As before, this can be achieved with the SP61 model in its current stage of development if outputs are made to match inputs. This can be seen in the definition of `NOTXOR' shown in Figure \ref{notxor_definition}. Here, the alternative `outputs' of the XOR proposition are contained in the field `A ... \#A' and the same two symbols are used to mark the possible `inputs' of the truth table for NOT (that changes `1' to `0' and {\em vice versa}).

\begin{figure}[!hbt]
\centering
\begin{tabular}{l}
NOTXOR XOR NOT R \#R \#NX \\
XOR 1 1 A 0 \#A \\
XOR 1 0 A 1 \#A \\
XOR 0 1 A 1 \#A \\
XOR 0 0 A 0 \#A \\
A 1 \#A NOT R 0 \#R \\
A 0 \#A NOT R 1 \#R \\
\end{tabular}
\caption{A set of patterns defining the `not XOR' proposition as described in the text.}
\label{notxor_definition}
\end{figure}

\begin{figure}[!hbt]
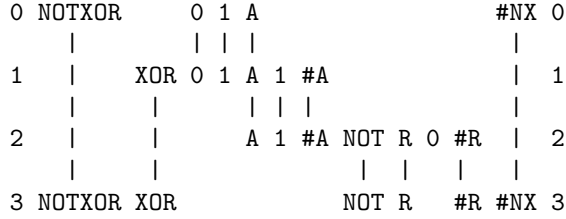

\centering
\begin{BVerbatim}
0 NOTXOR     0 1 A                 #NX 0
    |        | | |                  | 
1   |    XOR 0 1 A 1 #A             |  1
    |     |      | | |              | 
2   |     |      A 1 #A NOT R 0 #R  |  2
    |     |              |  |   |   | 
3 NOTXOR XOR            NOT R   #R #NX 3
\end{BVerbatim}
\caption{The best multiple alignment found by SP61 with `NOTXOR 0 1 A \#NX' in New and the 
patterns shown in Figure \ref{notxor_definition} in Old.}
\label{notxor_alignment}
\end{figure}

\sloppy{Figure \ref{notxor_alignment} shows the best multiple alignment found by SP61 with the pattern `NOTXOR 0 1 A \#NX' in New and the patterns from Figure \ref{notxor_definition} in Old. In the multiple alignment, the digits `0' and `1' in New yield `A 1 \#A' as the `result' of the XOR component. These three symbols match the first three symbols in the first of the two `NOT' patterns and this yields `R 0 \#R as the overall `result' of the NOTXOR proposition.}

As in the case of adding numbers containing two or more digits, the SP 
patterns that have been shown are somewhat contrived owing to the need to make the 
output at one stage match the input at the next stage. It is anticipated that more elegant 
solutions may be found when the SP61 model has been further developed.

\subsection{True and false in the SP system}\label{true_false_sp}

Although readers may accept that, in broad terms, the matching and unification of patterns in the SP system can model the kinds of inferences that can be made in propositional logic, an objection may be raised that, in propositional logic {\em per se}, every inference has only two values, TRUE or FALSE, whereas in the SP system, there may be and often are alternative multiple alignments for a given inference that suggests that there are uncertainties attaching to the inferences that the system makes.

The SP concepts apply most naturally to probabilistic reasoning (Chapter \ref{pr_chapter}) and this suggests that `exact' reasoning may be seen as reasoning where probabilities are very close to 0 or 1. In itself, this idea is not satisfactory owing to uncertainties about the precise meaning of `close to' in this context. A better idea is to use explicit TRUE and FALSE values in `output' fields (or `0' and `1' values as in the examples we have been considering), coupled with a focus always on the `best' multiple alignment for any set of patterns. If multiple alignments are created and interpreted in this way then the effect of two-valued reasoning can be achieved within the SP system without ambiguity. An example is shown in the next subsection.

\subsection{Syllogistic reasoning}\label{syllogistic_reasoning}

\index{reasoning!syllogistic|(}

Consider the following text book example of a {\em modus ponens} syllogism:

\begin{enumerate}

\item All humans are mortal.

\item Socrates is human.

\item Therefore, Socrates is mortal.

\end{enumerate}

\noindent In logical notation, this may be expressed as:

\begin{enumerate}

\item $\forall x: human(x) \Rightarrow mortal(x)$.

\item $human(Socrates)$.

\item $\therefore mortal(Socrates)$.

\end{enumerate}

The classical interpretation in logic of syllogistic reasoning requires 
explicit values (TRUE and FALSE) for propositions, in much the same style as the `output' 
fields of the XOR, NOT and NOTXOR functions shown above. 

In the SP system, the first of the propositions just shown may be expressed with the pattern `X \#X human true $\Rightarrow$ mortal true'. In the manner of Skolemization, the variable `X \#X' in the pattern represents anything at all and may thus be seen to be universally quantified. The scope of the variable may be seen to embrace the entire pattern, without the need for it to be repeated. The symbol `$\Rightarrow$' in the SP pattern serves simply as a separator between `human true' and `mortal true'.

If this pattern is included (with other patterns) in Old and if New contains the pattern `Socrates human true $\Rightarrow$' (corresponding to `human(Socrates)' and, in effect, a request to discover what that proposition implies), SP61 finds one multiple alignment (shown in Figure \ref{truth_value_socrates_alignment}) that encodes all the symbols in New. After unification to form the pattern `X Socrates \#X human true $\Rightarrow$ mortal true', we may read the multiple alignment as a statement that because it is true that Socrates is human it is also true that Socrates is mortal. In this reading, we ignore the `service' symbols, `X' and `\#X'.

\begin{figure}[!hbt]
\centering
\begin{BVerbatim}
0   Socrates    human true =>             0
       |          |    |   |             
1 X    |     #X human true => mortal true 1
  |    |     |                           
2 X Socrates #X                           2
\end{BVerbatim}
\caption{The best multiple alignment found by SP61 with `Socrates human true $\Rightarrow$' in New and patterns in Old that include `X \#X human true $\Rightarrow$ mortal true'.}
\label{truth_value_socrates_alignment}
\end{figure}

Notwithstanding the probabilistic nature of the SP system, there is no ambiguity in a multiple alignment like this because it is the only multiple alignment formed that encodes all the symbols in New. Given that we restrict our attention to multiple alignments of this kind, it is entirely possible for the framework to yield deductions that are either TRUE or FALSE and not something in between.

\subsubsection{Probabilistic `deduction'}

Apart from the interpretation of examples like the one above in terms of classical logic, there 
is, as we saw in Chapter \ref{pr_chapter}, a more relaxed `everyday' interpretation that allows statements and inferences to have levels of confidence or `probabilities'. This kind of interpretation would view a proposition like ``All humans are mortal'' to be an inductive generalisation about the world which, like all inductive generalisations, might turn out to be wrong. In the same vein, we may not be totally confident that Socrates is human. As a consequence of uncertainties like these, the proposition that ``Socrates is mortal'' may have some doubt attaching to it.%
\index{reasoning!syllogistic|)}

\subsection{Chains of reasoning}

A prominent feature of classical systems for logical reasoning is that we can develop `chains' of reasons: if A then B, if B then C, and so on. In the context of probabilistic reasoning, we have seen how this can be done in the SP system in Section \ref{probabilistic_decision_network}---and Section \ref{reasoning_with_rules_section} showed how probabilistic rules could be combined in the SP system to reach a conclusion. In this chapter, we have also seen how chains of additions can be modelled in the SP system (Section \ref{adding_two_bit_numbers}) and how propositions with explicit TRUE or FALSE answers can be combined (Section \ref{propositional_logic_and_ic}). So, although detailed work in this area has not yet been done, we have reason to be confident that the system will accommodate more complex kinds of composite reasoning with explicit values for TRUE and FALSE. 

\section{Discussion}\label{discussion_conclusion}

Before concluding this chapter, let us consider briefly a few issues that 
relate to the ideas that have been presented. There is insufficient space to discuss these issues fully---this section merely flags their relevance and indicates some lines of thinking.

\subsection{Could information compression be a foundation for mathematics and logic?}

Rather than suppose that mathematics is founded on logic or {\em vice versa}, this chapter 
has suggested that mathematics, logic and related disciplines may be seen to be founded on information compression, itself the product of primitive operations of matching and unification of patterns.

Possible objections to this thesis are that:

\begin{itemize}

\item Concepts of information that are the basis of methods for information compression are couched in 
mathematical terms and

\item Mathematics is used in methods for information compression, including the SP model.

\end{itemize}

A possible answer to the first point is that the Hartley-Shannon and algorithmic information theory concepts 
of information are themselves founded on a more primitive idea of discriminable changes (patterns of variation) across one or more dimensions and that, while mathematics provides useful precision, it is not the only way in which information may be expressed or measured (see below).

In a similar vein, an answer to the second point is that mathematics can be useful in 
methods for information compression but that the core operations of matching and unifying patterns do not, in themselves, use mathematical constructs.

Rather less certainly, it may be said that, although a concept like `frequency' (which 
has an important role in discriminating amongst alternative ways in which patterns may be 
matched and unified) may conveniently be expressed in the form of numbers (and these are 
clearly mathematical) the concept may equally well be expressed as a non-numeric 
`strength' of some kind of chemical or physical variable in a brain or computer. Much 
depends, of course, on what does or does not count as being `mathematical'.

In general, there are alternatives to mathematics in the expression or measurement of information, in compression of information and in the measurement of frequencies of occurrence. The proof of these assertions is that all of these things can be and often are done by brains and 
nervous systems in humans and other animals without the assistance of mathematical knowledge. It would be perverse to say that a fox's ability to learn where rabbits are most likely to be found is a mathematical ability. Capabilities of this sort may be less accurate than those that exploit mathematical techniques but they are useful for everyday living and probably essential for survival of species through millions of years.

\subsection{Mathematics, logic and human psychology}

Chapter \ref{foundations_chapter} reviewed some of the writings and research relating to the idea that information compression is fundamental in many aspects of perception and cognition in humans and animals. Given this evidence, and given that mathematics and logic is a product of human cognition, it should not be altogether surprising if we find that information compression is fundamental in mathematics and logic. This is tangential evidence that points in the same direction as the more direct evidence reviewed in this chapter that mathematics, logic and related disciplines may usefully be understood as information compression.

Here we simply note this suggestion as a matter for further consideration in the 
future. There are, no doubt, many implications and ramifications that may be explored.

\subsection{Mathematics and its applicability}\label{maths_science}

The arguments and examples in this chapter may provide some solution to the 
`mystery' of why mathematics is such a good ``language of science''. As Barrow says 
\citeyearpar[p. 247]{barrow_1992}, ``Science is, at root, just the search for compression in the world''. If the arguments in this chapter are accepted, then mathematics is a good medium for the expression of laws and other regularities in the world because mathematics provides the kinds of mechanisms that are standardly used for information compression, with matching and unification of patterns at their core.

Notice that the usefulness of information compression in this connection is not merely in reducing the volume of scientific information (which may facilitate storage or transmission of that information). At least as important is the intimate connection that appears to exist between information compression and the inductive prediction of the future from the past (Section \ref{probabilities_ic_section}).

\subsubsection{The world is compressible}\label{the_world_is_compressible}

Barrow goes on to say ``... the world is surprisingly compressible and the success of mathematics in describing its workings is a manifestation of that 
compressibility.'' ({\em ibid.}, p. 247).

It is interesting to reflect on what things would be like if the world (and 
the universe) were {\em not} compressible. In accordance with algorithmic information theory principles, such a world 
would be totally random. This would mean that there would be no `structure' of any kind, 
including the world itself, people, brains, computers and, indeed, the structures of 
mathematics, logic and related disciplines. So all of the things we have been discussing, 
including mathematics and science, would not exist. Science and the expression of 
scientific truths in mathematics can only exist {\em because} the world is compressible!

It is true that, out of all the possible patterns of information, compressible patterns 
(containing regularities) are relatively rare \citep{li_vitanyi_1997}. If surprisingness depends on rarity, then it is surprising that the world is compressible. But since, in accordance with the anthropic principle, we can only exist in a compressible world that accommodates persistent structures like ourselves, it is {\em not} surprising that the world is compressible!

\subsubsection{The scope of mathematics in science}

Mathematics is useful in many areas of science but it has limitations. The theory of 
evolution by natural selection, which is by any standards a SUPERB theory (to use Penrose's word and emphasis) is not, fundamentally, mathematical. Mathematics can certainly be used to 
analyse aspects of evolution but the theory in itself does not use any mathematical constructs (unless one adopts a definition of mathematics that is so wide that it becomes meaningless). Again, there are many aspects of the world---like distributions of plants and animals or patterns of political beliefs---that seem to be fundamentally `messy' and not amenable to `neat' compression into mathematical formulae.

In short, mathematics is a good language of science but it is not the only language 
for expressing regularities in the world. And there are many aspects of the world that 
have not yet been compressed into compact formulae or any other succinct representation and seem unlikely to yield to that treatment in the future.

\subsection{Mathematics, logic and the inductive prediction of the future from the past}\label{inductive_prediction}

\index{reasoning!inductive|(}

If the main thesis of this chapter is accepted, then it is accepted that mathematics and logic is founded 
on the same principles that appear to underlie concepts of probability and probabilistic 
inductive inference. As we saw in Section \ref{probabilities_ic_section}, there is a very close connection between information compression and concepts of probability. The SP system is based on minimum length encoding 
principles. And \citet{solomonoff_1986} has argued that the great majority of problems in science {\em and} mathematics can be seen as either `machine inversion' problems or `time limited optimization' problems, and that both kinds of problems can be solved by inductive inference using the minimum length encoding principle.

Of course, there are parts of mathematics to do with probability and there are types 
of logic that are probabilistic. But, for many mathematicians and logicians, the idea that 
mathematics or logic are founded on the same principles that underlie concepts of 
probability and probabilistic inductive inference may be hard to accept. It is, after all, the 
highly predictable `clockwork' nature of these disciplines that gives them their appeal.

There is no space here to discuss these issues fully. Five points are noted here briefly:

\begin{itemize}

\item Non-determinism and associated uncertainties can appear in the Turing model 
itself, depending on the transition function built into the model. Likewise, the Post canonical system can exhibit non-determinism, depending on the 
productions provided in the model \citep[see][chapters 10 to 14]{minsky_1967}.

\item Section \ref{true_false_sp} described briefly how the effect of two-valued logic might be obtained in a system that is otherwise geared to probabilistic reasoning.

\item As we saw in Section \ref{syllogistic_reasoning}, some kinds of syllogistic argument can be seen to have inductive origins.

\item That mathematics may not, fundamentally, be quite as `clockwork' as is commonly 
supposed has been suggested by \citet[p. 80]{chaitin_1988}: ``I have recently been able 
to take a further step along the path laid out by G\"{o}del and Turing. By translating a 
particular computer program into an algebraic equation of a type that was familiar 
even to the ancient Greeks, $I$ have shown that there is randomness in the branch of 
pure mathematics known as number theory. My work indicates that---to borrow 
Einstein's metaphor---God sometimes plays dice with whole numbers.''

\item The idea that we can only exist in a compressible world (Section \ref{the_world_is_compressible}, above) seems to provide an answer to the problem of finding a rational basis for inductive reasoning. Of course, we cannot justify such reasoning by saying that it has always worked in the past---because this simply invokes the principle we are trying to justify. A better answer seems to be as follows:

\begin{itemize}

\item The assumption that the future will repeat the past is equivalent to the assumption that there is information redundancy between past and future.

\item If we make that assumption and it turns out to be true, then we will reap the benefits of inductive reasoning---remembering where food or shelter can be found, avoiding dangers, and so on.

\item If we make that assumption and it turns out to be false, the universe will have exploded into total randomness (or exploded and reformed into {\em totally} unrecognisable patterns) and we are dead anyway!

\item In short, there is no downside risk to the assumption that the past is a guide to the future and a very substantial upside if it turns out to be true. Hence, it is a rational assumption to make.

\end{itemize}

\end{itemize}

\index{reasoning!inductive|)}

\subsection{Reconciling the inevitability of mathematics with its applicability}

\citet{potter_2000} writes that the chief philosophical problem posed by arithmetic is ``that of reconciling its inevitability---the impression we have that it is true no matter what---with its applicability.'' (p. 1). Presumably, the same can be said about mathematics as a whole and, indeed, logic.

This section has already suggested a reason why mathematics and logic is so applicable: because it is an embodiment of standard techniques for information compression and, as such, it provides a convenient means of describing the world in a succinct manner and, perhaps more importantly, it gives us a handle on the inductive prediction of the future from the past.

What about the `inevitability' of arithmetic and other aspects of mathematics and logic? The first point in this connection is that, as was noted above, the work of G\"{o}del, Turing, Chaitin and others has shown that arithmetic, at least, has a probabilistic aspect (``God sometimes plays dice with whole numbers''): it may not be as true as is commonly assumed that $2 + 2 = 4$, no matter what. That said, any `fuzziness' in arithmetic is not apparent in its everyday applications and it is still pertinent to ask why the laws of arithmetic are amongst the most certain things that we can know.

The view of mathematics and logic that has been sketched in this chapter suggests possible reasons for the certainties of arithmetic and, indeed, the uncertainties too:

\begin{itemize}

\item With the simple kinds of patterns that seem to form the basis of natural numbers (Section \ref{unary_numbers_and_sp}), the SP system often yields a multiple alignment that stands out very clearly as better in terms of compression than any of the alternative multiple alignments---because simple patterns impose relatively tight constraints on the number of alternative ways in which symbols may be matched and unified. This phenomenon may be responsible for our subjective impression of the inevitability of arithmetic truths. 

\item The uncertainties and paradoxes of mathematics and logic (G\"{o}del's theorems, the Halting Problem, Chaitin's work, Russell's paradox, etc) seem all to arise from the infinite regress of recursion that can occur in mathematical or logical systems. This sits comfortably with the SP view of mathematics and logic because recursion can easily arise in the matching and unification of patterns (as we saw in Section \ref{unary_numbers_and_sp}).

\end{itemize}

\section{Conclusion}

The view of mathematics and logic that is presented in this chapter does not prove that mathematics and logic is based on information compression but it may provide a perspective that stimulates new thinking about the nature of mathematics, logic and related disciplines, and new insights into those disciplines.%
\index{mathematics|)}\index{logic|)}

%% file: neural.tex
\chapter{Neural Mechanisms}\label{neural_chapter}

\index{neural!processing|(}

\section{Introduction}\label{introduction}

As we saw in Chapter \ref{foundations_chapter} (mainly Sections \ref{adaptation_and_inhibition}, \ref{perceptual_constancies}, \ref{oo_in_perception_and_cognition}, \ref{classes_subclasses_inheritance} and \ref{compression_in_nl}), there is a wealth of evidence for the significance of information compression in the workings of brains and nervous systems. Although this evidence has formed part of the motivation for developing the SP theory, the theory has to a large extent been developed in purely abstract terms without reference to the anatomy or physiology of neural tissue. The main purpose of this chapter is to consider, as far as current knowledge allows, possible ways in which the abstract concepts described in Chapter \ref{theory_chapter} may be mapped on to structures and mechanisms in the brain. It will be convenient to refer to these proposals as `SP-neural'.

The main focus of this chapter is on the storage of knowledge and the way in which sensory data may connect with previously-learned knowledge in perception and learning. These are areas where there is a relative paucity of neurophysiological evidence compared with what is now known about the processing of sensory data in sense organs, thalamus and cortex. The SP system, together with existing knowledge about the way in which neural tissue normally works, is a source of hypotheses about the organisation and workings of the brain.

To anticipate a little, it is proposed that `patterns' in the SP system are realised with structures resembling Hebb's \citeyearpar{hebb_1949} concept of a `cell assembly'\index{cell assembly}. By contrast with that concept, it is proposed here that any one neuron can belong in one assembly and {\em only} one assembly. However, any assembly may contain neurons that serve as `references', `codes' or `identifiers' for one or more other assemblies. This mechanism allows information to be stored in a compressed form, it provides a robust mechanism by which assemblies may be connected to form hierarchies and other kinds of structure, it means that assemblies can express abstract concepts, and it provides solutions to some of the other problems associated with cell assemblies.

\section{Global considerations: arguments from biology and engineering}\label{biology_engineering}

The immense difficulties in studying what is going on inside a living brain means that there are enormous gaps in our knowledge. Trying to piece together a coherent theory is a bit like reconstructing an ancient document from a few charred fragments of parchment or papyrus. Various kinds of inference are needed to build understanding in the sea of ignorance that surrounds the scattered islands of knowledge.

To the extent that the SP theory provides a coherent account of cognitive phenomena, it may itself suggest predictions about what is going on in the brain. If, as a working hypothesis, we assume that the theory is true, we may expect to find correspondences between elements of the theory and structures or processes in the brain.

The main aim in this chapter is to consider how the elements of the SP system described in Chapter \ref{theory_chapter} may be realised in neural tissue. But before we examine the nuts and bolts of the theory, there are some global considerations that have a bearing on how the SP theory might relate to concepts in neurophysiology.

Given that brains and nervous systems are the product of evolutionary processes of natural selection, we may evaluate the theory in these terms. Given that biological systems must survive the rigours of the world, we need to consider how the theory may relate to principles of engineering that govern all physical systems, both natural and artificial.

\subsection{Information compression and natural selection}

We would expect natural selection to have favoured organisms that process information in an efficient manner, minimising an organism's requirements for information storage (or maximising the amount of information that can be stored for a given `cost') and maximising the speed with which any given volume of information can be transmitted from one place to another within the nervous system (or minimising the biological cost of transmitting information at a given speed). These considerations may increase our confidence in the long-standing idea---incorporated in the SP theory---that information compression has a central r{\^o}le in perception and cognition (but see Section \ref{compression_revisited}, below).

Another reason---perhaps even stronger---for believing that information compression is a significant factor in the way nervous systems have evolved is the intimate connection that exists between information compression and the inductive prediction of the future from the past (Section \ref{probabilities_ic_section}). Arguably, this kind of prediction---in the processes of finding food, avoiding predators and so on---provides the entire {\em raison d'{\^e}tre} in biological terms for any kind of nervous system. If that is accepted, then we should expect information compression to have a central r{\^o}le in the way neural tissues operate.

\subsection{Is the `cost' of information compression significant?}

It may be argued that the benefits of information compression may not always justify the resources required to achieve it. May we not expect to find situations where organisms use information without compressing it?

If storage costs or speed of transmission were the only consideration, this kind of expectation would seem to be justified. But the very close connection that exists between inductive prediction and information compression seems to imply that we cannot have the former without the latter. If it is accepted that prediction is indeed a fundamental function of nervous systems, it seems that the costs of information compression cannot be avoided.

Compression of information may occur in ways that are not obvious. For example, it may on occasions be useful to store information from the environment in `raw', uncompressed form and this may suggest that information compression has no r{\^o}le. But this stored information is of no use unless it can be related to information that arrives later. And the process of relating the incoming information to information that has been stored means matching and unification of patterns---which are the elements of information compression.

\subsection{Information compression and the benefits of redundancy in the storage and processing of information}\label{multiple_copies}

\index{information!redundancy|(}

Although the SP theory sees information compression as fundamental in computing and cognition, it should not be forgotten that redundancy in information has a clear r{\^o}le to play in information systems in guarding against loss of information, in speeding up processing, or in the correction of errors that may occur during the processing or transmission of information (Section \ref{uses_of_redundancy}). 

It seems likely that exploitation of redundancies would be as important in brains and nervous systems as it is in artificial systems. For example, it seems very unlikely that evolution would have endowed us with information processing systems that allow us to keep only a single copy of our hard-won knowledge. And there is every reason to think that redundancies may be exploited to speed things up or to reduce errors in neural processing of information.

As we saw in Section \ref{uses_of_redundancy}, there is no conflict between these uses of redundancy in information processing and the idea that information compression is a key part of the foundations of computing and cognition.%
\index{information!redundancy|)}

\subsection{SP-neural and the modular structure of the brain}

A last issue to consider briefly before we proceed to the main proposals is the way in which the SP theory may relate to the structure of the brain.

The mammalian brain (and other parts of the nervous system) is divided into clearly-defined modules, many of which have a clear association with particular functions---Broca's areas is concerned with language, the visual cortex deals with vision, and so on. Furthermore, there are clear differences in anatomy and physiology from one module to another.

It might be thought that this modular structure---and the differences amongst modules---rules out any global theory of brain function such as the SP-neural proposals. An alternative view---which is the one adopted in this research---is that the SP-neural theory describes general principles that may seen to apply in a variety of different modules in different parts of the brain. It is entirely possible that the same principles may apply in different modules and, at the same time, there may be differences in how they are applied. It is clear for example, that two-dimensional visual images need to be handled in a way that is rather different from the time-ordered and predominantly one-dimensional patterns of hearing. And yet the matching and unification of  patterns---and the building of multiple alignments---are ideas that may be applied in both areas. 

\section{Neural mechanisms for the SP framework}\label{neural_mechanisms_overview}

This section presents an outline of how the SP concepts may be realised in neural tissue, while the next section considers aspects of the outline in more detail in the light of available evidence, highlighting areas of uncertainty and issues that need to be resolved.

In broad-brush terms, the proposals are these:

\begin{enumerate}

\item An SP {\em pattern} in Old may be realised by a network of interconnected cells, similar to Donald Hebb's \citeyearpar{hebb_1949} concept of a {\em cell assembly}\index{cell assembly} \citep[see also][pp. 103--107]{hebb_1958}. Because there are significant differences between the SP-neural proposals and Hebb's original concept, the neural realisation of an SP pattern in Old will be referred to as a {\em pattern assembly}.

\item An SP {\em pattern} in New may be represented in arrays of cells in the sensory systems or elsewhere. In the case of vision for example, these may be in the layers of the visual cortex concerned with early stages of processing (layers 3 and 6) or, perhaps, in the lateral geniculate body. Comparable structures may be found in other parts of the sensory cortex or thalamus. In the process of establishing long-term memories, the hippocampus appears to be involved \citep{kandel_2001}.

Each such structure must be able to respond to a succession of different patterns, as discussed below. In addition to its r{\^o}le in processing information, each such structure functions as a buffer within which a pattern from New may be stored pending further processing. Any structure like this that can receive a pattern and store it temporarily will be referred to as a {\em neural array}. Anything that is either a pattern assembly or an neural array may be referred to as a `neural pattern'.

\item Much of the discussion below will focus on the analysis of sensory data. But it should not be forgotten that the r{\^o}le of `New' may, on occasion, also be played by patterns that are in some sense `internal' to the system. As we saw in Section \ref{decompression_by_compression}, an internally-generated code pattern for a sentence or, perhaps, a semantic structure, may take the r{\^o}le of New. We shall suppose that there is some kind of `internal' neural array to store patterns of that kind, distinct from neural arrays for sensory data that will be referred to as `sensory' neural arrays. It is conceivable, of course, that sensory neural arrays may also serve as internal neural arrays.

\item Where an SP {\em symbol} represents a fine-grained detail, it may be realised as a single neuron. Where a symbol represents something a bit larger, it may be realised with a pattern assembly. It is envisaged that each such `neural symbol'---a single cell or a small pattern assembly---will respond preferentially to a particular pattern of stimulation received from elsewhere within the nervous system.

\item A neuron within a pattern assembly that corresponds to a C-symbol within an SP pattern will be referred to as a C-neuron. Likewise, a neuron that represents an ID-symbol will be called an ID-neuron.

\item Although sensory information is received initially in analogue form, it is widely accepted that the information is converted at an early stage into something like David Marr's \citeyearpar{marr_1982} concept of a `primal sketch' in the domain of vision or something equivalent in other domains. In vision, it seems that sensory data is converted into the digital language of on-centre cells, off-centre cells, cells that are sensitive to lines at particular angles, cells that detect motion, and so on \citep[see, for example,][]{nicholls_etal_2001, hubel_2000}. Other sensory modalities seem to be similar.

\item It is envisaged that, when the cells of a sensory neural array become active as a result of sensory input, the pattern of excitation will be transmitted to the C-neurons of a range of pattern assemblies. Through an interplay of excitatory and inhibitory processes to be described below, activity will spread to other pattern assemblies until sets of pattern assemblies have become active in a manner comparable with multiple alignments in the SP system.

\item In a similar way, cells of an internal neural array may broadcast excitation to ID-neurons of a range of pattern assemblies. From that point on, the spread of activity should achieve the formation of multiple alignments in precisely the same way as if the excitation had originated from sensory input.

\item In SP70, learning occurs by adding newly-created patterns to Old, including some derived relatively directly from New patterns and others derived from multiple alignments in which there is partial matching of patterns. When a range of patterns have been added, there is a process of evaluation to differentiate those that are good in terms of the economical encoding of patterns from New from those that are bad. It is envisaged that, in future versions of the model, bad patterns will be removed from Old, leaving only the relatively good ones. This kind of evaluation and purging may occur periodically as patterns from New are processed.

There seems no reason in principle why a comparable kind of process should not operate with pattern assemblies. It seems unlikely that cells would be literally added or removed from within a brain but it seems possible that groups of pre-existing cells could be brought into play as pattern assemblies, or the reverse, according to need.

\end{enumerate}

\subsection{A schematic representation}\label{schematic_representation}

\begin{figure}
\centering
\includegraphics[width=0.9\textwidth]{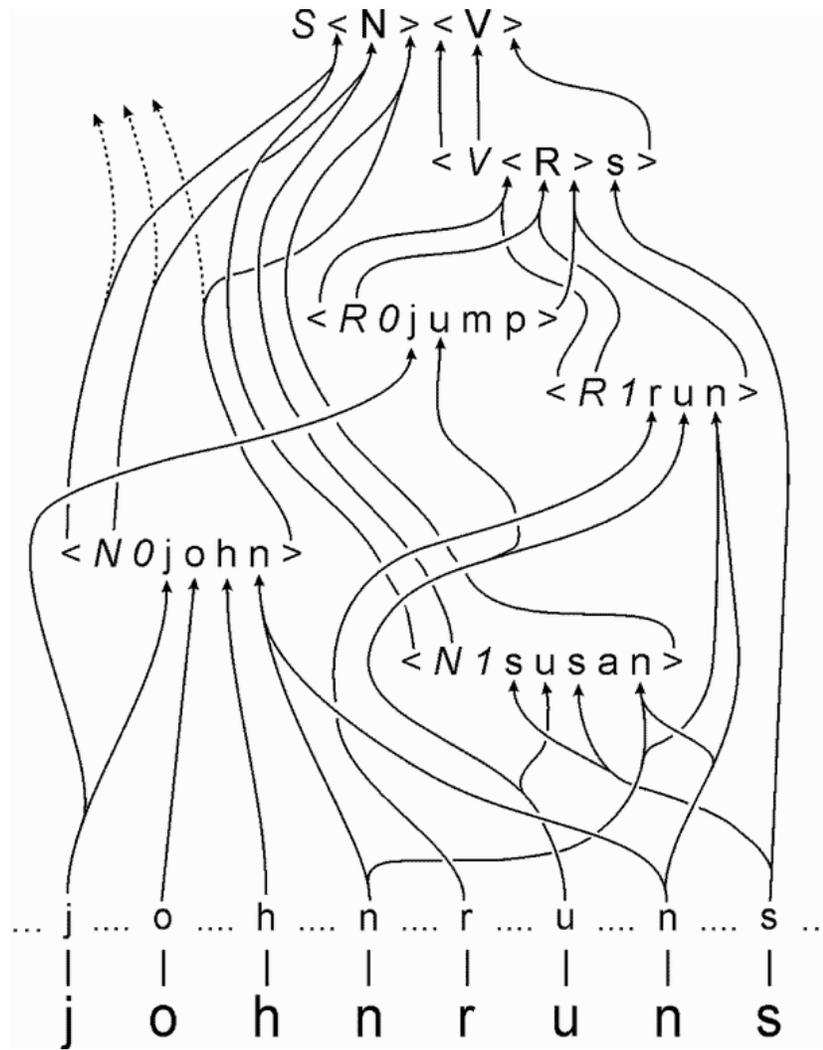}
\caption{Schematic representation of the proposed neural basis for SP concepts. The conventions used and other aspects of this figure are explained in the text.}
\label{neural_basis_figure}
\end{figure}

In order to summarise the proposals as they have been developed so far and to provide a focus for the discussion that follows, Figure \ref{neural_basis_figure} provides a diagrammatic representation of the way in which the neural substrate for SP concepts may be organised.

This example has a `linguistic' flavour but it should be emphasised that the proposals are intended to apply to all sensory modalities---vision, hearing, touch etc---both individually and in concert. Thus each of the patterns shown in the figure should be interpreted broadly as an analogue of information in {\em any} modality (or combination of modalities), not merely as a textual representation of some aspect of linguistic structure.

\subsubsection{Neural arrays}

The pattern `j o h n r u n s' at the bottom of the figure represents raw sensory data in analogue form. Each letter is intended to represent a low-level feature of the sensory input that will be recognised by the system.

The pattern shown immediately above, with the same letters but with dots in between them, is intended to represent a sensory neural array containing detectors (neurons) for low-level features of the image---lines at various angles, colours, luminances, elementary movements and the like. Physiologically, it seems that different kinds of features are processed in different areas of the visual cortex but conceptually they may be seen to belong to one neural array.

The entire alphabet of detectors is repeated many times across the neural array so that it can respond to any feature at any position within the neural array. The schematic representation in the figure is intended to convey the idea that, at each position, an appropriate detector is responding to the corresponding small portion of the sensory input.

\subsubsection{Pattern assemblies}

Each of the patterns shown in the figure above the neural array are intended to represent pattern assemblies located somewhere in the brain, most probably in the cortex. Each character within each pattern assembly represents a neuron. Angle brackets (`$<$' and `$>$') at the beginning and end of each pattern and characters with an oblique angle represent ID-neurons. All other characters represent C-neurons.

Although the pattern assemblies shown in the figure are all one dimensional, it seems likely that many pattern assemblies (e.g., in the domain of vision) will have two dimensions.

Contrary to what the figure may suggest, it is likely that the lowest-level pattern assemblies correspond to features of the sensory input that are relatively small (but larger than the features registered by detectors in the sensory neural array). In vision, these might be corners, angles or other small motifs. In speech, they may be structures such as vowels, consonants or diphthongs.

It seems likely that there will be many more `levels' of pattern assemblies than the few levels shown in Figure \ref{neural_basis_figure}. As indicated above, it seems likely that the higher-level pattern assemblies will represent associations bridging several different sensory modalities.

It seems likely that each pattern assembly will contain many more cells than is suggested by the examples in the figure.

Not shown in the figure are the connections envisaged between neurons within each pattern assembly. Each neuron will have connections to its immediate neighbours at least and, perhaps, beyond.

\subsubsection{Connections between neural patterns}

Each of the neurons within the sensory neural array is connected to zero or more C-neurons representing the same low-level feature in zero or more pattern assemblies. The figure shows only the connections from neurons within the neural array that have become active as a result of sensory input. There will be similar connections from each of the many neurons not shown within the sensory neural array.

Each of the pattern assemblies are connected in a similar way to other pattern assemblies. Each of the ID-neurons within each pattern assembly is connected to zero or more C-neurons representing the same symbol in zero or more pattern assemblies. It is possible for an ID-neuron within one pattern assembly to be connected to a C-neuron within the same pattern assembly but this `recursive' kind of connection is not shown in the figure.

The dotted arrows in the figure are included as an indication that any one pattern assembly (`$<$ N 0 j o h n $>$' in this example) may be connected to two or more higher-level pattern assemblies. The dotted arrows might, for example, connect with a higher-level pattern assembly representing the interrogative form of a sentence. 

Each connection comprises a single axon, or a chain of two or more neurons, with branches where necessary. Contrary to what the figure suggests, axons do not join together where they converge on one neuron. That kind of representation is a shorthand for two or more synapses onto the body or dendrites of the given neuron.

No attempt has been made to show any kind of internal neural array in the figure. It is envisaged that structures of that kind would have connections to ID-neurons such as `0' in `$<$ N 0 j o h n $>$' or `1' in `$<$ R 1 r u n $>$'.

Not shown in the figure are lateral connections between fibres, proposed and discussed in Section \ref{building_mas}, below.

\section{Representation of knowledge in the brain}\label{rep_knowl_brain}

The aim in this section and the ones that follow is to flesh out the proposals outlined in the previous section, filling in detail, considering relevant evidence and discussing associated issues.

The central idea---that knowledge is stored in the brain as SP-style assemblies of neurons---is broadly similar to Hebb's proposals but there are important differences, as we shall see.

The cells in a pattern assembly need not be physically contiguous although it seems likely that, in most cases, they will be. Whatever the spatial arrangement of the cells, they should be topologically equivalent to an array of cells in one or two dimensions, or possibly more in some cases.

It is assumed in these proposals that pattern assemblies exist largely in the cortex. It seems likely that those concerned with patterns in a particular sensory modality will be in the corresponding area of the sensory cortex. But many others, especially those concerned with a combination of sensory modalities, are likely to be located in the `association' areas of the cortex.

As in Hebb's proposals, it is envisaged that each cell in a pattern assembly is connected to other cells in the pattern assembly, at least its immediate neighbours.

As noted above, an SP symbol may translate into neural terms as a neuron or, if it represents a relatively large chunk of information, it may be realised as a pattern assembly. In the latter case, it needs to be connected to the pattern assembly representing the pattern in which the given symbol appears. This may be done in exactly the same way as a pattern assembly may be linked to another pattern assembly, as described below.

\subsection{Hierarchical organisation of knowledge}\label{hierarchical_organisation}

\index{knowledge, representation of|(}

Hebb \citeyearpar[p. 130]{hebb_1949} envisaged an hierarchical organisation for conceptual structures and the possibility of low-level concepts being shared by two or more higher-level concepts (see also the paragraph from \citet[p. 105]{hebb_1958} quoted in Section \ref{parts_wholes_assocs}, below). Hebb's descriptions imply that any one neuron may belong in two or more cell assemblies and this is commonly assumed (see, for example, \citet[p. 385]{huyck_2001a}, \citet[pp. 257--258]{pulvermuller_1999} and \citet[p. 213]{sakurai_1998}).

This kind of organisation is problematic. If, for example, the cell assembly\index{cell assembly} that recognises the letter `a' is embedded in the cell assembly for each word that contains that letter, it is hard to see how the cell assembly for `aardvark' would be organised. It is also hard to see how one could represent the abstract structure of a sentence, for example, where each `slot' in the pattern may receive a range of {\em alternative} elements.

These problems may be overcome by the use of {\em references} between assemblies instead of the literal embedding of one assembly in another.

By contrast with the generally accepted view of cell assemblies, it is envisaged in the SP-neural proposals that any one neuron may belong in one pattern assembly and {\em only} one pattern assembly. Hierarchical organisation and the sharing of structures is achieved by connections between ID-neurons and C-neurons that achieve the effect of pointers in software or references in books or articles. There is further discussion in Section \ref{parts_wholes_assocs}, below.

This kind of reference mechanism means that a given pattern assembly can link widely-dispersed areas of the cortex without itself needing to be large (see Section \ref{lengths_of_connections}, below).%
\index{knowledge, representation of|)}

\subsection{New information in brains and nervous systems}

\index{New|(}

As indicated above, sensory information is received by `neural arrays' such as the visual cortex or areas of the cortex concerned with other sensory modalities. Comparable structures may also exist within the thalamus. It is envisaged that internally-generated patterns that play the r{\^o}le of New information will be stored in some kind of internal neural array.

A sensory neural array, unlike a pattern assembly, must be able to capture any of a wide variety of patterns that arrive at the senses. As a rough generalisation, it seems that this is normally achieved by providing a range of different types of receptor cells, each type tuned to some relatively small feature that may be found in sensory data and the whole set repeated across the neural array. In the visual cortex, for example, there are cells that respond selectively to short line segments at different angles and the complete range of orientations is repeated within each of a large number of fairly small `orientation columns' \citep{barlow_1982, hubel_2000}. It seems likely that a similar organisation may exist in the somatosensory cortex, mirroring the way in which receptors that respond selectively to heat, cold, touch, pressure and so on are repeated across areas of the skin \citep[][Chapter 18]{nicholls_etal_2001}.

There are many complexities of course (e.g., `complex' cells in vision that can respond to a line at a given orientation in any of several different positions \citep{hubel_2000}\footnote{If complex cells could accommodate large variations of position, this would mean that useful information about the positions of visual entities would be lost at the level complex cells. It has been suggested \citep[][p. 422]{nicholls_etal_2001} that the main r{\^o}le of complex cells is to cope with the relatively small changes in position caused by the small saccadic eye movements that are continually made to prevent visual input being suppressed by adaptation in the retina. However, Milner (personal communication) suggests that the receptive fields of many complex cells are significantly bigger than would be needed for that function.}). But in general the overall effect seems to be to translate raw analogue data into a digital language composed with an alphabet of sensory features.

So far, we have assumed that a New pattern arrives all at once like an image projected on to the retina. But it is clear that streams of sensory information are being received in all sensory modalities throughout our lives and that we are very sensitive to patterns in the temporal dimension within those streams of information, especially in hearing. It is assumed here that temporal sequencing in hearing, vision and other modalities is encoded spatially in arrays of cells, probably in the cortex.%
\index{New|)}

\subsection{Connections between neurons}

As noted above, it is envisaged that, within each pattern assembly, neurons will be interconnected, at least to their nearest neighbours. If, as suggested above, the cells within any one pattern assembly lie close together, then lateral connections between the cells within any one pattern assembly will normally be quite short.

From each neuron with an neural array, there may be zero or more connections, each one to a corresponding C-neuron within a pattern assembly. And from each ID-neuron within a pattern assembly there may be zero or more connections, each one to a corresponding C-neuron within another pattern assembly. In this context, the word `corresponding' means that both neurons represent the same SP symbol. We may suppose that the connections from any one neuron (within an neural array or a pattern assembly) are formed from an axon with as many branches as are required. 

\subsubsection{The volume of connections between neural patterns}

There is a possible worry here that the number of connections might get out of hand and that the volume of axons would be too big to fit inside anyone's skull. This might be true if everything was connected to everything else but that is not necessary and seems an unlikely feature of the brain.

Consider, for example, how a neuron within an neural array may connect to pattern assemblies representing sensory patterns. A neuron representing a short line segment at a particular angle could, conceivably, make a connection to each one of thousands of pattern assemblies representing objects or scenes that contain that small feature. It seems more likely, however, that a neuron within an neural array would connect only to a relatively small number of pattern assemblies each one of which represents some relatively small `motif'---an angle in some orientation or some other small detail. In a similar way, each of those motifs may connect to a relatively small number of higher-level patterns and so on through several different levels, much like the various levels of description that are familiar in a linguistic context (illustrated in Section \ref{framework_examples_section}).

By breaking patterns down into parts and sub-parts in this kind of way, with connections chiefly between one level and the next, the number of connections that are required can be dramatically reduced.

As with SP patterns (or, indeed, the rules of a typical grammar), pattern assemblies do not necessarily fall into a strict hierarchy. It would be more accurate to call it a `loose heterarchy', where any one pattern assembly may receive connections from two or more lower-level pattern assemblies and may also connect to two or more higher level pattern assemblies. Connections need not always be to some strictly-defined `next level' and, where there are recursive relations, ID-neurons within a given pattern assembly may actually connect to C-neurons of the same pattern assembly (examples of this kind of recursion may be found in \citet{wolff_1999_comp}).

Any untidiness of this kind makes no difference to the basic idea that any one neuron need only be connected to a small subset of all the pattern assemblies and thus the volume of interconnections may be kept within reasonable bounds. This economy in interconnections is, indirectly, a reflection of the economy that can be achieved with an SP style of knowledge representation.

\subsubsection{The lengths of connections between neural patterns}\label{lengths_of_connections}

How long would the connections between neural patterns have to be? That depends on what kinds of things are being connected.

pattern assemblies representing concepts within one domain---some aspect of vision, for example---may all reside within one relatively small area of the sensory cortex and their interconnections need not be very long. However, many of our concepts involve associations across two or more modalities and this seems to demand relatively long interconnections in most cases. Thus, for example, Pulverm{\"u}ller \citeyearpar{pulvermuller_1999} has suggested that ``most word representations consist of two parts, a perisylvian part related to the word form and a part located mainly outside the perisylvian areas representing semantic word properties.'' (p. 275).

As another example, consider the concept of `our house'. This is likely to include the visual appearance of the house (patterns of luminance, colour and motion), sounds, smells and tactile aspects as well as emotional associations. Although the corresponding pattern assembly may be quite compact and located in one small part of the cortex (one of the `association' areas, perhaps), it must have connections to the areas of the brain concerned with each of the elements, sensory or otherwise, from which the concept is built. Given that each modality is typically processed in a localised area of the cortex and given that these areas are widely dispersed \citep{nicholls_etal_2001}, it seems that relatively long connections will normally be needed.

This has some bearing on learning processes, as we shall see in Section \ref{learning_section}.

\subsection{Are there enough neurons in the brain?}\label{storage_capacity}

The idea that the brain stores knowledge in the form of SP-style cell assemblies raises the question ``Are there enough neurons in the brain to support this style of representation''? This is a difficult question to answer because of various uncertainties. Here are some very rough estimates, made with rather conservative assumptions:

\begin{itemize}

\item Estimates of the number of neurons in the human brain range from $10^{10}$ \citep{sholl_1956} to $10^{11}$ \citep{williams_herrup_2001}.

\item If a neuron in the sensory cortex responds to some small feature of the world---a line at a certain angle or a note of a given pitch---and if there are, at a conservative estimate, about 30 different basic features across all sensory modalities, then each neuron represents $\lceil \log_2 30 \rceil = 5$ bits of information. It seems reasonable to assume that other neurons in the brain have a similar capacity. Erring on the side of caution, we shall make the very conservative assumption here that each neuron represents only 1 bit of information. On that basis, the `raw' storage capacity of the brain is between approximately $10^9$ to $10^{10}$ bytes or between 1000 MB and 10,000 MB.

\item Bearing in mind that stored knowledge will be heavily compressed using SP-style mechanisms, the effective storage capacity of the brain will be larger. With an ordinary Lempel-Ziv compression algorithm (e.g., PkZip or WinZip) it is possible to compress English text to one third of its original size (without loss of non-redundant information). This is probably a rather conservative estimate of the amount of compression that may be obtained across diverse kinds of knowledge:

\begin{itemize}

\item Lempel-Ziv algorithms are designed for speed on a conventional serial computer rather than optimal compression. More compression can be achieved when more processing power is available (as in the human brain).

\item Visual information typically contains more redundancy than English text. For example, a good quality compressed JPEG version of a typical photograph is about 5 times smaller than an uncompressed bitmap version of the same image. It is true that JPEG is a lossy compression technique but if the quality is good, the amount of non-redundant information that is lost is relatively small.

\item `Cinematic' information normally contains very high levels of redundancy because, in each sequence of frames from one cut to the next, each frame is normally very similar to its predecessor. This redundancy allows high levels of compression to be achieved.

\end{itemize}

If we adopt the conservative assumption that compression by a factor of 3 may be achieved across all kinds of knowledge, our estimates of the storage capacity of the brain will range from about 3000 MB up to about 30,000 MB.

\item Is this enough to accommodate what the average person knows? Any estimate here can only be very approximate. Let us assume that the average person knows only a relatively small proportion of what is contained in the Encyclopaedia Britannica. Clearly, each person knows lots of things that are {\em not} contained in that encyclopaedia---how to walk or ride a bicycle, information about friends, relatives and acquaintances, how the local football team is doing, and so on. If we assume that the `personal' things that we do know are very roughly equal to the things in the encyclopaedia that we do not know, then the size of the encyclopaedia provides an estimate of the volume of information that the average person knows.

\item The Encyclopaedia Britannica can be stored on two CDs in compressed form. Assuming that most of the space is filled, this equates to 1300 MB of compressed information or approximately 4000 MB of information in uncompressed form. This estimate of what the average person knows is the same order of magnitude as our range of estimates of what the human brain can store.

\end{itemize}

Even if the brain stores two or three copies of its compressed knowledge (to guard against the risk of losing it, as discussed in Section \ref{multiple_copies}), our estimate of what needs to be stored (12,000 MB, say) lies within the range of our estimates of what the brain can store. Bearing in mind the {\em very} conservative assumption that was made about how much information each neuron represents and other conservative assumptions that have been made, it seems that the proposals in this chapter about how the brain stores information are not wildly inconsistent with what is required.

\subsection{Comparisons}

In terms of the `static' forms of knowledge in the brain (ignoring learning), the SP-neural proposals extend or modify Hebb's original concept in the following ways:

\begin{itemize}

\item There is a distinction between neural arrays that, {\em inter alia}, serve as temporary buffers for information (from the senses or elsewhere) and pattern assemblies that provide for longer-term storage of information. 

\item Within each pattern assembly, it is envisaged that some cells (C-neurons) represent the `contents' or `substance' of the pattern while others (ID-neurons) will serve to identify the pattern assembly or otherwise define how it relates to other pattern assemblies. 

\item Connections between ID-neurons and C-neurons provide a mechanism by which one pattern assembly may be `referenced' from another pattern assembly. This is the key to the expressive power of the SP system.

\item As noted above, any one neuron may not belong in more than one pattern assembly.

\item Information is transmitted through the network of pattern assemblies by an interplay of excitatory and inhibitory processes as described in Section \ref{building_mas}, next.

\end{itemize}

\section{Neural processes for building multiple alignments}\label{building_mas}

Given a structure of pattern assemblies, neural arrays and interconnections as described above, how may the system work to build structures that are comparable with the kinds of multiple alignments created by the SP61 and SP70 computer models? The focus here is mainly on the way in which sensory data may connect with stored patterns, amplifying what was sketched in Section \ref{neural_mechanisms_overview}, discussing relevant issues and filling in details.

\begin{figure}[!hbt]
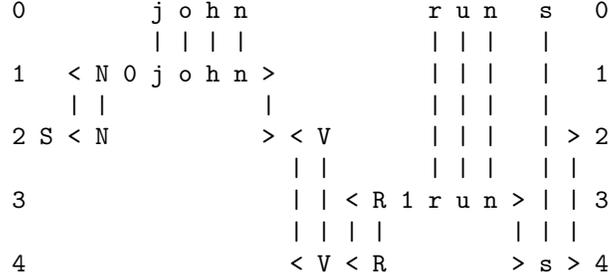

\centering
\begin{BVerbatim}
0         j o h n             r u n   s   0
          | | | |             | | |   |  
1   < N 0 j o h n >           | | |   |   1
    | |           |           | | |   |  
2 S < N           > < V       | | |   | > 2
                    | |       | | |   | |
3                   | | < R 1 r u n > | | 3
                    | | | |         | | |
4                   < V < R         > s > 4
\end{BVerbatim}
\caption{The best multiple alignment found by SP61 with patterns that are comparable to the pattern assemblies shown Figure \ref{neural_basis_figure}.}
\label{neural_alignment_1}
\end{figure}

In terms of the schematic representation shown in Figure \ref{neural_basis_figure}, we are interested in the kind of neural processing that would yield something equivalent to the multiple alignment shown in Figure \ref{neural_alignment_1}. This is the best multiple alignment found by SP61 with patterns that are comparable with the pattern assemblies shown in Figure \ref{neural_basis_figure}.

Given a scarcity of direct evidence about what is going on in this area of brain function, it has been assumed that the kinds of principles that have been seen to operate elsewhere in nervous tissue will also apply here. In particular:

\begin{enumerate}

\item Contrary to what one might assume, neurons do not normally rest quietly in the absence of stimulation. There is normally a background rate of firing and this may be increased by excitatory input or decreased by inhibitory input \citep[see, for example,][]{ratliff_hartline_1959, ratliff_etal_1963}.

\index{neural!inhibition|(}

\item Inhibition seems to be a prominent part of the mechanism by which nervous systems remove redundancy from information. For example, lateral inhibition in the retina operates when neighbouring parts of a visual image are the same---which means that there is redundancy in that part of the image. In a similar vein, \citet{barlow_1997} describes a `law of repulsion': ``when two stimuli frequently occur together their representations discourage or inhibit each other so that each is weaker than it is when each stimulus is presented alone.'' (p. 1144). Since frequent recurrence of two stimuli together is a manifestation of redundancy, this law seems to be in keeping with the idea that a major r{\^o}le of inhibition in neural tissue is to reduce redundancy in information.

\item Lateral inhibition may, in some situations, be seen as a way of forcing a decision amongst competing alternatives so that the `winner takes all'. In the innervation of the Basilar membrane, where position along the length of the membrane encodes the frequency of sound, lateral inhibition has the effect of strengthening the response where the signal is strongest and suppressing the response in neighbouring regions \citep{von_bekesy_1967}.

Mutual inhibition amongst competitors may also be seen as a means of reducing redundancy because some or all of the information expressed by each element in the competition is also expressed by other elements. Neighbouring regions of the Basilar membrane respond to sounds with similar frequencies---which means that there is information in common amongst those sounds.%
\index{neural!inhibition|)}

\item Where there is a mapping from one part of the nervous system to another (e.g., between the retina and each of the areas of the visual cortex), each point in the origin normally projects to an area in the destination and there is overlap between the projection areas of neighbouring points \citep{nicholls_etal_2001}.

\end{enumerate}

What follows is an outline of two possible ways in which the system may function, in keeping with the principles just described.

\subsection{Outline of processing}

When sensory input is received, this causes a subset of the neurons within each sensory neural array to increase their rate of firing. These signals are transmitted to corresponding C-neurons within a relatively small number of `low level' pattern assemblies like `$<$ N 0 j o h n $>$' and `$<$ R 1 r u n $>$' in the schematic representation in Figure \ref{neural_basis_figure}. At this stage, there seem to be two main possibilities, described in the next two subsections.

\subsubsection{`Excitatory' scheme}

\index{neural!excitation|(}

In the first scheme, which is similar to Hebb's original proposals, signals received by the C-neurons of a pattern assembly would have the effect of increasing the rate of firing of those neurons. This excitation would spread via the lateral connections to all the other neurons in the pattern assembly, including the ID-neurons and any C-neurons that had not received any stimulation on this occasion.

When the rate of firing of ID-neurons is increased as a result of excitatory input to the C-neurons of a pattern assembly, these signals will be transmitted to C-neurons in other pattern assemblies. As before, the effect will be to increase the rate of firing of those neurons. And, as before, when C-neurons within a pattern assembly have become excited, this will have the effect of increasing the rate of firing of ID-neurons within the pattern assembly and any C-neurons that did not receive any input.

This process may be repeated so long as there are pattern assemblies that can receive signals from other pattern assemblies in which the ID-neurons are active.

Although the emphasis here is on the excitation of neurons, inhibition may have a r{\^o}le to play in preventing excitation getting out of control \citep{milner_1957}. There might, for example, be inhibitory feedback between each C-neuron within a pattern assembly and the fibres that bring excitatory signals to that neuron. This would have the effect of dampening down the excitatory signals once they had been received. 

This kind of inhibitory feedback may be seen to have the effect of reducing or eliminating redundancy in the temporal dimension corresponding to the persistence of a given sensory input over a period of time.%
\index{neural!excitation|)}

\subsubsection{`Inhibitory' scheme}

\index{neural!inhibition|(}

Notwithstanding some evidence suggesting that the majority of nerve cells in the cortex are excitatory \citep[][p. 264]{pulvermuller_1999} an alternative `inhibitory' scheme is sketched here that is broadly consistent with the principles described above.

The first of the three principles suggests that, in the absence of stimulation, the neurons in each pattern assembly maintain a steady rate of firing.

Since each pattern assembly represents a repeating (redundant) pattern in the environment, it is further assumed---in accordance with the second principle---that there is mutual inhibition amongst the neurons of each pattern assembly, mediated by the lateral connections between the neurons. There will be a balance between the tendency of the neurons to fire spontaneously and the inhibitory effect that they have on each other.

We may further suppose that, when a neuron within a pattern assembly receives stimulation, the effect is inhibitory. Again, this seems to accord with the second principle because the incoming stimulation and the neuron receiving the stimulation represent the same small pattern or symbol, and this repetition of information means redundancy.

If the rate of firing of C-neurons within a pattern assembly is inhibited, this will have the effect of reducing the inhibitory signals which those neurons have been sending to their neighbours. Hence, the effect of inhibitory input to the C-neurons of a pattern assembly will be to {\em increase} the rate of firing of other neurons in the pattern assembly that are not receiving input. The neurons whose rate of firing will increase will be all of the ID-neurons (because they do not receive any input) and any C-neurons that may not have received any input on this occasion. The former will be the focus of attention here. The latter may be seen as inferences that can be derived from the multiple alignment, as described in Chapter \ref{pr_chapter}.

When the rate of firing of ID-neurons is increased as a result of inhibitory input to the C-neurons of a pattern assembly, these signals will be transmitted to C-neurons in other pattern assemblies. As before, the effect will be to inhibit the activity of those neurons. And, as before, when C-neurons within a pattern assembly have been inhibited, this will have the effect of increasing the rate of firing of ID-neurons within the pattern assembly and any C-neurons that did not receive any input.

This process may be repeated so long as there are pattern assemblies that can receive signals from other pattern assemblies in which the ID-neurons are active.%
\index{neural!inhibition|)}

\subsection{Discussion and elaboration}

Neither of the schemes just described is sufficient in itself to achieve an effect comparable with the building of multiple alignments in the SP system. This subsection considers some other aspects of the proposals and fills in some details that seem to be needed for the system to work as intended.

\subsubsection{Inputs to pattern assemblies}

In both of the schemes that have been outlined, it is natural to assume that the effect of inputs to each pattern assembly would be related to the amount or number of inputs received. If the pattern assembly `$<$ N 0 j o h n $>$' in Figure \ref{neural_basis_figure} were to receive inputs from only `j' and `h', one would expect the effect to be less than when the pattern assembly receives signals from all of `j', `o', `h' and `n'. Correspondingly, one would expect the increase in firing rate of the ID-neurons to be larger in the second case than the former.

It seems reasonable to assume that the strength of the responses would depend on the proportion of C-neurons in any pattern assembly that receive stimulation rather than their absolute numbers. 

\subsubsection{Divergence}\label{divergence}

What happens when the signals from a given neuron or combination of neurons in an neural array or a pattern assembly are transmitted to corresponding C-neurons in two or more other pattern assemblies? In the absence of any disambiguating context (see Section \ref{convergence}, next), all of those pattern assemblies should respond, and the strength of the response should be related to the proportion of C-neurons in each pattern assembly that have received an input.

This kind of response would naturally accommodate the kind of ambiguity described in Section \ref{ambiguity_in_parsing}.

\subsubsection{Convergence}\label{convergence}

What happens where two or more sets of fibres converge on one set of C-neurons in one pattern assembly? An example can be seen in Figure \ref{neural_basis_figure} where fibres from `$<$, `N' and `$>$' in the pattern assembly `$<$ N 0 j o h n $>$' are connected to `$<$ N $>$' in the pattern assembly `S $<$ N $>$ $<$ V $>$', and similar connections are made to the same three neurons from the pattern assembly `$<$ N 1 s u s a n $>$'.

In this example, the pattern assemblies `$<$ N 0 j o h n $>$' and `$<$ N 1 s u s a n $>$' are, in effect, competing for the r{\^o}le of noun within the pattern assembly `S $<$ N $>$ $<$ V $>$' representing the structure of a simple sentence. In the light of the third principle mentioned above, we may suppose that a decision about which of these two pattern assemblies should `win' the competition may be decided by inhibition.

Where fibres representing the same symbol converge on a single neuron, we may suppose that there are lateral connections between them that carry inhibitory signals when the main fibres are excited. This should ensure that the strongest signal gets through and weaker ones are suppressed.

At the highest levels in the hierarchy of pattern assemblies, we may suppose that there is a similar competition for the r{\^o}le of `best overall interpretation' of the sensory input. Where there is ambiguity of the kind mentioned earlier, a disambiguating context may tilt the balance in favour of one interpretation or the other.

Possible evidence in support of the proposed lateral connections between fibres carrying rival signals is the discovery of lateral connections in the visual cortex that connect columns in which the cells respond to edges or lines with similar orientations \citep[see][p. 440]{nicholls_etal_2001}. It is not clear whether or not these connections are inhibitory.

Lateral connections are also broadly consistent with the fourth principle described above (projection from neighbouring points to overlapping areas).

\subsubsection{Keeping track of order in perception}\label{order_relations_perception}

As the system has been described so far, a stimulus like `b o y' would produce the same level of response in a pattern assembly like `$<$ N 2 y o b $>$' as it would in a pattern assembly like `$<$ N 3 b o y $>$'. In other words, there is nothing in what has been proposed to keep track of the ordering or arrangement of elements in a pattern.

Our ability to recognise patterns despite some jumbling of their elements (e.g., solving anagrams) is consistent with the scheme that has been described. But the fact that we can see the difference between an anagram and its solution shows that something else is needed. Some kind of mechanism is required to distinguish between two kinds of situation:

\begin{enumerate}

\item Fibres leave a pattern assembly or one portion of an neural array and arrive together, in the same relative positions, at one portion of another pattern assembly. Fibres that conform to this rule will be described as {\em coherent}.

\item Fibres arrive at one portion of a pattern assembly from a variety of different sources or, if they come from one source, their relative positions are not preserved. This kind of arrangement may be described as {\em incoherent}.

\end{enumerate}

When signals arrive at a pattern assembly, they should produce a stronger response in the first case than in the second. A possible mechanism to ensure that this happens would be lateral connections amongst the fibres of a coherent bundle that would have the effect of increasing the rate of firing in the bundle if the majority of them are firing together. When any one fibre in the bundle is firing, then signals will be carried to neighbouring fibres via the lateral connections and these signals should lower the threshold for firing and thus increase the rate of firing in any of the neighbouring fibres that are already firing. These kinds of lateral connections would be broadly consistent with the fourth principle described above.

If signals arrive in a temporal sequence, then other possible mechanisms include `synfire chains' and temporal `sequence detectors', as described by \citet{pulvermuller_2002}. However, these would not do for keeping track of order in spatially-distributed patterns, seen in a single glance.

\subsubsection{Keeping track of order in production}\label{order_relations_production}

In speaking or writing or comparable activities not involving language, we need to do things in the right order. The spatial arrangement of neurons in a pattern assembly can provide information about that order but supplementary mechanisms would be needed to control the timings of actions and the use of sensory feedback to monitor progress. 

\subsection{Constancies}\label{constancies_neural}

\index{perception!constancies|(}

A possible objection to the SP-neural proposals is that they are inconsistent with the `constancy' phenomena in perception. These include:

\begin{itemize}

\item {\em Size constancy}. We can recognise an object despite wide variations in the size of its image on the retina---and we judge its size to be constant despite these variations.\footnote{Size constancy is similar to but distinct from `stimulus equivalence'---our ability to recognise a triangle, for example, as a triangle despite variations in size. Size constancy compensates for the effect of variations in distance while stimulus equivalence may be seen as a form of generalisation, comparable with our ability to recognise a triangle as a triangle despite variations in each of its three angles.}

\item {\em Brightness constancy}. We can recognise something despite wide variations in the absolute brightness of the image on our retina (and, likewise, we judge its intrinsic brightness to be constant).

\item {\em Colour constancy}. In recognising the intrinsic colour of an object, we can make allowances for wide variations in the colour of the light which falls on the object and the consequent effect on the colour of the light that leaves the object and enters our eyes.

\end{itemize}

If the pattern recorded in an neural array was merely a copy of sensory input there would indeed be wide variations in the size of the visual pattern projected by a given object and in brightness and colours. The mechanism described in the previous subsection for detecting coherence would not work for patterns whose elements were widely dispersed across the neural array. And recognising the elements of a pattern would be more complicated if there were wide variations in brightness or colour.

The suggestion here is that much of the variability of sensory input from a given object has been eliminated at a stage before the information leaves the neural array. As was noted in Section \ref{adaptation_and_inhibition}, lateral inhibition in the retina emphasises boundaries between relatively uniform areas. The redundant information within each uniform area is largely eliminated which means that it is, in effect, shrunk to the minimum size needed to record the attributes of that area. Since this minimum will be the same regardless of the size of the original image, the overall effect should be to reduce or eliminate variations in the sizes of images from a given object (see also Sections \ref{2d_patterns_boundaries} and \ref{encoding_2d_patterns_compressed}).

In a similar way, the `primal sketch' created by lateral inhibition should be largely independent of the absolute brightness of the original image. And adaptation in the early stages of visual processing should to a large extent prevent variations in brightness having an effect on patterns reaching the neural array.

Edwin Land's `retinex' theory suggests that colour constancy is achieved by processing in the retina and in the visual cortex \citep[see][pp. 437--439]{nicholls_etal_2001}. This is consistent with the idea that this source of variability has been removed at a stage before sensory input is compared with stored patterns.%
\index{perception!constancies|)}

\section{Neural mechanisms for learning}\label{learning_section}

\index{learning|(}

As we have seen in Chapter \ref{learning_chapter}, SP70---like SP61---tries to encode New information as economically as possible by encoding it in terms of Old information. Like SP61, it does this by trying to encode New information economically in terms of patterns stored in Old but, in addition, it builds the repository of Old patterns by adding patterns to Old as described in Chapter \ref{learning_chapter}.

\subsection{Creation and storage of patterns}\label{pattern_creation}

Patterns that are added to Old in the learning process would, of course, be realised as pattern assemblies. The SP70 style of learning implies that pattern assemblies may be created in response to patterns of incoming information that have occurred only once. This contrasts with Hebb's proposal that ``Any frequently repeated, particular stimulation will lead to the slow development of a `cell-assembly'...'' \citeyearpar[p. xix]{hebb_1949}. This difference between the two views of learning is discussed briefly in Section \ref{speed_of_learning}, below.

Another implication of the SP70 style of learning is that there is some kind of mechanism that can add ID-neurons to each newly-created pattern assembly. We need also to suppose that the system can create connections between pattern assemblies of the kind described earlier.

It seems unlikely that the neurons of each newly-created pattern assembly would be grown according to need. Although there is now evidence that new nerve cells can grow in mature brains \citep{shors_gould_2001}, it seems likely that the bulk of neurons in the brain are present at birth and that the creation of pattern assemblies is largely a process of adapting pre-existing neurons. We may also suppose that the postulated lateral connections between neurons within each pattern assembly are also largely present at birth. These suppositions accord with the observation that learning occurs in the marine snail {\em Aplysia} although the neural framework is largely fixed and essentially the same from one individual to another \citep{kandel_2001}.

It is possible that the boundaries of pattern assemblies may be established by breaking pre-existing lateral connections. Something like this is suggested by evidence for the progressive weakening of connections between cells that do not normally fire at the same time \citep[][pp. 254--255]{pulvermuller_1999}.

In a similar vein, it is not necessary to suppose that connections between pattern assemblies are grown from scratch according to need. In many cases, this would be a cumbersome process given that many of the connections between different areas of the cortex are fairly long (see also Section \ref{speed_of_learning}, below). It is now generally accepted that short-term memories are mediated by relatively rapid changes in the ability of synapses to conduct nerve impulses and that long-term memories are mediated by slower but more long-lasting changes in synapses and also by the growth of new synapses \citep{kandel_2001}.

\subsection{Finding good partial matches between patterns}

A key part of the learning process in SP70 is the formation of partial matches between patterns like the ones shown in Chapter \ref{learning_chapter}. The kind of neural processing described in Section \ref{building_mas} should be able to achieve this kind of partial matching automatically, without the need for any supplementary mechanisms.

In the `inhibitory' scheme, lateral inhibition should have the effect of marking the boundaries between parts of a pattern assembly that are receiving signals and parts that are not. And this marking of boundaries would presumably be helpful in the process of creating new pattern assemblies, derived from matched and unmatched parts of the given pattern assembly.

\subsection{Counting}\label{counting_section}

In the SP70 model, each pattern has an associated frequency of occurrence. These frequency values are used periodically to adjust the numbers and sizes of ID-symbols assigned to each pattern, in accordance with the Shannon-Fano-Elias coding scheme \citep[see][]{cover_thomas_1991}.

It seems possible that information about the frequency of occurrence of each pattern assembly may be recorded by the `strength' of the intra-assembly synapses (as in Hebbian cell assemblies) or, perhaps, by the predisposition of cells to fire in response to incoming stimulation or by the strength of their response when they do fire. It seems possible that this kind of information may lead to adjustments in the numbers of ID-neurons assigned to each pattern assembly.

\subsection{Speed of learning}\label{speed_of_learning}

In Hebb's theory of learning, cell assemblies are built up gradually by the progressive strengthening of connections between cells that regularly fire at the same time. Although this seems to accord with the slow build up of knowledge throughout childhood and beyond, it does not sit well with the undoubted fact that we can remember things that we have experienced only once---what we had for lunch today, our first day at school, and so on. It is true that unique events are probably encoded in terms of things we already know but we need to be able to store that encoded representation without the need for repetition.

Because the slow growth of cell assemblies does not account for our ability to remember things immediately after a single exposure, Hebb adopted a `reverberatory' theory for this kind of memory. But, as \citet{milner_1996} points out, it is difficult to understand how this kind of mechanism could explain our ability to assimilate a previously-unseen telephone number. Each digit in the number may be stored in a reverberatory assembly but the way in which we remember the {\em sequence} of the digits is unexplained.

In the learning scheme outlined above, new pattern assemblies can be created in response to a single sensory input. This is consistent with our ability to remember unique events and sequences of events.\footnote{It also suggests that our brains would soon become cluttered with thousands or millions or pattern assemblies, created in response to specific observations and events. But an important part of the learning process described in Chapter \ref{learning_chapter} is a mechanism for purging the system of patterns that are not proving useful. This should prevent the growth of new patterns or pattern assemblies getting out of hand.} And, if we suppose that synapses can be switched on or off according to need, we may be able to account for the speed with which immediate memories can be established. But an ability to lay down new pattern assemblies relatively quickly does not explain why it takes several years to learn something like a language.

The principles on which SP70 is based suggest why learning a language, and similar kinds of learning, take quite a lot of time. The abstract `space' of alternative grammars for any natural language is astronomically large and it takes time to search. Although exhaustive search is not feasible, the kinds of heuristic techniques used in the SP70 model---trading accuracy for speed---can bring search times within the bounds of practicality. But finding a tolerably good grammar is still a complex problem and it cannot be solved in an instant.

\subsection{Empirical validation}

Although SP70 is not well enough developed to be compared in detail with the way children learn to talk, it is anticipated that, in the sphere of learning, it will behave much like the earlier SNPR model \citep{wolff_1988, wolff_1982}. This is because both models are founded on MLE principles and it seems that these principles can account for many features of the way children learn language:

\begin{itemize}

\item Unsupervised learning of conjunctive segmental structures (words, phrases etc) from unsegmented linguistic input \citep[see also][]{wolff_1980, wolff_1977, wolff_1975}.

\item Unsupervised learning of disjunctive categories (nouns, verbs etc) from unsegmented linguistic input.

\item Unsupervised learning of hierarchical grammars with conjunctive and disjunctive structures at any level.

\item generalisation of grammatical rules and unsupervised correction of overgeneralisations.

\item The pattern of changes in the rate of acquisition of words as learning proceeds. The slowing of language development in later years.
 
\item The order of acquisition of words and morphemes.

\item Brown's (1973) Law of Cumulative Complexity.

\item The S-P/episodic-semantic shift.

\item The learning of 'correct' forms despite the existence of errors in linguistic data.

\item The learning of non-linguistic cognitive structures. 

\item The word frequency effect.

\end{itemize}

\noindent These things are discussed quite fully in \citet{wolff_1988}.%
\index{learning|)}

\section{Discussion}

This section considers a selection of other issues, including some difficulties with Hebb's original proposals \citep[described by][]{milner_1996} and how, in the SP-neural proposals, they may be overcome.

\subsection{Stimulus equivalence and discrimination}\label{stim_equiv_discrim}

An important motivation for the cell assembly\index{cell assembly} concept was to explain the phenomenon of `stimulus equivalence' or `generalisation'---our ability to recognise things despite variations in size, shape, position of the image on the retina, and so on. However, Milner writes that ``generalisation is certainly an important and puzzling phenomenon, but discrimination is equally important and tends to be ignored in theories of neural representation. Any theory of engram formation must take into account the relationship between category and instance---the ability we have to distinguish our own hat, house, and dog from hats, houses and dogs in general.'' ({\em ibid.}, p. 70).

The SP-neural proposals account for generalisation in four main ways:

\begin{itemize}

\item As with the original cell assembly concept, a pattern assembly can respond provided that a reasonably large subset of its neurons has received inputs. This accommodates our ability to recognise things despite omissions, additions or substitutions in the sensory input relative to the information stored in the pattern assembly. The effect is similar to the dynamic programming built into the SP models.

\item In the present scheme, each neuron or feature detector in an neural array is connected to each one of the lowest level pattern assemblies that contain the corresponding feature. Thus, each of these pattern assemblies may respond to appropriate input regardless of the position of the input on the neural array.

\item The way in which perceptual constancies may be accommodated in the present proposals was considered in Section \ref{constancies_neural}. 

\item Owing to the provision of a `reference' mechanism in the present proposals (discussed in Section \ref{hierarchical_organisation}, above), it is possible to create pattern assemblies that represent abstractions from sensory input. An example from Figure \ref{neural_basis_figure} is the pattern assembly `S $<$ N $>$ $<$ V $>$'. This pattern assembly represents the abstract structure of a simple sentence and it may be activated by a range of {\em alternative} inputs: any sentence that conforms to the noun-verb pattern.

\end{itemize}

The SP system provides a neat account of our ability to distinguish specific instances from the categories in which they belong. Given a pattern to be recognised (in New) and appropriate patterns in Old, the SP61 model can build multiple alignments which include several different levels of abstraction, including a level that represents a specific instance (Section \ref{class_part_inheritance}). The SP system also accounts for the variability of individual objects: any specific object such as `Tibs' is, in effect, a {\em class} of even more specific observations as described in Section \ref{objects_classes_metaclasses}.

If the SP concepts can be realised with neural mechanisms, we can account in neural terms for recognition at multiple levels of abstraction, including the levels corresponding to a specific instance and a specific instance in various contexts.

\subsection{Parts, wholes and associations}\label{parts_wholes_assocs}

With respect to Hebbian concepts, Milner raises some other questions: ``How do associations between cell assemblies differ from internal associations that are responsible for the assemblies properties? It does not seem likely that both these processes can be the result of similar synaptic changes as is usually assumed. If they were, the interassembly associations would soon become intraassembly loops. A related puzzle is that parts are not submerged in the whole. Doors and windows are integral parts of my concept of a house, but they are also robust, stand-alone concepts.'' ({\em ibid.}, p. 71). Later on the same page he writes: ``Perhaps the toughest problem of all concerns the fact that we have many associations with almost every engram....The brain must be a veritable rat's nest of tangled associations, yet for the most part we navigate through it with ease.''

\citet[p. 105]{hebb_1958} provides a possible answer to the way in which parts may be distinguished from wholes: ``If two assemblies A and B are repeatedly active at the same time they will tend to become `associated,' so that A excites B and vice versa. If they are always active at the same time they will tend to merge in a single systems---that is, form a single assembly---but if they are also active at different times they will remain separate (but associated) systems. (This means that exciting part of A, for example, has a very high probability of exciting all of A, but a definitely lower probability of exciting a separate assembly, B; A may be able to excite B only when some other assembly, C, also facilitates activity in B).''

The reference mechanism in the SP-neural proposals provides a neater solution to this problem and the others raised by Milner:

\begin{itemize}

\item Associations between pattern assemblies may be encoded by building a new pattern assembly containing references to the pattern assemblies that are to be associated. These associations are then internal to the new pattern assembly and the original pattern assemblies retain their identity.

\item In the same way, there can be a stand-alone pattern assembly for each type of component of a house while the pattern assembly for a house comprises a collection of {\em references} to the components. In this way, the concept of a house does not become muddled with concepts for doors, windows etc.

\item It is true that most concepts have many associations but the reference mechanism allows these to be encoded in a way that does not disturb the integrity of each concept. The coherence of a web page is not affected by links to that page from other pages and it makes no difference how many such pointers there may be.

\end{itemize}

\subsection{Direct and distributed neural representations}\label{direct_distributed}

In the terms described by \citet{gardner-medwin_barlow_2001}, the SP-neural scheme for representing knowledge may be classified as {\em direct} because each neuron serves one representation and only one representation. In this respect it contrasts with other {\em distributed} schemes in which any one cell may be involved in two or more different representations.

Arguments that have been made in favour of distributed representations include:

\begin{enumerate}

\item Large numbers of stimulus events may be distinguished with relatively few cells.

\item Provided the representations are `sparse' (only a small proportion of cells are involved in each recognition event), they can function as content-addressable memories that store and retrieve amounts of information approaching the maximum permitted by their numbers of modifiable elements.

\item Given that counting is needed for learning (see below), distributed representations have the flexibility to represent and count recognition events that have not been foreseen.

\end{enumerate}

Arguments in favour of direct representations include:

\begin{enumerate}

\item Starting from the reasonable premise that learning requires an ability to count the frequencies of perceptual entities, Gardner-Medwin and Barlow argue that ``compact distributed representations (i.e., ones with little redundancy) enormously reduce the efficiency of counting and must therefore slow reliable learning, but that this is not the case if they are redundant, having many more cells than are required simply for representation.'' ({\em ibid.}, p. 480). This need for redundancy offsets the first of the advantages listed for distributed representations. Gardner-Medwin and Barlow also acknowledge that this kind of redundancy is not required to achieve counting with direct representations.

\item As discussed in Sections \ref{stim_equiv_discrim} and \ref{parts_wholes_assocs}, there are problems associated with distributed representations but these can be overcome in the SP-neural version of direct representation.

\item In the SP-neural scheme, compression mechanisms allow knowledge to be represented in a direct manner with a minimum of redundancy.

\item In the SP-neural scheme, direct representations can function as content-addressable memory.

\item As discussed in Section \ref{pattern_creation}, it seems possible that pattern assemblies in the SP-neural proposals may be created relatively quickly in response to unforeseen events. And once created, they may count recognition events as described in Section \ref{counting_section}.

\end{enumerate}

\subsection{Compression revisited}\label{compression_revisited}

With respect to the r{\^o}le of information compression in neural functioning, \citet{barlow_2001_bbs} writes: ``There has been one major change in my viewpoint. Initially I thought that economy was the main benefit to be derived from exploiting redundancy, as it is for AT \& T and British Telecom. But ... the physiological and anatomical facts do not fit the idea that the brain uses compressed, economical, representations, and one can see that these would be highly inconvenient for many of the tasks that it performs, such as detecting associations. Therefore I now think the principle is redundancy {\em exploitation}, rather than {\em reduction}, since performance can be improved by taking account of sensory redundancy in other ways than by coding the information onto channels of reduced redundancy.'' (p. 604).

Elsewhere \citeyearpar{barlow_2001_network}, Barlow suggests that interest in the statistics of sensory messages ``was right in drawing attention to the importance of redundancy in sensory messages because this can often lead to crucially important knowledge of the environment, but it was wrong in emphasising the main technical use for redundancy, which is compressive coding. The idea points to the enormous importance of estimating probabilities for almost everything the brain does, from determining what is redundant to fuelling Bayesian calculations of near-optimal courses of action in a complicated world.'' (p. 242).

With respect to compressive coding (but not the importance of probabilistic reasoning), this revised view is somewhat at odds with what has been proposed in this chapter (apart from the suggested need for multiple copies of stored information (Section \ref{multiple_copies})). There is insufficient space here to discuss these issues fully but a few comments may be made:

\begin{itemize}

\item \citet{barlow_2001_bbs} writes that there is ``more than a thousand times as many neurons concerned with vision in the human cortex as there are ganglion cells in the two retinas ... on any plausible assumptions the capacity of the cortical representation is vastly greater than that of the retinal ganglion cells, so redundancy appears to be increased, not reduced.'' (p. 605). In the SP-neural proposals, the cortex is the repository of most of our stored knowledge. Even allowing for the compression that can be achieved using reference mechanisms, we need all that space to store even a small fraction of the huge amount of information that impinges on our senses (Section \ref{storage_capacity}, above).

\item As we saw in Section \ref{nature_of_counting}, counting, in itself, implies information compression by the unification of matching patterns.

\item As we saw in Section \ref{direct_distributed}, the need for efficient counting seems to demand the provision of redundancy in distributed representations. This may be seen as an argument for redundancy in stored information but equally it may be seen as an argument in favour of SP-style direct representation where no such redundancy is required.

\item Contrary to the suggestion that ``compressed, economical, representations ... would be highly inconvenient for ... detecting associations'', such representations in the SP-neural proposals provide a foundation for the detection of associations and they are indeed an expression of those associations.

\item As was argued in Section \ref{biology_engineering}, we would expect organisms that have evolved by natural selection to exploit information compression as a means of economising on the transmission or storage of information.

\item There is a lot of other evidence for the importance of redundancy reduction in brain function, some of which was reviewed in Chapter \ref{foundations_chapter}.

\end{itemize}

\subsection{Motivation, intention and attention}

\index{motivation|(}\index{intention|(}\index{attention|(}

The SP theory and the SP-neural proposals described in this chapter have little to say about such things as motivation, intention or selective attention. These things are clearly important in perception and learning \citep{milner_1999} but, so far, they have remained largely outside the scope of the SP theory apart from the brief comments in Section \ref{motivation_and_emotion_in_learning}.

Each of the SP models processes New information in sections, one pattern at a time. SP61 is also able to divide a New pattern into `windows' and process successive windows one at a time. Each of these things may be seen as a very crude version of selective attention. Their main function in the models is to provide a means of reducing the abstract `space' of alternative solutions that needs to be searched at any one time. This may indeed be one of the functions of selective attention in humans and other animals but more work is clearly needed to understand how concepts of motivation, intention and attention might relate to current or future versions of the SP theory.%
\index{motivation|)}\index{intention|)}\index{attention|)}

\section{Conclusion}

This chapter has attempted to show how the abstract concepts of the SP scheme may be realised 
with neural structures and processes. The potential payoff from this exercise is that the substantial explanatory power of the SP system may be imported into neuroscience.

What has been proposed may be seen as an extension and development of Hebb's original ideas. By incorporating ideas from the SP system, especially the mechanism which allows one pattern to `refer' to another, many of the weaknesses of the original cell assembly\index{cell assembly} concept may be overcome.

These proposals offer an heuristic framework and a source of hypotheses for further exploration and development.%
\index{neural!processing|)}

%% file: psychology.tex
\chapter[SP and Cognitive Psychology]{The SP Theory and Cognitive Psychology%
\protect\footnote{This chapter is based on a draft of a journal article prepared by Emmanuel Pothos of Edinburgh University. I am very grateful to Emmanuel for this work and for giving me permission to base this chapter on it. All errors and oversights are, of course, my responsibility.}}%
\label{psychology_chapter}

\section{Introduction}

Research on language learning provided the original inspiration for the SP theory and, as we saw in Chapter \ref{foundations_chapter}, psychological phenomena and aspects of the workings of brains and nervous systems provides much of the motivation for the theory. However, in most of the preceding chapters, the theory and its applications have been considered mainly from the perspective of computing and AI. In this chapter, the SP theory is considered in relation to research that is more explicitly concerned with cognitive psychology, excluding the neural dimension discussed in the previous chapter.

Given the very broad scope of the theory, I have only attempted to examine a selection of the areas to which it is relevant. Section \ref{recognition_categorization} is concerned with how we recognise things as belonging to this or that concept or category. Section \ref{psychology_of_reasoning} looks at psychological aspects of reasoning. Sections \ref{psychology_associative_learning}, \ref{artificial_grammar_learning} and \ref{psychology_language_learning} consider three aspects of learning: associative learning, experimental studies of how people learn artificial grammars, and the way children learn a first language. And Section \ref{analogy_structure_mapping} is an all-too-brief comparison between SP concepts and concepts developed in research on analogies, `structure mapping' and related topics.

\section{Recognition and Categorization}\label{recognition_categorization}

\index{perception!recognition|(}\index{class!classification|(}

Broadly speaking, there are four influential approaches to understanding concepts, classes or categories and the way in which we recognise things as belonging to this or that category \citep[see reviews by][]{chater_hahn_1997, komatsu_1992}:

\begin{enumerate}

\item According to the `classical' view, each mental category comprises a definition of what kinds of object belong in that category \citep[e.g.,][]{katz_1972, katz_fodor_1963}. For example, my concept of a chair may be defined as something like ``A piece of furniture composed of a seat, legs, back, and often arms, intended to seat a single person.'' Using this kind of information, I can decide whether or not any newly-observed object is a chair or not.

\item According to the `prototype' view, each concept is an abstract, summary representation of all the exemplars of that concept that we have come across. Thus, when I see a new object, I can decide whether it is a chair or not by examining how similar it is to the average, prototypical, representation I have for chairs \citep{homa_etal_1981, reed_1972}.

\item For `exemplar' models, each concept is just a collection of individual exemplars: I label a subset of the instances in my experience as `chairs' and this is what my concept of a chair corresponds to. Assignment of a new instance to a given concept depends on the similarity of that instance to the exemplars of that concept and likewise for rival concepts to which the instance may be assigned \citep[e.g.,][]{nosofsky_1989, nosofsky_1990}.

\item `Explanation' or `theory-based' views question the centrality of similarity in exemplar and prototype models and suggest that what drives the assignment of a new instance to any given category is a complex process based on general knowledge and na{\"i}ve (as opposed to scientific) understanding of the world. Thus, a chair is an object that has a certain functionality in a particular context and a specific relation to objects of similar categories (such as tables, armchairs, sofas etc. \citep{murphy_medin_1985, medin_schaffer_1978, medin_wattenmaker_1997}. More generally, some categories will be psychologically more intuitive than others, and the theory-based view of categorization has been the main theoretical approach to understand such differences in category coherence \citep[see also][]{pothos_chater_2002}.

\end{enumerate}

\subsection{Necessary, Sufficient and Characteristic Features---and Polythetic Categories}

In the definitional view of categorization and, to some extent in the other views, concepts are seen to have `features', meaning attributes or properties that have a role to play in the recognition process.

In some cases, a feature can be seen to be a `necessary' or `critical' part of a concept in the sense that the given feature must be present in every instance of the given concept.

In other cases, a feature (or combination of features) may be seen to be `sufficient' for assigning something to a given category in the sense that, if the feature is present in a given entity, then that entity must be a member of the category---but not all members of the category have the given feature (or combination of features).

More often, features of a concept are merely `characteristic', meaning that they are neither necessary nor sufficient but do have a role to play in the process of categorization. For example, most birds can fly but some birds such as penguins, ostriches and so on, cannot fly---and being able to fly is not an exclusive attribute of birds.

It is widely recognised that many of the `natural' categories that we use in everyday thinking are `polythetic'\index{class!polythetic} \citep{sokal_sneath_1973} or `family resemblance' categories (attributed to Wittgenstein) meaning that, for each such category, {\em all} its features are characteristic. This means that no single attribute is necessarily present in every member of the given category and there need be no single attribute that is exclusive to that category.

\subsection{Empirical Evaluation}

To date, the weight of evidence is against the definitional view of concepts as originally stated. However, several researchers have suggested that the available evidence is compatible with a modified version of the definitional view in which features are mainly `characteristic'. 

With regard to prototype and exemplar models, it has been shown that while certain types of prototype models are identical to certain types of exemplar ones, this is far from the case more generally \citep[see, for example,][]{mckinley_nosofsky_1995}. Unlike exemplar models, the prototype view predicts that experience with a set of instances leads to the creation of the prototype representation. This prediction has received some support from the observation that studying a set of exemplars apparently leads to a `memory' for the prototype of these exemplars, even if the prototype was not included among the training stimuli \citep[see the review by][]{komatsu_1992}. There is a corresponding prediction for exemplar models, whereby exemplar effects implicate memory for the corresponding instances \citep{nosofsky_zaki_1998}.

What we wish to conclude from this research is that there are, apparently, situations where the assignment of a new instance to a category is guided by the similarity which it has to some kind of prototype, and there are other situations in which similarity to exemplars appears to be more important. While these views have been traditionally presented as mutually exclusive, it is possible, as we shall see later, that they may reflect different aspects of a single categorization process.

With respect to the theory-based view, there have been some attempts to provide a principled computational model of the way in general knowledge may influence the process of classification. However, the incorporation of general knowledge into models of perception and cognition has been notoriously difficult \citep{pylyshyn_1980} and it is really only safe to conclude that classification of a new instance will partly be affected by general knowledge relating to the given instance and the categories to which it may be assigned.

A further observation that has been the subject of intensive study is that when people assign some entity to a category within a hierarchy of categories, they are normally biased in favour of some kind of `basic' or `default' level in the hierarchy \citep{rosch_mervis_1975, corter_gluck_1992, jones_1983, gosselin_schyns_1997}. For example, when we see a cat on the street we rarely think ``Look there is a mammal!'' or ``Look there is a Siamese cat!'' but rather ``Look there is a cat!''.

A variety of empirical evidence lends support to this idea of `basic-level' categories. For example, pictures can be named relatively rapidly if the name corresponds to a basic-level category and the process is slower if the required name corresponds to a subordinate or a superordinate category. In a similar vein, experimental subjects are less variable in their judgements of basic-level categories than they are in making judgements about categories at higher or lower levels \citep{rosch_etal_1976}. Again, \citet{mervis_crisafi_1982} and \citet{horton_markman_1980} showed that basic-level categories have a `privileged' status when children are asked to name things and likewise in other tasks that relate to mental categories. In general, the weight of evidence seems to favour the idea that each basic-level category describes the corresponding entities in a way that is, in some sense, optimal---neither too general nor too specific \citep{corter_gluck_1992, gosselin_schyns_1997}.

From the vast literature on categorization, some general conclusions emerge that are fairly widely accepted:

\begin{enumerate}

\item Concepts may have features that are necessary or sufficient but in many cases features are merely `characteristic'---and many of our `natural' categories are polythetic.

\item Classification of new instances is guided in some cases by prototypes and in other cases by similarity to category exemplars.

\item In many cases, general knowledge has an influence on the way classification is done.

\item Our concepts are often organized into hierarchies and within each hierarchy there is often a level that is in some sense more basic than other levels.

\end{enumerate}

In addition, the following features of human categorization are widely accepted:

\begin{enumerate}

\item There is often cross-classification in `natural' categories (see below).

\item We don't always recognise things unambiguously, as can be seen with the well-known `wife and mother-in-law' and `duck and rabbit' pictures.

\item Recognising something as belonging in a given category allows us to infer aspects of that thing that may not have been in our original observations. For example, recognising someone as a person enables us to make all sorts of inferences about what a surgeon would find inside if there was an operation.

\item Notwithstanding the phenomenon of basic-level categories, we often recognise things at two or more levels of abstraction simultaneously (see below).

\end{enumerate}

\subsection{The SP Theory and Categorization}

The SP theory provides an approach to categorization that is distinct from any of the views described above but, as we shall see, it exhibits features from all of them.

In the early stages of learning, the SP system stores sensory patterns in essentially the same form as they are received. This is because the repository of Old patterns contains relatively little information and there are limited opportunities to find good matches between patterns, to encode New information in an economical manner or to create any kind of abstract representation of the world.

At this stage of learning, patterns that are stored are relatively direct copies of patterns that have been observed in the world. At this stage, patterns may not correspond neatly to the discrete `objects' that we normally recognise because, in the SP scheme, the segmentation of the world into discrete entities is a process that depends on the matching and unification of patterns.

As Old accumulates raw patterns from the environment, there are increasing opportunities for the system to find good matches between patterns. After the early stages, the system begins to develop patterns in Old that represent a relatively abstract view of the system's sensory environment:

\begin{itemize}

\item It develops patterns (like `$<$ \%7 12 t h a t $>$' and `$<$ \%8 13 r u n s $>$' in Figure \ref{patterns_figure_1}), corresponding to words, objects or other discrete entities in the world.

\item It creates sets of patterns (like `$<$ \%9 14 b o y $>$' and `$<$ \%9 15 g i r l $>$' in Figure \ref{patterns_figure_1}) that are {\em alternatives} in a given context. For linguistic material, these sets of alternatives correspond to syntactic `classes' or `categories' like `noun', `verb' or `adjective'.

\item It creates patterns (like `$<$ \%10 16 $<$ \%7 $>$ $<$ \%9 $>$ $<$ \%8 $>$ $>$' in Figure \ref{patterns_figure_1}) each one of which is an abstract representation of a set of patterns that are similar to each other in the sense that they share certain features. In this very simple example, only two patterns are described, `t h a t g i r l r u n s' and `t h a t b o y r u n s'. These are similar to each other because they share the features `t h e' and `t r e e'. A set of patterns like this is essentially a concept, class or category of the kind we have been considering.

\end{itemize}

How do these ideas relate to the views of categorization that were outlined above? This is discussed in the subsections that follow.

\subsubsection{Exemplars}\label{exemplars_section}

As we have seen, the SP system is able to store patterns that are relatively direct copies of patterns that has been observed in the environment. Any such pattern may be regarded as an `exemplar' of the corresponding pattern in the world. Furthermore, the SP system is able to recognise when a New pattern is `similar' to an already-stored Old exemplar. This is because the system has an ability to find fuzzy matches between patterns. Thus each exemplar pattern that is stored in Old serves to define a set or category of patterns that are similar to the given pattern.

Although the SP theory captures these elements of the exemplar view, it differs from that view in one important respect. The system would not normally store a set of similar exemplars as discrete patterns without any relationship to each other. This is because the SP system is designed to look for redundancies amongst its stored patterns and it would find them in any set of patterns that are mutually similar. In this case, it would compress the set of patterns into an abstract representation (as described above) and individual exemplars would be encoded in terms of that abstract representation.

\subsubsection{Definitions and Features}\label{definitions_and_features}

Regarding the `classical' view of categorization, a pattern like `$<$ \%10 16 $<$ \%7 $>$ $<$ \%9 $>$ $<$ \%8 $>$ $>$', together with the other patterns in Figure \ref{patterns_figure_1}, may be regarded as a `definition' of the category that comprises the two original patterns. With regard to `features' in the definitional view, the SP theory can accommodate all of the notions described above:

\begin{itemize}

\item The system's ability to find fuzzy matches between patterns means that, in many cases, a category can be recognised from any reasonably large subset of its features. In each of these cases, the category is, in effect, polythetic and its attributes are characteristic rather than necessary or sufficient.

\item Even if the system were to be constrained to find only exact matches between patterns, it would still be possible to model polythetic categories. This is described in Section \ref{polythetic_categories}.

\item Although the system can accommodate polythetic categories and characteristic features, it can also model features that are necessary or sufficient or both. These things depend entirely on the particulars of the patterns that are stored in Old. To take an extreme example, if Old contains the one-letter pattern `X' and no other patterns contain the letter `X', then that letter is a feature that is both necessary and sufficient for the recognition of the category represented by the pattern `X'.

\end{itemize}

\subsubsection{Prototypes}

The pattern `$<$ \%10 16 $<$ \%7 $>$ $<$ \%9 $>$ $<$ \%8 $>$ $>$' in Figure \ref{patterns_figure_1}, together with the other patterns in the figure may be seen to be an abstract, summary representation of the original exemplars. This abstract pattern may be regarded not only as a definition of a concept (as discussed in Section \ref{definitions_and_features}) but also as a prototype.

Readers may object that the small grammar in Figure \ref{patterns_figure_1} captures all the non-redundant information in the original patterns and so it differs from the normal conception of a prototype in which there is significant loss of non-redundant information compared with the original exemplars.

Although this objection may be valid for this particular example, it is not a necessary feature of the SP system that it should always retain all the non-redundant information in the exemplars to which it is exposed. Indeed, the periodic purging from Old of patterns that are not proving useful in terms of minimum length encoding principles is an important part of the theory. It is entirely possible, within the SP system, to create an abstract representation of a set of exemplars which does not encode every detail of the originals. Given that a representation of this kind can be matched to New exemplars in a fuzzy manner, it can function very much in the same way as a prototype as normally understood.

\subsubsection{General Knowledge}

In the SP theory, `general' knowledge does not differ in any fundamental way from any other kind of knowledge. SP patterns can be used to express a wide variety of different kinds of knowledge, including knowledge of `functions', `processes', `events' or the contexts in which things typically appear. In broad terms, it appears that the SP theory is compatible with the kinds of observation that have motivated `explanation' or `theory-based' views of categorization.

\subsubsection{Hierarchies, Cross-Classification and Basic Levels in Categorization}

\index{class!hierarchy|(}\index{class!cross-classification|(}

A major strength of the SP theory is that it provides a powerful means of representing class-inclusion hierarchies and their integration with part-whole hierarchies (Section \ref{class_part_inheritance}). In accordance with the organisation of `natural' categories that we use in everyday thinking, the system can accommodate cross-classification---meaning that any entity or category may belong in two or more categories that are not themselves in a class-inclusion relationship. For example, a given person may be a `woman' and a `doctor' but the first of these is not a subclass of the second, or vice versa.

In the SP system, an unknown entity may be recognised at multiple levels of abstraction. The system may recognise something as a cat but, at the same time, it may recognise it as the lower-level category `my cat, Tibs' and as a member of the higher-level categories `mammal' and `animal'.

At first sight, the ability of the system to recognise things at multiple levels of abstraction seems to conflict the idea of basic-level categories considered above. However, the idea that, when we recognise things, there is a bias in favour of one level in the relevant class hierarchy does not mean that we are totally unaware of other levels. For example, when we recognise something as a cat, we will infer without conscious effort that, if it is alive, the creature will be breathing, that it is likely to be sensitive to various kinds of stimulation, that it is liable to get hungry, and so on. Since these attributes are ones that would naturally be recorded at the level of `animal' and not `cat', it is clear that our understanding that something is a cat does not rule out a recognition of the other categories in which the entity belongs.

It seems, then, that the SP theory is broadly consistent with the way in which recognition may involve multiple levels in a hierarchy of categories. However, given the empirical evidence in support of the concept of basic-level categories, it is pertinent to ask whether there is a place in the SP system for that kind of idea.

A tentative answer to this question is ``yes''. In the SP system, the process of recognition is modelled by the process of building multiple alignments, with each level of the relevant class hierarchy represented by a corresponding pattern in the multiple alignment. From each such multiple alignment, a code may be derived (as described in Section \ref{ma_evaluation}) that represents a compressed version of the pattern being recognised. Each level in the class hierarchy makes a contribution to the code and the contributions made by different levels are rarely equal. It seems possible that the level that makes the largest contribution---which means that it is most informative about the nature of the unknown entity---might correspond to the empirically-based concept of basic-level category. This idea is similar to proposals that have been made in other computational approaches to this issue \citep{corter_gluck_1992, gosselin_schyns_1997}.%
\index{class!hierarchy|)}\index{class!cross-classification|)}

\subsubsection{Summary}

The SP theory appears to be compatible with all the major features of human categorization that have been established in empirical studies. It exhibits features of all the main views of categorization that have been proposed. These features are `emergent' by-products of the way the SP system is organised, not merely the result of a simple marriage of the several views.

Apart from the `integrated' nature of the SP theory, the main differences between this approach to categorization and the main alternatives are:

\begin{itemize}

\item By contrast with the `exemplar' view, the system would not normally store sets of similar patterns without any modification. It would normally compress them in the kind of way that was outlined earlier.

\item The SP theory predicts that, in the course of learning, there should be a gradual transition from knowledge stored as exemplars of previously-observed patterns to knowledge in the form of relatively abstract definitions or prototypes.

\end{itemize}

Without some means of examining directly what is stored inside someone's head, it is hard to see how the first of these predictions might be checked empirically. However, there is some evidence that seems to lend support to the second prediction. It has been known for some time that, when a child learns his or her first language, irregular forms of verbs and irregular plural forms of nouns are often correct when they are first learned but are then displaced incorrectly by regular forms which themselves eventually give way to correct, adult forms \citep{slobin_1971}. This accords with the idea that our knowledge of the world is stored initially in essentially the same form that is received but may be broken down later as the system seeks to encode the information in terms of regular patterns. After regular patterns have been discovered, any over-regularisations may be corrected.%
\index{perception!recognition|)}\index{class!classification|)}

\section{Reasoning}\label{psychology_of_reasoning}

\index{reasoning|(}

For a long time it has been hypothesized that our everyday reasoning is mediated by the principles of classical logic or classical probability theory. Thus we may suppose that people draw logical `conclusions' about things in a deterministic manner from a set of `premises' and one or more logical `rules' that relate to the particular problem at hand. If, for example, we know that Tower Bridge is `in' London and that London is `in' the United Kingdom, then we may conclude that Tower Bridge is `in' the United Kingdom. For other kinds of problem, it is clear that some kind of probabilistic analysis is required. If, for example, we see black clouds in the sky, we may reason probabilistically that rain is likely.

These kinds of reasoning have been considered to be the `correct' or normative forms that distinguish rational cognitive agents (humans) from all other kinds of beings capable of some thought. Much of the research into human reasoning in the last hundred years has revolved around the issue of how the principles of classical logic and probability could be incorporated in cognitive structures \citep[e.g.,][]{evans_1991, evans_etal_1991, braine_etal_1995, rips_1989, rips_1990}.

The extent to which the reasoning process is indeed mediated by the rules of logic has been extensively investigated in the context of reasoning problems such as the Wason selection task and syllogistic reasoning \citep{wason_1960, johnson-laird_1983, johnson-laird_etal_1989}. In the Wason task, there are four cards and a conditional rule. The rule may say something like ``If there is an even number on one side of the card, then there must be a consonant on the other.'' The four cards are presented to participants in such a way that all the possibilities associated with the rule are shown: one card has a vowel on the side that is shown, another has a consonant, another an even number, and the last one an odd number. The question is which card or cards need be selected in order to check whether the rule is correct or not. Most participants select the card showing an even number---which is correct in normatively terms, since if there is a vowel on the other side of the card then the rule must be false. But the same majority of subjects also choose the card showing a consonant---which is a poor choice in terms of classical logic, since if there is an even number on the other side of the card then there is some confirmation of the rule, but there is nothing we can conclude if there is an odd number. In summary, participants fail to select the two cards (the even number card and the vowel card) that, according to classical logic, would allow for unambiguous falsification of the rule.

In probabilistic reasoning, the overall picture is somewhat similar. On the whole, people are sensitive to the laws of probability, but investigators have identified several examples where there are significant deviations from what one would predict from these laws \citep{osherson_1990, kahneman_tversky_1972}. A famous example is of the following form: Linda is described as a very active, outgoing, dynamic woman. Having read Linda's description, participants are asked to evaluate the relative probability of a set of statements regarding Linda. The critical comparisons concern the statements ``Linda spends a lot of time at home'' and ``Linda spends a lot of time at home and works at a bank.'' The description of Linda is clearly consistent with a person who might be working at bank but not with someone who likes to stay at home a lot. Thus, participants select the second statement as more probable for Linda compared to the first. But, the rules of probability tell us that the probability of a conjunctive statement will never be higher than the probability of either of the statement's constituents.

In summary, human reasoning appears to deviate significantly from what one would expect if it conformed to the principles of classical logic or classical probability theory. There are several ways in which investigators have attempted to explain these deviations:

\begin{itemize}

\item One possibility is that reasoning is based overall on rules of logic and probability, but na{\"i}ve participants comprehend reasoning problems in ways different from that assumed by the experimenter. Thus, it is the case that rules of logic are applied correctly to reach the normative conclusion, but that this conclusion corresponds to a problem different to the one intended by the experimenter.

\item A second possibility is that, along with a reasoning faculty based on classical logic and probability, people have developed a set of `heuristics' that provide speedy shortcuts to everyday reasoning problems. The theory is that these heuristics are developed to approximate the normative rules, but they are not of course guaranteed to reach the normative solution in all problems.

\item It is also possible that the reasoning process could be influenced by information that is irrelevant to its logical structure. Such information is described as `biases'. For example, the `belief bias' suggests that people would prefer to confirm a hypothesis rather than disprove it. General knowledge is a likely source of such biases.

\item Other investigators reject the relevance of classical logic and probability in reasoning but postulate other kinds of rules instead. For example, \citet{cheng_holyoak_1985} suggest that people may create for themselves and use `pragmatic reasoning schemas', each one geared to the demands of a particular domain.

\item Some investigators have pointed out that the nature of everyday reasoning (defeasibility, non-monotonicity) cannot be addressed within a formal system based on rules of logic. For example, Oaksford and Chater's \citeyearpar{oaksford_chater_1994} information-theoretic proposal states that people examine a hypothesis in terms of a prerogative to minimize uncertainty about the veracity of the hypothesis.

\item Models of reasoning based on exemplar similarity have been advocated within another influential tradition \citep[see, for example,][]{kolodner_1992, smith_etal_1992}. Here, it is hypothesised that if we have successfully solved a problem in the past then we would be able to address a new similar problem by analogy to the previous one. Thus, if our solution to a given problem involved rule X then on encountering a novel problem of the same type we could solve it not because we have abstracted rule X but because the two problems are similar (they both share the same overall structure determined by rule X). Whether the reasoning process is based on `similarity' or `rules' is a largely unresolved issue.

\end{itemize}

To summarise the foregoing, it seems that principles of classical logic and probability are relevant in at least some everyday reasoning situations. But it is also possible that other mechanisms and processes have a role to play such as misunderstandings between experimenter and subject, heuristics, biases, general knowledge, `pragmatic reasoning schemas', minimization of uncertainty, and similarities between problems.

\subsection{The SP Theory and Reasoning}

As we have seen in Chapter \ref{pr_chapter}, the SP system provides a powerful means of modelling several different kinds of probabilistic reasoning. And in Chapter \ref{maths_logic_chapter}, it was suggested that the framework may also be applied to classical kinds of reasoning with explicit values that are `true' or `false', with nothing in between.

How might the reasoning capabilities of the SP system relate to alternative theories about human reasoning and available empirical data?

With regard to the range of ideas about human reasoning that were outlined above, the SP theory is closest in spirit to the information-theoretic proposals of \citet{oaksford_chater_1994}. Both sets of proposals are founded on the pioneering work of \citet{solomonoff_1964, wallace_boulton_1968, rissanen_1978} and others but the multiple alignment concept, as it has been developed in the SP theory, is a distinctive feature of the system, not proposed elsewhere.

An important part of the empirical support for the SP theory is that it provides an account of several aspects of human reasoning that can be readily observed without the need for any formal study. With the possible exception of Bayesian networks \citep[see, for example,][]{pearl_1988}, it appears that no other theory can match the explanatory range of the SP theory in the domain of probabilistic reasoning.

It may be thought that the system's ability to model classical kinds of logic and mathematics is not relevant to the empirical validation of the theory. But it should not be forgotten that, until the 1940s, a `computer' was a person who did computations and the human brain was the sole vehicle for all aspects of logic and mathematics. Any theory of human reasoning should be able to account for the fact that, when we make the necessary effort, we can reason in accordance with established principles of logic and mathematics. Unlike some of the alternatives, the SP theory appears to satisfy this requirement.

How might the SP theory account for the fact that people do not always reason in ways that are strictly logical? In broad terms, this phenomenon may be explained by the way in which the system searches for good matches and unifications between patterns. For any realistically-large body of information, the abstract `space' of possible matches and unifications is far too large to be searched exhaustively. To be practical, any system must constrain the search in some way, using heuristic techniques or otherwise (the SP system uses a range of heuristic techniques). Without these constraints, the system will get bogged down and will achieve nothing. With the constraints, the system can often deliver results that are useful---but the penalty is that theoretically ideal results cannot be guaranteed and mistakes will be made. This seems to be the human condition!

Without detailed study, it is not possible to say whether or how the SP theory might account for the kinds of experimental results outlined above. To make a precise comparison between any cognitive model and human experimental subjects, it is necessary for the model and the subjects to be using the same knowledge. Herein lies a significant methodological problem: it is very hard to know exactly what a given person has or has not got stored in their heads! Notwithstanding this kind of difficulty, it may be possible to find ways of assessing the detailed reasoning behaviour of the SP model in comparison with people.

With regard to other ideas that have been proposed to account for human styles of reasoning:

\begin{itemize}

\item The SP theory is entirely compatible with the idea that experimental subjects might not fully comprehend what they are required to do. Any system that has the ability to make mistakes may fail in that way.

\item Given the way in which the SP system is designed to store recurrent patterns in its experience, it seems reasonable to suppose that some of these patterns might serve as `heuristics', `biases', `general knowledge' or `pragmatic reasoning schemas' with an influence on how reasoning is done.

\item Given that the detection of good matches between patterns lies at the heart of the SP system and that the system stores knowledge of recurrent patterns, it is not hard to imagine that experience with certain kinds of problems would facilitate the solution of similar problems which are encountered later.

\end{itemize}

\subsubsection{Summary}

In summary, the SP theory can model a wide range of different kinds of reasoning. Given that these different kinds of reasoning---including classical kinds of logic and mathematics---are all part of the human repertoire, they provide empirical support for the theory as a model of human reasoning.

Apart from its ability to model classical forms of reasoning, the theory suggests in general terms why human reasoning is not always correct and the theory is broadly compatible with other explanations that have been offered for the idiosyncrasies of human reasoning.

The empirical validity of the theory should, of course, be examined in more detail but, in this area, there are some difficult methodological problems to be solved.%
\index{reasoning|)}

\section{Associative Learning}\label{psychology_associative_learning}

\index{learning|(}

For many years, it has been recognized by psychologists that people have a tendency to notice and remember `associations' between objects and events. If I see that the entities A and B occur together frequently, then eventually I am likely to treat them as a single entity.

This tendency, which is a matter of observation, has led to the development of `associative' learning theories in which learning is achieved by connecting entities or events in a pairwise manner. This kind of mechanism is the basis of the MK10 computer model of the way in which children may learn the segmental structure of language \citep{wolff_1975, wolff_1977, wolff_1980}. Similar mechanisms have been incorporated in the SNPR model of syntax learning \citep{wolff_1982, wolff_1988}. More recently, ideas of this sort have been proposed as a `competitive chunking' hypothesis for learning \citep{servan-schreiber_1991, servan-schreiber_anderson_1990}.

In theories of this kind, chunks are the elements of an hierarchical structure in long-term memory, with higher-level chunks constructed recursively in a pairwise manner from lower-level chunks. For each node within such a hierarchy, there is some kind of measure of the strength of association, related to the frequency of occurrence of the association and perhaps also to the recency with which the association has been observed.

Some theories of associative learning embody a distinction between outcomes and cues that allows for explicit teaching of certain associations. For example, according to the Rescorla-Wagner model \citet{rescorla_wagner_1972}; see also \citet{mackintosh_1975}), different events are characterized by an association strength, which reflects how salient their relatedness is to an observer. For each occurrence of a particular outcome following a cue, the outcome-cue association is strengthened on the basis of a learning rule that is similar to the delta rule common in connectionist models \citep[e.g.,][]{plunkett_elman_1996}. The upshot of this process is that eventually some cues are more readily associated with certain outcomes.

\subsection{The SP Theory and Associative Learning}

Part of the inspiration for the SP theory was the hierarchical chunking models of language learning mentioned above. However, it was recognised at an early stage in the development of the SP theory that the pairwise chunking technique used in the earlier models would not provide a good model for such things as fuzzy pattern recognition and probabilistic reasoning that the new theory aimed to embrace. At the same time, it was intended that the new theory should be compatible with the empirical fact that people do indeed have a tendency to associate things that occur together frequently.

At the heart of the SP models is a mechanism for finding full and partial matches between patterns that are good in terms of minimum length encoding principles. This mechanism can form associations between any number of entities, not just two. For example, if the system finds repeated occurrences of a pattern like `p l a y', it will normally unify them to form a single instance of that pattern. This single instance represents an association amongst the four letters of the pattern. It may also be seen as a `chunk', as that term in normally understood.

Clearly, this kind of mechanism can form associations in a pairwise manner and it is thus compatible with the empirical observation that people do tend to associate pairs of things that regularly occur together. But the mechanism is not, in itself, a mechanism for forming pairwise associations---it is much more general than that. Like any associative learning theory, the SP theory can build hierarchical structures in a pairwise manner. But it can also build hierarchies in which associations amongst three or more entities are identified in a relatively direct manner. And the SP mechanisms for finding good full and partial matches between patterns provide a basis for modelling several other phenomena that are well beyond the scope of any associative learning theory.

\subsubsection{Frequency and Size}

Where does frequency fit into the picture? In associative learning theories, the frequency of associations is the central driver. In the SP theory, by contrast, the key principle is the amount of compression that can be achieved by the unification of matching patterns. Clearly, more compression can be achieved by unifying many instances of a pattern than by unifying only a few. And the SP theory also incorporates the Huffman principle that compression can be enhanced by using short identifiers for patterns that occur frequently and longer identifiers for patterns that are relatively rare.

So the SP theory, and the minimum length encoding principles on which it is based, are compatible with the observation that people tend to associate things that occur together frequently---but frequency is only part of the picture. The other major factor that affects the amount of compression that can be achieved is the {\em sizes} of patterns to be unified. Clearly, the unification of two or more large patterns yields more compression than the unification of the same number of small patterns.

For any given size of pattern, there is a minimum frequency below which no compression can be achieved. This is because each unified chunk requires some kind of label or identifier by which it can be referenced---and the information `cost' of these labels offsets the compression achieved by the unification of matching patterns. Unless that compression is greater than the amount of information required for the labels, there is no net saving in the number of bits that are needed.

In this connection, there is a trade-off between sizes of patterns and the minimum frequency that is needed for compression. With large patterns, useful compression can be achieved when frequencies are as low as 2 or 3. With small patterns, including patterns with only two elements, the minimum frequency needed to achieve useful compression is normally much larger than 2 or 3.

\subsubsection{Summary}

In summary, the SP theory can be seen to be compatible with the observation that people can form associations between pairs of entities and that frequently-occurring pairs are most likely to be learned. But the SP theory is much more general than any of the theories that propose that learning actually occurs in a pairwise manner, with frequency as the central driver.

The SP theory makes predictions that are different from those made by associative learning theories and, in principle at least, these predictions could be tested experimentally:

\begin{itemize}

\item The main difference between the two views of learning is that the SP theory predicts, with some qualification, that the amount of repetition required to isolate a large pattern as a discrete chunk is less than would be required for a small pattern, whereas associative learning theories make no such prediction. The qualification is that there is a limit to the sizes of patterns that can be matched in one operation. Patterns that are larger than this maximum must be processed in stages---which would explain why an actor takes longer to learn a large part than a small one. But for patterns that fall within the scope of one `percept', the isolation of large chunks should require fewer repetitions than small ones.

\item Another difference is that, in associative learning theories, large chunks are always built up in a pairwise manner from smaller components whereas the SP theory proposes that large chunks may be learned in a more direct manner---although pairwise learning is still possible. As before, the picture is more complicated in the case of patterns that are too large to be perceived in one operation.

\end{itemize}

\index{learning|)}

\section{`Rules', `Similarity' and Artificial Grammar Learning}\label{artificial_grammar_learning}

\index{learning!artificial grammar|(}

In this section we focus on the issue of whether experience is coded primarily as `rules' or in terms of `similarity'---an `opposition' of the kind that was criticised by Allen Newell (Section \ref{breadth_and_depth}). We do this in the context of artificial grammar learning since this is an experimental paradigm where many of the more popular theories of learning have been applied and compared.

Artificial grammar learning is a learning task that typically uses stimuli derived from a finite-state grammar. A finite-state grammar is a relatively simple system for the creation and recognition of sequences of symbols. This kind of system is quite widely used in computer applications but it was recognised many years ago that it has less expressive `power' than other systems such as PSG or transformational grammar and that it is not adequate for describing the complexity and subtleties of certain forms of knowledge such as natural language syntax \citep{chomsky_1957}. 

For any given finite-state grammar, symbol sequences that are consistent with its rules are described as `grammatical', while others that are not consistent with the rules are classified as `nongrammatical'. As we read a grammatical sequence from left to right, the range of symbols that we may see at any given point in the sequence is constrained by what has gone before. For example, one finite-state grammar might specify that the first symbol can only be `M' or `V', that if `M' is the first symbol then it must be followed by an `X' or another `M' or an `S'---and so on. Thus, while `MS' would be a valid prefix for a grammatical sequence for that finite-state grammar, `XM' would not be.

In a typical artificial grammar learning task, participants are first presented with a subset of the possible grammatical sequences, with instructions only to observe them. They are not told any of the rules of the grammar or what will be asked from them in the subsequent test phase. In the test phase, they are presented with grammatical and non-grammatical sequences and their task is to discriminate between the two. It has been found on repeated occasions that participants can do this kind of task with an accuracy that is significantly better than chance. The main focus of artificial grammar learning studies has been an examination of alternative hypotheses about the kinds of knowledge that may guide participants' selections.

It was thought originally that if experimental subjects are sensitive to the grammatical/non-grammatical distinction, then because that distinction is defined by the rule structure of a finite state language, the subjects must have learned something about those rules \citep{reber_1967, reber_1989, reber_allen_1978}. More specifically, it was assumed that participants develop an abstract, partial representation of the rule structure of the finite state language, which can be used to decide whether a given test sequence is grammatical or non-grammatical. A variant of Reber's account is the `microrules' hypothesis \citep{dulany_etal_1984, dulany_etal_1985} which proposes that the knowledge acquired by experimental subjects takes the form of heuristic-like tests of limited validity and scope. It is supposed that these tests can be used more or less independently of each other to determine the grammaticality of different sequences.

Several investigators have challenged the claims of Reber and his associates by showing that other types of knowledge covary with the grammatical/non-grammatical distinction. This implies that, even if participants can successfully discriminate between grammatical and non-grammatical sequences, their achievement may not necessarily be based on a knowledge of the grammatical rules. Vokey and Brooks \citeyearpar{vokey_brooks_1992, vokey_brooks_1994} suggested that, in Reber's original studies, grammatical items in the test part of his artificial grammar learning task were similar to the training items and they argued that the results that Reber obtained could be explained in terms of similarity rather than grammaticality.

Vokey and Brooks defined the similarity of a test item to a training item in terms of the number of symbols that would need to be changed in order to convert one into the other. Using this definition, they devised one set of test sequences in which the grammatical and non-grammatical items were all similar to the training sequences, and another set in which the grammatical and non-grammatical items were all rather different from the training set. Using these materials, they found that the performance of experimental subjects was affected both by the grammatical status of the test sequences and by their similarity to the training sequences.

\citet{pothos_bailey_2000} used a more general approach to similarity, one based on empirically-derived similarity ratings between the items, consistent with Nosofsky's generalised Context Model of similarity \citep{nosofsky_1989, nosofsky_1990}. As in the studies by Vokey and Brooks, these investigators found that the performance of experimental subjects was affected both by the grammatical status of the test items and by their similarity to the training sequences. Similarity seemed to have the strongest influence.

Results from artificial grammar learning experiments have also been interpreted in terms of associative theories of learning, discussed in the previous section. In the training phase of an artificial grammar learning experiment, it will normally happen that some pairs of symbols appear more frequently than others. Associative learning theories suggest that any such pairs should be recognised first as discrete `bigrams' and that larger structures may then be built in a pairwise manner with frequency as the driving heuristic.

\citet{perruchet_pacteau_1990} showed that participants in artificial grammar learning experiments were indeed sensitive to bigram information. Furthermore, they showed that knowledge of bigrams was sufficient to account for the observed levels of performance and they suggested that it was not necessary to postulate forms of knowledge such as `rules'. \citet{knowlton_squire_1996} presented a more sophisticated `chunking' model of artificial grammar learning based on associative learning principles. They concluded that the performance of experimental subjects could be accounted for by a hybrid of their chunking model and the operation of `rules'. Some doubt has been expressed about whether associative learning approaches to artificial grammar learning are really distinct from rules or from concepts of similarity \citep{dickinson_2001, cheng_1997}.

Part of the appeal of artificial grammar learning is that the range of different hypotheses about learning can be compared with each other by applying them to concrete examples. This has been done by \citet{johnstone_shanks_1999} and also by \citet{pothos_bailey_2000}. The main finding in both cases was that no single form of knowledge is sufficient to explain the selections made by experimental subjects in the test part of an artificial grammar learning experiment. To account for all the observations, it seems necessary to invoke the operation of rules and similarities and fragments of structure.

To complete this brief review of artificial grammar learning studies, mention should be made of so-called `transfer' artificial grammar learning experiments. By contrast with the artificial grammar learning studies in which the same symbols are used in the training phase and the testing phase, experiments in the `transfer' paradigm use different sets in the two phases, with a one-to-one mapping between the two. This means that, at a superficial level, there is no similarity at all between the materials used in the two phases of the experiment. Consequently, when investigators discovered that experimental subjects in these experiments displayed above chance sensitivity to the grammatical/non-grammatical distinction, it seemed natural to conclude that they must be basing their decisions on a knowledge of rules. However, it has been shown in later investigations that transfer performance could be guided by abstract analogies amongst stimuli. Thus, for example, the stimulus ``MSSSX'' has the same abstract structure as the stimulus ``HJJJL'' \citep{brooks_vokey_1991}. In a similar way, it can argued that experimental results may be explained by abstract relationships amongst fragments of stimuli \citep{redington_chater_1998}. Clearly, the transfer phenomenon is an interesting finding that needs to be considered in any discussion of artificial grammar learning. 

\subsection{The SP Theory and Artificial Grammar Learning}

What light might the SP theory throw on the issues and observations outlined above? In this connection, there are two aspects of the theory that invite attention:

\begin{itemize}

\item As was noted in Section \ref{representation_of_knowledge}, {\em all} kinds of knowledge in the system are expressed as `patterns'. There is no formal distinction between `rules' or `fragments' or any other format for knowledge. That said, the system can model a wide variety of representational schemes, including finite-state grammars, context-free phrase-structure grammars and context-sensitive grammars.

\item Although `similarity' has no formal status in the theory, concepts of that sort are implicit in the way the system works:

\begin{itemize}

\item A method of finding full matches and good partial matches between patterns lies at the heart of the system. In effect, the system is dedicated to finding similarities amongst patterns.

\item Abstract structures created by the system in the course of learning can be seen to represent `polythetic' concepts in which the constituent members have a `family resemblance' to each other (Section \ref{polythetic_categories}). In effect, the system is geared to the creation of knowledge structures that express similarities amongst patterns.

\end{itemize}

\end{itemize}

Thus, from the perspective of the SP theory, it is a false antithesis to ask whether experience is encoded in terms of `rules' or `similarity'. In terms of the theory, the answer is ``both'' and ``neither''. Likewise, it does not make sense to ask whether knowledge may be encoded in terms of `rules' or `fragments'. These `oppositions' are a ``form of conceptual structure [that] leads ... to an ever increasing pile of issues, which we weary of or become diverted from, but never really settle.'' \citep[][p. 289]{newell_1973}.

We should not forget that the systems people use for encoding knowledge are certainly more sophisticated than any finite-state grammar. Although the training stimuli in an artificial grammar learning experiment may have been derived from a finite-state grammar, it is likely that participants would use any or all of their repertoire of encoding techniques and it unlikely that the resulting knowledge structures would fit neatly into any of the categories that have been considered in the context of artificial grammar learning experiments. 

\subsubsection{Analysis of Experimental Results}

Although there may be some doubt about assumptions that lie behind some of the artificial grammar learning experiments, we may still ask whether or how the SP theory might account for the results that have been obtained in these experiments.

The theory looks promising in this connection because the SP70 model has already demonstrated an ability to learn sets of patterns that are equivalent to finite-state grammars. However, the SP70 model has recognised shortcomings and we really need to consider how it might operate when it is more fully mature.

At this point, we make a plea for caution. The mature version of the SP70 model is, of course, not fully specified. Armchair speculations about how a proposed mechanism is likely to behave are often falsified when it is implemented and run as a computer program. Thus, at this stage, it is probably not sensible to try to make detailed predictions about what the mature version of the SP70 model might do with the materials used in artificial grammar learning experiments.

It is pertinent to point out that very much the same can be said about any of the other `models' that have been proposed to account for the results of artificial grammar learning experiments. In general, these are verbal descriptions of possible mechanisms. Thus, they are at least as under-specified as the anticipated mature version of the SP70 model. And there is corresponding uncertainty about how the models would process the artificial grammar learning training stimuli.

\subsubsection{Transfer Experiments}

With regard to the `transfer' artificial grammar learning experiments, Figure \ref{agl_mapping_figure} shows one of several possible ways in which a mapping between two stimuli may be expressed as an SP multiple alignment. In this case, `MSSSX' and `HJJJL' of our earlier example are connected, symbol by symbol, via the `mapping' patterns `: M H :', `: S J :' and `: X L :'. Another possibility is that people see both patterns as something like ``one letter, followed by three instances of a different letter, followed by one letter that is different from any previous letters''.

\begin{figure}[!hbt]
\centering
\begin{BVerbatim}
0   M     S     S     S     X     0
    |     |     |     |     |    
1 : M H : |     |     |     |     1
      | | |     |     |     |    
2     | : S J : |     |     |     2
      |     | | |     |     |    
3     |     | | |   : S J : |     3
      |     | | |   |   | | |    
4     |     | : S J :   | | |     4
      |     |     |     | | |    
5     |     |     |     | : X L : 5
      |     |     |     |     |  
6     H     J     J     J     L   6
\end{BVerbatim}
\caption{One possible way in which a mapping from `M S S S X' to `H J J J L' may be expressed as a multiple alignment. This is the best multiple alignment found by SP61 with `M S S S X' in New and, in Old, the pattern `H J J J L' together with the `mapping' patterns `: M H :', `: S J :' and `: X L :'.}
\label{agl_mapping_figure}
\end{figure}

SP70 can create multiple alignments like the one shown in Figure \ref{agl_mapping_figure} but only if the mapping patterns have been supplied to Old by the user. In its current form, SP70 cannot create mapping patterns for itself. However, the model could in principle be extended to incorporate that kind of learning. Given that the mapping patterns have a clear role to play in the economical encoding of New data, this extension of the model would be entirely in keeping with the overall philosophy of the SP theory.%
\index{learning!artificial grammar|)}

\section{Language Learning}\label{psychology_language_learning}

\index{learning!language|(}

Although there has been substantial progress in understanding how a child learns his or her first language, this area remains a significant challenge. There is still no single model that can achieve the kind of learning that children demonstrate in their first few years.

It is clear that children do not learn by `imitation', simply storing portions of language that they have heard like a tape recorder. A wealth of evidence shows that acquiring `competence' in any given language means the development of something equivalent to a grammar that can generate an infinite range of utterances from the language, well beyond the finite sample of utterances that any child has ever heard.

Significant questions in this area include:

\begin{itemize}

\item How is it possible for children to develop a knowledge of the segmental structure of language (words, phrases etc) when people often talk in `ribbons' of speech, without reliable physical clues to the boundaries between one segment and the next?

\item How can children learn disjunctive categories such as `nouns', `verbs' or `adjectives'?

\item How can these things be combined in grammar-equivalent forms of knowledge with the kind of hierarchical structure that appears to be necessary in any adequate description of the structure of a natural language?

\item How can children learn to distinguish `correct' generalisations\index{grammar!generalisation} beyond the finite sample that they have heard from `incorrect' generalisations, given that both kinds of generalisation, by definition, have zero frequency in the child's experience?

\item A related question is how a child can develop an adult's ability to distinguish sharply between utterances that they feel are `correct' samples of their native language and other utterances that they judge to be `wrong', given that the language that a child normally hears is by no means totally correct. This problem is sometimes called the `dirty data' problem.

\item How can a child learn meanings for words and other grammatical forms and how can this `semantic' knowledge be integrated with the child's developing knowledge of syntax?

\end{itemize}

It has of course been argued that natural languages are far too complex to be learned from scratch by any kind of `empiricist' mechanism  \citep{chomsky_1965}. According to this influential `nativist' view, children are born with a knowledge of the main structures of natural languages and that the language environment into which any given child is born serves to set the parameters of this pre-established knowledge so that the child develops an ability to understand and speak the language that they are hearing. Given that a new-born baby has the potential to learn any of the huge number of different languages in the world, the challenge for the nativist school of thinking is to define the `linguistic universals' that are shared by all these different languages and that constitute the innate knowledge that each child supposedly brings to the language learning task. Despite much research, no satisfactory definition of linguistic universals has yet been established.

Nativist thinking was motivated in part by the unsatisfactory nature of early attempts to explain language learning in empiricist terms \citep[see, for example,][]{chomsky_1959}. However, with the development of empiricist learning mechanisms that are very much more sophisticated than any of those early pioneering proposals, there has been a quiet revolution in our understanding \citep[see, for example,][]{harris_1955, kiss_1972, wolff_1988, wolff_1982, wolff_1980, wolff_1977, wolff_1975, redington_chater_1998}. There is now an increasing realisation that it may, after all, be possible to explain language learning without the need to postulate the kind of inborn knowledge envisaged in nativist proposals.

\index{grammar!generalisation|(}

Learning to distinguish correct generalisations from incorrect ones would not be a problem if adults and older children were systematically to correct the errors that young children make or if there was some other source of evidence about `wrong' kinds of utterances. In the same vein, \citet{gold_1967} proved that grammatical inference conceived as ``language identification in the limit'' is not possible without one or more of: error correction by a `teacher', or the provision of language samples that are marked as `wrong', or the presentation of language samples in ascending order of their complexity.

However, there is good evidence that, while children may take advantage of corrections if they are available, they can learn language without the need for them. The evidence comes chiefly from relatively rare cases of children who have apparently learned their receptive knowledge of language at essentially normal speed despite the fact that, because of cerebral palsy or other physical handicap, they have not been able to speak intelligibly and have not, therefore, been able to benefit from any kind of error correction by people around them \citep[see, for example,][]{brown_1989}. And it seems unlikely that, in these cases, there was any significant grading of language samples from simple to complex.

Although this evidence appears to conflict with Gold's conclusions, his proof is only valid for a conception of grammatical inference in which there is a non-stochastic `target' grammar that the system may learn ``in the limit''. As we shall see, the concept of a target grammar is not a necessary part of the language learning problem and, without it, Gold's proofs do not apply. In addition, Gold's proofs do not apply to the learning of stochastic grammars and there is plenty of evidence for the stochastic nature of human linguistic knowledge.%
\index{grammar!generalisation|)}

As we saw earlier (Section \ref{syntax_semantics_learning}), the way in which we learn to attach meanings to linguistic labels seems puzzling because every utterance could apply to a multitude of aspects of the situation that the utterance describes (``the scandal of induction''). To help explain how linguistic labels may be attached to meanings, a number of `biases' have been proposed. For example, \citet{markman_1991b} suggested that, via a `principle of mutual exclusivity', children avoid applying the same two labels to the same referent. She also suggested that there is a `whole object bias', meaning that labels are usually assumed to apply to whole objects rather than any of their parts. \citet{markman_hutchinson_1984} and \citet{waxman_1990} found that labelling a set of objects has an effect on the way in which children generalise about those objects that is different from the way in which they generalise when no labels are used. 

Doubt has been expressed about whether these kinds of biases are `universal' or whether they may be restricted to the structural features of particular languages \citep[see, for example,][]{gathercole_etal_1999}. There has also been debate about whether such biases may be innately specified or whether they might arise from the workings of some kind of mechanism for statistical induction. 

\subsection{The SP Theory and Language Learning}

As we saw in Chapter \ref{introduction_chapter}, the immediate inspiration for the SP theory was an earlier programme of research developing the MK10 and SNPR models of first language learning. In developing the SP theory, the aim has been to capture the insights gained from the earlier research and, at the same time, to extend the scope of the theory to other areas. In what follows, we shall first briefly review what was achieved with the MK10 and SNPR models and then, in the subsections that follow, we shall assess the SP theory as a model of language learning.

\subsubsection{The MK10 and SNPR Models}

The chief merit of the MK10 model is that it demonstrates how it is possible for the segmental structure of language to be learned by purely `statistical' means without supervision and without the need for physical clues to the boundaries between successive segments. This has been demonstrated both at the word level \citep{wolff_1977, wolff_1975} and also at the level of phrases \citep{wolff_1980}. It is interesting that hierarchical chunking mechanisms that are broadly similar to the MK10 program have been have been re-invented on repeated occasions for different purposes \citep[see, for example,][]{dawkins_1976, rapp_et_al_1994, nevill-manning_witten_1997}.

The SNPR model \citep{wolff_1988, wolff_1982} can learn plausible grammars for artificial languages by statistical means without supervision and without the need for physical markers of boundaries between words, sentences or other segmental structure. This model, which works largely in accordance with minimum length encoding principles, has a bearing on several of the questions and issues outlined above.

Like the MK10 model, it demonstrates how the segmental structure of language can be learned without the need for physical markers. The model also demonstrates how grammatical classes of words or other linguistic structure may be learned by purely distributional means. And the model can form hierarchically-structured grammars with conjunctive (segmental) structures and disjunctive structures at any level.

Using purely `positive' information and without any of the sources of evidence postulated by Gold, the SNPR model can form plausible generalisations of grammatical rules\index{grammar!generalisation} and eliminate all of the generalisations that appear, intuitively, to be `incorrect'. The key to this discrimination appears to be the application of minimum length encoding principles: a grammar containing only `correct' generalisations and no `incorrect' generalisations is more efficient in minimum length encoding terms than any of the alternatives (see Section \ref{grammatical_inference_and_generalisation}).

The SNPR model also provides an answer to the `dirty data' problem. It can learn `correct' forms of grammars even when it is supplied with language samples containing a reasonable number of errors. As with the correction of overgeneralisations, minimum length encoding principles are the key. Because any one kind of error is relatively rare compared with the corresponding correct form, errors of all kinds are simply filtered out by the system because they do not make a big enough contribution to the economical encoding of the original data.

The model also provides an account of empirical phenomena that were not anticipated in the design of the model. These include:

\begin{itemize}

\item The pattern of changes in the rate of acquisition of words as learning proceeds. The slowing of language development in later years.

\item The order of acquisition of words and morphemes.

\item Brown's \citeyearpar{brown_1973} ``Law of Cumulative Complexity''.

\item The S-P/episodic-semantic shift.

\item The observation, noted earlier, that children initially learn the correct forms of irregular verbs and nouns, then regularise them incorrectly, and eventually return to the correct forms again.

\item The word-frequency effect.

\end{itemize}

\noindent These things are discussed quite fully in \citet{wolff_1988}.

\subsubsection{SP70 and the SP Theory}

To a large extent, SP70 can achieve the same kinds of things as the earlier models. Like the MK10 and SNPR models, it expresses minimum length encoding principles but it uses multiple alignment as its core mechanism rather than hierarchical chunking. A major strength of SP70 compared with the earlier models is that it provides an account of a very much wider range of cognitive phenomena, well beyond the relatively narrow sphere of language learning.

As we saw in Chapter \ref{learning_chapter}, the SP70 model, like SNPR, can learn plausible grammars from appropriate input without the need for error correction by a `teacher' or `negative' samples or samples that are graded from simple to complex. As with the earlier model, it demonstrates how the segmental structure of language may be learned by statistical means without the need for physical markers of the boundaries between one segment and the next, and it demonstrates how disjunctive categories may be learned by distributional means.

As it stands now, SP70 can develop grammars with two main levels of abstraction. It is not good at building grammars with three or more levels but it is anticipated that, with some reorganisation of the model, this weakness can be overcome.

The model has already demonstrated an ability to make `correct' generalisations whilst avoiding `incorrect' generalisations---but more work is required in this area. No attempt has yet been made to test how well the model can learn with corrupted data but, since it is founded on minimum length encoding principles, it should perform at least as well as the earlier models.

It is too soon to say whether mature versions of the SP70 model will mimic observed features of first language learning in the way that SNPR does. However, since many of these features appear to be by-products of the application of minimum length encoding principles, it seems likely that successors to the current model will behave in a similar way.

\subsubsection{Do Children Learn a `Target' Grammar?}

One of the arguments in support of nativist proposals is that the grammars that we learn are rather uniform, despite the variability of our experiences. Thus \citet{chomsky_1965} wrote:

\begin{quotation}

``A consideration of the character of the grammar that is acquired, the degenerate quality and narrowly limited scope of the available data, {\em the striking uniformity of the resulting grammars}, and their independence of intelligence, motivation, and emotional state, over wide ranges of variation, leave little hope that much of the structure of the language can be learned by an organism initially uninformed as to its general character.'' (p. 58, emphasis added).

\end{quotation} 

If everyone in a given language community learns essentially the same grammar, it is natural to assume that language learning is a process of discovering what that `target' grammar is. And this accords with the notion of a target grammar employed in Gold's proofs, mentioned above.

However, if language learning is viewed from an minimum length encoding perspective, there is no target grammar, there is merely a search for one or more grammars that are as good as possible in terms of the minimum length encoding criteria. Because the abstract space of possible grammars is too large to be searched exhaustively, one can never be sure that the best possible grammar has been found.

How can this view be reconciled with the posited uniformity of the grammars we learn? A search for good grammars should lead to a degree of uniformity but we would expect some variations. It turns out that, if our tests of grammatical knowledge are sufficiently sensitive, people are more variable in their knowledge than was originally thought (See, for example, Carol Chomsky \citeyearpar{chomsky_c_1969}).

\subsubsection{Learning Meanings}

As we saw in Section \ref{syntax_semantics_learning}, there are reasons to think that statistical  principles of the kind that are embodied in the MK10, SNPR and SP70 models, coupled with the use of heuristic techniques such as hill climbing, should be able to sift through the many alternatives that present themselves in the learning of word-meaning associations. What is less clear at present is whether the SP system in its current form would be able to accommodate this problem or whether some adaptation would be required.%
\index{learning!language|)}

\section{Analogy and Structure Mapping}\label{analogy_structure_mapping}

\index{analogy|(}\index{structure mapping|(}

In the last two decades or so there has been a lively interest in the psychological phenomenon of analogy and related topics such as metaphor, similarity, inference, and conceptual blending \citep[see, for example,][]{gentner_etal_2001}. In this area of research, it is widely accepted that analogy involves a process of mapping one structure (the `source') to another structure (the `target'), a process known as `structure mapping' or `structural alignment'.

This process and the results of the process are clearly related to the concept of `multiple alignment' which is so prominent in the SP theory and, as we saw in Section \ref{geometric_analogy_problems}, the SP system has at least some capability for processing analogies. Although the relationship between these two areas of research deserves a relatively full discussion, I shall here merely attempt a summary. The main similarities and differences are discussed in the subsections that follow.

\subsection{Flat patterns and Hierarchical Structures}

In structure mapping, the things being mapped to each other can have an hierarchical structure (e.g., `greater(temperature(coffee), temperature(ice))') whereas the SP theory deals only in flat patterns.

Superficially, the fact that the SP system deals only with flat patterns seems to be a disadvantage compared with structure mappings in which hierarchical structures can be mapped. However, within the SP system, an hierarchical structure can be represented as a flat pattern and it can be assigned an hierarchical analysis in a multiple alignment parsing as shown in Figure \ref{alignment_figure_1} and in many other examples in this book. As we have seen in earlier chapters, flat patterns can be used within the SP system to represent a wide range of different kinds of structure, a combination of simplicity and versatility that has no counterpart in structure mapping theory.

Another potential advantage of adopting flat patterns as the basis for knowledge rather than something more complex is that this relates quite neatly to the way in which neural tissue appears to be organised into flat `sheets' of neurons in such structures as the retina, lateral geniculate body and the cortex (as discussed in Chapter \ref{neural_chapter}).

\subsection{Matched and Unmatched Symbols}

In a structure mapping, an element of one structure can be aligned with an element of another even if they are not a match for each other, whereas in multiple alignments, symbols are aligned only if they match each other. At first sight, this difference may seem somewhat trivial because, within a multiple alignment, there may be an implicit alignment of unmatched symbols as shown in Figure \ref{learning_alignment_1}. However, the question of whether symbols match or not is not highlighted in structure mapping theory whereas it has a two-fold significance in the SP theory:

\begin{itemize}

\item The matching of symbols is the key to the compression of information that is so central in the SP theory. Unmatched symbols within a multiple alignment do not contribute to information compression or to the compression score for the multiple alignment.

\item It is the contrast between matched symbols and unmatched symbols that drives the process of learning in the SP theory (Chapter \ref{learning_chapter}).

\end{itemize}

\subsection{Alignment of Three or More Patterns}

In structure mapping, there are only two things to be aligned: the source and the target, somewhat like the alignment of a New pattern with one Old pattern. There is nothing comparable with the alignment of one New pattern with two or more Old patterns as is commonplace in multiple alignments.

This is, perhaps, the most significant difference between structure mappings and multiple alignments. The possibility of creating multiple alignments containing three or more patterns opens up a range of possibilities for multiple alignments which appear to be closed for structure mappings:

\begin{itemize}

\item With multiple alignments, it is possible to model chains of reasoning or inference with any number of steps.

\item In a similar way, it is possible with multiple alignments to model inheritance of attributes through multiple levels of a class hierarchy.

\item With multiple alignments, it is possible to share patterns amongst multiple alignments in a way that cannot be done with structure mappings. It is true that, with structure mappings, a given structure may appear in several different alignments but this can be done at only one level or possibly two. With multiple alignments, a pattern at any level in a hierarchy of arbitrarily many levels may appear in several different multiple alignments. This ability to share patterns at many different levels is the key to compression and the economical representation of knowledge.

\item With multiple alignments it is possible to model the workings of a universal Turing machine (Chapter \ref{computing_chapter}), something that appears to be outside the scope of structure mappings. Bearing in mind that the universal Turing machine is widely recognised as a definition of `computing' and that the ability to `compute' is part of human cognition, multiple alignments appear to have an explanatory edge over structure mappings in this area of cognitive science.

\item The ability to form multiple alignments with three or more patterns is the key to modelling nonmonotonic reasoning (Section \ref{nonmonotonic_reasoning_section}).

\end{itemize}

\subsection{Two or More Appearances of a Pattern within a Multiple Alignment}

Within one multiple alignment, a given Old pattern can appear two or more times within the multiple alignment, not as a copy but as a specific instance, with consequent constraints on matching as described in Section \ref{multiple_appearances}. There is nothing comparable in structure mappings.

Perhaps the main significance of this feature of SP-style multiple alignments is that it allows the system to model recursive structures, something that seems to be outside the scope of structure mappings. Since recursion is a prominent feature of many kinds of knowledge, this is another area where multiple alignments have an advantage compared with structure mappings.

\subsection{Minimum Length Encoding Principles}

Compression and minimum length encoding principles have a key r{\^o}le in the evaluation of multiple alignments in the SP theory whereas other criteria---such as `systematicity' and `structural consistency'---are used for the evaluation of structure mappings. Possible advantages of the minimum length encoding orientation include:

\begin{itemize}

\item It provides a single unifying criterion for the evaluation of knowledge structures rather than the diversity of criteria used in research on structure mappings.

\item It provides a unifying connection between research on multiple alignments (and their applications) and the long tradition that recognises information compression as a significant theme in the workings of brains and nervous systems (Chapter \ref{foundations_chapter}).

\item Likewise, it makes a connection with research on inductive inference and its intimate connection with information compression (Section \ref{probabilities_ic_section}).

\end{itemize}

\subsection{Discussion}

The organisation of SP-style multiple alignments and the way they are created and used within the SP theory appears to give them some advantages compared with structure mappings as currently conceived. With multiple alignments it is possible to model a range of things that appear to be beyond the scope of structure mappings, such as natural language parsing, inference through multiple steps, inheritance of attributes through multiple levels, nonmonotonic reasoning, recursive structures, and the process of computing. There may also be advantages for multiple alignments at more abstract levels such as the simplicity of representing all knowledge with flat patterns, realisation of the theory in terms of neural mechanisms, and integration of the theory with other theory in cognitive science.

However, the foci of interest in structure mapping research have been somewhat different from the main concerns in the SP programme and structure mapping ideas have been applied in areas such as metaphor and conceptual blending which have not, so far, received attention in the SP research. Although the SP system has been successful in processing one type of analogy, no attempt has yet been made to examine how the system might be applied to those other aspects of analogy that have been considered in the structure mapping programme. In general, the relative merits of structure mappings and multiple alignments need more evaluation across a range of different areas of application.%
\index{analogy|)}\index{structure mapping|)}

\section{Conclusion}

Only a selection of topics has been considered in this chapter and, in each area, the discussion has been relatively brief. Nevertheless, it is apparent that the SP theory provides a useful and distinctive perspective on a number of issues in cognitive psychology.

%% file: future.tex
\chapter{Future Developments and Applications}\label{future_chapter}

\section{Introduction}

The preceding chapters have described the SP theory as it has been developed to date and some of its potential applications. In this chapter, I will describe what further developments of the theory seem to be needed, how the theory may be embodied in an `SP machine', and ways in which the theory and the machine may be applied in the future.

In looking into the future for these ideas, it is tempting to be drawn into science fiction and to imagine how an SP system might mend your socks, make you cups of tea and be your tennis partner as well. Although this might be entertaining, I shall stick to the more humdrum possibilities that seem to me to be reasonable and feasible.

\section{Development of the theory}\label{development_of_theory}

In the last few years, many of the issues that have arisen in the development of the SP theory have been resolved and the framework has become relatively stable and mature. However, there are certain areas where further development seems to be needed. Two of them have been described in previous chapters:

\begin{itemize}

\item In Section \ref{alternative_method_for_unordered_patterns}, it was suggested that there might be a case for generalising the SP concept of multiple alignment to include multiple alignments containing two or more patterns from New. This should be useful for applications like medical diagnosis where New information may need to be treated as an unordered set of patterns or as an unordered or partially-ordered set of symbols (see Section \ref{ordering_of_symptoms}).

\item A programme of work is needed to overcome weaknesses in the SP70 model (Section \ref{learning_discussion}). It is anticipated that most of the current deficiencies can be remedied by some reorganisation of the model within the overall framework that has been developed.

\end{itemize}

\noindent Other planned developments are discussed in the subsections that follow.

\subsection{Patterns in two dimensions}\label{2d_patterns_section}

\index{perception!vision|(}

The main focus so far in the development of the SP theory has been on one-dimensional patterns and on the modelling of those kinds of knowledge that map straightforwardly on to 1-D patterns---such as natural language and grammatical rules for describing the structure of natural language. Relatively little has been said about possible applications of the theory to the description and processing of forms of knowledge that have a two-dimensional structure such as maps, diagrams and pictures (which will be referred to as `images').

In conventional computing, there is of course a well-developed tradition of vector graphics and related techniques for describing 2-D images with 1-D mathematical notations. Since it is envisaged that mathematical structures and processes may be modelled in the SP system (Chapter \ref{maths_logic_chapter}), this is a possible way forward, within the SP system, for the representation and processing of images.

However, as previously noted (Section \ref{representation_of_knowledge}), it is envisaged that the concept of {\em pattern} in the SP theory may be generalised to include patterns in two dimensions (and possibly more)---and this may provide a more direct approach to the modelling of images, with potential advantages over one that mimics established techniques (see Section \ref{maps_diagrams_pictures_films}, below).

\subsubsection{Matching, multiple alignment, learning and display}\label{matching_ma_learning_display}

There is, of course, no intrinsic difficulty in representing 2-D patterns in a successor to SP61 or SP70. All that is required is arrays of symbols in two dimensions instead of one. The immediate difficulty lies in developing a process for matching 2-D patterns with the same kind of flexibility as the matching process for 1-D patterns described in Appendix \ref{matching_appendix}. This would, in effect, be a 2-D analogue of dynamic programming with the same advantages as the current process: that it can be scaled up without excessive demands on processing time or memory; the ability to deliver two or more alternative matches for a pair of patterns; and scope for varying the `depth' of searching according to need. Although the development of such a matching process presents a significant challenge, there seems no reason in principle why existing techniques should not be generalised in this way.

Assuming that a 2-D analogue of dynamic programming could be developed, this would need to be incorporated within a process for building SP-style multiple alignments with 2-D patterns. The learning processes in SP70, or an improved successor to that program, would need to be generalised for 2-D patterns.

Apart from these problems, there is a cosmetic-cum-practical problem of displaying alignments of two or more 2-D patterns effectively. Alignments of that kind may be viewed from different angles using `virtual reality' graphical software---but some other technique would be needed to show these kinds of multiple alignments clearly on the printed page.

\subsubsection{Representation of two-dimensional patterns without encoding}\label{2d_patterns_no_encoding}

As was described in Section \ref{raw_and_encoded_information}, analogue information may be digitised `directly', by representing the amplitude of the analogue signal by the density of `0's and `1's. Using this scheme, a 2-D image that we would recognise as a white triangle surrounded by black, may be represented digitally by a triangular area of `1's surrounded by an area of `0's.

A `raw' image like the one just described contains significant amounts of redundancy. This may be reduced by the matching and unification of patterns (or individual symbols) and by the use of identifying `codes' as described in Sections \ref{ic_repetition_of_patterns} and \ref{techniques_for_ic}. When learning processes like those described in Chapter \ref{learning_chapter} have been further developed and then generalised for patterns in two dimensions, the SP system should be able to compress any raw digitised image, as tentatively sketched in the subsections that follow.

\subsubsection{First stage of encoding: marking boundaries}\label{2d_patterns_boundaries}

In the original digitised image of a triangle, the most obvious form of redundancy is that represented by the repetition of symbols in each of the uniform areas, one inside the triangle and the other surrounding it. These areas may be compressed by the unification of repeating symbols and by the use of ID-symbols.

In one-dimensional patterns, ID-symbols have a dual r{\^o}le: marking the left and right boundaries of the pattern and identifying each pattern uniquely amongst the other patterns in Old. With 2-D patterns, it seems necessary to provide ID-symbols to fulfil these two r{\^o}les but, with respect to the marking of boundaries, the entire outline of the pattern needs to be marked, not simply the left and right ends of the pattern.

\begin{figure}[!hbt]
\centering
\includegraphics[width=0.9\textwidth]{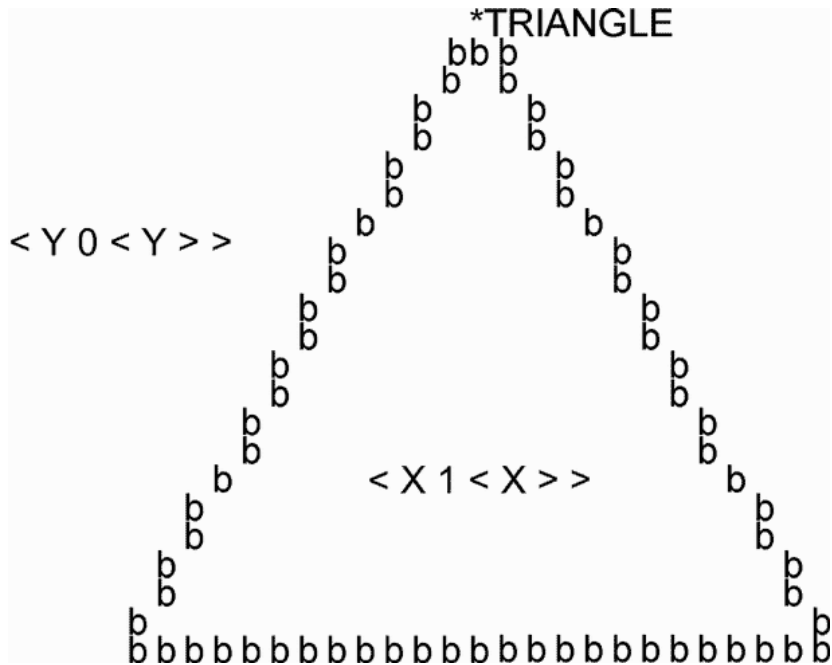}
\caption{A possible scheme for a first-stage encoding of a white triangle in black surroundings. The meanings of the symbols are explained in the text.}
\label{2d_image_1}
\end{figure}

Figure \ref{2d_image_1} shows a simple scheme for representing a white triangle. Here, `TRIANGLE' is an ID-symbol that may serve to distinguish the figure from others, `*' is an ID-symbol that serves as an anchor point for positioning the figure in relation to other figures and the `b' symbols are ID-symbols marking the boundary of the figure (also, if you wish, mnemonic for a `black' line around the figure).

Inside the triangle in Figure \ref{2d_image_1} is the pattern `$<$ X 1 $<$ X $>$ $>$' and outside the triangle is the pattern `$<$ Y 0 $<$ Y $>$ $>$'. The first of these patterns is intended as a schematic encoding of the repeating `1's inside the original `raw' triangle, and the second pattern is intended as a similar encoding of the repeating `0's that surround the triangle. Each pattern is a recursive structure modelled on the recursive pattern described in Section \ref{unary_numbers_and_sp}. However, neither pattern is entirely accurate in this application because it needs to be generalised for two dimensions. How that generalisation may be done is a matter for future investigation. 

\subsubsection{Further encoding}\label{encoding_2d_patterns_compressed}

The encoding shown in Figure \ref{2d_image_1} has removed much of the redundancy from the original `raw' data but some still remains. The straight lines marking the three sides of the triangle each contain redundancy in the form of repeating instances of `b' and the uniform trajectory from one `b' to the next. By contrast, the angles at each corner of the triangle, where a line in one orientation joins a line at another orientation, are places where there is non-redundant information that largely defines the nature of a triangle ({\em cf.} Attneave's \citeyearpar{attneave_1954} example showing how the form of a sleeping cat can be captured quite effectively with a few lines and angles).

\begin{figure}[!hbt]
\centering
\includegraphics[width=0.9\textwidth]{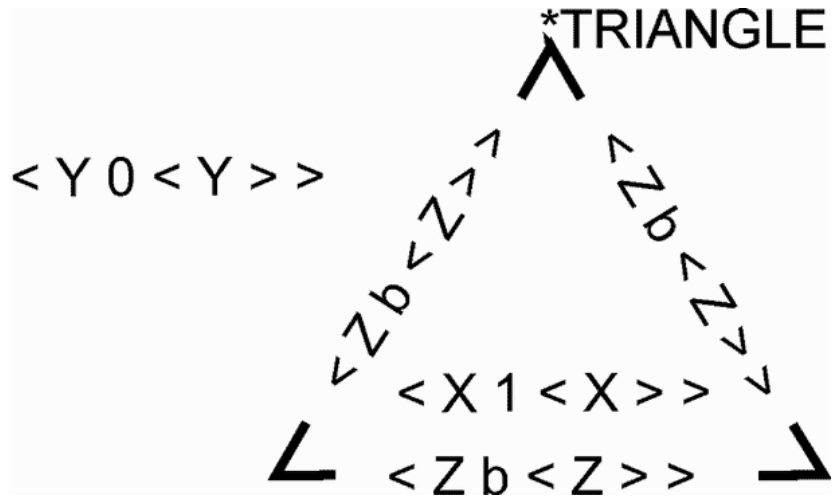}
\caption{A possible further encoding of a white triangle in a black surround. The interpretation of the figure is discussed in the text.}
\label{2d_image_2}
\end{figure}

Figure \ref{2d_image_2} shows how the residual redundancy in Figure \ref{2d_image_1} may be reduced. Instead of marking each side by a succession of `b' symbols, we may suppose that the system has recognised the recursive nature of each line and has created an encoding that is something like `$<$ Z b $<$ Z $>$ $>$', comparable with the recursive patterns in the body of the triangle and in the surrounding area. For each of the three corners of the triangle, we may suppose that the system has created three new ID-symbols, one for the top corner and one for each of the left and right lower corners. In the figure, the symbol `TRIANGLE' serves to distinguish the pattern from other patterns in the system and `*' serves as an anchor point, as before. 

An encoded representation like the one in Figure \ref{2d_image_2} has three interesting features:

\begin{itemize}

\item It goes some way to capturing John Locke's idea of an abstract, generalised concept of a triangle.\footnote{``For abstract ideas are not so obvious or easy to children... For, when we nicely reflect upon them, we shall find that general ideas are fictions and contrivances of the mind, that carry difficulty with them... For example, does it not require some pains and skill to form the general idea of a triangle... for it must be neither oblique nor rectangle, neither equilateral, equicrural [i.e., isosceles], nor scalenon; but all and none of these at once.'' John Locke, {\em Essay Concerning Human Understanding}, 1690, Book IV, Chapter 7, Section 9.}. Given an appropriate process for building 2-D multiple alignments, the representation in Figure \ref{2d_image_2} should serve for the recognition or construction of equilateral triangles of any size, not just the specific triangle shown in Figure \ref{2d_image_1}. With some further generalisation of the concept of `angle', a representation like this should serve for the recognition or construction of a range of triangles with a variety of angles at the corners.

\item Because the representation is compatible with triangles of many different sizes, it provides a possible solution to the problem of size constancy in perception, mentioned in Sections \ref{perceptual_constancies} and \ref{constancies_neural}.

\item The representation in Figure \ref{2d_image_2} does not encode the {\em absolute} sizes of the sides of the triangle but it does, in effect, encode the {\em relative} sizes of the sides of the triangle---via the positions of the symbols marking each angle and, perhaps, via the sizes of the angles encoded within those symbols. Encoding of relative sizes is sufficient for many purposes and it accords with a range of evidence showing that human perception is geared to relative judgements rather than absolute ones. This kind of system can make absolute judgments if a suitable reference is provided. 

\end{itemize}

\subsubsection{Hierarchical encoding of two dimensional patterns}\label{encoding_2d_patterns_hierarchical}

While it is true that we can recognise a triangle as a triangle regardless of its size or shape, it is also true that we can distinguish one triangle from another by size, shape, colour or other attributes. There seems no reason in principle why the recognition of images at two or more levels of abstraction should not be achieved in the SP system in much the same way as for 1-D patterns, as illustrated in Section \ref{class_part_inheritance}.

\begin{figure}[!hbt]
\centering
\includegraphics[width=0.9\textwidth]{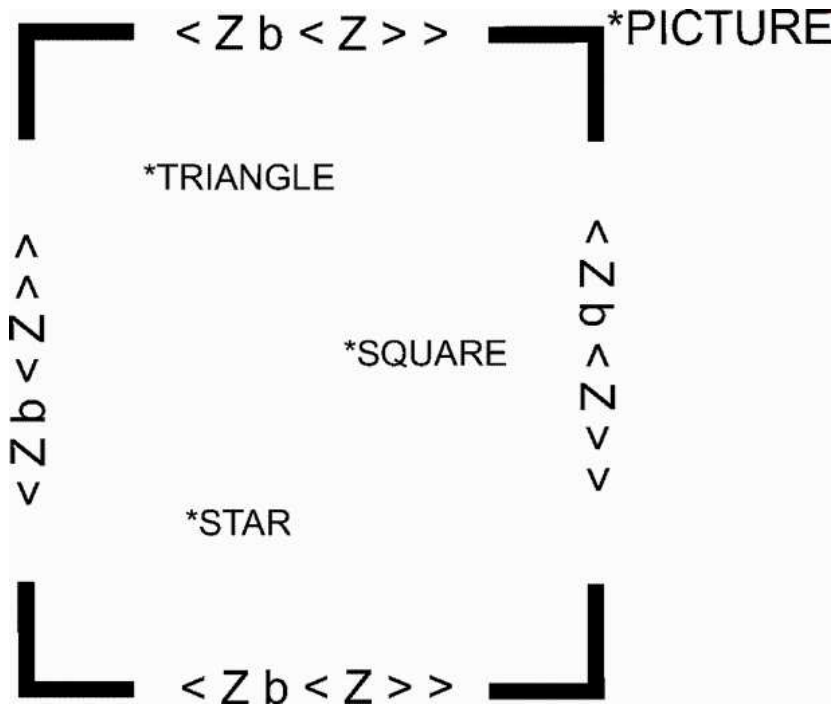}
\caption{A 2-D picture constructed as an assemblage of references to `lower' level patterns.}
\label{2d_image_3}
\end{figure}

The modelling of class hierarchies and part-whole hierarchies within the SP system is achieved quite simply by the provision of C-symbols in some patterns that serve as references to corresponding ID-symbols in other patterns. Figure \ref{2d_image_3} shows a schematic example of a pattern constructed as an assemblage of references to `lower' level patterns---a relationship that may be repeated recursively through any number of levels. When these ideas are more fully worked out, there will, no doubt, be a need for additional information to specify sizes, colours and textures of component patterns and, perhaps, relationships such as overlap between patterns.

\index{perception!vision|)}

\subsection{Developing the theory with patterns with three or more dimensions?}

Since we regularly deal with objects and spaces in three dimensions, it might seem natural to model these things in the SP system with patterns in three dimensions---and likewise for concepts with higher dimensions. Although this may seem the logical way to go, my current thinking is that this would probably be a mistake. Apart from the difficulty of imagining or representing alignments amongst patterns of that kind, there are two main reasons for caution in this area:

\begin{itemize}

\item There may be no particular need to develop the theory in this way. Patterns in one or two dimensions may serve well enough to represent concepts with three or more dimensions. There is a long tradition of representing buildings and other engineering artefacts with two-dimensional plans and elevations. To model concepts in four or more dimensions, there seems little prospect of improving on established mathematical techniques.

\item As we saw in Chapter \ref{neural_chapter}, there are reasons to think that arrays of neurons in one or two dimensions may provide the basis for our many kinds of knowledge:

\begin{itemize}

\item Although the human brain is clearly a three-dimensional structure, there is a distinct tendency for it to be organised into layers of cells such as those found in the cortex and in the lateral geniculate body. Although these are often folded into three dimensions, they are topologically equivalent to two-dimensional sheets. As was suggested in Chapter \ref{neural_chapter}, these sheets of neurons would provide a natural home for SP patterns.

\item In a similar way, there is a clear tendency for sensory neurons to be organised as arrays with two dimensions in the retina and in the skin or with a predominant single dimension in the case of cochlea. It seems possible that structures for the storage and retrieval of knowledge evolved from sensory structures and this might explain the similarities in their organisation. It is true that there is a temporal dimension to vision as well as the two dimensions of each momentary image but there seems no reason why these three dimensions should not be represented in two dimensions like a succession of images in a cinema film or video tape---but with high levels of encoding and compression. Similar principles may apply to hearing.

\end{itemize}

\end{itemize}

\subsection{Theoretical aspects of searching and constraints}\label{searching_and_constraints_theory}

\index{constraint|(}

Other aspects of the SP theory that need further exploration and development are methods for searching the space of alternative multiple alignments, the kinds of constraint that may be applied to the search, and inter-dependencies between the two. It is not entirely clear at present whether these issues are purely theoretical or whether they are at least partly concerned with the ways in which the SP theory may be implemented in a working  system (next section). Insofar as the distinction is sound, it seems likely that the issues straddle both areas.

In the context of the SP system, any generalised theory of searching would need to say something useful about building multiple alignments with 2-D patterns (Section \ref{2d_patterns_section}, above) and it should also throw light on ways in which parallel processing may be applied (Section \ref{searching_and_constraints_machine}, below).

It seems likely that, at some stage, the searching and matching processes will need to be generalised so that 1-D pattern can be matched in both possible directions and 2-D patterns can be matched in several possible orientations (see Section \ref{syntax_semantics_learning}).

With regard to constraints, an area that needs particular attention is the kinds of constraints that are embodied in classical systems for logic and mathematics. This is needed to improve our understanding of the ways in which those classical systems may (or may not) be understood within the SP theory. And it will also be needed in the design of a working system, as discussed in the following section.%
\index{constraint|)}

\subsection{Recency in learning}\label{recency_section}

When we learn from experience, it seems that recency of events is a factor in our thinking as well as frequency. If we are asked to guess who will score the most goals in a football match then, other things being equal, we are likely to choose the player with the best scoring record. But if that player had, in {\em recent} matches, failed to live up to his usual form, then we might choose a different player.

As the SP system is developed more fully for learning, there will be a need to take account of the effect of recency in learning. Of course, frequency and recency are linked because any measure of frequency must be relative to a particular `window' or time frame in which frequencies are measured, and those windows can vary in size and also in recency. These ideas will need to be more fully worked out in successors to the SP70 model.

\section{Development of an `SP' machine\index{SP machine}}

The SP theory provides an interpretation for computation in existing systems (Chapter \ref{computing_chapter}) but it also suggests new designs for computers, with potential advantages over the von Neumann model that has dominated the digital scene for so long. This section sketches some ideas for the development of an `SP' computer (or `SP' computer) that should, if required, be able to function like a conventional computer but which should in addition have strengths in artificial intelligence that are missing from the current generation of digital computers.

The `high level' design of the SP machine would be a realisation of the SP theory as it has been developed and expressed in software models. Although further developments in the theory are still needed (as described in Section \ref{development_of_theory}), the concepts embodied in the SP61 model have a range of potential applications and could provide a basis for a preliminary version of the SP machine.

A key feature of the proposed new machine is that it would incorporate high levels of parallel processing. This is because artificial intelligence applications are intrinsically `hungry' for computer power and because the SP theory offers opportunities to by-pass the `busy bee' style of conventional computing to achieve good results in a reasonable time by the application of parallel processing. Notice that the computational demands of artificial intelligence applications are intrinsic to those applications and not a consequence of their interpretation in terms of the SP theory.

\subsection{Implementation aspects of searching and constraints}\label{searching_and_constraints_machine}

The central process in the SP61 and SP70 models is the building of multiple alignments---and the core operation here is the process of searching for full matches and good partial matches between patterns, as described in Appendix \ref{matching_appendix}. 

Finding good matches between patterns presents opportunities to apply parallel processing. This is because the process of `broadcasting' symbols from one pattern to make yes/no matches with symbols in other patterns (Section \ref{matching_one_pattern_with_another}) is an intrinsically parallel operation that could be done as a single step in an appropriate parallel environment (in contrast to the sequence of operations required on an ordinary digital computer). There is also scope for matching sequences of two or more symbols against each other in parallel and this may have a place in the proposed new machine.

\subsubsection{Heuristic constraints}
\index{constraint|(}

Notwithstanding these opportunities for doing several things at the same time, it seems necessary to do some things in sequence. As we saw in Section \ref{matching_searching_and_constraints}, the constraints that are applied in heuristic search methods such as hill climbing or genetic algorithms are applied in a succession of stages, with a progressive narrowing of the search space at each stage. Although searching within each stage may be done in parallel, there seems no escape from the need for the stages of metrics-guided search to succeed each other in sequence.

In general, there is a need to explore a variety of methods for matching patterns in parallel, including the kinds of heuristic constraints that are needed to ensure that the process is tractable.

\subsubsection{Absolute constraints}

Apart from heuristic constraints, the SP machine should also support the application of absolute constraints. It should, for example, be possible to specify the maximum size of any sequence of unmatched symbols in a multiple alignment---including zero if `exact' matching of patterns is required. As in SP61 and SP70, it should be possible to enforce the rule that left and right brackets should be matched in such a way that they are always `balanced'.

As with heuristic constraints, there is a need to explore how parallelism can be applied to best effect in the matching of patterns and to combine this efficiently with the application of constraints. As a general rule, constraints of any kind should speed up processing by reducing the size of the search space. But care is needed to avoid inefficiencies in the way constraints are combined with searching, as seems to be the case with the `brackets' constraint in SP61 and SP70 (Section \ref{angle_brackets_constraint}).%
\index{constraint|)}

\subsection{Technological options}

The SP machine does not necessarily have to be developed in silicon. One futuristic possibility is to exploit the potential of organic molecules such as DNA or proteins---in solution---to achieve the effect of high-parallel pattern matching. This kind of `molecular computation'\index{computing!molecular} is already the subject of much research \citep[see, for example,][]{adleman_1994, adleman_1998} and techniques of that kind could, conceivably, form the basis of a high-parallel SP machine.

Another possibility is to use light for the kind of high-speed, high-parallel pattern matching that is needed at the heart of the SP machine.\index{computing!optical} Apart from its speed, an attraction of light in this connection is that light rays can cross each other without interfering with each other, eliminating the need to segregate one stream of signals from another \citep[see, for example,][]{cho_colomb_1998, louri_hatch_1994}.

In the field of optical and opto-electronic computing as it has developed to date, there is an apparent tendency for researchers to assume that the design of optical computers should imitate the design of the current generation of electronic computers and to concentrate on building optical versions of logic gates \citep[see, for example,][]{lee_dickson_2003, singh_roy_2003}. The SP theory provides an alternative approach to the development of optical computers, an approach which is founded on pattern matching and may by-pass many of the intricacies of computer design, as currently conceived.

On relatively short time scales, a silicon version of the SP machine would probably be the easiest option and it may be developed in one or more of three main ways:

\begin{itemize}

\item It should be feasible to design new hardware for the kind of high-parallel pattern matching that is needed.

\item Given that high-parallel machines are already in existence (in both `single instruction, multiple data' (SIMD) and `multiple instructions, multiple data' (MIMD) designs), an alternative strategy is to create the SP machine as a software virtual machine running on one of these platforms.

\item Systems like Google already provide high-parallel pattern matching of a kind that may be adapted for use within a software implementation of the SP machine. 

\end{itemize}

To some extent, the second and third options defeat the object of the exercise because the von Neumann architecture and much of its associated complexity would remain embedded in the system. However, these methods of developing the SP machine would probably be relatively cheap, they would probably yield useful insights into ways in which parallelism may be applied, and they should yield a version of the machine that could be used in practical applications.

\subsection{Other developments}\label{sp_machine_other_developments}

Other aspects of the SP machine that would need to be developed include:

\begin{itemize}

\item {\em Styles of Computation}. It should be possible to modify the operations of the machine to suite the needs of particular applications. For example, it should be possible to control the `depth' of searching as described in Section \ref{searching_and_constraints_in_sp61}. There may also be switches and parameters to control which constraints are in force and how they are applied (Section \ref{searching_and_constraints_machine}).

\item {\em Mathematics}. Although the system can, in principle, do arithmetic and other kinds of mathematics (Chapters \ref{computing_chapter} and \ref{maths_logic_chapter}), its current strengths lie in `symbolic' and non-mathematical kinds of computation. To do arithmetic, the system would need to be equipped with an appropriate set of patterns and the kinds of constraint mentioned in Section \ref{searching_and_constraints_theory} would also need to be developed. Pending these developments, there may be a case for exploiting existing technologies within the SP machine, in much the same way that `maths co-processors' are provided in conventional computers.

\item {\em User Interface}. A graphical user interface is needed to display data and results and to facilitate browsing and searching within that information. Since multiple alignments can grow very large, there is a need to be able to zoom in or out and to scroll through a multiple alignment (horizontally and vertically). It should be possible to see the unified version of any multiple alignment when required and, when viewing a unified version of a multiple alignment, it would probably be useful if ID-symbols could be concealed from view or revealed, according to need. Some kind of graphical control panel or dialogue box is needed so that switches and other parameters can easily be set.

\end{itemize}

\section{Applications of the SP theory and machine}\label{sp_future_applications}

Given the broad scope of the SP theory, its potential applications are legion. This section describes some of the possibilities, ranging from those that may be realised on relatively short time scales using existing technologies to those that depend on longer-term developments of the SP machine. The main emphasis here is on engineering applications but it should not be forgotten that the theory may also have heuristic uses in research, perhaps leading to new insights or suggesting new avenues of investigation in areas such as neuroscience, mathematics or logic.

\subsection{Information compression}

\index{information!compression|(}

Given the central importance of information compression in the SP theory, an obvious application for the SP machine would be in the compression of information. Possible arguments against that kind of development include:

\begin{itemize}

\item The capacities of computer memories and storage devices are increasing all the time, with no sign of a ceiling, and this will make compression irrelevant.

\item Likewise, it may be argued that communication bandwidths are constantly increasing and this will progressively reduce the need for information compression.

\item Information compression is already well-enough served by ZIP programs and the like and there is no need for anything different.

\item It may also be argued that too much compression is bad---that redundancy in information is needed to guard against errors or loss of information and to speed up processing.

\end{itemize}

\subsubsection{Compression will not be needed}

Considering the first and second of these arguments, similar things were said in the 1970s\footnote{These were the reasons given by the former National Research and Development Corporation (a United Kingdom government quango) for rejecting a proposal that I made at that time to develop the MK10 model for compression applications.} and there is still a need for information compression, despite vast increases in storage capacities and bandwidths. Streaming video over the internet is still a significant challenge as is the repeated archiving, at regular intervals, of the entire World Wide Web (http://www.archive.org/). Our demands for the storage and transmission of information seem always to outpace the available technologies.

\subsubsection{Current technologies are good enough}

Compression algorithms such as the Lempel-Ziv algorithms used in ZIP programs are designed for speed of compression and decompression on traditional digital computers. This means that the search for repeating patterns that is embodied in these algorithms is relatively simple and not designed to find as much redundancy as possible. The patterns that are found are always coherent substrings within the data and discontinuous subsequences are excluded. Patterns can contain references to other patterns but there is nothing equivalent to the multiple alignment concept and the scope that it offers for the encoding of information at several different levels of abstraction.

By contrast, it is envisaged that the SP machine will be designed so that it can, when required, do relatively thorough searches for redundancy in information, including redundancy in patterns that are both discontinuous and coherent, and redundancies at multiple levels of abstraction. High levels of parallelism should allow this relatively thorough search for redundancy---and correspondingly high levels of compression---to be achieved with reasonable speed. Although compression in itself might not justify the use of an SP machine, its capabilities in this area should be a welcome bonus on top of its other uses.

\subsubsection{Too much compression is bad}

The third argument should not detain us because, as we saw in Section \ref{uses_of_redundancy}, there is no real conflict between information compression in information processing and the ways in which redundancy can be exploited. Compression and redundancy have different r{\^o}les to play and they may co-exist in a single body of information.%
\index{information!compression|)}

\subsection{Representation and processing of graphical information}\label{maps_diagrams_pictures_films}

\index{graphics|(}

In the representation and processing of graphical information---maps, diagrams and pictures (both still and moving)---there is a well-developed tradition of object-oriented design and object-oriented programming. In the digital processing of a map, for example, it is natural to use a software `object' to represent each entity such as a church or a historic monument and to exploit class hierarchies with inheritance of attributes to achieve a clean and economical design (Sections \ref{classes_subclasses_inheritance} and \ref{oop_section}). In a similar way, each character in a computer game would naturally be programmed as a software object, inheriting methods and data structures from appropriate classes.

Although the main elements of object-oriented design flow naturally from the SP system, this in itself does not provide anything that is not already available in object-oriented languages such as Simula, Smalltalk or C++. The potential of the SP system in this area lies in the way it may combine object-oriented capabilities with others:

\begin{itemize}

\item {\em Recognition and Scene Analysis}. If the system can be developed as described in Section \ref{2d_patterns_section}, it should serve for `fuzzy' recognition of 2-D images at multiple levels of abstraction, with the flexibility to cope when parts of an image are obscured or when there are variations in sizes. It may also serve for the analysis of a scene into its component parts in a manner that is comparable with the parsing of natural language (Section \ref{framework_examples_section} and Chapter \ref{language_chapter}) but with provision for overlap amongst the components of a scene (Section \ref{scene_analysis_recognition_retrieval}). Although it is not yet clear how the system might best be developed for the representation and processing of 3-D objects (or 2-D images of 3-D objects), there seems to be potential here too.

\item {\em Unsupervised Learning of Objects and Classes}. At some point in the future it may be possible to use the system for the automatic or semi-automatic learning of `natural' objects and classes. It might, for example, be possible to scan an existing map and derive a collection of patterns corresponding to the kinds of entities and classes of entities that we would naturally recognise there.

\end{itemize}

\index{graphics|)}

\subsection{The Semantic Web}\label{semantic_web_section}

\index{Semantic Web|(}

The `Semantic Web' is envisaged as a successor to the current World Wide Web in which web pages contain structured descriptions of their contents that can be processed by computers to provide more `intelligence' in the operation of the web \citep{berners-lee_etal_2001}. One of the most popular approaches to the representation and processing of this `ontological' knowledge is a family of languages based on the Resource Description Framework (RDF), such as RDF Schema (RDFS), the Ontology Inference Layer (OIL) and DAML+OIL (see, for example, \citep{broekstra_etal_2002,mcguinness_2002,horrocks_2002,fensel_etal_2001}). RDF provides a syntactic framework, RDFS adds modelling primitives such as `instance-of' and `subclass-of', and DAML+OIL provides a formal semantics and reasoning capabilities based on `description logics', themselves based on predicate logic. For the sake of brevity, RDF and its associated languages will be referred to as `RDF+'.

Although RDF+ is popular, it has some shortcomings. Here, I will suggest that the SP system is a possible alternative with advantages compared with RDF+. In the spirit of `evolvability' in the web, the two approaches may be developed in parallel until their relative strengths and weaknesses are more fully understood.

Figure \ref{rdf_plus_example} shows an example of how the class `herbivore' may be described with one version of RDF+ \citep{broekstra_etal_2002}. In summary, this description says that a herbivore is a subclass of the class `animal' and that herbivores eat plants or parts thereof. This sample of RDF+ may be compared with the examples in Sections \ref{scene_analysis_recognition_retrieval} and \ref{cross_classification_multiple_inheritance} showing how class hierarchies and part-whole hierarchies may be represented in the SP scheme.

\begin{figure}[!hbt]
\fontsize{10.00pt}{12.00pt}
\centering
\begin{BVerbatim}
<rdfs:Class rdf:ID-"herbivore">
     <rdfs:subClassOf rdf:resource-"#animal"/>
     <oil:hasPropertyRestriction>
          <oil:ValueType>
               <oil:onProperty rdf:resource-"#eats"/>
               <oil:toClass>
          <oil:Or>
               <oil:hasOperand rdf:resource-"#plant"/>
               <oil:hasOperand>
                    <oil:HasValue>
                         <oil:onProperty
                              rdf:resource-"#is-part-of"/>
                         <oil:toClass rdf:resource-"#plant"/>
                    </oil:HasValue>
               </oil:hasOperand>
          </oil:Or>
               </oil:toClass>
          </oil:ValueType>
     </oil:hasPropertyRestriction>
</rdfs:Class>
\end{BVerbatim}
\caption{How the class `herbivore' may be described in one version of RDF+ \citep{broekstra_etal_2002}.}
\label{rdf_plus_example}
\end{figure}

A point of similarity between RDF+ and SP is the use of start and end markers such as `$<$oil:hasOperand$>$ ... $<$/oil:hasOperand$>$' in RDF+ and `NP ... \#NP' or `$<$ ... $>$' in SP. In RDF+, these markers---which are part of the XML foundations of RDF+---are universal, whereas in SP they are not always required.

Perhaps the main point of difference between RDF+ and SP is that the SP theory incorporates a theory of knowledge in which syntax and semantics are integrated and many of the concepts associated with ontologies are implicit. There is no need for the explicit provision of constructs such as `Class', `subClassOf', `hasPropertyRestriction', and similar constructs shown in Figure \ref{rdf_plus_example}, or others such as `Type', `hasPart', `Property', `subPropertyOf', or `hasValue' which are also used in RDF+. Classes of entity and specific instances, their component parts and properties, and values for properties, can be represented and integrated in a very direct, transparent and intuitive way, as described in Chapter \ref{rr_chapter}. In short, the SP approach allows ontologies to be represented in a manner that appears to be simpler and more comprehensible than RDF+, both for people writing ontologies and for people reading them.

Reasoning in RDF+ is based on predicate logic and belongs in the classical tradition of monotonic deductive reasoning in which propositions are either {\em true} or {\em false}. By contrast, the main strengths of the SP system are in probabilistic `deduction', abduction and nonmonotonic reasoning (Chapter \ref{rr_chapter}). There is certainly a place for classical styles of reasoning in the Semantic Web but probabilistic kinds of reasoning are needed too.

As we saw in Chapter \ref{rr_chapter}, the SP system copes easily with errors of omission, commission and substitution in information. This kind of capability is outside the scope of `standard' reasoning systems for RDF+ (see, for example, \citet{horrocks_2002}). It has more affinity with `non-standard inferences' for description logics (see, for example, \citet{baarder_etal_1999,brandt_turhan_2001}). Given the anarchical nature of the Web, there is a clear need for the kind of flexibility provided in the SP system. It may be possible to construct ontologies in a relatively disciplined way but ordinary users of the web are much less controllable. The system needs to be able to respond to queries in the same flexible and `forgiving' manner as current search engines. Although current search engines are flexible, their responses are not very `intelligent'. The SP system may help to plug this gap.

Apart from these possible advantages of the SP system compared with RDF+, the framework has potential in other areas---such as natural language processing, planning and problem solving, and unsupervised learning---that may prove useful in the future development of the Semantic Web.%
\index{Semantic Web|)}

\subsection[Software Engineering]{Software Engineering%
\footnote{Based on part of Chapter 6 in \citet{wolff_1991}.}}

\index{software engineering|(}

In spite of `fourth generation' languages and the slowly maturing CASE
tools and integrated project support environments, the process of developing new software and maintaining or
upgrading existing software still has problems:

\begin{itemize}

\item Software development is often {\em slow}. Delays can mean costs and inconvenience.

\item Software development is {\em labour-intensive} and correspondingly expensive.

\item {\em Incompatibilities} between machines, operating system and 
languages means difficulties in re-using software in new contexts 
and difficulties in integrating software systems. The resulting need 
to translate software or adapt it, or develop interfaces to 
translate information from one system to another, means complexity, 
delays, costs and new sources of bugs.

\item Software developers need extensive (and expensive) {\em training}.

\item {\em Verification}. Notwithstanding the use of formal methods in 
software development, it is impossible to be sure that any significant 
piece of software is correct in the sense that it does 
what its designers intended in all circumstances. Any 
significant body of software is likely to contain bugs.

\item {\em Validation}. System designers can, and often do, 
fail to understand fully what the users want and, consequently, 
fail to capture the users' requirements in the system.

\end{itemize}

In the following sub-sections, I outline some of the causes of these 
problems and how they may be solved or at least alleviated in the SP system.

\subsubsection{Procedural and declarative programming}

\index{programming!procedural|(}\index{programming!declarative|(}

In software design, there is a long-established distinction between `what' a software system is about and `how' the system will do what is required. The former is knowledge about the world that is encoded in the system (e.g., relevant laws and regulations for accountancy applications or knowledge about fonts, headings, paragraphs and the like in a program for desk-top publishing), while the latter is `lower level' knowledge about how to store information or find it using the underlying hardware.

With `procedural'\index{programming!procedural} programming languages like Fortran, Cobol and C, programmers need to consider both the `what' and the `how' of software design. It has been recognised for many years that the process of designing software would be simplified if programmers could concentrate on what the system should be able to do without  having to worry about how the machine would do it. Although this has been achieved to some extent with `declarative'\index{programming!declarative} languages like Prolog, it is still necessary to design software in a way that takes account of the step-by-step procedures that the machine will follow.

According to the SP theory, the `how' parts of computing are the processes of searching for matching patterns, building multiple alignments and encoding information economically that constitute the SP model. In conventional computers, it appears that SP processes are provided partly in the `core' of the computer and partly in the software (see Section \ref{simplification_of_computing_systems}, especially Figure \ref{conventional_and_sp_computers}). By increasing the generality and power of the SP processes that are provided in the core---as is envisaged in the proposed SP machine---it should be possible to eliminate the need for any `how' programming in the software.

To the extent that this vision can be realised, SP patterns that are processed by an SP computer would constitute a truly declarative language that may be used to specify what is to be done without concerns about how the machine will do it. This would mean a corresponding streamlining in the process of developing new software.%
\index{programming!procedural|)}\index{programming!declarative|)}

\subsubsection{One simple language and one model of computing for all purposes}

There are several different computer languages, specification formalisms
and design notations in common use today---including diagrammatic notations
and formalisms---and there are many other systems that are less commonly
used.

Associated with these various languages and notations are several different
perspectives or `philosophies' about how software and related knowledge
should be organised: `entity-relationship models', `data flow models',
`entity life histories', `object-oriented design', `logic programming',
`functional programming' and others.

Training software developers in these different languages, notations and
perspectives takes time and money. The lack of clear guidelines about which
formalism or perspective is appropriate in which contexts, and how they
should be related, if at all, causes confusion and wasted effort.

Given the `universal' nature of SP patterns (Section \ref{representation_of_knowledge}), they may serve as a `wide spectrum' language in software engineering with a wide range of 
potential applications.

Using one simple language for all types of software would mean less training
and less confusion about how software development should be done. Carried
to its logical conclusion it may eliminate all problems of compatibility
between systems.

\subsubsection{No refinement, verification or compiling}

In a typical development project, two, three or more different formalisms,
languages or notations will be used in different stages of the
development. There may be two or three translations between
formalisms---from analysts' models to formal specifications and from formal
specifications to program code. There is always some kind of compiling of
the code into a form which the machine can use. If proper standards of
quality are to be maintained, it will be necessary to check rigorously or
`prove' that each of these translations have been done correctly. 
The correctness of compilers must also be verified.

The whole process of translating between formalisms and proving the
correctness of translations is labour intensive and expensive. The
complexity of the process means there is plenty of scope for making
mistakes and there are corresponding problems in maintaining the quality of
the final product.

The SP machine has the potential to simplify the process of developing
software dramatically. SP patterns may be used in all stages
and phases of development. An SP data model prepared during systems
analysis may be incorporated directly in a software system without
modification. A formal specification of a software system expressed as SP patterns
may be run immediately without translation into a programming language. All
translations and all compiling may be eliminated. Since SP patterns are
intended to be executed `directly' without translation or compiling, there 
should be no need for `verification' or proof that an implementation conforms 
to its specification.

Eliminating all translation between formalisms and all verification that
translations have been done correctly can bring four major benefits:

\begin{itemize}

\item It can save large amounts of labour-intensive work and corresponding expense.

\item It can shorten the time needed to produce a working system. This is
useful in itself and it facilitates the use of iterative prototyping techniques, giving
prospective users good opportunities to validate the developing system and
ensure that it meets their needs. 

\item It should eliminate all errors arising from translation between formalisms during refinement and verification. This is a major source of errors in software and its total elimination would mean a dramatic improvement in the quality of software products.

\end{itemize}

\subsubsection{Software re-use}

The variety of languages and systems that are currently in use and the fact that, often, they do not
integrate properly, one with another, means significant difficulties in
re-using software in more than one context. Significant resources go into
translating between languages, porting software between machines and
operating systems or simply re-writing functions from scratch. It should
never again be necessary to re-write the standard arithmetic functions but
this does still happen.

The SP system may facilitate software re-use in four main ways:

\begin{itemize}

\item By supporting object-oriented class hierarchies with inheritance (Section \ref{class_part_inheritance}): any new class may re-use an existing class by making
the old class into a super-class of the new class. In this way, all the information in the old
class is inherited by the new class.

\item As in any `functional' programming language, an SP `function' can be re-used by `calling' it from two or more different contexts.

\item A significant problem in re-using software is finding what is needed
within a library of existing modules, functions or classes. The use of
meaningful names for functions can help as can the indexing and
cross-referencing of functions. The SP system provides the additional facility of
powerful pattern matching and searching. This can mean flexible retrieval
capabilities comparable with what is provided in the best systems for
bibliographic search.

\item The `universal' nature of SP patterns (Section \ref{representation_of_knowledge}) should facilitate the integration of diverse kinds of knowledge, making it easier to re-use software in different contexts without the incompatibilities that plague current systems.

\end{itemize}

\subsubsection{Configuration management}

A significant problem in system development is the `configuration
management' problem, meaning the problem of keeping track of and managing
the several versions of a software system which are typically produced in
the course of development. What gives the problem spice is that most
systems of any size come in parts, sub-parts and so on and each component
at any level may come in several versions. Furthermore, associations must
be maintained between source code, object code, requirements
specifications, designs and other documentation for all the versions and at
all the different levels within each version.

The ability of the SP system to integrate class hierarchies with part-whole hierarchies may serve for configuration management (Section \ref{making_haste_slowly}). Versions of a software system may be represented as classes as can the versions of its parts and sub-parts. And different combinations of components or associations between components can be tied together in a flexible manner according to need.

These capabilities should be useful in applications other than software engineering, wherever documents are to be developed comprising diverse kinds of knowledge divided into parts and sub-parts with one or more versions for each part.

\subsubsection{Automatic programming}

There is a well-established principle in software engineering that a
well-designed software system should model the domain which it is designed
to serve. This principle was cogently argued by \citet{jackson_1975} for data
processing, it underpins the practice of using entity-relationship
models, data-flow models and entity life histories as a basis for software
design and it is a key part of the philosophy of object-oriented design.

The anticipated ability of the SP machine to learn new concepts from experience means that, given
`raw' data about a domain, the system should be able to create a model of that
domain automatically. Given a stream of data, an SP machine should be able to derive a
well-structured Jackson diagram or grammar for the data. Information may
also be organised into class hierarchies with inheritance of attributes in
the manner of object-oriented design.

If the modularisation and structuring of software can be automated or semi-automated in this kind of way, this should greatly accelerate the process of designing new software systems.

\subsubsection{Modifiability}

Most software is far too inflexible. Although most packages give the
end-user some scope to set parameters or tailor the package to their
individual requirements, this flexibility is usually very limited. In
general, a software package is a take-it-or-leave-it item. Buying software
can mean a lot of effort in evaluating packages which are nearly but not
exactly what one needs. Short cuts in evaluation often mean discovering
shortcomings after a package has been bought. The same kinds of problem
apply to bespoke software: it is too difficult to modify software to meet
changing needs even when that software was originally designed to fit one's
specific requirements. All these problems would be reduced if it were
easier to modify software to meet one's needs or changes in one's needs.

It should be relatively easy to make changes to SP software for three main
reasons:

\begin{itemize}

\item {\em Simplicity} in use. As described earlier, the SP machine should eliminate all processes of
translation, compiling and verification. This should greatly simplify the process of changing software.

\item {\em Minimal redundancy} in software. The system should remove one of the 
main difficulties in making changes to conventional programs: a change 
made in one part of the program often means that corresponding changes 
must be made in other parts; it can be troublesome finding all the 
parts that must be kept in step and errors can easily creep in. 
As with other object-oriented systems which support class 
hierarchies and inheritance, the SP system should allow software to be designed with a 
minimum of redundancy; this means that any given change to a program 
can be made in only one part of the program without the risk of
inconsistencies.

\item If {\em automatic programming} can be developed as anticipated, this 
should assist the process of modifying software. Any inefficiencies 
in a modified software system may be corrected automatically.

\end{itemize}

\index{software engineering|)}

\subsection{Expert systems and intelligent databases}

\index{expert system|(}\index{database!intelligent|(}

One of the most promising areas of application for the SP machine is where stored knowledge is used as a basis for various kinds of reasoning---like an `expert system' or `intelligent database' running on a conventional computer. We have seen already how the system may be used in such applications as medical diagnosis (Section \ref{medical_diagnosis_section}), fault-finding with a discrimination network (Section \ref{probabilistic_decision_network}), reasoning in the context of a forensic investigation (Section \ref{reasoning_with_rules_section}), and causal diagnosis (Section \ref{causal_diagnosis_section}).

Compared with existing systems, the potential advantages of the SP machine include:

\begin{itemize}

\item Seamless integration of different kinds of knowledge. In a forensic investigation, for example, knowledge about a crime and knowledge about suspects may be expressed in the same format as `expert' knowledge about ways in which crimes are normally committed and `everyday' knowledge about the association between matches and fire, the need to keep dry in wet weather, and so on. This should facilitate the process of finding connections between different kinds of knowledge.

\item Seamless integration of different styles of reasoning: probabilistic `deduction', abductive reasoning, nonmonotonic reasoning, and others, as described in Chapter \ref{pr_chapter}, as well as classical kinds of logical reasoning considered in Chapter \ref{maths_logic_chapter}.

\item The ability to calculate probabilities for inferences fully in accordance with standard principles of probability theory.

\item The ability to learn new knowledge from `raw' data in the manner of `data mining', abstracting rules and organising knowledge into class hierarchies and part-whole hierarchies without the need for human intervention. Of course, there is still a substantial gap between these possibilities and what the current model can do (Chapter \ref{learning_chapter}) but there seems no reason in principle why the system should not be developed along these lines.

\end{itemize}

\index{expert system|)}\index{database!intelligent|)}

\subsection{The storage and retrieval of information and the relational model}

\index{information!retrieval|(}\index{database!relational model|(}

For applications where the emphasis is on the storage and retrieval of information, the SP system could be an attractive alternative to the relational model (\citet{codd_1970}, see also \citet{connolly_begg_2002}), which is currently the dominant model for database applications (if we discount the network model and its dramatic resurgence in the World Wide Web). This subsection compares the two.

\subsubsection{Representation of tables} 

One of the attractions of the relational model has been the simplicity of the idea of representing everything in tables. Although this idea is easy to grasp and has proved very practical and successful in a host of applications, it is not quite as simple as it may superficially appear. This is because a table in a relational database is not merely an array of columns and rows as it appears on the computer screen. In order to access individual cells efficiently, it is necessary to mark cells and rows with appropriate tags. Normally, these tags are hidden from view but they can be seen when a table is translated into something like XML, designed to facilitate the inter-change of information between computer systems in a form that can be read by people.

The difference can be seen by comparing Table \ref{periodic_table_example}---which shows two rows of the periodic table---with Figure \ref{xml_table_example} in which the same two rows are represented using XML. In the latter representation, each cell needs to be marked explicitly with tags like `$<$name$>$ ... $<$/name$>$' or `$<$symbol$>$ ... $<$/symbol$>$' corresponding to the column in which the cell appears. Tags like `$<$row$>$ ... $<$/row$>$' are also needed to mark each row as a distinct entity.

\begin{table}[!hbt]
\centering
\begin{tabular}{l l l l}
\em Name & \em Symbol & \em Atomic & \em Atomic \\
& & \em Number & \em Weight \\
\\
Hydrogen & H & 1 & 1.00794 \\

Helium & He & 2 & 4.0026 \\
\end{tabular}
\caption{Two rows of the periodic table.}
\label{periodic_table_example}
\end{table}

\begin{figure}[!hbt]
\fontsize{10.00pt}{12.00pt}
\centering
\begin{BVerbatim}
<periodic_table>
     <row>
          <name>Hydrogen</name> 
          <symbol>H</symbol> 
          <atomic_number>1</atomic_number> 
          <atomic_weight>1.00794</atomic_weight> 
     </row>
     <row>
          <name>Helium</name> 
          <symbol>He</symbol> 
          <atomic_number>2</atomic_number> 
          <atomic_weight>4.0026</atomic_weight> 
     </row>
</periodic_table>
\end{BVerbatim}
\caption{Table \ref{periodic_table_example} represented using XML.\protect\footnote{Adapted with permission from {\em The XML Bible} by Elliotte Rusty Harold, John Wiley \& Sons, 2001.}}
\label{xml_table_example}
\end{figure}

Although there are differences in details, the way in which a table is represented in XML---and more covertly in the implementation of a relational database---is essentially the same as it would be represented in the SP system. Thus, the SP system can represent information in the form of tables and it can do so in a manner which matches the simplicity of relational databases, neither more nor less.

\subsubsection{Class hierarchies}

\index{class!hierarchy|(}

Tables can be a convenient way to represent classes of entity. For example, in a typical data processing application, one might use a table to represent the class of employees, another table to represent the lorries that the company uses to distribute its products, and so on. But if tables are used in an unthinking way, they can lead to unwanted redundancies in a database. For example, the class of employees and the class of customers would both normally have fields for `title', `first name', `second name', `address', `telephone number' and, perhaps, others.

It has been recognised for some time that redundancies of that kind can be reduced by abstracting the repeated fields, putting them in a third table (which would, in this case, be called something like `person'), and linking the super-class with its subclasses by the use of reference numbers. In this kind of way, relational databases have provided a basis for the use of `entity-relationship' models in data processing applications \citep{chen_1976}, with hierarchies of classes which are very much like the class hierarchies of object-oriented design.

Given that class hierarchies can also be represented in the SP system (Section \ref{class_part_inheritance}), one might think that, with respect to class hierarchies, there was little to choose between the relational model and the SP system. But in this area the SP system has two advantages compared with the relational model:

\begin{itemize}

\item In the relational model, each attribute of a class---represented by a field or column in a table---is a {\em variable}. There is no provision in the relational model for representing the attributes of a class as literal values such as `furry', `warm-blooded', `has-backbone' or the like. In the SP system, by contrast, the attributes of a class can be any combination of literal values and variables, the latter represented by pairs of ID-symbols such as `title ... \#title', `first\_name ... \#first\_name', and so on. 

\item In the SP system, the connection between a given class and any one subclass is made via matching symbols in a corresponding pair of patterns. In the relational model, by contrast, the connection between a given class and any one of its subclasses is made via matching reference numbers in the tuples that represent individual instances of those classes. For tables containing hundreds or thousands of rows (which is not unusual in data-processing applications), this system for connecting a class with any one of its subclasses is considerably more clumsy than the comparative simplicity of the SP scheme where classes can be connected to each other directly and not via their individual instances. 

\end{itemize}

With respect to the first point, it may be argued that databases are all about storing information that varies from one instance of a class to another and that there is no need for any information about literal values. This may be true for conventional data processing applications where there is no uncertainty about what class a given entity belongs to. But in applications where one may wish to recognise some unknown entity on the basis of fragmentary information (as described in Chapter \ref{rr_chapter}), then it is important to know as much as possible about each class, including attributes that would most naturally be represented with literal values, not variables.%
\index{class!hierarchy|)}

\subsubsection{Part-whole hierarchies}

\index{part-whole hierarchy|(}

Very much the same can be said about part-whole hierarchies as has just been said about class-inclusion hierarchies. Part-whole hierarchies can be represented in a relational database but this is relatively awkward compared with the SP system and does not allow for parts to be described in terms of literal attributes as well as variables. With a relational database, part-whole relations cannot be integrated with class-inclusion relations in the seamless manner that SP allows.%
\index{part-whole hierarchy|)}

\subsubsection{Retrieval of information}

In the SP system, retrieval of information is achieved most naturally by the use of a New pattern representing the kind of Old information that is to be retrieved, with gaps where details are needed. This is essentially the same as `query-by-example' in relational databases, except that the SP system does not require the kind of `wild' symbols that are required in some relational systems (e.g., `?' or `*'), and the SP system is more tolerant of extraneous information than is the case in some implementations of relational databases.

Given appropriate patterns, the SP system allows information to be retrieved indirectly via `meanings' as well as via direct matches between the query and stored data (Section \ref{indirection_in_ir}). Relational databases do not lend themselves naturally to this kind of information retrieval.

Although current SP models are geared mainly to the representation and retrieval of symbolic information, it is envisaged that the SP machine will be developed to handle numeric information as effectively as conventional computers (Section \ref{sp_machine_other_developments}). In that case, it should be possible to store numeric information and to retrieve it with operators such as `more than' (`$>$'), `less than' (`$<$') and so on.

There is nothing in current SP models equivalent to SQL or similar languages for database retrieval. But if the system can cope with the syntax and semantics of natural languages (Chapter \ref{language_chapter}), there seems no reason in principle why, with appropriate patterns stored in Old, the SP system should not accommodate relatively simple artificial languages like SQL.%

\subsubsection{Organisation of knowledge}

\index{learning}

Currently, the organisation or `normalisation' of knowledge in a database is done by human judgement. Although the results are normally good, the process of designing a database can be a significant task, especially where the domain of application is complex.

In principle, some or all of this task may be automated. If learning processes like those in the SP70 model can be developed to their full potential, it should be possible to structure an SP database automatically when it is first established and to restructure it at any stage. This should save effort and corresponding expense, and it could conceivably overcome weaknesses in database structures that are created `manually'.

\subsubsection{Conclusion}

With respect to the storage and retrieval of information, the SP system can represent tables in essentially the same form as they are represented in relational databases, but the system has advantages over the relational model in the representation of class hierarchies, part-whole hierarchies and in aspects of information retrieval. If the SP machine can be developed as anticipated, it should be able to do everything that a relational database can do and more besides.

Since, in addition, the SP system can be applied in such areas as natural language processing, probabilistic reasoning, planning and problem solving, it has substantial attractions as an alternative to the relational model.%
\index{information!retrieval|)}\index{database!relational model|)}

\subsection{Other applications}

Other areas where the SP machine may be applied include:

\begin{itemize}

\item {\em Natural language processing}.\index{language!processing} In Chapter \ref{language_chapter}, we saw how the SP system lends itself to the representation of natural language syntax, to natural language parsing and production and, more tentatively, to the integration of syntax with semantic structures. On this evidence, the SP machine should be a promising vehicle for such applications as understanding and production of natural language in spoken or written form or translations between natural languages.

\item {\em Automatic learning of natural language grammars}.\index{learning} If the learning processes described in Chapter \ref{learning_chapter} can be developed as anticipated, automatic learning of natural language grammars, including semantics, may become feasible.

\item {\em Planning and problem solving}.\index{planning}\index{problem solving} With further work on planning and problem solving, as outlined in Chapter \ref{pps_chapter}, a range of new applications should open up.

\item {\em Project management}.\index{project management} On rather shorter timescales, an SP machine would be a convenient vehicle for the storage and retrieval of information associated with project management, as described in \citet{wolff_1989}.

\end{itemize}

\section{Conclusion}

Whether or not the SP system will be any good at mending socks remains to be seen. Meanwhile, there is a range of possible applications that promise to be useful on shortish to medium timescales. The SP theory provides a new perspective on many issues in computing and cognition and should have heuristic value in neurophysiology (Chapter \ref{neural_chapter}), in cognitive psychology (Chapter \ref{psychology_chapter}), and in mathematics and in logic (Chapter \ref{maths_logic_chapter}).

%% file: conclusion.tex
\chapter{Conclusion}\label{conclusion_chapter}

It has been a long but interesting journey since the first germ of an idea for this project about 1982 or 3. Serious work did not begin until about 1987 and for quite a long time after that the few areas of insight were overwhelmed by the many unanswered questions. A major step forward was the development of a version of the multiple alignment concept, but a lot more work has been needed to develop working models and to gain some understanding of ways in which the theory may be applied.

\section{Is the SP Theory a Good Theory?}

Although the theory is not yet `complete', it is pertinent to review its status in terms of the criteria discussed in Section \ref{creating_a_good_theory}.

\subsection{Falsifiability and the Making of Predictions}

If an empirical theory is simply a compressed version of observational data, it is not essential that it should make predictions that can be found to be `true' or `false'. But it is difficult to achieve this kind of compression without, at the same time, making predictions of one kind or another (as was noted in Section \ref{theory_criteria_integration}) and our willingness to believe a theory is certainly enhanced if it makes predictions that are born out by observation.

As a theory of human cognition and brain function, the SP theory makes the broad prediction that, if we could get a comprehensive and very detailed view of what is going on inside a living human brain, we would find structures like the `pattern assemblies' described in Chapter \ref{neural_chapter}, we would be able to observe the formation of associations amongst pattern assemblies that would be equivalent to multiple alignments, and we would be able to relate the building of multiple alignments to a person's ability to analyse or create natural language, to reason about things, to retrieve information from memory, to learn new knowledge, and so on. 
Clearly, the SP theory is not a vacuous catch-all theory and it can, in principle, be falsified.

Of course, the major difficulty is that, for very obvious ethical and methodological reasons, we {\em cannot} see the workings of a living human brain in sufficient detail to see immediately whether the SP theory is true or false. We are forced to use indirect clues to what is going on such as reaction times, error rates, measurements of brain waves, and so on. And this raises the further difficulty that the SP theory, like many other theories in this area, does not (yet) specify precisely how the neural processing would be done. Without this detail, it is very difficult to make sensible predictions of the kind that are needed to test the theory experimentally. Notwithstanding these difficulties, Chapter \ref{psychology_chapter} does contain some suggestions about ways in which the SP theory might be differentiated experimentally from alternative theories in cognitive psychology.

It should not be forgotten that many of the topics in artificial intelligence---including those that have been considered in previous chapters---are founded on everyday observations of what people can and cannot do. To the extent that the SP theory can accommodate them, these observations form part of the empirical support for the theory even though they are not derived from formal experiments. 

The SP theory is not set in stone. As we saw in Section \ref{cross_serial_dependencies}, the phenomenon of cross-serial dependencies in language is a potential problem for the theory, although it is not yet clear whether or how any adjustment to the theory is required. As work proceeds in exploring the potential applications of the theory, it is likely that changes will be made.

\subsection{Breadth, Power, Depth and Simplicity}

Assuming that explanatory `breadth' or `power' are equivalent, how does the SP theory fair? As a theory of computing, mathematics and logic, which is also a theory of perception and cognition, its scope is unusually wide.

A sceptic might object that Turing's theory of computing and information theory (Shannon's version and algorithmic information theory) are theories relating to information processing that also have very wide scope. Do we really need the SP theory as well? In answer, the theory is an alternative to Turing's theory of computing with much more to say about the nature of intelligence and with other advantages described in Chapter \ref{computing_chapter}. The SP theory incorporates significant insights from Shannon's theory and algorithmic information theory but says much more about the nature of information processing, especially the kinds of human-like processing of information that have proved so challenging to understand.

Apart from wide explanatory scope, a good theory needs to have sufficient depth or substance to do what it purports to do and, at the same time, it needs to be as simple as possible. With regard to depth, the use of computer models in the development of the SP theory has ensured that it is far removed from the kinds of sketchy descriptions that are sometimes dignified with the title `theory' when a more accurate appellation would be `proto-theory' or `first tentative thoughts'.

Is the theory simple? It is certainly more complex that the Turing theory but, if we take into account the software that is needed to make a universal Turing machine (or an ordinary computer) do anything sensible, then the SP theory almost certainly yields an overall simplification, as discussed in Sections \ref{sp_and_other_models_of_computing} and \ref{simplification_of_computing_systems}. This global simplification could be very substantial.

\begin{center}
\rule{3in}{2pt}
\end{center}

As we saw in the previous chapter, there are areas of the SP theory that are still in need of development, especially in relation to unsupervised learning. But the overall framework has now matured to the point where it should provide a reasonably firm foundation for further work as described in that chapter. I hope readers will feel inspired to pick up some of these threads and take the ideas forward.

%% file: matching.tex
\chapter[Finding Good Matches]{Finding Good Matches Between Two Sequences of Symbols}\label{matching_appendix}

This appendix, based on part of \citet{wolff_1994_scaleable}, describes the process for finding full matches and good partial matches between two sequences that lies at the heart of the SP61 and SP70 models. This process was first implemented in SP21 (a precursor of SP61 and SP70) that was designed for best-match information retrieval. Much of the discussion is couched in those terms.

\section{The hit structure}

Figure \ref{pattern_matching_figure} illustrates the main concepts introduced in the description that follows. In this description, the query and the database are both sequences of atomic symbols, assumed to be characters in the discussion, and the database may be divided into sections such as sentences or paragraphs. Here is the process:

\begin{figure}
\begin{center}
\includegraphics[width=0.9\textwidth]{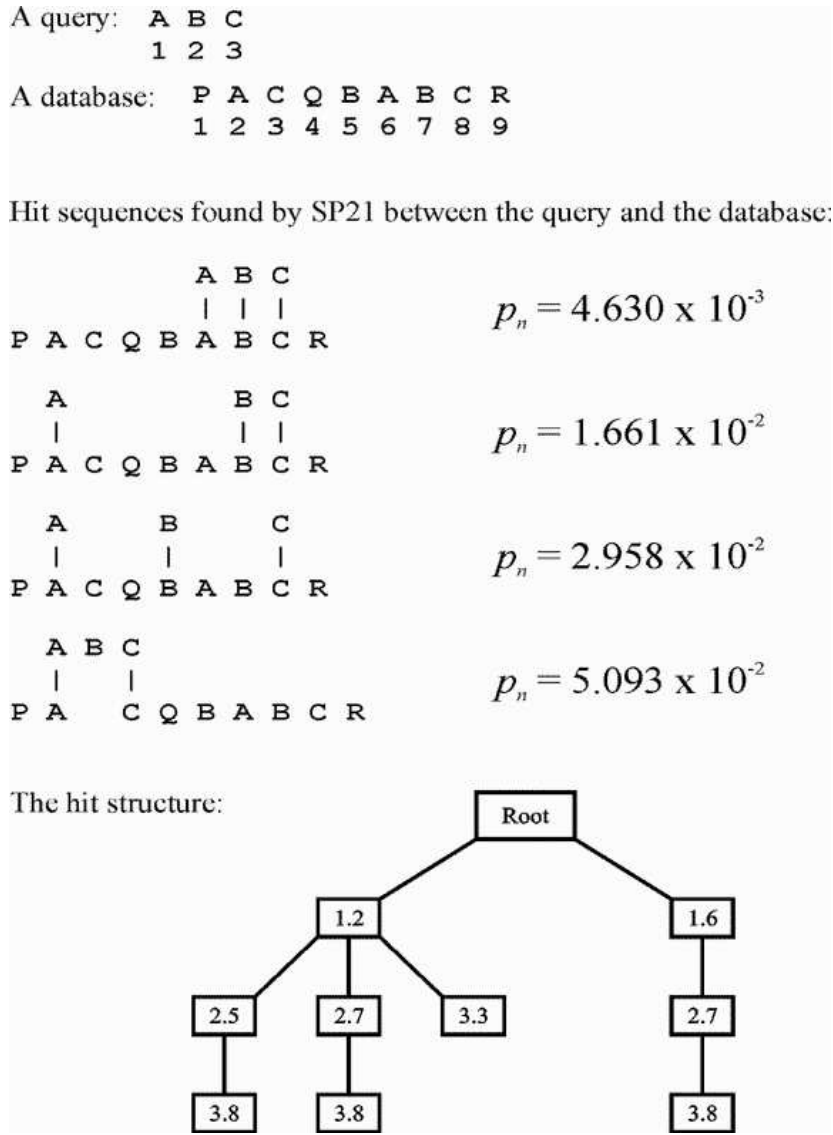}
\end{center}
\caption{Concepts in pattern matching and search. A `query' string and a `database' string are shown at the top with the ordinal positions of characters marked. Sequences of hits between the query and the database are shown in the middle with corresponding values of $p_n$ (described in the text). Each node in the hit structure shows the ordinal position of a query character and the ordinal position of a matching database character. Each path from the root node to a leaf node represents a hit sequence.}
\label{pattern_matching_figure}
\end{figure}

\begin{enumerate}

\item The query is processed left to right, one character at a time.

\item Each character in the query is, in effect, broadcast to every character in the database to make a yes/no match in each case.

\item Every positive match (hit) between a character from the query and a character in the database is recorded in a data structure which will be referred to as the {\em hit structure}.

\begin{enumerate}

\item The hit structure stores sequences of hits. In each such hit sequence, the order of the matched query characters is the same as the order of the matched database characters. But there may be unmatched query characters anywhere within the sequence of matched query characters and there may be unmatched database characters anywhere within the sequence of matched database characters.

\item The hit structure is implemented as a tree:

\begin{itemize}

\item Each path from the root of the tree to any other node records a left-to-right hit sequence, one hit on each node.

\item The root of the tree is a dummy node which does not record any hit.

\end{itemize}

\item Although a given query character may match two or more characters in the database, only one of these hits is recorded in any one hit sequence. Likewise for database characters.

\item For each hit recorded in a node in the tree, there is a record of the position of the character in the query, the position of the matching character in the database, and a measure of probability of the sequence of hits up to and including the given hit (as described below).

\item If the database is divided into sections, then it can be convenient to apply a rule that all the hits recorded in one path must all come from one section. If this rule is in force (and this may be a decision made by the user) then the system will not attempt to find match sequences which cross from one section to another.

\end{enumerate}

\item The hit structure is updated every time a hit is found. One or more new nodes for this hit (which will be referred to as the current hit) is added to the hit structure in the following way:

\begin{itemize}

\item The tree is examined to identify each hit which immediately precedes the current hit. In this context, the meaning of `precede' is that the database character for the given hit precedes the database character for the current hit and likewise for the query characters. The meaning of `immediately' is that, if the node for the given hit has children, then the hits in the child nodes do not precede the current hit. There may, of course, be unmatched query characters or unmatched database characters between the two hits.

\item For each of the `immediately preceding' hits identified in this way, a probability is calculated for the sequence of hits comprising the path up to and including the current hit.

\item For each hit sequence or path which has been identified in this way (if any), a new node for the current hit is added to the hit structure as the leaf node for that path. In each case, the probability value for the path is recorded in the new node.

\item If there are no paths identified in this way then a new path is started with a node for the current hit as a child of the root and an initial probability value as described below.

\end{itemize}

\item If the memory space allocated to the hit structure is exhausted at any time then it is `purged': the leaf nodes of the tree are sorted in reverse order of their probability values and each leaf node in the bottom half of the set is extracted from the hit structure, together with all nodes on its path which are not shared with any other path. The recording of hits may then continue using the space which has been released.

\item After the last query character has been processed, the paths from the root to the leaf nodes are displayed in order of their probability in a convenient form for inspection by the user.

\end{enumerate}

\section{Probabilities}

\index{probability|(}

The search process, just described, uses a measure of probability, $p_n$, as its metric. This metric provides a means of guiding the search which is effective in practice and appears to have a sound theoretical basis. To define $p_n$ and to justify it theoretically, it is necessary first to define the terms and variables on which it is based:

\begin{itemize}

\item For each hit sequence $h_1 ... h_n$, there is a corresponding series of gaps, $g_1 ... g_n$. For any one hit, the corresponding gap is $g = g_q + g_d$, where $g_q$ is the number of unmatched characters in the query between the query character for the given hit in the series and the query character for the immediately preceding hit; and $g_d$ is the equivalent gap in the database, $g_1$ is taken to be 0.

\item $A$ is the size of the alphabet of character types used in the query and the text.

\item $p_1$ is the probability of a match between any one character in the query and any one character in the database on the null hypothesis that all characters are equally probable at all locations. Its value is calculated as: $p_1 = 1 / A$.

\end{itemize}

Using these definitions, the probability of any hit sequence of length $n$ is calculated as:

\[p_n = \prod_{i=1}^{i=n}(1 - (1 - p_1)^{g_i + 1}), \hspace{3mm} g_1 = 0\].

It should be clear from this formula that it is easy to calculate the probability of the hit sequence up to and including any hit by using the stored value of the hit sequence up to and including the immediately preceding hit.

\subsection{Justification}

The thinking behind the formula is straightforward. In accordance with established practice in statistics, the method aims to calculate the probability that the observed distribution of hits, or better, could have occurred by chance on the `null hypothesis' that all the characters in the alphabet are equi-probable at every location in the query and the database, i.e. that the distribution of characters is random. In this context, a distribution of hits which is `better' than an observed distribution is one which has more hits within the same range, or the hits fall into clumps, or both these things.

A hit sequence with a low probability is more `significant' than one with a high probability and may be taken as evidence that the null hypothesis should be rejected. A hit sequence with a low probability is normally more interesting to the user than a high probability hit sequence which could be merely the result of chance.

It is important to stress that this approach to the analysis does not in any way prejudge the statistical properties of the query or the database. It is, of course, very well known that the distribution of characters in any natural language (e.g. English) is not random and that such texts contain large amounts of redundancy in the sense of Shannon's information theory \citep{shannon_weaver_1949}. The null hypothesis provides a reference point or baseline for measuring how far an observed distribution of hits departs from randomness.

The probability measure that has been described is an inverse measure of redundancy. In the case under discussion, where two strings are being compared, $p_n$ measures the redundancy between the two strings. It says nothing about redundancy that may exist within one string or the other.
Under the null hypothesis, the probability of an observed hit sequence or better depends on three main factors:

\begin{itemize}

\item There is a better chance of finding a hit sequence which is at least as good as the observed sequence if the query or the database or both of them are large.

\item If the $n$ hits of a hit sequence are scattered across a relatively long part of the query, or a relatively long part of the database, or both, then the associated probability is higher than for a `closely packed' hit sequence which is confined to portions of the query and the database which are as short as n or only a little longer.

\item Other things being equal, the probability of an observed hit sequence or better decreases as $n$ increases.

\end{itemize}

For the purposes of information retrieval, the size of the query and the size of the database should not be factors in deciding whether a given hit sequence is significant. What is of interest is the probability of an observed hit sequence after the effects of query size and database size have been abstracted. For this purpose, hit sequences which are closely packed and relatively long are the most significant, independent of the sizes of the query and the database, and independent of where the hit sequences occur within the query and the database.

On this basis, the probability of the first or only hit in a sequence $(p_1 = 1 / A)$ is the same as the probability that any given side of an $A$-sided unbiased die will appear on any one throw of the die. This formula for the first or only hit in a sequence can be derived from the main formula if $n$ is 1 and $g_1$ is 0. In the main formula, $(1 - p_1)$ is the probability of a non-match between any one character in the query and any one character in the database. If there are $g$ non-matches between one hit and the next, then the probability of finding one or more hits over a distance of $(g + 1)$ is $(1 - (1 - p_1)^{g + 1})$. This probability is then multiplied by the probability for the hit sequence up to and including the preceding hit, to give $p_n$.

As an illustration, consider the fourth hit sequence shown in Figure \ref{pattern_matching_figure}. The size of the alphabet, $A$, is 6, so $p_1$ is $1 / 6$ which is 0.1666666, and $(1 - p_1)$ is 0.8333333. As with any other hit sequence, the first hit in the sequence has $g_1 = 0$ so that its probability is $(1 - (1 - p_1)^{0 + 1})$ which is the same as $p_1$. For the second hit, $g_q$ is 1 and $g_d$ is 0 so that $g$ is 1. The corresponding value of $(1 - (1 - p_1)^{1 + 1})$ is 0.3055556. This is multiplied by the probability of the hit sequence up to and including the previous hit giving an overall value for $p_n$ of 0.0509259.

The analysis which has been presented assumes an alphabet of fixed size. This is plausible if the atomic symbols for yes/no matches are characters but may seem less plausible with larger units such as words or phrases because of their variety in natural languages. However, in any one combination of query and database there is a finite (if large) number of different words or phrases. This means that the analysis can be applied even with these larger units.%
\index{probability|)}

\section{Discussion of the search technique}

The technique which has been described incorporates the principles of metrics-guided search like this:

\begin{itemize}

\item The hit structure plots a set of alternative paths through the search space.

\item The probability metric is used (during purging) to prune leaves and branches from the tree of paths.

\item The method may be classified as beam search because the search proceeds along several paths at once. This reduces the risk of getting stuck on a local peak. Increasing the size of the hit structure increases the number of paths and thus increases the chance of finding `good' hit sequences.

\end{itemize}

If the database is divided into sections, and if a rule is applied that hits in a sequence must all come from the same section, this has the effect of blocking some paths through the search space, thus reducing the number of possibilities which need to be considered and saving some processing time.

There is a trade-off between the maximum size of the hit structure and the ability of the system to find partial matches. When the maximum size of the hit structure is small, processing times are short but the system may get stuck on local peaks and miss partial matches that people can see. When the maximum size of the hit structure is large, the system finds partial matches more effectively but processing times are longer. It seems reasonable that, in a fully-developed version of the system, this trade-off between search time and level of performance should be under the control of the user.

The idea of broadcasting symbols is not in itself especially new and has been described elsewhere \citep{carroll_etal_1988, lee_1986}. The novelty of the technique which has been described is in the way the broadcasting of symbols is combined with a technique for keeping track of partial matches between the query and the database and in how the system calculates probabilities and uses this information to select amongst the many possible paths through the search space.

In a serial processing environment, the broadcasting of symbols must be done serially, but this kind of operation lends itself very well to the application of parallel processing.

The advantages of this technique compared with the basic dynamic programming method are:

\begin{itemize}

\item The space complexity of the process is O$(D)$, better than O$(Q \cdot D)$ for the basic dynamic programming method.

\item The method appears to be better suited to parallel processing although, for the approximate string matching problem, an adaptation of dynamic programming for parallel processing has been described \citep{bertossi_etal_1992}.

\item The technique for pruning the search tree may be applied, however large the search space may be. In general, the `depth' or thoroughness of searching can be controlled by specifying the maximum size of the hit structure.

\item Unlike the standard dynamic programming method, this method can deliver two or more alternative alignments of two patterns.

\end{itemize}

\section{Computational complexity}\label{sp21_computational_complexity}

\index{complexity, computational|(}

Given the `combinatorial explosion' of possible matches which was described earlier, a key question about any system of this kind is the demand which it makes on processing time and computer memory when the quantities of data are increased. This section describes analytic and empirical evidence on these points.

\subsection{The best, worst and typical cases}

The core of the search process is the broadcasting of characters from one string (the query) to each of the characters in another string (the database). From the perspective of absolute running times and computational complexity, the best case is when none of the symbols in the query match any of the symbols in the database. In this case, there are no hit sequences to be stored and the search is completed very quickly.

The worst case is when all the symbols in the query and the database are the same. In principle, this yields the largest possible number of hit sequences although, in practice, SP61 will purge many of them from its hit structure.

The typical case, somewhere between the two extremes, is where the query and the database both contain a range of symbol types distributed in the kind of way that letters are distributed in natural languages.

Since the worst case is unlikely to occur in practice, the typical case has been assumed in what follows.

\subsection{Time complexity in a serial processing environment}

In a serial processing environment, it is clear that the processing time for this operation is proportional to the length of the query string and, independently, it is proportional to the length of the database. In other words, the time complexity for this operation is O$(n \cdot m)$, where $n$ is the number of characters in the query and $m$ is the number of characters in the database.

\subsubsection{Updating the hit structure}

The process of updating the hit structure includes the time required to search the hit structure for the best hit sequences and the time required to add new nodes. The time required to search the hit structure will vary, depending on whether the hit structure is full or has recently been purged; but, apart from a small effect at the start of processing as the space available for the hit structure is filled, the time required for this operation should be independent of $n$ or $m$.

Since the updating operation occurs only for hits, and since the proportion of hits amongst the yes/no matches should be independent of $n$ or $m$, we may conclude, overall, that our initial assessment of the algorithm remains valid. In short, an analysis of the algorithm shows that its time complexity in a serial processing environment should be O$(n \cdot m)$. This analysis is independent of the size of the hit structure.

The foregoing analysis remains valid when the database is divided into sections with the exclusion of hit sequences from one section to another. This kind of constraint can save overall processing time by reducing the variety of hit sequences and thus reducing the number of purges of the hit structure; but the constraint does not change the relationship between processing time and $n$ or $m$.

\subsection{Time complexity in a parallel processing environment}

As previously noted, the search process lends itself well to parallel processing:

\begin{itemize}

\item The process of broadcasting a query character to every character in the database is an intrinsically parallel operation.

\item If the database is divided into parts, each part with its own small hit structure, then updating of the hit structures may be performed in parallel.

\end{itemize}

If finding hits and updating the hit structure takes unit time independent of the size of the database, as seems possible in a parallel processing environment, then the time complexity of the process should be O$(n)$.

\subsection{Space complexity}

The space required to store the database is independent of any retrieval mechanism and is therefore excluded from this analysis of the space complexity of the search process. At this level of abstraction, there is no distinction between `main memory' and `secondary storage', both kinds of memory being assumed to function as a unified `virtual memory'. The memory required specifically for the search process is mainly the space required to store the hit structure.
Although the hit structure varies in size as the program runs, it never exceeds a pre-defined limit because it is purged whenever the limit is reached. If the user requires all hit sequences down to a fixed level of `quality' then, for typical data, the size of the hit structure should be increased in proportion to $m$ and the space complexity of the process would be O$(m)$.

\subsection{Empirical evidence}

Running times for SP21 have been plotted to show the effect of varying the size of the query (with a database of constant size) and also to show the effect of varying the size of the database (with a query of constant size) (see \cite{wolff_1994_scaleable}). The queries and the databases were all samples of English.

In each case, the relationship is approximately linear. These results lend support to the analytic conclusion that the time complexity of the SP21 process in a serial processing environment is O$(n \cdot m)$.%
\index{complexity, computational|)}

%% file: physics.tex
\chapter{The SP Theory and Physics}\label{physics_appendix}

This appendix is intended for some speculative thoughts about ways in which the SP theory may suggest a reinterpretation of some concepts in physics, especially in quantum theory.